\documentclass{thesis-uwf}

\makeatletter
\def\section{\@startsection{section}{1}{\z@}{1.5ex plus 1.5ex minus 0.5ex}%
{1.5ex plus 0.5ex minus 0.5ex}{\normalsize\scshape\MakeUppercase}}
\makeatother

\justifyleft

\usepackage{graphicx} 
\usepackage{amsmath} 
\usepackage{soul}
\usepackage[english]{babel}
\addto\extrasenglish{%
}
\usepackage{csquotes}
\usepackage[backend=bibtex,style=ieee,natbib=true]{biblatex}
\usepackage{setspace}
\usepackage{xspace}
\usepackage{mathtools}
\usepackage{array}
\usepackage{xcolor}
\usepackage{colortbl}
\usepackage{fvextra}
\usepackage{indentfirst}
\usepackage{enumitem}
\usepackage{ragged2e}
\usepackage{caption}
\usepackage{float}
\usepackage{titlesec}
\usepackage{tikz}
\usepackage[edges]{forest}
\usepackage{pgfplots}
\usetikzlibrary{arrows.meta,positioning,fit,calc,matrix,shapes.geometric,backgrounds,intersections}
\usepgfplotslibrary{groupplots}
\pgfplotsset{compat=1.18}

\newcommand{\inputthesisTikZ}[1]{%
\begingroup
\makeatletter
\newcommand{\thesisfig@setroman}[2]{%
  \fontencoding{T1}\fontfamily{ptm}\fontseries{m}\fontshape{n}\selectfont
  \@setfontsize#1{#2}}%
\def\normalfont{\thesisfig@setroman\normalfont{10}{12}}%
\normalfont
\let\textnormal\normalfont
\def\small{\thesisfig@setroman\small{9}{10.95}}%
\def\footnotesize{\thesisfig@setroman\footnotesize{8}{9.6}}%
\def\scriptsize{\thesisfig@setroman\scriptsize{7}{8.4}}%
\def\ttfamily{\fontencoding{T1}\fontfamily{pcr}\selectfont}%
\def\@listi{%
  \leftmargin\leftmargini
  \labelwidth\leftmargini
  \advance\labelwidth-\labelsep
  \parsep \z@
  \topsep \z@
  \itemsep \z@
  \small
}%
\AtBeginEnvironment{itemize}{\small}%
\setlist[itemize,1]{topsep=0pt,itemsep=0pt,parsep=0pt,partopsep=0pt}%
\tikzset{%
  execute at begin picture={%
    \normalfont
  },%
}
\makeatother
  \input{#1}%
  \endgroup
}

\setlist[description]{labelindent=0.4cm}

\titlespacing{\chapter}{0pt}{-14.4pt}{0pt}
\titleformat{\chapter}[display]
  {\centering\normalfont\fontsize{12}{14.4}\bfseries}
  {\thechapter}
  {1em}
  {}

\titlespacing{\section}{0pt}{2.4pt}{0pt}
\titleformat{\section}
  {\normalfont\fontsize{12}{14.4}\bfseries}
  {\thesection}
  {1em}
  {}

\titlespacing{\subsection}{0pt}{2.4pt}{0pt}
\titleformat{\subsection}
  {\normalfont\fontsize{12}{14.4}\itshape\bfseries}
  {\thesubsection}
  {1em}
  {}

\newcommand{\x}{\boldsymbol{x}}
\newcommand{\y}{\boldsymbol{y}}

\definecolor{myred}{RGB}{150, 10, 10}

\makeatletter
\renewcommand{\@biblabel}[1]{[#1]\hfill}
\makeatother

\title{A System for Fast, Resilient, and Adaptable Loco-Manipulation Behaviors on Humanoid Robots}

\author{Duncan William Calvert}

\pastdegrees{
B.S. Computer Science, University of West Florida, US, 2014
}

\department{Department of Intelligent Systems and Robotics}

\college{Hal Marcus College of Science and Engineering}

\email{dwc8@students.uwf.edu}

\orcid{0009-0001-5255-6882}

\committee{ %
\signature{Robert Griffin, PhD, Committee Chair}
\signature{Jerry Pratt, PhD, Committee Member}
\signature{Stephen Hart, PhD, Committee Member}
\signature{Matt Johnson, PhD, Committee Member}
}
\depchair{\signature{K. Brent Venable, PhD, ISR Program Director}}
\dean{\signature{Mohamed Khabou, PhD, Dean HMCSE}}
\graddean{\signature{Melissa Webb, EdD, Director, Office of Graduate Studies}}

\showcopyright
\showverification
\showlistoftables
\showlistoffigures
\showlistofalgorithms

\acknowledgments{\par
I would like to thank my Dad for buying me a LEGO Mindstorms robotics kit in 1998, when I was 6 years old.
It was the beginning of a lifelong robotics journey.
Later, when I was in high school, I still remember the day in 10th grade when an afternoon announcement sounded through the intercom for a Robotics club interest meeting.
I went straight there and every day since I've been focused on my career in robotics.
17 years!
I would like to thank my lifelong friend, Madison Fortenberry, and his Dad, Mr. Fortenberry, for teaching me fundamental concepts of engineering, providing support and resources, and coaching us through many competitions.
I have kept each lesson close to the heart.
We also had a ton of fun along the way.
I am especially grateful for Mr. Fortenberry's letter of recommendation for my acceptance to this PhD program.

I would also like to thank Dr. Matt Johnson for attending my senior project presentation in 2014 at the end of my B.S. in Computer Science at UWF.
He followed up with me afterwards, took a further look at my work, and recommended me for the job at IHMC.
I would like to thank Dr. Jerry Pratt and Dr. Peter Neuhaus for hiring me for the IHMC Robotics team during preparations for the DARPA Robotics Challenge.
Every day at IHMC Robotics has been a challenging and inspiring experience.
I have had the opportunity to work with over a hundred passionate and expert roboticists.
Many of them are now lifelong friends.
I would like to thank them all for the good times, thoughtful conversations, and hard teamwork.

Thank you to Jerry for encouraging me to pursue a PhD in this program, for recommending me, and for agreeing to be my advisor.
This was a once in a lifetime opportunity and I am forever grateful.
Thank you for all the engaging courses, conversations, and sailing instruction.

Thank you to Dr. Luigi Penco for an endless amount of ideas, feedback, and support.
Thank you to Dexton Anderson and Tomasz Bialek for helping build the technical foundations of this work, increasing performance, and helping with design.

Thank you to Dr. Robert Griffin for agreeing to be my advisor when Jerry left, for not only keeping the lab alive but giving it new life, and for the thoughtful feedback and encouragement over the years.
Thank you to Dr. Stephen Hart for being on my committee, being supportive and helpful, for the engaging conversations, and for inspiring the world with the Affordance Template Framework.
Thank you to Dr. Matt Johnson for being on my committee, for your three questions, and for challenging my scientific rigor.
Thank you to Jerry, Twan Koolen, Sylvain Bertrand, Georg Widebach, Dexton Anderson, Tomasz Bialek, Jesper Smith, Robert, John Carff, and Brandon Shrewsbury for inspiring me with technical innovations.
Thank you to Beomyeong Park and Robert for debugging logs of failing robots and providing controls fixes.
Thank you to Beomyeong for staying late to help me film one of the coolest loco-manipulation demos.
Thank you to Billy Howell and Jordan Accardo for filming and good vibes, including one night at the lab until like 2am.
Thank you to everyone else throughout the years who contributed to IHMC Robotics software and hardware.
Thank you to Dr. Brent Venable for being supportive, helpful, and kind.
The work in this thesis would not have been remotely possible without you.

Thank you to my best friend and beautiful wife, Jessica, for her unconditional love and support for me.
She agreed to a four year PhD program which ended up taking seven.
I kept having to rewrite that line, but it's ending for sure at seven.
Thank you to my smart and beautiful daughters Allie and Isla for being silly and having tons of fun.
Thank you to Allie for designing icons which were used in this work.
I love you guys.

This research was supported in part by ONR Grants N000-14-19-1-2023 (Fast Behaviors), N00014-22-1-2593 (SquadBot v2), Contract N00014-25-C2410 (SquadBot v3), and Collaborative Agreement W911NF2220201 (Breaching).
Many thanks to Dr. Thomas McKenna for his vision and support for our lab over the years.

I would also like to acknowledge the role of modern AI tools in this work.
Perplexity AI, JetBrains AI Assistant, Google Gemini, Anthropic Claude, OpenAI Codex, and Cursor were used extensively in both the development and presentation of this thesis.
}

\abstract{
    \parindent 0.5in
    \par
    \par
Humanoid robots could take on physically demanding, hazardous, and repetitive work in spaces built for humans, including shipyards, energy infrastructure, construction sites, and factories.
However, a useful robot for these spaces must coordinate locomotion, whole body motion, perception, contact, and operator supervision.
This thesis presents a robot-local, runtime-editable behavior authoring and runtime system.
We argue that behavior architecture can be a primary enabler of robot capability, task execution speed, and reliability, and that runtime editability enables fast behavior creation, adaptation, extension, and combination.
This means if the operator can ``dream it'' using the available behavior nodes, in a short amount of time, they can get the robot to ``do it'' repeatably, autonomously, and quickly.

The system strives to be maximally observable, predictable, and directable following Coactive Design principles developed during the DARPA Robotics Challenge.
Our operator interface remains continuously synchronized to the robot for runtime authoring, monitoring, and repair.
Our behavior architecture uniquely combines object-centric Affordance Templates, organization and logic inspired by Behavior Trees, and runtime-editable perception through a behavior scene and primitive scene actions.
Action primitives build on a whole-body controller that supports moving the arms while walking, and use a concurrent action layering algorithm for speed.

Door traversals are used as the main benchmark task because they expose the full coordination problem in a compact and repeatable setting.
They require approach walking, body placement, mechanism perception, grasp selection, handle actuation, interaction with a moving obstacle, and a transition back to locomotion.

The behavior library developed during this work covers more than twenty real-robot task variants, including push and pull doors with knob, push-bar, and lever-handle mechanisms, multi-step exploration sequences, obstacle clearing, and reactive table-to-table manipulation tasks.
This behavior system has been deployed on many humanoid robots, such as Boston Dynamics' DRC Atlas, NASA's Valkyrie, IHMC and Boardwalk Robotics' Nadia, Unitree's H1-2, and IHMC's Alex.

We evaluate our system across capability, speed, reliability, and speed of behavior creation, adaptation, extension, and combination.
We executed our fastest door traversal in 14 seconds and, in a manipulation demonstration, sorted 6 balls by color in 42 seconds under human disturbance.
We demonstrated 11 successful push door approach-and-opening repetitions in a row and 12 successful pull door approach-and-opening repetitions in a row.
Measured authoring sessions show that we can scratch-author new loco-manipulation behaviors such as push door traversal in hours.
Our experiments also demonstrate that we can adapt, extend, and combine existing behaviors to create novel loco-manipulation behaviors in minutes or hours.
Finally, we compare our results against measured results from the literature and find our approach to be competitive with the latest results from learned systems.
Videos of the work presented in this dissertation are available online at \url{https://www.youtube.com/playlist?list=PLJK5CTyotYqsfgfnXb-09YNFeBose6uEY}.

}

\begin{document}

\chapter{Introduction}
\label{ch:introduction2}

\section{Introduction}

There is tremendous value in humanoid robots taking on the dull, dirty, and dangerous work in spaces built for humans.
However, a useful robotic system must coordinate locomotion, whole body motion, perception, contact, and operator supervision.
It must also support adaptation to new tasks.
This chapter introduces the problem addressed by this dissertation and situates the proposed behavior architecture in the relevant literature.

The central claim of this thesis is that our behavior architecture enables fast, resilient, and adaptive humanoid robot behaviors.
Different architectural choices affect how quickly a behavior executes, how robustly it tolerates task variation and disturbance, and how much effort is required to adapt an existing behavior to a new variant.
The work therefore focuses on the runtime structure that makes the task faster, more resilient, and easier to modify than the DRC-era behavior stack~\cite{Johnson_2017} and our prior published behavior architecture~\cite{calvert2024behavior}.
In \autoref{ch:building}, we tell the story of how we arrived at our current design choices.
In \autoref{ch:architecture2}, we'll present our current behavior system architecture.

\begin{figure}[H]
    \centering
    \includegraphics[width=0.62\textwidth]{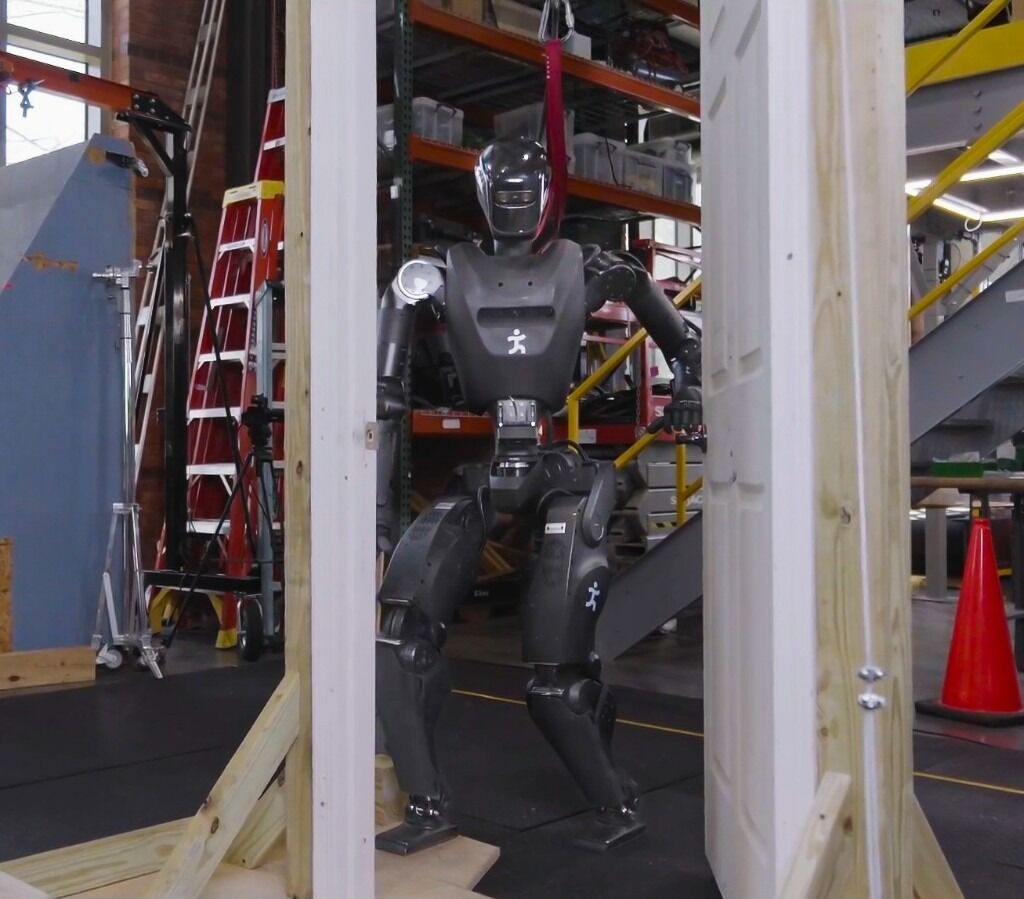}
    \caption{IHMC's fully electric Alex humanoid robot traversing a right pull door.
    Alex is the primary platform in the final dissertation evaluations.
    A video is available at \url{https://youtu.be/fP-9DfGFvW8}.}
    \label{fig:20260309_AlexRightPullDoorProfessionalStill}
\end{figure}

This thesis demonstrates its relevance through performant real-robot demos on a variety of humanoid robot platforms, including IHMC's recently developed Alex, a fully-electric humanoid robot with 29 degrees of freedom, as shown in \autoref{fig:20260309_AlexRightPullDoorProfessionalStill}.
Alex uses the PSYONIC Ability Hands, which are anthropomorphic 5-finger hands with 6 degrees of freedom each.
It also perceives the world on-board, using just two passive stereo color cameras in the head with a human-like interpupillary distance.
We present an evaluation of our results in \autoref{ch:defense}.

We have run our system with several humanoid robots over the years, including the Boston Dynamics DARPA Robotics Challenge Finals-Era Atlas, NASA's Valkyrie, IHMC and Boardwalk Robotics' Nadia, Unitree's H1-2, and IHMC's Alex.
This reflects the generality of our approach: it can run on a variety of humanoid robots.
We describe the prerequisites in more detail in \autoref{ch:tutorial}.

\section{Problem Statement and Scope}

This dissertation addresses the problem of local robot behavior authoring and execution for humanoid loco-manipulation tasks.
The specific setting considered here is a robot operating in human scale environments, where it must walk, reach, manipulate objects, perceive doors and tables, and respond to operator input without relying on external tracking infrastructure.
The scope is centered on a robot-local behavior system in which operators can compose, edit, and retarget behaviors.

The work presented here focuses on a behavior architecture that unifies task execution, authorable perception, and operator-robot teaming.
In this architecture, behaviors are authored as reusable structures that can be executed on the robot and inspected and modified at runtime.
This design can improve task speed, robustness under variation, and the time required to adapt a behavior to a new task variant such as a door or station.
In \autoref{ch:tutorial}, we provide a guide on composing and editing humanoid robot loco-manipulation behaviors using our system.

We compare our system to prior versions over a 10-year development period and to published results in the literature.
However, we do not experimentally reproduce results from the literature.
We also do not experimentally evaluate off-the-shelf alternatives such as MoveIt \cite{MoveIt_2019}, MoveIt Pro \cite{moveit_pro}, BehaviorTree.CPP \cite{behaviortree_cpp}, and Groot \cite{groot1, groot2}.
We present related work in \autoref{ch:prior_work}.

\section{Research Questions}

This scope can be defined more formally by asking three research questions:
\begin{enumerate}
    \item What concrete behavior architecture is sufficient to enable fast and robust performance across a range of humanoid loco-manipulation behaviors?
    \item How does this architecture compare with prior IHMC baselines and reported reinforcement learning door systems on overlapping metrics such as traversal time, reliability, and task variation coverage?
    \item How does runtime-editable behavior structure change the time and sequence of steps required for an expert operator to create new behaviors and adapt, extend, and compose existing ones?
\end{enumerate}

\section{Research Hypotheses}

To answer these questions, we establish the following research hypotheses, which are essentially the characteristics of the architecture we committed to building.
\begin{enumerate}
    \item Robot-local execution with synchronized UI state, concurrent action layering, reactive tree logic, and behavior-time semantic perception yield door behaviors that are faster and more reliable than prior IHMC baselines and competitive with reported reinforcement learning systems on overlapping door tasks.
    \item Runtime-editable behaviors and perception modules reduce the iteration loop required to diagnose failures, modify logic, and re-test on the robot, relative to redeploy, restart, or retrain workflows.
    \item Decomposing behaviors into reusable primitives, subtrees, and scene actions allows new door and loco-manipulation variants to be brought up by editing a small part of a working behavior rather than rebuilding it from scratch.
\end{enumerate}

\section{Three Pillars}

More simply, and as the core metrics by which we will judge and organize this work, we present the three pillars of this thesis: Speed, Resilience, and Adaptability.
Speed means fast execution of loco-manipulation behaviors.
Resilience means robust and reactive behavior under disturbance and task variation.
Adaptability means runtime-editable behaviors for rapid task retargeting.
These pillars provide the main lens for evaluating the architecture and the results presented throughout the dissertation.
In \autoref{ch:metrics}, we cover the full set of desirable characteristics of a humanoid robot behavior architecture.

\section{Contributions}

The main contributions of this dissertation are:
\begin{itemize}
    \item \textbf{Architecture:} We present a robot-local behavior architecture that unifies runtime-editable task logic, synchronized operator UI state, and behavior-time perception for humanoid loco-manipulation, evaluated through repeated real-robot demonstrations on multiple humanoid platforms, with Alex as the primary evaluation robot.
    \item \textbf{Speed:} We report humanoid door traversals among the fastest timed results in the published literature, including sub-20-second full traversals on real hardware and performance competitive with recent learned humanoid door policies on overlapping tasks.
    \item \textbf{Combined evaluation:} To our knowledge, this is the first door-traversal study to report competitive speed and repeated-trial reliability on a humanoid while also measuring behavior authoring and adaptation time on the same runtime-editable stack.
    \item \textbf{Adaptability:} We show that an expert operator can bring a novel humanoid door behavior from an empty sequence to first fully autonomous success in under two hours of measured active authoring time, and we report comparable adaptation durations for retargeting existing behaviors to new doors and tasks.
    \item \textbf{Perception:} We introduce behavior-time perception modules for generalized door traversal and table approach, including visual door-state estimation across diverse door configurations and a depth-based table-edge detection method that yields a reusable approach frame for precise humanoid positioning.
\end{itemize}

\section{Door Traversals as a Benchmark Task}

\begin{figure}[ht]
    \centering
    \includegraphics[width=\textwidth]{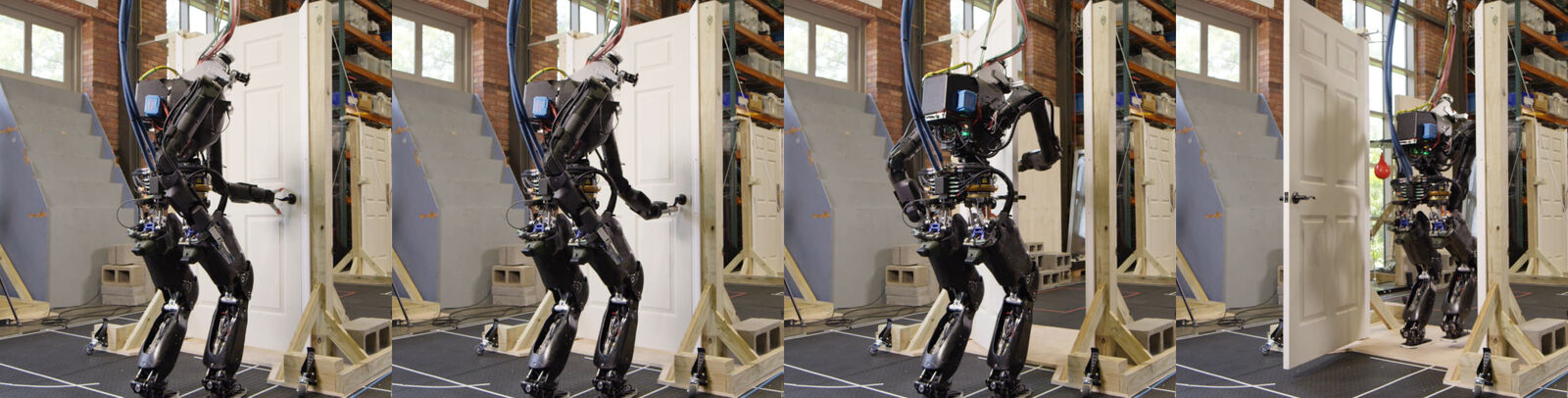}
    \caption{The Nadia humanoid robot performing a right pull lever handle door traversal using cycloidal drive forearms and Sake grippers.}
    \label{fig:2024NadiaRightPullLeverHandleCycloid}
\end{figure}

We use door traversal as a benchmark task because it exposes the full coordination problem in a compact and repeatable setting.
There are three main phases of door traversal: the approach, the opening, and the traversal walk.
For the approach, footsteps must be precise to provide arm reachability while avoiding collisions between the door and the robot.
Since the doors often have spring closers, a complex manipulation sequence is required to open the door all the way without failure.
Finally, the traversal walk is difficult because the robot must maintain balance while avoiding collisions with the door frame and resisting lateral impacts from the spring loaded door panel.
\autoref{fig:2024NadiaRightPullLeverHandleCycloid} shows a circa-2024 door behavior being executed on the Nadia humanoid robot.

There are many publications on door opening with wheeled base robots and arms, such as \cite{Jain2008dooropening}, \cite{2010ChittaDoorOpening}, \cite{xiong2024adaptive}, and \cite{kang2024door}.
By contrast, there is relatively little published work on door traversal with bipedal humanoid robots \cite{xue2025opening}.
That gap matters because humanoids must manage foot placement, balance, reachability, body orientation, and environmental contact at the same time.
The humanoid form is also especially well suited for navigating complex spaces designed for humans. \cite{Johnson_2017}.

\begin{figure}[H]
    \centering
    \small
        \begin{tikzpicture}[
        node distance=1.8cm and 0.5cm,
        category/.style={
            rectangle,
            rounded corners=3pt,
            draw=black!60,
            line width=0.8pt,
            text width=3.6cm,
            minimum height=1.0cm,
            align=center,
            font=\bfseries\small,
            inner sep=4pt
        },
        behavior/.style={
            rectangle,
            rounded corners=2pt,
            draw=black!40,
            line width=0.5pt,
            text width=9.2cm,
            align=left,
            font=\small,
            inner sep=5pt,
            fill opacity=0.18,
            text opacity=1
        },
        arrow/.style={
            -{Stealth[scale=0.8]},
            line width=0.8pt,
            draw=black!50
        }
    ]

    \node[category, fill=teal!25] (cat1) {Door traversal\\families};
    \node[behavior, fill=teal!20, right=of cat1] (beh1)
        {Left push and right push push-bar traversals;\\
         right push knob traversal;\\
         right pull and left pull lever-handle traversals;\\
         hands-free pull-handle traversal;\\
         standing right pull lever-handle opening-only behavior.};
    \draw[arrow, draw=teal!60!black] (cat1.east) -- (beh1.west);

    \node[category, fill=orange!25, below=of cat1] (cat2)
        {Multi-stage door and\\exploration behaviors};
    \node[behavior, fill=orange!20, right=of cat2] (beh2)
        {Three-door autonomous sequence on Nadia;\\
         ONR mock-building exploration across four rooms with\\
         successive door traversals and room-search decisions;\\
         received object from person and place on door;\\
         bottle pickup with push-door carry-through.};
    \draw[arrow, draw=orange!60!black] (cat2.east) -- (beh2.west);

    \node[category, fill=violet!25, below=of cat2] (cat3)
        {Obstacle clearing and\\breaching behaviors};
    \node[behavior, fill=violet!20, right=of cat3] (beh3)
        {Recycling-bin / trash-can doorway clearing;\\
         couch moving;\\
         kick-door-down breaching behavior.};
    \draw[arrow, draw=violet!60!black] (cat3.east) -- (beh3.west);

    \node[category, fill=olive!25, below=of cat3] (cat4)
        {Ball and table\\manipulation behaviors};
    \node[behavior, fill=olive!20, right=of cat4] (beh4)
        {Reactive single-table ball sorting by color;\\
         long-duration public open-house ball-return loop;\\
         two-table loco-manipulation ball sorting with\\
         table-to-table transport.};
    \draw[arrow, draw=olive!60!black] (cat4.east) -- (beh4.west);

    \end{tikzpicture}
    \caption[Behavior-library taxonomy by category.]{Distinct real-robot behavior types developed during this work, grouped by category.}
    \label{fig:behavior_library_taxonomy}
\end{figure}
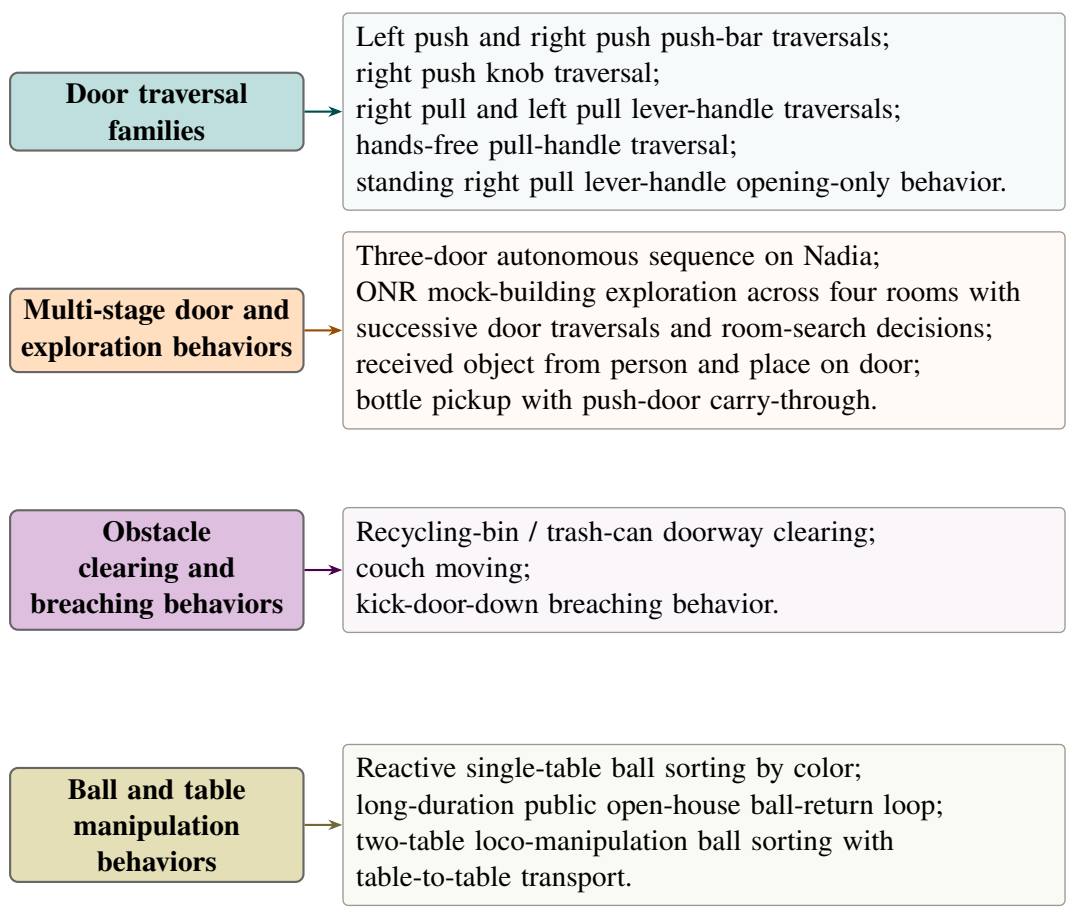

Although doors are the primary benchmark, the architecture is intended for loco-manipulation more broadly.
The behavior library developed during this work spans more than twenty real robot task variants.
In one demonstration, the robot walks between two tables and sorts colored balls into the correct containers.
\autoref{fig:behavior_library_taxonomy} groups the distinct behavior types by category.

\section*{Contributed Works}
\addcontentsline{toc}{section}{\textbf{Contributed Works}}

We contributed to the following publications while working on this thesis.
The first paper in this list received the Humanoids 2022 Best Oral Paper Award \cite{humanoids2022_awards}.
The second paper in this list earned a Bronze Medal in Benjie Holson's Humanoid Olympic Games for a round knob push door traversal timed at 18~s \cite{holson2025_ihmc_bronze}.

\begin{enumerate}

    \item \textbf{Duncan Calvert}, Bhavyansh Mishra, Stephen McCrory, Sylvain Bertrand, Robert Griffin and Jerry Pratt (2022). A Fast, Autonomous, Bipedal Walking Behavior over Rapid Regions. \emph{2022 IEEE-RAS 21st International Conference on Humanoid Robots (Humanoids)} \cite{lookandstep}.

    \item \textbf{Duncan Calvert}, Luigi Penco, Dexton Anderson, Tomasz Bialek, Arghya Chatterjee, Bhavyansh Mishra, Geoffrey Clark, Sylvain Bertrand and Robert Griffin (2024). A Behavior Architecture for Fast Humanoid Robot Door Traversals. \emph{Robotics and Autonomous Systems} \cite{calvert2024behavior}.

    \item \textbf{Duncan Calvert}, Luigi Penco, Dexton Anderson, Tomasz Bialek, Arghya Chatterjee, Beomyeong Park, and Robert Griffin (2026). A System for Fast, Resilient, and Adaptable Loco-Manipulation Behaviors on Humanoid Robots. \emph{IEEE Robotics and Automation Letters} (in preparation) \cite{calvert2026resilient}.

    \item Bhavyansh Mishra, \textbf{Duncan Calvert}, Sylvain Bertrand, Jerry Pratt, Hakki Erhan Sevil, and Robert Griffin (2024). Efficient Terrain Map Using Planar Regions for Footstep Planning on Humanoid Robots. Accepted for publication in the \emph{Proceedings of the 2024 IEEE International Conference on Robotics and Automation (ICRA)} \cite{Mishra_2023_SKIPR}.

    \item Bhavyansh Mishra, \textbf{Duncan Calvert}, Brendon Ortolano, Max Asselmeier, Luke Fina, Stephen McCrory, Hakki Erhan Sevil, and Robert Griffin (2022). Perception engine using a multi-sensor head to enable high-level humanoid robot behaviors. Published in the \emph{Proceedings of the 2022 International Conference on Robotics and Automation (ICRA)} \cite{Mishra_2022_PerceptionEngine}.

    \item Bhavyansh Mishra, \textbf{Duncan Calvert}, Sylvain Bertrand, Stephen McCrory, Robert Griffin, and Hakki Erhan Sevil (2024). GPU-Accelerated Rapid Planar Region Extraction for Dynamic Behaviors on Legged Robots. Published in the \emph{Proceedings of the 2021 IEEE/RSJ International Conference on Intelligent Robots and Systems (IROS)} \cite{Mishra_2021_RapidRegions}.

    \item Luigi Penco, Kazuhiko Momose, Stephen McCrory, Dexton Anderson, Nicholas Kitchel, \textbf{Duncan Calvert}, and Robert J Griffin (2024). Mixed reality teleoperation assistance for direct control of humanoids. \emph{IEEE Robotics and Automation Letters}, 9(2), 1937-1944 \cite{penco2024mixed}.

    \item Sylvain Bertrand, Luigi Penco, Dexton Anderson, \textbf{Duncan Calvert}, Valentine Roy, Stephen McCrory, Khizar Mohammed, Sebastian Sanchez, Will Griffith, Steve Morfey, Alexis Maslyczyk, Achintya Mohan, Cody Castello, Bingyin Ma, Kartik Suryavanshi, Patrick Dills, Jerry Pratt, Victor Ragusila, Brandon Shrewsbury, and Robert Griffin (2024). High-Speed and Impact Resilient Teleoperation of Humanoid Robots. \emph{2024 IEEE-RAS 23rd International Conference on Humanoid Robots (Humanoids)} \cite{bertrand2024highspeed}.

    \item Stephen McCrory, Sylvain Bertrand, Achintya Mohan, \textbf{Duncan Calvert}, Jerry Pratt, and Robert Griffin (2023). Generating humanoid multi-contact through feasibility visualization. \emph{2023 IEEE-RAS 22nd International Conference on Humanoid Robots (Humanoids)} \cite{mccrory2023generating}.

    \item Sylvain Bertrand, Inho Lee, Bhavyansh Mishra, \textbf{Duncan Calvert}, Jerry Pratt, and Robert Griffin (2020). Detecting Usable Planar Regions for Legged Robot Locomotion. \emph{2020 IEEE/RSJ International Conference on Intelligent Robots and Systems (IROS)} \cite{Bertrand_2020}.

\end{enumerate}

\chapter{Desirable Characteristics of a Behavior Architecture}
\label{ch:metrics}

In this chapter, we'll discuss some of our dream requirements for a behavior architecture to establish our goals and the desired properties of a particular implementation.
In this thesis we focus on behavior systems that require human expertise to dream up, create, adapt, and modify.
It is conceivable that a generally intelligent AI could replace this role in the future, but nevertheless we do not consider that here.
This list of characteristics is therefore in the realm of Operator-Robot teams, be it any number of humans and robots.
For complex tasks that require domain expertise, it may be desirable for many humans, at times, to manage a single robot. Generally though, our preference would be for a few humans to manage a fleet of robots.

\section{Capability}
The goal of a behavior system should be to support doing as many tasks as possible to help it achieve maximum utility.
The whole point of a robot behavior system, as we are concerned with it in this thesis, is to fill in for the dull, dirty, and dangerous work humans do.
We define capability as how many different tasks and their variations can be performed successfully.
For example, a system that only supports door traversals is not as capable as one that supports exploring buildings.

\section{Feasibility}
Any implementation of a behavior system must be feasible given real-world constraints.
It is desirable to not require overly expensive computers or ones that are not readily available.
If the robot needs to function autonomously in comms-degraded scenarios, the behavior should not rely on external comms or compute to operate.
The behaviors cannot require robot actuation hardware that is not reliable, readily available, or that does not exist.
For example, there should be no jetpack flying requirement for behaviors if robots with jetpacks and controllers for them do not exist or only exist as prototypes.
The behaviors should not rely on control software that does not exist.
For example, modern whole-body controllers do not do much in the way of planning out how to achieve complicated bracing positions and techniques to avoid falling in dynamic scenarios, so the behavior system should support getting the robot into positions and authoring at a primitive action level that allows the operator to reason about what the controller can handle while authoring the behavior actions.

\section{Speed}
Behaviors should be watchable at 1x speed.
Computational components of the system need to run within their allotted time boundaries, not causing any pauses.
The robot hardware and the whole-body controller that the behavior system relies on should be capable of decently fast motions.
We don't mean multiples of human speed, just approaching casual human speed in performing day-to-day chore-like tasks.
We want robots to be a drop-in for human work without an immediately huge tradeoff or question mark on speed.

\section{Parallelizability}
The system should support moving multiple parts of the body at once and the ability to walk while doing that.
This is particularly useful when doing manipulation.
Before performing many manipulations, the robot will need to prepare the grasping arm in a pre-grasp-ready pose while putting the other arm in a collision-avoidance pose.
Having to get into these ready poses sequentially would cause an unnecessary delay.
This goes towards the speed metric, but is also important for capability.
For example, for traversing spring-loaded doors, the robot must walk while keeping its arm out in the correct locations to prevent the door from closing on the robot.

\section{Reliability}
Robots should be able to execute tasks repeatably without failures.
More formally, for a given task, given similar environmental conditions, the robot should consistently perform the task.
If it succeeds, it should consistently succeed (and if it fails, it should consistently fail).
In other words, operators should be able to count on the robot to perform a task repeatedly without random failures that have little to do with environmental variance. The system should approach being deterministic but there is no need to formally satisfy a determinism claim.

\section{Robustness}
The robot and behavior system should be robust to environmental disturbances.
These can be both physical and visual.
For example, slight pushes to the robot, unmodeled friction and inertias in manipulated items, and changes in lighting due to time of day should not cause task failures.
Robustness mechanisms should be present via the whole-body controller or in the behavior system to address these.
Any vision models should be trained using data from varied times of day and lighting conditions to prepare them for round-the-clock work.

\section{Resilience}
The behavior system should support resilience to changing task conditions and attempt to recover or work around gaps in reliability and robustness.
We define resilience to mean being responsive and creative when facing task failures.
For example, when the robot is trying to turn a door handle, perhaps a human is present and trying to test the reactivity of the system.
In this case, the behavior system should be able to identify that the task is not proceeding nominally and enact some retry strategy.
Retry strategies can include simply retrying the action sequence, mutating the pose-grasp sequence, or aborting the mission entirely and doing something else.
Resilience ultimately means surviving day-to-day unexpected events and failures.
Attaining resilience is a long-tail robotics problem, but the prior examples are good places to start.

\section{Independence from External Systems}
Ideally, the robot can execute behaviors without being dependent on external compute or network connections, given the behaviors have been authored and set up ahead of time.
This mirrors animals in nature (they aren't digital) and supports a level of robustness by removing an unintuitive dependence on network communication.
It also allows the robot to operate in more environments, including inside buildings with thick concrete walls and rural areas.

Some behavior systems rely on external perception.
It is desirable to perceive the world only via the robot and not be dependent on motion capture systems or fiducial markers, which are common in laboratory setups.
We want the robots using our behavior system to thrive beyond the lab environment and provide useful service in the real world.
It is also desirable to have humanoid robots be a drop-in replacement for human workers without having to make robot-specific adjustments to the environment, such as placing fiducial markers.
It would be better if robots were to read the same signage and maps as humans.

\section{Dependence on Only Passive Color Vision}
By using only passive color vision, the robot mirrors human nature and is more robust to varied lighting conditions.
For example, structured light projection sensors can have degraded performance outdoors and in the presence of certain frequencies of light.
Additionally, it can be more intuitive and understandable when the robot's vision modality is similar to human vision. For example, when it is too dark for humans to see, it's obvious that it's too dark for the robot to see.
Since we are building humanoid robots to fill in for humans, it could be a more surefire drop-in replacement by matching the mode of vision.

\section{Adaptability of the Operator-Robot Team}
Given the near infinite world of tasks robots could help us with, we want a behavior system that can support creating new and adapting existing behaviors to tackle them.
Adaptability means being able to survive in a changing environment and, for robots, this means operators must be able to readily adapt robot behaviors to changing needs.
One of the selling points of humanoid robots is their generality and similarity to humans, which means one application of them is to fill in for human work.
Given the adaptiveness of humans and the existence and competitiveness of purpose-built machinery, the value-add for robots must exist in the realm of being an adaptable generalized form.
Therefore, it is desirable to be able to create and modify behaviors to tackle various dull, dirty, and dangerous work tasks in a quick time frame.

The following three characteristics are inherently required in building the adaptability components, as defined by Coactive Design~\cite{Johnson_2014}.

\section{Observability}
This means knowing the current state of the system in order to understand what is going on.
For a robot, there is a lot of information to take in at any particular moment and there are different levels of granularity in doing so.
Knowing the current state is required for a human operator to reason about the behavior to modify or adapt it.
It is also required to monitor what the robot is accomplishing and determine if it needs help.
Let's list some of the biggest ones:
\begin{enumerate}
    \item Seeing the current configuration of the robot's body and hands, visual elements that indicate current forces on the environment, robot hardware status, motor temperatures, and joint faults.
    \item Seeing what the robot sees such as the current robot view video stream(s) and current semantic object identification.
    \item Getting a feel for the robot's immediate environmental surroundings and how the robot is situated in it.
          For example, this can be done via a colored depth point cloud in the 3D view with the robot configuration.
          If the robot doesn't have 360-degree vision, mapping may be required or the robot could move the head around to rescan.
    \item Knowing the current state of the behavior system and whole-body controller.
          For example, what state is the behavior in?
          Is anything currently executing?
          Has anything failed?
          What have we done in the recent past and was it successful?
    \item Knowing the robot's current model of the environment.
          Which objects does it know about?
          Where does it think they are in 3D?
          Does the robot know where it is on a map?
          Is it aware of the major obstacles nearby?
\end{enumerate}

\section{Predictability}
This one is about a sense of what is going to happen next both with the robot and the environment.
This is required for a human operator to create, adapt, and diagnose behaviors.
For example, when authoring the next action(s), it is desirable to see a preview of the motion of the robot and environment as a way of verifying that action.
The preview can be inspected for collisions or bad inverse kinematics solutions to avoid failures before executing it on the real robot.
Predictability also goes hand-in-hand with authoring at runtime and in mission-critical field scenarios.
For example, in the DARPA Robotics Challenge, many tasks, such as getting out of the car, were a step-by-step sequence of predefined robot motions.
The robot was being teleoperated live and if the robot fell, the competition would be lost.
When doing this task, a preview of the next motions allowed the team to inspect the plan before executing it, increasing confidence and the reliability of the operator-robot team.

On a technical level, it is possible to provide predictability of whole-body motions by playing back a planned animation of future motion, as a transparent colored robot.
Primitive graphics like footsteps can be shown to convey where the robot will step next.
Color can be used to convey feasibility.
For example, a blue transparent graphic of an arm can represent a feasible solution and it can turn red to notify the operator of an infeasible or hard-to-reach configuration.
On a longer horizon, a browsable list of actions could be shown which lists all future actions and sub-sequences in the behavior.

\section{Directability}
The last of Johnson's three characteristics in Coactive Design is directability.
It is a measure of how expressive the operator can be in commanding the robot to do things.
For a humanoid robot, at the basic level, it means being able to command the robot to take steps, walk, move its hands, look around with its head, and generally pose the whole body.
At a higher level, the availability of planners increases expressiveness.
Good examples of high expressiveness would be the ability to ask the robot to clean a room or to fetch a particular object.
We also extend the scope of directability to include non-direct ways of commanding the robot, such as tuning parameters of primitive actions, behavior logic, perception, or scene management.
In this way, we want to not only directly command the robot's physical actions, but also its cognitive model of the world and its plan.

\section{Learnability (Operator Learning Curve)}
The operator interface should be designed in a way that facilitates a novice operator in learning how to use it.
The behavior operator interface should be interactive and guide the user with cues to point them in the right direction and give them confidence that what they did is what they wanted to do.
Nielsen's 10 Usability Heuristics for User Interface Design is a good reference point for designing a user interface~\cite{nielsen1994ten}.
These 10 heuristics include: visibility of system status, match between the system and the real world, user control and freedom, consistency and standards, error prevention, recognition rather than recall, flexibility and efficiency of use, aesthetic and minimalist design, help users recognize, diagnose, and recover from errors, and help and documentation.

\section{Understandability (of the Implementation)}
It would be nice if it were easy to learn how the behavior system works by reviewing the code and observing behaviors in operation.
Viewing a behavior's composition in the user interface should give a good idea of what the behavior does by being organized and supporting abstractions.
The use of hierarchical abstractions, for example, can allow the reader to understand the high level at a glance and dive deeper where they want to learn more.

\section{Usability}
The user interface should be easy to use, even for an expert operator.
Functionality should be organized in meaningful ways, such as grouping like functionalities, scene objects by category, and organizing primitive actions by part of the body.
When behaviors get large, there should be mechanisms to abstract their contents into high-level parts.
One way to do this, for example, is to structure behaviors hierarchically, such that the higher level layers are more generic, like ``navigate to room C'', and lower level layers are more specific like ``move hand forward 5 centimeters''.
Functionalities should be organized into menus.
Buttons and checkboxes should be easy to click.
Text and widgets should be easy to read and size-adjustable.

\section{Ability to Analyze in Post}
A lot can happen in the course of a robot run and sometimes it can happen very fast.
When there are failures or potential improvements, it is often useful to do a post-mortem analysis of what happened.
This characteristic is desirable especially because running and supervising robots is stressful and requires attention.
We want the ability to log all the data for a robot run and dive into that data later, without the cognitive overhead of the live run.
This also gives the operator or behavior engineer the opportunity to view the system in non-live ways.
They do not necessarily need the same observability data; they can choose to deep dive into control or logic data, using screen real estate for that instead of live monitoring.
The logged data should include (ideally lossless) recordings of the robot's sensors, behavior state and parameter data over time, controller variables over time, robot configuration state over time, and more.
This also implies the availability of post-mortem analysis software, which would allow the logged data to be explored in a rich and interactive way.
Examples of this include a slider to scrub data over time, a 3D reconstruction of robot configuration state and 3D depth data, and time plots of controller and behavior logic variables.

\section{Debuggability}
The system should provide outputs that assist in debugging when things go wrong or while bringing up new functionalities.
Examples of this include good print statements in the robot processes and logging them, sending log messages from the robot to the user interface at runtime, and coloring the log messages by severity and importance.
Dynamic user interface elements can also be helpful, for example, when an action fails, making it blink red to draw the operator's attention.
Another way to support debugging is to carefully select a representative set of state variables to log in a time-dependent buffer.
These buffers can be streamed live or stored to disk and viewed as scrubbable plots.

\section{Testability}
It is desirable to be able to test the system in an automated way.
This could be with the real robot, virtually with real data, or using fully simulated data.
For example, having test fixtures available for code continuous integration tools to perform simulated behaviors and inspect the results for success and performance characteristics would be helpful to ensure quality and prevent regressions.
Testing often requires significant resources as in the case of real robot automated testing.
To support these cases, the behavior system should be able to be operated in an automated way and not just by a human operator.

Another case that requires significant resources is fully simulated testing.
It can be very difficult to reduce the sim-to-real gap for loco-manipulation behaviors that need realistic vision and physics.
Tasks often need to be rigged as articulated simulation assets as in the case of doors, which is a manual process that requires expertise.
However, there is also a middle ground in which tradeoffs can be made and components replaced with dummies.
For example, poses of objects in the scene could be given via ground truth knowledge, bypassing the vision system entirely or partially.

\section{Extendability}
We're still firmly in the early stages of humanoid robots starting to work well.
Any system for running behaviors on humanoid robots should be easy to extend, functionality-wise, to keep pace with the state of the art and maintain competitive usefulness.
For example, given the availability of a new footstep planner, it should be a straightforward process to include it as an option.
Likewise, if a new comms protocol is adopted, it should not require a complete redesign of the architecture to switch over.
Some ways that could help in achieving extendability are keeping the code well tested and maintaining separation of concerns in the design and implementation.

There are a lot of different ways to achieve these characteristics, and in some sense there are tradeoffs depending on the specific requirements.
The tradeoffs could be in engineering time or they could be theoretical.
For example, if a system will only be used by trained expert operators, more engineering time can be invested in the functional and utilitarian aspects like reducing number of clicks or relying more heavily on keyboard shortcuts.
However, if a system needs to be usable by a more general audience, more engineering time needs to be spent on Nielsen's 10 heuristics and a user study may even be warranted.

An example of a more theoretical tradeoff would be what to show to the operator at any given time.
There is only so much screen real estate and operator attention that can go around, so hard decisions need to be made about the value of information.
We think this will vary from system to system and ultimately is based on the confidence levels of the particular subsystems.
For example, if there is high trust in the controller's ability to walk and balance, balance information may not need to be shown to the operator.
Conversely, if the system depends on a semantic object detection subsystem that is always failing to detect objects, that subsystem will likely need to be visible in high detail at all times so operators can monitor and learn how to either exploit its properties or make informed improvements.

Now that we have defined some desirable characteristics of a good behavior architecture, we'll tell the story of our journey in navigating the tradeoffs and building one from near-scratch that met our requirements.

\chapter{Related Work}
\label{ch:prior_work}

\subsection{Introduction}
In this chapter we'll cover literature that relates to this dissertation on reusable action structure, reactive task coordination, human-machine coordination, perception, and real robot door task demonstrations.
Architectural references include CLARAty~\cite{Volpe_2001_CLARAty}, Affordance Templates~\cite{Hart_2014, Hart_2015}, Affordance Primitives~\cite{Pettinger_2020, Pettinger_2022}, Coactive Design~\cite{Johnson_2014}, Director~\cite{Marion_2017}, FlexBE~\cite{Schillinger_2016}, RAFCON~\cite{Brunner_2016}, Drawing Board~\cite{Senft_2021}, and the CENTAURO behavior tree systems~\cite{2023_centaur_bt, 2024_wang_grounded_lm}.
Door task references span from classical and model-based mobile-manipulator and humanoid systems to recent learned legged and humanoid policies.
The systems with metrics that overlap our benchmark are compared with each other and this thesis in \autoref{ch:defense}.

\autoref{fig:design_space_map} maps every system reviewed in this chapter across four dimensions: execution boundary, runtime editability, perception model, and door evidence.
The execution boundary refers to where the process that controls the robot runs.
It can be off-board, operator-in-the-loop, or purely on-board/robot-local.
This is one part of qualifying our ``Independence from External Systems'' characteristic in \autoref{ch:metrics}.

In \autoref{fig:design_space_map}, each row is one system.
The filled cells place that row on the execution-boundary axis (left band) and the runtime-editability axis (right band).
The two rightmost columns encode the perception model used at task time, distinguishing behavior-time authored perception from a fixed onboard pipeline, learned end-to-end perception, and reliance on external measurement, alongside the strength of the system's door-task evidence on a four-level scale from none to repeated-trial real-world.
This thesis is the only row that combines robot-local execution, runtime structural and perception edits, behavior-time authored perception, and repeated-trial real-world door evidence.
DoorMan is the only off-board entry, and the learned door systems concentrate on the policy-retrain column rather than runtime structural edits.

\begin{figure}[p]
    \centering
    \input{tikz/DesignSpaceMapTikz.tex}
    \caption[Design-space placement of prior systems in this chapter.]{%
        Design-space placement of every prior system reviewed in this chapter.
        }
    \label{fig:design_space_map}
\end{figure}
\clearpage

\section{Architectural Foundations}

Affordance Templates, Behavior Trees, and Coactive Design form the strongest theoretical foundations of the work in this thesis.
Affordance Templates define an architecture for reusable behavior with respect to recurring tasks in the environment.
Behavior Trees provide a data structure for organizing and orchestrating behavior.
Coactive Design poses three questions that drive the design of a system to leverage the synergistic opportunities of human-robot teams.

\subsection{Affordance Templates}
\label{sec:ats}

The Affordance Template Framework~\cite{Hart_2014, Hart_2015} was developed and used for the DARPA Robotics Challenge Trials in 2013.
Affordance templates were first presented in the literature in 2014.
The work was performed on NASA's Valkyrie humanoid robot, where it demonstrated a valve turning task.
The Affordance Template framework provides a way to parameterize and reuse robot loco-manipulation behaviors with respect to environmental affordances.
Hart et al.~\cite{Hart_2014, Hart_2015} developed a set of affordance templates at NASA Johnson Space Center for Valkyrie during the 2013 DARPA Robotics Challenge Trials, including templates for door opening, walking, hose mating, valve turning, and ladder and stair climbing.

The word ``affordance'' comes from James Gibson's 1979 book, \emph{The Ecological Approach to Visual Perception}~\cite{Gibson_1979}.
It is used in the sense that an environmental feature ``affords'' an action.
For example, the ground is an affordance for walking and a handle is an affordance for turning.
An affordance template is a robotics construct providing a theoretical foundation for defining and parameterizing robot behavior with respect to environmental affordances.
This makes it one of the most consequential design inspirations for our work.

\begin{figure}[H]
    \centering
    \includegraphics[width=.7\columnwidth]{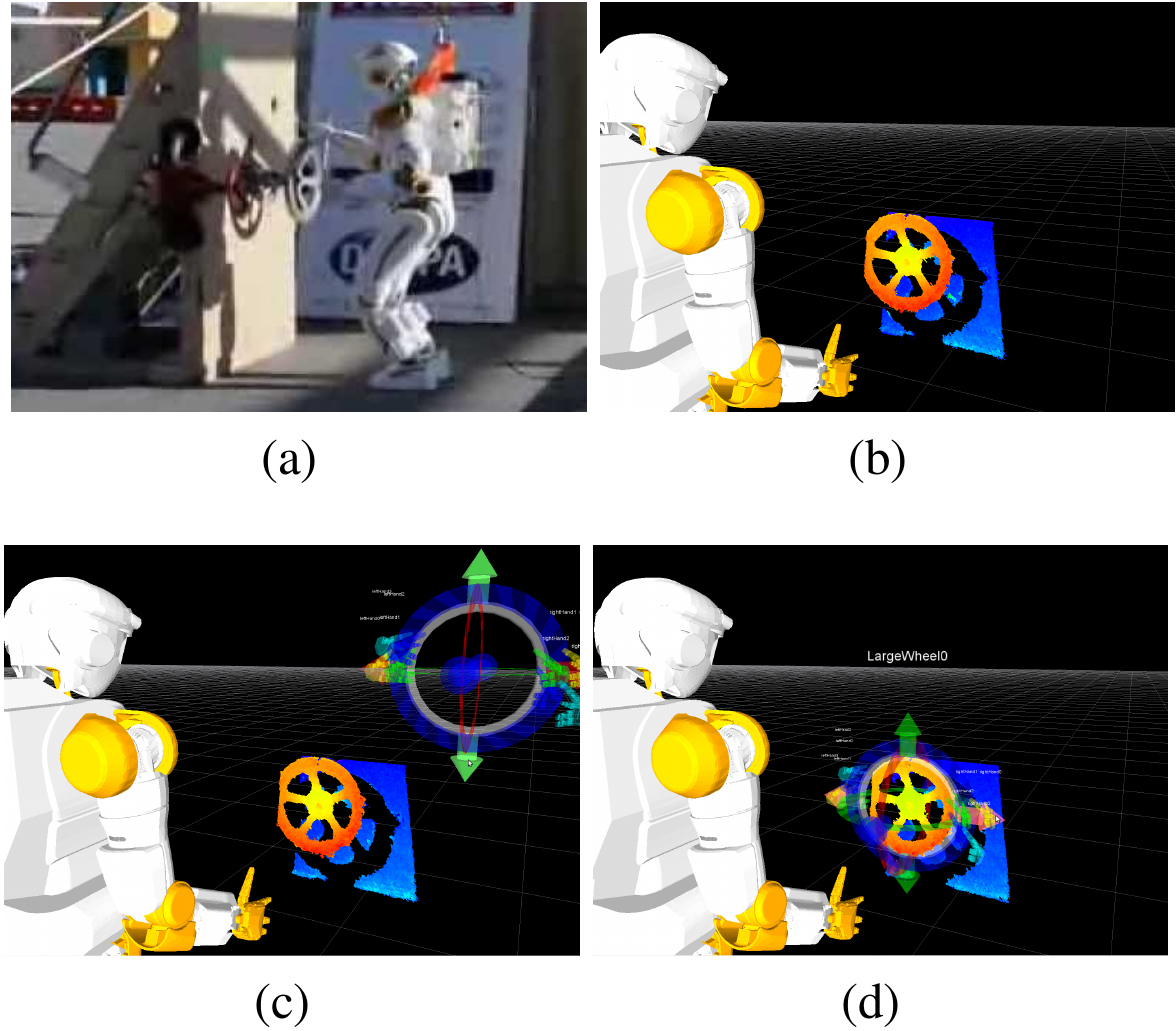}
    \caption{``Placing a wheel-turning template in RViz.'', taken from~\cite{Hart_2014}.
        (a) shows NASA Valkyrie turning a valve using an affordance template in the DARPA Robotics Challenge Trials.
        (b) shows the perceptual 3D point cloud scan, colored by depth.
        (c) shows a virtual affordance template before it is aligned to the object.
        (d) shows the affordance template after it has been aligned to the object by a human operator.
    }
    \label{fig:affordance_template_reference}
\end{figure}

\begin{figure}[H]
    \centering
    \includegraphics[width=.7\columnwidth]{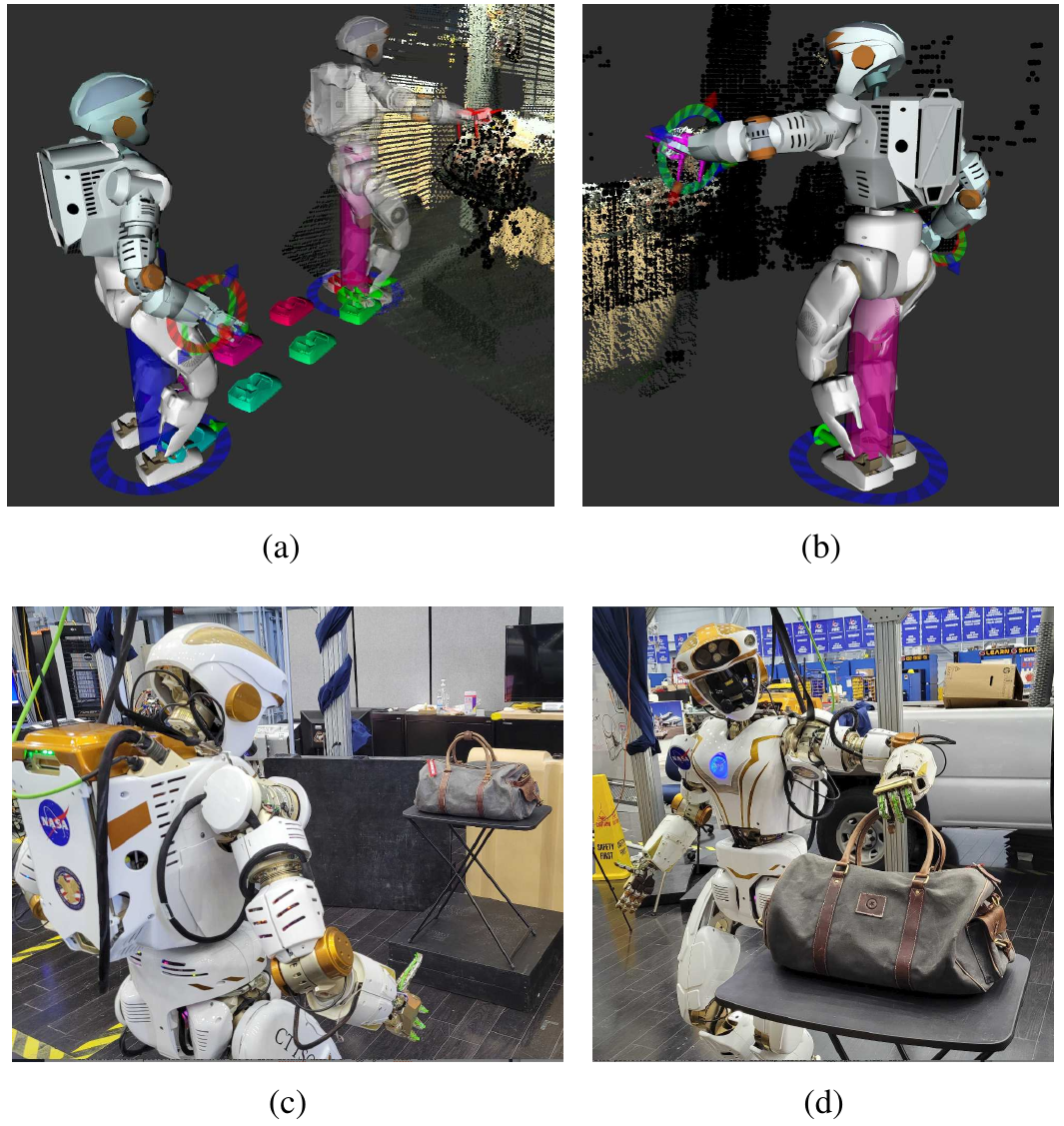}
    \caption{``NASA Valkyrie using an AT to remove a bag from a table'', taken from~\cite{Hart_2022}.
        (a) shows the alignment of an affordance for picking up a bag to the point cloud.
    Tools are then invoked for footstep planning, stance generation, and manipulation and visualized as shown.
        (b) shows the robot's state at the conclusion of the bag pickup task.
        (c) shows the state of the robot and the environment at the start of the task and (d) shows it at the end.
    }
    \label{fig:affordance_template_2}
\end{figure}

In 2022, Hart et al.~\cite{Hart_2022} generalized and further demonstrated the power of this approach, as shown in \autoref{fig:affordance_template_2}.
Tools for footstep planning, motion planning, stance generation, and grasping were incorporated to enable increased autonomous functionality.
This was demonstrated on NASA Valkyrie for integrated task execution, including car-door interaction and explosive-device handling~\cite{2019_Jorgensen_valEOD}.
In that work, the operator could load, edit, and execute templates at runtime.

Related works on affordance primitives~\cite{Pettinger_2020, Pettinger_2022} emphasize constrained interaction motions such as turning a valve or closing a drawer.
The screw primitive in particular shows how a reusable low-level action can capture a class of constrained manipulation motions without hard coding them.

\subsection{Behavior Trees}

Behavior Trees~\cite{2018_colledanchise, 2022_iovino_behavior_trees} are a data structure that coordinates low-level actions through structured and reactive task logic.
They define logical operator nodes such as sequence, fallback, and parallel which work to control the execution flow of a behavior.
Actions are performed by the leaf nodes which command the robot and gather environmental data.
They provide reactivity through a ticking system in which each tick starts at the root node.
This is in contrast to state machines in which each tick starts from the previous state.
By continuously re-evaluating from the top down, from tick to tick there are pathways to end up in very different parts of the tree without explicit connections between those parts.

A Behavior Tree is ticked from the root node at a constant frequency.
The tick is a signal which propagates through the tree, causing nodes to evaluate or execute.
Control nodes redirect the tick and condition and action nodes return results: success, failure, or running.
A node is executed if and only if it receives ticks.
If an action is underway, it returns running to the parent.
When it is done it returns success or failure.

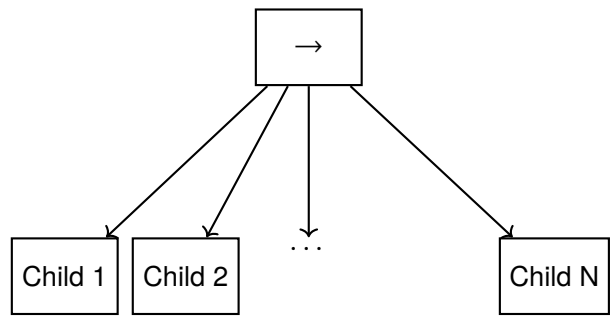
\begin{figure}[H]
    \centering
    \small
            \begin{tikzpicture}[
            node distance=1.5cm and 1.2cm,
            every node/.style={font=\sffamily},
            btnode/.style={
                draw,
                minimum width=1.4cm,
                minimum height=1.0cm,
                align=center,
                fill=white,
                thick
            }
        ]

            \node[btnode] (seq) {$\rightarrow$};

            \node[btnode, below left=2.0cm and 1.8cm of seq] (c1) {Child 1};
            \node[btnode, below left=2.0cm and 0.2cm of seq] (c2) {Child 2};
            \node[below=2.0cm of seq] (dots) {$\cdots$};
            \node[btnode, below right=2.0cm and 1.8cm of seq] (cn) {Child N};

            \draw[->, thick] (seq) -- (c1);
            \draw[->, thick] (seq) -- (c2);
            \draw[->, thick] (seq) -- (dots);
            \draw[->, thick] (seq) -- (cn);

        \end{tikzpicture}
    \caption{A sequence node with N children.}
    \label{fig:behavior_tree_sequence}
\end{figure}

A sequence node is the most fundamental Behavior Tree control node.
It runs each node in order while they are successful, as shown in \autoref{fig:behavior_tree_sequence} and \autoref{alg:sequence_node}.

\begin{algorithm}[H]
\caption{Pseudocode of a Sequence node with $N$ children}
\label{alg:sequence_node}
\For{$i \gets 1$ \KwTo $N$}{
    $\textit{childStatus} \gets \text{Tick}\left(\text{child}(i)\right)$\;
    \uIf{$\textit{childStatus} = \text{Running}$}{
        \Return Running\;
    }
    \uElseIf{$\textit{childStatus} = \text{Failure}$}{
        \Return Failure\;
    }
}
\Return Success\;
\end{algorithm}

\begin{figure}[H]
    \centering
    \small
            \begin{tikzpicture}[
            node distance=1.5cm and 1.2cm,
            every node/.style={font=\sffamily},
            btnode/.style={
                draw,
                minimum width=1.4cm,
                minimum height=1.0cm,
                align=center,
                fill=white,
                thick
            }
        ]

            \node[btnode] (fb) {?};

            \node[btnode, below left=2.0cm and 1.8cm of fb] (c1) {Child 1};
            \node[btnode, below left=2.0cm and 0.2cm of fb] (c2) {Child 2};
            \node[below=2.0cm of fb] (dots) {$\cdots$};
            \node[btnode, below right=2.0cm and 1.8cm of fb] (cn) {Child N};

            \draw[->, thick] (fb) -- (c1);
            \draw[->, thick] (fb) -- (c2);
            \draw[->, thick] (fb) -- (dots);
            \draw[->, thick] (fb) -- (cn);

        \end{tikzpicture}
    \caption{A fallback node with N children.}
    \label{fig:behavior_tree_fallback}
\end{figure}
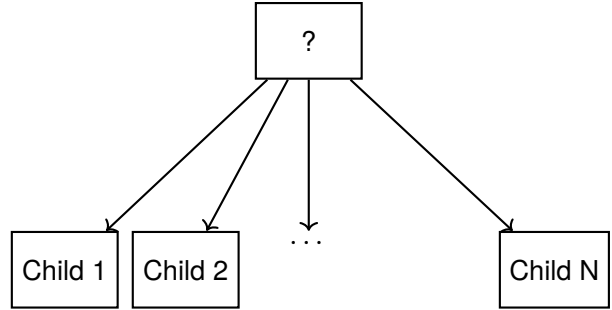

\vspace{1cm}
A fallback node runs each node in order while they are failing.
This is shown in \autoref{fig:behavior_tree_fallback} and \autoref{alg:fallback_node}.

\begin{algorithm}[H]
\caption{Pseudocode of a Fallback node with $N$ children}
\label{alg:fallback_node}
\For{$i \gets 1$ \KwTo $N$}{
    $\textit{childStatus} \gets \text{Tick}\left(\text{child}(i)\right)$\;
    \uIf{$\textit{childStatus} = \text{Running}$}{
        \Return Running\;
    }
    \uElseIf{$\textit{childStatus} = \text{Success}$}{
        \Return Success\;
    }
}
\Return Failure\;
\end{algorithm}

\vspace{.4cm}
The remaining types of nodes are summarized in \autoref{fig:behavior_tree_nodes}.
\vspace{-.5cm}

\begin{figure}[H]
    \centering
    \small
    \begin{tikzpicture}[
    font=\sffamily\small,
    header/.style={font=\bfseries\sffamily\small},
    cell/.style={
        align=center, anchor=center, inner sep=1.5pt
    },
    sym/.style={
        draw, font=\small, anchor=center,
        inner xsep=5pt, inner ysep=2.5pt, 
        text height=1.8ex, text depth=0.4ex
    },
    action/.style={
        draw, font=\small, anchor=center,
        inner xsep=7pt, inner ysep=2pt, 
        text height=1.4ex, text depth=0.3ex
    },
    condition/.style={
        ellipse, draw, font=\small, anchor=center,
        inner xsep=6pt, inner ysep=0pt,
        text height=1.4ex, text depth=0.5ex
    },
    decorator/.style={
        diamond, draw, minimum size=0.5cm,
        inner sep=0pt, text height=0pt, text depth=0pt,
        anchor=center
    }
]

\matrix (m) [
    matrix of nodes,
    row sep=0pt,
    column sep=0pt,
    nodes in empty cells,
    nodes={cell},
    column 1/.style={nodes={text width=1.7cm, minimum height=3.0em}},
    column 2/.style={nodes={anchor=center}},
    column 3/.style={nodes={text width=3.5cm, minimum height=3.0em}},
    column 4/.style={nodes={text width=3.7cm, minimum height=3.0em}},
    column 5/.style={nodes={text width=3.7cm, minimum height=3.0em}},
] {
    |[header]| Node type & |[header, minimum width=1.8cm]| Symbol & |[header]| Succeeds & |[header]| Fails & |[header]| Running \\
    Fallback & |[sym]| ? & If one child succeeds & If all \mbox{children} fail & If one child \mbox{returns} Running \\
    Sequence & |[sym]| $\rightarrow$ & If all \mbox{children} succeed & If one child fails & If one child \mbox{returns} Running \\
    Parallel & |[sym]| $\rightrightarrows$ & If $\geq M$ \mbox{children} succeed & If $> N-M$ \mbox{children} fail & else \\
    Action & |[action]| text & Upon completion & If impossible to complete & During completion \\
    Condition & |[condition]| text & If true & If false & Never \\
    Decorator & |[decorator]| & Custom & Custom & Custom \\
};

\draw[thick] (m-1-1.north west |- m-1-1.north) -- (m-1-5.north east |- m-1-1.north);
\foreach \i in {2,...,7} {
    \draw (m-\i-1.north west |- m-\i-1.north) -- (m-\i-5.north east |- m-\i-1.north);
}
\draw[thick] (m-7-1.south west |- m-7-1.south) -- (m-7-5.south east |- m-7-1.south);

\foreach \j in {1,...,4} {
    \draw (m-1-\j.north east |- m-1-1.north) -- (m-1-\j.north east |- m-7-1.south);
}
\draw (m-1-1.north west |- m-1-1.north) -- (m-1-1.north west |- m-7-1.south);
\draw (m-1-5.north east |- m-1-1.north) -- (m-1-5.north east |- m-7-1.south);

\end{tikzpicture}
    \caption{The different types of Behavior Tree nodes.}
    \label{fig:behavior_tree_nodes}
\end{figure}

\subsection{Coactive Design}

The Coactive Design method~\cite{Johnson_2014} is an iterative process comprised of three main subprocesses: identification, selection and implementation, and evaluation of change.
The most complex is the identification process, in which requirements, alternatives, and interdependence relationships are explored.
A set of desired interdependence relationships are determined and selected for implementation.
The result is then evaluated using human feedback and performance analysis.
This methodology was used for the design and development of Team IHMC's operator interface in the 2015 DRC~\cite{Johnson_2014}.
A later analysis details how that methodology led to success in the competition~\cite{Johnson_2017}.

Coactive Design treats human-machine systems as interdependent and emphasizes the value in making the system observable, predictable, and directable, with interdependence analysis charts used to document those relationships.
The architecture in this thesis is meant to be developed, debugged, and adapted by expert operators on real hardware.
Coactive Design directly motivates the authoring and supervision interfaces in \autoref{ch:architecture2}. An interdependence analysis was done in~\cite{calvert2024behavior}.

\section{Architectural Influences}

\subsection{MIT Director}
MIT's DRC system Director~\cite{Marion_2017} is an earlier example of a structured autonomy framework that integrated operator supervision, planners, a 3D scene, and behavior-level scripting.
It combined locomotion and manipulation planning with an operator-in-the-loop execution pipeline and an embedded Python editor for writing task scripts.
It is a relevant precedent for integrating high-level autonomy and operator tooling in one environment, but its behavior representation remained scripting-heavy and it did not target the combination of robot-local runtime editing, synchronized state sharing, and fast loco-manipulation studied here.

\subsection{FlexBE}
Schillinger et al. introduced FlexBE, a high-level control framework for rescue robotics built on hierarchical state machines, adjustable-autonomy guards on state outcomes, and runtime modification of executing behaviors~\cite{Schillinger_2016}.
FlexBE is a particularly important precedent because it treats runtime behavior adaptation as a normal operational capability rather than an offline development step.
The paper reports qualitative success in example scenarios and competition use, but it does not provide overlapping numerical measures of door traversal speed, door-task reliability, or authoring effort.
FlexBE therefore contributes as an architectural design influence rather than as a direct quantitative baseline.

\subsection{DLR RAFCON}
In the same year, Brunner et al. introduced RAFCON, a DLR tool for engineering robotic tasks as hierarchical state machines with first-class concurrency~\cite{Brunner_2016}.
RAFCON exposes preemptive and barrier concurrency states alongside hierarchy and library states, allows the structure of a state machine and the Python code inside execution states to be modified while it is running, and supports stepping, an execution history with full data context, and backwards stepping for debugging.
The state machine is persisted as a folder of JSON files that mirrors the tree, which makes it readable, version controllable, and amenable to multi-developer collaboration.
The most complete demonstration is the SpaceBotCamp 2015 mission, in which a state machine of more than 700 states across 8 hierarchy levels orchestrated navigation, exploration, perception, and manipulation; it is therefore a strong precedent for runtime-modifiable graphical authoring at scale, but, like FlexBE, it does not provide overlapping numerical evidence on door-task speed, reliability, or authoring effort, and it remains a host-side orchestration tool over ROS components rather than a robot-local synchronized runtime.

\subsection{Drawing Board}
Senft et al.~\cite{Senft_2021} present \emph{Drawing Board}, a task-level authoring interface for remote teleoperation of a tabletop Franka Emika Panda arm.
The system is built around four principles relevant here: interleaving observation and planning, action-level robot control, a unified augmented-reality interface, and graphical specification of actions.
An 18-participant study showed that novices produced longer and more frequent autonomous periods with task-level authoring than with direct or point-and-click control.
The architecture in this thesis shares the action-level authoring stance and unified-interface principle, and extends them to humanoid loco-manipulation, robot-local execution under degraded communications, and reactive tree structure with behavior-time perception authored as scene actions.

\subsection{Behaviors on CENTAURO}
More recent CENTAURO work shows how structured task coordination can extend into perception-aware execution.
De Luca et al. used BehaviorTree.CPP~\cite{behaviortree_cpp} and Groot~\cite{groot1, groot2} to manage online replanning and recovery for rough-terrain navigation on the CENTAURO wheeled-legged robot~\cite{2023_centaur_bt}.
They have a robot centaur with four legs with wheels, two arms with hands, and a head.
That system is a useful reference point for online planning using Behavior Trees, but it remains navigation only and does not demonstrate fast execution, with real-robot runs reported between 225~s and 335~s.
A later paper by Wang et al.~\cite{2024_wang_grounded_lm} moved into simple loco-manipulation by combining a predefined behavior library with task graphs executed as behavior trees.
That paper is especially relevant because perceptual operations such as object detection, grip-force sensing, and visual question answering are inserted into the task structure as authored behaviors, which is a precedent for the scene action idea developed here.
The reported real-world execution remains slow, with pick-and-place times of 160.6~s in the nominal case and 203.2~s with failure recovery.
Together, these structured systems show important pieces of the design space, but they do not establish fast humanoid loco-manipulation with editable authored structure and task-local perception.

\section{Door Traversal Systems}

Door traversal is the benchmark task for this dissertation, so the most relevant prior systems execute a substantial portion of the full sequence from approach through passage.
The literature spans behavior-based reactive systems, motion-planning formulations, and learned policies.
The comparison is bounded to systems that overlap enough with the benchmark task to clarify the architectural design space.

\subsection{Classical and Model-Based Systems}

\subsubsection{2008-2010: Jain and Kemp}
Before recent learned door systems, much of the door-opening literature focused on mobile manipulators.
Jain and Kemp address two complementary halves of the task across a pair of papers.
The 2008 system, El-E~\cite{Jain2008dooropening}, is a statically stable mobile manipulator with a 5-DOF Katana arm and custom force-sensing fingers; it decomposes the push-side task into a serial chain of behaviors, with branches to explicit failure states, for locating the handle, deciding whether the door is locked, twisting the handle, deciding whether the door can be pushed, and pushing through the doorway.
Across 30 trials on 6 doors, 5 per door (1 locked, 4 unlocked), the robot completed the full unlocked task in 21/24 trials (87.5\%) and correctly detected the locked condition in 6/6 trials, stopping safely on every failure rather than requiring intervention.
The 2010 system~\cite{Jain2010EPC} switches platform to a hooked compliant manipulator on an omnidirectional base and addresses the pull-side task using equilibrium point control to coordinate the base and the compliant arm without a prior kinematic model of the mechanism.
Across 40 trials on 10 mechanisms (7 cabinet doors and 3 drawers, four trials each), a trial succeeded if the robot reached the handle and opened the door more than 60\degree{} or pulled the drawer more than 30~cm. The robot met those criteria on 37/40 trials overall, including 26/28 door trials and 11/12 drawer trials.
These are important early benchmark-oriented references: they take on the whole task on real hardware, report substantial repeated-trial evidence on diverse mechanisms, and treat handle perception, force-based contact, and recovery as first-class parts of the behavior.
The platforms remain statically stable wheeled mobile manipulators and the task still begins from an externally provided handle cue, by laser pointer in 2008 and by 3D handle pose plus hook orientation in 2010, rather than from authored task-time perception.
The behavior set in each paper is also fixed at deploy time rather than runtime-editable, so changing recovery logic, swapping perception, or adding a new variant requires returning to source.

\subsubsection{2010: Chitta}
Chitta et al.~\cite{2010ChittaDoorOpening} address coordinated base-and-arm motion planning for autonomous door opening on the Willow Garage PR2.
Their key idea is to partition the planning problem rather than search a full whole-body state space: a graph search in $(x, y, \theta, d)$ plans the omnidirectional base trajectory, where $d$ is a binary ``door interval'' variable indicating whether the door is connected to the closed or open configuration, while the arm trajectory is recovered by inverse kinematics at each base waypoint so that the gripper stays on the handle and the door remains within the arm's reachable workspace.
The system was evaluated for 5/5 push and 5/5 pull trials on the PR2, demonstrating that opening and passage can be posed as a coupled base-arm planning problem rather than a hard-coded motion sequence.
The contribution is planning-focused and assumes that the door has already been grasped and unlatched, that an initial door model is available from a building map, and that the world is static during execution; the paper explicitly does not handle disturbances such as a person holding the door closed.
By contrast, this dissertation focuses on editable reactive task structure operating from task-time perception, and treats grasp acquisition, retries, and disturbance recovery as authored parts of the same behavior rather than as preconditions of the planner.

\subsubsection{2015: Axelrod and Huang}
Axelrod and Huang~\cite{Axelrod_2015} report autonomous door opening and traversal on an iRobot 510 PackBot, a tracked skid-steer base with a 5-DoF arm and no wrist yaw joint.
A custom Honeybee Robotics gripper provides a passive 2-DoF compliant wrist and fingertip Takktile tactile sensors~\cite{Tenzer_2014_takktile} in place of a wrist force/torque sensor, and a 2D laser rangefinder on the arm base tracks door pose throughout the task.
The system is shared autonomy: the operator drives to a starting pose, specifies door type and handle type, confirms the visual handle detection, and then lets the robot execute the rest of the sequence.
It is the most variation-broad classical reference here, covering push and pull, knobs and levers, and crucially the pull lever with a self-closing mechanism, where the robot uses its flipper to cage the door open before re-grasping from the inside and driving through.
Reported task times are 63~s for push knob, 128~s for pull knob, 83~s for push lever with closer, and 118~s for pull lever with closer, measured from first robot motion until the back of the robot clears the doorway, with the robot starting 1~m from the door.
Reliability is reported only qualitatively at approximately 60\% because, as the authors note, any phase failure aborts the run and no recovery behavior is authored.
The behavior runtime executes on a laptop tethered to the robot, the operator-classified door and handle inputs gate which fixed sub-behavior runs, and the per-handle perception is hard coded with Hough circles and Sobel-filtered Hough lines.
Axelrod and Huang therefore extend the variation coverage of behavior-decomposed door traversal in classical mobile manipulation, while reinforcing the same architectural distance from this thesis: fixed shared-autonomy decomposition, off-robot execution, and no editable recovery structure.

\subsubsection{2015: Banerjee et al.}
Banerjee et al.~\cite{Banerjee_2015}, WPI-CMU's DARPA Robotics Challenge entry, is a particularly relevant classical humanoid door traversal reference.
It reports a human-supervised semi-autonomous system in which Atlas detects a door, approaches it, opens it, and walks through it in the DRC setting.
The system uses an event-driven finite-state machine executed on the robot, with human validation at critical transitions, motion planning for the manipulation phases, and both autonomous and operator-aided door detection.
This is an important example of full-task humanoid door traversal under shared autonomy.
Reported execution is slow by modern standards, at 9~min~23~s for the pull door case and 7~min~40~s for the push door case, and the DRC Finals deployment used the operator-aided detection mode for reliability.
Relative to the system developed here, the key adaptability difference is that Banerjee et al. execute a pre-defined supervised FSM rather than a runtime-editable one, so it does not address fast behavior iteration, tighter perception-behavior integration, or repeated reactive execution.

\subsubsection{2023: Jang et al.}
Jang et al.~\cite{2023JangDoorTraversal} extend the partitioned base-plus-arm planning idea from Chitta et al. to the full navigation problem on a Husky-based mobile manipulator with a Franka Emika Panda arm.
A graph search in $(x, y, \theta, a)$ plans the base pose and an \emph{area indicator} $a \in \{0,\dots,4\}$ that records where the robot is relative to the door, from approaching, to opening, to crossing the doorsill, to closing, to navigating to a goal beyond the door, and an inverse kinematics solver then recovers the arm path along the planned base trajectory.
The integer area indicator generalizes the binary door interval from Chitta et al. and lets approach, opening, traversal, closing, and goal navigation be solved in a single search.
In simulation across 25 push and 25 pull trials, the framework reaches 100\% planning success against 76--80\% for the equivalent separate-planning baseline, and is 8.7$\times$ faster on pull doors and 2.3$\times$ faster on push doors while producing shorter, lower-cost paths.
The framework is offline and assumes the door type, joint range, and handle position as inputs, so adapting to a new door, recovering from a failed grasp, or reacting to scene change at task time falls outside the planner; the authors note closed-loop replanning as future work.

\subsubsection{2023: Sleiman et al.}
Sleiman et al.~\cite{Sleiman_2023} take a planning-centered approach to loco-manipulation on a quadrupedal mobile manipulator, an ANYmal with a 6-DoF arm.
Their framework casts multi-contact loco-manipulation as a Task and Motion Planning problem solved by sampling-based bilevel optimization combined with informed graph search, and it is demonstrated on the real robot for opening and closing a heavy dishwasher and for traversing a spring-loaded pull door using both prehensile and non-prehensile contacts.
This is the strongest planning-based reference for spring-loaded door traversal in the recent literature, and it complements the learned door results discussed below by showing that multi-contact loco-manipulation can also be produced by holistic offline planning.
The system is a quadruped rather than a humanoid, the planner runs offline from a known object model, and the executed behavior is not represented as an editable task graph that an operator can modify at runtime.

\subsubsection{2024: Thamrongaphichartkul and Vongbunyong}
Thamrongaphichartkul and Vongbunyong~\cite{Thamrongaphichartkul_2024} pair a mobile manipulator door-traversal task with a behavior tree implemented over ROS~2 and evaluated in Gazebo.
The platform is a differential-drive base with a 6-effective-DoF arm augmented by two added base-side degrees of freedom, and the authored tree covers both an initially closed door and an initially open door, including approach, handle grasp, opening, traversal, and closing as reusable subtrees.
This is the closest published precedent that explicitly combines a behavior-tree task model with the full door-traversal task on a mobile manipulator and reports honestly on the limitations of classical behavior trees: the authors find that small base-positioning errors propagate into incorrect tree decisions because the leaf actions do not adapt at task time.
That observation lines up directly with the reactive structure and behavior-time perception emphasized in this dissertation, and it explains why a behavior tree alone is not sufficient for fast humanoid loco-manipulation.

\subsubsection{2024: Kang et al.}
Kang et al.~\cite{kang2024door} is a more recent system-level reference because it presents a complete door opening and passage pipeline on a wheeled mobile manipulator rather than an isolated perception or control component.
The system combines handle segmentation and pose estimation, exploratory force-based identification of the door opening direction, an adaptive position-force controller, and an SAC-based reinforcement learning controller for the opening and passing phase.
It targets a complete benchmark task with varied handle types, door widths, and both push and pull doors, and explicitly compares a classical adaptive controller against a learned alternative inside one integrated system.
The platform is a wheeled mobile manipulator rather than a legged humanoid, and the RL portion was trained only for a Push-CCW door while the other phases remained fixed procedural modules.
In the real-world results, the RL controller was evaluated only on a single 0.9~m Push-CCW case, where it completed ST4 19\% faster than the adaptive controller, while the broader real-world door-variation coverage came from the adaptive position-force controller.
Kang et al. concentrate adaptability inside a controller layer within an otherwise fixed procedural pipeline, whereas this dissertation focuses on editing the task structure itself, including perception, coordination, and recovery logic, directly on the robot during development.
Kang et al. are therefore prior art for full-sequence door opening and passage with a mobile manipulator and a benchmark-oriented comparison point, but not for fast editable loco-manipulation behavior structure on a humanoid robot.

\subsubsection{2025: Schulze et al.}
Schulze et al.~\cite{Schulze_2025} report a deployed transport-and-messaging service on a SCITOS~G5 differential-drive base with a Kinova Gen~II 7-DoF arm, integrated to traverse closed doors and ride elevators in a populated multi-story environment.
Across long-term field tests in an elderly-care facility and a university office building, the full system reported overall task success of 88.6\% across 79 runs in one site and 80.0\% across 40 runs in the other, with door manipulation alone exceeding 88\% in both sites.
This is one of the few real-world deployment references that integrates door manipulation with a longer mission and reports honest failure analysis at the system level, which makes it a useful precedent for the multi-step exploration and three-door composite behaviors discussed in \autoref{ch:defense}.
The platform is a wheeled differential-drive service robot rather than a bipedal humanoid, and the authored skill set is not a runtime-editable behavior tree, but the operating point and failure-mode discussion are directly relevant to deploying door behaviors in cluttered, populated environments.

\subsection{Learning-Based Systems}

\subsubsection{2024: Zhang et al.}
Zhang et al.~\cite{zhang2024learningopentraversedoors} present a teacher-student reinforcement learning policy for an ANYmal-based legged manipulator to open and traverse doors.
The paper targets the combined task of opening and passing through the doorway on a legged platform rather than stopping at door opening alone.
It claims a single learned policy that handles both push and pull doors without being given the opening direction a priori, and reports repeated-trial results on a single spring-loaded door of 20/20 traversals on the pull side and 18/20 on the push side, for an overall success rate of 95.0\%.
The two failures were not failures to open the door itself, but push-side traversal failures in which the robot got stuck on protruding doorway geometry that was not represented in the simulation model.
This is a strong recent reference point for learned door traversal on a legged robot.
The method is a monolithic learned controller rather than a runtime-editable behavior architecture, and the hardware experiments rely on externally provided handle and doorway measurements from motion capture or AprilTags rather than the onboard authored perception pipeline studied here.
That failure mode also illustrates a low-adaptability workflow: because the issue arises from a simulation mismatch inside a learned monolithic policy, addressing it requires changing the simulation or training setup and retraining rather than making a runtime behavior edit, a turnaround on the order of at least a day in our own development terms.
Zhang et al. serve as a learned door traversal comparator on a legged platform, but they do not address the behavior authoring speed, editable task structure, or robot-local behavior coordination that are central here.

\subsubsection{2025: Xue et al.}
The closest learned humanoid comparison point is Xue et al.~\cite{xue2025opening}, which presents DoorMan, a teacher-student-bootstrap sim-to-real learning pipeline for humanoid door loco-manipulation from pure RGB perception.
It closes several gaps left by earlier learned door systems: it uses a humanoid platform rather than a quadruped with arm or a wheeled manipulator, it does not rely on externally provided door measurements as in Zhang et al., and it evaluates real-world door interaction across three categories: push lever, pull lever, and push bar.
DoorMan reports an overall task success rate of 83\% and an average completion time of 15.40~s across those categories, while outperforming expert teleoperators on the same whole-body controller in task completion time.
That places it in roughly the same coarse speed regime targeted in this dissertation and makes it the most important learned humanoid baseline.
The reported runtime deployment is not robot-local: DoorMan policy inference runs on a desktop workstation with an Intel i9-14900K CPU and an NVIDIA RTX 4090 GPU rather than on the humanoid itself~\cite{xue2025opening}, so the robot depends on active external communications during execution and cannot continue functioning autonomously if that link is lost.
The method is a monolithic learned policy trained through privileged-state teacher learning, DAgger-based RGB distillation, GRPO fine-tuning, and large-scale simulation randomization, rather than a runtime-editable behavior architecture.

The rough pipeline for training a behavior is:
\begin{enumerate}
    \item Build an ultra realistic and randomized physics simulation of the task.
    \item Tune a PPO policy to get desired whole body motion with ground truth information.
    \item Run DAgger to convert to a vision based model.
    \item Use GRPO to wean dependence from ground truth to vision.
    \item Test on the real robot.
\end{enumerate}

DoorMan uses multi-stage decomposition to shape rewards and resets during training, but the demonstrated runtime result is a single end-to-end door-interaction policy, with no reported task-level branching across multiple authored behaviors or composition of heterogeneous behaviors into longer autonomous sequences.
Adaptation in DoorMan is therefore retraining-centered rather than edit-centered: changing the policy behavior requires revisiting simulation assets, reward shaping, and the training pipeline rather than modifying the executing task logic on the robot.
Xue et al. serve as a major comparison point for high-performance RGB door execution on a humanoid, but not a baseline for robot-local autonomy, authoring effort, or targeted runtime adaptation.

There is also broader recent work on learning for articulated-object manipulation, such as Xiong et al.~\cite{xiong2024adaptive}.
That work is relevant at the level of general manipulation direction, but the detailed comparison in \autoref{ch:defense} stays bounded to door systems with overlapping task scope and reported speed or success metrics.

\subsubsection{2026: Zhang et al.}
Zhang et al.~\cite{Zhang_2026_Sumo} present Sumo, a hybrid loco-manipulation framework that uses sample-based model-predictive control at runtime to steer a pre-trained whole-body reinforcement-learning policy.
Sumo is most extensively evaluated on a Boston Dynamics Spot quadruped on dynamic whole-body tasks such as uprighting a tire and dragging a crowd-control barrier, and it also demonstrates the same hybrid framework in simulation on a Unitree G1 humanoid for door opening, table pushing, and box pushing.
Architecturally it inverts the more common high-level RL plus low-level MPC stack by keeping the high-level decision in MPC and the low-level locomotion in RL, which lets the system adapt to new objects or new task objectives at deployment by changing only the planner's cost function and object model rather than retraining the policy.
This is closer in spirit to the edit-centered adaptation argued for in this thesis than a monolithic policy is, but the published humanoid door evidence is in simulation only, so Sumo is a recency-aware design-space comparison rather than a real-world humanoid door baseline alongside Xue et al.

\section{Design Space Summary}

\autoref{fig:design_space_map} makes the gap explicit.
The prior work reviewed here contributes ingredients used in this dissertation, but not their combination.
Affordance Templates and Affordance Primitives provide precedent for reusable authored action structure.
Behavior Trees provide precedent for reactive task coordination.
Coactive Design, Director, FlexBE, RAFCON, and Drawing Board provide precedent for making autonomy inspectable and adaptable under human supervision.
The CENTAURO papers show that perception can be elevated into task structure rather than left as a global background service.
Classical and learned door systems demonstrate several ways to achieve substantial portions of the benchmark task, from shared-autonomy humanoid FSMs to mobile-manipulator planners to monolithic learned door policies.

We are not aware of a prior system that combines reusable authored task structure, reactive execution, robot-local runtime behavior state synchronized to an operator interface, behavior-time perception configuration, and fast real-robot humanoid door behaviors in one architecture with an edit-centered adaptation workflow.
Some prior systems are structured and editable but do not demonstrate fast humanoid loco-manipulation.
Others are strong door-performance references but depend on fixed procedural pipelines, external measurements, heavy supervision, off-board inference with active communications, or retraining-centered adaptation.
This is the gap the thesis addresses, evaluated in later chapters on speed, robustness across task variation, and adaptation time.

\chapter{Building Our Behavior Architecture}
\label{ch:building}

In this chapter we'll tell the story of how our behavior system was inspired, designed, and built over roughly a 10-year time period.
Ultimately, we'll reveal the current design as a snapshot in time in the year 2026, as it is still in active development.
The story arc of the behavior system spans from roughly 2014 in the run-up to the DARPA Robotics Challenge (DRC) finals~\cite{DRC}, in which our DRC Atlas robot could perform 8 tasks in 1 hour under scrupulous direct teleoperation, to 2026 when our IHMC Alex humanoid robot can perform door traversals and multi-station loco-manipulation sorting tasks automatically in the tens of seconds regime and with high levels of operator neglect.

\section{2014-2015 DARPA Robotics Challenge Era}
\label{sec:drc_era}

\subsection{Teleoperation and the DRC UI}
The autonomous behavior system wasn't developed heavily until around 2016, after Team IHMC won 2nd place in the DARPA Robotics Challenge Finals~\cite{Johnson_2017}.
In the DARPA Robotics Challenge, teams were given 1 hour to teleoperate humanoid robots in accomplishing 8 tasks in a mock-urban setting.
The tasks were designed such that a humanoid robot might be the most suitable form factor for completing them.
They included driving a car, getting out of the car, walking on uneven terrain, using hand tools, going through a door, up a flight of stairs, and operating various machinery.
Comparatively, a human would be able to accomplish this set of tasks in under 5 minutes.
Most teams were unsuccessful in completing all 8 tasks within the hour and those that did took the greater portion of the hour to do it.
Participation in this challenge taught us a lot about the problems involved in getting loco-manipulation behaviors to work.

Though we ended up using teleoperation, we had tried adding degrees of autonomy to increase speed and reliability.
The challenge featured low bandwidth communication with the robot at all times and high bandwidth communication only periodically.
The communication blackouts were designed to encourage robot autonomy.
A robot that did not have to wait for human input would be able to advance without waiting, reducing the team's run duration which was used as a tie-breaker after points scored.
Team MIT, for example, had autonomous functionalities~\cite{Marion_2017}.
Reliance on human-guided perceptual registration was shown to slow down run times because it required waiting through multiple communications blackouts.

\begin{figure}[t]
    \centering
    \includegraphics[width=.95\columnwidth]{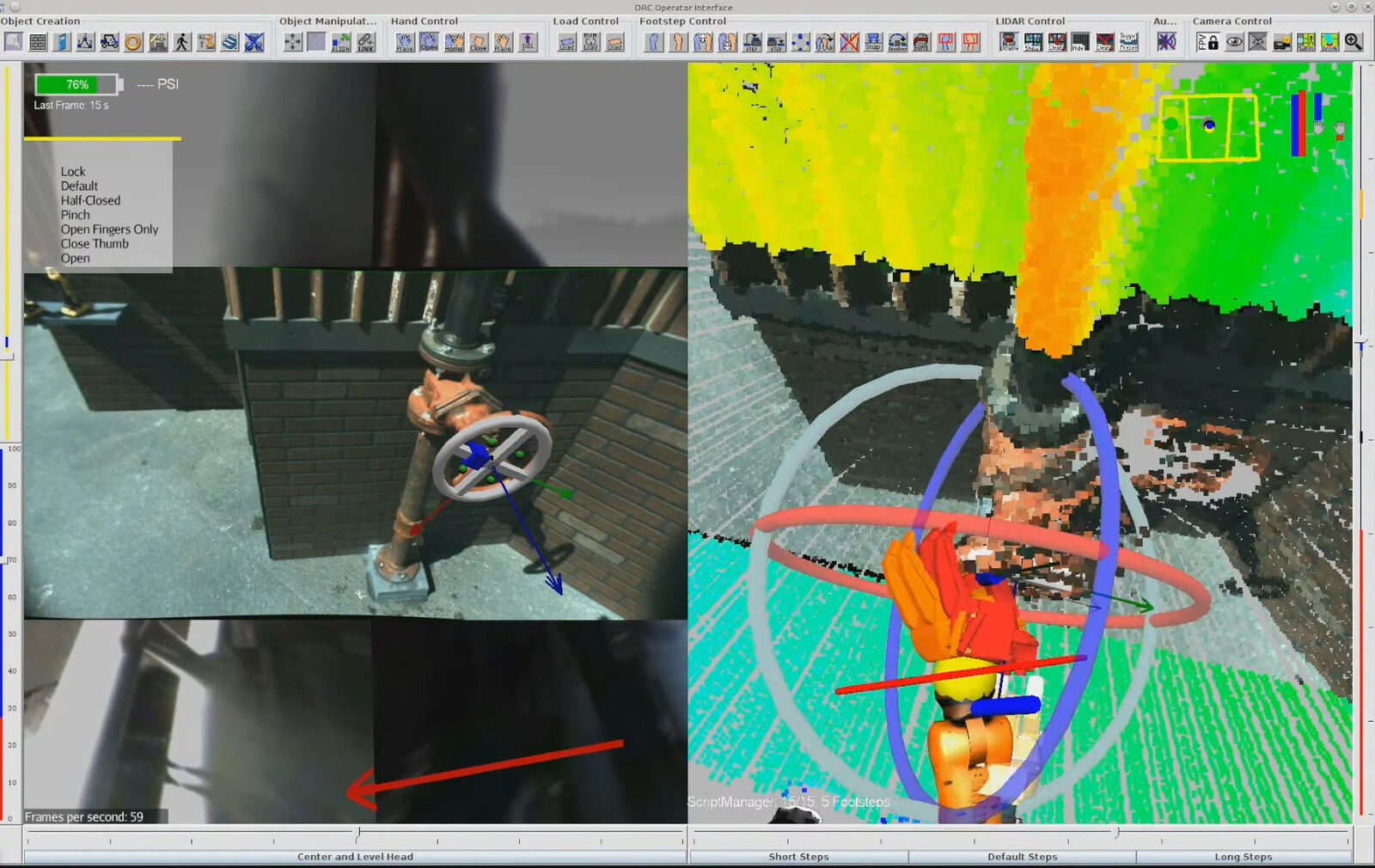}
    \caption{The IHMC DRC operator interface during the Valve Task in the DARPA Robotics Challenge Finals.
    On the left side, a virtual valve has been placed in the scene on top of the point cloud to register the task.
    On the right side the current and next hand poses can be seen as two separate graphics.
    Also, the control gizmo for the hand can be seen as the large colored tori that revolve around the hand.
    A video is available at \url{https://youtu.be/TstdKAvPfEs}.
    }
    \label{fig:drc_ui_valve_task}
\end{figure}

\begin{figure}[t]
    \centering
    \includegraphics[width=.95\columnwidth]{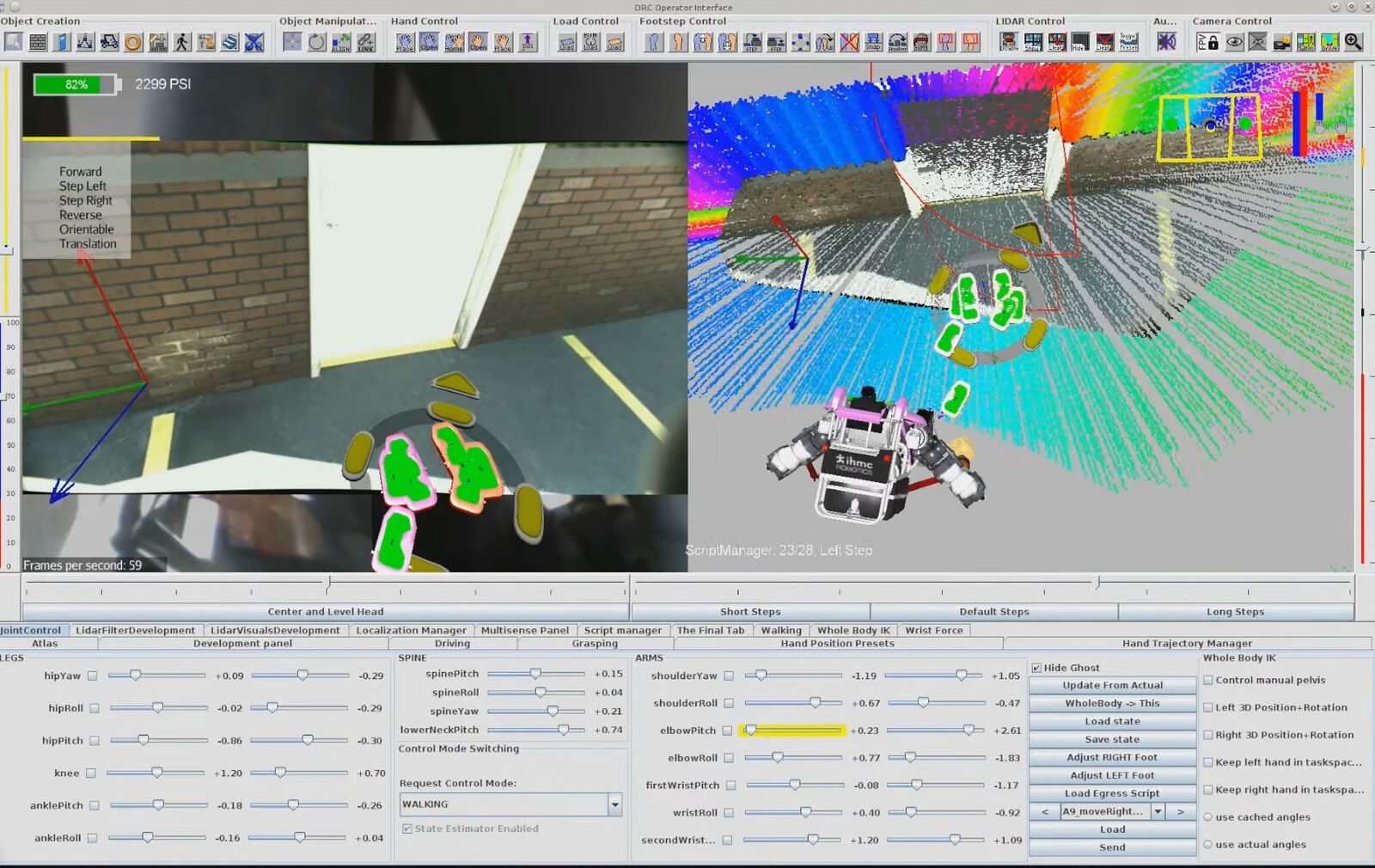}
    \caption{The IHMC DRC operator interface during the Door Task in the DARPA Robotics Challenge Finals.
    In the first-person view on the left, virtual planned footsteps can be seen.
    On the right, the virtual door modifiable has been placed at the pose of the real door.
    The small red arc shows the door panel edge arc if it were to swing open.
    A video is available at \url{https://youtu.be/TstdKAvPfEs}.}
    \label{fig:drc_ui_door_task}
\end{figure}

Even through teleoperation, the challenge taught us a lot about the effectiveness of a particular style of user interface for humanoid robot-operator teaming.
In the challenge, Team IHMC used a hybrid operator interface comprised of two 3D views: a third-person view with a posable viewport camera and a first-person view that looks at the world from the perspective of the robot's head camera, such that virtual graphics could be situated within the scene to provide an augmented reality experience, as shown in \autoref{fig:drc_ui_valve_task} and \autoref{fig:drc_ui_door_task}.

The DRC UI introduced a lot of highly useful user interface mechanisms for managing a humanoid robot in accomplishing tasks.
It was based on the Java Swing~\cite{Java_Swing} widget library and jMonkeyEngine~\cite{jMonkeyEngine} for the 3D graphics engine.
The DRC UI had a collection of competition-hardened interaction features that worked very well for operating humanoid robots.
These included a unique 3D scene camera control and orbit system, quick access buttons with icons, 3D interactable pose gizmos, virtual modifiable plan footsteps, a walk path control ring, sliders for head and spine joint-space commands, virtual object placement, and object action templates.

\begin{figure}[t]
    \centering
    \includegraphics[width=.95\columnwidth]{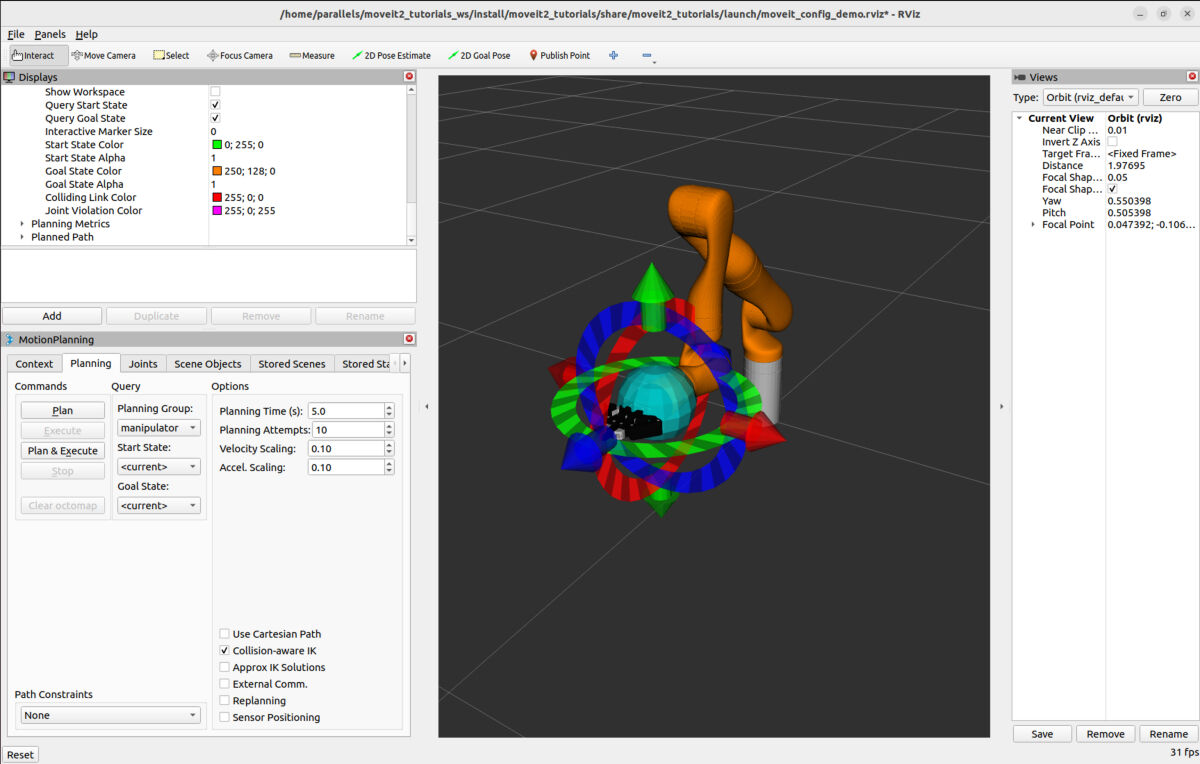}
    \caption{A still from the MoveIt tutorial~\cite{MoveIt_quickstart_rviz_tutorial} showing the red, green, and blue pose gizmo used to move the robot's end effector.
    The user can drag the circular planes to reorient the pose and drag the arrows to translate it.}
    \label{fig:rviz_gizmo}
\end{figure}

Gizmo has become a colloquial term in robotics to describe a virtual interactable 3D set of widgets that assist a user in defining or modifying a 3D position, orientation, or pose.
There are a few implementations out there such as in RViz~\cite{Rviz} and ImGuizmo~\cite{Imguizmo}, with a representative MoveIt pose-gizmo example shown in \autoref{fig:rviz_gizmo}.
The DRC UI featured a custom one built by John Carff in jMonkeyEngine (JME) using colored tori for controlling orientation.
The DRC UI gizmo was mainly controlled by keyboard shortcuts, with the F keys being used to select the axis and the arrow keys being used to move the gizmo along or about that axis.
It featured orientation and position control and was used to specify poses of the virtual interactable objects such as hands, footsteps, torso motions, and scene objects.

\subsection{Virtually Placeable Objects}
Virtually placeable objects included a door, a hand drill, and variable-sized valves.
These virtually placed objects had hand-coded, steppable action scripts in code that were executable with respect to the object.
This was very similar, unknowingly, to the Affordance Template framework being designed at TracLabs at around the same time by Steve Hart~\cite{Hart_2014}.
Hart's use of the word affordance was to establish a moniker for the concept of reusable behaviors for common things you might encounter in the environment.
The word affordance comes from Gibson's book, \emph{The Ecological Approach to Visual Perception}~\cite{Gibson_1979}.
It is used in the sense that an environmental feature affords an action.
For example, the ground is an affordance for walking and a handle is an affordance for turning.

\subsection{Scripted Behaviors}
The DRC-era behaviors were scripts executed from the user interface as seen in \autoref{fig:drc_ui_door_task}.
The operator would step through the actions one by one.
They also did not close the loop on perception.
Tasks were aligned to the scene manually by the operator using the point cloud and images.

\begin{figure}[t]
    \centering
    \includegraphics[width=.95\columnwidth]{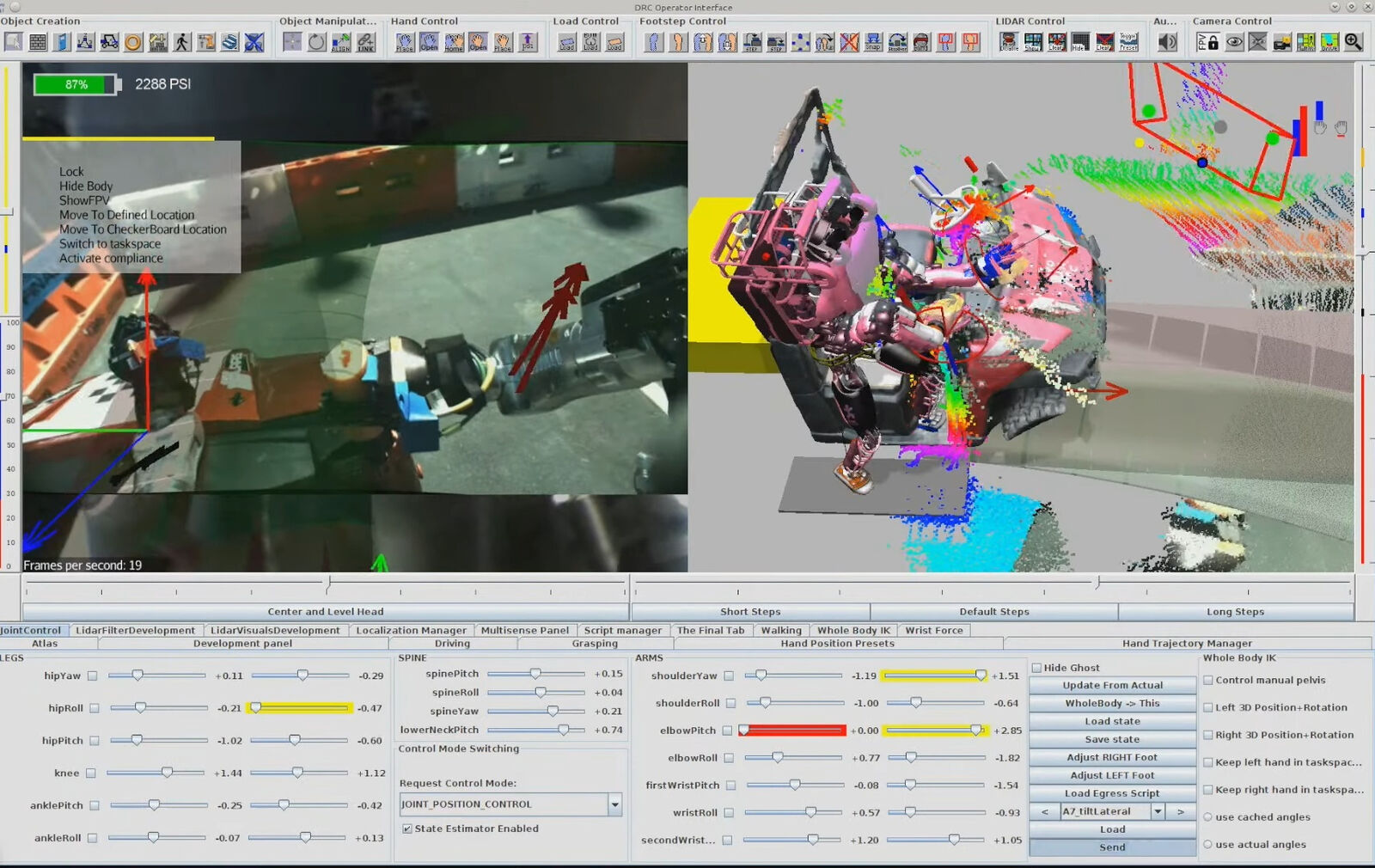}
    \caption{The IHMC DRC operator interface during the Car Egress Task in the DARPA Robotics Challenge Finals.
    This is an example of a scripted behavior.
    The current state is ``A7\_tiltLateral'' as seen in the bottom right.
    Scripts were loadable and savable locally to the operator computer.
    Script actions could be previewed before executing them.
    A video is available at \url{https://youtu.be/TstdKAvPfEs}.
    }
    \label{fig:drc_ui_egress_scripting}
\end{figure}

\section{2016-2019 Atlas, Hard-Coded, JavaFX Era}

\subsection{State Machines and Pipelines}
The next generation of behaviors at IHMC started around 2016, where we implemented code-defined behaviors as state machines and pipelines.
These behaviors used the whole-body controller API directly, which was developed for the NASA Space Robotics Challenge~\cite{Hambuchen_2017_SRC} and is largely the same today.
It allows asynchronous commands for the various body parts to be submitted in real time.
The controller maintains a list of active commands that it is actively controlling the robot to satisfy.
For example, you can send a list of footsteps for the robot to take, a 3D hand pose for the robot to reach to, or a list of joint angles for the spine, a leg, an arm, or the neck.
Pipelines allowed a level of hard-coded action concurrency by branching out into parallel states before rejoining.

\begin{figure}[t]
    \centering
    \includegraphics[width=.95\columnwidth]{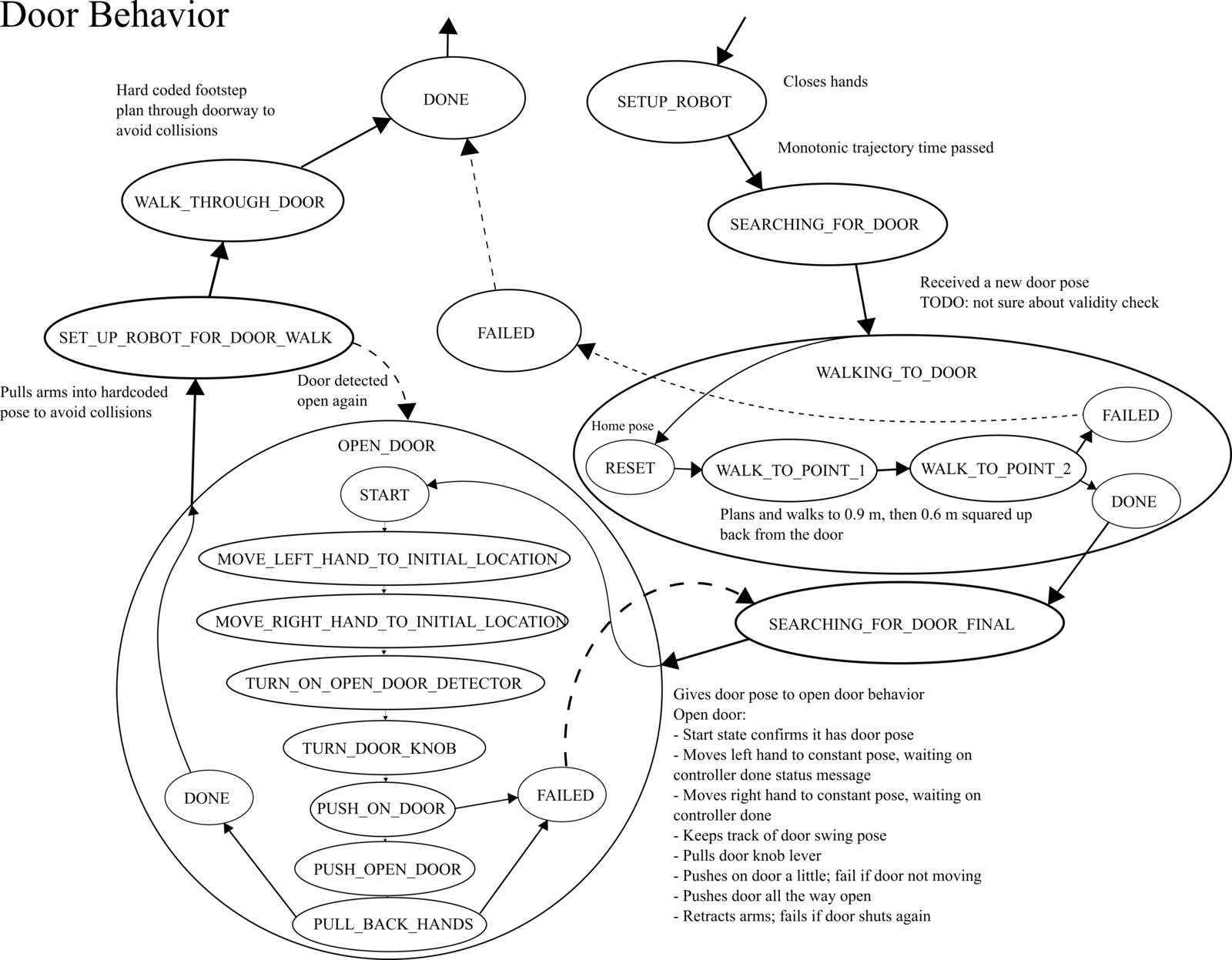}
    \caption{The door traversal behavior hierarchical state machine from the 2019 era.
    This behavior was hard-coded in Java.
    A video of a 2018 door behavior is available at \url{https://youtu.be/-XBlT1H8ZZ4}.
    }
    \label{fig:door_diagram}
\end{figure}

This generation of behaviors started to use perception in the form of basic OpenCV color filters, blob detectors, fiducial detectors, and simple lidar processing.
This enabled us to perform behaviors such as pull-door traversals (\autoref{fig:door_diagram}) and autonomous ball pickup behaviors.
These behaviors were slow in the several minutes range for door traversals and maybe 5 to 10 minutes for picking up colored balls off the floor.
They were performed with the Boston Dynamics DRC Atlas.
Some of the slowness was due to reaching the relatively slow maximum speeds of this hardware platform.

\subsection{System Limits}
Perception was a big limitation in these behaviors.
We relied on a MultiSense SL~\cite{multisense_sl} sensor package which featured a spinning Hokuyo lidar scanner~\cite{hokuyo_utm30lx} which took some time to gather sufficient data.
We also didn't have good software for buffering and querying the point cloud quickly.
This made it difficult to achieve fast behavior based on perception.

Another key limitation was the architecture.
The behaviors were coded by hand, with magic numbers in the code defining the poses of the hands and footsteps relative to objects.
This meant that iterating on behaviors required code changes for every tweak.
In practice, this was done by running the behavior off-board on the operator computer and using a code-reloading tool called JRebel~\cite{kabanov2014thousand, jrebel}.
The iteration loop required magic number tweaking, code commenting, reordering, and restructuring.
This authoring process was an unguided expert code tweaking exercise.

\subsection{Autonomous Locomotion with JavaFX UIs}
\begin{figure}[H]
    \centering
    \includegraphics[width=.95\columnwidth]{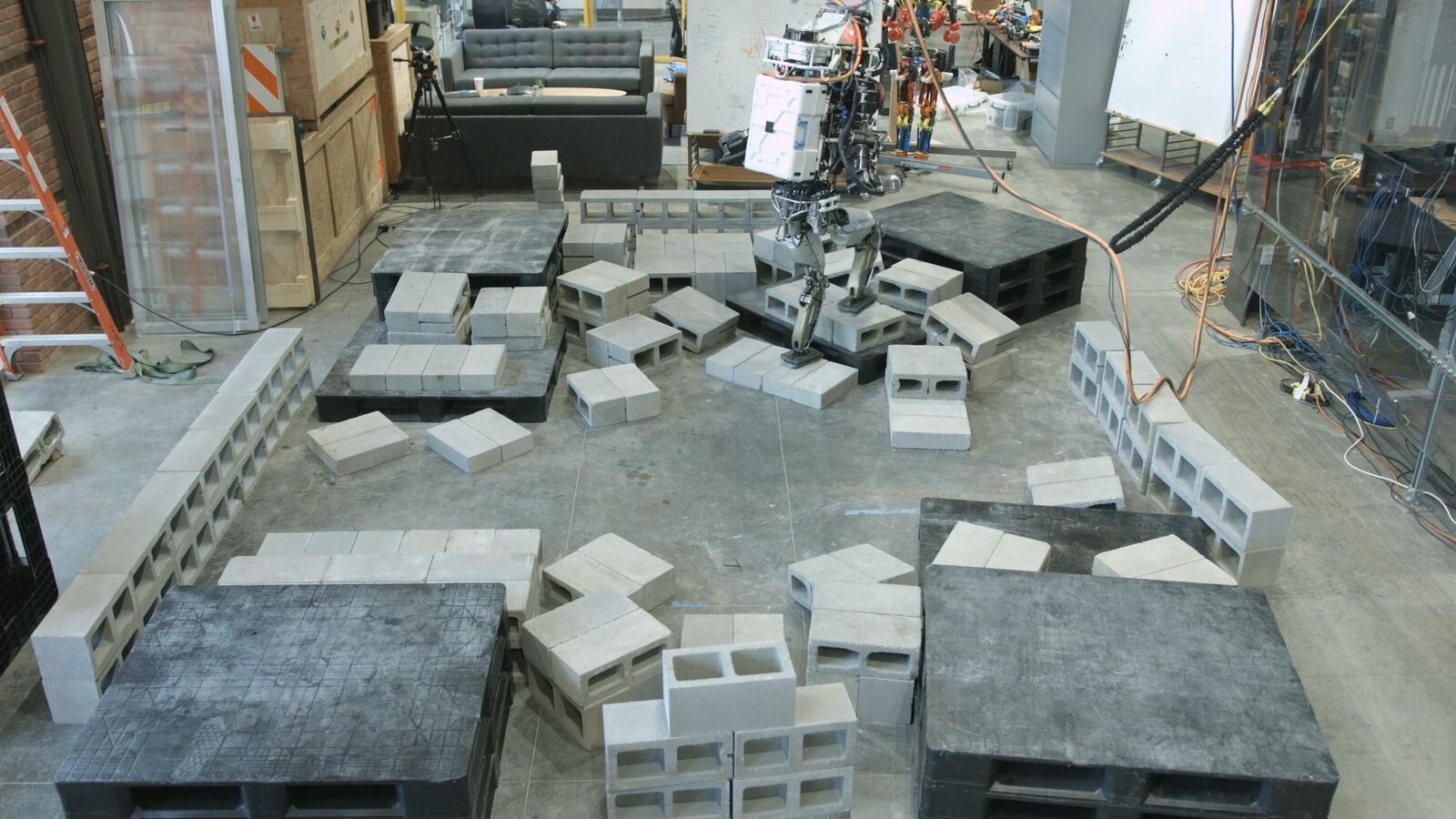}
    \caption{A rough terrain autonomous exploration behavior that sought out high and low flat regions, turning and scanning the terrain in between.
    A planar region map was built over time and used to traverse the rough terrain.
    A video is available at \url{https://youtu.be/cwnrZzzueFA}.
        (May 2019)}
    \label{fig:rough_terrain_autonomy_atlas}
\end{figure}

\begin{figure}[H]
    \centering
    \includegraphics[width=.95\columnwidth]{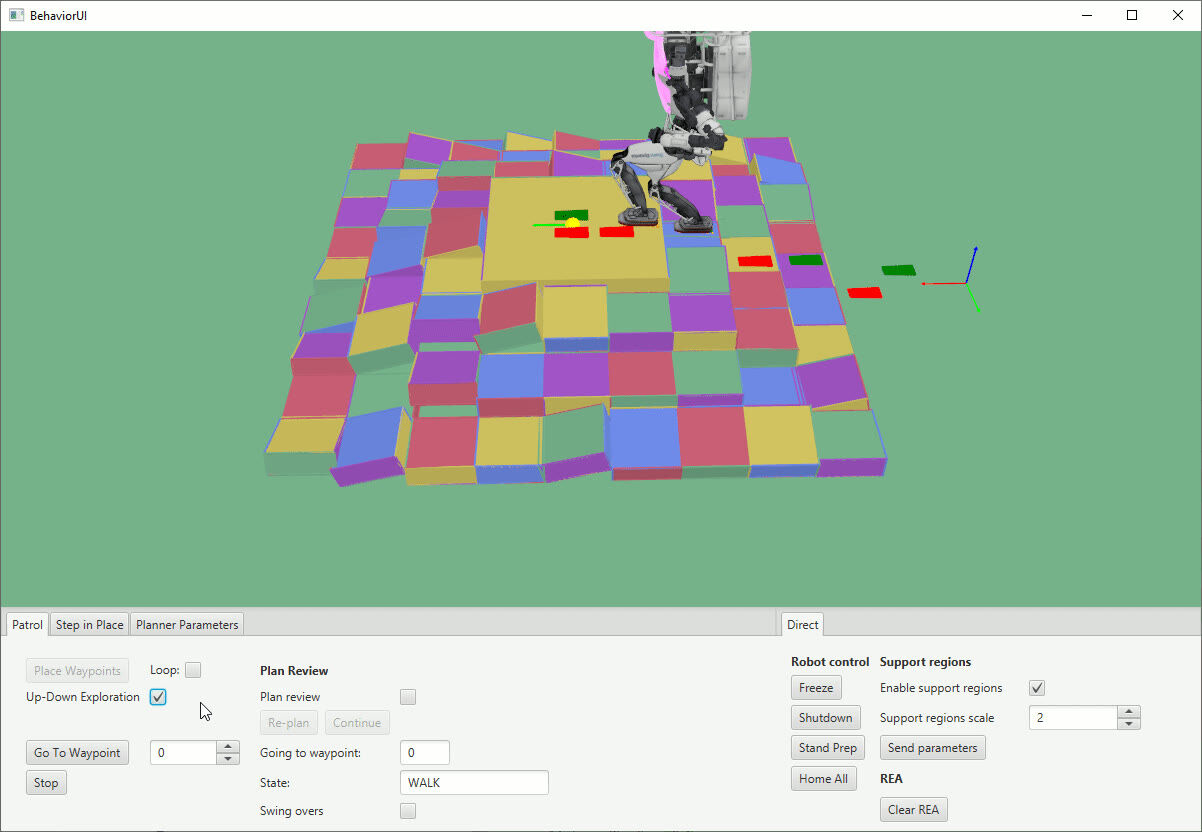}
    \caption{A simulation example of the rough terrain up and down exploration behavior.
    This behavior was a hard-coded state machine for perceptive locomotion over planar regions operated via the widgets shown.
    The robot would plan one or several steps at a time while perceiving and traversing rough terrain.
    It allowed for manual or automatic waypoint placement represented as the yellow ball and green arrow.
    The operator could execute one step at a time or allow automatic execution.
    The current state was rendered in the UI for monitoring.
    On the bottom right, direct robot management controls can be seen alongside planar regions and perception operations.
    REA stands for Robot Environment Awareness~\cite{Bertrand_2020}, which is the module that generates planar regions from the MultiSense SL lidar scanner.
    The clear function was used by the operator and the behavior to maintain usable planar regions, as the data would get messy or stale.
    This also highlights the JavaFX behavior user interface at the time. (May 2019)
    A video is available at \url{https://youtu.be/MHvjn6B9hQ4}.
    }
    \label{fig:rough_terrain_autonomy_sim}
\end{figure}

In 2017 and 2018, we started to develop more planning and perception capability, but using a separate UI framework from the DRC UI.
We developed an A* footstep planner~\cite{griffin2019footstepplanner} and a planar region segmentation perception algorithm~\cite{Bertrand_2020} that worked together.
Using the MultiSense SL lidar scanner, we could wait for 20 seconds or so, pool an OctoMap~\cite{Hornung_2013_OctoMap} terrain representation, segment it into planar regions, and plan footsteps over it.
This facilitated a leap in robot autonomy.
The robot was now able to traverse sections of rough terrain autonomously, as shown in \autoref{fig:rough_terrain_autonomy_atlas} and \autoref{fig:rough_terrain_autonomy_sim}.

These planning and perception features departed from the DRC UI, opting to use JavaFX for the algorithm visualization.
This meant that there was now a separate UI framework for algorithm development and robot operation.
A complication was that JavaFX could not render point clouds efficiently like the DRC UI could, causing somewhat of a rift when deciding how to visualize robot data.

\subsection{Navigation with Planar Regions}
\begin{figure}[H]
    \centering
    \includegraphics[width=.95\columnwidth]{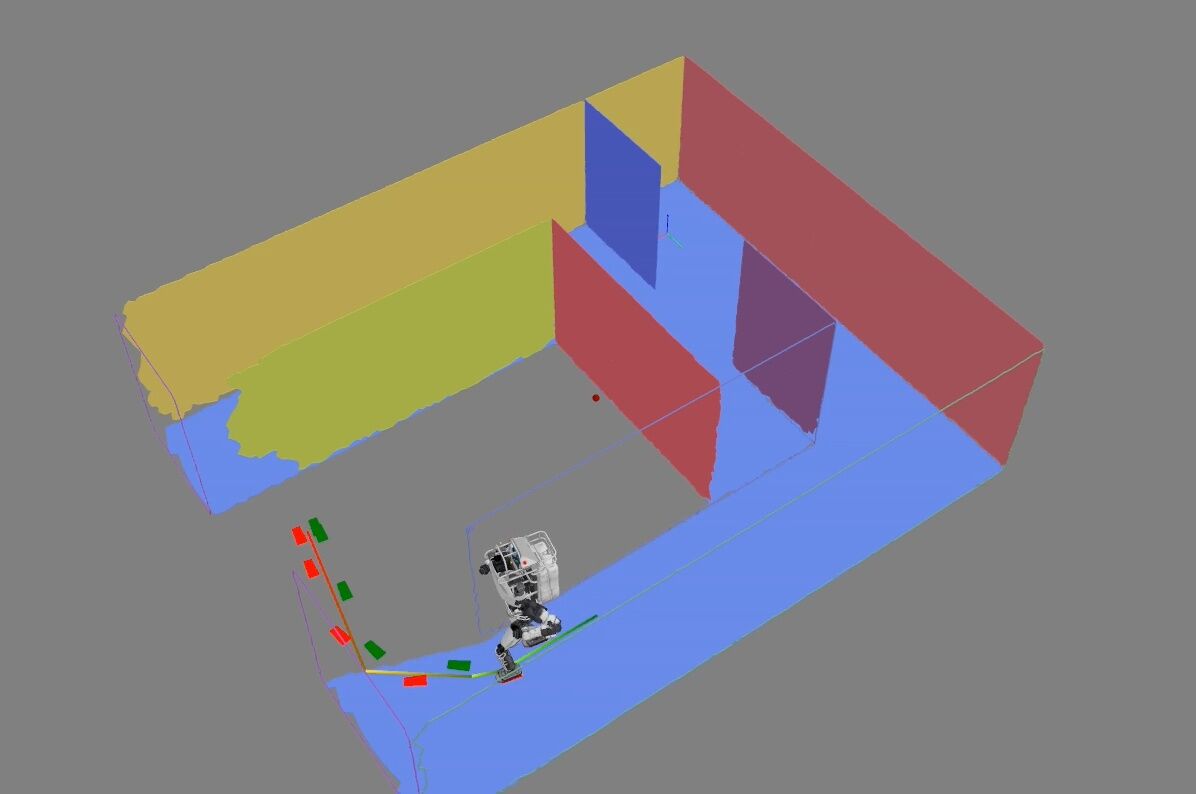}
    \caption{A simulation of Atlas performing a navigation behavior in the JavaFX-based behavior system user interface.
    This phase of the behavior system heavily relied on planar regions for locomotion.
    A video is available at \url{https://youtu.be/0Ygti7aOxYM}.
        (December 2019)
    }
    \label{fig:atlas_maze}
\end{figure}

Around 2019, we started to develop room-to-room navigation tools in the JavaFX ecosystem (\autoref{fig:atlas_maze}), while also starting the development of a rough terrain traversal behavior system that used active perception for each footstep.
It was in 2019 that the work on this thesis started directly.
One aspiration was to try to unify the tools used for robot operation and planning and perception algorithm development.
The reason this was important was to simplify the decision of what tools to use when developing a feature and also reduce code duplication.
For example, the JavaFX UI could not render point clouds efficiently and it was cumbersome to create UI widgets.
Conversely, the DRC UI code was aging and it was time to evaluate a more modern set of building blocks while cleaning up the code.

Another issue with the DRC UI was the friction in creating standalone applications.
When developing planners and perception algorithms, it is desirable to create a test or demo app just for that, while maintaining the usability of those components in larger apps, such as algorithm visualization elements.

\subsection{Virtual Reality Teleoperation}
Around this time, we had also developed some virtual reality (VR) applications based on the DRC UI, which allowed for teleoperating the robot.
This virtual reality interface made authoring one-time-use footstep plans really fast.
We ran a demonstration where the VR operator was able to manually place footsteps over a rough terrain cinder block field in the range of a few minutes.
One caveat to this implementation was that the operator had to either launch the 2D DRC UI or the VR one and could not switch between VR and mouse and keyboard modalities without restarting the user interface.
Restarting the user interface is a frustrating operator action because you can lose any persistent state you may have in setting up robot commands.
Additionally, it can take several minutes to relaunch the user interface.
This can really start to eat up time.

\subsection{Simulation Construction Set}
\begin{figure}[t]
    \centering
    \includegraphics[width=.9\columnwidth]{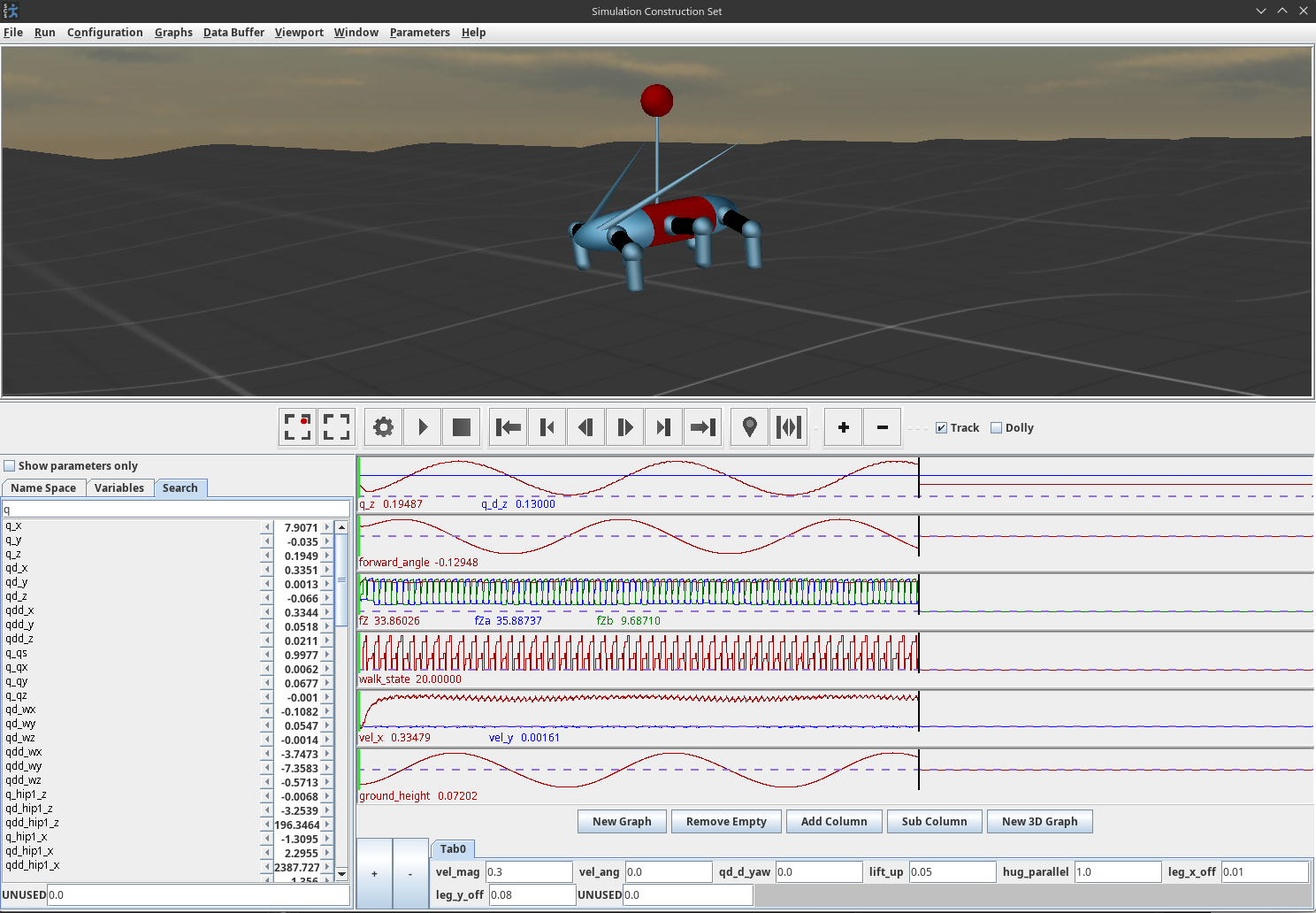}
    \caption{The Agile Hexapod simulation in Simulation Construction Set.
    In the upper area, the simulated robot and terrain is visible.
    In the center, simulation and buffer controls are available.
    The gear runs the simulation forward and the play button plays it back via the buffered data.
    In the bottom left there is an area to search for variables to plot or tune.
    In the bottom right variables are plotted or rendered as widgets where the variables can be modified.
    Modifying variables can be used to tune parameter values.
    Screenshot taken in 2026.}
    \label{fig:agile_hexapod}
\end{figure}

We also had a simulation and data visualization engine called Simulation Construction Set (SCS), which allows users to programmatically create rigid-body dynamics simulations and robot controllers.
On top of that, it featured a 3D view of the robot and scrubbable plots of buffered variable data, as seen in \autoref{fig:agile_hexapod}.
SCS was implemented using an older version of jMonkeyEngine and features were not interchangeable with our DRC UI.
For UI widgets, SCS used Java Swing, but using a very different code path than the Java Swing used in the DRC UI.
The takeaway here is that while SCS and the DRC UI used the same underlying libraries, functionality could not be shared between them.

\section{2020-2021 Atlas, RDX, Behavior Tree Era}

\subsection{Evaluating Next-Generation Tooling}
A Slack message from December 18, 2020, reads: ``[...] because of the graphics adapters for 2D and 3D you can't easily use the underlying engine features and you can't share graphics code with even our other JME apps like the DRC UI. [...] The JavaFX UIs, the (DRC) Operator UI, and SCS are basically not even close to being compatible with each other.''

In late 2020, we did a survey of software libraries in an attempt to build a framework that could act as a sandbox for robotic data exploration and planner development but also be extendable to state-of-the-art robot teleoperation.
Additionally, since the promise of VR was now clear, we wanted to include VR support as a base function in all applications, such that switching between mouse and keyboard and VR was as simple as enabling VR and putting on the headset.
We wanted the 3D scene to be shared between VR and the mouse and keyboard interface and treat VR as just another interaction method in the same class as monitor, mouse, and keyboard.

Another desirable characteristic of this new framework was ease of creating user interface widgets.
Often for designing, tuning, and interacting with experimental planners, controllers, and perception algorithms, the engineer wants to tweak variables quickly and easily to see the effects on the process and the data.
This is in stark contrast to considering user experience for every UI element.
In our survey, we found a library called ImGui~\cite{dearimgui} which fit this bill perfectly, allowing user interface elements to be rendered in application code, in the same place as the interaction logic is handled, and in very few lines of code.
It turned out to be a great way to provide base functionalities quickly while still allowing for custom rendered widgets that consider user experience.

Another goal of the new framework was to stay within the current software ecosystem so lab developers could seamlessly use it and contribute to it and so that existing lab tools could be used and not rewritten.
Our existing code was in Java so this meant we wanted to stick with Java.
Our thought was also to build on the most popular open-source tools in the community in order to avoid any rug pulls and also ride the wave of community maintenance and development.

For 3D graphics engines, there were two major competitors: jMonkeyEngine and libGDX~\cite{libgdx}.
The DRC UI was implemented using jMonkeyEngine but was on an old version.
libGDX had more stars on GitHub, a vibrant community, and an API that was closer to plain OpenGL, the underlying library.
Because we needed to maintain the DRC UI usability as a legacy app and it was somewhat tangled in an aging codebase, it was unclear if bringing it directly forward to modern libraries would work.

When designing the new framework, a few things were clear:
\begin{enumerate}
    \item We wanted to adopt the pattern used by Eclipse IDE~\cite{eclipseide}, where the application was a set of dockable panels.
    These panels should be showable and hideable, and the docking configurations should be easily savable and loadable.
    This system of UI design was shown to be very adaptable to different workflows and tasks.
    It is also similar to how other popular and versatile apps work, such as Blender~\cite{blender}.
    \item We wanted to build in virtual reality (VR) support at the base level.
    If you didn't have a VR headset, the app would operate normally.
    However, if you did, there was a checkbox to enable VR and you were able to jump into the main 3D view and interact with the same elements you could with mouse and keyboard.
    An element of the VR support was that the 3D view on the monitor would remain visible while in VR, and others could view the poses of the VR user's headset and controllers in 3D by standing next to the computer.
    \item We wanted to implement the main features of the DRC UI such as the gizmos, the walk path control ring, modifiable footsteps, and affordance templates.
\end{enumerate}

ImGui~\cite{dearimgui} included support for the dockspaces of panels, multi-window support, and even saving those entire layouts to .ini files.
OpenVR~\cite{openvr} was available for Java and supported any VR headset that would work in SteamVR, which is nearly all of them.
We decided to rewrite the new framework from scratch and use libGDX, ImGui, and OpenVR.

\subsection{Robot Data eXplorer}
\begin{figure}[t]
    \centering
    \includegraphics[width=.9\columnwidth]{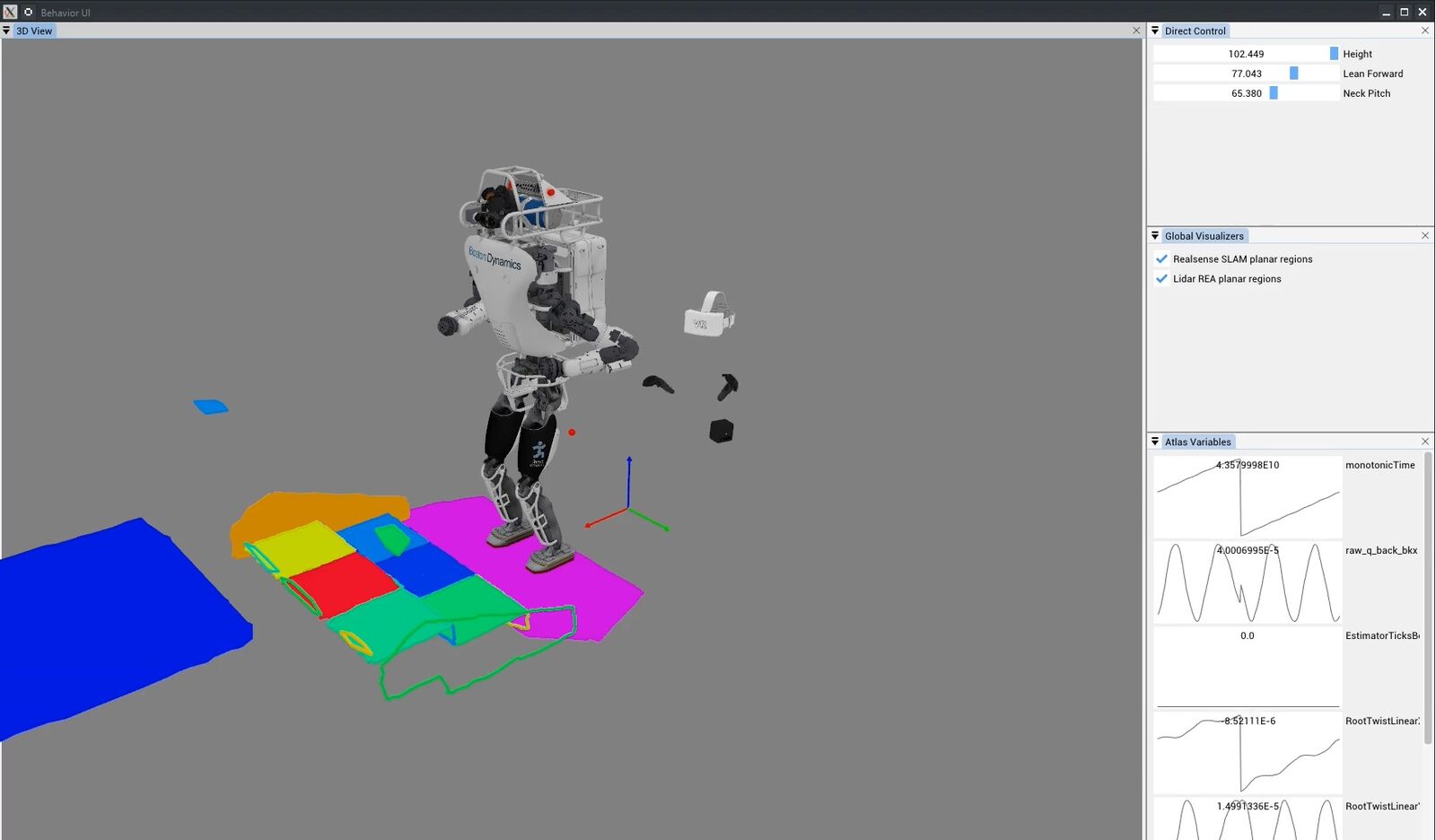}
    \caption{A screenshot of one of the first Robot Data eXplorer (RDX) applications on January 13, 2021.
    It features an Atlas simulation with planar regions.
    RDX features dockable panels of widgets, an interactive 3D view, and VR support by default.
    It is designed as a sandbox for general robotics data computation and exploration while also being extendable to more purpose-specific applications such as teleoperation and simulation.
    A video is available at \url{https://youtu.be/APSf1y6VryQ}.
    }
    \label{fig:rdx_early}
\end{figure}

This effort ultimately culminated in the robotics sandbox application suite later named Robot Data eXplorer (RDX), shown in \autoref{fig:rdx_early}.
We wanted this set of tools to be useful for teleoperation, simulation, and perception and control algorithm development.
It was designed in a similar spirit to SCS in that the user would first create a main entry point for their project and implement what they needed programmatically.
For example, the application could be a test of a point cloud renderer, displaying it in 3D while also providing tuning widgets for point size, color, etc.
We also wanted RDX to be versatile enough such that our primary applications could use it, such as our teleoperation interface (like the DRC UI).
This meant that it needed to support not just utilitarian development widgets, but also custom rendered icons and interactive elements.
ImGui and libGDX supported all of this and we now use RDX in 2026 as our primary teleoperation interface.

The DRC UI mainly supported keyboard controls for the gizmo.
In RDX, we wanted to make it more intuitive to use by implementing click-and-drag controls for both orientation and translation.
We implemented a mouse ray projection into the 3D scene in RDX which was generally available to any component in RDX.
For the gizmo, this was used to intersect with the gizmo orientation control tori and translation control cylinders and cones.
The gizmo's axes are colored as Red, Green, and Blue, to match X, Y, and Z, and Roll, Pitch, and Yaw.
The way to remember it is ``RGB -> XYZ''.
We also re-implemented the walk path control ring and virtual interactable footsteps, as shown in a 2022 screenshot in \autoref{fig:nadia_interactables}.

The VR mode in RDX also featured ImGui widgets.
The same panels available in the monitor, mouse, and keyboard modalities were also available as panels mounted on the wrist in VR, and they were operable via point and click with the VR controller.
This continued the RDX design goals of reusability and versatility into the VR modality.
We wanted VR to be simply another interaction method for the same data, not a separate framework.
Additionally, we wanted switching between the monitor, keyboard, and mouse and virtual reality to be as simple as putting on and taking off the headset and picking up and setting down the controllers.

\subsection{Perceptive Locomotion}
\begin{figure}[H]
    \centering
    \includegraphics[width=.95\columnwidth]{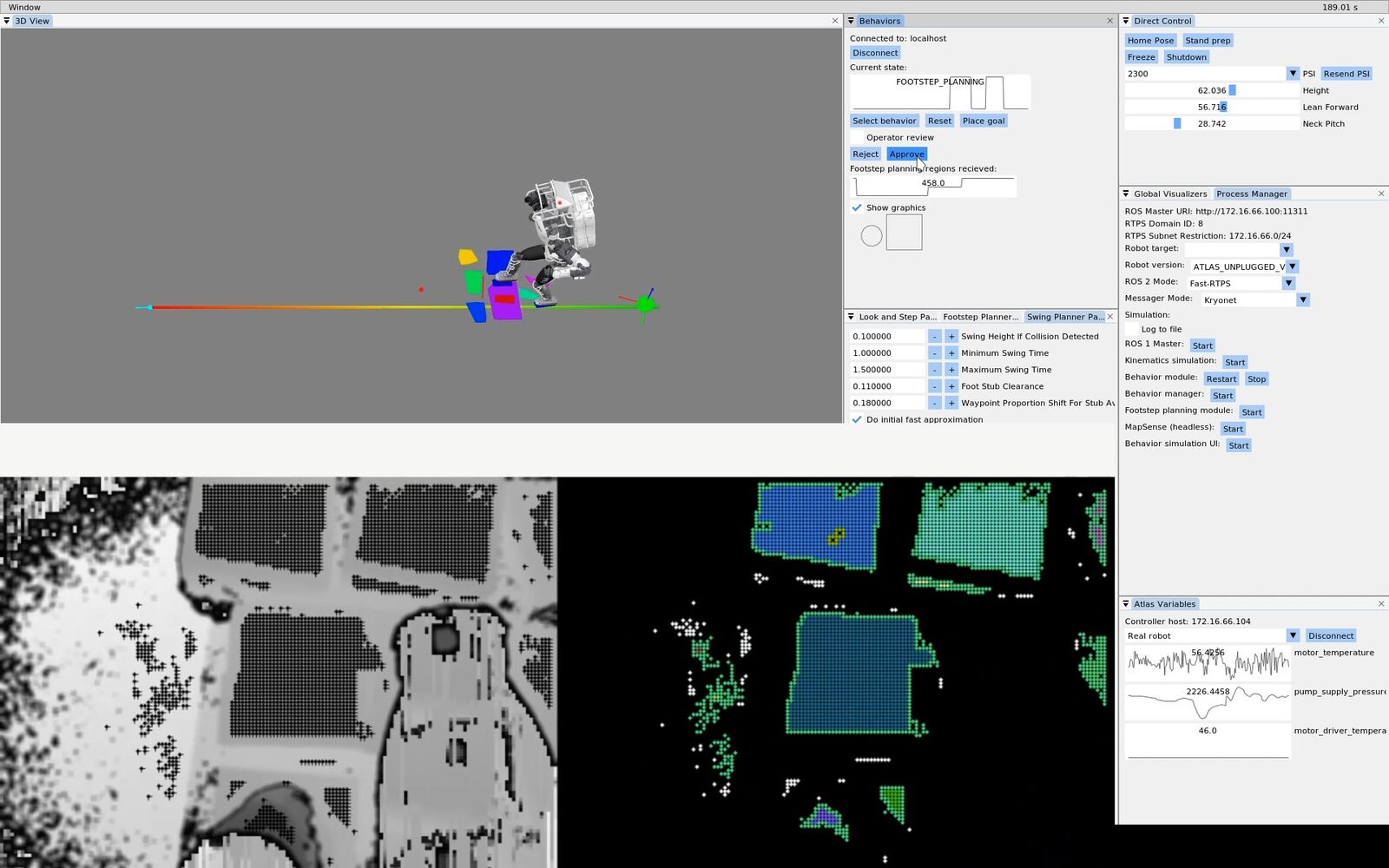}
    \caption{A perceptive locomotion behavior on Atlas in 2021 using RDX for operation.
    This behavior used the rapid planar region extractor from~\cite{Mishra_2021_RapidRegions} as an input to a hard-coded reactive perceptive locomotion behavior called ``look and step''.
    In the top left, the 3D scene is shown featuring the live robot configuration, the perceived 3D planar regions, and the walk path.
    The input to the behavior was a waypoint, as shown by the small teal ball and arrow at the end of the green, yellow, and red path.
    The behavior would continuously walk along the path, planning 1-3 steps at a time while doing a percept on every step.
    In the top right, the ``Behaviors'' panel is shown, which offers the ability to place a goal and toggle whether the operator was required to review each step.
    Additionally, a plot of the current behavior state is shown.
    A video is available at \url{https://youtu.be/qAtfV7hTzYg}.
    }
    \label{fig:atlas_rdx_perceptive_locomotion_demo}
\end{figure}

In 2020, we started using the RealSense L515 depth sensor~\cite{l515} for perceptive locomotion.
Bhavyansh Mishra developed a planar region extractor using OpenCL~\cite{Mishra_2021_RapidRegions} which enabled the development of a perceptive locomotion behavior with active perception while walking.
This led to one of the first autonomous behaviors that used RDX as the user interface, as shown in \autoref{fig:atlas_rdx_perceptive_locomotion_demo}.
This work is covered in the 2022 paper~\cite{lookandstep}.

\subsection{Behavior Trees}
\begin{figure}[t]
    \centering
    \includegraphics[width=.5\columnwidth]{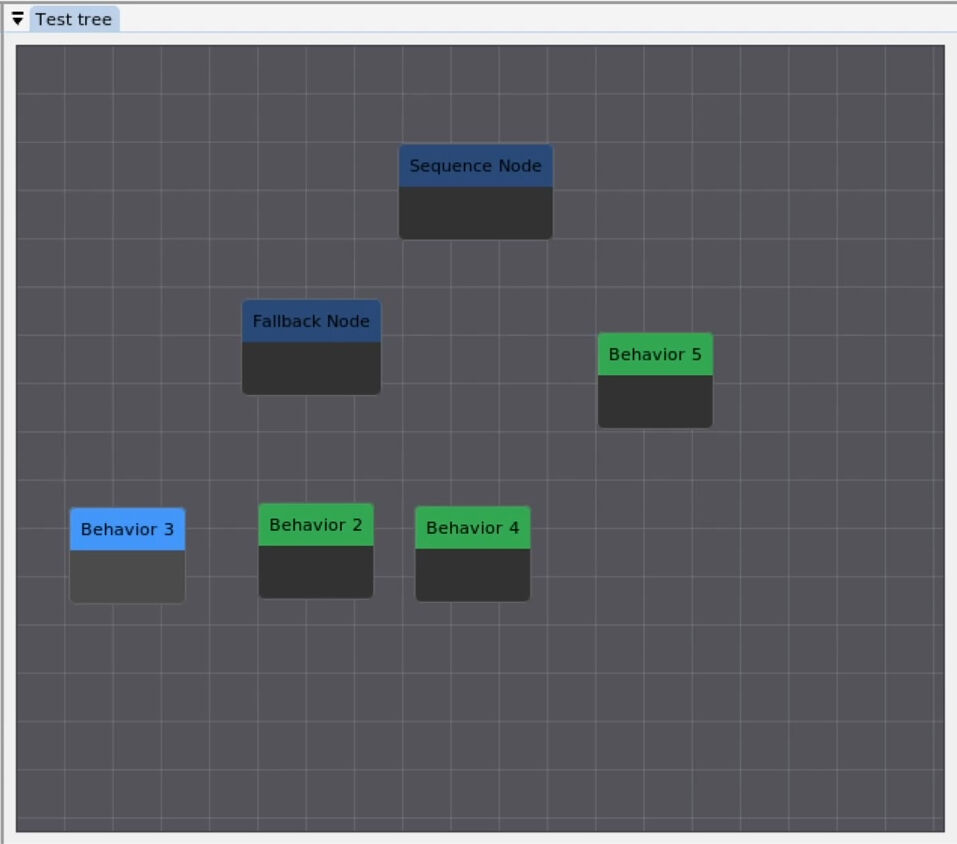}
    \caption{Development of the 2D canvas behavior tree view began in May 2021.
    A video is available at \url{https://youtu.be/kpWE3UVIIvM}.}
    \label{fig:atlas_2d_behavior_tree_view}
\end{figure}

After the initial development of the perceptive locomotion behavior, ``look and step'', we sought to build out multi-task behaviors as part of our building exploration demo.
To assemble a hybrid teleoperated and autonomous multi-stage behavior, we looked to behavior trees.
\autoref{fig:atlas_2d_behavior_tree_view} shows the very beginning of that effort in May of 2021.
We pulled in an ImGui library called ImNodes and started rendering a tree of behaviors in a 2D canvas.

\subsection{Semi-Autonomous Building Exploration}
\begin{figure}[H]
    \centering
    \includegraphics[width=.95\columnwidth]{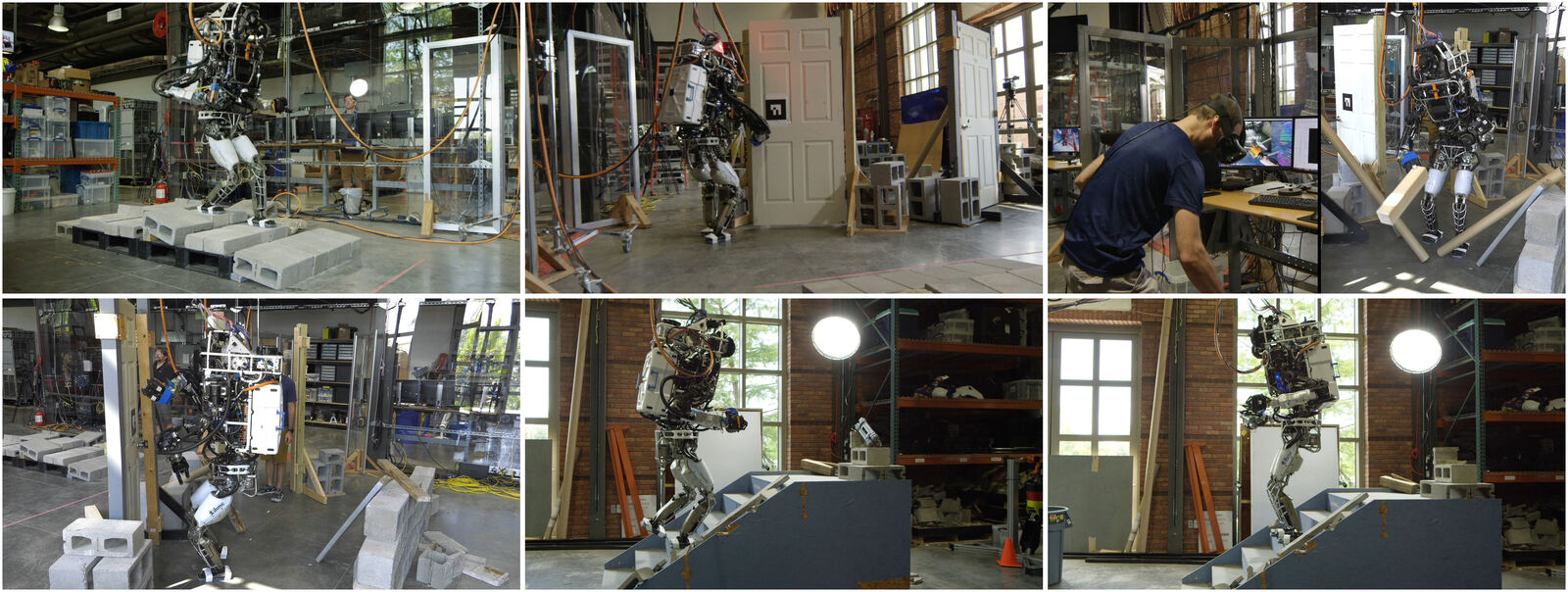}
    \caption{The June 23, 2021 building exploration demo consisting of seven tasks in a hybrid autonomy model.
    A video is available at \url{https://youtu.be/CFiFaO-ENPw}.}
    \label{fig:2021_building_exploration_demo}
\end{figure}

In June 2021, we worked on a building exploration demo which consisted of seven tasks with a hybrid autonomy model: an autonomous rough terrain traversal using perceptive locomotion, an automatic pull door behavior, VR teleoperated debris clearing, and automatic push door traversal, obtaining a mock pipe bomb from the top of a flight of stairs, walking back down, and placing the mock pipe bomb in a trash can.
The tasks are shown in \autoref{fig:2021_building_exploration_demo}.
We successfully performed the demo on June 23, 2021, in 22 minutes.

\begin{figure}[t]
    \centering
    \includegraphics[width=.95\columnwidth]{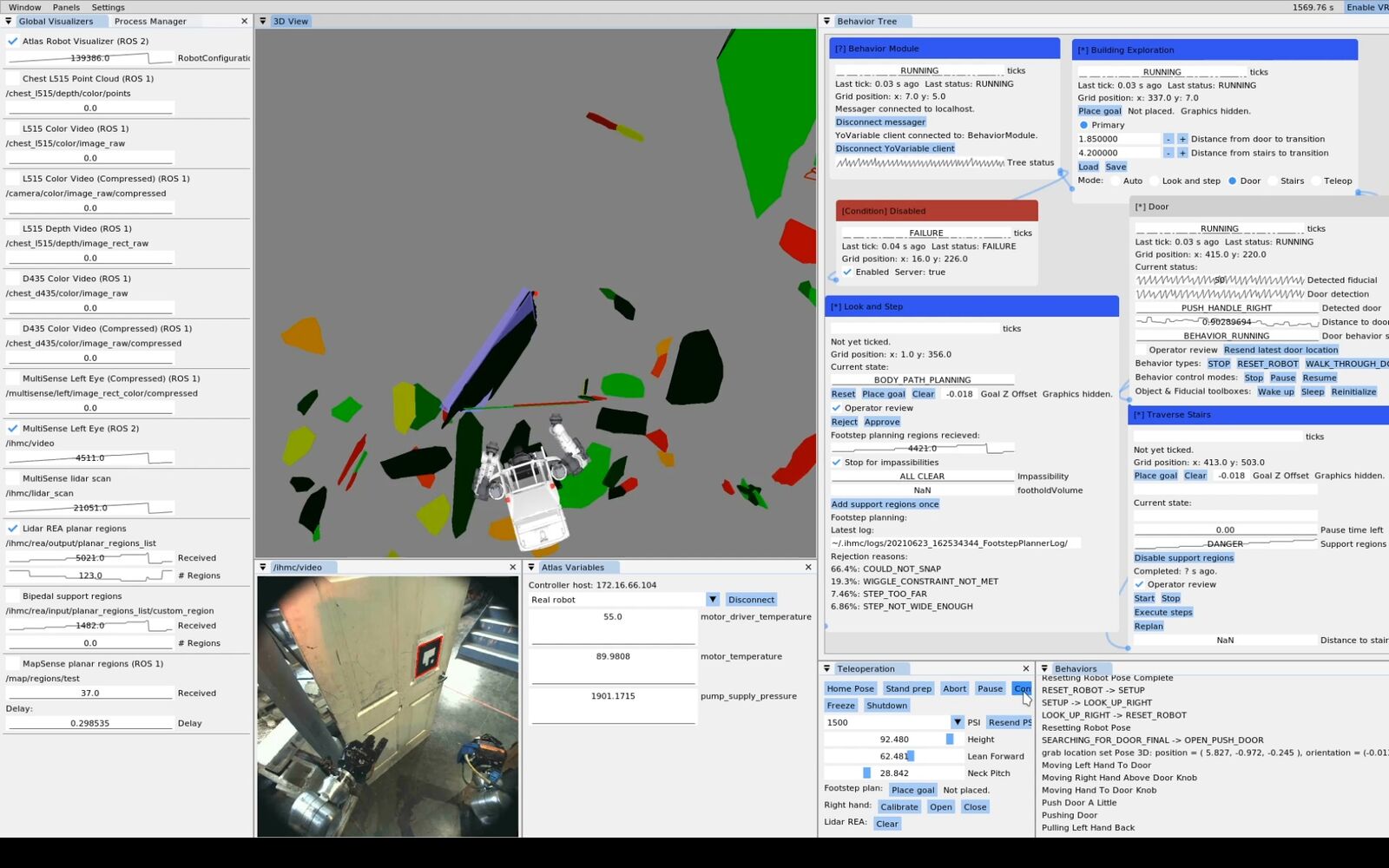}
    \caption{The June 23, 2021 building exploration demo user interface.
    In the center, a 3D view with the robot and the planar region environment is shown.
    In the bottom center, a video panel shows the first person view of the robot with an overlay that outlines the ArUco fiducial marker.
    On the upper right, a 2D canvas behavior tree view is shown.
    It consists of the root node, named ``Behavior Module'', a ``Building Exploration'' node, a ``Disabled'' node, a ``Door'' node, a ``Look and Step'' node, and a ``Traverse Stairs'' node.
    This hard-coded but interactable behavior tree allows the operator to monitor and guide the building exploration behavior at several levels of autonomy.
    A video is available at \url{https://youtu.be/CFiFaO-ENPw}.}
    \label{fig:2021_building_exploration_demo_ui}
\end{figure}

This demo tested our ability to incorporate autonomous behaviors with operator supervision and teleoperation.
Major innovations in the demo with respect to the DRC were VR kinematics streaming, continuous perceptive locomotion, and automatic push and pull door behaviors.
However, the behavior structure for the locomotion, the door traversals, and the building exploration assembly were still hard-coded and not runtime editable, and perception of semantic objects was still reliant on fiducial markers.
Push and pull door and stairs behaviors were selected by the fiducial IDs.
The UI for this demo can be seen in \autoref{fig:2021_building_exploration_demo_ui}.



%
%

\subsection{Perception Sensor Upgrades}
\begin{figure}[t]
    \centering
    \includegraphics[width=.95\columnwidth]{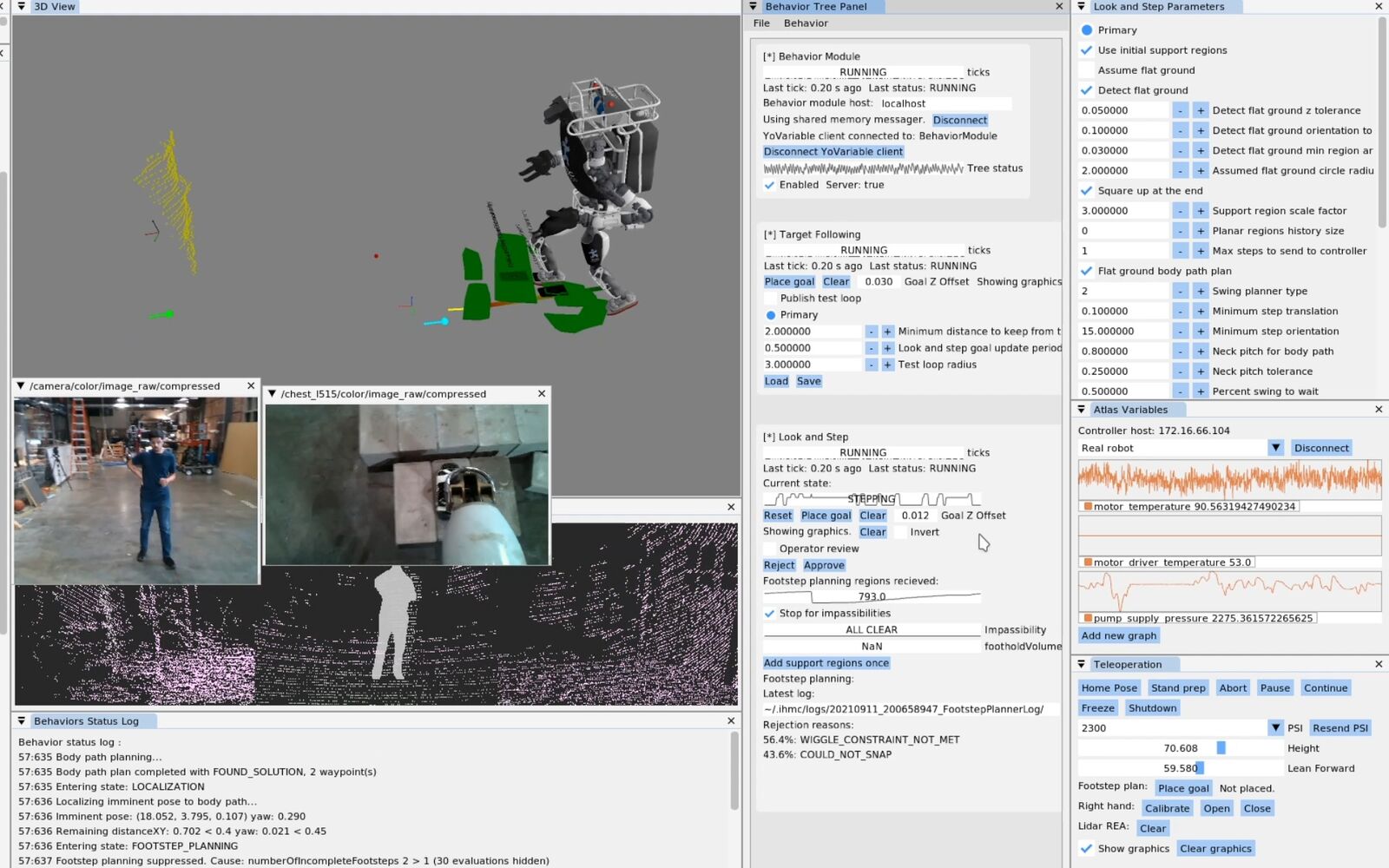}
    \caption{A person following behavior which used a YOLO~\cite{redmon2016yolo} person detection and segmentation model in combination with the Ouster point cloud to provide a continuously updating position of a person.
    The existing ``Look and Step'' locomotion behavior was adapted to walk towards the person in a loop.
    This demo was conducted in September, 2021.
    A video is available at \url{https://youtu.be/z51BHiq3OQs}.}
    \label{fig:atlas_follow_person_over_rough_terrain}
\end{figure}

After the building exploration demo, we switched from the MultiSense SL to the Ouster OS0-128 lidar sensor~\cite{ouster_os0_lidar_sensor} in combination with a stereo pair of Blackfly cameras with fisheye lenses.
A paper about this sensor suite is available at~\cite{Mishra_2022_PerceptionEngine}.
With the new sensor configuration, we had instantaneous dense lidar scans instead of needing to wait for the MultiSense SL to spin.
This enabled us to develop the person following behavior shown in \autoref{fig:atlas_follow_person_over_rough_terrain}.

%
%

\section{2022-2023 Nadia, Runtime-Editable Sequences Era}

\subsection{Runtime-Editable Sequences}
\begin{figure}[t]
    \centering
    \includegraphics[width=.95\columnwidth]{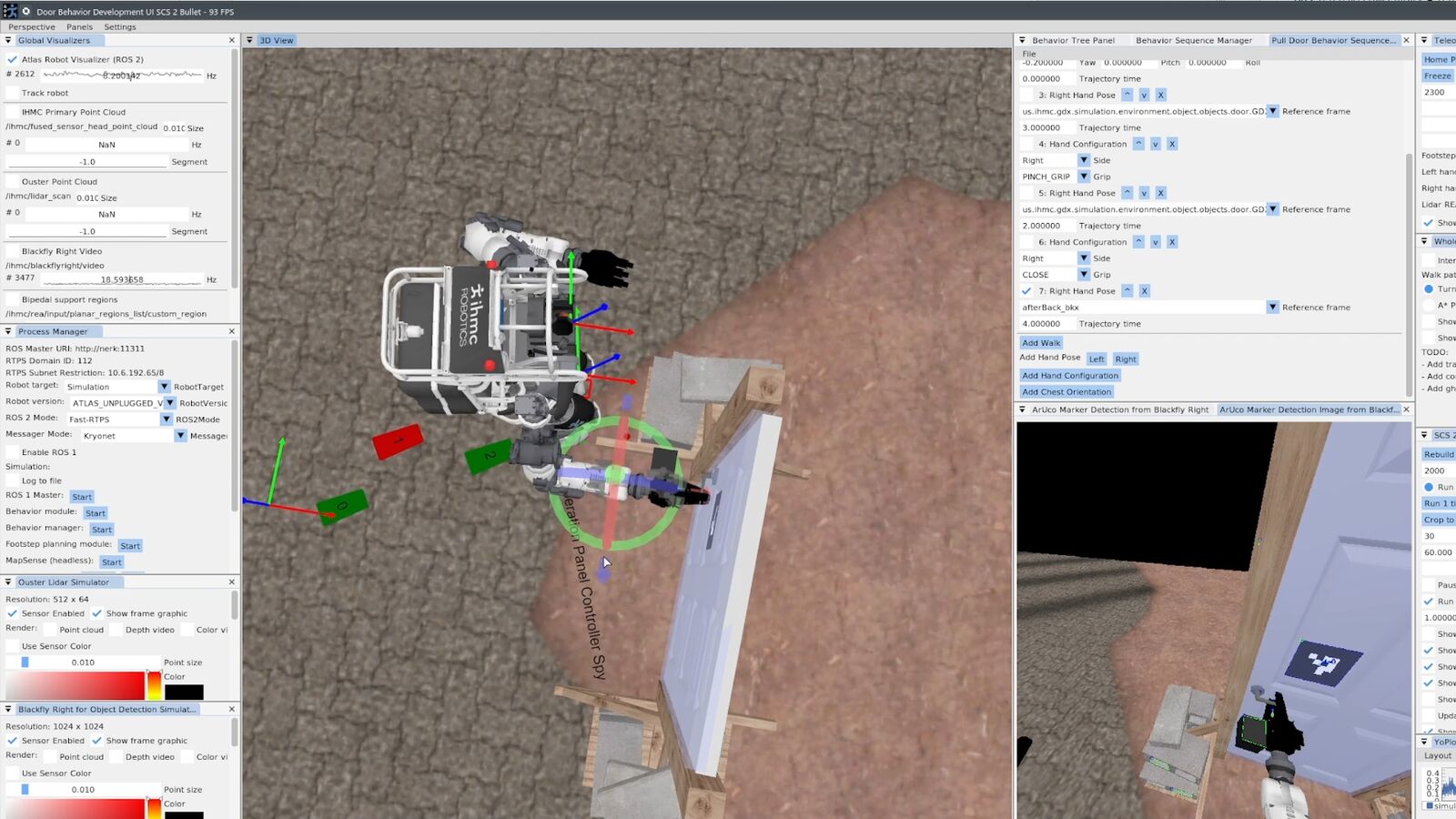}
    \caption{A Bullet~\cite{coumans2021} physics simulation of Atlas with a new runtime behavior authoring implementation.
    A pull door opening behavior is being authored.
    At this time there were only 4 action types: Walk, Hand Pose, Hand Configuration, and Chest Orientation.
    A video is available at \url{https://youtu.be/KfUFNM7SWz8}.
    April 6, 2022.
    }
    \label{fig:atlas_sim_early_editable_sequence}
\end{figure}

In 2022, inspired by the Affordance Template framework, we started developing an interactive and runtime editable behavior authoring pipeline.
This was the beginning of the architecture we still use in 2026.
The initial goal was to replicate the door behaviors we had before, but in a runtime editable way.
To do this, the behavior state was implemented as a synchronized Conflict-Free Replicated Data Type (CRDT).
This allows the operator and robot to share and co-modify the behavior action definitions and run modes.
A screenshot from an early demo can be seen in \autoref{fig:atlas_sim_early_editable_sequence}.
This version of the behavior system had 4 action types: Walk, Hand Pose, Hand Configuration, and Chest Orientation.
The Walk and Hand Pose actions were defined in task frames.
It saved and loaded behaviors to and from JSON.
However, it only supported a linear sequence of actions and they were only executable one at a time.

\subsection{Teleoperated Building Exploration}
\begin{figure}[H]
    \centering
    \includegraphics[width=.9\columnwidth]{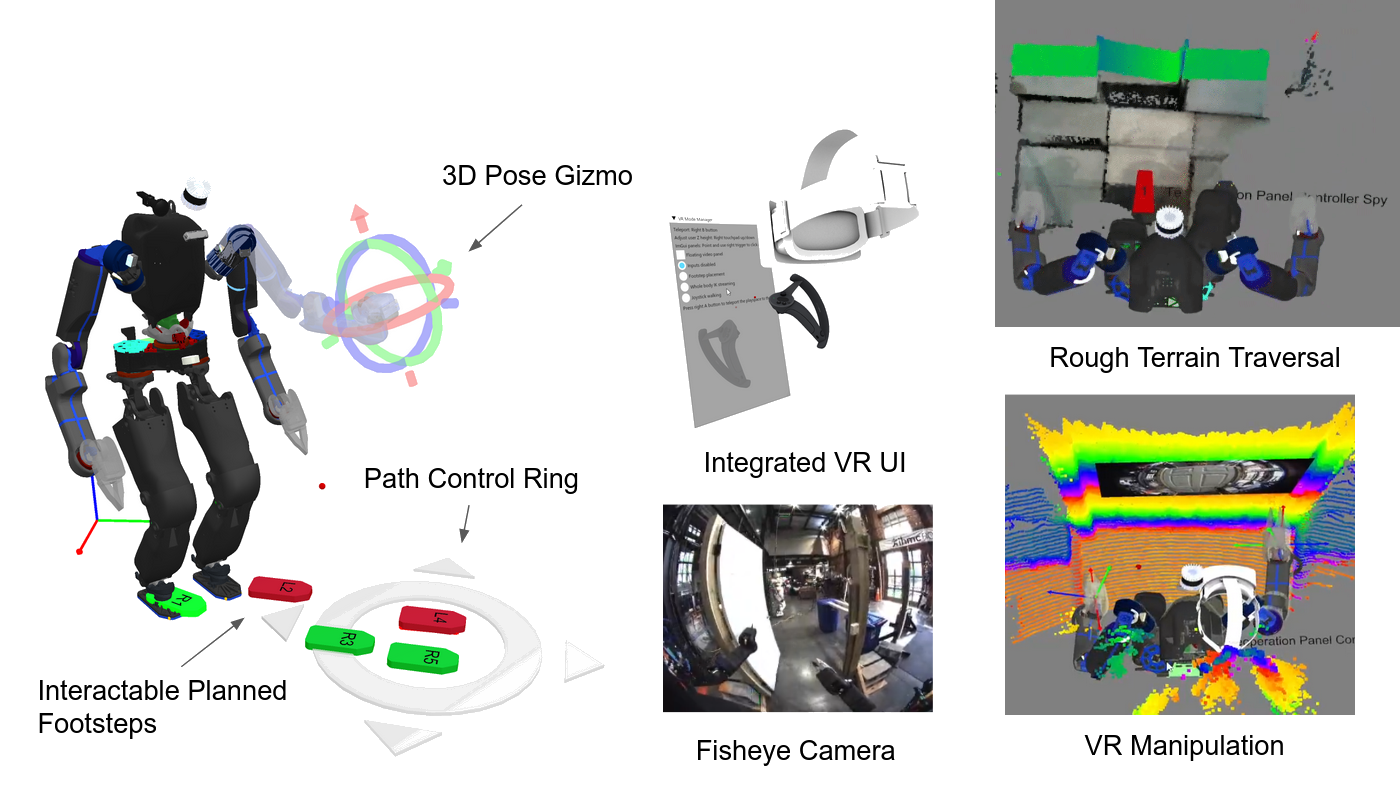}
    \caption{A presentation of teleoperation features in 2023 for the Nadia humanoid robot in RDX.
    On the left, a hand 3D pose gizmo is activated, showing a kinematic arm preview.
    At the robot's feet are planned but modifiable footsteps and the walk path control ring.
    In the center top, an RDX screenshot shows a VR user interacting with a 3D-situated ImGui panel, providing the same panels available to the mouse and keyboard operator.
    In the center bottom, a view of the fisheye camera on Nadia's head can be seen.
    In the top right, we show our mouse and keyboard footstep placement interface.
    In this interface, the camera is positioned over the robot and follows the robot while the operator hovers virtual footsteps which snap to the point cloud and clicks the mouse to place a footstep.
    The operator then presses the spacebar to execute however many footsteps were placed.
    In the bottom right, the VR kinematics streaming mode can be seen, which allows the operator to dynamically pilot the robot's arm motions as their own while being able to open and close the grippers using the VR controller triggers.
    This can be used to clear debris and open push doors while standing still.
    The operator was able to walk the robot with the joysticks in this mode, but not while streaming the arms.
    The joysticks commanded x, y, and yaw velocities.
    }
    \label{fig:nadia_interactables}
\end{figure}

\begin{figure}
    \centering
    \includegraphics[width=.95\columnwidth]{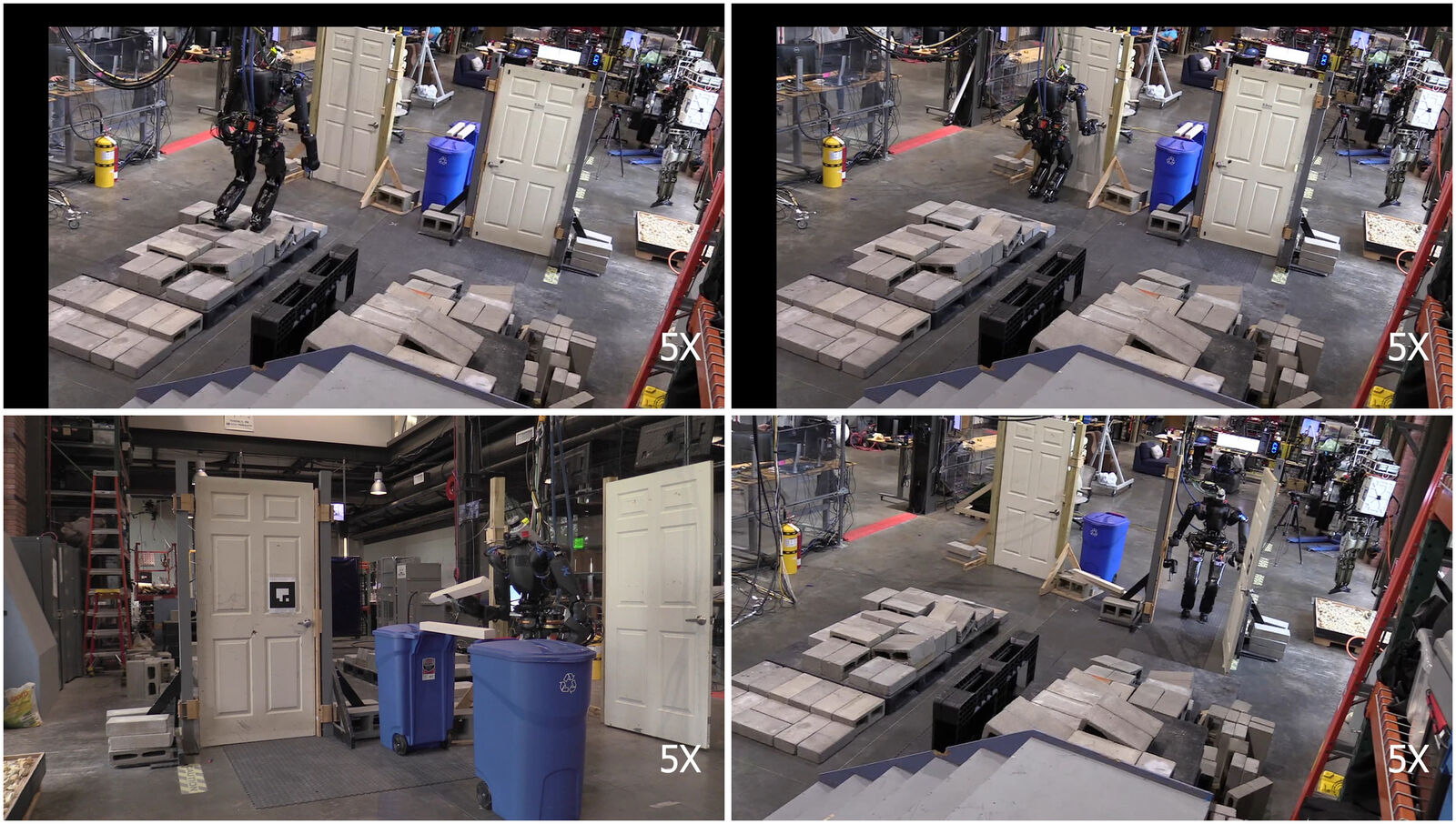}
    \caption{The Milestone 1 tasks from April 2023: rough terrain (top left), push door (top right), debris (bottom left), and push door (bottom right).}
    \label{fig:milestone_1_tasks}
\end{figure}

Also in 2022, we switched to IHMC and Boardwalk Robotics's Nadia humanoid robot and set some milestones for the next few years.
Milestone 1 was a teleoperation-only multi-task demo.
Milestone 1 was performed on the robot on September 9, 2022 using the RDX teleoperation UI, using an upgraded set of teleoperation features shown in \autoref{fig:nadia_interactables}.
In that demo, Nadia traversed rough terrain, traversed a push door, cleared debris, and traversed another push door in 7 minutes.
The tasks for Milestone 1 can be seen in \autoref{fig:milestone_1_tasks}.
During this demo and with this set of functionality we found that some methods of teleoperation were faster in VR and some were faster with mouse and keyboard.
RDX supported switching between them quickly.
We used mouse and keyboard to place individual footsteps over a cinder block field and then switched to VR for debris clearing and the push doors, where we used the VR controllers to continuously stream the hand poses and the VR controller trigger buttons to open and close the hands.
VR was also used to drive the robot with the joysticks.

This user interface also supported multiple operators as a side effect.
Often, the VR operator would have another operator observing the monitor and sometimes clicking buttons to put the robot in pre-defined arm configurations, like the home and door avoidance.
This helped us overcome gaps in the VR interface quickly and allowed for ``copilot'' supervision and monitoring, which doesn't usually work well for a VR system.

\subsection{Semi-Autonomous Pick and Place}
A primary goal of the Milestone 1 demo was to get base functionalities in place with the plan to start adding autonomous components in later milestones.
To that end, in 2023, we developed the next generation of behaviors, which were entirely authorable at runtime.
We started with a can pick and place behavior and the push and pull door traversals.

\begin{figure}[H]
    \centering
    \includegraphics[width=.95\columnwidth]{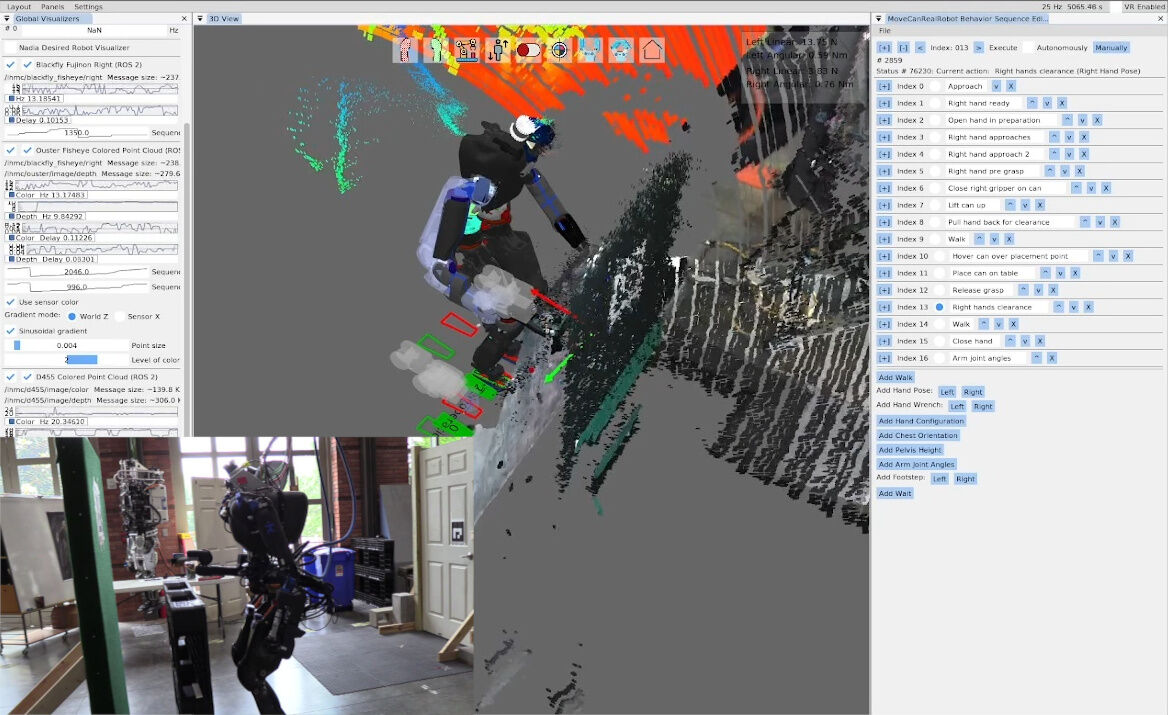}
    \caption{Nadia executing a can of soup pick and place behavior on June 20, 2023.
    Videos are available at \url{https://youtu.be/V8jMvhVdP8k} and \url{https://youtu.be/ZBj8zs1wzik}.
    }
    \label{fig:nadia_pick_soup}
\end{figure}

On June 20, 2023, we executed a can of soup pick and place behavior in 1 minute and 50 seconds as shown in \autoref{fig:nadia_pick_soup}.
This behavior was slow, not autonomous, and used an ArUco marker instead of detecting the can directly.
Our control of the SAKE Robotics EZGripper~\cite{sakerobotics} in Nadia at this time was unreliable and there was no fallback node yet so the operator had to keep trying the gripper open and close actions.
The operator had to visually check the state of the gripper fingers and re-execute the action if necessary.
There was no way to get the state of the gripper from the user interface at this time.
The operator also had to wait for each action to complete before starting the next because action completion conditions were not finished.
Regardless, it was a good real-robot test of a savable/loadable and runtime editable affordance template behavior sequence.

\subsection{Runtime-Editable Door Behaviors}
\begin{figure}[H]
    \centering
    \includegraphics[width=.95\columnwidth]{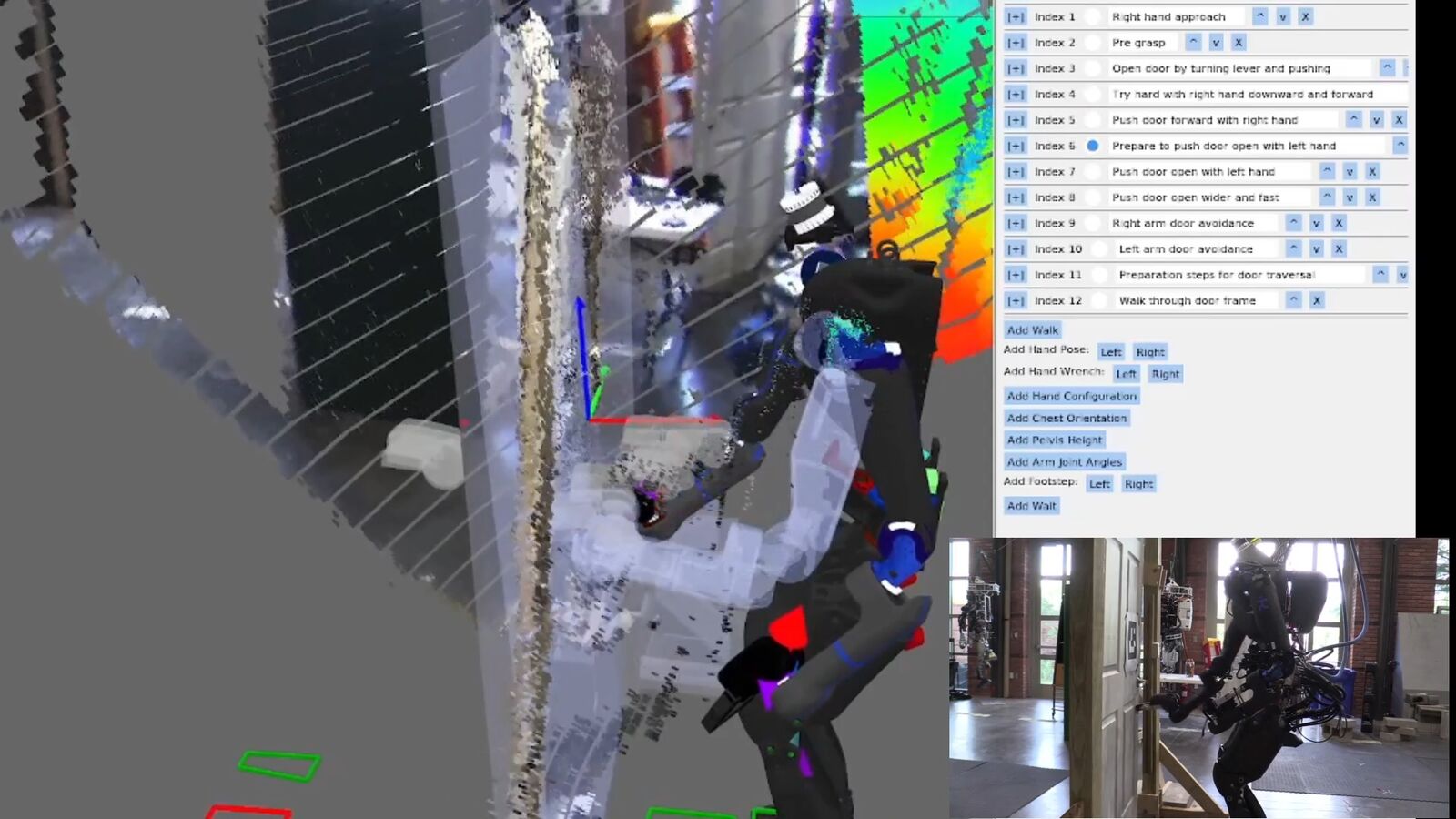}
    \caption{Nadia executing an autonomous push door behavior in 36 seconds on June 27, 2023.
    This was the first autonomous door traversal on Nadia.
    A video is available at \url{https://youtu.be/sMw1A72bqNo}.
    }
    \label{fig:nadia_first_auto_push_door}
\end{figure}

Our first autonomous push door traversal on Nadia using this system was on June 27, 2023, as shown in \autoref{fig:nadia_first_auto_push_door}.
UI improvements were made to condense each action node's settings into collapsible areas so we could display larger sequences.
Several new action types were added: hand wrench, pelvis height, arm joint angles, manual footsteps, and wait.
We were able to autonomously traverse the push door in 36 seconds with this version of the system.
The behavior was based on the pose of an ArUco marker detection.
In this generation of behaviors we used the Ouster point cloud colored using the left Blackfly fisheye camera image for situational awareness in the operator UI.

The hand wrench action was used for a box pickup behavior which didn't end up working very well.
The wrenches were used for squeezing the box from the sides to hold it.
However, the box wasn't fully constrained and would wobble.
Later, we decided to change the box picking strategy to using the grippers to pinch and lift from the top edges of a tote.

\begin{figure}[H]
    \centering
    \includegraphics[width=.95\columnwidth]{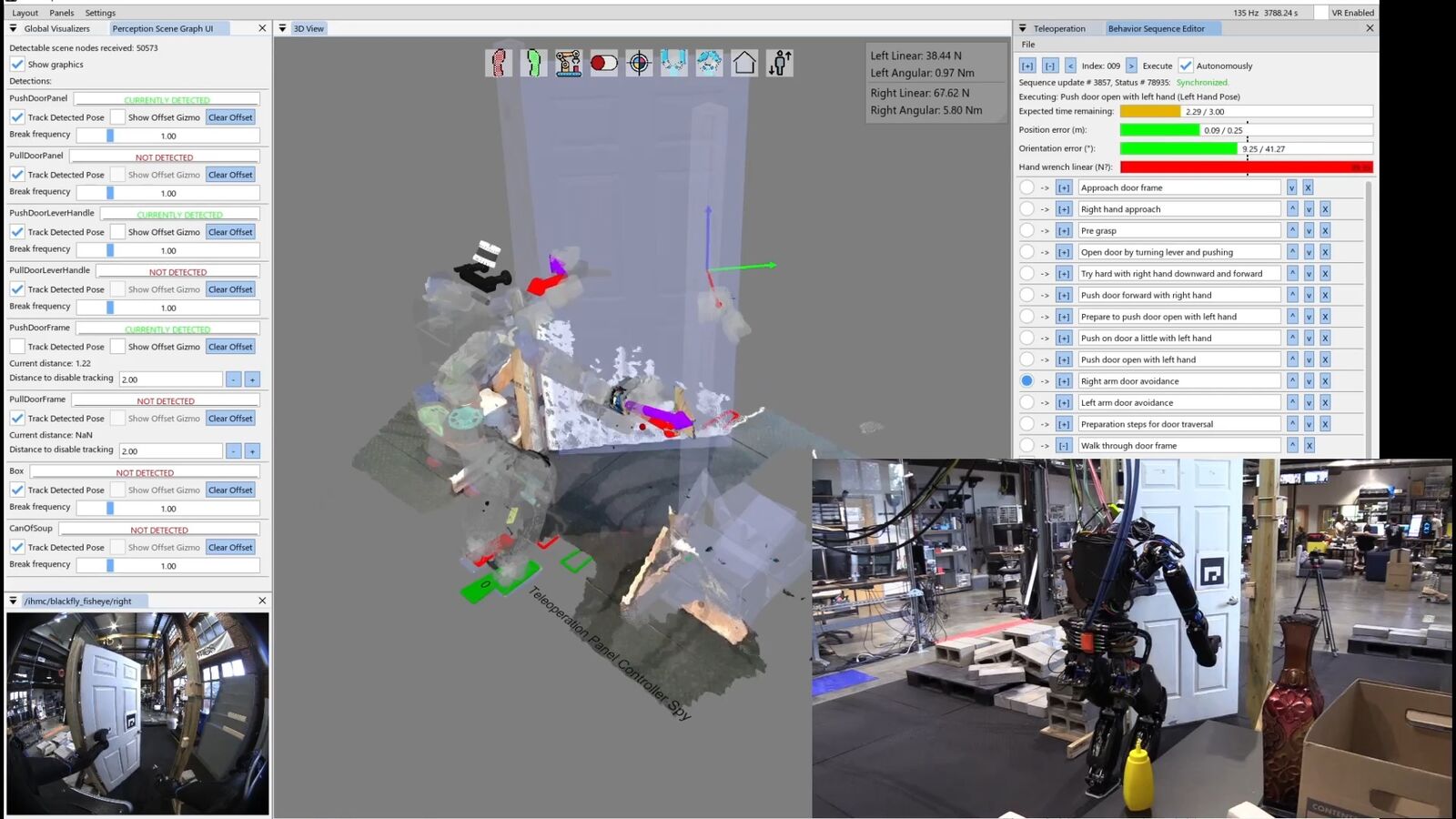}
    \caption{An autonomous push door behavior on Nadia on August 22, 2023.
    A video is available at \url{https://youtu.be/XY7We5PYQds}.
    }
    \label{fig:aug_push_door_behavior}
\end{figure}

\autoref{fig:aug_push_door_behavior} shows an improved version of the push door behavior executing on August 22, 2023.
In this version, there are several enhancements to the behavior user interface.
The first person view was included in the interface, allowing the operator to monitor the robot's view of the door and the hands without needing line of sight.
Other enhancements included 3D force and torque indicators on the hands.
The idea behind this was to experiment with force based control by first taking readings on the current measurements.
The force and torque data was pretty noisy and we weren't sure how to handle it.
Another enhancement was the action progress bar indicators.
This allows the operator to monitor the execution progress of the actions and give the operator confidence that action nominal duration and termination conditions were working correctly.
The speed of the behavior was the same as the June run at 36 seconds.

\subsection{Footstep Plan Authoring}
\begin{figure}[H]
    \centering
    \includegraphics[width=.95\columnwidth]{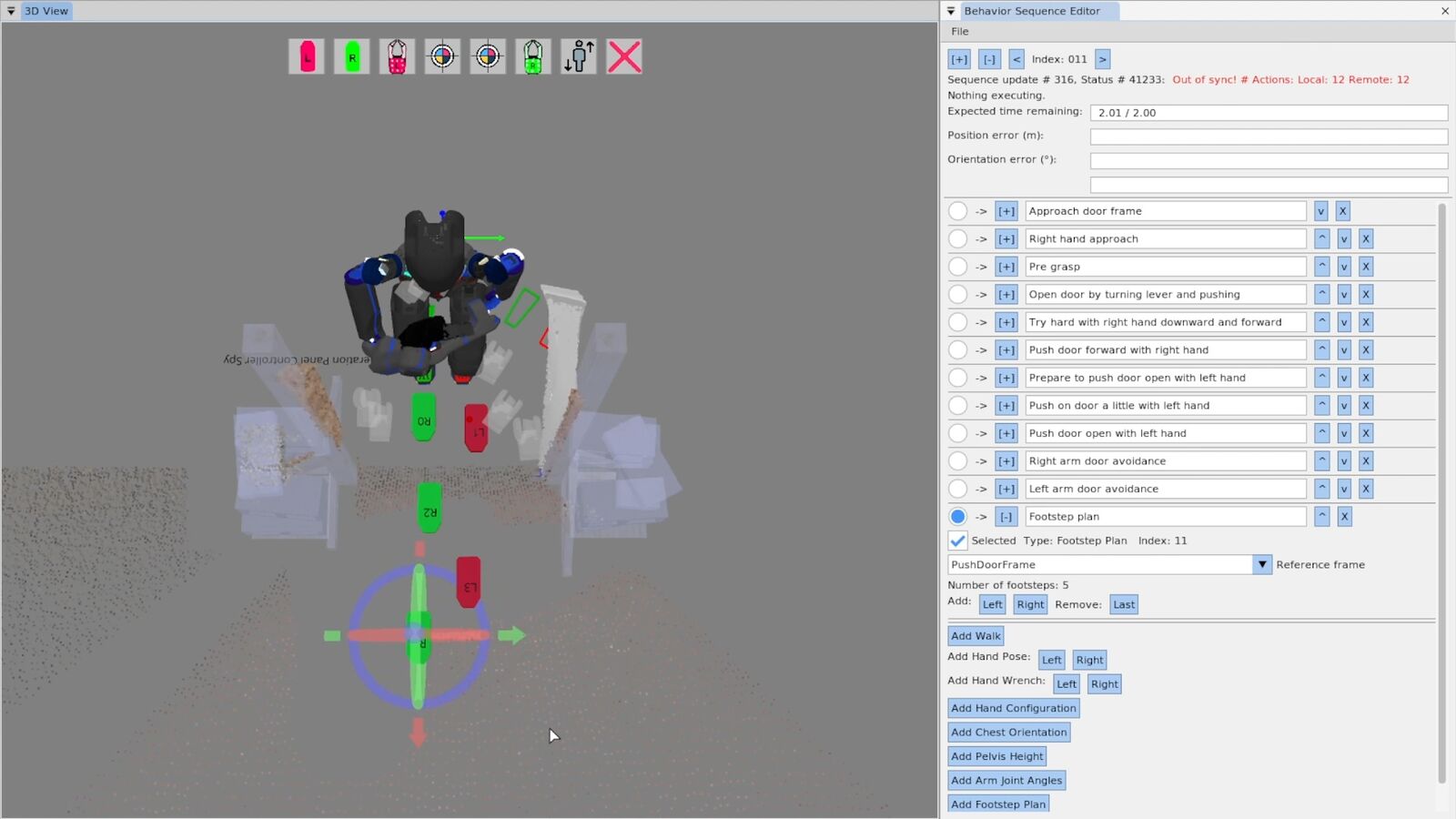}
    \caption{September 1, 2023 screenshot of a new behavior authoring feature.
    The behavior author is able to append footsteps to the walk action footstep plan and tune their poses with a gizmo.
    A video is available at \url{https://youtu.be/FZrVUsMcG7s}.
    }
    \label{fig:2023_footstep_authoring}
\end{figure}

In September 2023 we made enhancements to the walk action as seen in \autoref{fig:2023_footstep_authoring}, allowing the behavior author to interactively edit a multi-step footstep plan.
This was especially useful for authoring door traversal walk-throughs, as a very specific set of footsteps are required to satisfy controller stability limits while avoiding collisions between the robot's shoulders and the door frame.

\subsection{Behavior Node Decomposition}
\begin{figure}[H]
    \centering
    \includegraphics[width=.95\columnwidth]{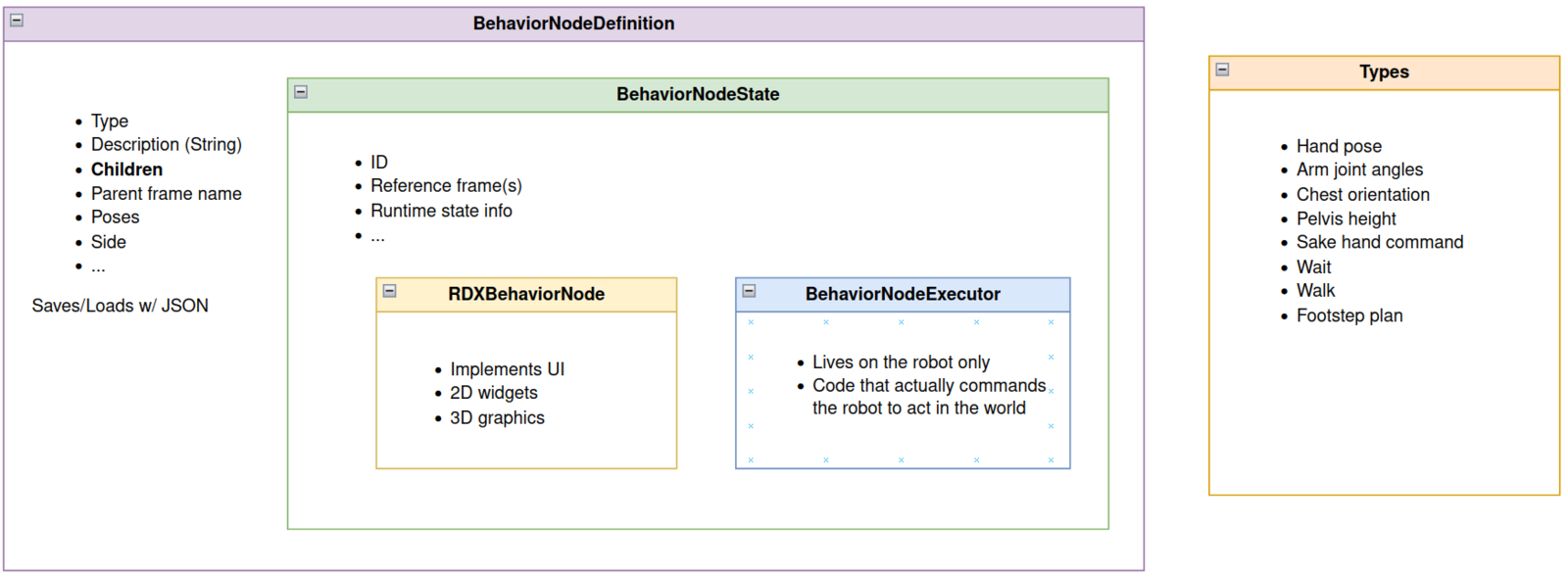}
    \caption{An architecture diagram for the code structure of behavior nodes that was established in September 2023.}
    \label{fig:2023_code_refactor}
\end{figure}

Also in September of 2023, we established the code structure for the behavior nodes as seen in \autoref{fig:2023_code_refactor}.
This structure separated the concerns of a behavior node into four layers: the definition, the state, the executor, and the user interface (RDX) implementation.
Each of the four layers was implemented in a separate code file to keep similar code in similar files.
The definition layer was responsible for defining and serializing the data that defines the node to and from JSON files.
The state layer contains the data that only exists once the node is instantiated and is common between the UI and executor implementations.
It can also contain any data helper functions needed in the UI and the executor.
The executor layer is the instantiated type of the node on the robot.
It is the only layer that commands the robot directly through its ROS 2 API.
It is also the only layer that has access to full resolution and frequency perception data, as it is co-located with the perception sensors.
Finally, the UI layer is for rendering the authoring interface elements in RDX.
The UI layer is also contained in a UI-specific folder that has access to the graphics engine and widget libraries.

\subsection{Runtime-Editable Behavior Tree}
\begin{figure}[H]
    \centering
    \includegraphics[width=.5\columnwidth]{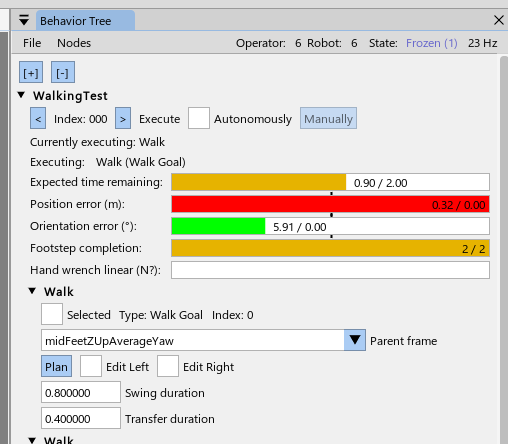}
    \caption{An early version of the behavior tree version of the runtime-editable behavior system.
    Screenshot from November 3, 2023.}
    \label{fig:2023_tree_view}
\end{figure}

In November 2023, we started to convert the sequence into a tree, as seen in \autoref{fig:2023_tree_view}.
Here, the root node ``WalkingTest'' can be seen at the top with an expanded arrow and the root node control widgets.
Two child walk action nodes can be seen below it.

\subsection{Next Action to Execute Pointer}
\begin{figure}[H]
    \centering
    \includegraphics[width=.6\columnwidth]{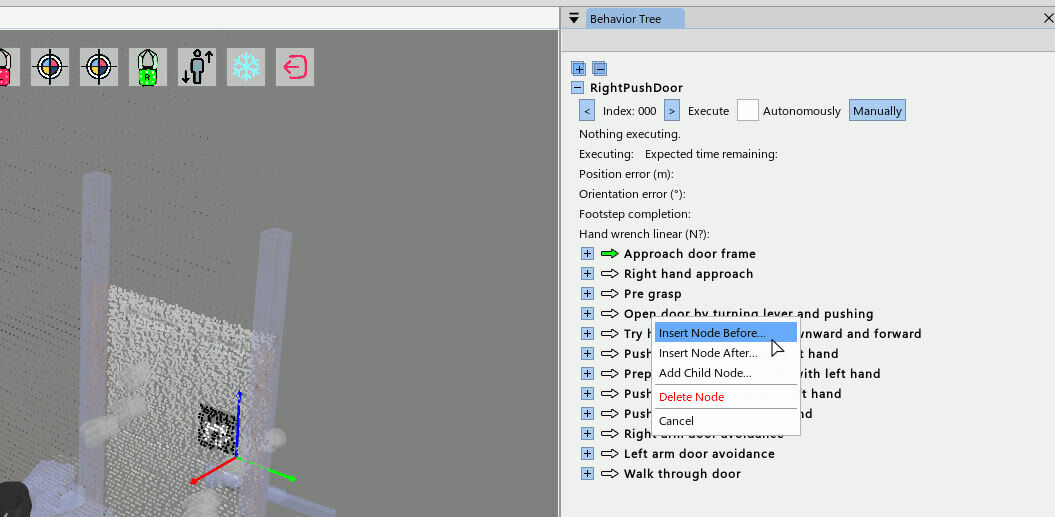}
    \caption{A screenshot from November 14, 2023, showing the introduction of the green arrow to mark the action's next execution index as well as a right-click context menu for inserting nodes.}
    \label{fig:2023_green_arrow_context}
\end{figure}

Further enhancements to the behavior system were made in November of 2023.
As seen in \autoref{fig:2023_green_arrow_context}, a green arrow was introduced to highlight the currently selected next node to execute.
This created a uniquely identifiable marker for the operator to focus on that was distinct from the default ImGui widgets.
Clicking the green arrow is used to set the index.
Later the arrow would blink blue to signal the action was currently executing.

\subsection{Moving and Reordering Nodes}
\begin{figure}[H]
    \centering
    \includegraphics[width=.6\columnwidth]{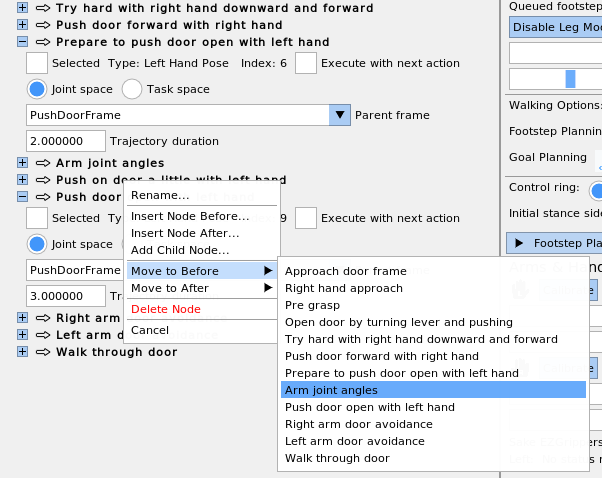}
    \caption{A screenshot from November 16, 2023, showing the introduction of a move nodes context menu, which was used to reorder the nodes and place them under different parents.}
    \label{fig:2023_move_nodes}
\end{figure}

\autoref{fig:2023_move_nodes} shows a context menu feature used to reorder nodes and place them under different parents.
This was especially useful if a node needed to be moved far away.
Since switching to the tree-based view, this was the first time the nodes could be reordered or moved under new parents.

\subsection{Nesting File-Backed Subtrees}
\begin{figure}[H]
    \centering
    \includegraphics[width=.6\columnwidth]{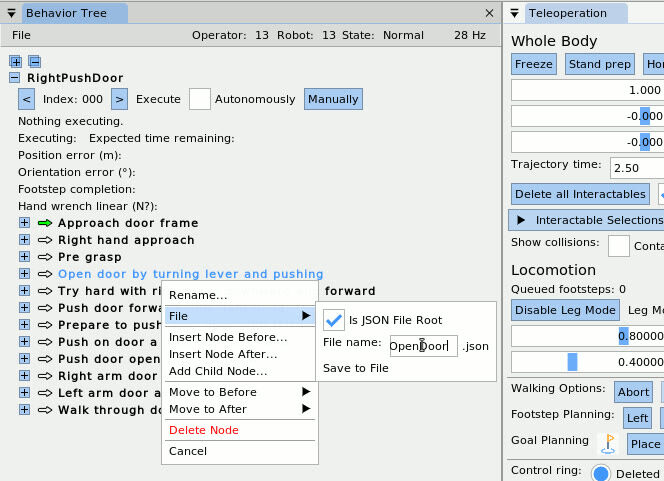}
    \caption{A screenshot from November 16, 2023, showing how nodes could be converted to nested JSON files.
    A gif is available at \url{https://giphy.com/gifs/N1sLeoz3z1zJeJfw8T}.
    }
    \label{fig:2023_nesting_json}
\end{figure}

A feature developed on November 16, 2023, can be seen in \autoref{fig:2023_nesting_json}, which shows how a node can be converted into a nested JSON file via the context menu.
The JSON behavior loader now supported recursively loading behavior trees with nested JSON files.
This supported the separation of skills into their own JSON files and kept larger tree files from getting too large.

\subsection{Screw Primitive Action}
\begin{figure}[H]
    \centering
    \includegraphics[width=.9\columnwidth]{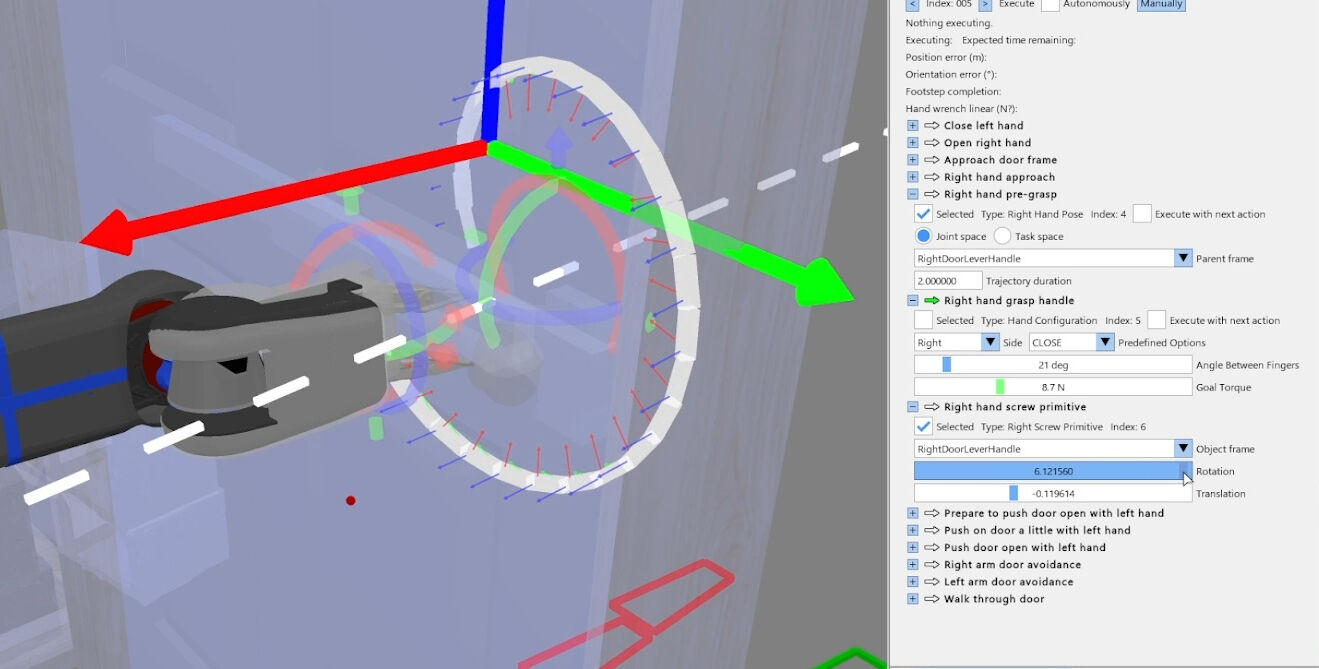}
    \caption{A screenshot from December 7, 2023, showing a new screw primitive action type.
    The dotted line through the center represents the axis to revolve around or through.
    The white helix graphic represents the desired action trajectory.
    On the right, under the ``Right hand screw primitive'' action, sliders for rotation and translation are available to adjust the screw motion.
    A video is available at \url{https://youtu.be/_9qtUpvAAso}.
    }
    \label{fig:2023_screw_primitive}
\end{figure}

In December of 2023, we developed a screw primitive action to open doors, which can be seen in \autoref{fig:2023_screw_primitive}.
This parameterized helical screw trajectory option was inspired by~\cite{Pettinger_2022}.
It allows the operator to define revolving motions useful for turning handles and swinging door panels.
It is defined by a 3D axis (dashed white line), an angle of revolution, and a translation amount to move along the axis.
The trajectory is also grasp-invariant, as it generates the trajectory online from the hand's pose just before execution.

\section{2024 Nadia, Semantic Perception, Fast Behaviors Era}

\subsection{Semantic Perception for Manipulation}
\begin{figure}[H]
    \centering
    \includegraphics[width=.95\columnwidth]{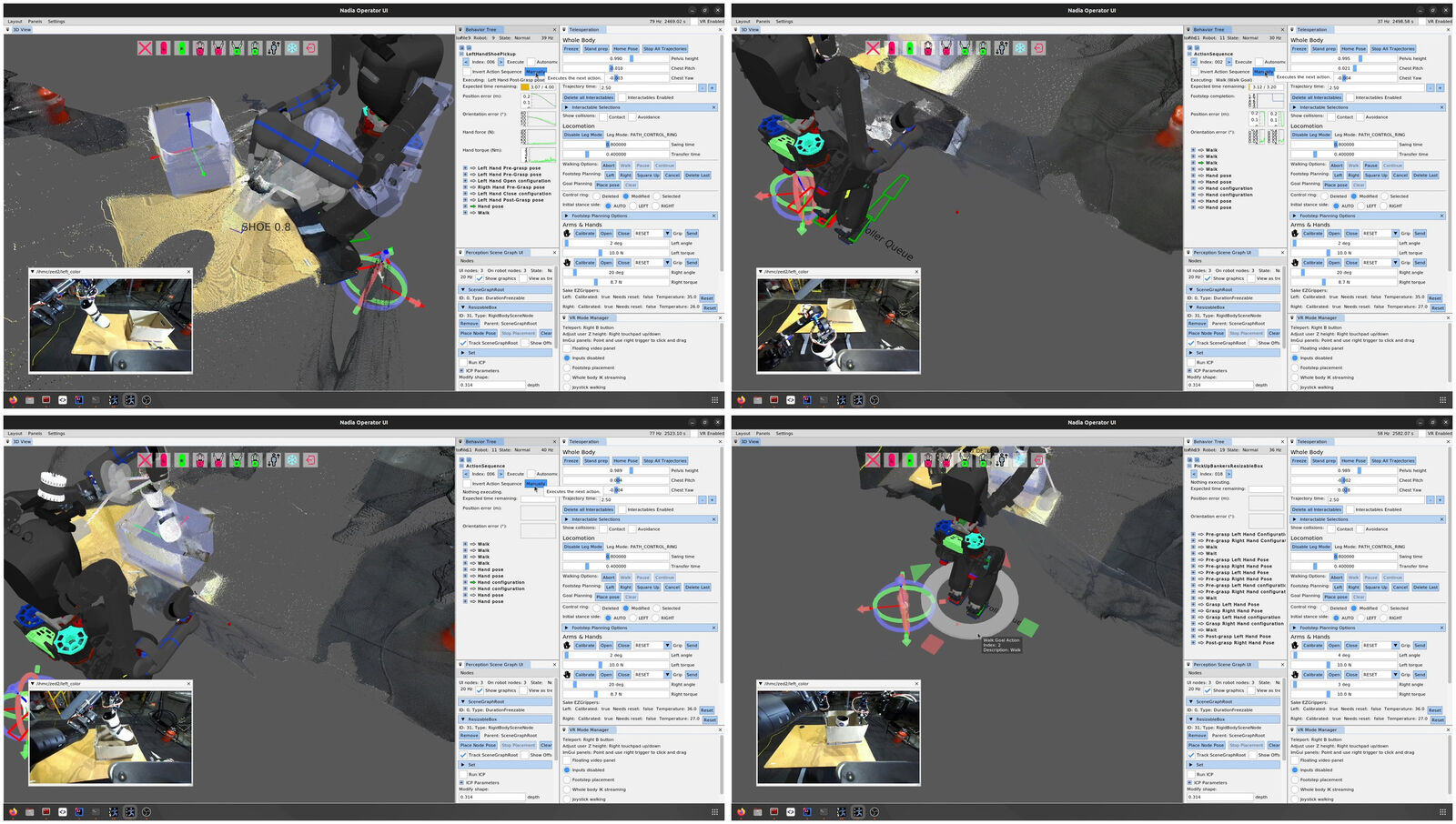}
    \caption{A behavior for picking up a shoe (top left), walking to the side (top right), placing the shoe in the box (bottom left), and picking up the box (bottom right) executed on January 21, 2024.
    A video is available at \url{https://youtu.be/BV9vFMgMc3A}.
    }
    \label{fig:2024_shoe_in_box}
\end{figure}

In January 2024, in an effort to demonstrate advancements in manipulation capability, the behavior system was used in a demo that picks up a shoe off a table, sidesteps, places the shoe in a banker's box, picks up the box, and backs away with it as shown in \autoref{fig:2024_shoe_in_box}.
This behavior was authored by Dexton Anderson and Dhruv Thanki and was a good example of the behavior system being used by someone other than Duncan Calvert or Luigi Penco, who had been the primary users until this point.
The behavior was executed using manually stepping as opposed to the automatic mode and lasted 3 minutes and 9 seconds.
This was also the first integration of a CenterPose~\cite{lin2022icra:centerpose} model for the shoe and the first behavior that did not use an ArUco marker for manipulation of an object.
The behavior was split into three separate JSON files which were executed separately: ``LeftHandShoePickUp.json'', ``ShoeDropOff.json'', and ``PickUpBankersResizableBox.json''.

This behavior was very difficult to get to work due to the unreliability of the components.
The Nadia T-Motor arms and lower body EtherCAT bus would often fault, causing the robot to fall.
CenterPose was very unreliable in detecting the shoe.
There was no onboard compute so a laptop was tethered to the robot to run the perception and behavior processes.
On top of those, the behavior system was still new and buggy.
We also had a new implementation of a scene graph that worked with the behavior system, but separate.
This scene graph had synchronization bugs.

\subsection{Behavior System User Interface}

\begin{figure}[H]
    \centering
    \includegraphics[width=.95\columnwidth]{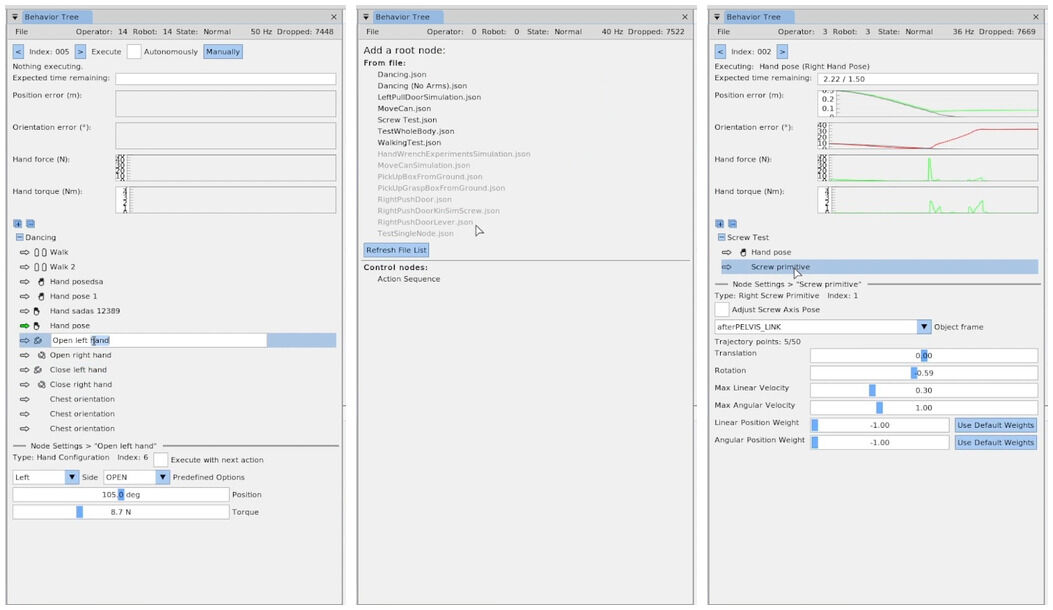}
    \caption{A screenshot from January 30, 2024 showing a refreshed user interface for the behavior system.
    Icons were now present for action types, actions could be renamed with a double-click, the node settings were consolidated to a modal area in the lower portion of the panel (left), the root node selection menu was visually refreshed (center), and plots were now an available option for action progress (right).
    A video is available at \url{https://youtu.be/RQ8N9wGhW7s}.
    }
    \label{fig:2024_refreshed_ui}
\end{figure}

Also in January 2024, a major refresh of the behavior system user interface was completed as shown in \autoref{fig:2024_refreshed_ui}.
The general visual improvements in this version are still present in 2026 at the time of writing.
The major changes included the following.
\begin{enumerate}
    \item Icons were added for the action types to aid in visual navigation.
    \item Actions could be renamed by double clicking them, typing the new name, and hitting enter.
    \item The settings for each node were moved to a modal area at the bottom of the Behavior Tree panel instead of being inlined next to each node.
    This was an important change that simplified the user experience, as only one node should be modified at a time.
    \item The root node selection area was refreshed visually and used selectable underlined text buttons.
    \item Action progress was now plotted at the top as an option instead of the progress bars.
    The user could toggle between these two types of progress information.
\end{enumerate}

\begin{figure}[H]
    \centering
    \includegraphics[width=.95\columnwidth]{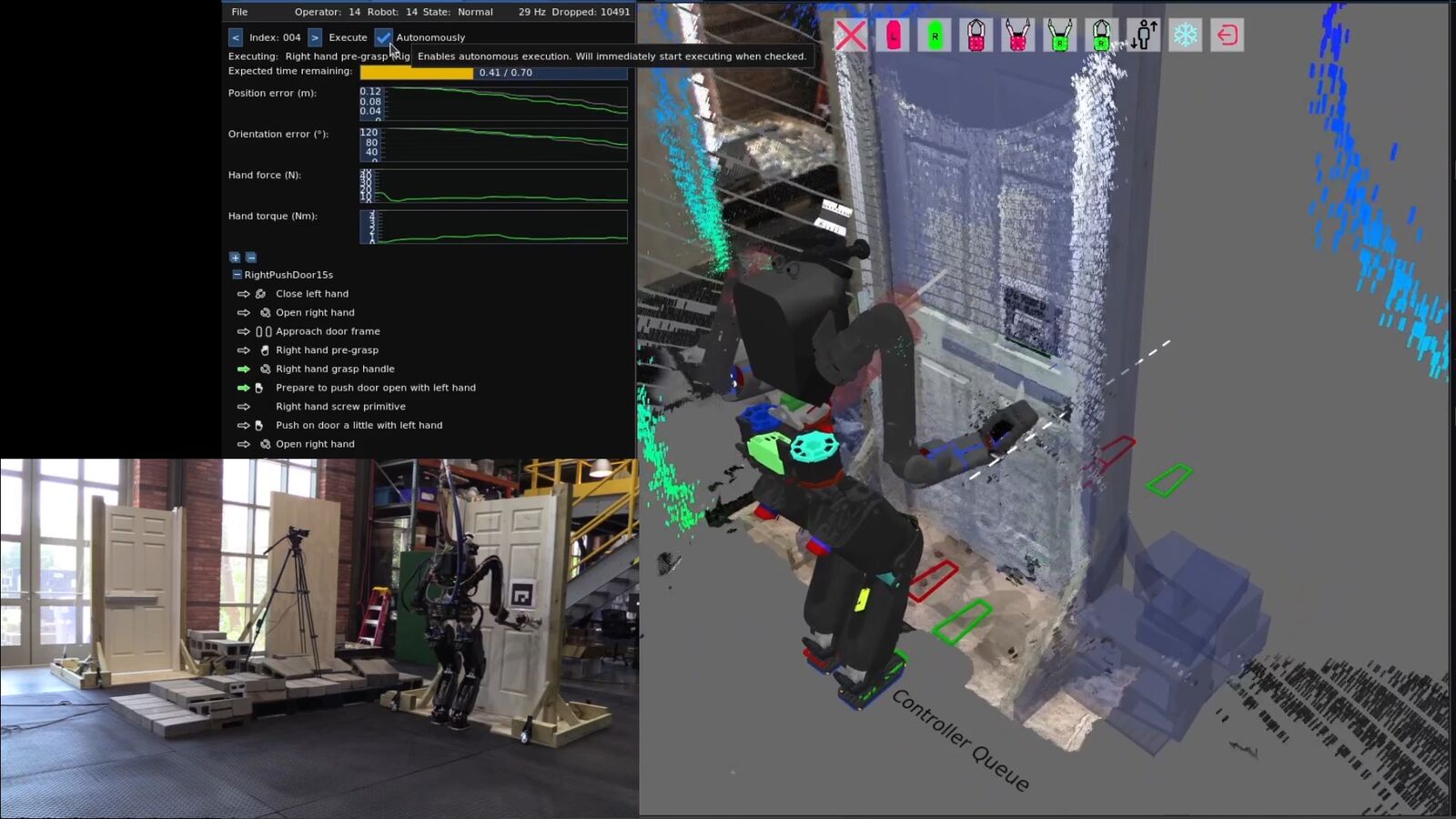}
    \caption{A screenshot of the February 4, 2024, 17-second push door behavior.
    A video is available at \url{https://youtu.be/c929cDimaCY}.
    }
    \label{fig:2024_17s_push_door}
\end{figure}

\subsection{Fast Behaviors}
On February 4, 2024, we achieved the fastest push door behavior to that date on Nadia, executing in 17 seconds, shown in \autoref{fig:2024_17s_push_door}.
By this time, Nadia's upper arms had been upgraded to a cycloid drive based design which enabled faster motions without hardware faults.
It also used a new ``execute with next action'' boolean option for action nodes which was added back in September 2023.
This feature supported executing multiple nodes at once, contributing to the speed of this push door behavior.
It did this by allowing subsequent actions to proceed without waiting for the currently executing action to complete.

\begin{figure}[H]
    \centering
    \includegraphics[width=.5\columnwidth]{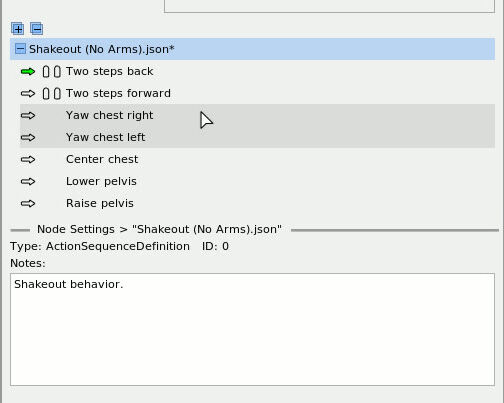}
    \caption{A screenshot from February 28, 2024, showing the introduction of the ``mark modified'' feature, which illustrated to the operator that a node has been modified but not saved, using the asterisk symbol (*).
    A gif is available at \url{https://giphy.com/gifs/kqTj94XEMbo9l6gdDl}.
    }
    \label{fig:2024_mark_modified}
\end{figure}

Also in February 2024, we added a feature that shows an asterisk next to the behavior node names if they have been modified, as shown in \autoref{fig:2024_mark_modified}.
This was an important usability feature in behavior authoring, because accidental behavior modifications were common at this stage in the implementation.
It was, therefore, important to understand when a modification was made so the operator could undo it if necessary.
This feature involved some extra boilerplate in each node's definition implementation but was well worth it for the usability improvement.

\subsection{``Execute After'' Concurrency Mechanism}
\begin{figure}[H]
    \centering
    \includegraphics[width=.8\columnwidth]{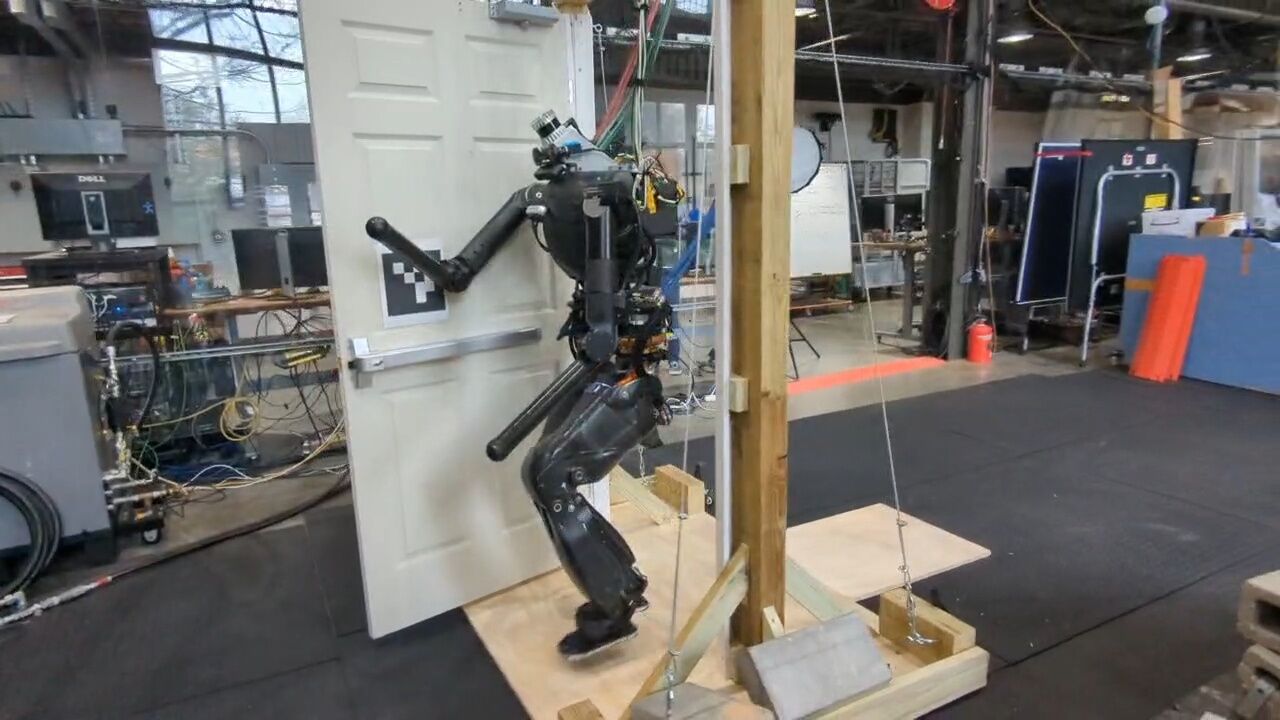}
    \caption{The 14 second continuous walking spring-loaded push bar door traversal on March 15 2024.
    This behavior used the new ``execute after'' concurrency mechanism to achieve scheduled concurrent arm motions while walking.
    A video is available at \url{https://youtu.be/DrR2Ng3ft5Y}.
    }
    \label{fig:2024_14s_push_door}
\end{figure}

\begin{figure}[H]
    \centering
    \includegraphics[width=.8\columnwidth]{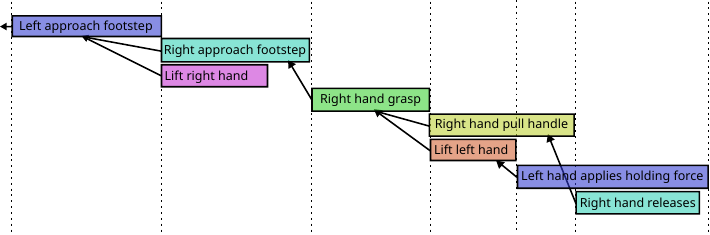}
    \caption{An illustration of the ``execute after'' concurrency mechanism.
    Arrows represent a dependency on action completion.
    This has the effect of scheduling the dependent action later or allowing it to run earlier.}
    \label{fig:execute_after_concurrency}
\end{figure}

In March of 2024, the ``execute with next action'' concurrency feature was replaced with an ``execute after'' node pointer.
This allowed each action to declare a dependency on which action to wait for, as illustrated in \autoref{fig:execute_after_concurrency}.
By setting the ``execute after'' field to something earlier than the previous action, actions could start along with prior actions.
This also allowed for action scheduling using the Wait action in combination with the ``execute after'' field.
For example, arm motions could be scheduled during walking.
This feature enabled our fastest ever door traversal, performed on March 15 2024, as shown in \autoref{fig:2024_14s_push_door}.
In this behavior, Nadia traversed a spring-loaded push door without stopping walking in 14 seconds, using nubs for hands and an ArUco marker for detecting the pose of the door.

\subsection{Bimanual Box Pick and Place}


\begin{figure}[H]
    \centering
    \includegraphics[width=.95\columnwidth]{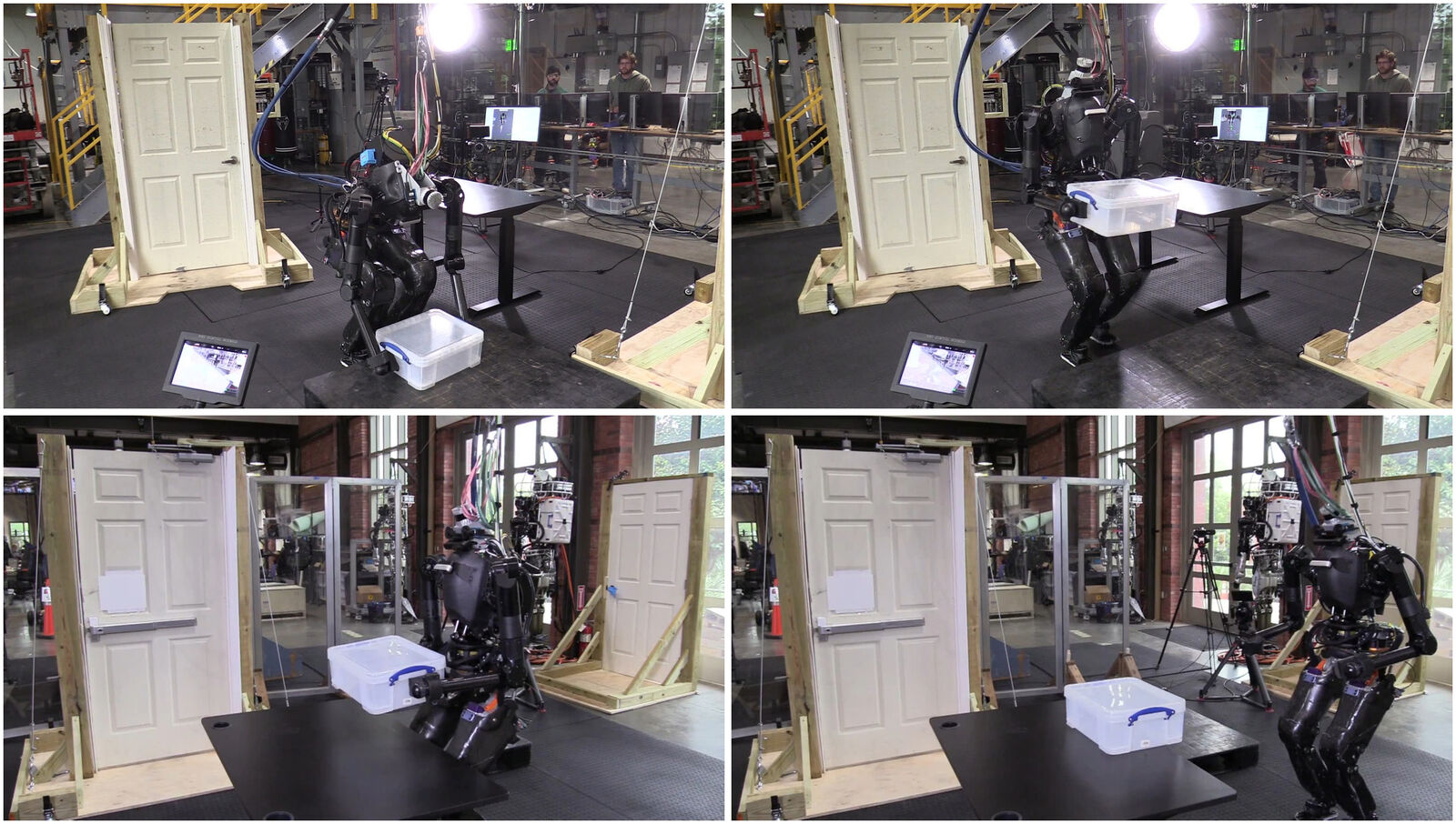}
    \caption{The April 9, 2024 bimanual box pick and place behavior.
    A video is available at \url{https://youtu.be/m0NC2EV4XB4}.
    }
    \label{fig:2024_box_pick_and_place}
\end{figure}

On April 9, 2024, we demonstrated a bimanual box pick and place behavior as shown in \autoref{fig:2024_box_pick_and_place}.
With some padding attached to the nub forearms, Nadia executed a behavior that picked a box up from a pallet on the ground, carried it to a table, set it down, and backed away.

\subsection{Hard-Coded Disturbance Recovery}
\begin{figure}[H]
    \centering
    \includegraphics[width=.95\columnwidth]{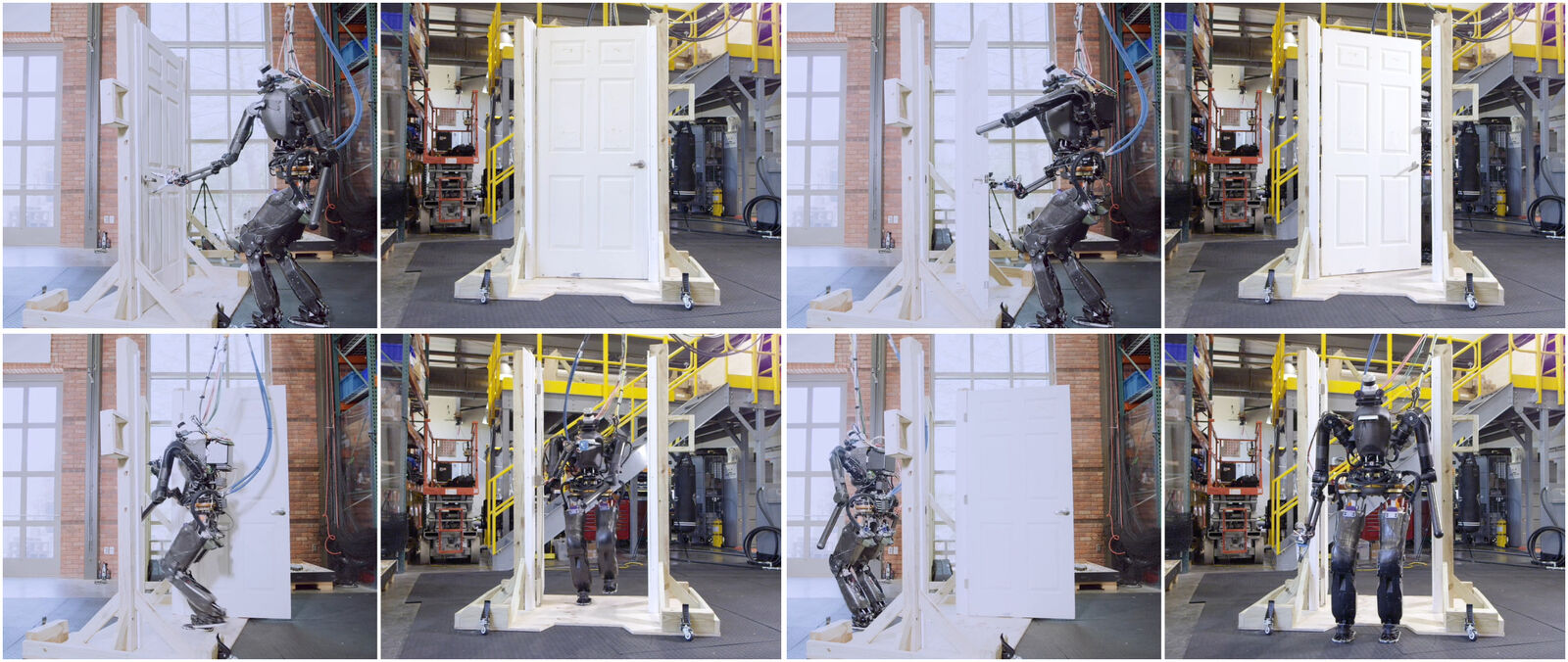}
    \caption{A 19 second pull door behavior executed on April 12, 2024.
    A video is available at \url{https://youtu.be/jSM8mdmOZlc}.
    }
    \label{fig:2024_pull_door_tmotor}
\end{figure}

On April 12, 2024, we demonstrated a pull door behavior in 19 seconds, shown in \autoref{fig:2024_pull_door_tmotor}.
This behavior used a new YOLO model for door opening mechanisms and did not use an ArUco marker.

Earlier, in late March 2024, we had started implementing a heuristic door traversal node as a place to hard-code door-specific logic while continuing to use the runtime editable actions.
The purpose of this was to implement reactive elements like retries on failure before figuring out how to generalize it with fallback nodes.
Pseudocode for the initial pull door lever retry is presented in \autoref{alg:pull_door_screw_primitive_retry}.

\begin{algorithm}[H]
\SetAlgoLined
\caption{Retry door opening if gripper slips off handle and closes all the way}
\label{alg:pull_door_screw_primitive_retry}

\If{$pullDoorScrewPrimitiveIsExecuting \land gripperFingerOpenAngle < 20^\circ$}{
    $executionNextIndex \gets$ index of the wait action before opening the hand and doing pre-grasp\;
}
\end{algorithm}

\begin{figure}[H]
    \centering
    \includegraphics[width=.95\columnwidth]{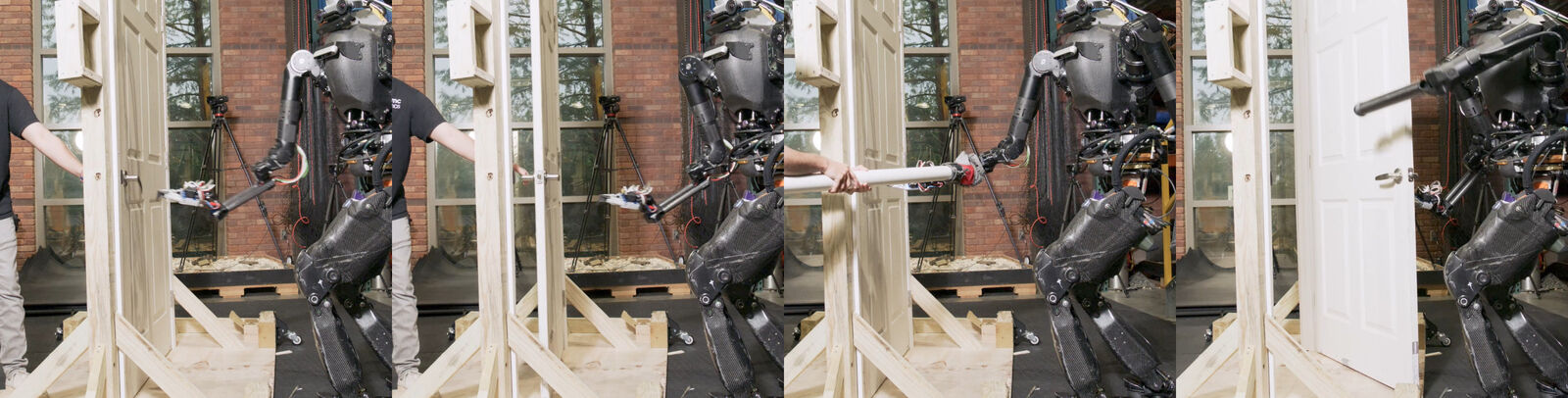}
    \caption{A pull door behavior is disturbed during the opening phase 5 times.
    The robot retries the opening 5 times, ultimately leading to task success.
    This demonstration was executed on April 12, 2024.
    A video is available at \url{https://youtu.be/j_rzh5cAP2E}.
    }
    \label{fig:2024_pull_door_reactive_dhruv}
\end{figure}

On April 12, 2024, we used this strategy to demonstrate a reactive version of the pull door behavior, as shown in \autoref{fig:2024_pull_door_reactive_dhruv}.
A human repeatedly holds the door closed, pulls the door closed, and pushes the robot's hand back as the robot tries to open the door.
The door traversal node detects the action failure, rewinds to the grasp approach action, and tries again until successful.

In \autoref{alg:door_traversal_node_reactive}, we present an improved hard-coded heuristic which was used to detect three types of failures specific to opening a door.
This yielded an improvement in reliability over \autoref{alg:pull_door_screw_primitive_retry}.
At the end of the grasp action, if the hand did not reach the handle, we assume the grasp failed.
At the end of the pull door open action, we check the pose of the door handle and the pose of the robot's hand.
In the case the handle did not move far from its original position, we assume that the unlatch failed.
In the case the handle pose is far from the hand pose, we assume the hand slipped off.
On any type of failure detected by this node, we rewind the action sequence to the grasp approach action, where the robot immediately begins to retry the door opening.
This approach worked well for the human holding the door shut, pulling the door out of the robot's hand, and pushing the robot's hand away from the door handle.
In all three cases, the robot was able to recover and successfully execute the full door traversal without operator intervention.

\begin{algorithm}[H]
\SetAlgoLined
\caption{Door Traversal Node Reactive Algorithm}
\label{alg:door_traversal_node_reactive}

\SetKwProg{Fn}{Function}{:}{end}

\Fn{updateNode()}{
    $a_{name} \gets$ action(name)\;
    $P_{rigidbody} \gets$ rigidbodyPose()\;
    $failureDetected \gets \textbf{false}$\;
    \If{finishedExecuting($a_{graspHandle}$)}{
        \If{$\lVert P_{hand} - P_{handDesired} \rVert > \epsilon_{grasp}$}{
            $failureDetected \gets \textbf{true}$\;
        }
    }
    \If{finishedExecuting($a_{pullDoor}$)}{
        \If{$\lVert P_{handle} - P_{hand} \rVert > \epsilon_{slip}$}{
            $failureDetected \gets \textbf{true}$\;
        }
        \If{$\lVert P_{handle} - P_{handleInitial} \rVert < \epsilon_{unlatch}$}{
            $failureDetected \gets \textbf{true}$\;
        }
    }

    \If{$failureDetected$}{
        setNextActionToExecute($a_{graspApproach}$)\;
    }
}

\end{algorithm}

During this demo, we noted the friction with hard-coded reactive logic.
Iterating from \autoref{alg:pull_door_screw_primitive_retry} to \autoref{alg:door_traversal_node_reactive} took a lot of time and guesswork.
Our metric for evaluation was the reliability of the reactive retry mechanism.
Since this test required the full system to be online and two people to operate, one at the operator computer and one to disturb the robot,
experiment time was scarce.
Having to change code, implement debugging output, redeploy, and restart the behavior to modify the reactive logic consumed a lot of that time.
It was clear that a fallback node, proximity condition, and improved action success criteria would streamline this process in the future.

\subsection{Three Door Traversals in a Row}
\begin{figure}[H]
    \centering
    \includegraphics[width=.95\columnwidth]{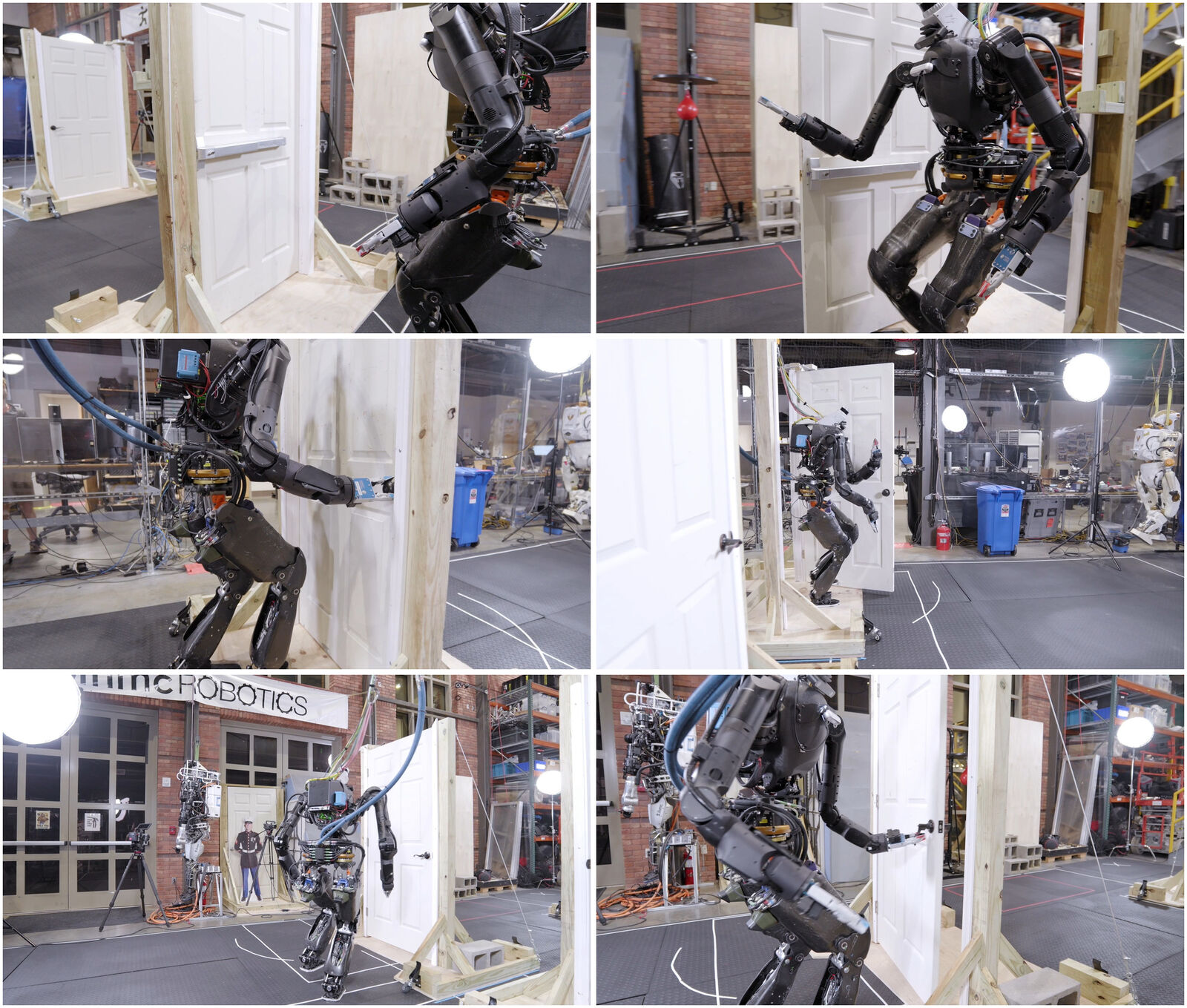}
    \caption{Our July 3, 2024 run where we attempted to traverse three doors in a row with continuous autonomy.
    We traversed the first two in 48 seconds, but ultimately fell after opening the third.
    A video is available at \url{https://youtu.be/cd-lo-l7pPI}.
    }
    \label{fig:2024_3_doors}
\end{figure}

On July 3, 2024, we wanted to show reliability and behavior composition in a building exploration type environment.
We put together a behavior to traverse three doors of different types consecutively in one continuous automatic execution.
We built three lab doors for this: a door with spring closer with a push bar on one side and a pull handle on the other, a knob handle door with a spring closer, and a lever handle door without a spring closer.
We traversed the first two doors in 48 seconds but fell after opening the third door.
This run is shown in \autoref{fig:2024_3_doors}.

There were a couple of failure modes that made achieving this behavior difficult.
Firstly, the EZGripper gripper, having only two fingers and limited grippiness, resulted in slipping off the pull lever handle very often.
Additionally, having only two fingers made getting a solid grasp on the knob handle difficult.
For example, if the two fingers are not positioned at opposite vertices of the circular knob handle, it would slip off to the side.
This was a key moment in realizing that having at least three fingers would make door behaviors more reliable.

Another failure mode, and the one that ultimately prevented us from getting the third door completed, was the result of a combination of hardware and control issues.
The Nadia robot was tethered to a hydraulic pump across the room and the hydraulic hoses were very stiff.
Also, Nadia's hip pitch actuators in their configuration were barely strong enough to lift the leg when the arms were also out in front of the robot, as required for door traversals in avoiding collisions with the door frame.
When the robot was positioned to traverse the third door, the hydraulic hoses put enough extra torque on the robot to make walking through the door un-achievable for the robot and it repeatedly fell.

\subsection{ONR Building Exploration Demo}
\begin{figure}[H]
    \centering
    \includegraphics[width=.95\columnwidth]{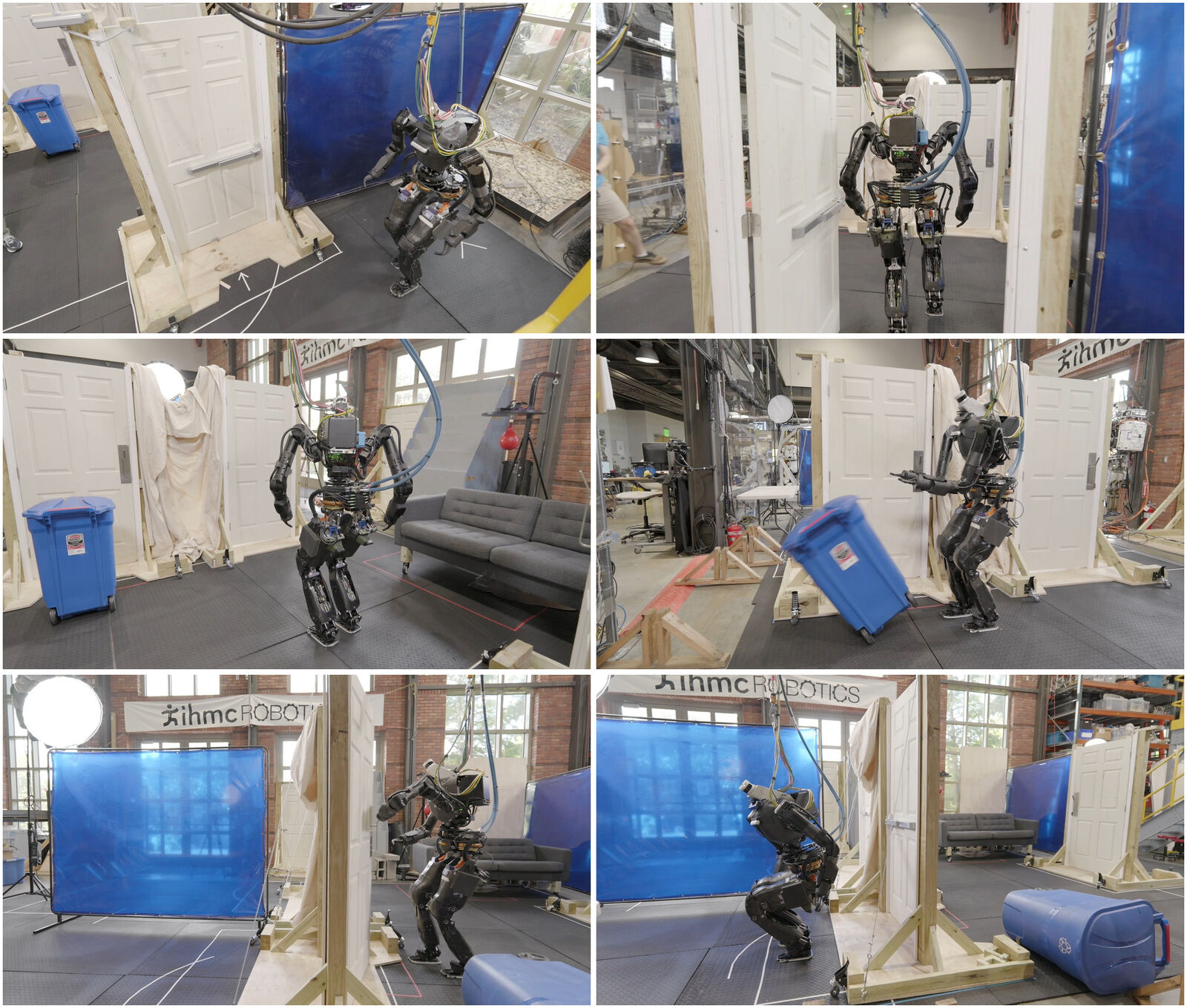}
    \caption{
    The ``ONR Demo'' run where we entered a mock building through a push door (top), searched the room for couches and recycling bins (center left), navigated to the door being blocked by the recycling bin and cleared it out of the way (center right), entered that room (bottom left), and searched for a person (bottom right).
        A video is available at \url{https://youtu.be/GQZbcyediI0}.
    }
    \label{fig:2024_onr_demo_partial}
\end{figure}

\begin{figure}[H]
    \centering
    \includegraphics[width=\columnwidth]{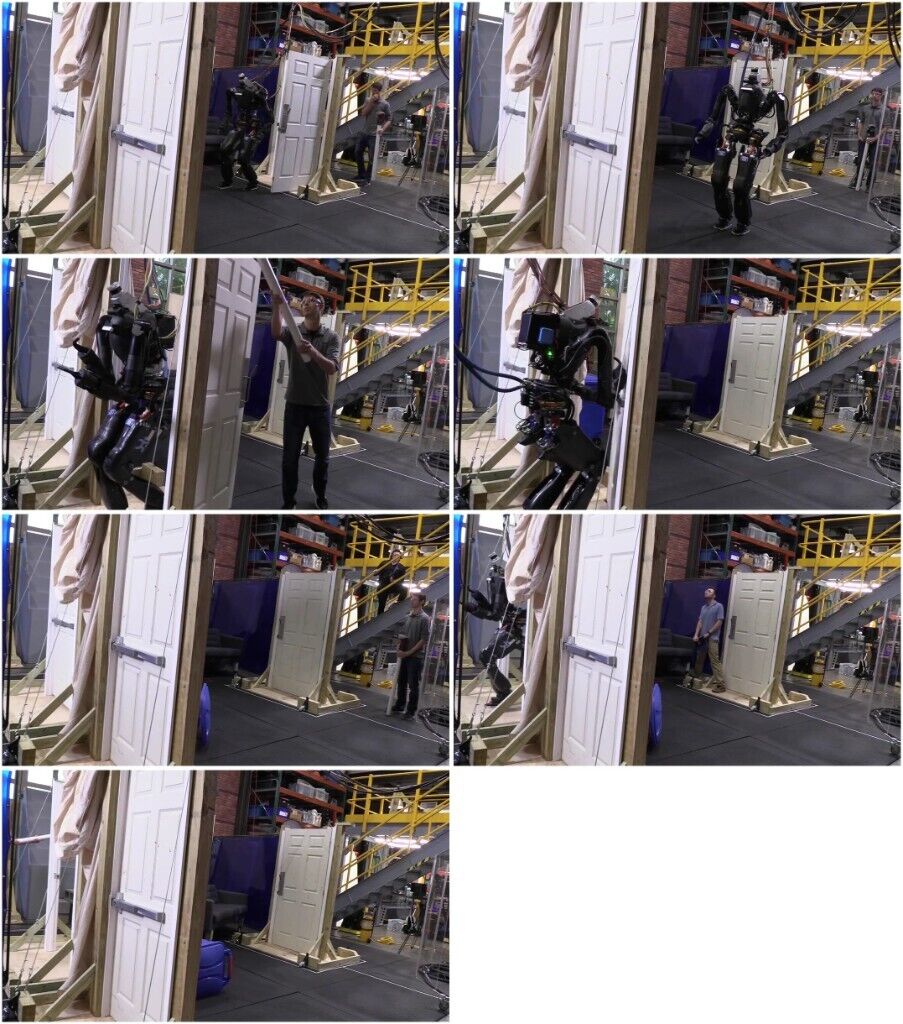}
    \caption{The more complete July~19,~2024 ONR Demo run~2 collage.
    This run entered through a push-bar door, searched multiple rooms, cleared a blocked doorway, traversed additional doors, moved furniture to recover a hidden object, and ended with a salute behavior in 7 minutes and 45 seconds.
    A video is available at \url{https://youtu.be/D0gylAJEdZw}.}
    \label{fig:2024_onr_demo_run_2}
\end{figure}

Later that month, we attempted an extended version of that behavior in which we turned the behavior into a multi-room search.
Dubbed the ``ONR Demo'', this behavior was demonstrated on July 19, 2024.
\autoref{fig:2024_onr_demo_partial} shows a partial test run that entered through a push-bar door, searched the first room, cleared a recycling bin blocking a doorway, and entered a second room.
Later the same day, we also executed a more complete second run in 7 minutes and 45 seconds, shown in \autoref{fig:2024_onr_demo_run_2}.
That run searched all three rooms, cleared a blocked doorway, moved the couch to recover a hidden object, and ended with a salute behavior.

We have printed the structure of the behavior which consists of 178 nodes, generated from the actual JSON in \autoref{fig:onr_demo_tikz} and \autoref{fig:onr_demo_tikz2}.

\begin{figure}[H]
    \centering
    {\def\ONRDEMOTIKZINCLUDE{1}\ifdefined\ONRDEMOTIKZINCLUDE
\else
\documentclass[10pt, landscape]{article}
\usepackage[margin=0.5in]{geometry}
\usepackage[edges]{forest}
\begin{document}
\fi

\ifdefined\TIKZWEBINCLUDE
    \begin{minipage}[t]{\textwidth}
\else
    \begin{minipage}[t]{0.25\textwidth}
\fi
        \begin{forest} baseline,
            for tree={
            folder,
            grow'=0,
            fit=band,
            minimum height=0.35cm,
            inner ysep=0.03cm,
            s sep=-0.0cm,
            font=\sffamily\scriptsize,
            text=black
        }
            [ONRDemo\_BuildingExploration.json
            [START OF DEMO (Wait)]
            [Walk to first door (Footstep plan - dynamic plan)]
            [ONRDemo\_RightPushBarDoorSequence.json (Sequence)
            [Set static for approach (Wait)]
            [Home left (Hand pose - jointspace)]
            [Home right (Hand pose - jointspace)]
            [Approach door (Footstep plan - dynamic plan)]
            [Wait after approach (Wait)]
            [Set static for grasp (Wait)]
            [Traverse door (Footstep plan - 13 pre-planned steps)]
            [Left arm to push panel (Hand pose - jointspace)]
            [Right arm to push bar (Hand pose - jointspace)]
            [Wait to home right arm (Wait)]
            [Home right arm (Hand pose - jointspace)]
            [Wait to push panel (Wait)]
            [Left arm keep panel open (Hand pose - jointspace)]
            [Home left arm (Hand pose - jointspace)]
            ]
            [END FIRST DOOR (Wait)]
            [START SCAN (Wait)]
            [ONRDemo\_ScanArea.json (Sequence)
            [Yaw left (Chest orientation)]
            [Wait before yawing (Wait)]
            [Yaw right (Chest orientation)]
            [Wait again before yawing (Wait)]
            [Home chest yaw (Chest orientation)]
            [Pitch pelvis (Pelvis height orientation)]
            [Wait after pelvis (Wait)]
            [Home pelvis (Pelvis height orientation)]
            [Wait (Wait)]
            ]
            [END SCAN (Wait)]
            [START walk door A (Wait)]
            [Walk towards door A (Footstep plan - dynamic plan)]
            [Pitch pelvis to see door A (Pelvis height orientation)]
            [END walk door A (Wait)]
            [-------------------------------------- (Wait)]
            [START walk door B (Wait)]
            [Walk towards door B (Footstep plan - dynamic plan)]
            [Hoist maneuver turn to face door (Footstep plan - dynamic plan)]
            [Pitch pelvis to see door B (Pelvis height orientation)]
            [END walk door B (Wait)]
            [START turn door A (Wait)]
            [Turn in place door A back (Footstep plan - dynamic plan)]
            [Turn to face door (Footstep plan - dynamic plan)]
            [END turn door A (Wait)]
            [START turn door B (Wait)]
            [Start turning (Footstep plan - dynamic plan)]
            [turn more (Footstep plan - dynamic plan)]
            [END turn door B (Wait)]
            [GET closer door A (Wait)]
            [Approach door A (Footstep plan - dynamic plan)]
            [END get closer door A (Wait)]
            [GET closer door B (Wait)]
            [Approach door B (Footstep plan - dynamic plan)]
            [END get closer door B (Wait)]
            ]
        \end{forest}
    \end{minipage}%
\ifdefined\TIKZWEBINCLUDE
    \par\vspace{1em}
    \begin{minipage}[t]{\textwidth}
\else
    \hspace{4cm}
    \begin{minipage}[t]{0.49\textwidth}
\fi
        \begin{forest} baseline,
            for tree={
            folder,
            grow'=0,
            fit=band,
            minimum height=0.35cm,
            inner ysep=0.03cm,
            s sep=-0.0cm,
            font=\sffamily\scriptsize,
            text=black
        }
            [ONRDemo\_BuildingExploration.json (cont.)
            [START PULL DOOR (Wait)]
            [Wait a bit (Wait)]
            [Wait more (Wait)]
            [Pelvis up (Pelvis height orientation)]
            [Wait (Wait)]
            [Disable (Wait)]
            [ONRDemo\_RightPullDoorHandleSequence.json (Sequence)
            [Set static for approach PULL (Wait)]
            [Home left arm (Hand pose - jointspace)]
            [Home right arm (Hand pose - jointspace)]
            [Approach (Footstep plan - dynamic plan)]
            [Wait after approach (Wait)]
            [Grasping stance (Footstep plan - 2 pre-planned steps)]
            [Left arm up and ready (Hand pose - jointspace)]
            [Wait after step (Wait)]
            [Set static for grasp PULL (Wait)]
            [Yaw chest (Chest orientation)]
            [Left hand pre-grasp (Hand pose - taskspace)]
            [Left hand engage (Hand pose - taskspace)]
            [Wait a bit (Wait)]
            [Pull door screw primitive (Screw primitive)]
            [Prepare right hand (Hand pose - jointspace)]
            [CHECK POINT Unlatched door (Wait)]
            [Home chest (Chest orientation)]
            [Right hand block panel (Hand pose - jointspace)]
            [Left hand disengage (Hand pose - jointspace)]
            [Left hand along body away from panel (Hand pose - jointspace)]
            [CHECK POINT Door secured with right arm (Wait)]
            [Chest yaw to fully open panel (Chest orientation)]
            [Sneak left hand inside panel (Hand pose - jointspace)]
            [CHECK POINT Door fully open (Wait)]
            [Left arm along panel (Hand pose - jointspace)]
            [Chest orientation (Chest orientation)]
            [Right arm door avoidance (Hand pose - jointspace)]
            [Left arm against panel (Hand pose - jointspace)]
            [Tuck up left arm (Hand pose - jointspace)]
            [Traverse door (Footstep plan - 11 pre-planned steps)]
            [Wait to home arms (Wait)]
            [Right hand pose home (Hand pose - jointspace)]
            [Left hand pose home (Hand pose - jointspace)]
]
            [Enable (Wait)]
            [END PULL DOOR (Wait)]
            [START TRASHCAN (Wait)]
            [Wait 1s (Wait)]
            [Disable1 (Wait)]
            ]
        \end{forest}
    \end{minipage}%

    \ifdefined\ONRDEMOTIKZINCLUDE
    \else
\end{document}
\fi}
    \caption{The ONR Demo behavior tree (Part 1).}
    \label{fig:onr_demo_tikz}
\end{figure}
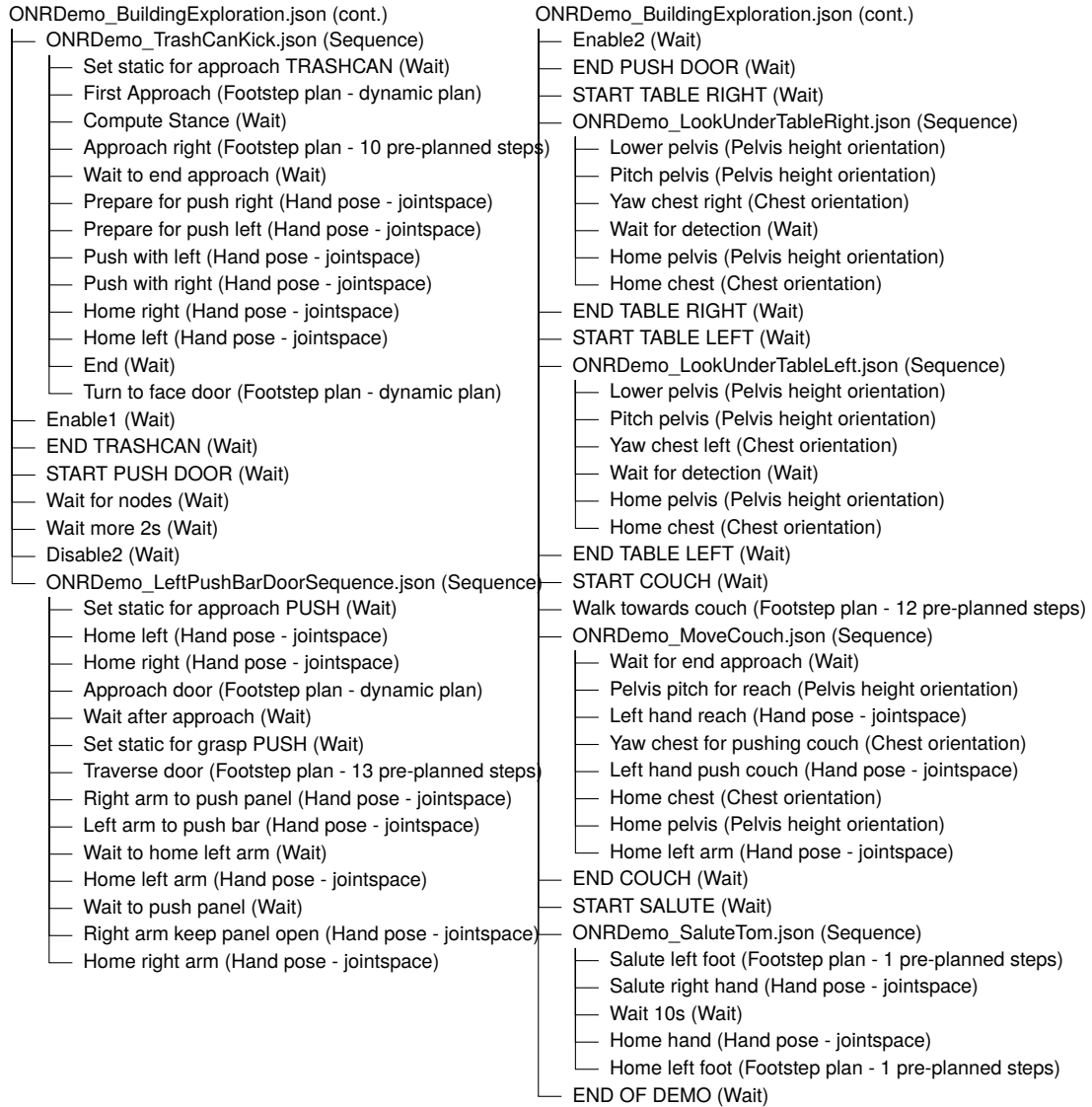
\begin{figure}[H]
    \centering
    {\def\ONRDEMOTIKZINCLUDE{1}\ifdefined\ONRDEMOTIKZINCLUDE
\else
\documentclass[10pt, landscape]{article}
\usepackage[margin=0.5in]{geometry}
\usepackage[edges]{forest}
\begin{document}
    \fi

    \begin{center}
        \begin{minipage}[t]{0.30\textwidth}
            \begin{forest} baseline,
                for tree={
                folder,
                grow'=0,
                fit=band,
                minimum height=0.35cm,
                inner ysep=0.03cm,
                s sep=-0.0cm,
                font=\sffamily\scriptsize,
                text=black
            }
                [ONRDemo\_BuildingExploration.json (cont.)
                [ONRDemo\_TrashCanKick.json (Sequence)
                [Set static for approach TRASHCAN (Wait)]
                [First Approach (Footstep plan - dynamic plan)]
                [Compute Stance (Wait)]
                [Approach right (Footstep plan - 10 pre-planned steps)]
                [Wait to end approach (Wait)]
                [Prepare for push right (Hand pose - jointspace)]
                [Prepare for push left (Hand pose - jointspace)]
                [Push with left (Hand pose - jointspace)]
                [Push with right (Hand pose - jointspace)]
                [Home right (Hand pose - jointspace)]
                [Home left (Hand pose - jointspace)]
                [End (Wait)]
                [Turn to face door (Footstep plan - dynamic plan)]
                ]
                [Enable1 (Wait)]
                [END TRASHCAN (Wait)]
                [START PUSH DOOR (Wait)]
                [Wait for nodes (Wait)]
                [Wait more 2s (Wait)]
                [Disable2 (Wait)]
                [ONRDemo\_LeftPushBarDoorSequence.json (Sequence)
                [Set static for approach PUSH (Wait)]
                [Home left (Hand pose - jointspace)]
                [Home right (Hand pose - jointspace)]
                [Approach door (Footstep plan - dynamic plan)]
                [Wait after approach (Wait)]
                [Set static for grasp PUSH (Wait)]
                [Traverse door (Footstep plan - 13 pre-planned steps)]
                [Right arm to push panel (Hand pose - jointspace)]
                [Left arm to push bar (Hand pose - jointspace)]
                [Wait to home left arm (Wait)]
                [Home left arm (Hand pose - jointspace)]
                [Wait to push panel (Wait)]
                [Right arm keep panel open (Hand pose - jointspace)]
                [Home right arm (Hand pose - jointspace)]
                ]
                ]
            \end{forest}
        \end{minipage}%
        \hspace{2cm}
        \begin{minipage}[t]{0.48\textwidth}
            \begin{forest} baseline,
                for tree={
                folder,
                grow'=0,
                fit=band,
                minimum height=0.35cm,
                inner ysep=0.03cm,
                s sep=-0.0cm,
                font=\sffamily\scriptsize,
                text=black
            }
                [ONRDemo\_BuildingExploration.json (cont.)
                [Enable2 (Wait)]
                [END PUSH DOOR (Wait)]
                [START TABLE RIGHT (Wait)]
                [ONRDemo\_LookUnderTableRight.json (Sequence)
                [Lower pelvis (Pelvis height orientation)]
                [Pitch pelvis (Pelvis height orientation)]
                [Yaw chest right (Chest orientation)]
                [Wait for detection (Wait)]
                [Home pelvis (Pelvis height orientation)]
                [Home chest (Chest orientation)]
]
                [END TABLE RIGHT (Wait)]
                [START TABLE LEFT (Wait)]
                [ONRDemo\_LookUnderTableLeft.json (Sequence)
                [Lower pelvis (Pelvis height orientation)]
                [Pitch pelvis (Pelvis height orientation)]
                [Yaw chest left (Chest orientation)]
                [Wait for detection (Wait)]
                [Home pelvis (Pelvis height orientation)]
                [Home chest (Chest orientation)]
]
                [END TABLE LEFT (Wait)]
                [START COUCH (Wait)]
                [Walk towards couch (Footstep plan - 12 pre-planned steps)]
                [ONRDemo\_MoveCouch.json (Sequence)
                [Wait for end approach (Wait)]
                [Pelvis pitch for reach (Pelvis height orientation)]
                [Left hand reach (Hand pose - jointspace)]
                [Yaw chest for pushing couch (Chest orientation)]
                [Left hand push couch (Hand pose - jointspace)]
                [Home chest (Chest orientation)]
                [Home pelvis (Pelvis height orientation)]
                [Home left arm (Hand pose - jointspace)]
]
                [END COUCH (Wait)]
                [START SALUTE (Wait)]
                [ONRDemo\_SaluteTom.json (Sequence)
                [Salute left foot (Footstep plan - 1 pre-planned steps)]
                [Salute right hand (Hand pose - jointspace)]
                [Wait 10s (Wait)]
                [Home hand (Hand pose - jointspace)]
                [Home left foot (Footstep plan - 1 pre-planned steps)]
]
                [END OF DEMO (Wait)]
                ]
            \end{forest}
        \end{minipage}
    \end{center}

    \ifdefined\ONRDEMOTIKZINCLUDE
    \else
\end{document}
\fi}
    \caption{The ONR Demo behavior tree (Part 2).}
    \label{fig:onr_demo_tikz2}
\end{figure}

On the day of the ONR Demo, we were able to accomplish each task without a failure.
However, in the run-up to the demo, we noted an important property of behavior composition.
Our full demo was composed of seven main loco-manipulation behaviors:
\begin{enumerate}
    \item Traverse a push bar door.
    \item Move the recycling bin out of the way.
    \item Traverse a push knob handle door.
    \item Traverse a pull knob handle door.
    \item Traverse a push lever handle door.
    \item Traverse a pull lever handle door.
    \item Move the couch.
\end{enumerate}

Our observation was that even if each behavior had a success rate of 90\% (which they didn't), the full demo reliability is calculated as
\[
    0.9^7 = 48.7\%
\]
which is an alarmingly low (less than 50\%) full-demo success rate even when the sub-behaviors have achieved very good reliability.
On the day of the final demo, where we were ultimately successful, our success likely hinged on sufficiently configuring and testing the initial conditions such that the chance of success was much higher.

\section{2024-2026 H1-2, 5 Fingers, Manipulation Era}

\subsection{Large Language Model Integration}
\begin{figure}[H]
    \centering
    \includegraphics[width=.95\columnwidth]{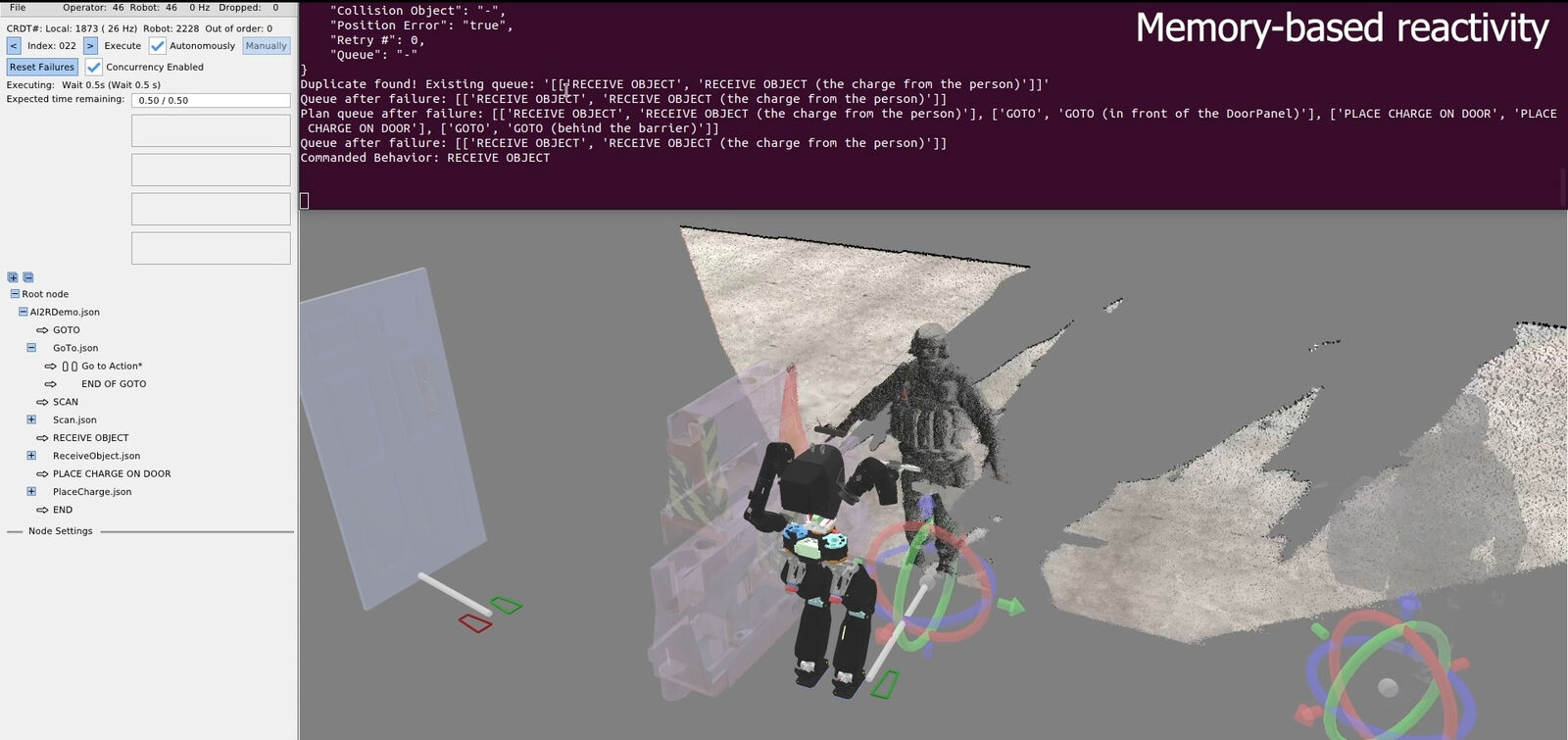}
    \caption{
        A demo using the ``AI to Robot'' (AI2R) node to work with an external LLM process to coordinate behavior using reasoning in natural language.
    }
    \label{fig:2025_ai2r}
\end{figure}
In late 2024, we started looking into integrating our behavior tree system with external AI systems, such as Large Language Models (LLMs) and Vision Language Models (VLMs).
For this, we created another heuristic control node called the AI to Robot (AI2R) node.
In a similar spirit to our door traversal node, this node would contain hard-coded interfacing to external AI processes.
The goal was to use LLMs and VLMs to generate behavior and make decisions.
By June 2025 we had a working demo in simulation, as shown in \autoref{fig:2025_ai2r}.
The goal of this demo was to receive an object from a person, stick it to the door, and retreat.

In this demo, the robot scans the area and reports the detected scene objects to the LLM, including people.
The LLM is given a list of objects and poses, works out which person has the object, then tells the robot to go to the person with the object.
There is also failure handling.
If the robot goes to the wrong person, the one who is not holding the object, the LLM will detect this and an operator can, in natural language, correct the behavior to go to the other person.
The system also supports collision avoidance.
In the case a person is blocking the robot's path, the behavior will pause and the operator can, in natural language, ask the robot to go behind the last person.
This hybrid behavior system suggests that it is possible for LLMs to work in place of some classical components.
However, we do not present any results from these experiments in this thesis.

\subsection{Attempt to Integrate VLAs}
For a seven-month period from March to October 2025, we tried to get Vision Language Action Models (VLAs) working for task sub-components like the door opening, but ultimately failed to reproduce the reported results and diverted our efforts back to our classical route.
Our conclusion was that at this time, VLAs were not an ``off-the-shelf'' solution and still required expertise and possibly even an ``artistic'' talent to get working well.
However, it may have also been that we had a bug somewhere.

In October of 2025, we did some restructuring of the internals of the behavior system, making the root node reference final and accessible directly to all nodes.
This allowed more direct access to behavior-wide components.
For example, for all executor nodes, the controller ROS 2 node, the behavior scene, and the root node were all made final and accessible directly without needing to navigate boilerplate.
This made the code a little nicer to deal with but didn't have any meaningful effects on operator usability or robot performance.

\subsection{Runtime-Editable Logic}
\begin{figure}[H]
    \centering
    \includegraphics[width=.95\columnwidth]{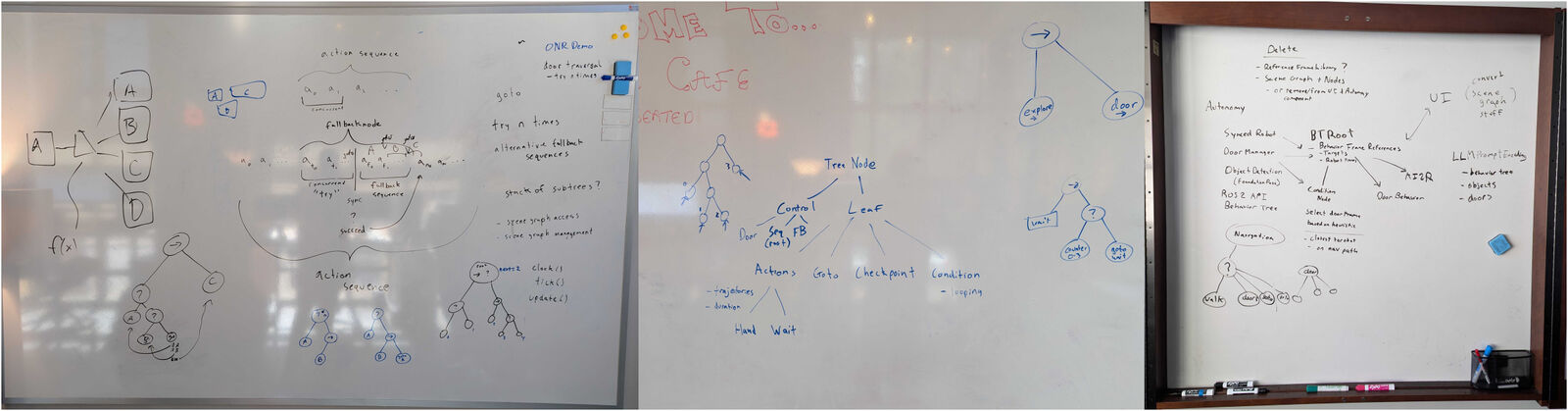}
    \caption{
        Whiteboards from late 2024 and early 2025 behavior system design meetings.
        Left: ``Design of the fallback node'' from December 6, 2024, Center: ``A tree with fallback, condition, and goto nodes'' from January 31, 2025, and Right: ``A design diagram showing how the scene graph would be deleted and a simpler `behavior frame references' component would be built'' from March 3, 2025.
    }
    \label{fig:2025_whiteboards}
\end{figure}

In late 2024 and early 2025 we had a series of design meetings focused on two major areas.
The first was how to extract the reactive logic from the door traversal node into a runtime-editable version that would extend to other loco-manipulation behaviors.
The second was how to resolve reliability and code architecture issues we were facing between the behavior tree and the scene graph.
Whiteboards from these meetings are shown in \autoref{fig:2025_whiteboards}.

We had been looking to Behavior Trees (BTs)~\cite{2022_iovino_behavior_trees,2018_colledanchise} in the literature for some time, and had used an implementation of them in the 2021 building exploration demo on Atlas, as we showed in \autoref{fig:2021_building_exploration_demo}.
However, that 2021 implementation did not include runtime-editable behaviors.
We wanted to figure out how to support behavior tree control nodes with our novel implementation of runtime-editable concurrent-capable action sequences.
We decided to look for small, incremental modifications that made the biggest impact on reactivity.

In the above design meetings, we decided that adding a fallback node, a condition node, and a goto node would take us a long way without having to do a more significant redesign.
Our idea was that our existing design of ``everything is a single sequence'' would still work and that the sequence implementation would monitor the fallback node, condition node, and goto node children and manage the index of the next action to execute accordingly.

\begin{figure}[H]
    \centering
    \small
    \resizebox{\columnwidth}{!}{\begin{tikzpicture}[
    node distance=1.4cm and 1.6cm,
    every node/.style={font=\sffamily},
    btnode/.style={
        draw,
        minimum width=1.5cm,
        minimum height=1.0cm,
        align=center,
        fill=white,
        thick
    },
    actionnode/.style={
        btnode,
        minimum width=2.2cm
    },
    condition/.style={
        ellipse,
        draw,
        thick,
        fill=white,
        align=center,
        inner xsep=6pt,
        inner ysep=3pt,
        minimum width=2.4cm
    },
    flow/.style={
        ->,
        thick
    },
    nextexec/.style={
        ->,
        very thick,
        blue!60!black
    },
    gotoflow/.style={
        -{Stealth[length=2.8mm, width=2.2mm]},
        thick,
        dashed,
        draw=black!65,
        line cap=round
    },
    succeedflow/.style={
        -{Stealth[length=2.8mm, width=2.2mm]},
        very thick,
        green!60!black,
        line cap=round
    },
    region/.style={
        draw,
        thick,
        rounded corners=3pt,
        inner sep=8pt
    },
    tryregion/.style={
        region,
        draw=blue!55!black,
        fill=blue!6
    },
    catchregion/.style={
        region,
        draw=orange!70!black,
        fill=orange!6
    },
    reglabel/.style={
        font=\scriptsize,
        align=center
    }
]

\newcommand{\btrcrossedzero}{%
    \tikz[baseline=-0.55ex, scale=0.9]{
        \node[inner sep=0pt, font=\sffamily] at (0,0) {0};
        \draw[line width=0.8pt] (-0.18,-0.14) -- (0.18,0.14);
    }%
}

\node[btnode] (root) {\btrcrossedzero};
\node[btnode, below=1.6cm of root] (seq) {$\rightarrow$};

\node[actionnode, below left=2.0cm and 2.6cm of seq] (actionA) {Action A};
\node[btnode, below right=2.0cm and 0.8cm of seq] (fb) {?};
\node[actionnode, below right=2.0cm and 5.8cm of seq] (actionB) {Action B};

\node[actionnode, below=2.0cm of fb, xshift=0.8cm] (catchAct) {Catch action};
\node[actionnode, right=0.55cm of catchAct] (gotoAct) {Goto action};
\node[condition, below=2.0cm of fb, xshift=-3.4cm, yshift=-0.06cm] (tryCond) {Try condition};

\begin{scope}[on background layer]
    \node[catchregion, fit=(catchAct)(gotoAct)] (catchBox) {};

    \coordinate (tryNW) at ([xshift=-8pt]tryCond.west |- catchBox.north);
    \coordinate (trySE) at ([xshift=8pt]tryCond.east |- catchBox.south);
    \node[tryregion, fit=(tryNW)(trySE), inner sep=0pt, label={[reglabel, yshift=2pt]above:try}] (tryBox) {};
\end{scope}

\draw[flow] (root) -- (seq);
\draw[flow] (seq) -- (actionA);
\draw[flow] (seq) -- (fb);
\draw[flow] (seq) -- (actionB);
\draw[flow] (fb) -- (tryCond);
\draw[flow] (fb) -- (catchAct);
\draw[flow] (fb) -- (gotoAct);

\draw[nextexec] ([xshift=-1.1cm]actionA.west) -- (actionA.west);

\coordinate (succeedTarget) at (actionB.west);
\coordinate (succeedIn) at ([xshift=-0.75cm]succeedTarget);
\draw[succeedflow]
    (tryBox.east)
    .. controls +(0.9cm, 0.45cm) and (succeedIn) ..
    (succeedTarget)
    node[pos=0.5, font=\scriptsize, fill=white, inner sep=1pt, sloped, above]
        {succeed};

\coordinate (gotoTarget) at (actionA.east);
\coordinate (gotoIn) at ([xshift=0.75cm]gotoTarget);
\draw[gotoflow]
    (gotoAct.north)
    .. controls +(-0.35cm, 0.65cm) and (gotoIn) ..
    (gotoTarget)
    node[pos=0.5, font=\scriptsize, fill=white, inner sep=1pt, sloped, above]
        {goto};

\node[reglabel, fill=white, inner sep=2pt, anchor=south]
    at ([yshift=2pt]catchBox.north) {catch sequence};

\end{tikzpicture}}
    \caption[Fallback and goto node operation within the action sequence.]{%
        An illustration of fallback and goto node operation within the depth-first action sequence.
        A blue arrow represents the next action to execute, which proceeds among leaves in depth-first order.
        The fallback node's first concurrent sequence is the try (here just a single condition node); remaining children form the catch sequence.
        If the try fails, the catch sequence executes. If it succeeds, the catch is skipped.
        Goto nodes specify the action to execute next, enabling retry loops or strategy changes.}
    \label{fig:fallback_goto_mechanism}
\end{figure}
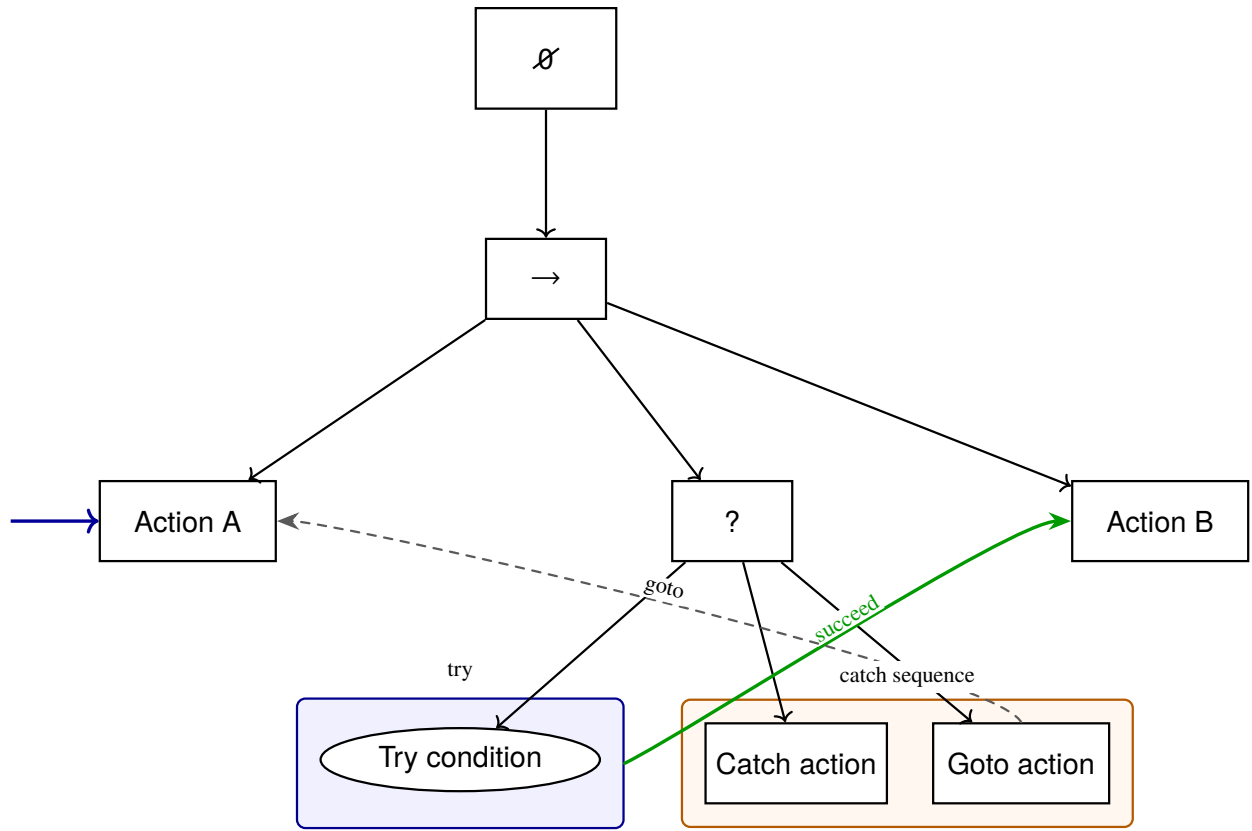

In December 2024, we added a fallback node that supported concurrent actions.
It has two sections: a ``try'' and a ``catch'', as illustrated in \autoref{fig:fallback_goto_mechanism}.
It is implemented such that the first concurrent sequence in the ordered children represents the ``try'' and the rest represent the ``catch''.
The ``try'' can be one or more actions and the fallback node waits for them to finish.
If one or more ``try'' actions fail, the entire ``catch'' sequence executes.
There is no limit on the ``catch'' sequence length.
Else, if the entire ``try'' is successful, the ``catch'' is skipped and the node after the fallback node will be executed next.

In January 2025, we added a goto node.
This node simply holds a settable reference to the next node to goto.
It stores this reference simply as a node name in the JSON.
When executed, it sets the ``next execution index'' accordingly.
A goto node is also shown in \autoref{fig:fallback_goto_mechanism}, which implements a ``retry''.

For a while, actions were simply placed as the ``try''.
In this way, if an action failed, a goto could be placed in the catch, pointing back to the ``try'', to retry the action.
For example, if a hand pose action does not achieve its goal pose within the tolerance setting, the node would fail.
This already effectively made a certain amount of reactivity possible.
A push door opening action could react to an unlatching failure through the ``push door open'' action failing to achieve the pushed open pose.

However, it would be better to perform perceptual checks than to attempt and fail physical actions.
For example, when pulling a door open, a slipped grasp would not cause the hand to fail to reach its goal pose.
Instead, it would be better to detect that the door did not open with the hand by observing the point cloud.

It took longer to implement a condition node that had a meaningful impact on behavior capability.
In February 2025, a simple counter condition node was implemented to execute loops with iteration limits.
This was ultimately not very useful.

\subsection{LLM Condition Node}
\begin{figure}[H]
    \centering
    \includegraphics[width=.7\columnwidth]{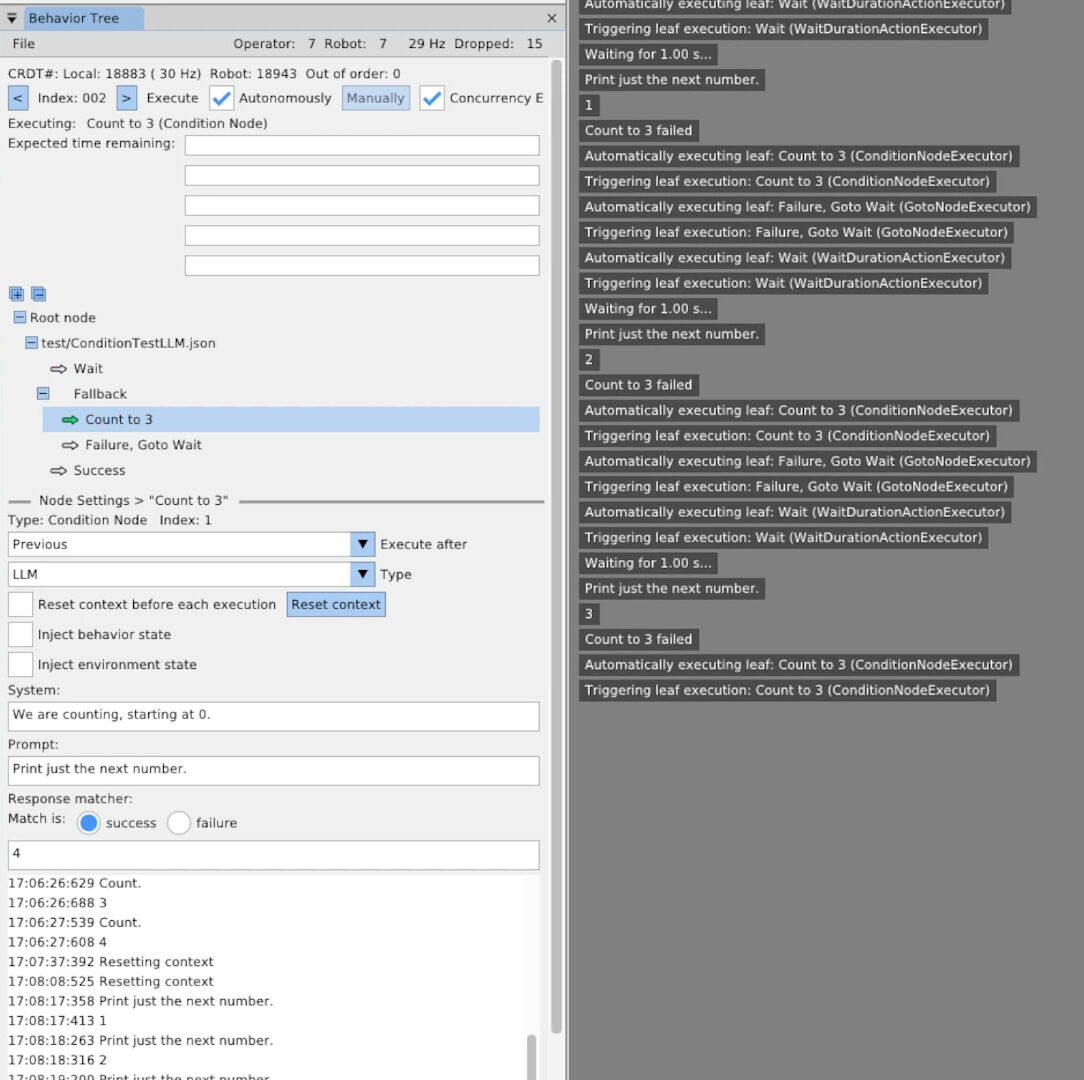}
    \caption{
        A condition node backed by an LLM.
        This test on March 6, 2025, replicated a simple persistent counter node using prompting and a regular expression matcher.
        In this example, the system prompt starts the count at 0, and the repeating prompt asks for the next number.
        The response matching returns success when the response is 4.
        A video is available at \url{https://youtu.be/OQhm2xduzuk}.
    }
    \label{fig:2025_llm_condition_node}
\end{figure}

In March 2025, we implemented an LLM condition node.
It was based on the Llama 3.2 3B Instruct model with 8-bit quantization.
As shown in \autoref{fig:2025_llm_condition_node}, we replicated the counter condition node type with the LLM type.
After experimentation, we found that the LLM's response format varied.
To create a simple binary pass/fail as required by the condition node, we structured the node as three parts: the system prompt, a repeating prompt, and a response matcher.
The system prompt would be input to the model once and the repeating prompt would be input to the model each time the condition node is executed.
The response matcher is implemented with a regular expression that either matches or does not match the output from the LLM.
The behavior author is provided with an option for whether a match means a success or a failure of the condition node.

This worked okay, but we had trouble with such a small LLM model generating meaningful results.
This development did not result in any meaningful impact on the capabilities of loco-manipulation behaviors.
Our focus quickly drifted to approaches that queried paid online VLM models via the AI2R node.

\subsection{Behavior Scene}
\begin{figure}[H]
    \centering
    \includegraphics[width=.7\columnwidth]{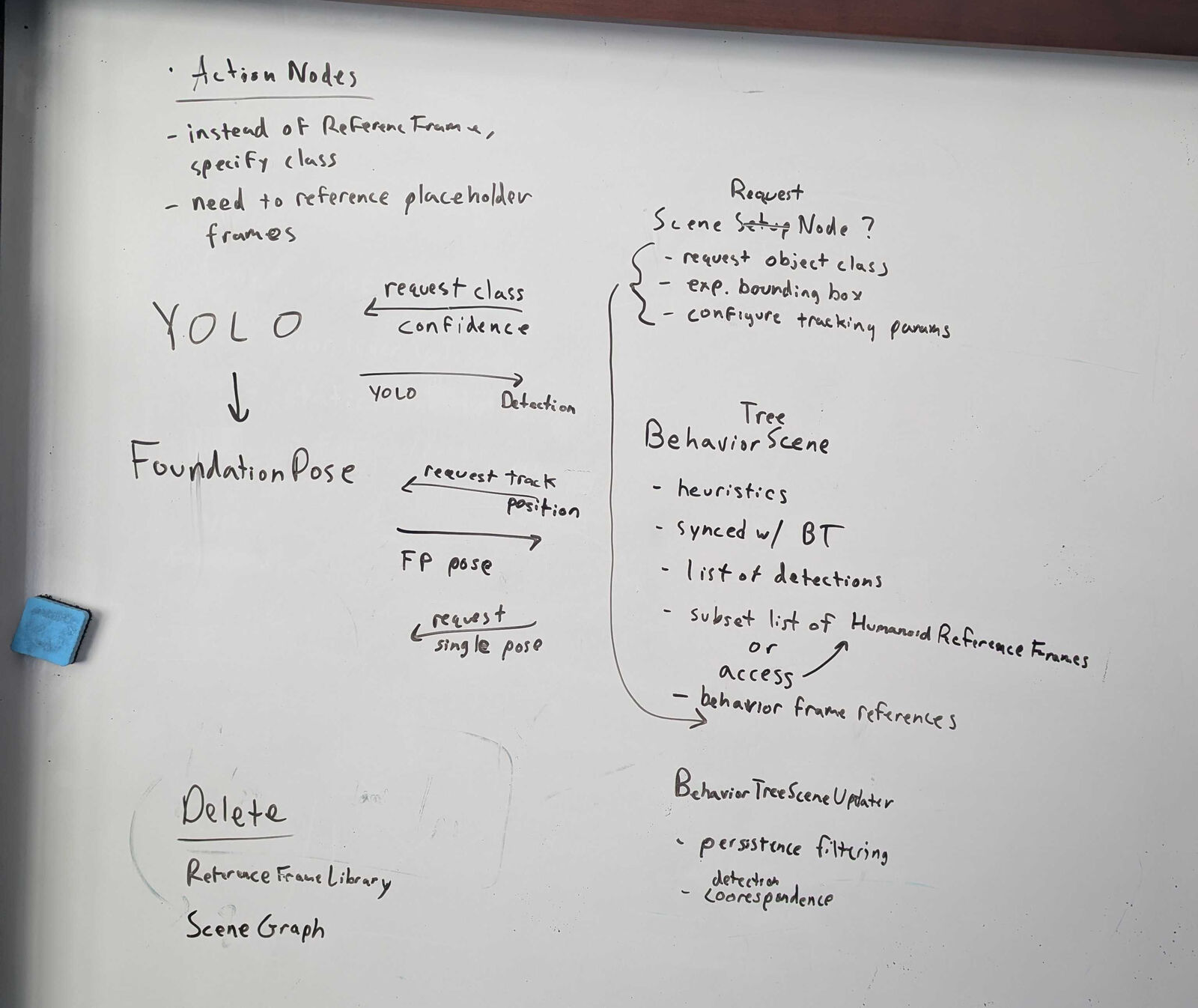}
    \caption{
        An October 21, 2025 behavior design meeting where the behavior scene and the concept of a scene action was solidified.
    }
    \label{fig:2025_scene_action_whiteboard}
\end{figure}

In late October of 2025, we finally acted on the March 3rd, 2025 design meeting, shown in \autoref{fig:2025_whiteboards}, and had another meeting shown in \autoref{fig:2025_scene_action_whiteboard}.
At this time we introduced a behavior scene that is owned by the behavior system alongside the tree, as opposed to relying on an external scene graph system.
The October 21, 2025 meeting is where we solidified the concept of a scene action node.
We pulled the instant and persistent detection components from the scene graph and left everything else.
We referenced the detection manager that was part of the scene graph and created a condensed and revised version.

Instant detections are immutable objects created from each run of the YOLO and FoundationPose models.
They contain all the information about the detection such as the semantic object class name and the timestamp.
Instant detections are extended by the type of the originating detector.
For example, the YOLO instant detection contains the YOLO specific information, like the region-of-interest coordinates, the confidence value, and the segmentation image mask.
The FoundationPose instant detection contains the pose of the object and the 3D bounding box corners.

The YOLO instant detections also pull in and store the corresponding depth points from the ZED X Mini depth image.
This feature required impressive technical wizardry from Tomasz Bialek and Dexton Anderson to maintain performance and avoid memory leaks.
However, this capability of associating 3D depth data with semantic detection masks is essential to making perceptive behaviors work.
The association gives a high frequency 3D position of the centroid of the detected object, which is usable to act on constrained objects like door handles and symmetrical objects like tennis balls.

\begin{figure}[H]
    \centering
    \includegraphics[width=.95\columnwidth]{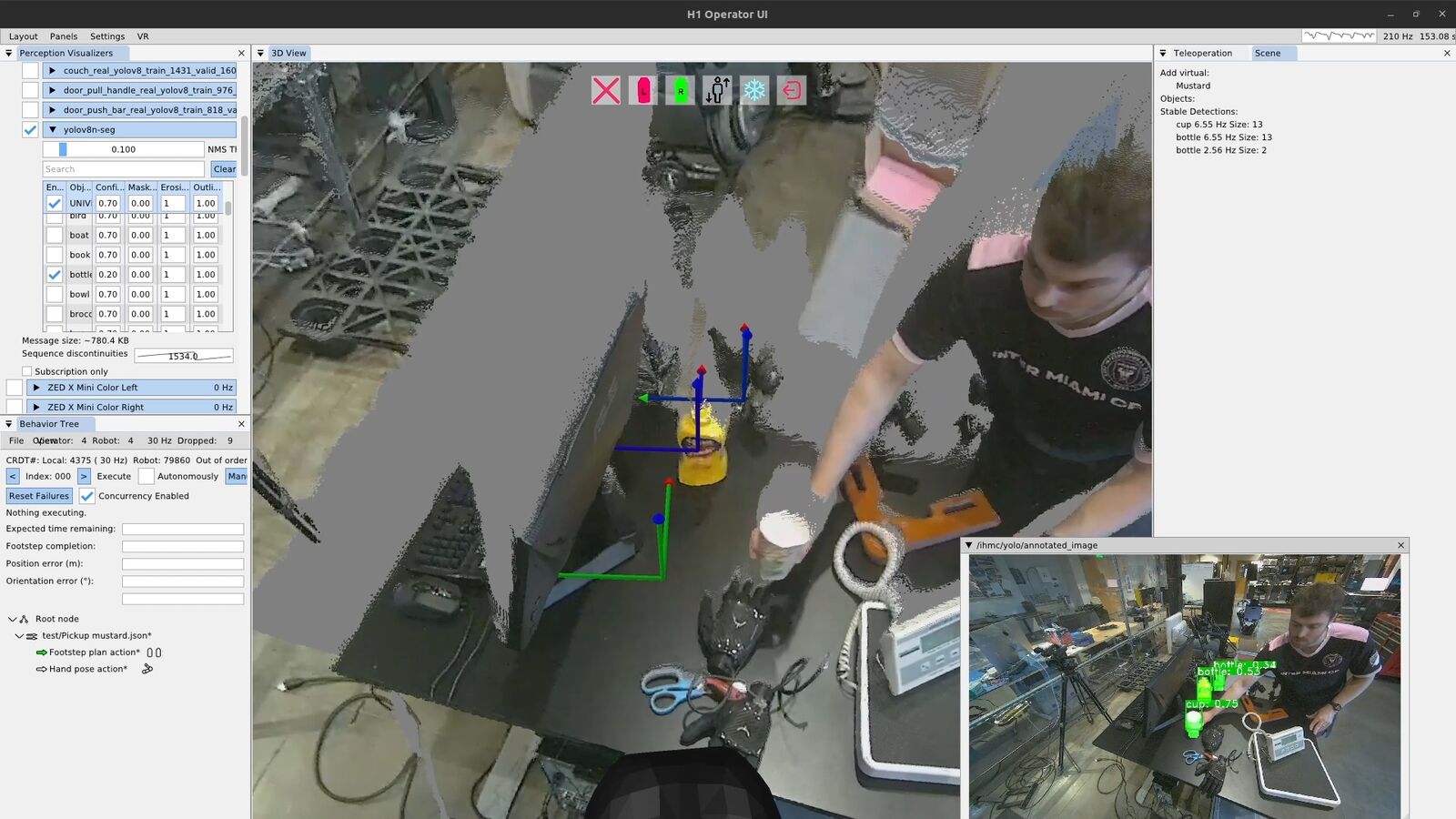}
    \caption{
        A November 4th, 2025 demo of the new behavior scene.
    On the right, the behavior scene gets its own panel, with the persistent detections listed as ``Stable Detections''.
    The persistent detection entries show the semantic class (i.e. cup), the incoming instant detection frequency, and the current history size, which is the number of instant detections held by the persistent detection.
        A video is available at \url{https://youtu.be/NLDHnASOzCc}.
    }
    \label{fig:2025_scene_demo}
\end{figure}

Persistent detections are simply a collection of instant detections over time.
An early demo is shown in \autoref{fig:2025_scene_demo}.
The new behavior scene implementation carried over a similar logical structure to the detection manager that was in the scene graph.
This algorithm sorts incoming instant detections into existing persistent detections or new ones based on location.
If an incoming instant detection is the same semantic class as an existing persistent detection and is close by, it is sorted into that one.
If not, a new persistent detection is created.
This system implements persistent object tracking which can be used to track objects for use in our frame-relative behaviors/affordance templates.

\subsection{Scene Action}
Another innovation in November of 2025 was the introduction of the scene action.
Now that we had an active list of persistent detections, we needed a mechanism for the behaviors to select one and use it.
We had a list of issues we didn't know how to fix from the prior scene graph implementation.
One was that persistent detections would get distorted once they got occluded, such as what happens when grasping an object.
The hand starts to cover the object and the centroid of the segmentation points moves away from the hand, corrupting the prior position estimate.
Another issue was in object selection.
When there are multiple instances of the same class of object, which one do you choose?

Both of these issues require careful coordination with the behavior in order to solve and are context dependent.
So, we decided that maybe it was best if the responsibility to handle these was just handed off to the expert operator.
With that thought in mind, we decided to make the tools that made it possible for the behavior author to handle these issues.
This led to the idea of an authorable scene action, which would have distinct types to do different scene related things.

To solve the object selection issue, we provide a ``setup object'' scene action type which, when executed, chooses the \textit{closest} persistent detection of the defined semantic class and puts that persistent detection in a privileged area called the list of ``objects''.
This privileged list of objects is where attachments to the behavior actions are made.
Any object's pose in this list is immediately available to all behavior actions as a reference frame that can be used to define actions relative to.
This implementation limits the list to contain one instance of a particular semantic class at a time so behavior actions can be defined relative to the class of object -- not the specific instance of one.
Another property of the ``objects'' list is that the persistent detection does not automatically expire.
If instant detections stop coming in for it, it merely becomes stale, indicated as greyed out in the UI, but the frame is still available for the behavior actions.
This is actually an important and intentional feature, as this can be used to ``dead reckon'' actions with respect to something you perceived before but can no longer see.
For example, for door behaviors, the traversal walk through is actually defined with respect to the stale opening mechanism reference frame.
We planned to have selection heuristics for the ``setup object'' scene action type, but haven't yet ended up with a burning need for anything other than \textit{closest}.

To solve the hand-object occlusion issue, we gave the operator a ``freeze object'' scene action type.
This allowed a ``freeze'' to be performed on a scene object just before grasping it.
For example, a scene object can be set up, the hand moved towards it, but not occluding it, while still tracking the object, and a freeze object action can be executed just before finally occluding the object.
The freeze operation has an effect equivalent to the object becoming ``stale'' as described earlier.
The object's frame, similarly, can still be used for physical action definitions.

\subsection{Ability Hand and Reliability}
\begin{figure}[H]
    \centering
    \includegraphics[width=.95\columnwidth]{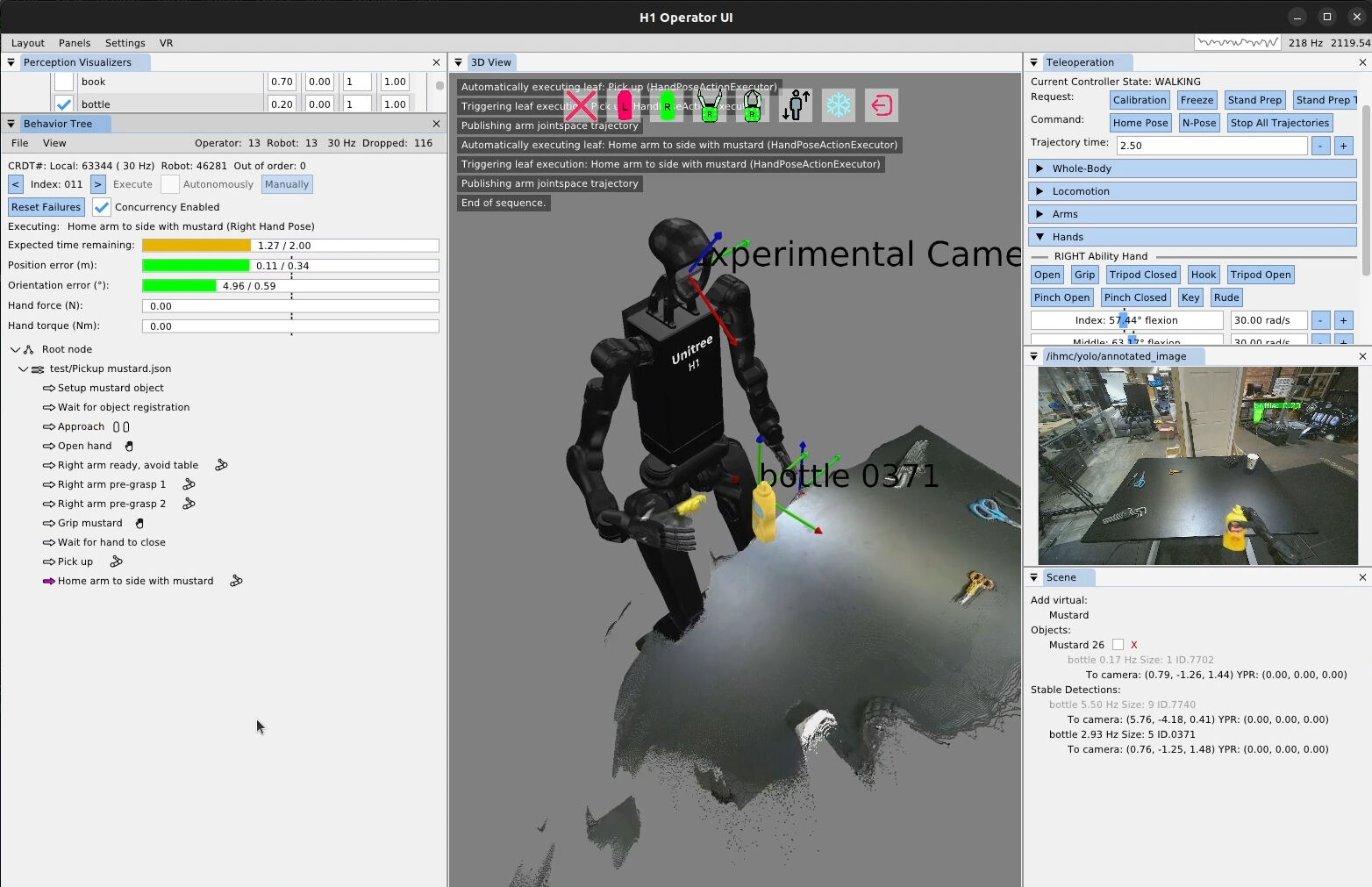}
    \caption{A November 10th, 2025 demo where our Unitree H1-2~\cite{unitree_h1_2} robot picked up a mustard bottle automatically using the new scene action node.
    A video is available at \url{https://youtu.be/PVtL7KYMC9g}.
    }
    \label{fig:2025_mustard_pickup}
\end{figure}

A demonstration on November 10th, 2025, showed the utility of the new behavior scene and scene action node.
This was the first time we picked up an object from a table in the fully automatic mode, as shown in \autoref{fig:2025_mustard_pickup}.
It was also the first manipulation behavior we executed on the Unitree H1-2~\cite{unitree_h1_2} robot we got that year.
There were still lots of little bugs and issues in this demo, but nothing super theoretical.
For one, our whole body controller was not great at walking with the H1-2 so the approach stance was not accurate, resulting in a low quality IK solution for reaching the mustard due to not approaching the table closely enough.
A low quality IK solution with our system has always meant a high error bar on the resulting hand pose.
Another of the biggest issues was that our finger control was unreliable.
We ended up fixing that on the hand controller side in the coming months.

\subsection{Proximity Condition}
\begin{figure}[H]
    \centering
    \includegraphics[width=.95\columnwidth]{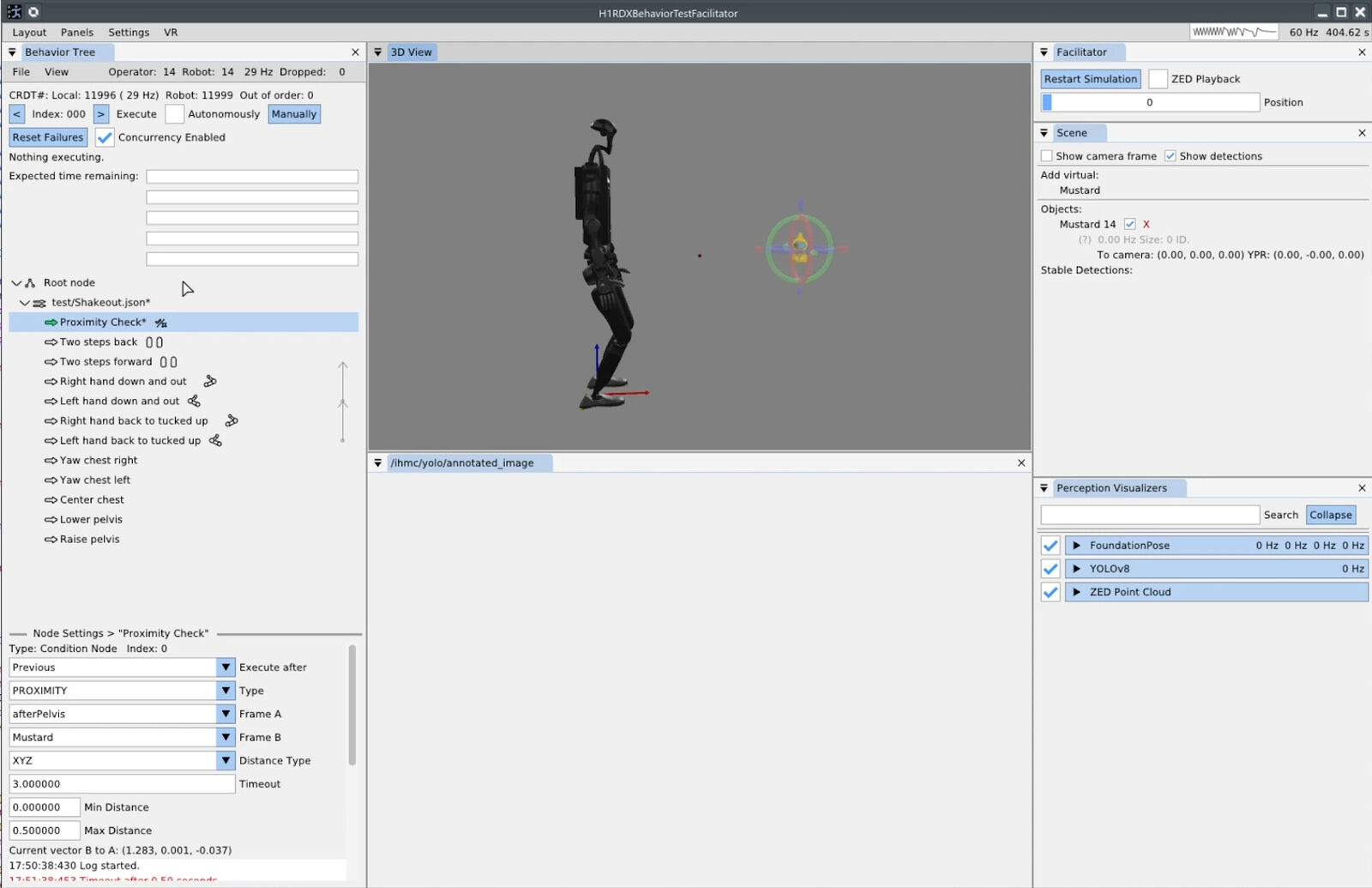}
    \caption{A November 19, 2025 test of our new generalized proximity condition.
    A video is available at \url{https://youtu.be/CeHAXjiN4hw}.
    }
    \label{fig:2025_proximity_condition}
\end{figure}

On November 19, 2025, we generalized an initial implementation of the proximity condition type, as seen in \autoref{fig:2025_proximity_condition}.
In the bottom left of the screenshot, the options can be seen.
The proximity condition type compares the positions of two reference frames.
In the case shown, frame A is selected as ``afterPelvis'' (the pelvis frame) and frame B is selected as the mustard object, which we've instantiated and posed virtually in simulation.
The distance type field selects the comparison criteria: XYZ, which compares the pure Euclidean distance, XY, which compares the distance as projected onto the XY plane, and Z, which simply compares the height of the frames.
A min and max distance is specified to define the success criteria.
When the condition executes, it continues executing until the criteria are met (success) or the timeout is reached (failure).
The current distance is also presented in the operator interface to help with authoring.
Observing the current values can help the operator reason about which min and max values are best.

\subsubsection{Concurrency Visualization}
\begin{figure}[H]
    \centering
    \includegraphics[width=.5\columnwidth]{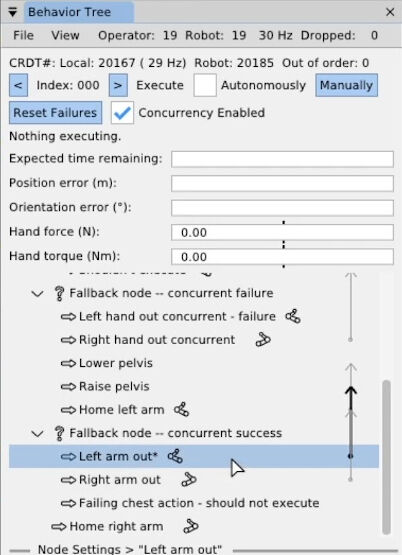}
    \caption{
        A November 13th, 2025 development -- showing arrows on the right side which represent the ``execute after'' dependencies visually.
        A video is available at \url{https://youtu.be/4_26TwmRYr4}.
    }
    \label{fig:2025_execute_after_arrows}
\end{figure}

Later in November, we added a visualization element for the ``execute after'' field for action nodes, as shown in \autoref{fig:2025_execute_after_arrows}.
Prior to this feature, to understand the concurrency of sequences you would select each action in the list and check the ``execute after'' field.
You could also click the little green arrows and see if the ones after it were also green, indicating that they would execute together.
However, we wanted to reduce uncertainty and increase understandability of the behavior at a glance.
We came up with the arrow design, seen in the figure, which starts at the defining action and extends upwards, pointing to the action specified by the ``execute after'' setting.

On December 1, 2025, we conducted our first behavior shakeout on IHMC's new fully electric humanoid robot, Alex.
We would soon transition behavior development fully to our new robot.
The behavior shakeout consists of a sequence of open-loop walking and arm motions to test the system's basic mechanics.

\subsection{Logging Behavior Data}
\label{sec:logged_data}
\begin{figure}[H]
    \centering
    \includegraphics[width=.95\columnwidth]{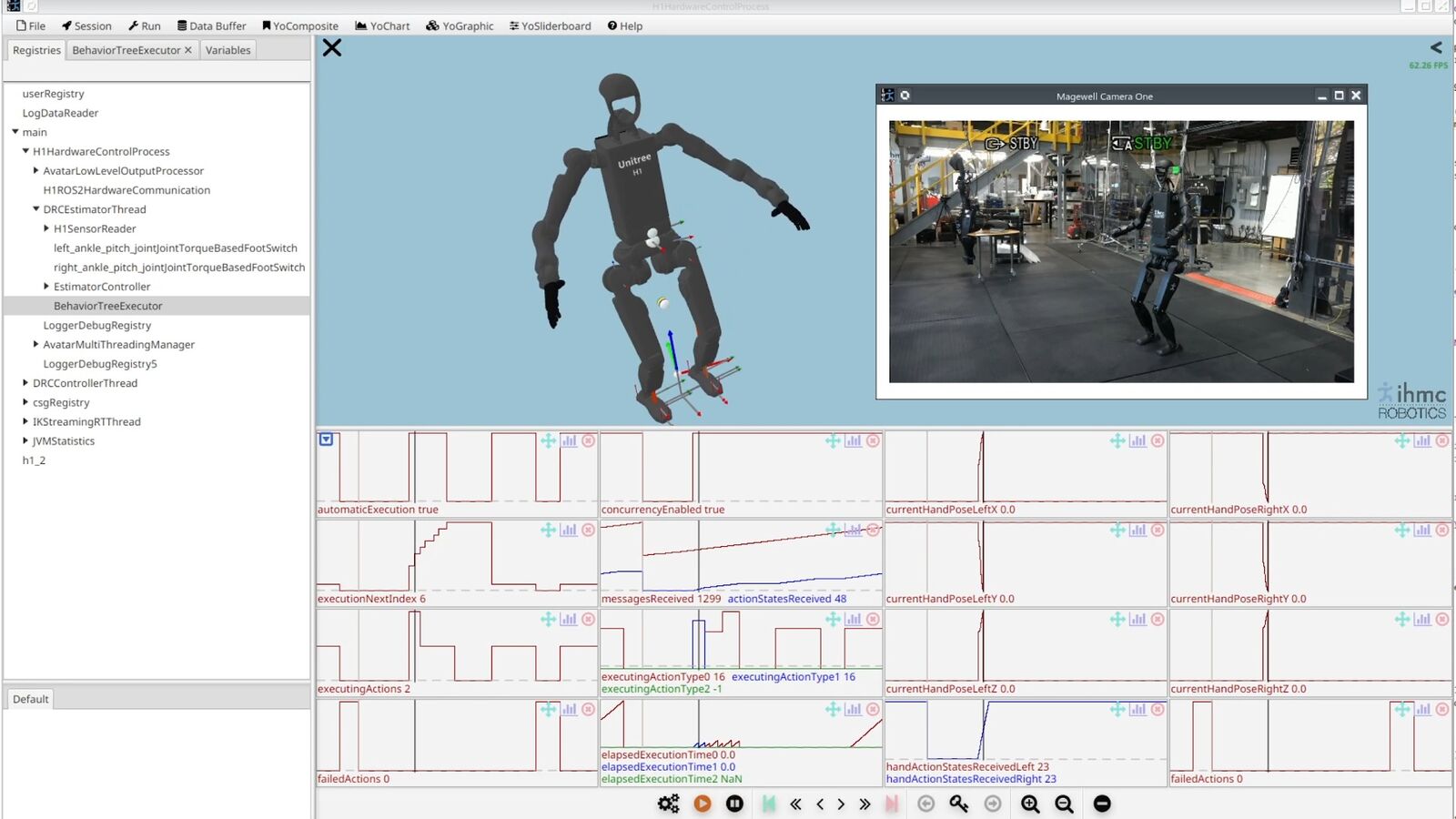}
    \caption{A December 10, 2025 screenshot that shows the BehaviorTreeExecutor YoVariable registry in the tree on the left-hand side.
    Behavior variables such as ``automaticExecution'' and ``executionNextIndex'' are plotted over time.
    A video is available at \url{https://youtu.be/mcS6PkfpIWE}.
    }
    \label{fig:2025_behavior_yovariables}
\end{figure}

Later in December, we implemented behavior data logging in a way that is synchronized and viewable alongside whole body controller data.
In this system, the behavior system process sends a periodic ROS 2 message containing a registry of behavior YoVariables, the buffered data types supported in SCS.
A subscriber is set up in the whole body controller process to subscribe to these variables and inject them into the whole body controller's registry, which then gets logged in the normal way.
A screenshot of a loaded log in SCS is displayed in \autoref{fig:2025_behavior_yovariables}, showing the behavior variables embedded in and synchronized with the whole body controller log.

\subsection{FoundationPose Integration}
\begin{figure}[H]
    \centering
    \includegraphics[width=.95\columnwidth]{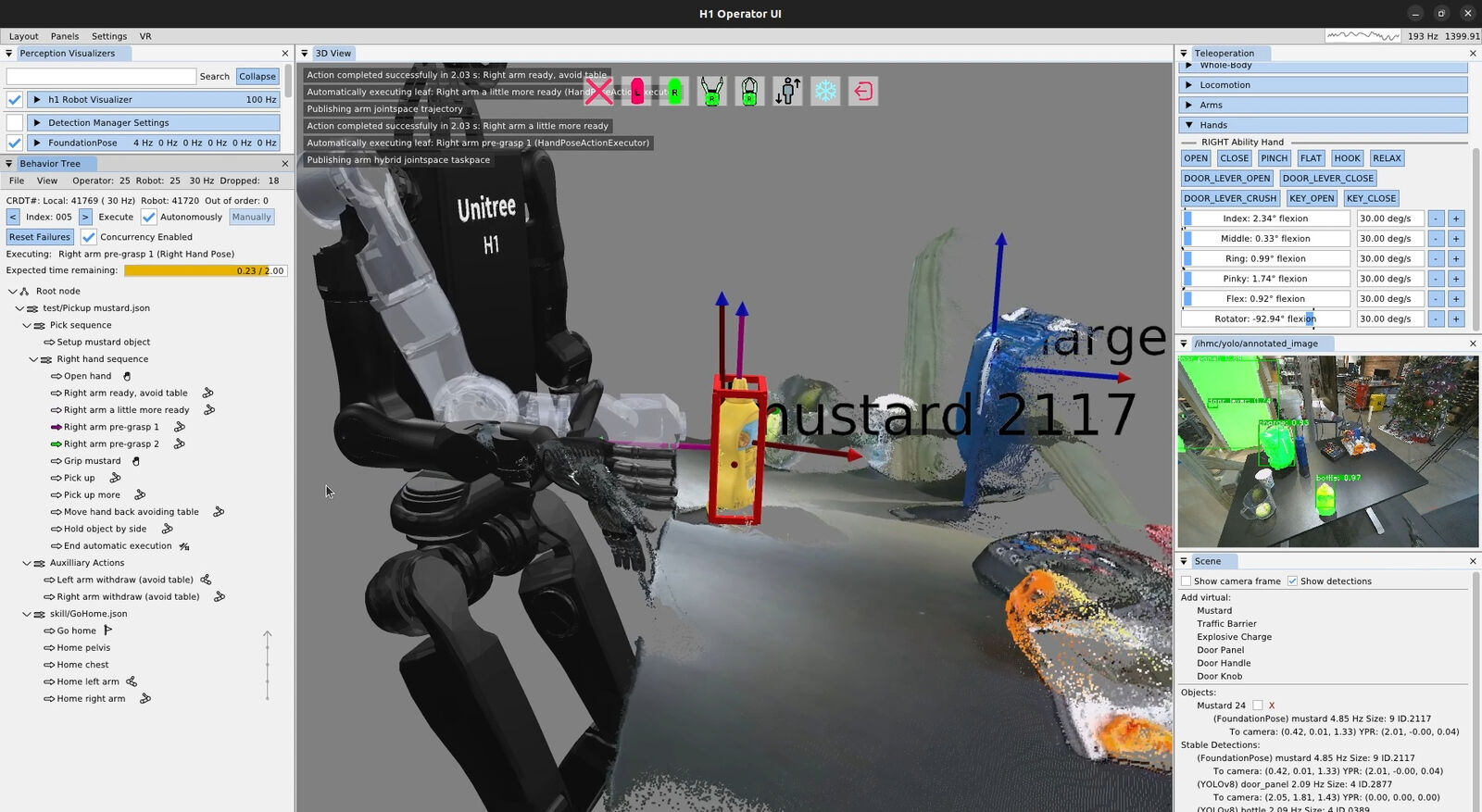}
    \caption{A December 11, 2025 demo where we used FoundationPose~\cite{wen2024foundationpose} to detect the pose of the mustard bottle and execute an automatic behavior to pick it up.
    A video is available at \url{https://youtu.be/tP2ZvOkEjFk}.
    }
    \label{fig:2025_fp_mustard_pickup}
\end{figure}

On December 11, 2025, we integrated FoundationPose~\cite{wen2024foundationpose} into the behavior scene.
We used this to create an improved version of the mustard pickup task, as shown in \autoref{fig:2025_fp_mustard_pickup}.
This added the capability to grasp asymmetric objects as FoundationPose detects an object's orientation and bounding box in addition to the position.
The red box in the figure shows the FoundationPose bounding box.
In the bottom right, the FoundationPose scene object for the mustard can be seen.
At this point we had two types of scene objects: FoundationPose and YOLOv8.
Our scene was now designed to handle different types of objects abstractly and be extended easily to support more types.

\subsection{Ability Hand Control}
This demo was still somewhat unreliable due to the low level control code of our new 5-finger PSYONIC Ability Hand~\cite{akhtar2022touch} grippers.
The hands would often not respond to behavior commands and, worse, retrying the command did not work.
By December 15, 2025, after trying at least five different control strategies, we fixed this issue and finger control was very smooth and reliable.
The on-board Ability Hand controller wants an alpha-filtered position setpoint input that is generated from an ideal velocity-limited trajectory, as presented in \autoref{alg:ability_hand_pos_vel_limit}.
This helps the hand reliably move at the desired velocity to the desired setpoint and hold the position without jittering.

\begin{algorithm}[H]
    \SetAlgoLined
    \caption{Ability Hand Velocity-Limited Position Control Update}
    \label{alg:ability_hand_pos_vel_limit}

    \KwIn{Time step $\Delta t$, target positions $\mathbf{q}_{goal}$, maximum velocities $\mathbf{\dot{q}}_{max}$}

    $f_c \gets 1.0$ \tcp*{Filter break frequency in Hz}
    $\tau \gets \frac{1}{2 \pi f_c}$\;
    $\alpha \gets \frac{\Delta t}{\tau + \Delta t}$\;

    \For{$i = 1 \dots 6$}{
        \tcp{Apply first-order low-pass filter on velocity and position commands}
        $\dot{q}_{filt}[i] \gets \alpha \dot{q}_{raw}[i] + (1 - \alpha) \dot{q}_{filt_prev}[i]$\;
        $q_{filt}[i] \gets \alpha q_{cmd}[i] + (1 - \alpha) q_{filt_prev}[i]$\;

        \tcp{Calculate velocity-limited position step}
        $d \gets \text{sgn}(q_{goal}[i] - q_{cmd}[i])$\;
        $\Delta q \gets d \cdot \mathbf{\dot{q}}_{max}[i] \cdot \Delta t$\;

        \tcp{Update commanded state}
        $q_{cmd}[i] \gets q_{cmd}[i] + \Delta q$\;
        $\dot{q}_{cmd}[i] \gets 0$\;
    }
\end{algorithm}

\subsection{Hand and Finger Previewing}
\begin{figure}[H]
    \centering
    \includegraphics[width=.95\columnwidth]{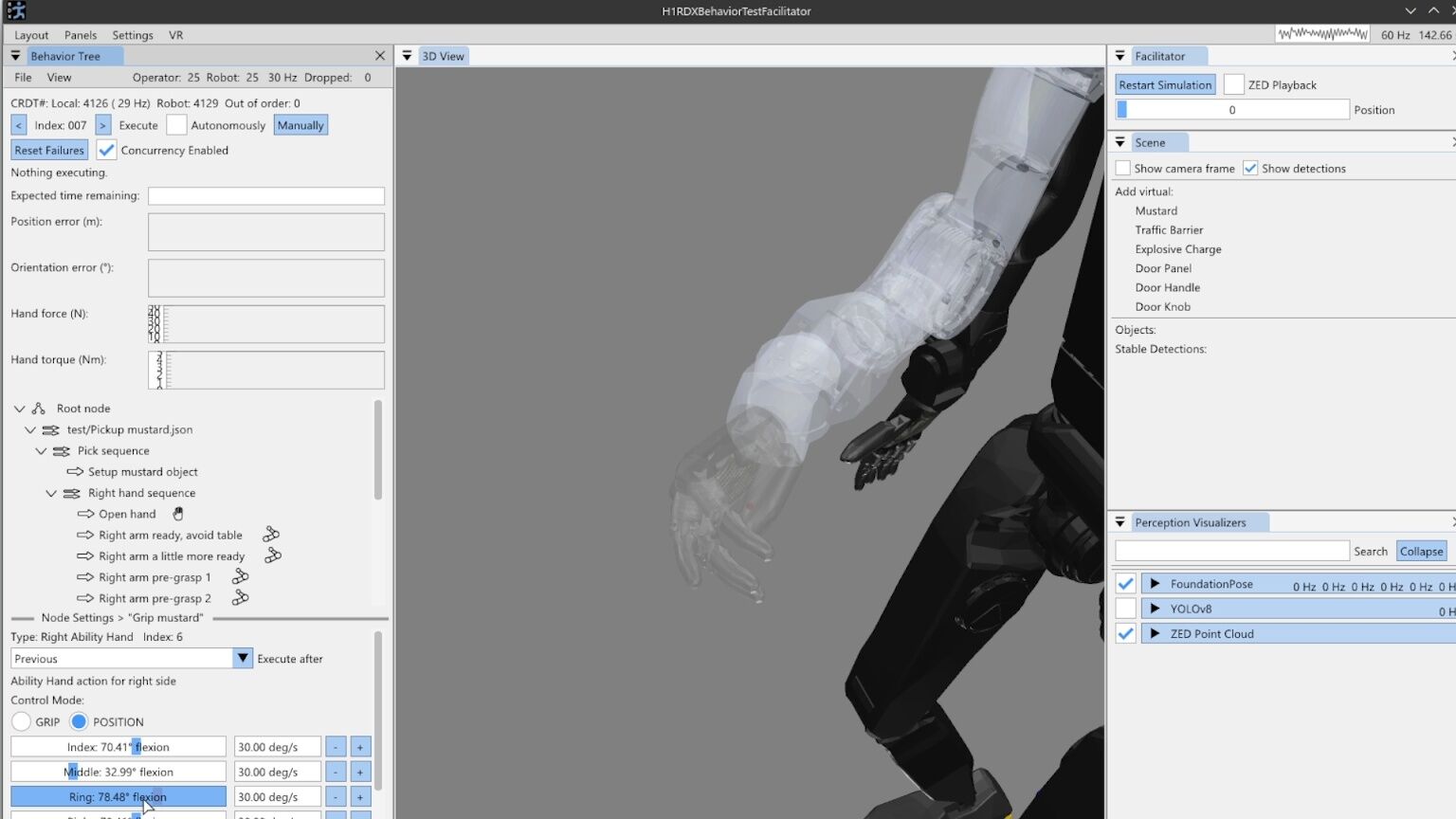}
    \caption{A screenshot of the hand previewing feature from December 13, 2025.
    In the center, the transparent hand indicates the future grasp configuration.
        A video is available at \url{https://youtu.be/ioqLeNys_Mc}.
    }
    \label{fig:unitree_hand_previewing}
\end{figure}

Later in December, we added hand and finger previewing in the behavior editor, as shown in \autoref{fig:unitree_hand_previewing}.
This feature was designed to help the operator understand where the fingers will be during authoring.
When a hand pose action is selected, the finger previews show up at the end of that arm pose.
Since the pose of the arm and the configuration of the fingers are specified in separate actions and can be interwoven with other actions, this feature looks backwards in the sequence to find the most recent Ability Hand action.
This action is used for the finger preview.
If there is no prior Ability Hand action, the finger preview will not be shown.

This feature could probably be improved in a few ways.
First, we could also show the preview when an Ability Hand action is selected, utilizing prior arm actions for the arm and hand locations.
Secondly, if an arm action is selected and there are no prior Ability Hand actions, we could use the current configuration of the hand for the preview.

\begin{figure}[H]
    \centering
    \includegraphics[width=.3\columnwidth]{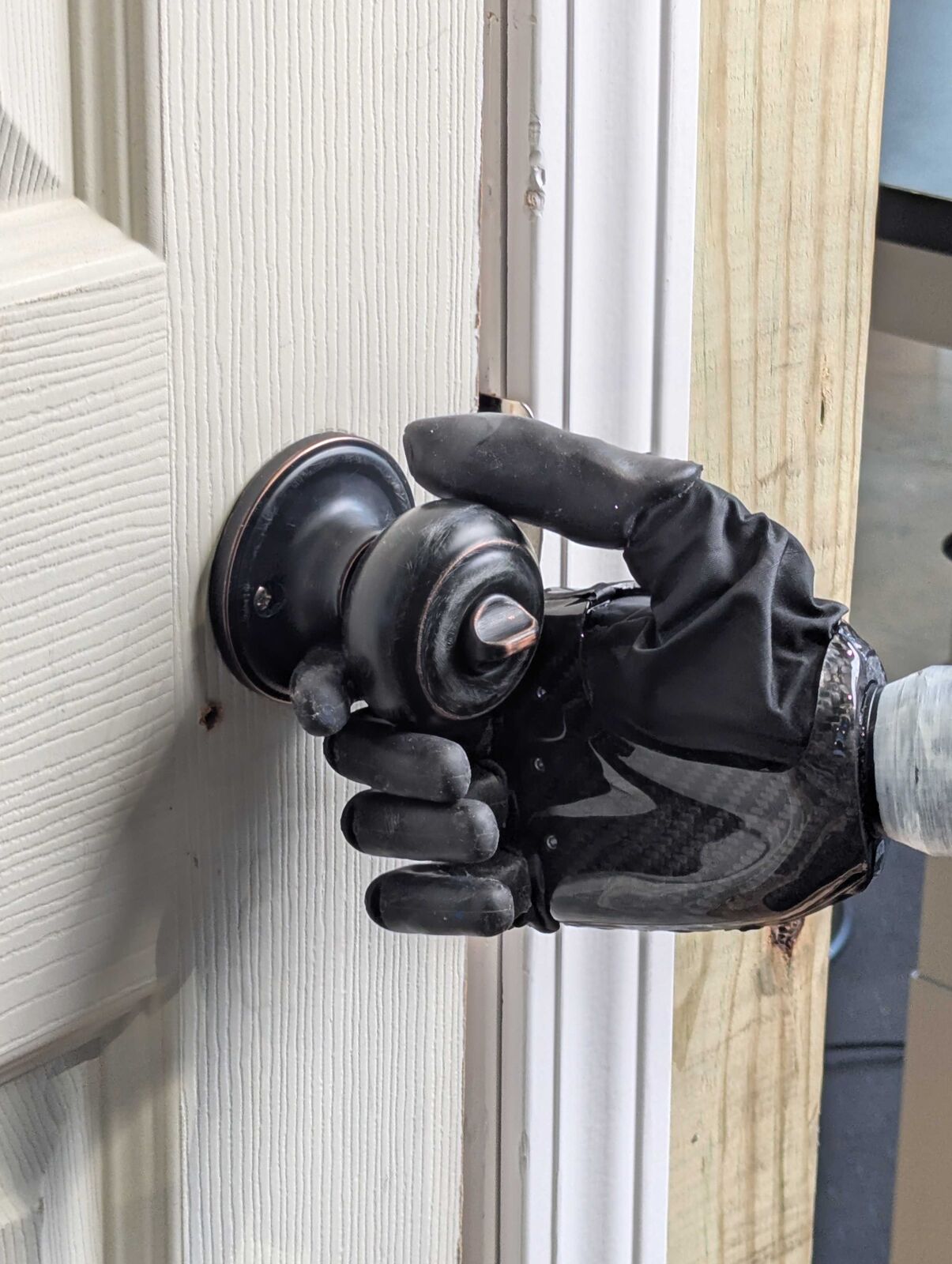}
    \caption{Unitree H1-2 with Ability Hand performing a pinch grasp on the knob door handle.
    This is a very secure and reliable grasp, mimicking nature, and provides a wide error margin for pre-grasp hand poses.}
    \label{fig:unitree_ability_hand_knob_grasp}
\end{figure}

\subsection{Door Knob Turning}
\begin{figure}[H]
    \centering
    \includegraphics[width=.95\columnwidth]{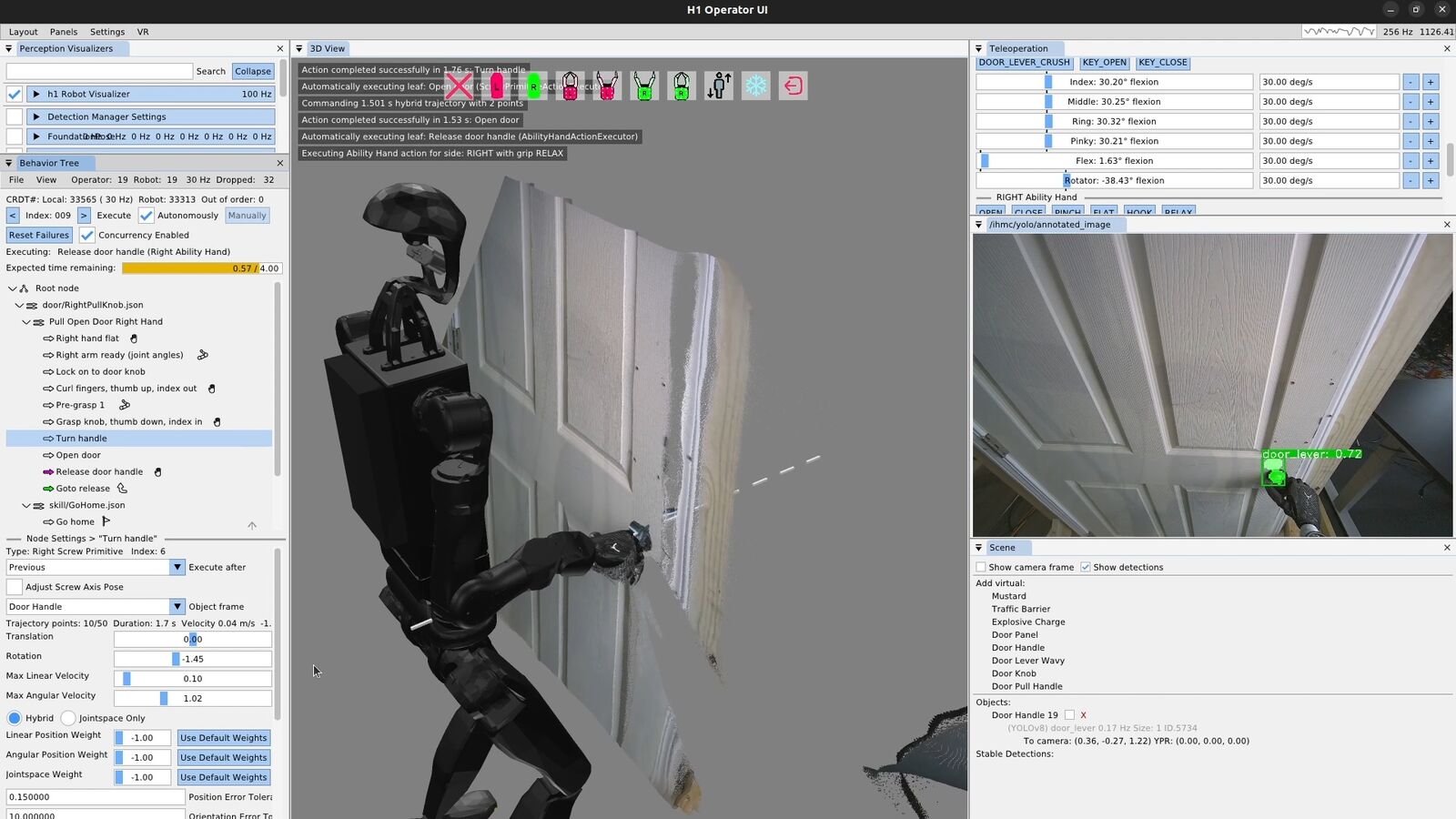}
    \caption{Our December 23, 2025 demonstration of a knob door handle repeated opening demo.
    A video is available at \url{https://youtu.be/3hRLBV4dZ0Q}.
    }
    \label{fig:unitree_knob_23x}
\end{figure}

On December 23, 2025, using the pinch grasp shown in \autoref{fig:unitree_ability_hand_knob_grasp}, we demonstrated a pull knob door handle opening task automatically 23 times in a row, as shown in \autoref{fig:unitree_knob_23x}.
On the 24th attempt, the knob was not turned enough, and the hand slipped off on the pull.
This was our first return to doors in over a year.
Using now higher-confidence and more robust YOLO models for the door mechanisms in combination with the scene actions and the 5-finger Ability Hand gripper, door handle manipulation reliability has drastically increased.

\subsection{Lever-Handle Door Opening Demo}
\begin{figure}[H]
    \centering
    \includegraphics[width=.95\columnwidth]{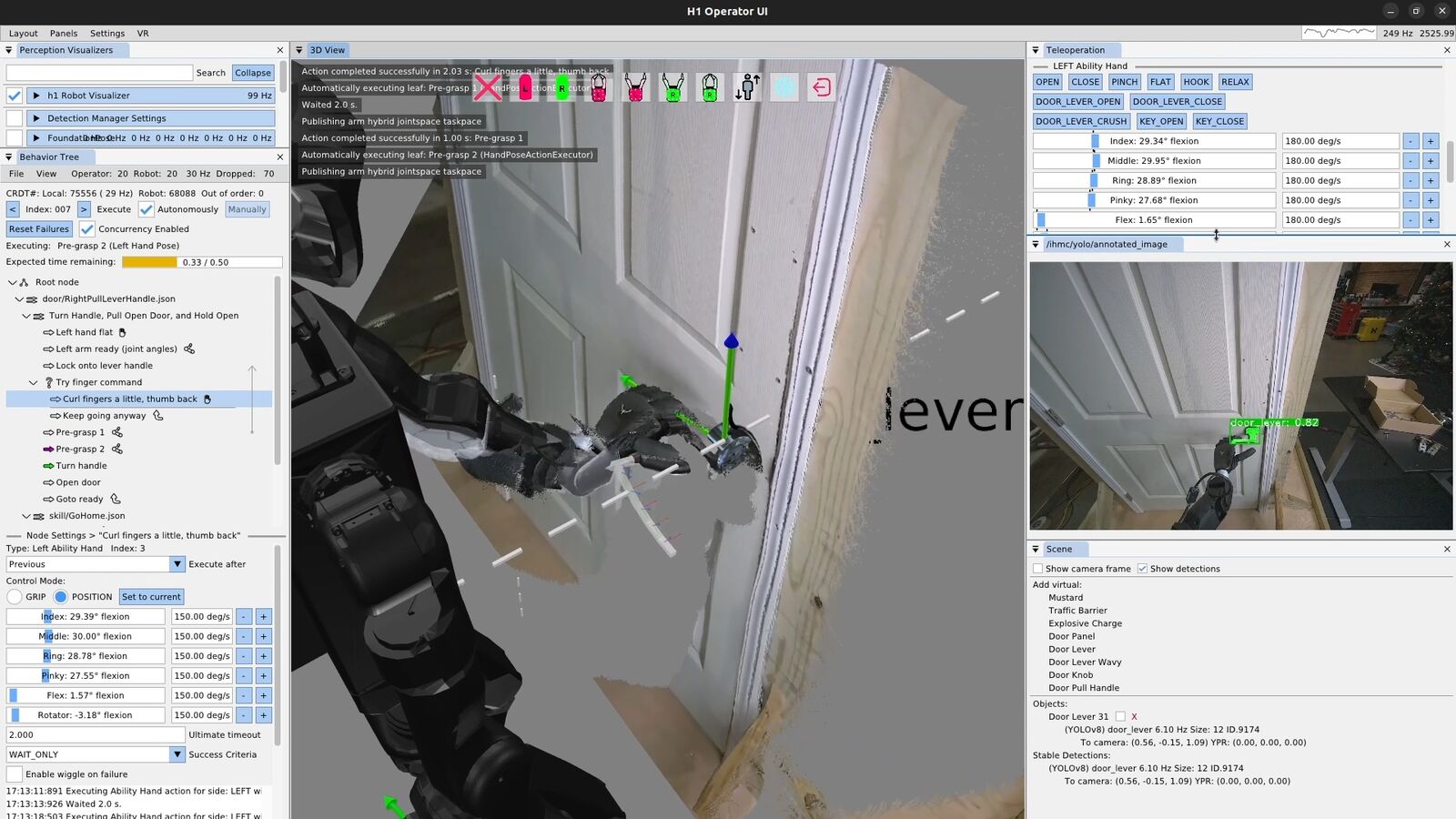}
    \caption{The 32-time pull lever door opening reliability demo run on January 2nd, 2026.
    A video is available at \url{https://youtu.be/fQCNuXEF9xg}.
    }
    \label{fig:unitree_lever_32x}
\end{figure}

On January 2, 2026, a similar demonstration with a door lever handle achieved 32 pull door openings in a row, as shown in \autoref{fig:unitree_lever_32x}.
This time, the behavior was not run until failure.
It was just run until we got bored.
Similarly to the pull knob behavior, we think the Ability Hand being grippy and in an anthropomorphic 5-finger configuration contributed to the reliability along with the scene action and the maturation and stabilization of our behavior system as a whole.

\begin{figure}[H]
    \centering
    \includegraphics[width=.5\columnwidth]{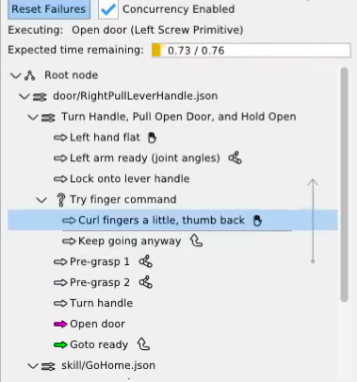}
    \caption{The hand action fallback workaround from January 4th, 2026.
    The behavior would try the ``Curl fingers a little, thumb back'' command, but keep going regardless of the result.}
    \label{fig:unitree_fallback}
\end{figure}

However, this demo was not issue-free.
There was some kind of bug in the hand action unrelated to hand control that was causing the action to fail.
In this case, simply ignoring the failure would work to get the behavior to succeed, because this finger curling motion only slightly increased reliability, but was not necessary.
We were able to use a fallback node to keep the behavior going regardless of failure, as shown in \autoref{fig:unitree_fallback}, marking our first use of the fallback node in a real robot behavior.
The use of the fallback node helped in this case not to overcome an environmental failure, but to overcome a bug in the code, which was not a use case we expected.
We learned that the fallback node was also useful for working around bugs in the system, reducing the chance that we would need to stop the robot experiment and go fix code bugs.

\subsection{Scene Action Improvements}
\begin{figure}[H]
    \centering
    \includegraphics[width=.5\columnwidth]{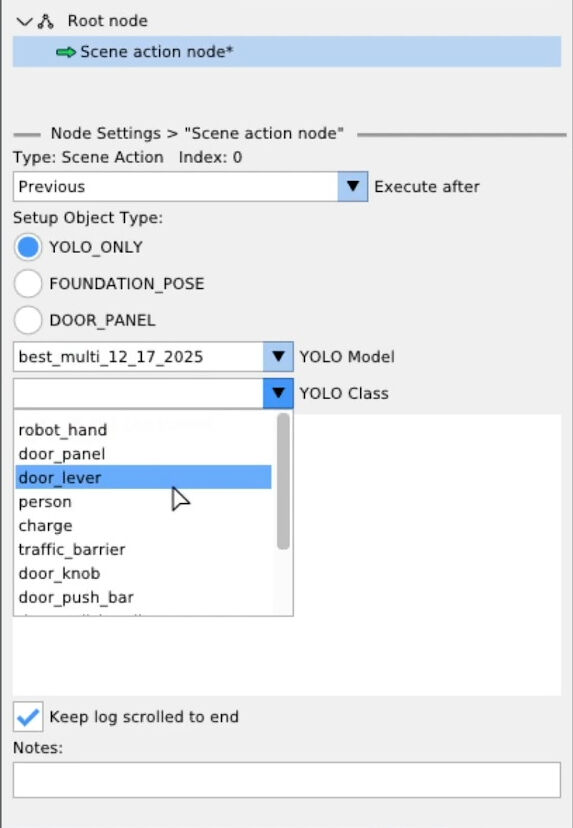}
    \caption{The scene action settings for setting up a YOLO object.
    Dropdown combo boxes are available to select the YOLO model and object class.
        A video is available at \url{https://youtu.be/TPXugNdRx6E}.
    }
    \label{fig:unitree_scene_setup_type}
\end{figure}

On January 8, 2026, the scene action was extended to fully support our full library of FoundationPose and YOLO models as shown in \autoref{fig:unitree_scene_setup_type}.
The operator was now able to select any semantic object type from either model architecture via drop-down menus.
We also extended the scene object library to include heuristic objects, such as the door panel, shown here.

\subsection{Door Panel Tracking}
\begin{figure}[H]
    \centering
    \includegraphics[width=.95\columnwidth]{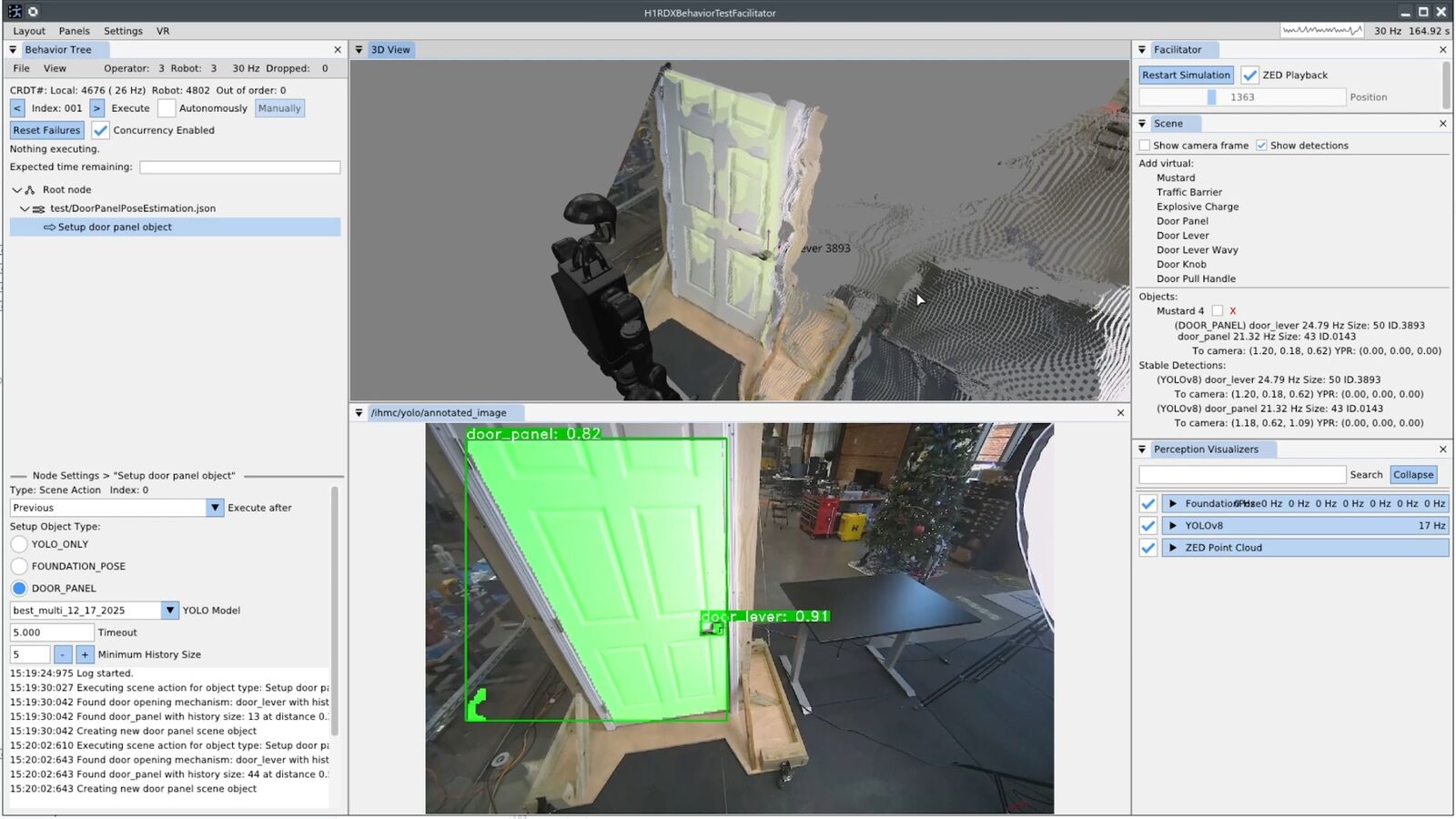}
    \caption{Our door panel detector in a screenshot from January 20, 2026.}
    \label{fig:unitree_door_panel}
\end{figure}

On January 20, 2026, we implemented that door panel heuristic scene object, as shown in \autoref{fig:unitree_door_panel}.
It uses two YOLO persistent detections, one for the handle and one for the panel, and draws a line between the centroids to identify the orientation of the panel.
This is possible with the assumption that door panels swing on a vertical hinge.

\section{2026 Alex, Resilience, Adaptability Era}

\subsection{Alex Pull Door Traversal}
\begin{figure}[H]
    \centering
    \includegraphics[width=.95\columnwidth]{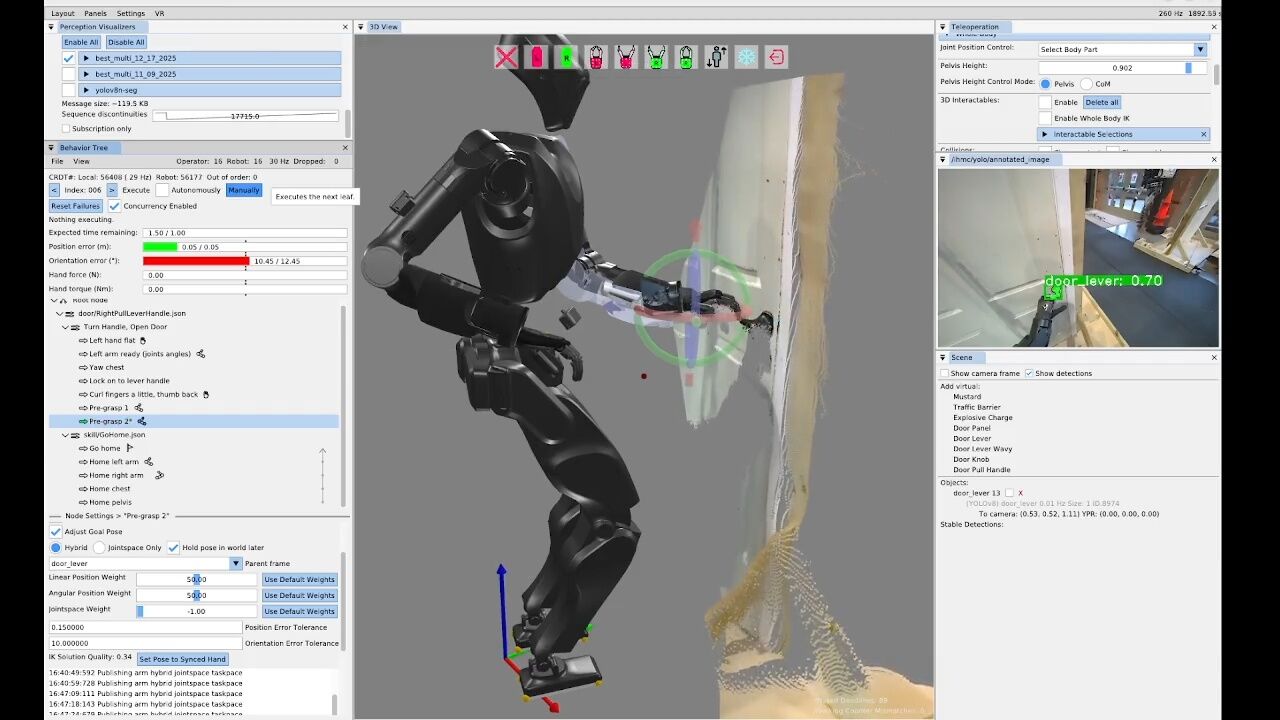}
    \caption{Our first manipulation behavior on the IHMC Alex robot on January 20, 2026.
    A video is available at \url{https://youtu.be/t4vXJIXfsl8}.
    }
    \label{fig:alex_door_opening}
\end{figure}

On the same day, in a 33 minute authoring session, we demonstrated a stand-in-place door opening behavior on the new IHMC Alex robot, as seen in \autoref{fig:alex_door_opening}.
This was the first autonomous manipulation behavior to run on the Alex platform.
The process for getting this to work was essentially the same as the prior door opening behaviors on the Unitree H1-2, with the exception that we needed to point the head down, which wasn't supported as a behavior action yet.
Alex features a neck with pitch and yaw and the default neck pitch does not provide sufficient visibility of the manipulation zone in front of the robot.
To perform manipulation, we need to pitch the head down by 30 degrees.

\begin{table}[H]
    \caption[First Alex right pull authoring milestones.]{Normalized authoring timeline for the January~20--24,~2026 first Alex loco-manipulation behavior, the right pull lever-handle door traversal.
    The time column is cumulative across days but excludes inactive gaps between daily sessions.}
    \centering
    \footnotesize
    \setlength{\tabcolsep}{4pt}
    \renewcommand{\arraystretch}{0.97}
    \begin{tabular}{>{\RaggedRight\arraybackslash}p{1.4cm} >{\RaggedRight\arraybackslash}p{1.95cm} >{\RaggedRight\arraybackslash}p{\dimexpr\textwidth-3.35cm-6\tabcolsep\relax}}
        \hline
        Day & Time & Authoring milestone \\
        \hline
        Day~1 & 0:00:00 & Create \texttt{door/RightPullLeverHandle.json}. \\
        Day~1 & 0:32:59 & Closed unlatch-and-open loop from a squared-up stance. \\
        Day~2 & 1:02:50 & Approach sequence created; footstep-planning issues encountered. \\
        Day~3 & 1:27:57 & Approach finalized; pre-grasp work begins. \\
        Day~3 & 3:15:35 & Knee-clearance check against the swinging panel. \\
        Day~3 & 3:57:05 & First open from a staggered approach stance. \\
        Day~4 & 5:32:44 & Pull-and-hold sequence begins. \\
        Day~4 & 6:14:21 & Door opens farther but not fully by end of session. \\
        Day~5 & 7:52:57 & Door fully open; first traversal step attempted. \\
        Day~5 & 9:52:47 & Full traversal footstep plan authored. \\
        Day~5 & 10:54:06 & Hold-open fallback added after force-related trouble. \\
        Day~5 & 11:01:58 & Pre-authored \texttt{Pull right arm in} clearance pose added. \\
        Day~5 & 11:10:17 & First successful full pull-door traversal. \\
        \hline
    \end{tabular}
    \label{tab:alex_first_right_pull_authoring_timeline_story}
\end{table}

As presented in \autoref{tab:alex_first_right_pull_authoring_timeline_story}, the authoring of the first pull door behavior on Alex was a five day process spanning 11 hours and 10 minutes of authoring time.

\begin{figure}[H]
    \centering
    \includegraphics[width=.95\columnwidth]{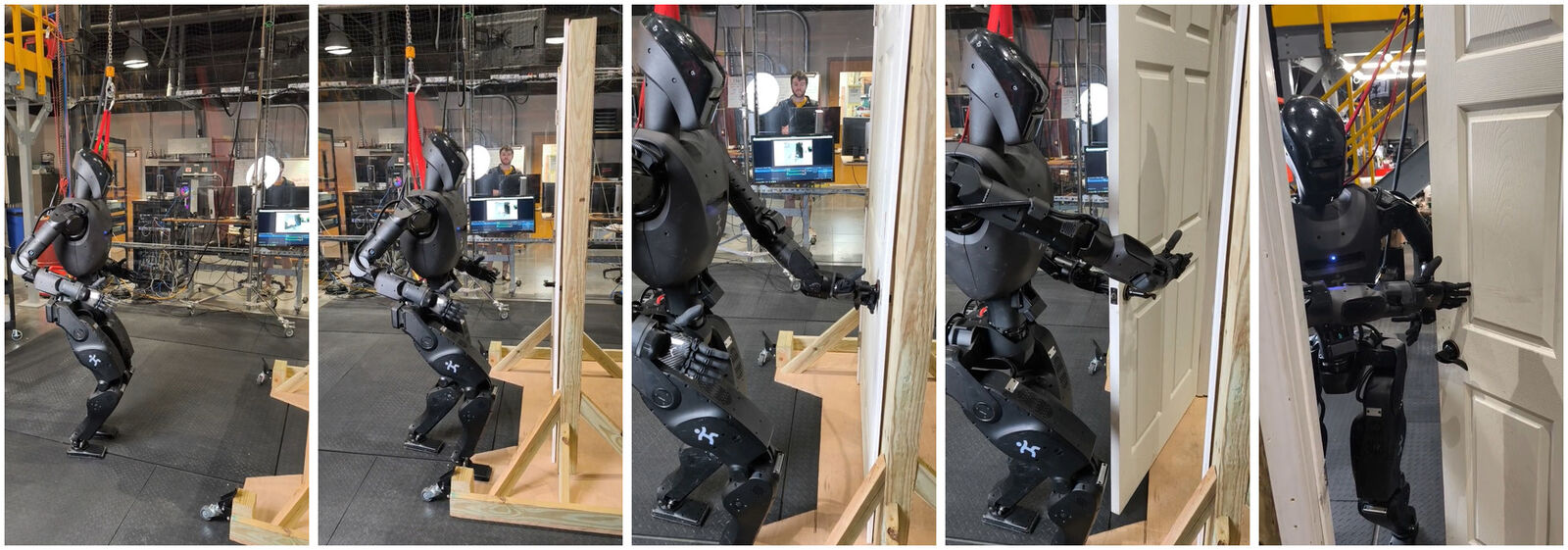}
    \caption{Alex executing a pull door approach and opening behavior on January 23, 2026 as the behavior was being developed.
    A video is available at \url{https://youtube.com/shorts/0_joZBgrBsQ}.
    }
    \label{fig:alex_door_approach_opening}
\end{figure}

On January 23, 2026, three days after the squared up stance opening behavior, we achieved the approach and opening portion of the pull door behavior, as shown in \autoref{fig:alex_door_approach_opening}.
This took a while to get working because Alex's arms are shorter than Nadia's were and the spine range of motion was more limited, meaning the same strategy did not work.
Nadia's spine yaw range of motion of +/- 60 degrees was twice that of Alex's +/- 30 degrees.
Also on Nadia, we took a double support stance for opening that was further from the door, using the spine yaw and the longer arms.
This allows Nadia's pull door opening motions to be simpler and faster, due to the robot being farther from the door and having more space to work with.
Part of Alex's pull door behavior involved ``sneaking'' the left arm in to hold and pull the door open, which was complicated not only by space but also through trying to reduce risk of damage to the Ability Hands.
These items contributed to Alex's door behaviors being significantly slower than Nadia's.
However, we also think that, given some more time, we could speed Alex's behaviors up significantly.

\begin{figure}[H]
    \centering
    \includegraphics[width=.95\columnwidth]{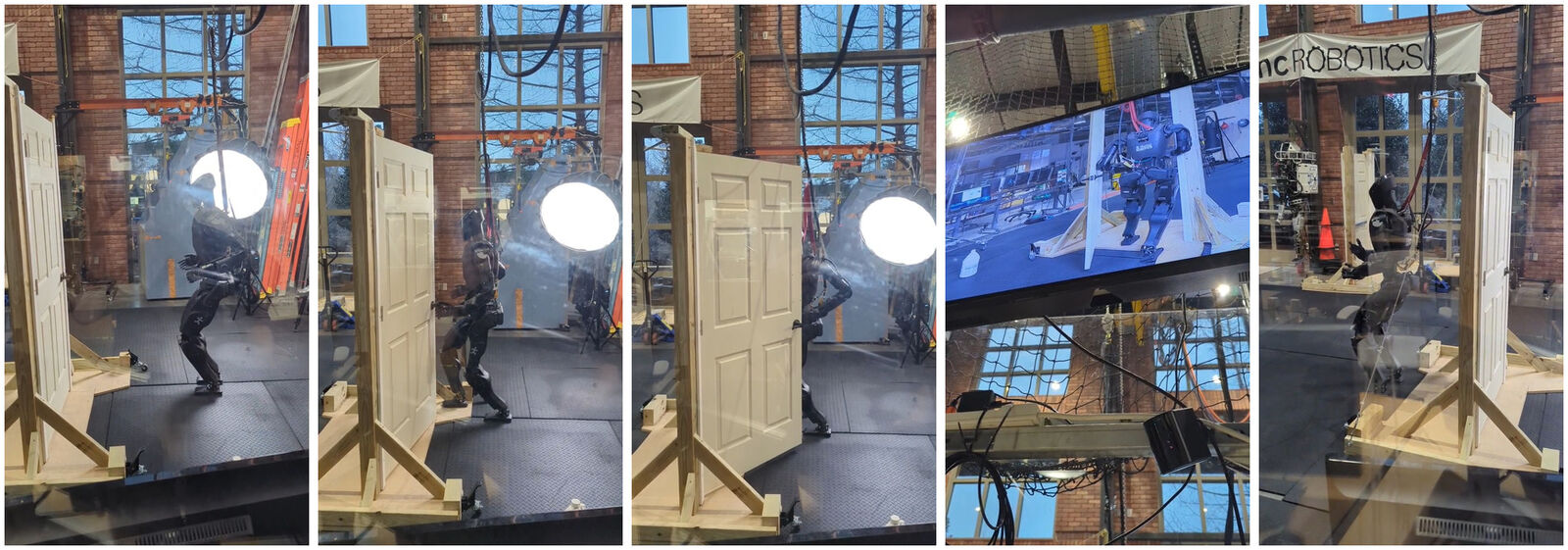}
    \caption{Alex executing its first full door traversal, a pull lever door, on January 24, 2026.
    A video is available at \url{https://youtube.com/shorts/xp6Zs7By5tw}.
    }
    \label{fig:alex_first_pull_door}
\end{figure}

The next day, as shown in \autoref{fig:alex_first_pull_door}, we authored the sequence for walking through the door and ran the whole behavior.
This one was not fully automatic, but it was the first successful door traversal on Alex.
Some difficulty was encountered in the preparation steps for the final traversal steps.
While taking these preparation steps, Alex had to hold the door open with the left arm, as this door had a spring closer.

\begin{figure}[H]
    \centering
    \includegraphics[width=.9\columnwidth]{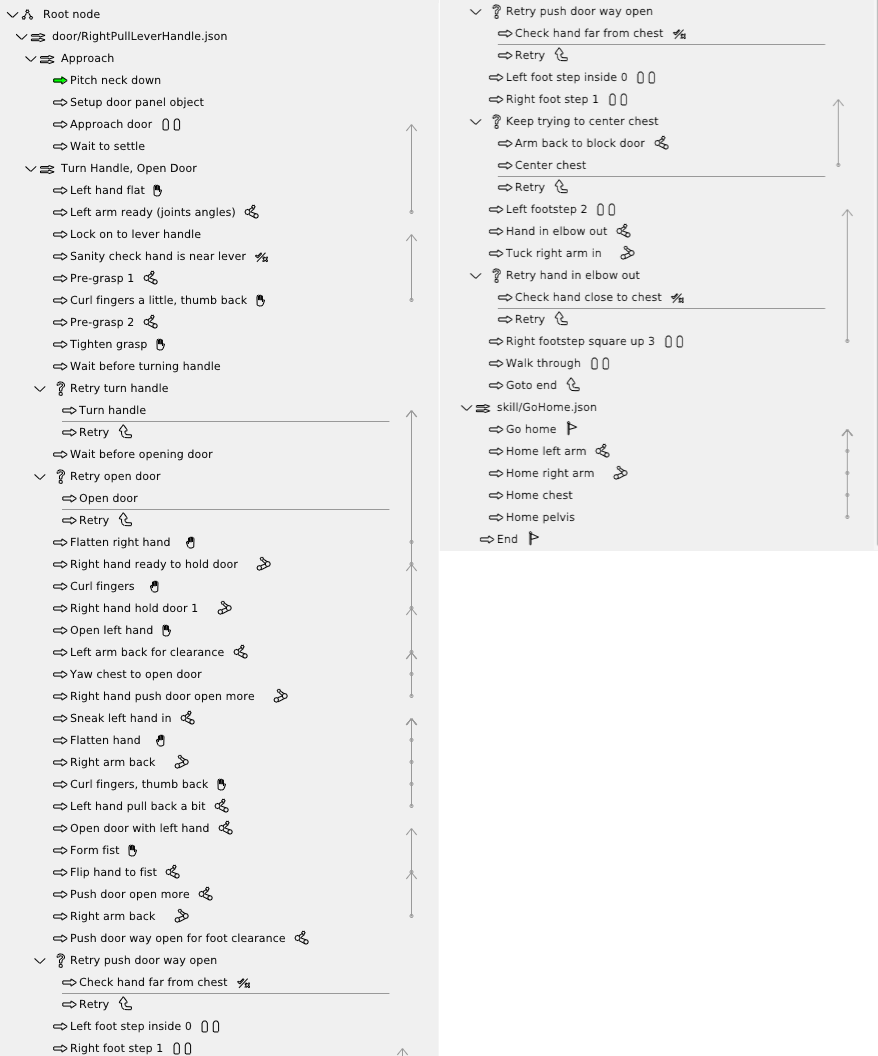}
    \caption{The behavior tree used for Alex's first pull door traversal on January 24, 2026.}
    \label{fig:alex_first_pull_door_tree}
\end{figure}

When we first tried, we had falls for at least two reasons.
One was that the whole body controller was not well tuned on Alex for walking with the arms out, as was required for holding the door open.
Another was that when the robot would take a step toward the door while holding it open, if the holding arm is kept still, when the robot puts its weight on the stance foot, the upper body and arm shifts towards the robot and the foot swing would get caught on the bottom of the door, causing a fall.
The solution was to put the arm farther out before or during the traversal preparation steps.
This solution can be seen in the tree in \autoref{fig:alex_first_pull_door_tree} as the ``Push door way open for foot clearance'' and ``Retry push door way open'' nodes.

\subsection{Shape Contains Condition}
\begin{figure}[H]
    \centering
    \includegraphics[width=.9\columnwidth]{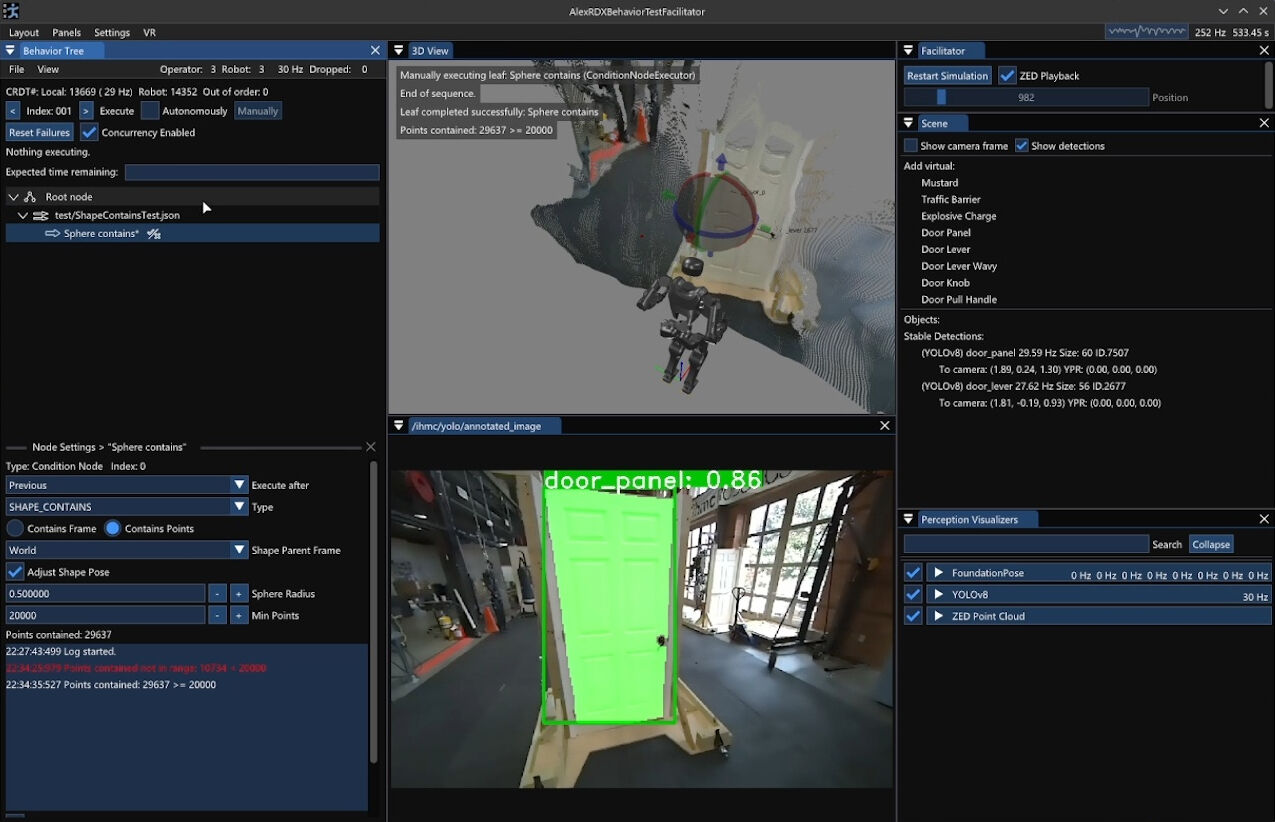}
    \caption{A demonstration of the ``shape contains'' condition node type.
    On the left, a condition node is placed in the tree.
    In the bottom left, the condition node settings area can be seen, with the \texttt{SHAPE\_CONTAINS} type selected.
    The sphere radius, min points, and current number of points contained are displayed.
    In the center, the 3D view shows the sphere, intersecting the door, with a red tint that indicates a high number of contained points.
    A video is available at \url{https://youtu.be/tbTrKuGGmqk}.
    }
    \label{fig:alex_shape_contains_condition}
\end{figure}

The ``shape contains'' condition was implemented on January 27, 2026 and is shown in \autoref{fig:alex_shape_contains_condition}.
This new behavior condition returns success if either a reference frame or some minimum number of points from the point cloud lies within a 3D shape.
We supported just spheres initially which can be sized and placed with respect to any behavior frame, just like the taskspace actions.

For example, we can use it to check if the whole body controller actually achieved a commanded hand goal pose.
Without a fallback node, this condition can be used to stop automatic execution by failing in a sequence.
When combined with a fallback node, it can be used in the ``try'' to branch the sequence of execution.
We use it in this way for the door opening retry mechanism, by placing the virtual sphere where the door panel should be with respect to the robot's chest after opening.
If there are no or few points in the sphere at that point, the fallback catch executes a goto node, returning the behavior back to door handle pre-grasp.
Else, the behavior skips the fallback catch and continues opening and eventually traversing the door.

\begin{figure}[H]
    \centering
    \includegraphics[width=.9\columnwidth]{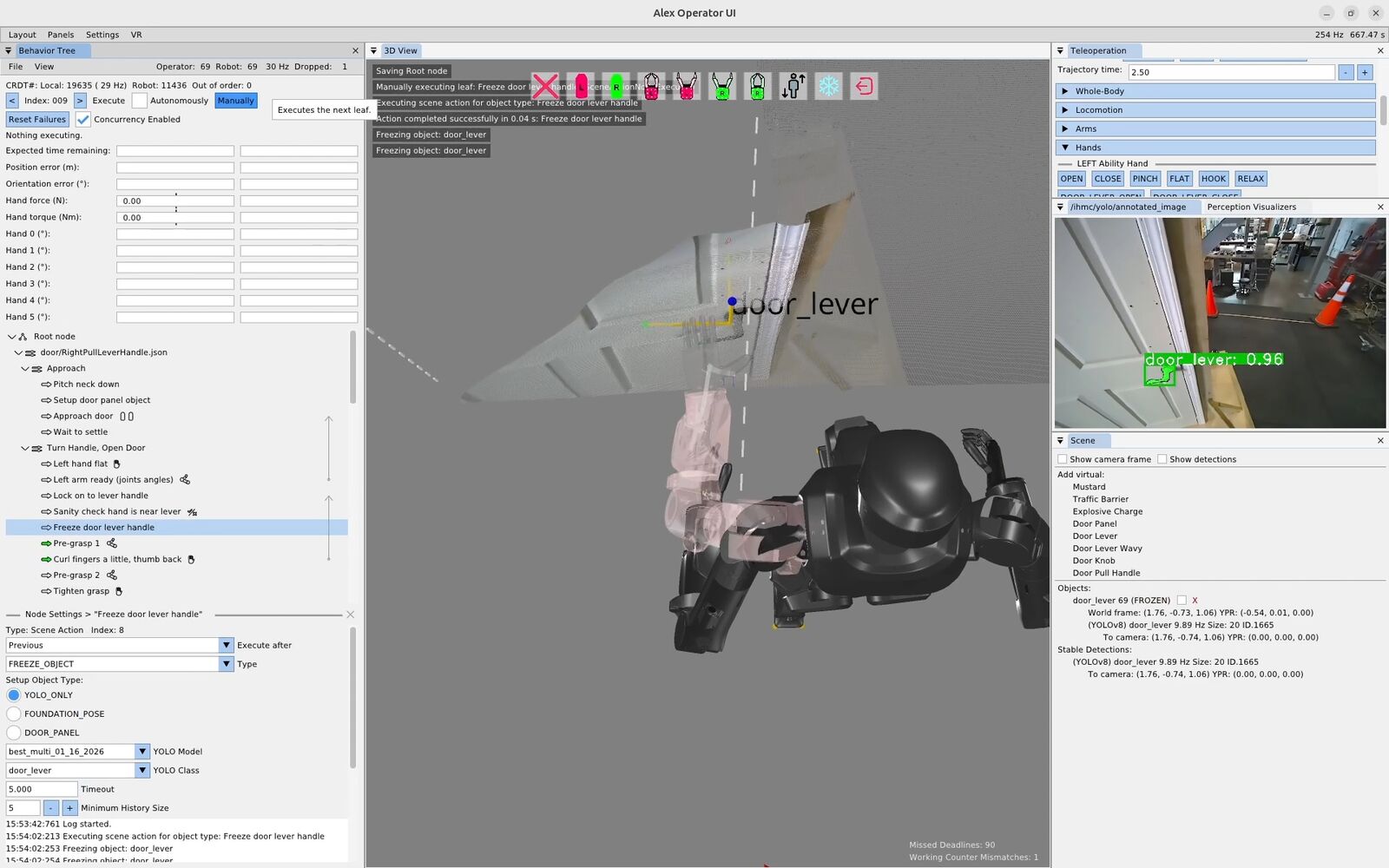}
    \caption{A demonstration of the \texttt{FREEZE\_OBJECT} scene action type.
    On the left, a scene action can be seen in the tree named ``Freeze door lever handle''.
    In the bottom left, the scene action settings area is shown with the \texttt{FREEZE\_OBJECT} type selected.
    In the center, the 3D scene has text overlaid reading ``Freezing object: door\_lever''.
    On the right, in the Scene panel, the door\_lever object indicates ``FROZEN'', meaning its pose will no longer be updated by active tracking.
    A video is available at \url{https://youtu.be/dTM4Rw_912Q}.
    }
    \label{fig:alex_freeze}
\end{figure}

\subsection{Freeze Scene Action Type}
On January 29, 2026, we developed the planned freeze object scene action type, as shown in \autoref{fig:alex_freeze}.
As mentioned previously, the freeze action helps with manipulation by preventing a partial hand occlusion of objects while actively tracking.
By freezing the object frame just before occlusion by the hand, we prevent a corrupted object frame during the grasp.
Another use case for the freeze action is to dead reckon from the frame of an object we saw in the past, as we do for door handles and the door traversal footsteps.



\subsection{Fully-Automatic Repeatable Door Traversals}
On February 12, 2026, we had our first fully automatic pull door traversal on Alex, that is, there were no gaps in autonomous execution from start to finish.
The first run traversed the door in about 45 seconds.
In about 10 minutes of speed-focused behavior tuning, we were able to shave that speed down to around 30 seconds.
This was done by reducing action trajectory durations, reducing wait times, and increasing the concurrency of robot motions.

As an anecdotal statement on the robustness of our Alex pull door behavior, we ran the pull door behavior from these runs again on February 17, without modification, and it worked the first try.
We ran it again on February 18, and it had only one slight arm tolerance issue, which didn't cause a fall, only a brief gap in autonomy, before completing the traversal successfully.
This was a good indication that our behaviors worked independently of slight variations in environmental conditions, such as careful placement of the lab door and the natural lighting coming through the windows, which varies based on time of day.


\subsection{From-Scratch Push Door Authoring}
\begin{figure}[H]
    \centering
    \includegraphics[width=.95\columnwidth]{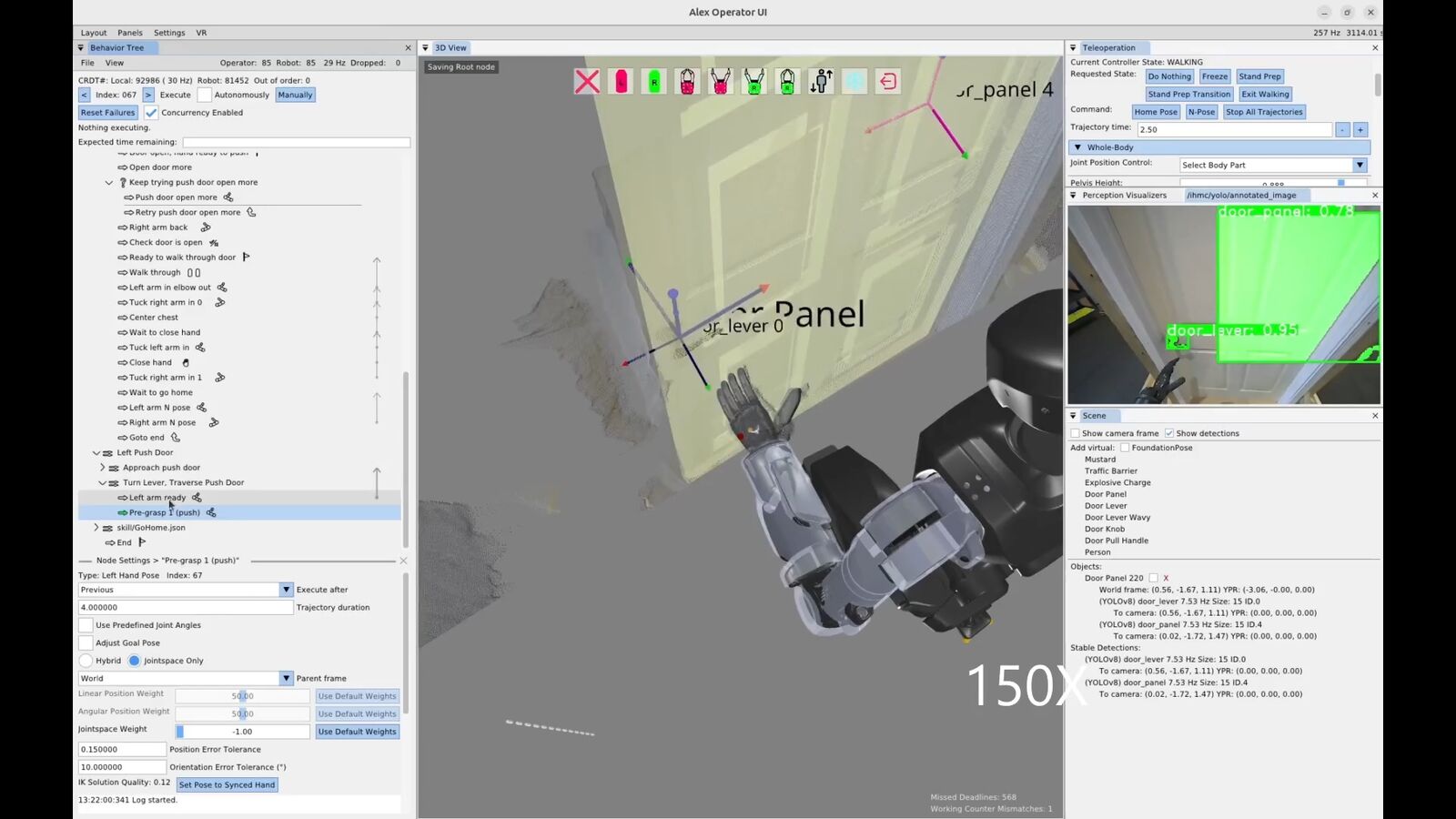}
    \caption{A screenshot from the authoring session on February 22, 2026 for the push door behavior.
    A video is available at \url{https://youtu.be/bLHQmV4GEFA}.
    }
    \label{fig:alex_push_door_authoring}
\end{figure}

On February 22, 2026, we authored our first push door traversal behavior on Alex in under 2 hours.
A screenshot from the session is shown in \autoref{fig:alex_push_door_authoring} and the timeline of authoring is documented in \autoref{tab:left_push_scratch_authoring_timeline_story}.
The push door traversal executed in 21 seconds.

\begin{table}[H]
    \caption[Scratch left push door authoring milestones.]{Normalized scratch-authoring timeline for the February~22,~2026 left push door behavior, from creation of the \texttt{Left Push Door} sequence to first automatic success.}
    \centering
    \footnotesize
    \setlength{\tabcolsep}{4pt}
    \renewcommand{\arraystretch}{0.97}
    \begin{tabular}{>{\RaggedLeft\arraybackslash}p{1.9cm} >{\RaggedRight\arraybackslash}p{\dimexpr\textwidth-1.9cm-6\tabcolsep\relax}}
        \hline
        Elapsed time & Authoring milestone \\
        \hline
        0:00:00 & Create the \texttt{Left Push Door} sequence. \\
        0:14:33 & Door-panel scene action and squared-up pre-approach stance added. \\
        0:24:22 & Staggered approach stance executed. \\
        1:21:14 & First handle turn. \\
        1:24:37 & Door opened. \\
        1:39:21 & Right arm pushes door fully open. \\
        1:49:24 & Full traversal footstep set authored. \\
        1:50:11 & Traversal taken; arm task-space error sends the right arm backward. \\
        1:51:13 & Right-arm fix: bring the arm in for traversal. \\
        1:58:15 & \texttt{shape contains points} stop condition added if the door does not open. \\
        1:59:48 & First fully automatic push-door success. \\
        \hline
    \end{tabular}
    \label{tab:left_push_scratch_authoring_timeline_story}
\end{table}

\subsection{Quick Footstep Planner}
Around this time, we started developing the Quick Footstep Planner to more reliably get footsteps for task approaches on flat ground.
This footstep planner uses a procedural geometric heuristic instead of a search algorithm.
This helped us by providing an alternative to the existing A* and turn-walk-turn planners.
Since the author is able to choose the planning type for each walk action, alternative planning options increase the chance of finding a workable solution.

\subsection{Node Duplication Options}
On February 28, 2026, we introduced a duplication option for all behavior tree nodes.
This allows the operator to right-click any node and select ``Duplicate'' to create a copy.
This is useful to speed up authoring, as the author can mutate an existing node if a similar one is needed.
Copying and mutating an existing node is often faster than creating a new one from scratch, especially for nearby nodes in a similar context.

\subsection{New Behavior Node Icons}
\begin{figure}[H]
    \centering
    \includegraphics[width=.7\columnwidth]{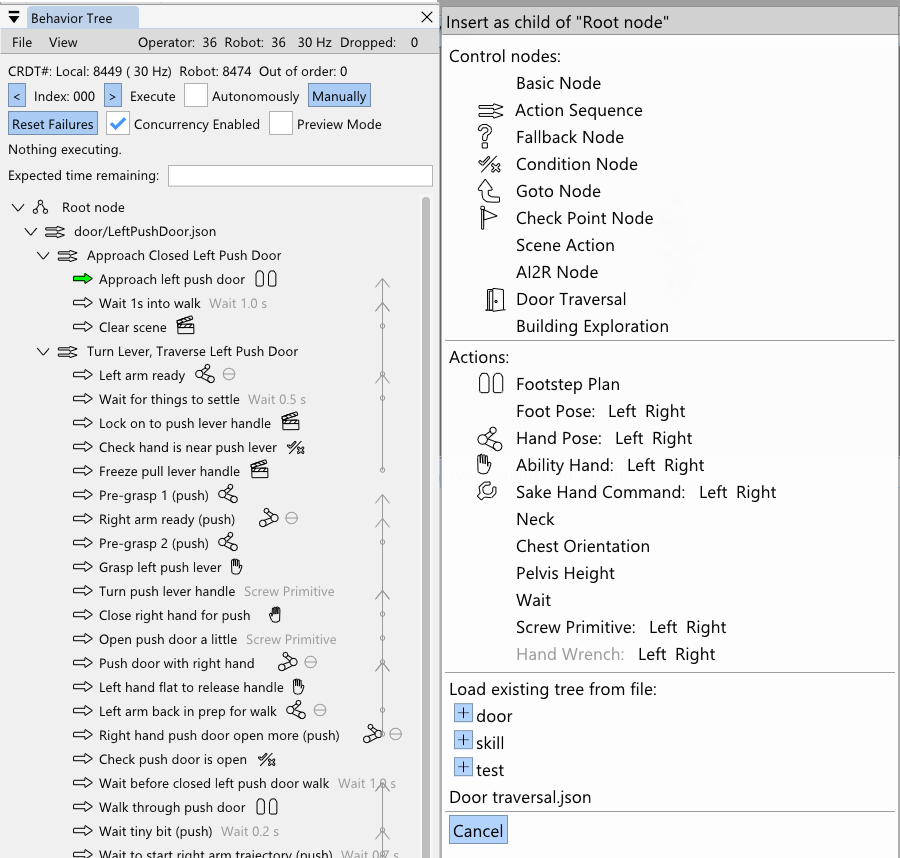}
    \caption{New icons for behavior tree nodes on March 3, 2026.}
    \label{fig:alex_behavior_tree_icons}
\end{figure}

On March 3, 2026, we introduced some new icons, shown in \autoref{fig:alex_behavior_tree_icons}.
At this point in time, the most common actions had an icon.
Icons are useful when working with the behavior tree, because they help you locate things much easier and also give at-a-glance verification of node type.
This takes the burden off the behavior author to include the action type in the name, which also helps reduce the amount of text in the view.

\subsection{Node Referencing Improvements}
In March, we made a modification that allows the ``execute after'' field of actions and the ``node to goto'' field of the goto node to point to non-leaf nodes.
This helps avoid needing to create checkpoint nodes just for the purpose of creating a concurrent sequence.
It also preps the goto node to eventually become a ``gosub'', where it can point to a sequence with the guarantee that control will be returned when that sequence is complete.

\subsection{Jointspace Indicator}
On March 3, 2026, we added a little theta indicator to arm actions when it is defined as joint angles rather than a taskspace pose.
This was important because we would often tune the arm configurations using taskspace and forget to switch it back to jointspace.
For arm configurations that are frame invariant, such as door frame avoidance or table surface avoidance configurations, the joint angle definition is important.
However, after tuning these configurations, if the arm action definition was accidentally left in taskspace in world frame, when executed later, the arm would go crazy trying to reach whatever point in world we were at during authoring.
We made the default frame of arm actions chest frame to mitigate this issue, but the visual indicator helps the operator to verify this important action option at a glance.
Since jointspace actions are generally safer to execute, it also gives the operator some confidence in executing the action.

\subsection{Improved Frame Names}
On the same day, we cleaned up the default frame names available to behavior actions.
They were now more human readable like ``Left Hand'' instead of ``afterGripperZLeft'' and ``Pelvis'' instead of ``afterPelvisLink''.
This could help new operators climb the learning curve faster and help reduce cognitive burden for expert operators.



\subsection{Reactive Pull Door Behavior}
On March 9, 2026, we conducted a reactivity test of the pull door behavior with Alex.
In addition to the fallback node for retrying the door opening, we added one that waits for the doorway to be clear of obstacles before walking through.
The test was successful.
The robot was disturbed while opening the door three times and retried each time, succeeding on the fourth try.
Then, the robot waited for the human to move out of the doorway before starting the walk-through.
Unfortunately, the robot fell during the walk-through due to loss of balance, but the reactivity test was a success.


\subsection{Preview Mode and Nominal Frames}
Earlier in March, we implemented a behavior ``Preview Mode'', which allows the operator to execute the full behavior against a kinematics simulation robot instead of the real one.
On March 12, 2026, we developed a new element to the scene action ``setup object'' type node called the ``nominal frame''.
This nominal frame field specifies a pre-defined pose of the object for use in preview mode.
At this point in time we do not have a photo-realistic or physics-accurate simulation of the robot, but we wanted to preview the robot's motions when doing a full loco-manipulation behavior.
When the behavior is running in preview mode, setup type scene actions will use a JSON-saved nominal pose to place the object in the scene at that location.
This enables the behavior to run through its nominal motions without perception.
This feature turned out to also be useful for recording reference motions for training reinforcement learning mimic behaviors.






\subsection{Collision Avoidance with RRT-Connect}
On March 20, we introduced a navigational mode for the Quick Footstep Planner which uses RRT-Connect~\cite{kuffner2000rrt} to avoid YOLO detected obstacles.
It plans from the start to the goal but avoids any YOLO objects by maintaining a radius of avoidance.
We didn't end up using it in an actual behavior yet.
We think it might be better to use a capsule point check with the point cloud to make it more general.
However, RRT-Connect may be too slow when the collision checks have to query the point cloud.
It may be a good application of an occupancy map.

\subsection{Behavior Timeline}
\begin{figure}[H]
    \centering
    \includegraphics[width=.95\columnwidth]{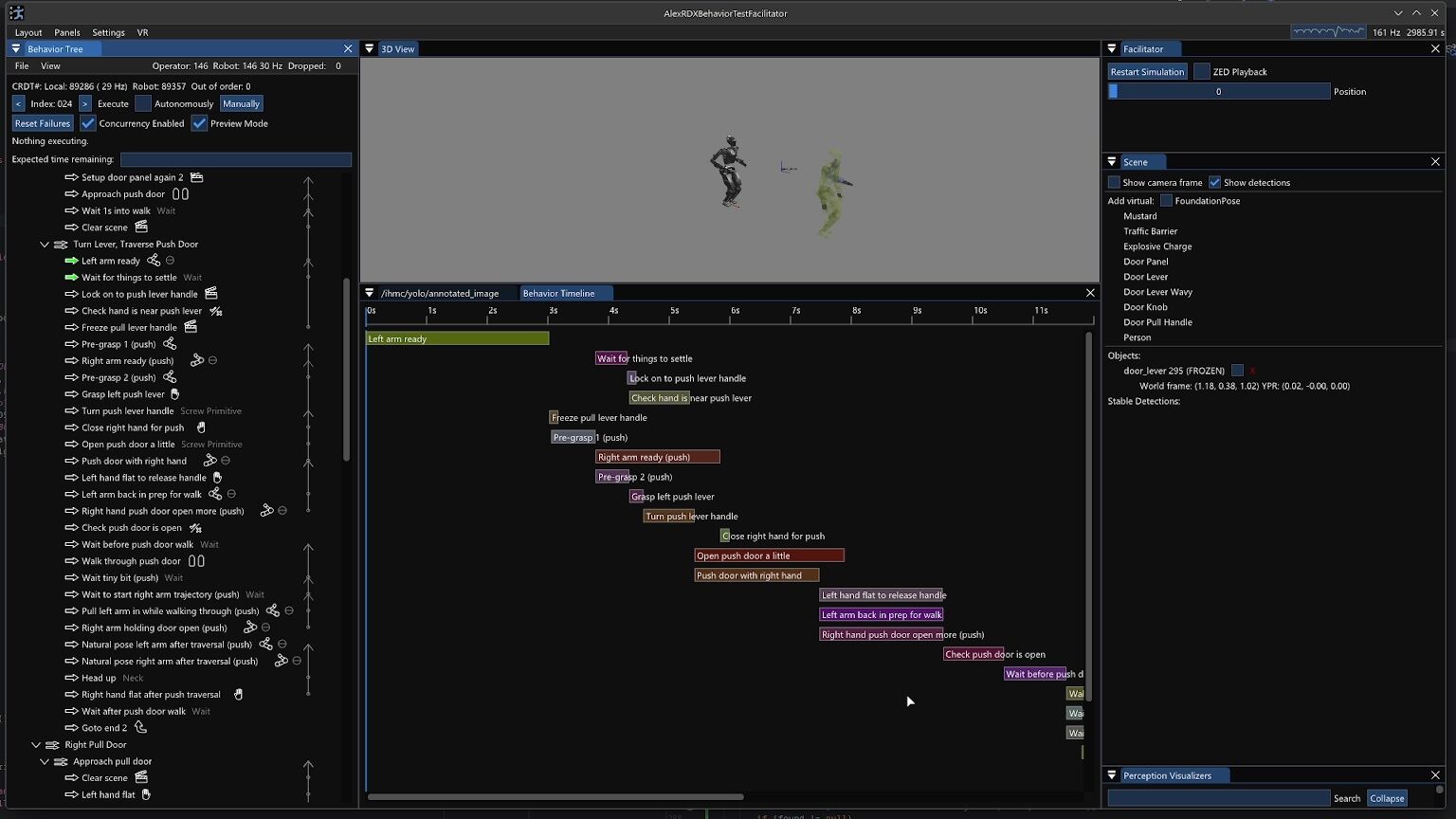}
    \caption{A view-only behavior timeline implementation from March 22, 2026.
    In the bottom center, the horizontal bars show the concurrency and durations of actions from a portion of our push door behavior as previewed in simulation.
    A video is available at \url{https://youtu.be/QmiHA9GlQjY}.
    }
    \label{fig:alex_behavior_timeline}
\end{figure}
We had talked over the years about how to do a video-editor-like horizontal bars implementation where the behaviors can be viewed by concurrency, start time, and end time.
On March 22, we took a stab at an initial behavior timeline implementation as seen in \autoref{fig:alex_behavior_timeline}.
One immediate issue with rendering such a view is that action timings are not always known ahead of time.
They are often dependent on real-time events, such as the scene node obtaining a stable detection.
When the behavior is run in preview mode or regular mode, the actual action durations are stored in the bars, to provide a retrospective on the behavior's actions over time.

\subsection{Action and Subtree Mirroring}
In trying to build out our general door traversal behavior, we needed mirrored versions of our right pull and left push door behaviors.
This became a priority on March 25, 2026 because we were about to attempt our first real-world door traversal -- our break room door.
Our break room door was a left pull door.
Having added the duplicate node feature back in February, we decided to implement a ``Mirror'' operation as a similarly general feature.
Some of our actions were able to be mirrored between left and right invariantly: jointspace arm actions, neck yaw actions, and spine yaw actions.

Other actions required additional information to mirror such as the door-relative approach footsteps.
To address our immediate needs, we simply hard-coded a door specific mirror option for five of our action types with a ``Mirror (door)'' option in addition to the invariant ones.
This door-specific mirroring option applies to condition nodes, scene action nodes (for the nominal object poses), arm action nodes, screw primitive action nodes, and walk action nodes.

We also added the option to mirror subtrees, i.e. (``Mirror Subtree'' and ``Mirror Subtree (door)''), which would try its best to perform the mirroring on the subtree recursively using the functionality we just described on each node or no mirroring if it was not covered.
This actually worked really well to create a mirrored version of the pull door behavior which we then took to the break room door to try it for the first time.
We were able to preview the mirrored behavior in simulation to verify the general motions before deploying the real-robot.

\subsection{Traversal of a Real Door}
\begin{figure}[H]
    \centering
    \includegraphics[width=.95\columnwidth]{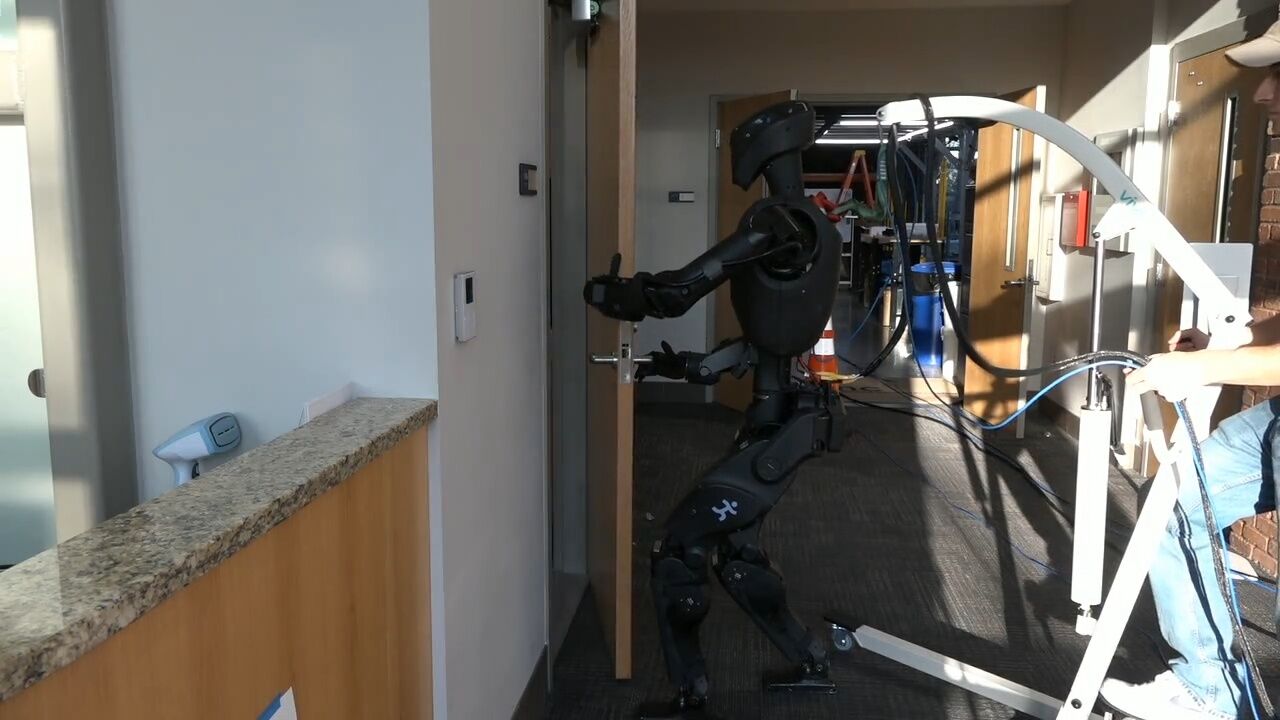}
    \caption{Alex opening the break room door on March 26, 2026.
    A video is available at \url{https://youtu.be/sv6r1ZBn6FA}.
    }
    \label{fig:alex_break_room_door}
\end{figure}

On March 26, we attempted to traverse the break room door, which would be the first time an IHMC robot has traversed a real-world door as opposed to a lab-constructed one.
In less than two hours of tweaking the behavior and attempting the behavior, we got a full door traversal to execute successfully in about 33 seconds.
A frame from this run is shown in \autoref{fig:alex_break_room_door}.

\subsection{Ball Pick and Place}
\begin{figure}[H]
    \centering
    \includegraphics[width=.95\columnwidth]{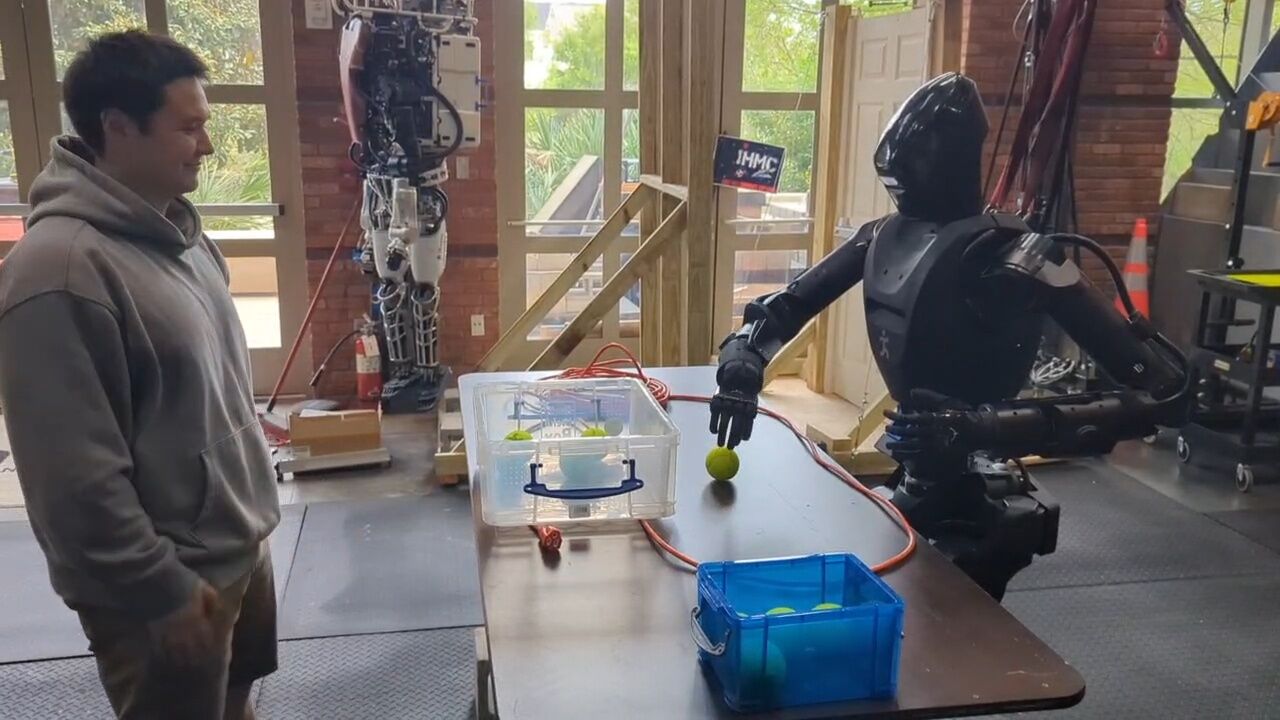}
    \caption{Our first ball pick and place behavior on March 31, 2026.
    A video is available at \url{https://youtu.be/hq0y2LSdQ4A}.
    }
    \label{fig:alex_jared_ball_pick_and_place}
\end{figure}
In late March 2026, we decided to focus on demonstrating the versatility of our approach by manipulating objects on tables.
Since FoundationPose tracking was not working so well at this time, we picked balls as an object to manipulate since the grasp of a ball does not depend on the ball's orientation.
Since we did not have any orientation-invariant graspable objects in our in-house trained YOLO models, we had to find a generally available model to use.
We picked the default ``yolov8n-seg'' model that comes out of the box with YOLOv8.
It has a ``sports ball'' object class that just barely worked for our use case, detecting our colored tennis balls and baseballs with widely varying confidence levels.
Our balls could be detected with high confidence in the 70-80\% range, but we could only rely on the confidence levels to be above 2\% or so, as they were very often that low.

As seen in \autoref{fig:alex_jared_ball_pick_and_place}, on March 31, 2026 we had our first ball pick and place behavior running.
We used our ``sphere contains'' condition check to only pick up balls that were in a reachable region on the table, to avoid catastrophic unreachable grasp attempts.
This version would often miss the grasps and had some unnatural looking arm configurations and slow trajectories.
We also had some issues with the reliability of the YOLO persistent detections.
Another problem was that we could not detect the storage containers and the balls at the same time because they were supported by two different YOLO models.

\subsection{Round-Robin YOLO Model Inference}
By April 3 2026, just three days later, we had worked through a lot of these issues and achieved a much more resilient behavior.
To start, we corrected unnatural arm configurations and sped up the motions, so the behavior was faster.
Secondly, we added the ability to run more than one YOLO model at a time, round-robin style, so we could perceive the balls and the container at the same time.
Thirdly, we implemented YOLO model and persistent detection settings management to the scene action node.
This meant that the behavior author could decide which YOLO models were running, the enabled object classes within them, and the confidence thresholds for specific object classes for both the YOLO model and the persistent detections.
This meant that the behaviors could configure the sports ball class to allow persistent detections of very low confidence, such as the 2\% setting we used, to get much higher reliability.

\subsection{Reactive Colored Ball Sorting}
\begin{figure}[H]
    \centering
    \includegraphics[width=.95\columnwidth]{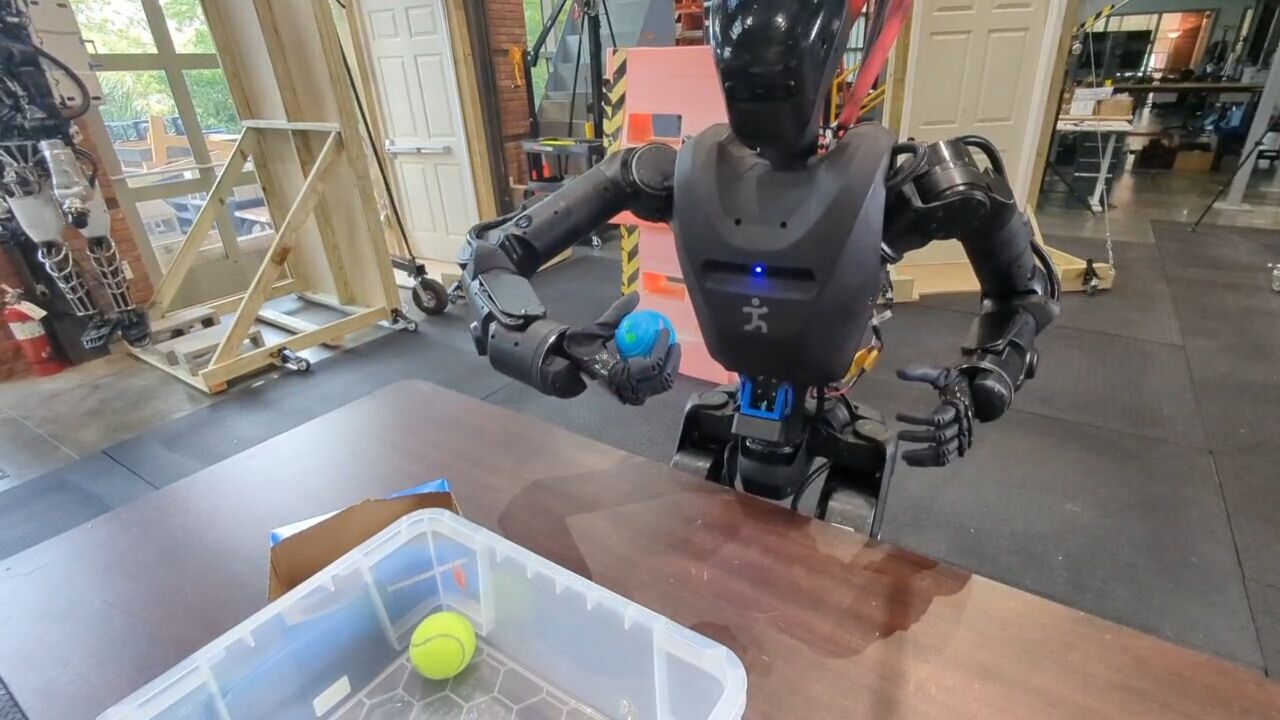}
    \caption{A reactive and robust ball sorting behavior on April 4, 2026, in which the robot sorted 5 balls successfully while gracefully handling a human-caused disturbance.
    A video is available at \url{https://youtu.be/9DhgbjIRynM}.
    }
    \label{fig:alex_reactive_robust_ball_sorting}
\end{figure}
In order to demonstrate online decision-making and reactivity, we then sought to sort the balls by color into separate containers.
We also planned, for our annual robotics lab open house where we share our work with the public, to trigger specific behaviors based on ball color.
For example, picking up a green ball would cause the robot to go and deliver that green ball to a box on a table beyond a door.
On April 4, 2026, as seen in \autoref{fig:alex_reactive_robust_ball_sorting}, we achieved a demonstration in which the robot successfully sorts five balls by color into two different containers, while successfully handling a disturbance where a ball was removed from the table just before the grasp.
In this demo, the behavior was authored to pick up the balls and inspect them while still in hand for grasp success and color.
If there were no points within the hand, we assumed the grasp failed and returned to pre-grasp.
If there were points in the hand, we checked if they were yellow, in which case they would be placed in container A, else they would be placed in container B.

\subsection{RL Mimic Action}
Around this time we were also exploring options to increase the reliability of our door traversal walk-throughs, as the robot would often lose balance during them.
Since we had been working with RL mimic policies for boxing and martial arts at the time, we decided to try to incorporate the door walk-throughs as a robustified RL mimic policy trained on the whole-body motion preview available in the behavior system.
We trained push and pull door policies and integrated a behavior node called the ``Mimic action'' which would transition from our model based controller to the mimic controller, perform the mimicked motion, and transition back to the model based controller.
We haven't yet run the mimic policy for the door traversals at the time of writing, but we did train and demonstrate several dance moves as behaviors triggered by picking up different colors of balls, such as an ``I love you!'' dance, a ``Dab'' dance, and stretching dance.

\subsection{Hybrid and Composite Frames}
\begin{figure}[H]
    \centering
    \includegraphics[width=.95\columnwidth]{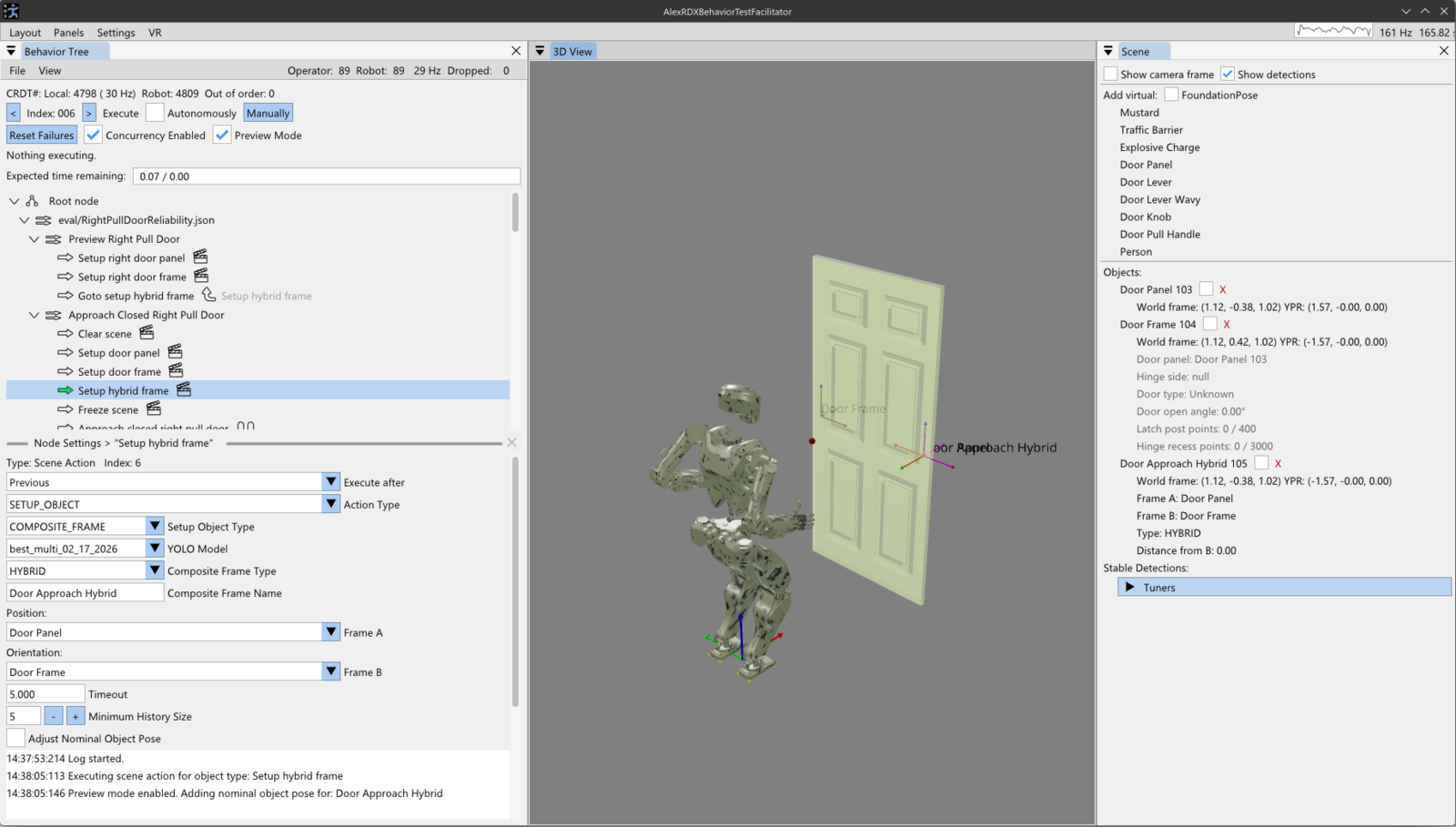}
    \caption{The composite frame feature.
    On the left, a series of scene actions can be seen: Clear scene, Setup door panel, Setup door frame, and Setup hybrid frame.
    On the bottom left, the settings for the composite frame scene action type can be seen.
    In the settings, the hybrid frame is named ``Door Approach Hybrid'' and takes the position of the Door Panel object and the orientation of the Door Frame object.
    All three of the resulting scene objects can be seen in the scene panel on the right.
    This example uses nominal object poses in simulation.
    }
    \label{fig:alex_composite_frame}
\end{figure}
Around this time we ran into a theoretical limitation with approaching tasks from a distance.
Given that perceptual capability degrades with distance to a task and we first just need to navigate and approach it, we need a reference frame to define the approach in.
Since the pose estimation of the object has the object's orientation, it does not work as a frame to specify the approach in.
Nominally, we would want the robot to approach the task by taking the shortest path from the robot's current location to the task.
Since we don't want to run into the task, we need to ``back off'' the approach stance from the object toward the robot.
To do this, on April 4, 2026, we decided to add a type of behavior scene object called a \textit{composite frame}.
This composite frame would exist as a privileged object in the behavior scene such that it would be usable in the same way to define actions.

There are currently two types of composite frames: an approach frame and a hybrid frame.
They are both generalized to be named frames as a derivative computation of two pre-existing behavior frames.
In \autoref{fig:alex_composite_frame}, this feature is shown.
The approach frame's orientation is defined to face from frame A to frame B and its position is defined to be on the line segment from frame B to frame A at some tunable distance from frame B.
This makes the approach frame suitable for walking directly towards an object, but stopping before getting too close.

The hybrid frame is similar.
It takes the position of frame A and the orientation of frame B.
The hybrid frame is useful for approaching ajar door panels, where you want to approach with the orientation of the door frame but the position of the door opening mechanism.

Composite frames also can be layered.
For example, the ajar door hybrid frame could be used in a subsequent approach frame to approach an ajar door's handle from a distance.
This composite frame mechanism is designed to be extended and generalized further based on encountered real world applications.

\subsection{Crucial Scene Bugfix}
On April 7, 2026, we fixed a pretty major race condition bug between the scene actions and the behavior scene.
We had been having to put wait durations after scene actions because the bug would cause the subsequent physical actions to use outdated scene object frames when run in automatic mode.
We fixed this through safer and more thorough scene management and synchronization.
The behavior tree and scene are managed on the same thread, but it took multiple update ticks for the scene object's pose to reflect the updated persistent detection's pose.
We mention this bug fix because it marks a leap in trust of the system, prevents user frustration, and avoids robot damage from actions reaching to stale, unreachable object poses.

\subsection{Approaching Tables with Active Perception}
\begin{figure}[H]
    \centering
    \includegraphics[width=.95\columnwidth]{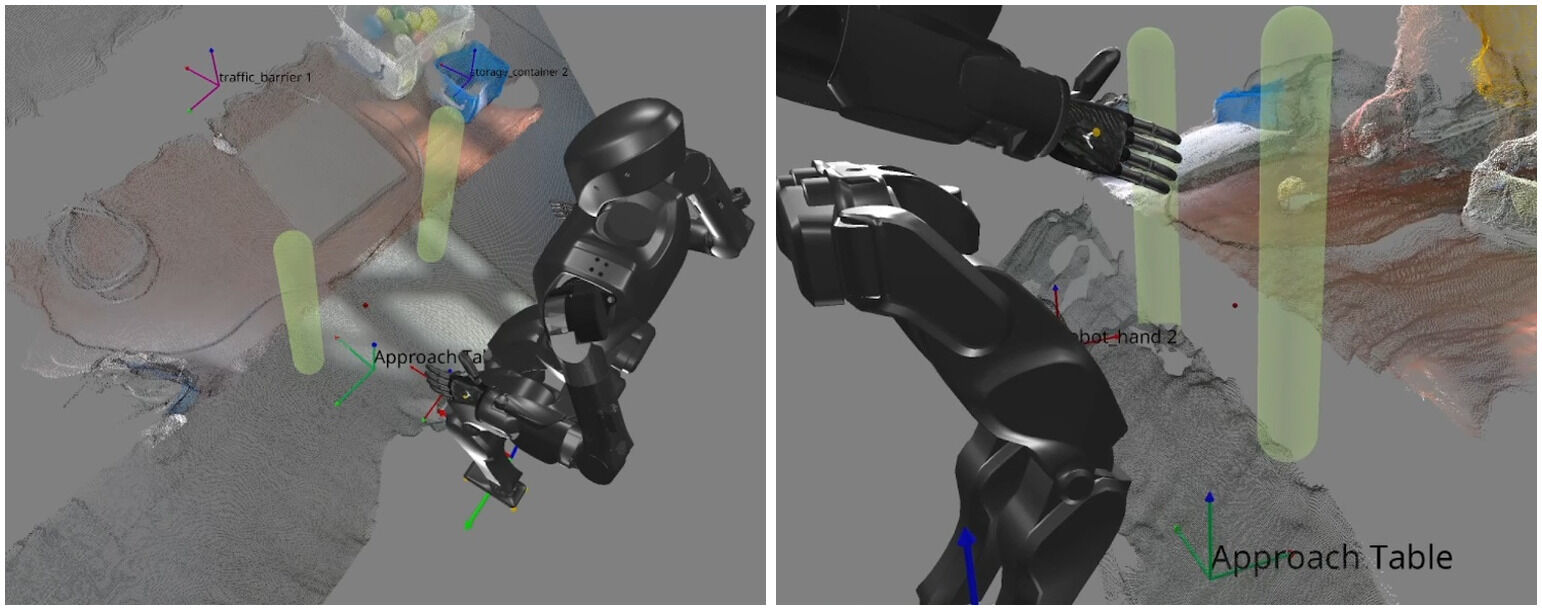}
    \caption{The table approach algorithm visualization.
    The approach reference frame available to the physical actions is seen on the floor between the capsules.
        A video is available at \url{https://youtu.be/iBPPJW1Pbmc}.
    }
    \label{fig:alex_table_approach}
\end{figure}
In the same week we developed a novel algorithm for humanoid robot table approach.
Using our point-in-shape counting CUDA kernel, we designed a special heuristic scene object called the ``Approach Table'' object.
This was similar in spirit to the heuristic door panel object discussed previously, but differs in that it does not use any semantic detections.
Instead, we sweep two vertical capsules forward from the robot's hips with the intention of colliding with the table's edge, as shown in \autoref{fig:alex_table_approach}.
A tunable threshold for the number of points to be considered constituting the table edge is defined in the settings of this type of scene node.
Typically a value of 300-400 points seemed to be good.
The capsules start around knee height and end just below chest height with the intention of handling tables of various heights.
When a capsule collides with the table, it stops the forward sweep.
The two capsules sweep independently.
The result is that a line segment in the X-Y plane is now identified as the table edge.
This line segment and the current stance height are used to form a reference frame on the ground with the orientation of the table edge.
This reference frame can then be used by a subsequent walk action to perform a squared-up approach to the table edge.

Anecdotally, we found this technique to be very reliable and able to approach our tables within a few centimeters of accuracy.
This was also an important capability milestone for our behavior system.
The ability to approach tables with this degree of accuracy is a necessary part of obtaining the reachability of the items on the table and avoiding failures caused by running into the table.

\subsection{High-Volume Pick and Place Demo}
\begin{figure}[H]
    \centering
    \includegraphics[width=.95\columnwidth]{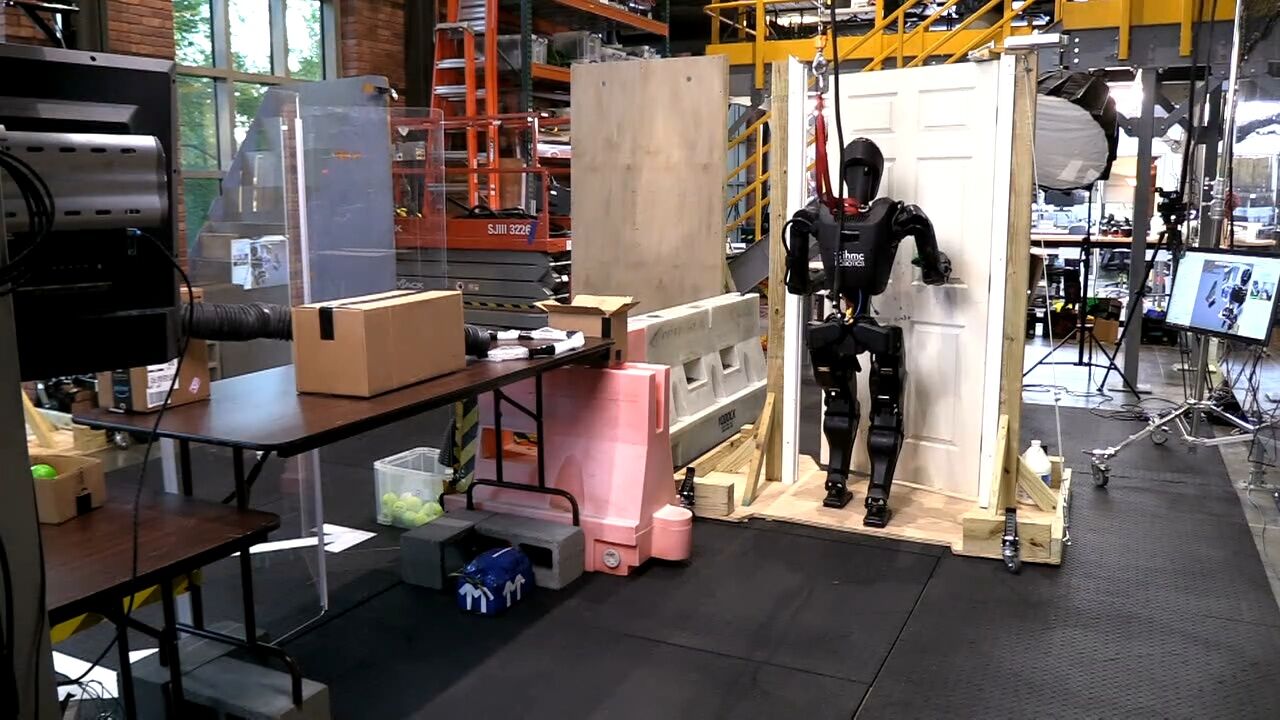}
    \caption{A still frame from our open house demo behavior preparation on April 9, 2026, where Alex has picked up a green ball from the table and is executing a door traversal behavior as part of the delivery sub-behavior.
    A video is available at \url{https://youtu.be/NmmnQui-uu0}.
    }
    \label{fig:alex_open_house_setup}
\end{figure}

On April 9 and 10, 2026, we designed and rehearsed a demo for our open house where the robot would have a table station where balls of different colors would be fed to the robot through a tube system.
The robot would be tasked with picking up the balls, determining their color, and executing a specific behavior for each color, in an infinite loop.
Yellow balls would simply be placed in a chute to deliver the ball back to the visitor's side.
Green balls would trigger a delivery behavior where the robot would traverse a door and deliver the green ball to a box on a table beyond that door, as presented in \autoref{fig:alex_open_house_setup}.
The rest of the colors would trigger different RL mimic dances.
For example, red balls would trigger the ``I love you!'' RL mimic dance.

We did some partially successful rehearsals of this demo, but on April 10, 2026, just before the demo, the robot's legs fell off its body in a dramatic fall.
We were able to repair the robot, but avoided doing any more walking that day.
During the demo, we picked and placed 199 balls at the table station.

\subsection{Multi-Station Reactive Ball Sorting}
\begin{figure}[H]
    \centering
    \includegraphics[width=.95\columnwidth]{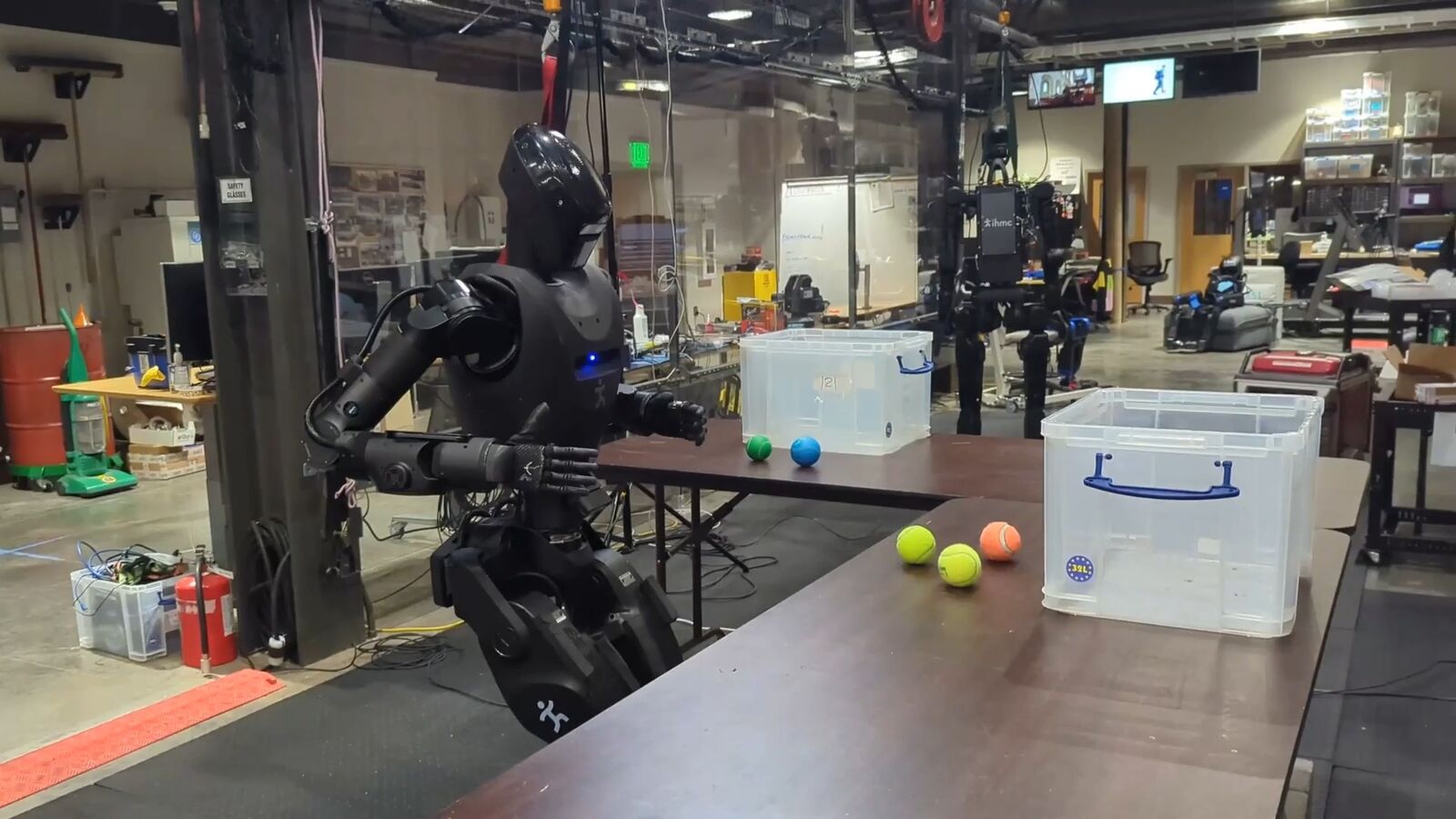}
    \caption{The two-station ball sorting task scenario, as filmed on April 4, 2026.
    2 yellow balls, 1 orange ball, and container A sit on table A and a green ball, a blue ball, and container B sit on table B.
    A video is available at \url{https://youtu.be/KxIihoDiMtc}.
    }
    \label{fig:alex_two_table_sorting}
\end{figure}
Our last day of testing before the time of writing this thesis yielded some important results.
On April 14, 2026, we extended our reactive and robust ball sorting behavior to a multi-station ball sorting behavior.
The intention behind this demonstration was to show a compelling loco-manipulation task beyond door traversals.
The robot and behavior author were tasked with sorting colored balls between two containers on two different tables, requiring walking between the tables.
A still frame from this behavior is shown in \autoref{fig:alex_two_table_sorting}.
In a 1 hour and 50 minute time period we were able to extend the stationary sorting behavior to multi-station sorting.
In the final demonstration run, the robot sorted 9 balls correctly, with three table approaches including two table-to-table transitions.
However, the robot did not perceive the containers and there was one pause where the human experimenter had to shift one of the balls to get YOLO to detect it.

\subsection{Repeated-Run Reliability Tests}
On that same night, we conducted two reliability tests for door approach and opening on both push and pull door variants.
We were able to achieve 11 push door approach and openings in a row and 12 in a row for the pull door.
This final demonstration in our story marked a reliability milestone in loco-manipulation.

\section{Reflections on a Decade of Development}

Over the course of nearly a decade, we think we made dramatic progress in robot autonomy and capability using our behavior system.
In the DARPA Robotics Challenge Finals in 2015, the most successful teams had expert roboticists crowded around operator interfaces, meticulously managing every detail of an hour-long 8-task run.
In 2026, we are able to somewhat casually author and run neglect-tolerant loco-manipulation behaviors with just a few expert roboticists.
In addition, any video you will see of the DARPA Robotics Challenge is sped up by 5, 10, or 30 times realtime speed.
Our automatic loco-manipulation behaviors are watchable and interesting when played at 1x speed.

In the 2016-2021 era with Atlas, behaviors were hard-coded, developed features in themselves and only used perception for locomotion.
In 2021, we introduced Robot Data eXplorer (RDX), a novel development tool suite for data visualization, algorithm development, and robot operation.
In 2022, we invented the runtime-editable behavior sequence for humanoid robots, which allowed for an explosion of the size of our behavior library, given they could be authored in an operator interface instead of being hand-engineered with code.
In 2024, with Nadia, we made the behaviors fast and started using perception for manipulation, executing some of the fastest door traversals ever done by humanoid robots and weaning our dependence on ArUco markers to direct environmental perception.
In 2025, with H1-2 and the PSYONIC Ability Hand, we upped our game with 5-finger anthropomorphic manipulation.
Finally, in late 2025 and early 2026, with Alex, we made perceptual reactivity runtime-editable and demonstrated the ability to author novel loco-manipulation behaviors in hours.

A few items of significance in the timeline are of note.
Firstly, in 2025, we had decided to explore vision-language-action models starting in March.
By October, we hadn't gotten it working, so we decided to abandon the effort.
What's interesting is the impressive pace of development that followed.
The October 2025 - April 2026 period yielded some of the most significant advancements in the entire system, such as the points-in-shape counting capability.
Also in that time period, the advancement of manipulation sky-rocketed, from the mustard picking with H1-2, to the 199 ball sorting open house demo, to the 9/9 multi-station colored ball sorting demo.

The pace of development is currently very strong.
We are extremely proud of this work and the very large number of team members that contributed to this system in the last decade did an incredible job.
At the same time, we are excited for the future.
We want to see this system extended further and deployed on our new tetherless version of Alex for real-world use cases, not just tech demos.

In the next chapter, we'll describe the current architecture of the behavior system.
Then, in \autoref{ch:tutorial}, we'll dive into hands-on examples of creating behaviors using our system.



\chapter{Current Architecture}
\label{ch:architecture2}

In \autoref{ch:building}, we told the story of building our system and went over the components in some detail.
In this chapter, we'll cover the current structure again to round off the full current design.

Our central design choice is to enable the robot to be autonomous but utilize human expert knowledge to set up the autonomous behavior.
We do this by including a human operator in the loop through tight synchronization between a user interface and the robot.
Once behaviors are set up sufficiently, our architecture, which keeps all perception and control on board the robot, allows the robot to function autonomously.

\section{Domain}
\begin{figure}[H]
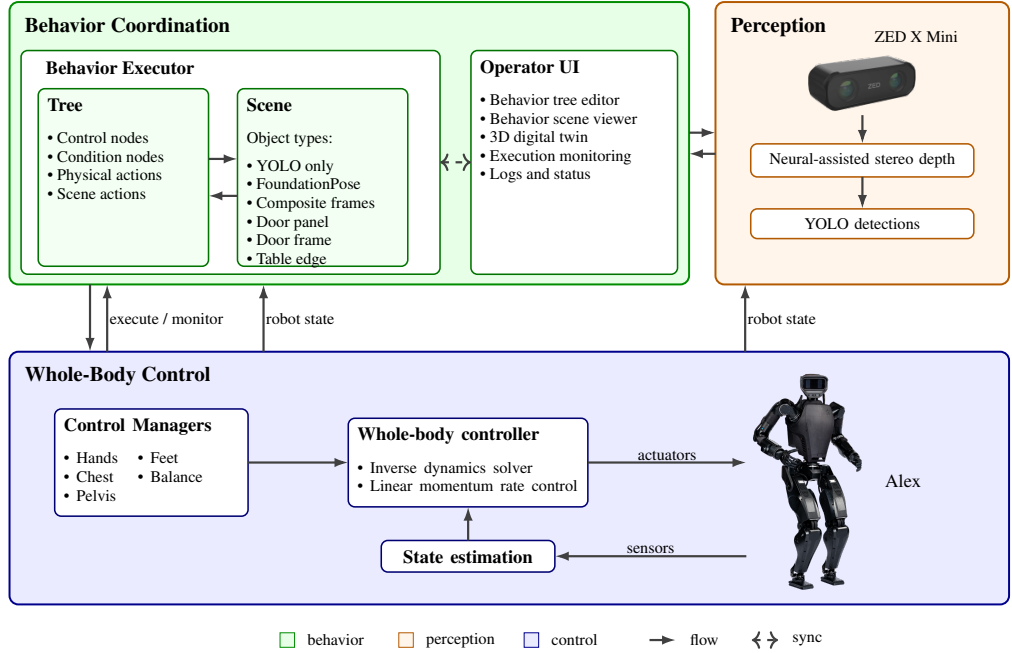

    \centering
    \newsavebox{\behaviorOverviewBox}
    \sbox{\behaviorOverviewBox}{%
        \resizebox{\textwidth}{!}{%
  \input{tikz/ThesisFigureFontBegin.tex}%
  \input{tikz/BehaviorOverviewTikzDissertation.tex}%
  \input{tikz/ThesisFigureFontEnd.tex}%
}%
    }
    \makebox[\textwidth][c]{\hspace*{-0.225\wd\behaviorOverviewBox}\usebox{\behaviorOverviewBox}}%
    \caption[Domain of the behavior system.]{%
        This figure illustrates the domain of our behavior system.
        Our thesis centers on behavior coordination and how it integrates with perception and whole-body control.
        Our behavior executor owns and manages a tree and scene in coordination with a human operator.
        Perception is provided via a ZED X Mini and YOLO detections.
        Physical behavior actions are executed via a whole-body controller which can achieve objectives asynchronously and simultaneously.
        Control objectives include moving or locking parts of the body and walking.}
    \label{fig:Behavior_Overview}
\end{figure}

There are many perspectives from which to view our system.
\autoref{fig:Behavior_Overview} summarizes the primary components by domain: Behavior Coordination, Perception, and Whole-Body Control.
This highlights that our system is built separately from depth and semantic perception and from whole body control.
Our work sits firmly in the top-left area and describes how it interfaces with the others.
The behavior executor owns the behavior tree and a behavior-specific scene.
The operator UI owns the behavior editor, 3D digital robot twin, and system monitoring.

We use a ZED X Mini~\cite{zedxmini}, which is a purely stereo color vision depth sensor, mirroring that of a human.
It provides a high-fidelity, neural-assisted depth point cloud.
A key dependency of the behavior system is YOLO~\cite{redmon2016yolo}, which is a neural net that provides high-frequency semantic object detection and segmentation.
It is crucial to our door behaviors and loco-manipulation behaviors, which perceive the world naturally, with no external sensing or fiducial markers.

The other key dependency is a whole body controller that can reliably walk, pose, and exert meaningful forces on the world.
Ours is set up uniquely to accept asynchronous commands for footsteps and the different body parts, triaging the requests into whole body motions while balancing~\cite{Koolen_2016}.

\section{Process Structure}

\begin{figure}[H]
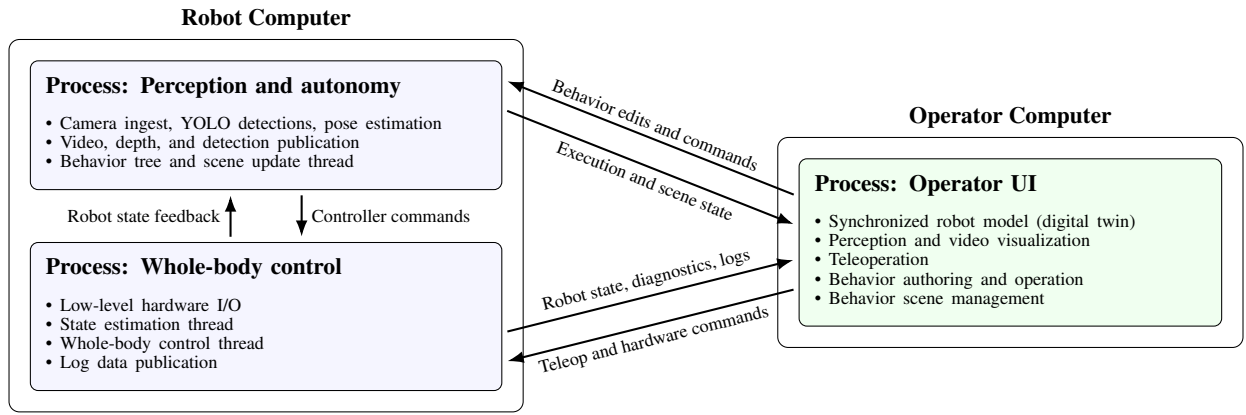

    \centering
    \resizebox{\textwidth}{!}{%
  \input{tikz/ThesisFigureFontBegin.tex}%
      \begin{tikzpicture}[
        font=\small,
    proc/.style={draw, rounded corners=3pt, align=left, inner sep=6pt, text width=0.36\textwidth, fill=blue!4},
    op/.style={draw, rounded corners=3pt, align=left, inner sep=6pt, text width=0.32\textwidth, fill=green!6},
    flow/.style={
        -{Latex[length=2.2mm,width=1.6mm]}, thick,
        shorten >=2pt, shorten <=2pt
    },
    flowlabel/.style={
        sloped, midway, fill=white, inner sep=1.5pt,
        font=\scriptsize
    },
    vlabel/.style={
        midway, fill=white, inner sep=1.5pt, font=\scriptsize
    },
    group/.style={draw, rounded corners=4pt, inner sep=8pt}
        ]
        \node[proc] (autonomy) at (0, 0) {
            \textbf{Process: Perception and autonomy}\\[3pt]
            {\scriptsize
            \begin{tabular}[t]{@{}l@{}}
                \textbullet\ Camera ingest, YOLO detections, pose estimation\\[-1pt]
                \textbullet\ Video, depth, and detection publication\\[-1pt]
                \textbullet\ Behavior tree and scene update thread
            \end{tabular}}
        };

        \node[proc, below=0.7cm of autonomy] (control) {
            \textbf{Process: Whole-body control}\\[3pt]
            {\scriptsize
            \begin{tabular}[t]{@{}l@{}}
                \textbullet\ Low-level hardware I/O\\[-1pt]
                \textbullet\ State estimation thread\\[-1pt]
                \textbullet\ Whole-body control thread\\[-1pt]
                \textbullet\ Log data publication
            \end{tabular}}
        };

        \node[group, fit=(control) (autonomy), label={[font=\small]above:\textbf{Robot Computer}}] (robotgroup) {};

        \node[op, right=4.0cm of control, yshift=1.0cm] (operator) {
            \textbf{Process: Operator UI}\\[3pt]
            {\scriptsize
            \begin{tabular}[t]{@{}>{\raggedright\arraybackslash}p{\linewidth}@{}}
                \textbullet\ Synchronized robot model (digital twin)\\[-1pt]
                \textbullet\ Perception and video visualization\\[-1pt]
                \textbullet\ Teleoperation\\[-1pt]
                \textbullet\ Behavior authoring and operation\\[-1pt]
                \textbullet\ Behavior scene management
            \end{tabular}}
        };

        \node[group, fit=(operator), label={[font=\small]above:\textbf{Operator Computer}}] (operatorgroup) {};

        \def\robotLinkX{0.48cm}
        \def\linkSep{0.20cm}
        \def\autonomyLinkY{0.42cm}
        \def\controlLinkY{-0.42cm}

        \draw[flow] ([xshift=\robotLinkX]autonomy.south) --
            node[vlabel, right=2pt]{Controller commands}
            ([xshift=\robotLinkX]control.north);
        \draw[flow] ([xshift=-\robotLinkX]control.north) --
            node[vlabel, left=2pt]{Robot state feedback}
            ([xshift=-\robotLinkX]autonomy.south);

        \draw[flow] ([yshift={\autonomyLinkY+\linkSep}]operator.west) --
            node[flowlabel, above=1pt]{Behavior edits and commands}
            ([yshift={\autonomyLinkY+\linkSep}]autonomy.east);
        \draw[flow] ([yshift={\autonomyLinkY-\linkSep}]autonomy.east) --
            node[flowlabel, below=1pt]{Execution and scene state}
            ([yshift={\autonomyLinkY-\linkSep}]operator.west);

        \draw[flow] ([yshift={\controlLinkY+\linkSep}]control.east) --
            node[flowlabel, above=1pt]{Robot state, diagnostics, logs}
            ([yshift={\controlLinkY+\linkSep}]operator.west);
        \draw[flow] ([yshift={\controlLinkY-\linkSep}]operator.west) --
            node[flowlabel, below=1pt]{Teleop and hardware commands}
            ([yshift={\controlLinkY-\linkSep}]control.east);
    \end{tikzpicture}%
  \input{tikz/ThesisFigureFontEnd.tex}%
}
    \caption[Runtime structure.]{%
        The runtime structure used in this thesis.
        Two processes run on the robot, and an operator process supports authoring, supervision, and visualization.}
    \label{fig:RuntimeStructureAlex}
\end{figure}

In \autoref{fig:RuntimeStructureAlex} we show another view of the system, where the parts are grouped by process and location.
This is meant to illustrate that the behavior system and perception run on board the robot in a process alongside the whole-body controller, whereas the operator UI is on a separate computer.
This structure is designed to tightly integrate the perception, behavior, and control stack on the robot for performance and for autonomous capability.
This enables high-rate color, depth, and scene data to stay local.
The operator UI inspects, edits, pauses, single-steps, and triggers autonomous execution but does not own task progression.

The decoupling of the user interface and the on-robot stack matters operationally.
If the UI crashes or communications degrade, the runtime keeps the state needed to continue the current task or stop in a controlled way according to its authored logic.
It also matters during development.
The operator can reconnect, inspect the same execution, and resume working without reconstructing behavior state from an external script or a lost UI session.
The split also isolates responsibilities cleanly: autonomy and perception focus on decision-making and scene state, control focuses on real-time balance, walking, and hardware I/O, and the operator process can be rich and inspectable without sitting in the critical execution path.

\section{Node Structure}

\begin{figure}[H]
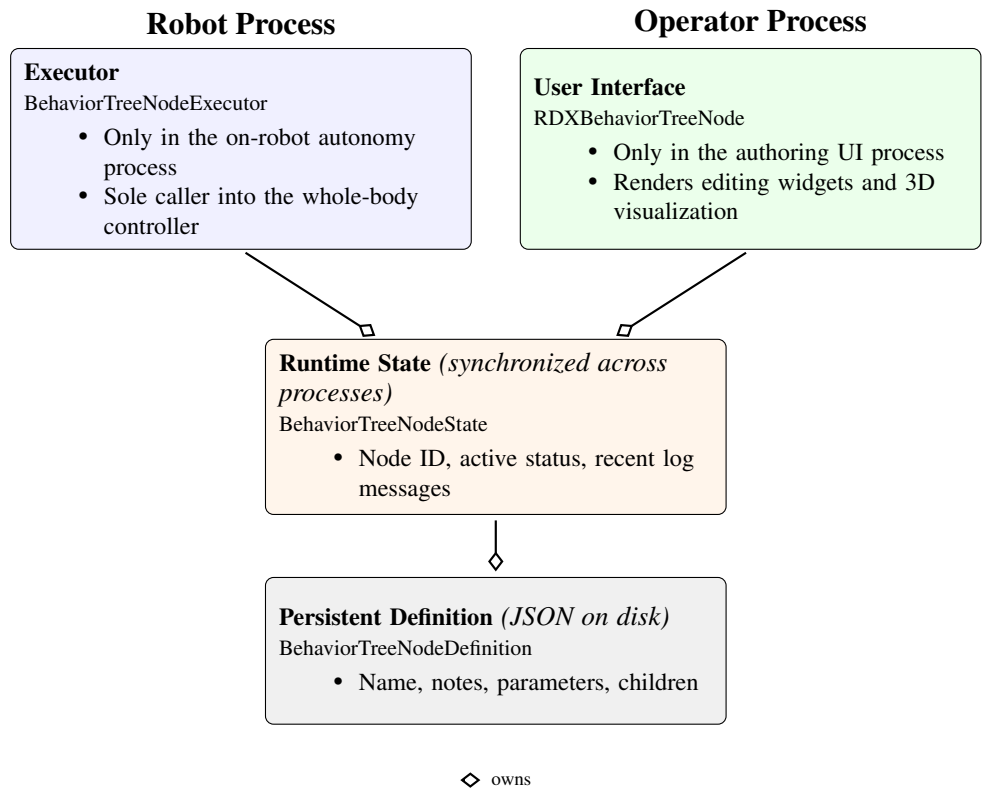

    \centering
    \resizebox{0.8\textwidth}{!}{%
  \input{tikz/ThesisFigureFontBegin.tex}%
\ifdefined\TIKZWEBINCLUDE
    \setlength{\textwidth}{469.75502pt}
    \setlength{\linewidth}{469.75502pt}
\fi
    \newcommand{\nodetitle}[1]{{\small\bfseries #1}}
    \newcommand{\nodecode}[1]{{\ttfamily\footnotesize #1}}
    \begin{tikzpicture}[
        font=\small,
        wrapperbox/.style={
            font={\small\setstretch{1.0}},
            draw, rounded corners=3pt, align=left, inner sep=5pt,
            text width=0.34\textwidth, minimum height=2.6cm
        },
        sharedbox/.style={
            font={\small\setstretch{1.0}},
            draw, rounded corners=3pt, align=left, inner sep=5pt,
            text width=0.34\textwidth, minimum height=1.9cm
        },
        ui/.style={wrapperbox, fill=green!8},
        exec/.style={wrapperbox, fill=blue!6},
        state/.style={sharedbox, fill=orange!8},
        definition/.style={sharedbox, fill=gray!12},
        owns/.style={-{Diamond[open=false, length=2.6mm, width=2.0mm]}, thick,
        shorten >=2pt, shorten <=2pt},
        colheader/.style={font=\fontsize{11.5}{13.8}\selectfont, fill=white, inner sep=1pt}
    ]
    \def\colx{3.3}
    \def\statey{-3.6}
    \def\defy{-6.5}
    \def\headery{1.62}

    \node[colheader] at ( \colx,  \headery) {\textbf{Operator Process}};
    \node[colheader] at (-\colx,  \headery) {\textbf{Robot Process}};

    \node[ui] (rdx) at (\colx, 0) {%
        \nodetitle{User Interface}\\
        \nodecode{RDXBehaviorTreeNode}\\[2pt]
        \begin{itemize}
        \item Only in the authoring UI process
        \item Renders editing widgets and 3D visualization
        \end{itemize}};
    \node[exec] (execn) at (-\colx, 0) {%
        \nodetitle{Executor}\\
        \nodecode{BehaviorTreeNodeExecutor}\\[2pt]
        \begin{itemize}
        \item Only in the on-robot autonomy process
        \item Sole caller into the whole-body controller
        \end{itemize}};

    \node[state] (state) at (0, \statey) {%
        \nodetitle{Runtime State \normalfont\textit{(synchronized across processes)}}\\
        \nodecode{BehaviorTreeNodeState}\\[2pt]
        \begin{itemize}
        \item Node ID, active status, recent log messages
        \end{itemize}};

    \node[definition] (def) at (0, \defy) {%
        \nodetitle{Persistent Definition \normalfont\textit{(JSON on disk)}}\\
        \nodecode{BehaviorTreeNodeDefinition}\\[2pt]
        \begin{itemize}
        \item Name, notes, parameters, children
        \end{itemize}};

    \draw[owns] (rdx.south)   -- ([xshift= 1.5cm]state.north);
    \draw[owns] (execn.south) -- ([xshift=-1.5cm]state.north);
    \draw[owns] (state.south) -- (def.north);

    \node[font=\scriptsize, anchor=north] at (0, \defy - 1.45)
        {\tikz[baseline=-0.6ex]\draw[owns] (0,0) -- (2mm,0);~owns};
    \end{tikzpicture}%
  \input{tikz/ThesisFigureFontEnd.tex}%
}
    \caption[Four-part decomposition used for each behavior-tree node.]{%
        Four-part decomposition used for each behavior-tree node.
        Two process-specific layers, UI and executor, each own an instance of a runtime state, which in turn owns a persistent definition.
        The synchronization layer keeps the runtime state and the live-edited portion of the definition consistent across processes, while the full definition is persisted as JSON.}
    \label{fig:BehaviorNodeDecomposition}
\end{figure}

Each behavior tree node is decomposed into four parts as shown in \autoref{fig:BehaviorNodeDecomposition}: a robot-side executor, an operator-side UI, a synchronized runtime state, and a persisted definition.
The persisted definition stores authored content (name, notes, parameters, children).
The runtime state instantiates the definition and adds runtime information (node identity, active status, recent log messages).
The UI and execution implementations wrap the state and definition layers, allowing the separation by process while sharing synchronized runtime state and authored structure.
Operator interaction, on-robot execution, synchronized state, and definition are each isolated in separate code files.

\section{Robot-Operator Data Synchronization}

\begin{figure}[H]
    \centering
    \resizebox{\textwidth}{!}{\providecommand{\syncbullet}{\textbullet\ }
\newcommand{\synchead}[1]{{\footnotesize\bfseries #1}}
\newcommand{\synccode}[1]{{\ttfamily\scriptsize #1}}
\begin{tikzpicture}[
    font=\small,
    panel/.style={
        draw, rounded corners=4pt, align=left, inner sep=6pt, line width=0.8pt
    },
    sidepanel/.style={
        panel, text width=4.45cm
    },
    op/.style={sidepanel, draw=green!45!black, fill=green!8},
    robot/.style={sidepanel, draw=blue!45!black, fill=blue!8},
    mid/.style={panel, text width=5.20cm, draw=black!45, fill=gray!10},
    aux/.style={panel, text width=7.35cm, draw=orange!70!black, fill=orange!10},
    groupline/.style={draw=black!45, rounded corners=5pt, line width=0.9pt},
    localflow/.style={<->, line width=0.9pt, draw=black!70},
    cmdflow/.style={-{Latex[length=2.2mm,width=1.5mm]}, line width=1.0pt, draw=green!50!black},
    statusflow/.style={-{Latex[length=2.2mm,width=1.5mm]}, line width=1.0pt, draw=blue!55!black},
    clockflow/.style={<->, dashed, line width=0.95pt, draw=orange!70!black},
    grouphead/.style={font=\small\bfseries},
    note/.style={font=\scriptsize, align=center, fill=white, inner sep=1pt}
]

\node[op] (opTree) at (-6.1, 1.55) {%
{\synchead{Local Tree Instance}}\\[-1pt]
{\synccode{RDXBehaviorTreeNode + BehaviorTree}}\\[4pt]
{\scriptsize
\begin{tabular}[t]{@{}l@{}}
\syncbullet authored edits\\[-1pt]
\syncbullet mirrored robot and task info
\end{tabular}}};

\node[robot] (robotTree) at (6.1, 1.55) {%
{\synchead{Local Tree Instance}}\\[-1pt]
{\synccode{BehaviorTreeNodeExecutor + BehaviorTree}}\\[4pt]
{\scriptsize
\begin{tabular}[t]{@{}l@{}}
\syncbullet executor and scene logic\\[-1pt]
\syncbullet task state remains robot-local
\end{tabular}}};

\node[op] (opSync) at (-6.1, -1.55) {%
{\synchead{Per-Tick Wrapper}}\\[-1pt]
{\synccode{ROS2BehaviorTree}}\\[4pt]
{\scriptsize
\begin{tabular}[t]{@{}l@{}}
\syncbullet \texttt{updateSubscription()}\\[-1pt]
\syncbullet local UI tick\\[-1pt]
\syncbullet \texttt{updatePublication()}
\end{tabular}}};

\node[robot] (robotSync) at (6.1, -1.55) {%
{\synchead{Per-Tick Wrapper}}\\[-1pt]
{\synccode{ROS2BehaviorTree}}\\[4pt]
{\scriptsize
\begin{tabular}[t]{@{}l@{}}
\syncbullet \texttt{updateSubscription()}\\[-1pt]
\syncbullet local executor tick\\[-1pt]
\syncbullet \texttt{updatePublication()}
\end{tabular}}};

\node[mid] (message) at (0, -1.55) {%
{\synchead{30 Hz Tree Snapshot}}\\[-1pt]
{\synccode{BehaviorTreeStateMessage}}\\[4pt]
{\scriptsize
\begin{tabular}[t]{@{}l@{}}
\syncbullet sequence ID and next node ID\\[-1pt]
\syncbullet root and topology modification records\\[-1pt]
\syncbullet depth-first typed snapshot\\[-1pt]
\syncbullet full or partial node payloads
\end{tabular}}};

\node[aux] (clock) at (0, -4.75) {%
{\synchead{Clock-Offset Support for CRDT Merge}}\\[-1pt]
{\synccode{ROS2PeerClockOffsetEstimator}}\\[-1pt]
{\synccode{ROS2PeerClockOffsetEstimatorPeer}}\\[4pt]
{\scriptsize
\begin{tabular}[t]{@{}l@{}}
\syncbullet 5 Hz ping/reply between publisher GUIDs\\[-1pt]
\syncbullet half round-trip time estimates peer clock offset
\end{tabular}}};

\node[groupline, draw=green!45!black, fit=(opTree) (opSync), inner sep=9pt] (opGroup) {};
\node[groupline, draw=blue!45!black, fit=(robotTree) (robotSync), inner sep=9pt] (robotGroup) {};
\node[grouphead, anchor=south] at (opGroup.north) {Operator UI Process};
\node[grouphead, anchor=south] at (robotGroup.north) {Robot Autonomy Process};

\draw[localflow] (opTree.south) -- (opSync.north);
\draw[localflow] (robotTree.south) -- (robotSync.north);

\draw[cmdflow] ([yshift=0.52cm]opSync.east) -- ([yshift=0.52cm]message.west);
\draw[cmdflow] ([yshift=0.52cm]message.east) -- ([yshift=0.52cm]robotSync.west);
\draw[statusflow] ([yshift=-0.52cm]robotSync.west) -- ([yshift=-0.52cm]message.east);
\draw[statusflow] ([yshift=-0.52cm]message.west) -- ([yshift=-0.52cm]opSync.east);

\draw[clockflow] ([xshift=1.10cm]opSync.south) -- ([xshift=-2.30cm]clock.north);
\draw[clockflow] ([xshift=-1.10cm]robotSync.south) -- ([xshift=2.30cm]clock.north);

\end{tikzpicture}}
    \caption[Behavior-tree synchronization between the operator UI and robot executor.]{%
        Behavior-tree synchronization between the operator UI and robot executor.
        Each side maintains a local tree instance and a per-tick synchronization wrapper.
        A 30~Hz depth-first tree snapshot carries either full or partial per-node payloads.
        Bidirectional CRDT fields use clock-offset-corrected last-writer arbitration; robot-owned status fields are mirrored one way to the UI.}
    \label{fig:BehaviorTreeSynchronization}
\end{figure}
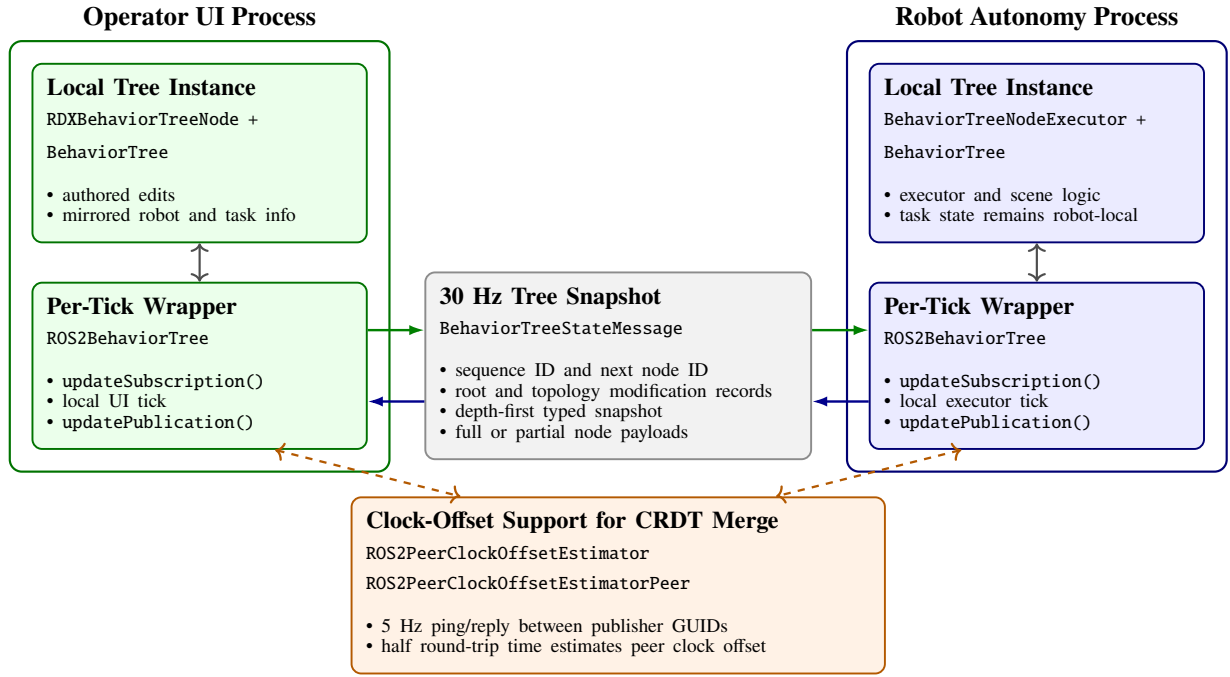

The cost of our process structure is that it necessitates a complex synchronization mechanism between the robot and the operator to facilitate rich runtime editability.
We looked to the literature to find algorithms to synchronize our data, where we found Conflict-Free Replicated Data Types (CRDTs)~\cite{shapiro2011crdt}.
We decided to embrace the concept and formulate our behavior tree and scene state as a CRDT and synchronize it at 30 Hz, as shown in \autoref{fig:BehaviorTreeSynchronization}.

We use the ROS 2 DDS middleware to transmit the synchronization messages.
Its autodiscovery feature streamlines the connections between operator UIs and robots.
The robot and the operator are allowed to concurrently edit data types such as footstep goal poses, which may be moved by the robot to update their pose with respect to the parent frame while the operator may wish to modify the definition of those goal poses.
We use a latest-timestamp algorithm to resolve potential conflicts.
The latest modification to a data field persists.
A clock-offset estimator, shown at the bottom of \autoref{fig:BehaviorTreeSynchronization}, supports this design by providing a means for comparing data modification times.

The operator interface and the robot-side behavior system thread symmetrically synchronize on each tick.
At the start of each tick, the wrappers apply any received data updates.
After the UI or executor performs its local work, it republishes the current tree and increments an update number that is used to maintain order.

The serializer packs the tree in depth-first order using a node type table and per-type message arrays.
To save bandwidth, each node is sent as a compact partial-data payload unless a full type-specific payload is needed.
Full data are sent when the node definition has changed, when a peer has requested retransmission, or when the node reports fresh status.
The compact version carries only the generic state and CRDT metadata.
This mechanism is summarized in \autoref{fig:BehaviorTreeSynchronization}.

Topology is synchronized separately from per-node content.
The tree root reference, whole-tree metadata including the next available node ID, and each node's child list each carry their own modification metadata.
On receipt, the subscriber reconstructs an intermediate message tree, matches nodes by ID, and applies topology operations only where the incoming root or child-list modification is newer.
A topology change queue is used to apply all topology changes at once, later, to ensure tree consistency.
New nodes are replicated locally through the node builder, moved or reordered nodes are reattached through the topology queue, and nodes that disappear from the incoming tree are destroyed.
If a process comes online late or a dropped message leaves only partial data available, the receiver flags the node as needing full data and the peer resends the full payload on a later publication.

Some node fields, such as execution status and visualization data, are unidirectional and only modifiable by one actor.
Examples include leaf execution information such as ``is next for execution'', ``can execute'', ``is executing'', and ``has failed'', which are owned by the robot side.
These are mirrored to the other side without arbitration.

Other fields are concurrently modifiable by all actors, which we refer to as bidirectional fields.
Examples include the root node's ``automatic execution'' gate, ``next execution index'' selection, manual-step requests, ``enable concurrency'', preview mode, and ``reset failures''.

The bidirectional CRDT fields use a last-modified record that stores author identity, a monotonic modification number, and a timestamp.
A local write records the writer GUID, modification number, and timestamp.
On receipt, a newer modification number wins immediately.
Equal-number races are resolved by comparing timestamps after transforming the peer timestamp into the local clock frame using a peer clock-offset estimator.
This estimator derives the time offset from ROS 2 ping-reply messages under a symmetric-delay assumption.

\section{Persistence Storage of Behaviors}

To enable authored structure to be saved, loaded, copied, and versioned, we support saving to and loading from JSON files.
The saved JSON represents the definition layer from \autoref{fig:BehaviorNodeDecomposition}.
It contains names, notes, parameter values, child hierarchy, and node type specifics.
Each node is serialized with a \texttt{type} field and a \texttt{children} array that mirrors the behavior tree.
\autoref{fig:BehaviorDefinitionJsonCondensed} shows a condensed excerpt illustrating nested arm and walk action definitions.

\begin{figure}[H]
    \centering
    \begin{minipage}{0.72\columnwidth}
  \input{tikz/ThesisFigureFontBegin.tex}%
  \input{tikz/BehaviorDefinitionJsonCondensed.tex}%
  \input{tikz/ThesisFigureFontEnd.tex}%
    \end{minipage}
    \caption[Condensed behavior JSON excerpt.]{%
        Condensed excerpt from a simple behavior JSON file.
        Each node records its definition \texttt{type}, parameters, and nested \texttt{children}.
        Ellipses mark omitted sibling nodes and fields.}
    \label{fig:BehaviorDefinitionJsonCondensed}
\end{figure}

We have tried to make these files readable by humans, but they are not intended to be edited.
For example, instead of saving a number with unnecessarily high precision (i.e.\ ``0.1349159123''), we round it (i.e.\ ``0.135'').
We also use degrees instead of radians because it's easier to reason about.
Still, we recommend viewing behaviors in the operator interface for the best understanding.
A step-by-step tutorial example is given in \autoref{sec:simple_behavior_json_file}.

\section{Whole-Body Controller}

The behavior coordination layer interfaces with a whole-body and walking controller that executes arbitrary trajectory requests and footsteps.
The behavior runtime sends requests to the controller and monitors execution through status messages.

\begin{figure}[H]
    \centering
    \includegraphics[width=0.8\textwidth]{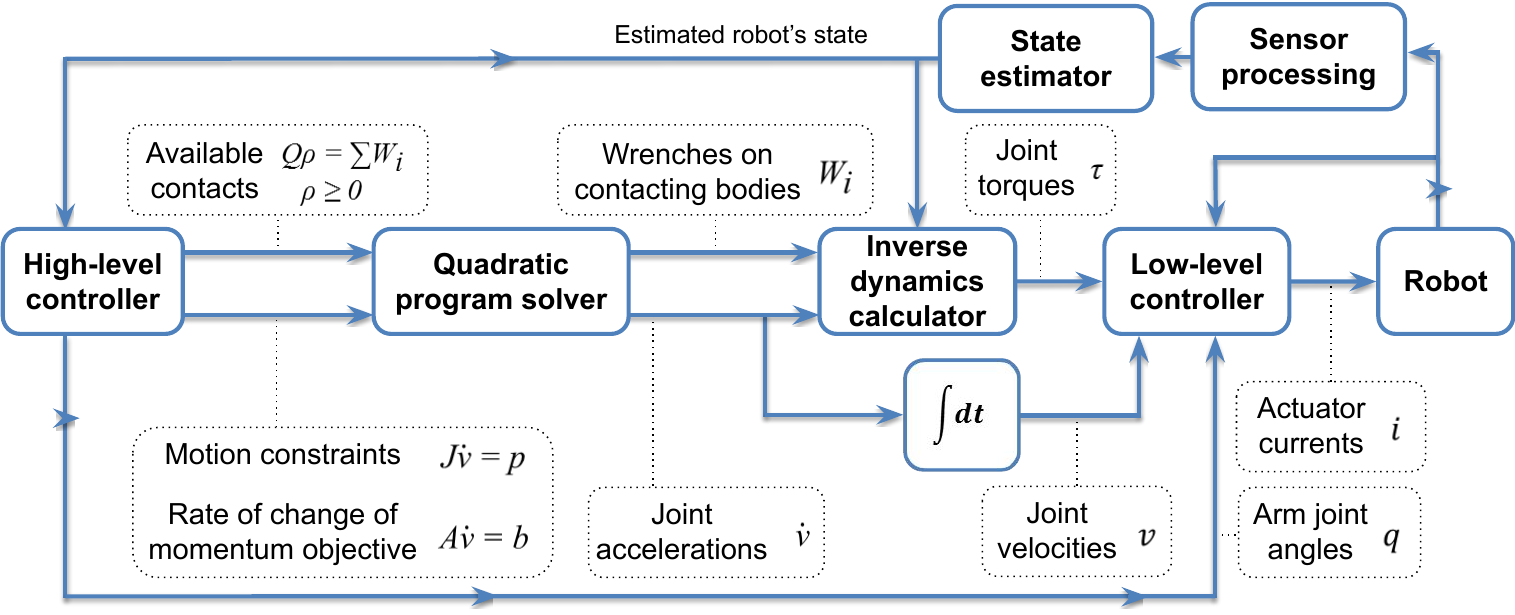}
    \caption[IHMC whole-body control framework.]{IHMC overall whole-body control framework.}
    \label{fig:ControllerOverview}
\end{figure}

\begin{figure}[H]
    \centering
    \includegraphics[width=0.7\textwidth]{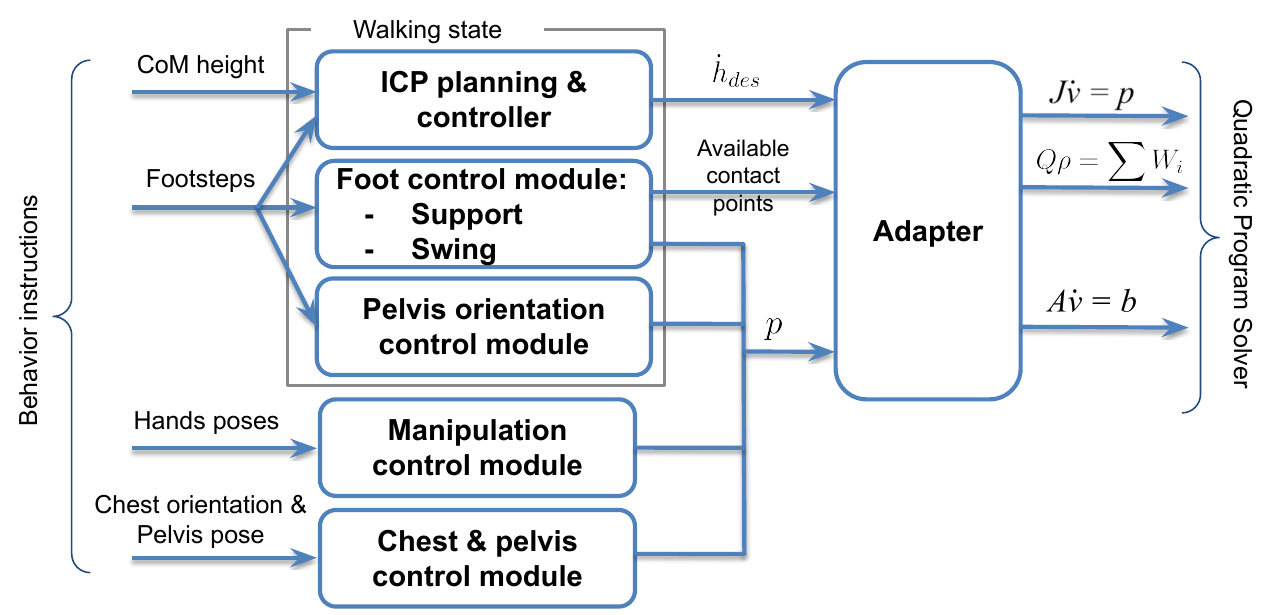}
    \caption[IHMC humanoid walking controller.]{%
        IHMC humanoid walking controller, highlighting control modules that regulate mobility and manipulation.
        Outputs of these modules become inputs to the QP solver in the whole-body control framework.}
    \label{fig:highLevelControllerOverview}
\end{figure}

We use the whole-body momentum-based control framework developed by Koolen et al.~\cite{Koolen_2016}.
The general structure and control flow of the whole-body controller and walking implementation are illustrated in \autoref{fig:ControllerOverview} and \autoref{fig:highLevelControllerOverview}.
It is a model-based torque-control scheme centered on a quadratic program (QP) that reconciles weighted motion-task objectives on the generalized joint-acceleration vector, with QP constraints for dynamic feasibility, foot-ground contact, and force-limited grasping.
The action primitives use joint-space and spatial task-space objectives provided by this controller.

The QP~\cite{Koolen_2016} solves for the desired generalized joint-acceleration vector $\dot{v}_{\text{d}}$ and basis-vector multiplier $\rho$, minimizing weighted momentum and spatial motion-task errors subject to the equations of motion and contact friction:

\begin{equation}
    \begin{aligned}
        \underset{\dot{v}_{\text{d}}, \rho}{\text{minimize}} \quad & (A\dot{v}_{\text{d}} - b_h)^T C_h (A\dot{v}_{\text{d}} - b_h) + (J\dot{v}_{\text{d}} + j - p_{\text{d}})^T C_J (J\dot{v}_{\text{d}} + j - p_{\text{d}}) + \rho^T C_\rho \rho, \\
        \text{subject to} \quad & A\dot{v}_{\text{d}} + \dot{A}v = W_g + Q\rho + \sum_i W_{\text{ext},i}, \\
        & \rho_{\text{min}} \leq \rho \leq \rho_{\text{max}},
    \end{aligned}
\end{equation}
\noindent where $Av$ is centroidal momentum and $\dot{A}v + A\dot{v}$ is its rate of change, $b_h = \dot{h}_{\text{d}} - \dot{A}v$ is the desired momentum-rate error, $C_h$, $C_J$, and $C_\rho$ are cost-function weighting matrices, $J$ is a Jacobian that maps generalized joint accelerations to spatial accelerations, $j = \dot{J}v$ is the convective term, $p_{\text{d}}$ is the desired spatial acceleration, $W_g$ is the wrench from gravity, $Q\rho$ is the ground-reaction wrench, and $W_{\text{ext},i}$ is an external wrench.
The resulting ground reaction, pre-specified wrenches, and desired joint accelerations are used as input to an inverse-dynamics solver to compute the desired joint torques $\tau_d$.

We use the dynamic walking behavior described in~\cite{Koolen_2016}, composed of standing, transfer, and swing states with associated motion tasks and active contacts.
The controller executes a queue of footstep poses submitted by the behavior layer and uses instantaneous capture point (ICP)~\cite{pratt2006capturepoint} to model balance dynamics.
Foot pose, pelvis height, and pelvis orientation trajectories are generated to achieve the desired footsteps while maintaining balance.
During transfer, an ICP trajectory is generated and heel lift is triggered by selectively enabling heel contact points as the stance foot pitches toward toe-off.
The centroidal moment pivot (CMP) can also be used to regulate angular momentum during balance control.
More details can be found in~\cite{Pratt2019_bookchapter}.

Foot swings use smoothed trajectories that vary by terrain type.
On rough terrain, swings may be planned to avoid tripping; on flat ground, default conservative trajectories with low swing height are used.
To recover from external disturbances and instabilities, swings can also be sped up during execution to more quickly catch balance.
State estimation combines kinematic foot-ground-contact proprioception for pelvis position and velocity with a pelvis-mounted IMU for pelvis orientation.
When a foot is in contact (determined by a glitch-filtered virtual foot switch), we assume it does not slide, and pelvis position and velocity are estimated by fusing leg kinematics with IMU acceleration data.
The positional estimate is therefore susceptible to drift from inaccurate foot-contact detection, foot slipping, and lack of both feet on the ground.

\section{Operator Interface}

Our operator interface is based on a robotics software sandbox called Robot Data eXplorer (RDX), as introduced in \autoref{ch:building}.
Hands-on examples of how RDX is used to create and modify behaviors using our system are covered in \autoref{ch:tutorial}.
RDX provides dockable panels of widgets, an interactive 3D scene, overlays, toolbars, and persisted layout and configuration handling.
Its primary third-party libraries are Dear ImGui, libGDX, and OpenVR~\cite{dearimgui,libgdx,openvr}, and the same framework treats virtual reality as a core visualization and interaction mode rather than a separate tool.
This single UI framework covers behavior authoring, digital-twin visualization, perception debugging, graph-based logging, and direct teleoperation modes such as whole-body streaming, footstep placement, and joystick walking.
The operator tooling co-evolved with the runtime and shares the same data structures, so authoring, execution, and diagnosis all operate over a common substrate.

See \autoref{ch:tutorial} for extensive coverage of operator interface features.

\section{Object-Centric Action Definition}

In our system, many of our physical actions can be defined in the coordinate frame of a part of the robot or a scene object.
This includes hand poses, footsteps, spine action, the screw primitive axes, and the neck action.
When approaching objects like tables and doors, the footsteps are defined with respect to the perceived object frame or door frame.
When grasping an object, the pre-grasp hand poses are defined with respect to the perceived object.
This important architectural design element is inspired by both the IHMC DARPA Robotics Challenge user interface presented in \autoref{sec:drc_era} and the Affordance Template Framework~\cite{Hart_2014, Hart_2015} discussed in \autoref{sec:ats}.

This is also referred to as ``object-centric'' action, which is acting with respect to the object, versus ``ego-centric'' action, which is acting with respect to the robot.
Object-centric action definition is actually essential and core to the reusability of behaviors.
It serves two main purposes.
The first is to enable varied starting conditions of the robot.
For example, task approach should be specified with respect to the task.
The other is to reduce compounding errors.
Ego-centric actions in sequence are subject to state estimator drift and other control errors.

If you played back a door traversal behavior fully ego-centrically, assuming the robot started at the same initial pose as during authoring, it would likely miss the handle grasp or run into the door panel or frame.
As an example of why, the door traversal approach stance may not achieve the desired footsteps accurately due to bad balance and recovery during the final steps.
This would result in a stance that is slightly different than planned.
If it were to then try to grasp the handle with respect to that stance (self) instead of the handle, even though it may have worked before, it will likely not work this time.
The pre-grasp hand pose would be misaligned.
Instead, by using the actively perceived or re-perceived door handle pose to define the pre-grasp action, the grasp should succeed, overcoming the prior error using the inverse kinematics solver to achieve the correct pose.

Object-centric action definition is a way of doing sequential composition in the sense proposed by~\cite{1999_burridge}.
The example above may be viewed as Lyapunov funnels, where task approach is funnel A\@.
The opening of funnel A is the space of possible starting locations of the robot.
Funnel B is the handle grasp action.
The size and shape of the opening of funnel B would represent the tolerance to different robot stance locations.
It would be mainly based on arm reachability from the various stances.
In this way, you could say that object-centric actions are a way to achieve task stability given a system with accruing errors.

This is implemented using the Euclid math library~\cite{ihmc_euclid}, which provides a reference frame tree with world frame as the root and the robot and all scene objects as subtrees and leaves.
Euclid makes it easy to create new reference frames and calculate transformations between any two frames in the tree.

We show how to define a hand pose with respect to a perceived object in \autoref{sec:authoring_a_frame_based_arm_action}.

\section{Behavior Tree Structure}

Our behaviors are structured as a tree of nodes.
The tree organization serves a dual-purpose.
It allows for organization of behavior and also a way to author logic in a simple way.

We reuse the concept of a filesystem in computing for behaviors.
Robot behaviors can be abstracted across many layers.
Useful tasks for robot work are defined as completable goals which usually consist of higher level abstractions like ``sort the objects into containers'' or ``deliver this package to room 40''.
These high level task specifications ultimately need to be triaged to a sequence of low-level primitive actions.
High level task specifications live towards the root of the tree and low level primitive actions live as the leaves.

\begin{figure}[H]
    \centering
    \small
    \begin{tikzpicture}[
    node distance=1.4cm and 1.6cm,
    every node/.style={font=\sffamily},
    btnode/.style={
        draw,
        minimum width=1.5cm,
        minimum height=1.0cm,
        align=center,
        fill=white,
        thick
    },
    actionnode/.style={
        btnode,
        minimum width=2.2cm
    },
    condition/.style={
        ellipse,
        draw,
        thick,
        fill=white,
        align=center,
        inner xsep=6pt,
        inner ysep=3pt,
        minimum width=2.4cm
    },
    flow/.style={
        ->,
        thick
    },
    nextexec/.style={
        ->,
        very thick,
        green!60!black
    },
    gotoflow/.style={
        ->,
        thick,
        dashed,
        draw=black!65
    }
]

\newcommand{\btrcrossedzero}{%
    \tikz[baseline=-0.55ex, scale=0.9]{
        \node[inner sep=0pt, font=\sffamily] at (0,0) {0};
        \draw[line width=0.8pt] (-0.18,-0.14) -- (0.18,0.14);
    }%
}

\node[btnode] (root) {\btrcrossedzero};

\node[btnode, below=1.6cm of root] (seq) {$\rightarrow$};

\node[actionnode, below left=2.0cm and 2.4cm of seq] (actionA) {Action A};
\node[btnode, below right=2.0cm and 2.4cm of seq] (fb) {?};

\node[condition, below left=2.0cm and 2.0cm of fb] (tryCond) {Try condition};
\node[actionnode, below=2.0cm of fb] (catchAct) {Catch action};
\node[actionnode, below right=2.0cm and 2.0cm of fb] (gotoAct) {Goto action};

\draw[flow] (root) -- (seq);
\draw[flow] (seq) -- (actionA);
\draw[flow] (seq) -- (fb);
\draw[flow] (fb) -- (tryCond);
\draw[flow] (fb) -- (catchAct);
\draw[flow] (fb) -- (gotoAct);

\draw[nextexec] ([xshift=-1.1cm]actionA.west) -- (actionA.west);

\draw[gotoflow]
    (gotoAct.north west)
    .. controls +( -1.8cm, 1.6cm) and +( 2.4cm, -1.4cm) ..
    (actionA.south east)
    node[pos=0.42, font=\scriptsize, fill=white, inner sep=1pt, sloped, above]
        {Action A};

\end{tikzpicture}
    \caption[Representative sample behavior tree.]{%
        A representative sample tree.
        At a glance, the structure looks similar to a behavior tree; however, there are some important differences.
        The next action to execute is held as state between updates, as indicated by the arrow pointing to Action A.
        In our tree, ``everything is a sequence'' by default in the depth-first ordering of the leaves.
        The fallback node catch is a sequence instead of a list of additional things to try.
        Goto nodes are used to modify the next action to execute index.}
    \label{fig:BehaviorTreeStructureSample}
\end{figure}
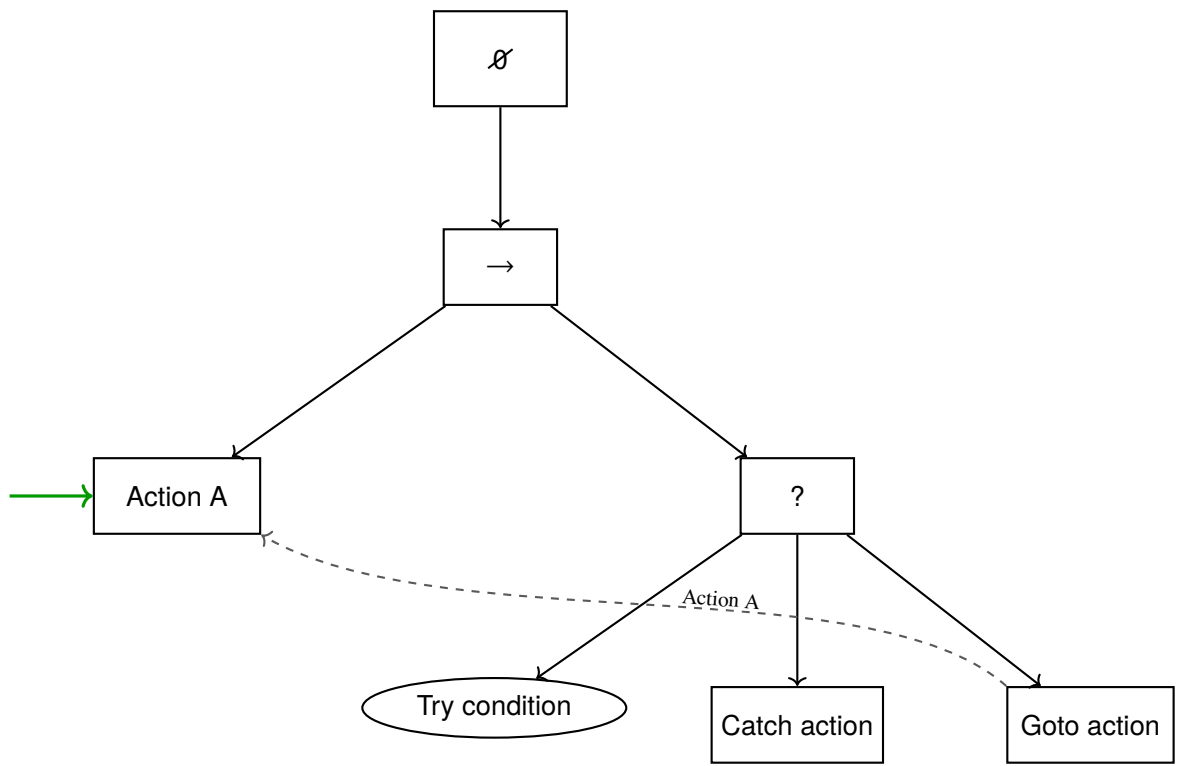

Our system is not a Behavior Tree as in the literature~\cite{2018_colledanchise} (we will use capitalized Behavior Tree to reference the literature version).
It is rather a tree of behavior implemented in a way we found most useful.
However, we do adopt Behavior Trees' concepts of sequence, fallback, condition, and action nodes.
\autoref{fig:BehaviorTreeStructureSample} summarizes the main structural elements.

Our runtime updates all nodes on each tick and does not reduce control flow to a fixed set of return types such as ``success,'' ``failure,'' and ``running.''
The node library is freer in how it represents and exposes execution state, and the model stays closer to how we actually authored and debugged humanoid door traversals, where action overlap, retries, manual stepping, and scene-state inspection mattered more than adherence to a standard BT interface.

In our implementation, the whole tree is one big sequence.
The order of the sequence is the depth-first ordering of the leaves.
\autoref{alg:simplified_execution} outlines the basic sequence algorithm.

A unique part of our architecture, which diverges even further from Behavior Trees, is that we keep a next execution index.
This is in contrast to restarting at the root node on each tick.
In this way, we are more like a state machine.
The next execution index is a pointer to the next node to execute, which can be changed by the operator, the sequence executor, a goto node, or a fallback node.

We also have a unique implementation of a fallback node, which is not a list of steps to try, but rather one thing to try and a sequence as a catch.
We cover the fallback node in detail in \autoref{sec:fallback_node}.

The tree view in the operator interface is closer to a hierarchical file browser than to a dense 2D graph editor.
That choice matched the rest of the operator workflow because the behavior view had to coexist with a 3D scene, first-person sensor imagery, scene-object panels, robot-status panels, and teleoperation widgets.

In summary, our implementation is neither a Behavior Tree, a state machine, nor a hierarchical state machine.
Instead, it combines elements from each to suit the needs of a runtime-editable behavior authoring system.

\section{Generalized Door Traversal Behavior}

As an example, \autoref{fig:TreeStructure} abstracts the authored structure of \texttt{door/Door\-Traversal.json} and the expanded right-pull subtree in \texttt{door/Right\-Pull\-Door.json}.
The top-level file handles the general approach, constructs the panel-, frame-, and hybrid-frame scene objects used for door-specific execution, and dispatches to left-push, left-pull, right-push, and right-pull branches.
The right-pull branch is expanded because it makes the authored strategy clear: state-specific approach nodes feed a detailed lever-turn and door-opening sequence with explicit fallback loops before the final traversal.

\begin{figure}[H]
    \centering
    {\singlespacing
    \resizebox{\textwidth}{!}{%
  \input{tikz/ThesisFigureFontBegin.tex}%
  \begin{tikzpicture}[
    font=\small,
    x=1cm,
    y=1cm,
    every node/.style={outer sep=0pt, text=black},
    flow/.style={-{Latex[length=2.2mm,width=1.5mm]}, line width=0.95pt, draw=black!75},
    flowhi/.style={-{Latex[length=2.2mm,width=1.5mm]}, line width=1.0pt, draw=blue!60!black},
    loopflow/.style={-{Latex[length=2.0mm,width=1.4mm]}, line width=0.85pt, draw=red!65!black},
    topbg/.style={rounded corners=5pt, draw=black!22, fill=black!2, line width=0.7pt},
    bottombg/.style={rounded corners=5pt, draw=blue!28!black, fill=blue!3, line width=0.7pt},
    title/.style={font=\small\bfseries, inner sep=1pt},
    root/.style={
        rounded corners=4pt,
        draw=teal!55!black,
        fill=teal!10,
        text width=4.5cm,
        align=center,
        inner sep=6pt,
        line width=0.8pt
    },
    phase/.style={
        rounded corners=4pt,
        draw=green!50!black,
        fill=green!9,
        text width=5.6cm,
        align=left,
        inner sep=6pt,
        line width=0.8pt
    },
    decide/.style={
        rounded corners=4pt,
        draw=orange!70!black,
        fill=orange!10,
        text width=5.6cm,
        align=left,
        inner sep=6pt,
        line width=0.8pt
    },
    branch/.style={
        rounded corners=3pt,
        draw=black!35,
        fill=black!4,
        text width=2.75cm,
        align=center,
        inner sep=5pt,
        line width=0.7pt
    },
    branchhi/.style={
        rounded corners=3pt,
        draw=blue!58!black,
        fill=blue!11,
        text width=2.95cm,
        align=center,
        inner sep=5pt,
        line width=0.9pt
    },
    detailroot/.style={
        rounded corners=4pt,
        draw=blue!58!black,
        fill=blue!10,
        text width=4.7cm,
        align=center,
        inner sep=6pt,
        line width=0.8pt
    },
    entry/.style={
        rounded corners=3pt,
        draw=blue!45!black,
        fill=blue!7,
        text width=3.4cm,
        align=left,
        inner sep=5pt,
        line width=0.8pt
    },
    entrymuted/.style={
        rounded corners=3pt,
        draw=black!32,
        fill=black!3,
        text width=3.2cm,
        align=left,
        inner sep=5pt,
        line width=0.7pt
    },
    detail/.style={
        rounded corners=4pt,
        draw=blue!42!black,
        fill=white,
        text width=3.2cm,
        align=left,
        inner sep=6pt,
        line width=0.8pt
    },
    endnode/.style={
        rounded corners=3pt,
        draw=black!35,
        fill=black!4,
        text width=1.8cm,
        align=center,
        inner sep=4pt,
        line width=0.7pt
    },
    note/.style={font=\scriptsize, align=left, text=black!72},
    loopnote/.style={
        font=\scriptsize\bfseries,
        text=red!65!black,
        fill=white,
        inner sep=1pt
    },
    faint/.style={font=\scriptsize\itshape, text=black!65}
]

\path[topbg] (0.1, 7.9) rectangle (16.7, 17.7);
\path[bottombg] (0.1, -0.85) rectangle (16.7, 7.6);

\node[title, anchor=north west] at (0.35, 17.4) {Top-level door traversal tree};
\node[title, anchor=north west] at (0.35, 7.4) {Expanded right-pull subtree};

\node[root] (doorroot) at (8.4, 16.6) {%
    {\bfseries \texttt{DoorTraversal.json}}\\[2pt]
    top-level traversal root};

\node[phase] (approach) at (8.4, 14.3) {%
    {\bfseries General approach}\\[2pt]
    acquire the door panel\\
    build the approach frame\\
    skip pre-approach when already close};

\node[decide] (decide) at (8.4, 11.2) {%
    {\bfseries Type determination and dispatch}\\[2pt]
    create panel-, frame-, and hybrid-frame objects\\
    freeze the scene and choose the matching door subtree};

\node[branch] (leftpush) at (2.2, 8.7) {%
    {\bfseries Left push}\\[-1pt]
    {\scriptsize \texttt{LeftPushDoor.json}}};
\node[branch] (leftpull) at (5.9, 8.7) {%
    {\bfseries Left pull}\\[-1pt]
    {\scriptsize \texttt{LeftPullDoor.json}}};
\node[branch] (rightpush) at (9.6, 8.7) {%
    {\bfseries Right push}\\[-1pt]
    {\scriptsize \texttt{RightPushDoor.json}}};
\node[branchhi] (rightpull) at (13.7, 8.7) {%
    {\bfseries Right pull}\\[-1pt]
    {\scriptsize \texttt{RightPullDoor.json}}};

\draw[flow] (doorroot.south) -- (approach.north);
\draw[flow] (approach.south) -- (decide.north);
\draw[flow] (decide.south) -- (leftpush.north);
\draw[flow] (decide.south) -- (leftpull.north);
\draw[flow] (decide.south) -- (rightpush.north);
\draw[flowhi] (decide.south) -- (rightpull.north);

\node[detailroot] (rproot) at (8.1, 6.25) {%
    {\bfseries \texttt{RightPullDoor.json}}\\[2pt]
    state-specific entry points for the right-pull strategy};

\draw[flowhi] (rightpull.south) -- ++(0,-0.35) -| (rproot.north);

\node[entry] (closed) at (2.8, 3.8) {%
    {\bfseries Closed door entry}\\[1pt]
    hybrid-frame stance\\
    clear scene and prepare lever interaction};

\node[entry] (ajar) at (7.9, 3.9) {%
    {\bfseries Ajar door entry}\\[1pt]
    door-frame approach\\
    shorter opening path};

\node[entrymuted] (open) at (13.5, 3.9) {%
    {\bfseries Open door entry}\\[1pt]
    checkpoint branch\\
    present but not detailed here};

\draw[flow] (rproot.south) -- (closed.north);
\draw[flow] (rproot.south) -- (ajar.north);
\draw[flow] (rproot.south) -- (open.north);

\node[detail] (lever) at (2.2, 0.95) {%
    {\bfseries Lever lock and pre-grasp}\\[1pt]
    freeze the handle\\
    align the hand and fingers};

\node[detail] (handle) at (6.3, 0.95) {%
    {\bfseries Turn handle and pull}\\[1pt]
    lever-turn primitive\\
    initial pull-open motion};

\node[detail] (clearway) at (10.6, 0.95) {%
    {\bfseries Open wider and clear doorway}\\[1pt]
    push the door farther open\\
    wait until the passage is clear};

\node[detail] (traverse) at (14.7, 1.2) {%
    {\bfseries Traverse doorway}\\[1pt]
    walk through\\
    tuck arms and reset posture};

\node[endnode] (end) at (14.7, -0.55) {%
    {\bfseries End}};

\draw[flow] (closed.south) -- (lever.north);
\draw[flow] (lever.east) -- (handle.west);
\draw[flow] (handle.east) -- (clearway.west);
\draw[flow] (clearway.east) -- (traverse.west);
\draw[flow] (traverse.south) -- (end.north);

\draw[flow] (ajar.south east) to[out=-35,in=150] (traverse.north west);
\draw[flow] (open.south) -- (traverse.north);

\draw[loopflow] (handle.north) to[out=120,in=60] node[loopnote, above, yshift=0.1cm, pos=0.48] {retry if still closed} (lever.north east);
\draw[loopflow] (clearway.north east) .. controls +(-0.9,0.8) and +(0.9,0.8) .. node[loopnote, above, pos=0.5] {repeat until clear} (clearway.north west);

\end{tikzpicture}%
  \input{tikz/ThesisFigureFontEnd.tex}%
}}
    \caption[Authored tree structure for top-level door traversal with expanded right-pull subtree.]{%
        Authored tree structure for the top-level door traversal behavior and the expanded right-pull subtree.
        The upper portion abstracts \texttt{door/DoorTraversal.json}: a general approach phase is followed by door-type determination and dispatch to one of four door-specific subtree files.
        The lower portion expands \texttt{door/RightPullDoor.json}, showing its closed-, ajar-, and open-door entry points and the detailed closed-door path through lever acquisition, handle turning, door opening, fallback retries, doorway-clear checks, and traversal.
        Intermediate action nodes are grouped into phase-level blocks for readability.}
    \label{fig:TreeStructure}
\end{figure}

\section{Node Library}

\begin{figure}[H]
    \centering
    \resizebox{\columnwidth}{!}{\begin{tikzpicture}[
    font=\small,
    x=1cm,
    y=1cm,
    scale=1.0,
    transform shape,
    every node/.style={outer sep=0pt, text=black},
    region/.style={rounded corners=4pt, line width=0.9pt},
    entry/.style={region, draw=black!45, fill=black!4},
    control/.style={region, draw=green!55!black, fill=green!10},
    action/.style={region, draw=blue!55!black, fill=blue!8},
    subpanel/.style={draw=black!30, fill=white, rounded corners=3pt, line width=0.7pt},
    controlpanel/.style={subpanel, draw=green!45!black},
    actionpanel/.style={subpanel, draw=blue!45!black},
    card/.style={
        draw=black!35,
        fill=white,
        rounded corners=3pt,
        line width=0.7pt,
        minimum width=2.70cm,
        minimum height=0.56cm,
        align=center,
        inner sep=2.4pt,
        font=\scriptsize
    },
    entrycard/.style={card, draw=black!45},
    controlcard/.style={card, draw=green!45!black},
    actioncard/.style={card, draw=blue!45!black},
    disabledcard/.style={card, draw=black!20, dashed, fill=black!2, text=black!55},
    title/.style={font=\small\bfseries, inner sep=1.5pt},
    subtitle/.style={font=\scriptsize, text=black!70, inner sep=1pt},
    grouplabel/.style={font=\scriptsize\bfseries, text=black!80}
]

\path[entry] (0.0, 8.55) rectangle (15.60, 10.35);
\path[control] (0.0, 4.55) rectangle (15.60, 8.25);
\path[action] (0.0, 0.70) rectangle (15.60, 4.20);

\node[title, anchor=north west] at (0.18, 10.25) {Creation Entry Points};
\node[subtitle, anchor=north east] at (15.40, 10.22) {Start from scratch or reuse an existing tree};
\node[entrycard, minimum width=3.25cm] at (4.20, 9.38) {Root Node};
\node[entrycard, minimum width=4.60cm] at (11.35, 9.38) {Load Existing Tree\\From File};
\node[subtitle] at (7.80, 8.78) {The remaining entries are shown when inserting relative to an existing node.};

\node[title, anchor=north west] at (0.18, 8.15) {Control Nodes};
\node[subtitle, anchor=north east] at (15.40, 8.12) {Structural composition plus specialized task nodes};
\path[controlpanel] (0.25, 4.82) rectangle (7.55, 7.52);
\path[controlpanel] (8.05, 4.82) rectangle (15.35, 7.52);
\node[grouplabel, anchor=north west] at (0.45, 7.34) {Structural control};
\node[grouplabel, anchor=north west] at (8.25, 7.34) {Task / scene nodes};

\node[controlcard] at (2.10, 6.58) {Action Sequence};
\node[controlcard] at (5.70, 6.58) {Fallback Node};
\node[controlcard] at (2.10, 5.88) {Condition Node};
\node[controlcard] at (5.70, 5.88) {Goto Node};
\node[controlcard] at (3.90, 5.18) {Checkpoint Node};

\node[controlcard] at (9.90, 6.58) {Scene Action};
\node[controlcard] at (13.50, 6.58) {AI2R Node};
\node[controlcard] at (9.90, 5.88) {Door Traversal};
\node[controlcard] at (13.50, 5.88) {Building\\Exploration};

\node[title, anchor=north west] at (0.18, 4.10) {Action Primitives};
\node[subtitle, anchor=north east] at (15.40, 4.07) {[L/R] = separate Left and Right menu entries};
\path[actionpanel] (0.25, 1.05) rectangle (7.55, 3.55);
\path[actionpanel] (8.05, 1.05) rectangle (15.35, 3.55);
\node[grouplabel, anchor=north west] at (0.45, 3.44) {Side-agnostic};
\node[grouplabel, anchor=north west] at (8.25, 3.44) {Side-specific [L/R]};

\node[actioncard] at (2.10, 2.68) {Wait};
\node[actioncard] at (5.70, 2.68) {Mimic};
\node[actioncard] at (2.10, 2.03) {Neck};
\node[actioncard] at (5.70, 2.03) {Spine};
\node[actioncard] at (2.10, 1.40) {Pelvis};
\node[actioncard] at (5.70, 1.40) {Walk};

\node[actioncard] at (9.90, 2.68) {Arm};
\node[actioncard] at (13.50, 2.68) {Screw Primitive};
\node[actioncard] at (9.90, 2.03) {Ability Hand};
\node[actioncard] at (13.50, 2.03) {EZGripper};
\node[actioncard] at (9.90, 1.40) {Leg};

\end{tikzpicture}}
    \caption[Available behavior node library.]{%
        Available behavior node library in the RDX behavior editor.
        The menu is organized into creation entry points, reusable structural control nodes, specialized task and scene nodes, and primitive action nodes.}
    \label{fig:BehaviorNodeLibrary}
\end{figure}
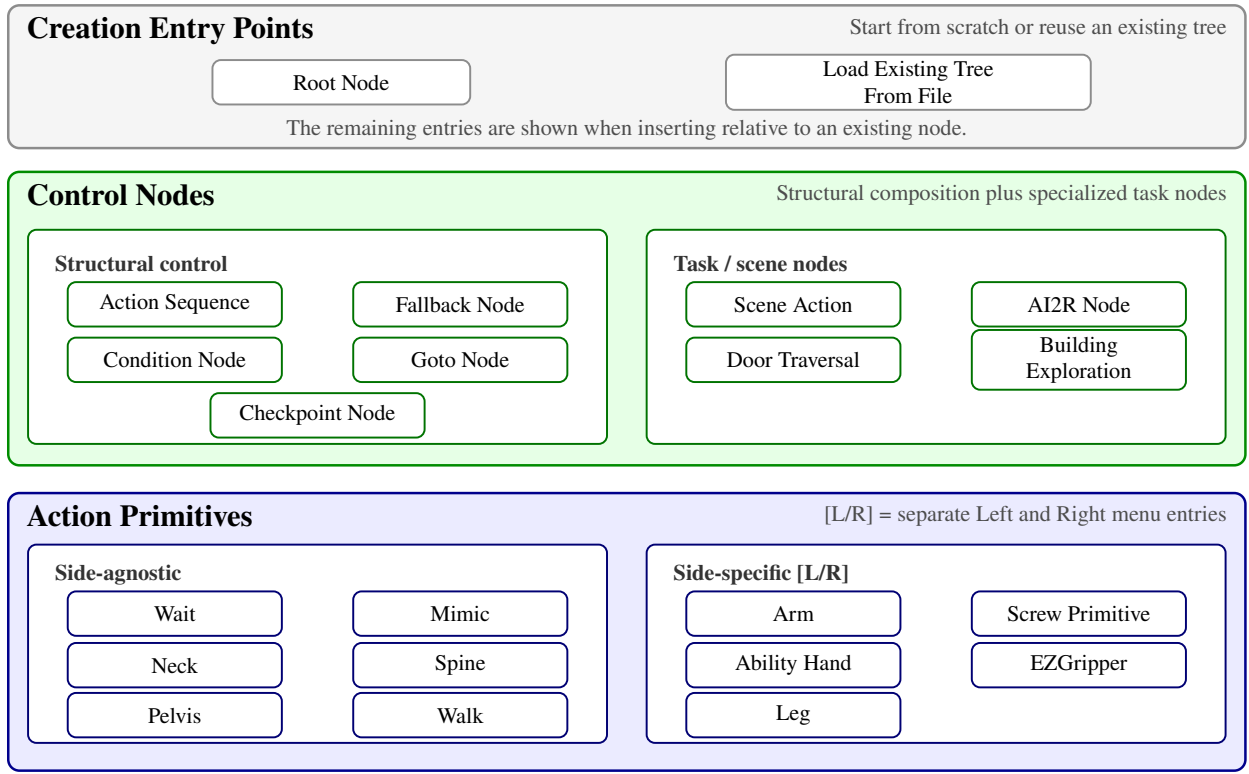

A full list of our node types is shown in \autoref{fig:BehaviorNodeLibrary}.
The functionalities of these nodes are covered in detail in \autoref{ch:tutorial}.

Here is a description of all node types:

\begin{itemize}
    \item \textbf{Action Sequence.} Sequential backbone of a behavior. Groups child leaves into a named routine without adding execution logic beyond standard tree traversal.
    \item \textbf{Fallback.} Splits child leaves into a primary try path and a recovery path. If the try path succeeds, recovery leaves are skipped; if it fails, execution jumps to the first catch leaf.
    \item \textbf{Condition.} Tests that gate or redirect execution without commanding motion. Counters, frame-to-frame proximity checks, shape containment checks, and explicit success or failure cases.
    \item \textbf{Goto.} Redirects execution to the next leaf or to a named target leaf elsewhere in the tree. Completes immediately, making retries, skips, and nonlocal jumps authorable at runtime.
    \item \textbf{Checkpoint.} Instantaneous named landmarks. Provides stable entry points, progress markers, and jump targets for external triggers or goto-based control flow.
    \item \textbf{Scene Action.} Manipulates behavior-time scene state rather than robot joints. Creates, refreshes, freezes, or deletes scene objects, clears the scene, or reconfigures detection and perception models.
    \item \textbf{Mimic Action.} Replays a recorded policy trajectory or requests a transition out of policy control. Replay is aligned to the robot's current mid-feet pose so the same recording can be reused from different starts.
    \item \textbf{AI2R.} Connects the tree to an external reasoning module through command and status topics. Publishes available behaviors, scene state, and failures, and can randomize repeated goto actions for data collection.
    \item \textbf{Door Traversal.} Domain-specific wrapper that inspects detected door geometry to choose the correct authored branch (closed versus ajar, push versus pull) and manages logging around specific walking segments.
    \item \textbf{Building Exploration.} Structural hook for larger-scale exploration behaviors, with little node-specific runtime logic beyond standard tree update.
    \item \textbf{Neck Action.} Commands head yaw and pitch over an authored duration for directing perception or the operator's viewpoint.
    \item \textbf{Spine Action.} Commands the torso by explicit spine joint angles or by an authored chest pose relative to a frame. Tracks joint or orientation error and can keep the torso pose anchored after completion.
    \item \textbf{Walk Action.} Main locomotion primitive. Executes manually placed footsteps or planner-generated routes from waypoints and goal frames, with preview planning and runtime progress tracking.
    \item \textbf{Arm Action.} Commands one arm using predefined joint targets or a hand palm pose relative to an authored frame. Solves IK, sends jointspace and/or taskspace commands, and monitors pose or joint error.
    \item \textbf{Screw Primitive Action.} Expresses hand motion as translation and rotation along a screw axis attached to an object frame, useful for constrained interactions such as turning or sliding. Builds a smooth trajectory while enforcing velocity limits.
    \item \textbf{Pelvis Action.} Adjusts body pose with an authored pelvis frame and duration to reshape reachability or stance.
    \item \textbf{Ability Hand Action.} Commands the six-actuator Ability Hand using grip presets or explicit joint targets and velocities. Completion uses waiting, per-joint tolerance, or cumulative motion, and the node can wiggle the hand if a grasp stalls.
    \item \textbf{EZ Gripper Action.} Commands the Sake gripper to a target opening with a grip-force limit. Checks calibration and aperture, then tracks knuckle motion until the requested configuration is reached.
    \item \textbf{Wait Action.} Timed pause synchronized to robot time, useful for settling and for spacing controller requests.
    \item \textbf{Leg Action.} Commands a single foot pose relative to an authored parent frame for direct swing-foot placement outside the full footstep planning pipeline.
\end{itemize}

\section{Behavior Scene}

The last major architectural component of our system is the behavior scene.
We maintain a set of persistent object detections local to the behaviors and for use only by the behaviors.
This exclusivity allows behaviors to have full control over the perception they depend on.
This is also important given the technical limitations of robot perception systems.

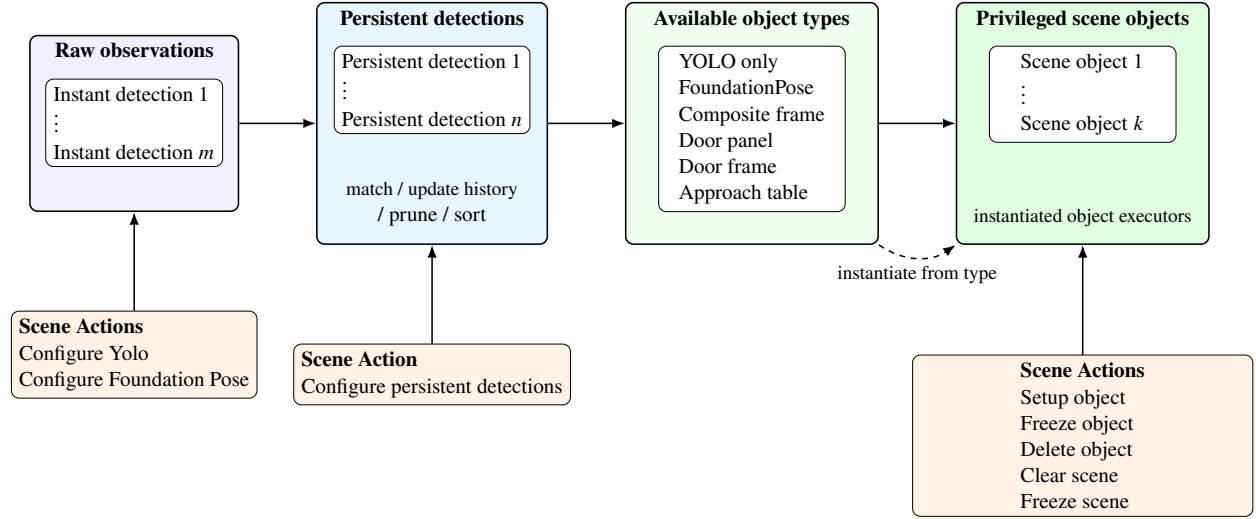
\begin{figure}[H]
    \centering
    \resizebox{\columnwidth}{!}{\begin{tikzpicture}[
    font=\small,
    >=Latex,
    node distance=6mm and 8mm,
    box/.style={draw, rounded corners, align=center, inner sep=4pt},
    area/.style={draw, rounded corners, thick, inner sep=6pt},
    flow/.style={->, thick},
    action/.style={draw, rounded corners, fill=black!5, align=left, inner sep=4pt},
    listbox/.style={draw, rounded corners, align=left, inner sep=4pt, fill=white},
    title/.style={align=center, font=\small\bfseries}
]

    \node[area, fill=blue!6, minimum width=38mm, minimum height=32mm] (raw) {};
    \node[title, anchor=north] at (raw.north) {Raw observations};

    \node[listbox, anchor=north, minimum width=28mm] (rawlist)
    at ([yshift=-8mm]raw.north)
        {Instant detection 1\\
    $\vdots$\\
    Instant detection $m$};

    \node[area, fill=cyan!8, minimum width=42mm, minimum height=44mm, right=14mm of raw] (persist) {};
    \node[title, anchor=north] at (persist.north) {Persistent detections};

    \node[listbox, anchor=north, minimum width=30mm] (plist)
    at ([yshift=-8mm]persist.north)
        {Persistent detection 1\\
    $\vdots$\\
    Persistent detection $n$};

    \node[align=center, anchor=south] at ([yshift=2mm]persist.south)
        {\footnotesize match / update history\\ / prune / sort};

    \node[area, fill=green!6, minimum width=46mm, minimum height=44mm, right=14mm of persist] (types) {};
    \node[title, anchor=north] at (types.north) {Available object types};

    \node[listbox, anchor=north, minimum width=34mm] (typelist)
    at ([yshift=-8mm]types.north)
        {YOLO only\\
    FoundationPose\\
    Composite frame\\
    Door panel\\
    Door frame\\
    Approach table};

    \node[area, fill=green!12, minimum width=46mm, minimum height=44mm, right=14mm of types] (objs) {};
    \node[title, anchor=north] at (objs.north) {Privileged scene objects};

    \node[listbox, anchor=north, minimum width=34mm] (objlist)
    at ([yshift=-8mm]objs.north)
        {Scene object 1\\
    $\vdots$\\
    Scene object $k$};

    \node[align=center, anchor=south] at ([yshift=2mm]objs.south)
        {\footnotesize instantiated object executors};

    \node[action, fill=orange!10, below=18mm of raw, minimum width=34mm] (rawact)
    {
        \textbf{Scene Actions}\\
        Configure Yolo\\
        Configure Foundation Pose
    };

    \node[action, fill=orange!10, below=18mm of persist, minimum width=34mm] (pconfig)
    {
        \textbf{Scene Action}\\
        Configure persistent detections
    };

    \node[action, fill=orange!10, below=20mm of objs, minimum width=62mm] (objsceneact)
    {
        \textbf{Scene Actions}\\
        Setup object\\
        Freeze object\\
        Delete object\\
        Clear scene\\
        Freeze scene
    };

    \draw[flow] (raw.east) -- (persist.west);
    \draw[flow] (persist.east) -- (types.west);
    \draw[flow] (types.east) -- (objs.west);

    \draw[flow] (rawact.north) -- ([xshift=0mm]raw.south);
    \draw[flow] (pconfig.north) -- ([xshift=0mm]persist.south);
    \draw[flow] (objsceneact.north) -- ([xshift=0mm]objs.south);

    \draw[flow, dashed] (types.south east) to[out=-35,in=-145] node[below, sloped] {\footnotesize instantiate from type} (objs.south west);

\end{tikzpicture}}
    \caption[Behavior scene overview.]{%
        The behavior scene consists of a list of active persistent detections and a privileged list of objects.
        Object types can be implemented to derive structured information about the scene.
        Scene actions are used to manage the privileged objects and the underlying perception models.}
    \label{fig:BehaviorScene}
\end{figure}

Since leveraging perception to accomplish tasks is difficult, we leave the creativity up to the human operator.
To do this, we have a library of scene actions which may be placed in the behavior tree and executed in the same way as the physical actions like moving a hand.
\autoref{fig:BehaviorScene} shows where the scene actions affect the behavior scene perception pipeline.
It also lists the current set of object and scene action types.

Our scene object types, listed in \autoref{fig:BehaviorScene}, include directly detected object types, such as YOLO-only and FoundationPose objects, but also derived types that compute new frames from persistent detections, existing scene objects, or depth data.
These specialized scene object types are a demonstration of how advanced perception and world modeling can be incorporated into our behavior system to extend capability to novel tasks.
In the next several sections, we will describe these derived types.

\section{Door Panel Object}

Our door panel object is a derived scene object built from two stable YOLO persistent detections: one for the opening mechanism and one for the door panel.
A setup object type scene action can be set up to create a door panel object after those detections have passed the usual stability and proximity checks.
An example is shown in \autoref{fig:unitree_door_panel}.


The object executor draws a line between the mechanism centroid and the panel centroid and uses that direction to define panel orientation.
This exploits the assumption that door panels swing on a vertical hinge.
The resulting frame is centered on the mechanism and establishes the mechanism side and the panel interior direction.
\autoref{fig:DoorPanelObjectAlgorithm} summarizes the computation and the resulting top-down frame convention.
Subsequent actions can then use this frame for grasp, unlatch, and opening motions.

\begin{figure}[H]
    \centering
    \resizebox{\columnwidth}{!}{\begin{tikzpicture}[
    font=\small,
    >=Latex,
    node distance=5mm and 7mm,
    flow/.style={->, thick},
    stepbox/.style={draw, rounded corners, align=center, inner sep=4pt, minimum width=28mm, fill=orange!8},
    outbox/.style={draw, rounded corners, align=center, inner sep=4pt, minimum width=30mm, fill=green!10},
    title/.style={font=\small\bfseries, align=center},
    note/.style={font=\footnotesize, align=center, text=black!70}
]

\node[stepbox] (mech) at (-2.4, 3.85) {Stable mechanism\\YOLO detection};
\node[stepbox, right=9mm of mech] (panel) {Stable door panel\\YOLO detection};
\node[stepbox, below=11mm of $(mech.south)!0.5!(panel.south)$] (centroids)
    {Centroids $C_m$, $C_p$};
\node[stepbox, below=7mm of centroids] (line)
    {Panel direction\\$\hat{d} = \widehat{C_p - C_m}$};
\node[outbox, below=7mm of line] (frame)
    {Panel object frame\\origin at $C_m$, $\hat{x}$ along $\hat{d}$};

\draw[flow] (mech.south) -- ++(0,-2.5mm) -| (centroids.north);
\draw[flow] (panel.south) -- ++(0,-2.5mm) -| (centroids.north);
\draw[flow] (centroids) -- (line);
\draw[flow] (line) -- (frame);

\node[note, anchor=north] at ($(frame.south)!0.5!(frame.south)+(0,-2.5mm)$)
    {Vertical-hinge assumption: panel swings about an axis normal to the plane};

\begin{scope}[shift={(5.5, 1.15)}]
    \node[title, anchor=south] at (2.30, 3.45) {Top-down schematic};

    \fill[gray!18, draw=gray!55, line width=0.7pt, rounded corners=2pt]
        (-0.45, 0.18) rectangle (5.05, 3.12);
    \coordinate (Cm) at (1.35, 0.95);
    \fill[orange!80!red] (Cm) circle (2.6pt);
    \node[font=\footnotesize, anchor=south east, text=orange!80!black] at ($(Cm)+(-1mm,1mm)$) {$C_m$};

    \coordinate (Cp) at (3.35, 1.25);
    \fill[blue!70!black] (Cp) circle (2.6pt);
    \node[font=\footnotesize, anchor=south west, text=blue!60!black] at ($(Cp)+(1mm,1mm)$) {$C_p$};

    \draw[blue!65!black, line width=0.9pt, -{Latex[length=2.2mm]}] (Cm) -- (Cp);
    \node[font=\footnotesize, anchor=south, text=blue!60!black] at ($(Cm)!0.55!(Cp)+(0,2.5mm)$)
        {$\hat{d}$};

    \draw[red!75!black, line width=0.85pt, -{Latex[length=2mm]}] (Cm) -- ++(0.95, 0.18);
    \draw[green!50!black, line width=0.85pt, -{Latex[length=2mm]}] (Cm) -- ++(-0.12, 0.95);
    \node[font=\footnotesize, text=red!70!black, anchor=west] at ($(Cm)+(1.0,0.18)$) {$\hat{x}$};
    \node[font=\footnotesize, text=green!40!black, anchor=south] at ($(Cm)+(-0.12,1.0)$) {$\hat{y}$};

    \node[note, anchor=north, text width=38mm] at (2.30, -0.09)
        {$\hat{x}$ points toward panel interior; used by grasp, unlatch, and opening actions};
\end{scope}

\end{tikzpicture}}
    \caption[Door panel object algorithm.]{%
        Door panel object algorithm.
        Stable YOLO persistent detections for the opening mechanism and door panel yield centroids $C_m$ and $C_p$.
        The vector from $C_m$ to $C_p$ defines panel orientation under the vertical hinge assumption.
        The privileged panel object frame is placed at $C_m$ with $\hat{x}$ pointing toward the panel interior.}
    \label{fig:DoorPanelObjectAlgorithm}
\end{figure}

\section{Door Frame Object}

The door frame object extends the door panel object with semantic door geometry estimated from the live depth image.
A door frame setup object type scene action requires an existing door panel object to be established in the scene.
It then derives the door frame pose, push or pull type, and door open angle from that panel state and the current point cloud.

\begin{figure}[H]
    \centering
    \resizebox{0.7\columnwidth}{!}{\begin{tikzpicture}[
    font=\small,
    >=Latex,
    post/.style={fill=gray!38, draw=gray!65, line width=0.8pt},
    panel/.style={fill=gray!16, draw=gray!55, line width=0.75pt},
    sweep/.style={dotted, line width=1pt, gray!55},
    searchcap/.style={draw=purple!55, fill=purple!12, line width=0.85pt},
    detectcap/.style={draw=green!50!black, fill=green!14, line width=0.85pt},
    sweeparrow/.style={->, line width=0.9pt, gray!65}
]

\def\panelW{4.0}
\def\panelT{0.22}
\def\openAngle{45}
\def\outerR{4.55}
\def\postSize{7mm}
\def\postGap{0.06}

\coordinate (H) at (0, 0);

\node[post, minimum size=\postSize, inner sep=0pt, anchor=south east] (hingepost)
    at (-\postGap, 0.42-7mm) {};
\node[post, minimum size=\postSize, inner sep=0pt, anchor=south west] (latchpost)
    at (\panelW+\postGap, 0.42-7mm) {};
\node[font=\footnotesize, anchor=east, text=black!65] at ($(hingepost.west)+(-1.5mm,0)$)
    {Hinge post};
\node[font=\footnotesize, anchor=west, text=black!65] at ($(latchpost.east)+(1.5mm,0)$)
    {Latch post};

\begin{scope}[rotate around={\openAngle:(H)}]
    \draw[panel] (0, -0.5*\panelT) rectangle (\panelW, 0.5*\panelT);
\end{scope}

\draw[sweep] (H) ++(8:\panelW) arc[start angle=8, end angle=82, radius=\panelW];

\coordinate (Cap) at ($(H)+(22.5:\outerR)$);
\draw[searchcap] (Cap) circle (0.27);
\node[font=\footnotesize, anchor=west, text=purple!60!black] at ($(Cap)+(0.32,0)$)
    {Search capsule};

\coordinate (DetectCap) at ($(H)+(0:\outerR)$);
\draw[detectcap] (DetectCap) circle (0.30);
\foreach \dx/\dy in {0/0, 1.4/0.9, -1.1/1.2, 1.6/-0.8, -1.4/-0.6, 0.7/1.4, -0.6/-1.1} {
    \fill[green!45!black] (DetectCap) ++(\dx mm, \dy mm) circle (0.6pt);
}

\draw[sweeparrow]
    (H) ++(\openAngle:\outerR) arc[start angle=\openAngle, end angle=8, radius=\outerR];
\draw[sweeparrow]
    (H) ++(\openAngle:\outerR) arc[start angle=\openAngle, end angle=82, radius=\outerR];

\end{tikzpicture}}
    \caption[Latch-side frame post search.]{%
        Latch-side frame post search (overhead view).
        The door panel is shown ajar at $45^\circ$ between hinge and latch frame posts.
        The latch-side edge sweeps along the dotted arc; curved arrows indicate search in either direction along but outside that arc.
        At each candidate angle a vertical search capsule (circle, top-down) counts contained depth points to detect the latch-side post.}
    \label{fig:DoorFrameObjectAlgorithm}
\end{figure}
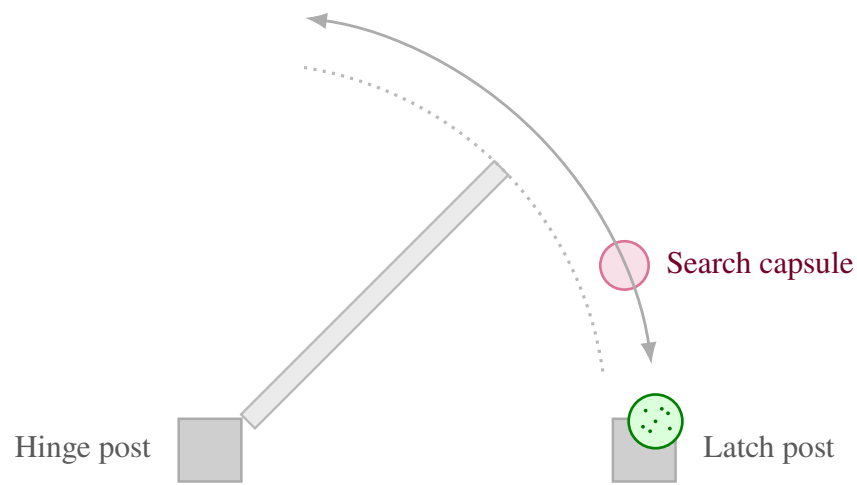

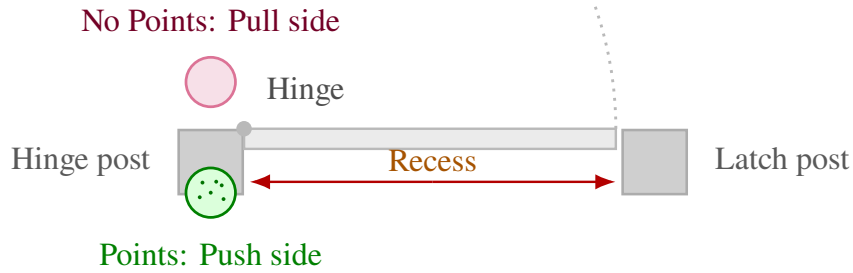
\begin{figure}[H]
    \centering
    \resizebox{0.7\columnwidth}{!}{\begin{tikzpicture}[
    font=\small,
    >=Latex,
    post/.style={fill=gray!38, draw=gray!65, line width=0.8pt},
    panel/.style={fill=gray!16, draw=gray!55, line width=0.75pt},
    recessarrow/.style={->, line width=0.75pt, red!70!black},
    checkcap/.style={draw=purple!55, fill=purple!12, line width=0.85pt},
    hingecap/.style={draw=green!50!black, fill=green!14, line width=0.85pt},
    openarc/.style={dotted, line width=0.9pt, gray!55}
]

\def\capR{0.27}

\def\panelW{4.0}
\def\panelT{0.22}
\def\postSize{7mm}
\def\postGap{0.06}
\def\recessH{0.12}
\def\openAngle{20}

\node[post, minimum size=\postSize, inner sep=0pt, anchor=south east] (hingepost)
    at (-\postGap, 0.42-7mm) {};
\node[post, minimum size=\postSize, inner sep=0pt, anchor=south west] (latchpost)
    at (\panelW+\postGap, 0.42-7mm) {};

\coordinate (InnerL) at ($(hingepost.north east)+(\postGap,0)$);
\coordinate (InnerR) at ($(latchpost.north west)+(-\postGap,0)$);

\node[font=\footnotesize, anchor=east, text=black!65] at ($(hingepost.west)+(-1.5mm,0)$)
    {Hinge post};
\node[font=\footnotesize, anchor=west, text=black!65] at ($(latchpost.east)+(1.5mm,0)$)
    {Latch post};

\coordinate (UpperCap) at ($(hingepost.center |- hingepost.north)+(0,{\capR+0.06+2*\capR/3})$);
\coordinate (LowerCap) at ($(hingepost.center |- hingepost.south)+(0,{-\capR+0.12+2*\capR/3})$);
\draw[checkcap] (UpperCap) circle (\capR);
\draw[hingecap] (LowerCap) circle (\capR);
\foreach \dx/\dy in {0/0, 1.2/0.8, -0.9/1.1, 1.4/-0.7, -1.2/-0.5, 0.6/1.2} {
    \fill[green!45!black] (LowerCap) ++(\dx mm, \dy mm) circle (0.6pt);
}
\node[font=\footnotesize, anchor=south, text=purple!60!black]
    at ($(UpperCap)+(0,\capR+4mm)$) {No Points: Pull side};
\node[font=\footnotesize, anchor=north, text=green!50!black]
    at ($(LowerCap)+(0,-\capR-4mm)$) {Points: Push side};

\coordinate (Hinge) at (hingepost.north east);
\coordinate (PanelBotR) at ($(InnerR |- Hinge)+(0,-\panelT)$);
\draw[panel] (Hinge) rectangle (PanelBotR);

\coordinate (LatchTop) at (InnerR |- Hinge);
\draw[openarc]
    let \p1=(Hinge), \p2=(LatchTop),
        \n1={veclen(\x2-\x1,\y2-\y1)},
        \n2={atan2(\y2-\y1,\x2-\x1)} in
    (\p1) ++(\n2:\n1) arc[start angle=\n2, end angle=\n2+\openAngle, radius=\n1];

\coordinate (RecessMid) at ($(InnerL)!0.5!(InnerR) + (0,{-0.04-0.5*\recessH-3*\panelT+1.5*\recessH})$);
\node[font=\footnotesize, anchor=south, text=orange!65!black] at ($(RecessMid)+(0,-0.03)$)
    {Recess};

\draw[recessarrow] (RecessMid) -- (InnerL |- RecessMid);
\draw[recessarrow] (RecessMid) -- (InnerR |- RecessMid);

\fill[gray!55] (Hinge) circle (2.4pt);
\node[font=\footnotesize, anchor=south west, text=black!70] at ($(Hinge)+(1mm,1mm)$)
    {Hinge};

\end{tikzpicture}}
    \caption[Closed-door recess check.]{%
        Push-pull side detection on a closed door (overhead view).
        On the push side, the panel is inset within the frame posts.
        On the hinge/pull side, there is no such recess.
        Looking for depth points in key positions relative to the hinge, as shown, can reveal which side of the door the robot is on.}
    \label{fig:DoorFrameRecessCheck}
\end{figure}

\begin{figure}[H]
    \centering
    \includegraphics[width=1.0\columnwidth]{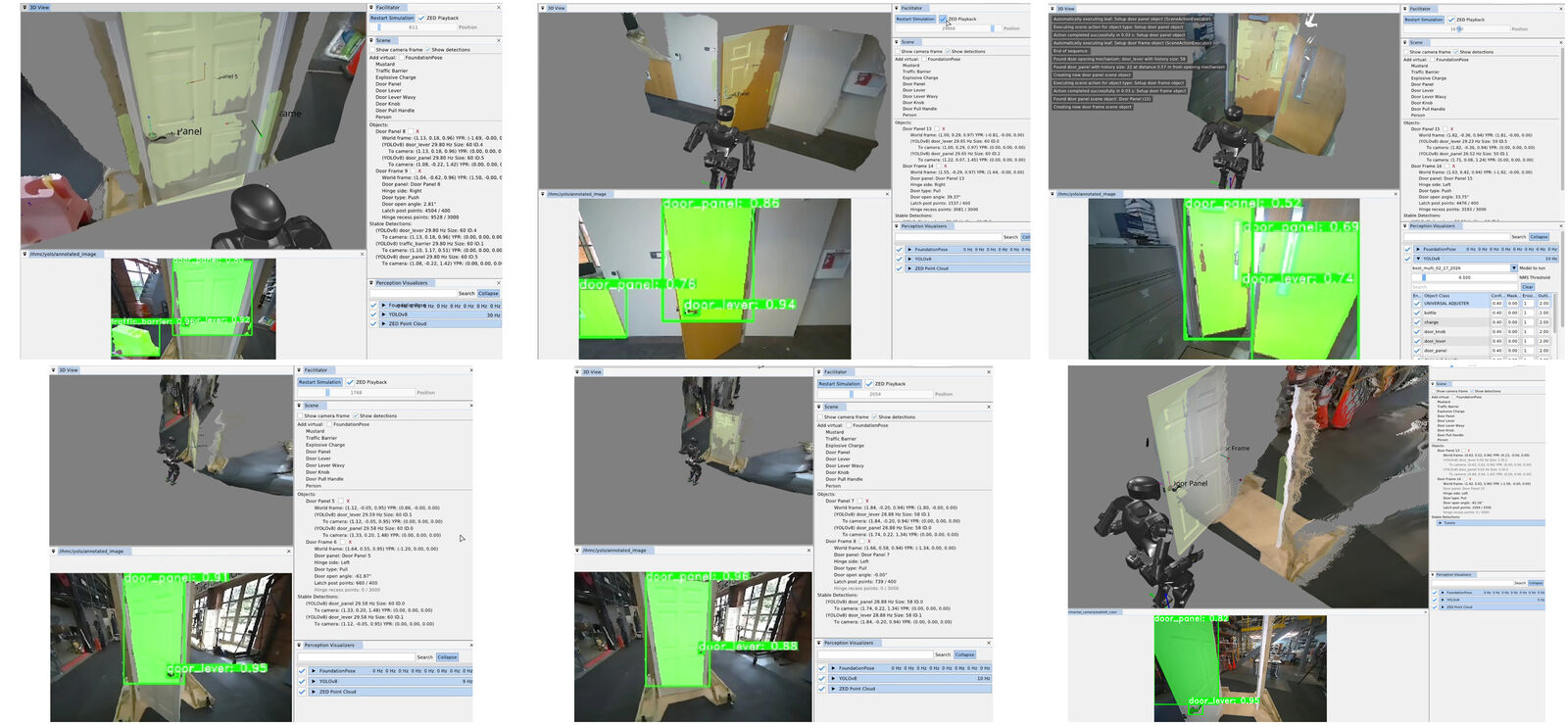}
    \caption[Visual door state estimation across six door configurations.]{%
        Visual door state estimation across six configurations.
        A transparent yellow-green model shows the estimated panel pose, a coordinate frame is rendered at the opening mechanism, and vertical 3D capsules visualize the post and recess checks.
        Top: a closed lab left-push door, a real left-pull door at $39^\circ$, and a real right-push door of a double door at $33^\circ$.
        Bottom: a lab right-pull door at $61^\circ$, the same closed, and at $81^\circ$ viewed from the back side.}
    \label{fig:DoorStateEstimation}
\end{figure}

The estimator places a nominal hinge location at a fixed offset from the mechanism along the panel width.
\autoref{fig:DoorFrameObjectAlgorithm} illustrates our algorithm, which sweeps a vertical 3D capsule around a frame existence hypothesis in the point cloud and, at each candidate angle, uses a CUDA points-in-shape counter to search for the latch-side frame post.
As soon as a tunable sufficient number of points are detected within the capsule, that angle is used to define the door frame plane.
Hinge side is inferred from the panel frame relative to the robot viewpoint.
When the panel is clearly open, the sign of the opening angle relative to the hinge side determines push or pull side.
When the door is closed, we use a second vertical capsule depth point check to detect the presence of a door panel recess, which indicates a push door when the recess is present and a pull door when it is not, as illustrated in \autoref{fig:DoorFrameRecessCheck}.
\autoref{fig:DoorStateEstimation} shows the estimator across closed, ajar, and widely open doors, both hinge configurations, push and pull, and a backside view of the panel.

\section{Composite and Hybrid Frames}

A composite frame is a perception-less object derived from two existing scene frames.
The reason for this is to perform geometric calculations to provide more useful reference frames.
Like other privileged objects, it is created by a setup-object scene action and can be referenced by walk, arm, spine, and other physical actions in the same way as a detected object frame.
\autoref{fig:alex_composite_frame} shows a representative door setup sequence: clear scene, setup door panel, setup door frame, and setup hybrid frame, with the resulting privileged objects visible in the scene panel.

There are currently two composite frame subtypes: an approach frame and a hybrid frame.
Both are parameterized by two source frames by name.
The approach frame orients its $x$ axis to face from frame~A toward frame~B and places its origin on the line segment between them at a tunable distance from frame~B.
This makes it suitable for walking toward a task while stopping short of collision with the object being approached.
A hybrid frame takes its position from frame~A and its orientation from frame~B.
It is useful for approaching an ajar door panel, where the robot should walk with the orientation of the door frame but approach the position of the opening mechanism.

Composite frames can be layered.
For example, an ajar-door hybrid frame can serve as one input to a subsequent approach frame that defines a distant handle approach stance.
The top-level door traversal behavior in \autoref{fig:TreeStructure} constructs panel, frame, and hybrid-frame objects in this way before selecting a door specific subtree.

\section{Approach Table Object}

The approach table object applies the same scene object pattern outside the door domain.
It is a heuristic derived object that does not rely on semantic detections.
Instead, it operates directly on the live depth image to produce a reusable table edge approach frame.
This is a novel algorithm that we have demonstrated to allow humanoid robots to approach tables with centimeter accuracy.
This is important because it enables the robot to get close enough to enable arm reachability on the table's workspace without colliding with the table.

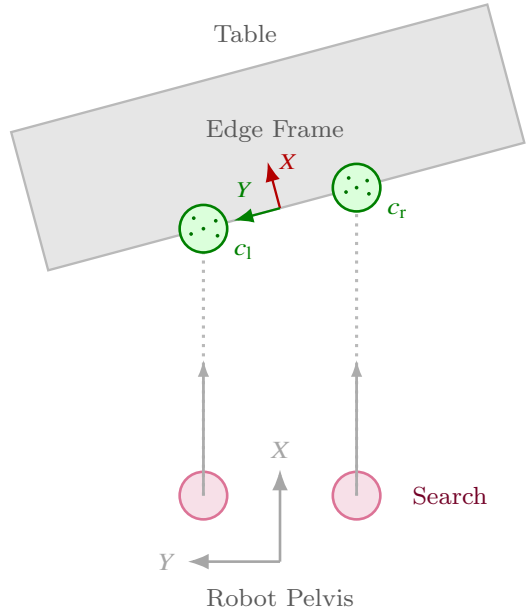
\begin{figure}[H]
    \centering
    \resizebox{0.5\columnwidth}{!}{
\newcommand{\TableEdgeFigFont}{\fontsize{9}{10.8}\selectfont}
\newcommand{\TableEdgeFigLabelFont}{\fontsize{8}{9.6}\selectfont}
\newcommand{\TableEdgeFigAxisFont}{\fontsize{7}{8.4}\selectfont}

\begingroup
\fontencoding{OT1}\fontfamily{cmr}\selectfont
\fontsize{10}{12}\selectfont
\begin{tikzpicture}[
    font=\TableEdgeFigFont,
    >=Latex,
    table/.style={fill=gray!20, draw=gray!55, line width=0.75pt},
    sweep/.style={dotted, line width=1pt, gray!55},
    sweeparrow/.style={-{Latex[length=2.0mm,width=1.3mm]}, line width=0.9pt, draw=gray!65},
    searchcap/.style={draw=purple!55, fill=purple!12, line width=0.85pt},
    detectcap/.style={draw=green!50!black, fill=green!14, line width=0.85pt},
    axis/.style={->, line width=0.85pt, gray!70},
    approachaxis/.style={->, line width=0.75pt}
]

\def\capR{0.27}
\def\tableW{5.6}
\def\nearHalfD{0.275}
\def\farHalfD{1.35}
\def\tableAngle{15}
\def\tableCenterY{4.3}
\def\searchForward{0.75}
\def\searchSide{0.87}
\def\approachAxisLen{0.546}
\def\labelClear{6pt}
\def\sweepArrowLen{1.55}

\coordinate (Pelvis) at (0, 0);

\draw[axis] (Pelvis) -- ++(0, 1.05) node[anchor=south, font=\TableEdgeFigLabelFont] {$X$};
\draw[axis] (Pelvis) -- ++(-1.05, 0) node[anchor=east, font=\TableEdgeFigLabelFont] {$Y$};
\node[font=\TableEdgeFigLabelFont, anchor=north, text=black!65] at ($(Pelvis)+(0,-0.18)$)
    {Robot Pelvis};

\coordinate (TableCenter) at (0, \tableCenterY);

\pgfmathsetmacro{\tableAngleRad}{\tableAngle}
\pgfmathsetmacro{\halfW}{0.5*\tableW}
\pgfmathsetmacro{\cosA}{cos(\tableAngleRad)}
\pgfmathsetmacro{\sinA}{sin(\tableAngleRad)}
\coordinate (TableNearL) at ($(TableCenter)+({-\halfW*\cosA+\nearHalfD*\sinA},{-\halfW*\sinA-\nearHalfD*\cosA})$);
\coordinate (TableNearR) at ($(TableCenter)+({\halfW*\cosA+\nearHalfD*\sinA},{\halfW*\sinA-\nearHalfD*\cosA})$);
\path[name path=TableNearEdge, overlay] (TableNearL) -- (TableNearR);

\begin{scope}[rotate around={\tableAngle:(TableCenter)}]
    \draw[table] ($(TableCenter)+(-0.5*\tableW,-\nearHalfD)$)
        rectangle ($(TableCenter)+(0.5*\tableW,\farHalfD)$);
    \node[font=\TableEdgeFigLabelFont, anchor=south, text=black!65]
        at ($(TableCenter)+(0,\farHalfD+0.18)$) {Table};
\end{scope}

\coordinate (StartL) at ($(Pelvis)+(-\searchSide,\searchForward)$);
\coordinate (StartR) at ($(Pelvis)+(\searchSide,\searchForward)$);
\draw[searchcap] (StartL) circle (\capR);
\draw[searchcap] (StartR) circle (\capR);
\node[font=\TableEdgeFigLabelFont, anchor=west, text=purple!60!black]
    at ($(StartR)+(\capR+5mm,0)$) {Search};

\path[name path=SweepL, overlay] (StartL) -- ++(0, 8);
\path[name path=SweepR, overlay] (StartR) -- ++(0, 8);
\path[name intersections={of=SweepL and TableNearEdge, by=DetectL}];
\path[name intersections={of=SweepR and TableNearEdge, by=DetectR}];

\draw[sweep] (StartL) -- (DetectL);
\draw[sweep] (StartR) -- (DetectR);
\draw[sweeparrow] (StartL) -- ++(0, \sweepArrowLen);
\draw[sweeparrow] (StartR) -- ++(0, \sweepArrowLen);

\draw[detectcap] (DetectL) circle (\capR);
\draw[detectcap] (DetectR) circle (\capR);
\node[font=\TableEdgeFigLabelFont, anchor=north west, text=green!50!black]
    at ($(DetectL)+(\capR+\labelClear,-\capR-1.2pt)$) {$c_\mathrm{l}$};
\node[font=\TableEdgeFigLabelFont, anchor=north west, text=green!50!black]
    at ($(DetectR)+(\capR+\labelClear,-\capR-1.2pt)$) {$c_\mathrm{r}$};
\foreach \cap/\dx/\dy in {DetectL/0/0, DetectL/1.4/0.9, DetectL/-1.1/1.2, DetectL/1.6/-0.8, DetectL/-1.4/-0.6, DetectR/0/0, DetectR/1.2/0.8, DetectR/-0.9/1.1, DetectR/1.4/-0.7, DetectR/-1.2/-0.5} {
    \fill[green!45!black] (\cap) ++(\dx mm, \dy mm) circle (0.6pt);
}

\coordinate (ApproachOrigin) at ($(DetectL)!0.5!(DetectR)$);
\begin{scope}[shift={(ApproachOrigin)}, rotate=\tableAngle]
    \node[font=\TableEdgeFigLabelFont, anchor=south, text=black!65, yshift=12pt] at (0, 0.22) {Edge Frame};
    \draw[approachaxis, red!70!black] (0, 0) -- (0, \approachAxisLen)
        node[font=\TableEdgeFigAxisFont, anchor=west, text=red!70!black] {$X$};
    \draw[approachaxis, green!50!black] (0, 0) -- (-\approachAxisLen, 0)
        node[font=\TableEdgeFigAxisFont, midway, above=2pt, xshift=-3.85pt, text=green!50!black] {$Y$};
\end{scope}

\path[use as bounding box]
    ([xshift=-4mm,yshift=-3mm]current bounding box.south west)
    rectangle
    ([xshift=4mm,yshift=3mm]current bounding box.north east);

\end{tikzpicture}
\endgroup}
    \caption[Table edge detection search.]{%
        Table edge detection (overhead view).
        Two vertical search capsules start left and right of the robot pelvis and sweep forward along $+X$ until each contains enough depth points on the table near edge at $c_\mathrm{l}$ and $c_\mathrm{r}$.
        The red and green axes show the computed approach frame from \autoref{eq:approach_table_frame}.}
    \label{fig:TableEdgeDetectionAlgorithm}
\end{figure}

The algorithm is as follows.
Two vertical capsules, one on each side of the robot, sweep forward from the pelvis in the mid-feet-under-pelvis frame at a height where table edges are expected.
The capsules start near knee height and end just below chest height so that tables of different heights can be handled with one parameterization.
Our CUDA point counter measures depth point containment inside each capsule at every step.
Each side continues sweeping until enough points are found or the search limit is reached.
Once both sides have located the edge, the approach frame is computed from the left and right detection points $c_\mathrm{l}$ and $c_\mathrm{r}$.
Let $c_\mathrm{l}, c_\mathrm{r} \in \mathbb{R}^2$ denote the ground-plane positions of the left and right capsule centers in the mid-feet-under-pelvis frame after each sweep finds enough table points.
In this frame, $+Y$ points to the robot's left, so the left capsule is placed at $+Y$ and the right capsule at $-Y$.
The edge origin and orientation are
\begin{equation}
    \begin{aligned}
        p &= \tfrac{1}{2}(c_\mathrm{l} + c_\mathrm{r}), \\
        \hat{y} &= \frac{c_\mathrm{l} - c_\mathrm{r}}{\|c_\mathrm{l} - c_\mathrm{r}\|}, \\
        \hat{x} &= \begin{bmatrix} \hat{y}_y \\ -\hat{y}_x \end{bmatrix},
    \end{aligned}
    \label{eq:approach_table_frame}
\end{equation}
where $\hat{y}$ points along the table edge toward the robot's left and $\hat{x}$ is the in-plane perpendicular chosen to point into the table.
In three dimensions, the translation uses $p$ with $z$ set to the current mid-feet height, and the rotation aligns to $\hat{x}$, $\hat{y}$, and world-up $\hat{z}$.
This enables tracking the table edge regardless of the table height.
\autoref{fig:TableEdgeDetectionAlgorithm} summarizes the overhead search geometry.

\begin{figure}[H]
    \centering
    \begin{minipage}[t]{0.485\columnwidth}
        \centering
        \includegraphics[width=\linewidth]{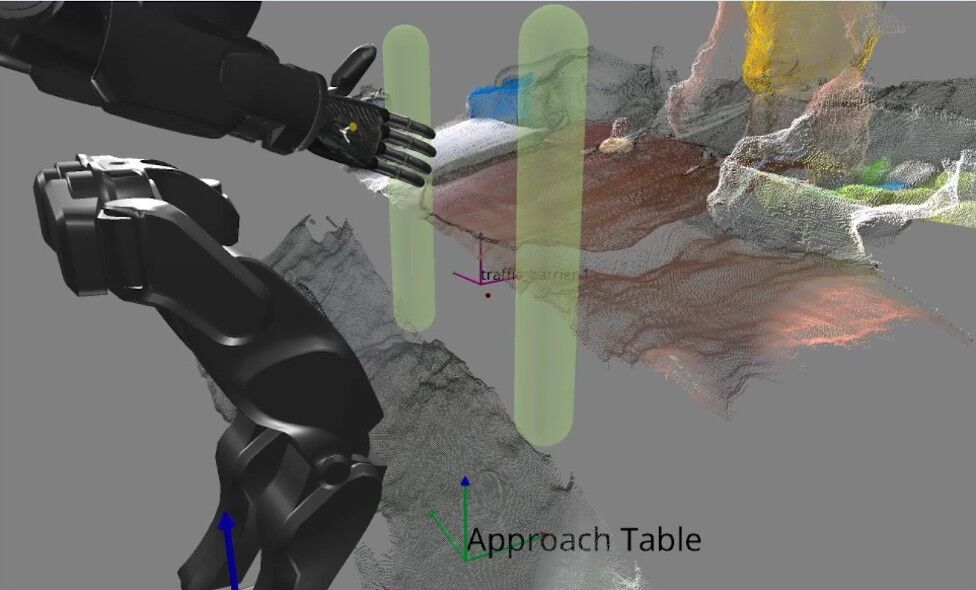}
    \end{minipage}\hfill
    \begin{minipage}[t]{0.485\columnwidth}
        \centering
        \includegraphics[width=\linewidth]{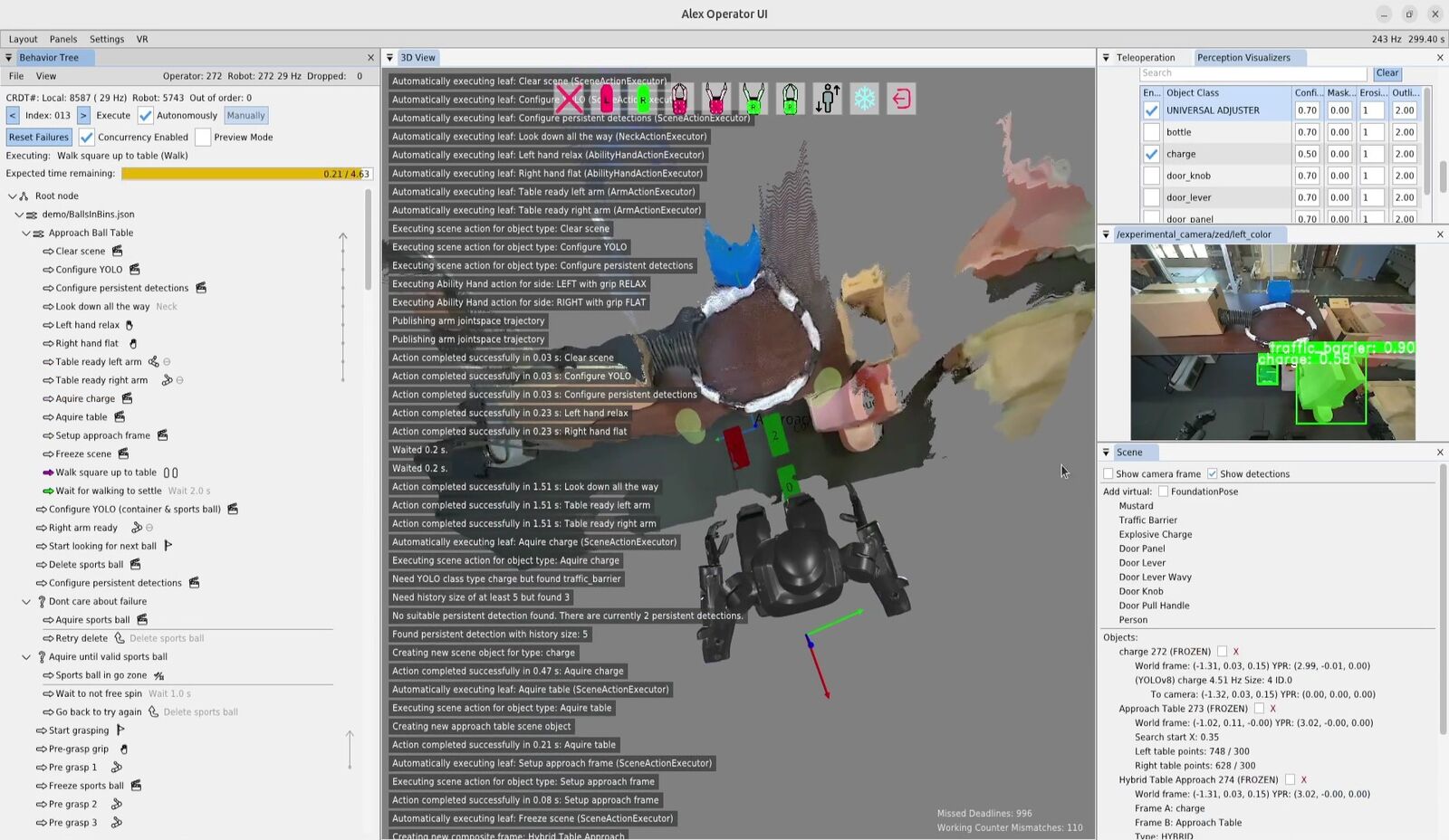}
    \end{minipage}
    \caption[Table approach scene object and affordance frame.]{%
        The table approach scene object and affordance frame.
        Left: two vertical capsules sweep forward at the expected table edge height until they contain enough depth points to locate the edge and determine its orientation.
        The edge pose is projected onto the ground to create a table approach frame that is invariant to table height for footstep authoring.
        Right: the operator UI view showing the detected capsules, the generated approach frame, and a footstep plan authored relative to that frame to square up to the table.}
    \label{fig:TableApproachAffordance}
\end{figure}

The point threshold and search limits are editable in the scene-action settings.
Subsequent walk actions use the resulting frame to square up to the table edge before manipulation.
This feature supports multi-station sorting, ball return, and other loco-manipulation tasks where the robot must approach a work surface reliably before reaching for objects on it.
\autoref{fig:TableApproachAffordance} shows the detected capsules, the generated approach frame, and a footstep plan authored relative to that frame.

In the next chapter, we'll dive into the specific functionalities of all these architectural features with a hands-on guide to creating and modifying behaviors using our system.

\chapter{Usage Guide}
\label{ch:tutorial}

\subsection{Overview}
The behavior system presented in this thesis can be used to author fast, resilient, and adaptable loco-manipulation behaviors for humanoid robots.
Examples include door traversals and sorting objects on tables.
This is in the pursuit of automating useful work and providing robotic drop-in replacements for humans for dull, dirty, or dangerous jobs.
In this chapter, we'll cover the system's limitations, prerequisites, and how to use it.

\section{Limitations}

\subsection{Perception}
We've made a lot of progress and performed some compelling demonstrations, but there is still a way to go.
In 2026, we've had an uptick in the expressiveness of our behaviors demonstrated on Alex through authorable perceptive reactivity and an expanded set of YOLO models.
However, we still do not have reliable object pose perception, which is why we have done so many demos with colored balls.
We have integrated FoundationPose for this, but it still needs a bit of work.
It is still possible to manipulate asymmetric objects if assumptions can be made about their initial state.

\subsection{Platform Bringup}
Another main limitation is that, to get this working on a new robot platform, there is significant bringup work.
Computer software and hardware engineering is required to bring up support for different sensors, hands, or controllers that your robot may have.

Another major limitation of the current implementation is the lack of navigational components.
It is possible to create room searching behaviors by turning in place and walking through doors or walking to objects if you see them, but there is no functionality in the scene for storing semantic topological maps or high-fidelity detailed map models.

\subsection{Behavior Tree Limits}
There are some smaller limitations and user frustrations that are present as well.
One of the bigger ones is the lack of the ``GOSUB'' functionality, where a sequence can be put on the stack and execution returned when it completes.
All we have at the moment is GOTOs.
Another is that when nested JSON-backed behaviors are included in multiple places in the tree, modifying one does not modify the other, and they will overwrite each other when saved.
We would like to fix these soon.

\section{Prerequisites}

\subsection{Robot Platform and Controller}
We have run our system with several humanoid robots over the years, including the Boston Dynamics DARPA Robotics Challenge Finals-Era Atlas, NASA's Valkyrie, IHMC and Boardwalk Robotics' Nadia, Unitree's H1-2, and IHMC's Alex.
Our system is centered on the humanoid form factor with footstep, walk, arm, leg, spine, and neck actions.
It is also centered on the presence of a whole body controller that can achieve those actions.
Ideally, the whole body controller can triage and queue asynchronous commands of those actions, to support concurrently tracked motions of different parts of the body that may start and end at different times.
For communication, we have been using a ROS 2 compatible DDS implementation, but other communication protocols could be subbed in with engineering work.
The controller could also be operated through a programmatic API and some of the interprocess comms avoided.

To get the best performance, the robot's controller should have push-tolerant standing and walking.
Even better if the robot can walk with the arms held up in front of the body.
If it can do this, it would be sufficient for basic multi-station loco-manipulation tasks and door traversals.

\subsection{Motion and Reachability}
We think longer arms generally work better than shorter ones.
A human-proportioned robot arm yields significantly less reachability than a human arm since humanoid robots typically don't have clavicle joints which extend a human's arm reach.
We tend to think that robot arms that are long enough such that the hands touch the knees when they are hanging are a good balance.
DRC Atlas and Nadia had these longer-arm proportions, but Alex didn't and we felt that it limited reachability and increased task complexity for the pull door behaviors and when reaching across tables.

Similarly, a high degree of freedom spine can help simplify behaviors.
For example, being able to yaw 90 degrees to the left and right can allow for faster scanning behaviors, increase arm reachability, and avoid unnecessary footsteps in tight spaces.

\subsection{Perception and Teleoperation}
On the perception side, the most significant requirements are a color image, a high-accuracy depth image, and the availability of high quality YOLO models for any objects you want to interact with.
There are many sensors available on the market that satisfy this requirement.
We have tried the ZED X Mini and the RealSense D457~\cite{intel2022d457}.
We think both of those would work.

Our preference and the one we currently use is the ZED X Mini because it has a stereo RGB camera pair which is useful for direct VR teleoperation.
To author complicated behaviors, an idea ahead of time of how the robot will accomplish the task is desirable.
With humanoid robots, this is helped by our natural intuition since the robot shares our form factor, but how the robot's capabilities differ from our own can be unintuitive.
This is why teleoperating a task can be a good way to quickly experiment with task strategies that work within the bounds of the robot's capabilities, and why we chose the ZED X Mini.
The ZED X Mini also has a human-like inter-pupillary distance, which makes for a natural robot embodiment by a VR teleoperator.


The other major thing you will need is a library of YOLO models for everything you need to interact with.
We have been using YOLOv8 because it does detections and segmentation for objects of interest.
Arghya Chatterjee crafted the models we used, but it took him weeks and weeks of hand labeling and a year or so of experimentation to achieve high confidence detections.
It might be good to find a vendor that provides professionally trained models.
Typically the data needed is around five videos of the task, each on a different day and time, light and dark, where the videos orbit around the object and get close and far from it.

\subsection{Hands}
For robot hands, we recommend a 5-finger anthropomorphic hand.
We had a great experience with the PSYONIC Ability Hand~\cite{psyonic_ability_hand}.
One major advantage of a 5-finger anthropomorphic hand is that, similar to the rest of the robot's body, it mirrors the human form, which allows for an increased level of intuition in task planning.
For example, a good start for planning how to get the robot to do a task is to just do it yourself and look at how you naturally do it.

Furthermore, this extends to teleoperation, where you can constrain the set of possible solutions to what is achievable by the robot while still leaning on natural intuition.
We use the Valve Index Controller's~\cite{valve_index_controllers} API to teleoperate the Ability Hands and it works great.
In fact, the OpenVR software maps the Valve Index Controller's finger estimation to six degrees of freedom which happens to match perfectly to the six degrees of freedom on the Ability Hand.

\subsection{Compute}
Another thing you'll need are NVIDIA GPUs if you want to use the ZED camera or if you want to use CUDA kernels.
The ZED SDK requires NVIDIA GPUs to use its most essential features such as its neural-assisted stereo depth estimation.
Additionally, we use CUDA heavily throughout the system, including for counting points and averaging color inside virtual shapes in the 3D point cloud, faster OpenCV functions, processing the output tensor of YOLO, non-maximum suppression of YOLO bounding boxes, and extraction of 3D points from the YOLO segmentation masks.

We use a Jetson AGX Orin on-board the robot and a desktop computer with an NVIDIA Turing GPU for the operator computer.
We run Linux on both except when doing VR teleoperation, when we use Windows.
We have had success with VR on Linux in simulation tests, but not on the real robot.
For VR headsets we've been using the Valve Index and the HTC Vive Focus 3.


In the next several sections we will walk through some token behavior authoring examples while explaining how system components work as we encounter them.
This is a guide to how the system works, but it's meant for reading and understanding more than it's meant to be literally followed as a tutorial.
We'll first use simulation examples to illustrate and explain some basic mechanisms without noise.
Then, we'll go through a real-robot authoring session: our 32 minute authoring session to get Unitree H1-2 opening a door repeatedly.

\section{Basic Example: Move the Arms and Walk}

\subsection{Simulation Setup}
To start, we'll be working in a simulation environment that we call the ``behavior test facilitator''.
It allows an operator to exercise system functionality without a vision and physics accurate simulator and without addressing any sim-to-real gaps.
We use a kinematics-only simulation of the robot that plays back the nominal motions without feedback from a physical environment.
In effect, this means that the desireds ultimately get set to the actuals for free motions and a special heuristic is used to hold the feet in place when ground contact is expected.

\begin{figure}[H]
    \centering
    \includegraphics[width=.95\columnwidth]{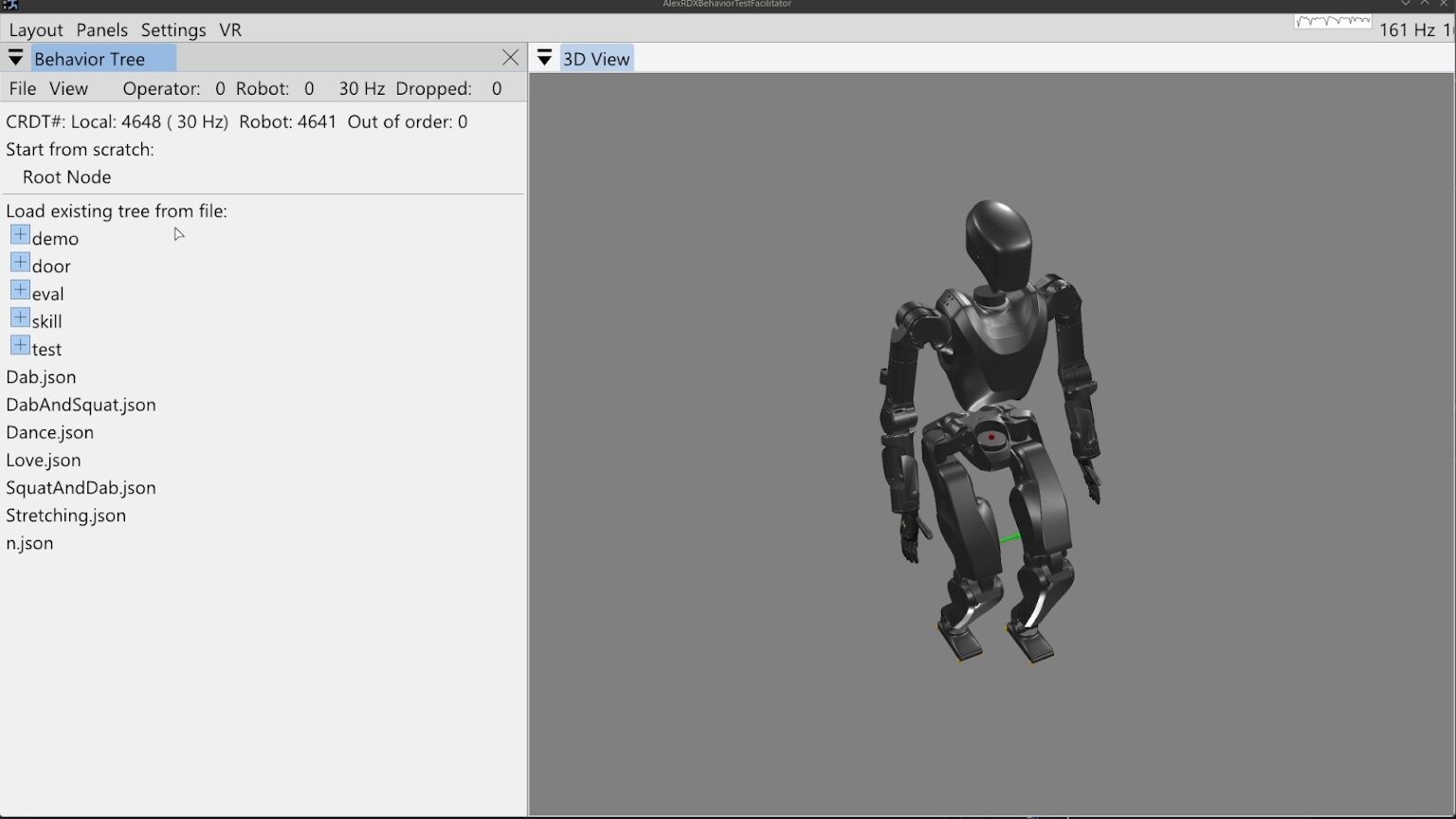}
    \caption{The behavior test facilitator on startup.
        A video of this example is available at \url{https://youtu.be/e3fk-EAJ5EQ}.
    }
    \label{fig:guide_sim_initial}
\end{figure}

To open the program, we will run the Java program \texttt{\href{https://github.com/ihmcrobotics/ihmc-open-robotics-software/blob/53f47de86106e996b657dae1c6a6fd0bd76e63ec/zulu/src/test/java/us/ihmc/zulu/rdx/ZuluRDXBehaviorTestFacilitator.java}{AlexRDXBehaviorTestFacilitator}}.
The initial view is shown in \autoref{fig:guide_sim_initial}.
The behavior tree panel is on the left and the 3D view is on the right.
The behavior tree panel is where an interactive model of the currently loaded behavior tree is rendered.
The 3D view renders the current, live robot state as an opaque, realistically colored robot model alongside virtual interactable elements that model planned actions and virtual scene elements.

\subsection{3D Viewport Camera}
The 3D view camera is controlled using the scheme developed and used in the DARPA Robotics Challenge era user interface.
We use this ``focus based'' camera view control algorithm because it simplifies getting the camera where you need it so it can be done quickly.
Since robots typically exist in 2D spaces under gravity, the most common translation camera movements are in the X-Y plane.
Additionally, our tasks most often focus on controlling a robot or inspecting specific things positioned in the environment.
For this reason, our ``focus based'' camera is based on a movable focus point in 3D space, meant to be located at the thing you are inspecting or monitoring.
The camera always faces this focal point and in the user interface it is represented as a small red sphere which is resized dynamically in 3D to be a constant small size in screen space, so it doesn't get huge or too small to see.
The keyboard controls, W, A, S, and D, translate the camera on the current X-Y plane that the focus point resides on.
The Q and Z keys move the focal point up and down along the world frame Z axis.
The camera orbits the focal point using a longitude and latitude.
The longitude freely loops 360 degrees around the focal point about its world Z axis.
The latitude, however, is bounded by +/- 90 degrees above and below the focal sphere's ``equator''.
For example, the limits are top down and bottom up, preventing the view from going upside down on the other side.
The full set of 3D view camera controls are presented in \autoref{tab:focus_based_camera_keys}.

\begin{table}[H]
    \centering
        \renewcommand{\arraystretch}{1.2}
        \begin{tabular}{ll}
            \hline
            \textbf{Focus based camera} & \textbf{Key} \\
            \hline
            Drag to orient camera & Left mouse \\
            Drag to pan camera & Middle mouse \\
            Fine adjustment & Shift \\
            Move camera back & S \\
            Move camera down & Z \\
            Move camera forward & W \\
            Move camera left & A \\
            Move camera right & D \\
            Move camera up & Q \\
            Zoom camera in & C \\
            Zoom camera in / out & Mouse scroll \\
            Zoom camera out & E \\
            \hline
        \end{tabular}
        \caption{Focus based camera keyboard shortcuts}
        \label{tab:focus_based_camera_keys}
\end{table}

\subsection{Loading the Tree}
On startup, no behavior tree is actively loaded, so the ``Load existing tree from file menu'' is shown.
There are two options from here: starting from scratch with a new root node or loading an existing behavior from file.
In this example, we'll start from scratch and click ``Root Node''.

\begin{figure}[H]
    \centering
    \includegraphics[width=.95\columnwidth]{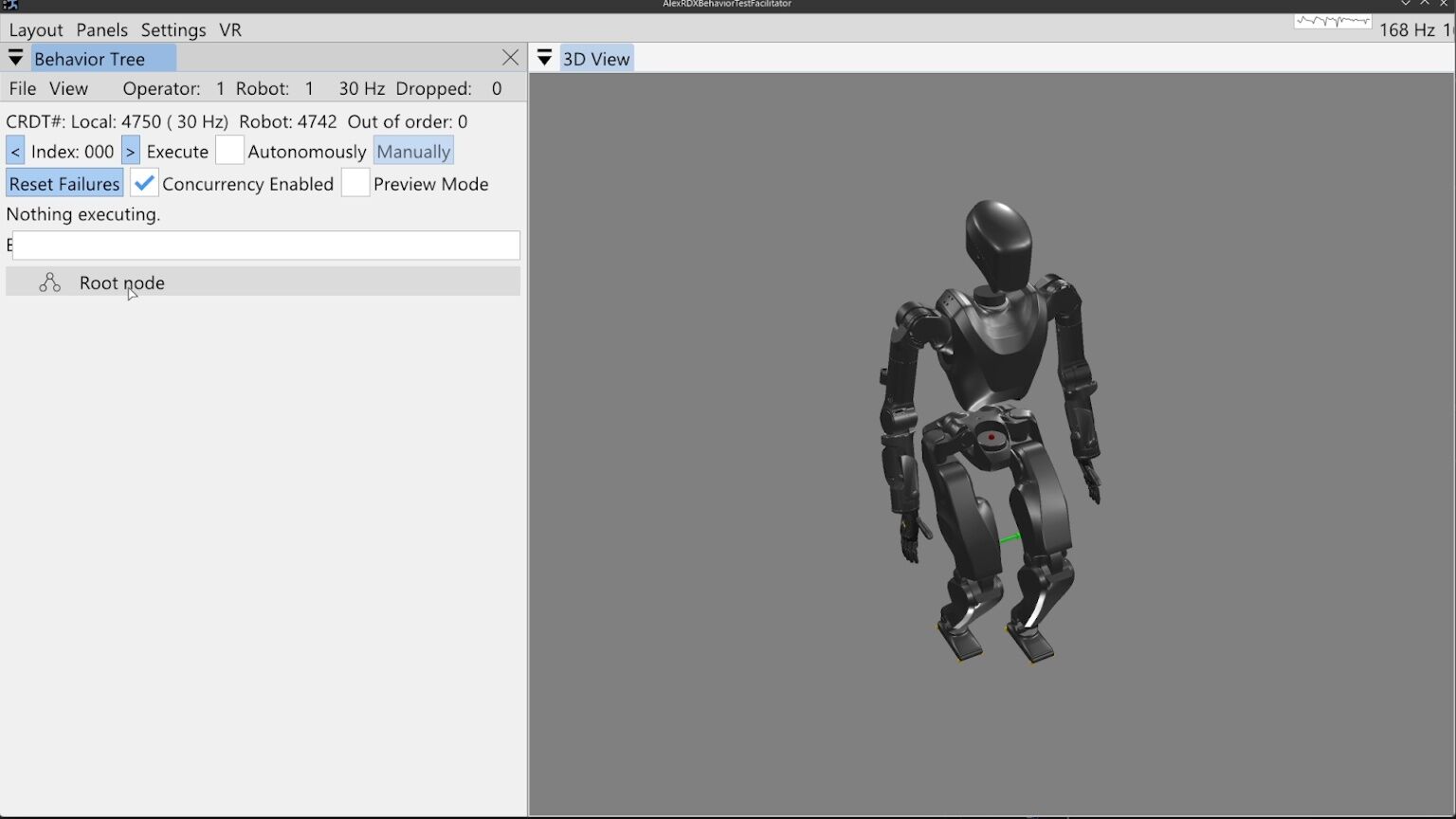}
    \caption{
        An empty behavior with just a root node.
        A video is available at \url{https://youtu.be/e3fk-EAJ5EQ}.
    }
    \label{fig:guide_sim_root_only}
\end{figure}

\subsection{Behavior Operation}
At this point, the view has transitioned into the primary working area for a loaded tree, as shown in \autoref{fig:guide_sim_root_only}.
At the top of the behavior tree panel, Conflict-Free Replicated Data Type (CRDT) statistics are shown.
This helps keep the operator informed about whether data is synchronizing correctly between the user interface and the robot-side autonomy process.
Next to the file and view menus, the node count for the operator-side and robot-side are shown.
These two numbers should be the same.
The frequency rendered on that line should be approximately 30 Hz, which is the desired rate of synchronization.
On the next line, a count of updates for the UI side (local) and robot side are shown.
These should be monotonically increasing but don't need to be the same.
There is an out of order message counter which should be 0 and incrementing a little is likely okay, but going up quickly is indicative of a problem.

The next lines regard execution operation.
The left and right pointing arrows decrement and increment the next node index to execute, which is printed here as ``Index: 000'' for our new tree.
On that line is a checkbox that toggles automatic execution and a button that manually executes the next concurrent action set.
This checkbox and button will cause the robot to immediately begin executing the behavior and possibly move!
In other words, these are the ``go'' buttons.
Automatic execution is started by checking the box and stopped by unchecking the box, so keep your mouse near it.
Automatic execution will also stop if any action fails that is not handled by a fallback catch or the end of the tree is reached.

On the next line, leaf node failures can be reset with the ``Reset failures'' button.
It is not normally necessary to do this, but can be nice if you want the blinking red to stop.
Also on this line, concurrency can be enabled and disabled with a checkbox.
This disables the concurrent functionality and every action will be run sequentially.
This is sometimes useful during authoring if it is desirable to run only one action but it is a part of a concurrent sequence.
The third widget on this line toggles on and off the preview mode, which retargets the executed behavior to a kinematics-only simulation robot.
When preview mode is enabled, a transparent preview robot will appear in the scene and execute the behavior in place of the real robot.
The preview mode is useful to verify compound motions in themselves and with respect to the scene.

The area that currently displays ``Nothing executing'' will display live information about the currently executing action or actions.
The final element in the view is the root node, with a representative icon with three circles connected by two lines.

\subsection{Building the First Behavior}
\begin{figure}[H]
    \centering
    \includegraphics[width=.95\columnwidth]{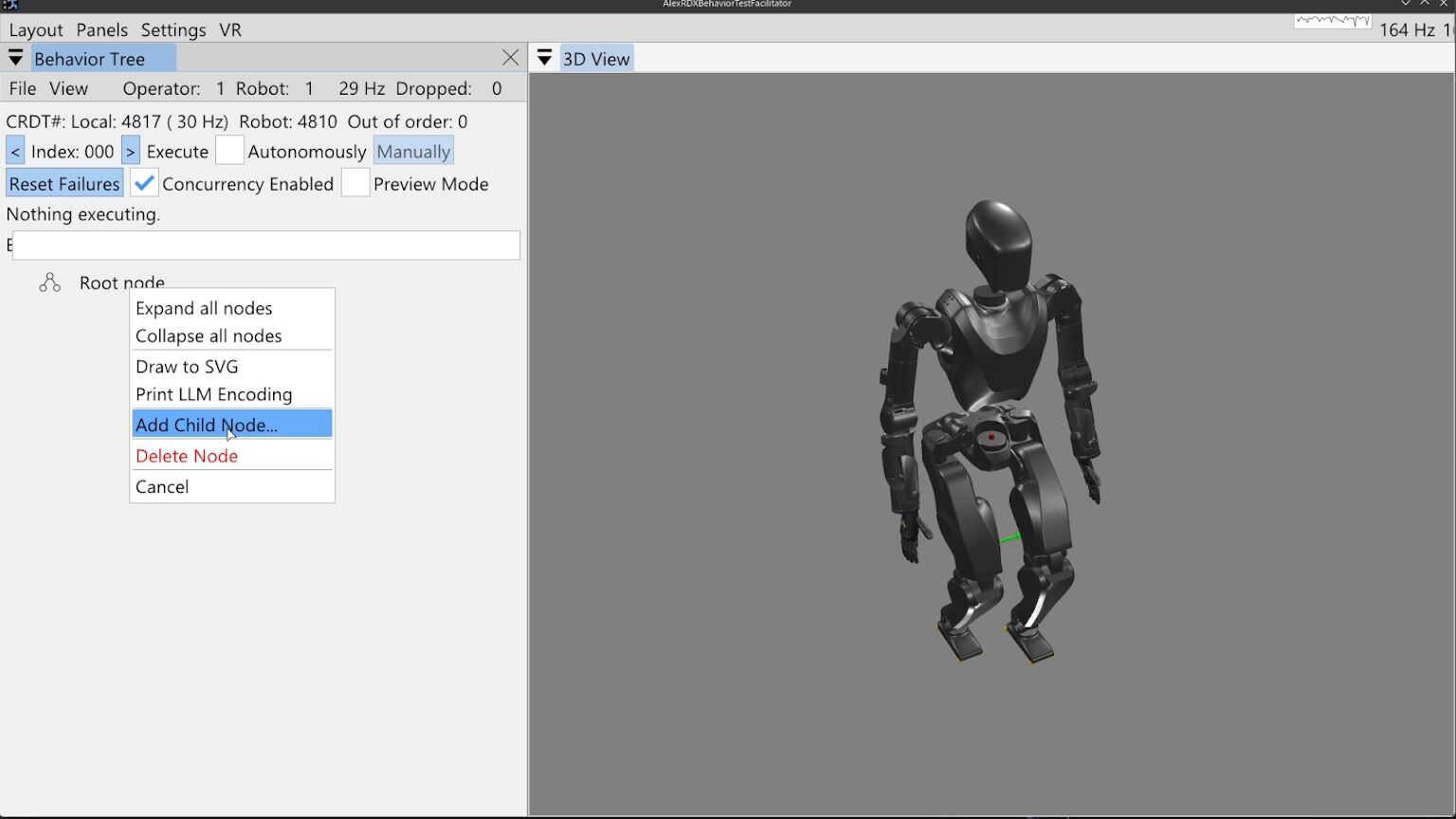}
    \caption{
        The root node context menu where we are adding a child node.
    }
    \label{fig:guide_sim_add_child}
\end{figure}

Right clicking the root node prompts a context menu which offers the option to add our first child node, as shown in \autoref{fig:guide_sim_add_child}.
In this first example, we'll build a small, simple behavior that moves the arms and walks.

\begin{figure}[H]
    \centering
    \includegraphics[width=.95\columnwidth]{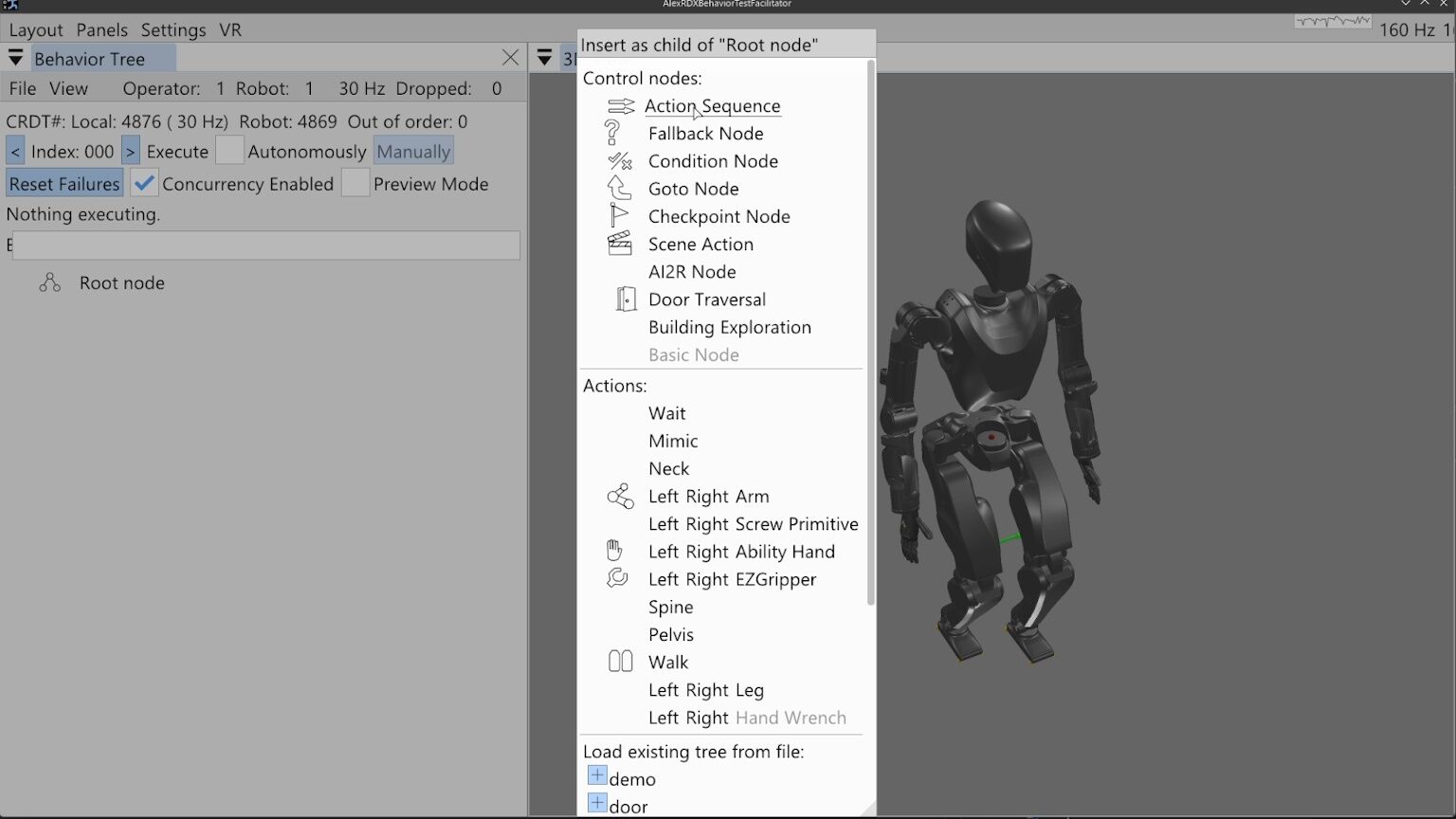}
    \caption{
        The node creation menu being used to create an action sequence node.
    }
    \label{fig:guide_sim_node_creation}
\end{figure}

\autoref{fig:guide_sim_node_creation} shows the node creation menu, which allows the operator to create a new node of any available type or loading an existing tree from file.
The available node types are sorted into control nodes and action nodes.
For this behavior, we will select the action sequence type, which serves as an organization element and a mechanism by which we save the behavior to file.

\begin{figure}[H]
    \centering
    \includegraphics[width=.95\columnwidth]{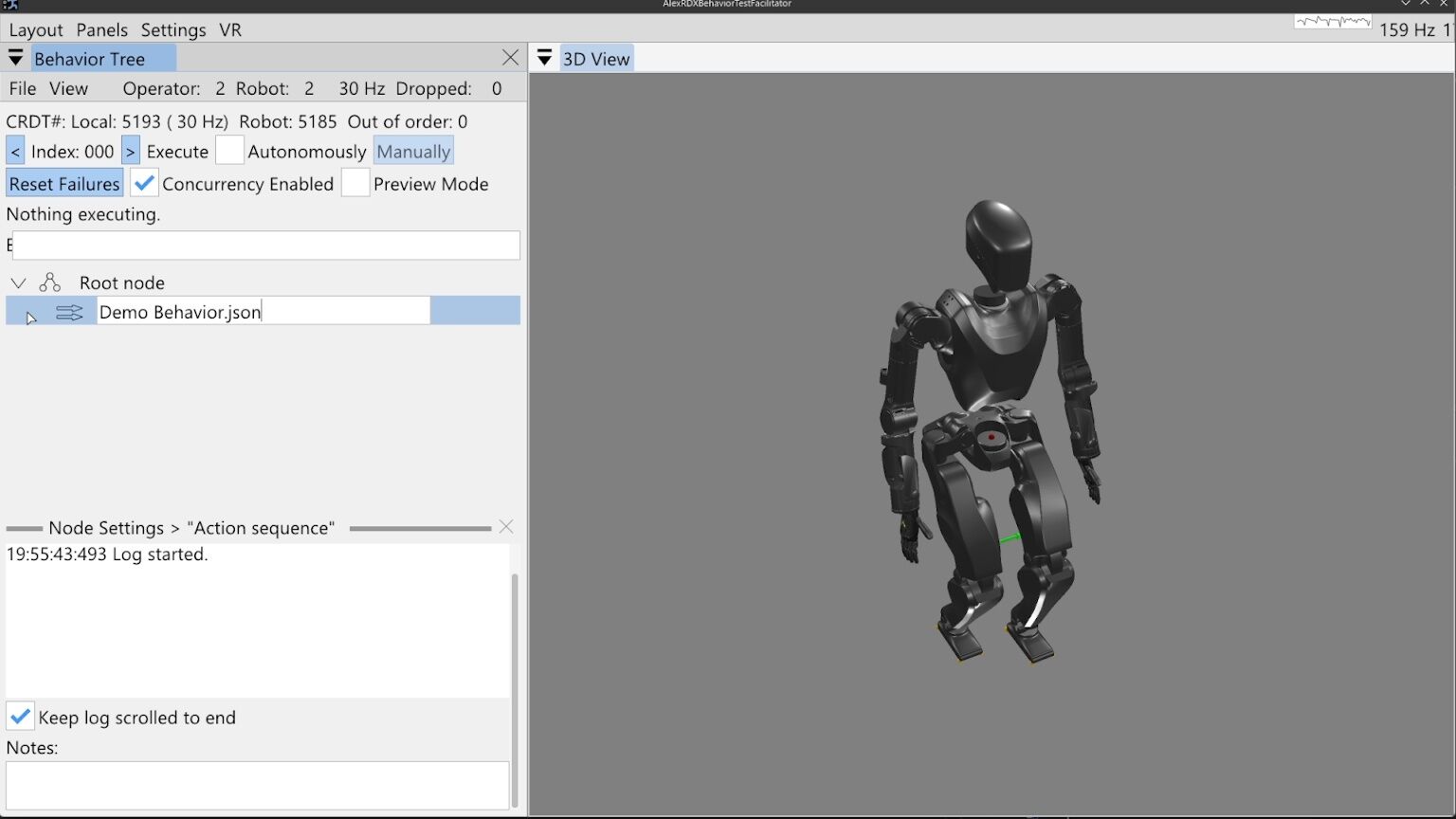}
    \caption{The sequence node has been added and renamed to ``Demo Behavior.json''.}
    \label{fig:guide_sim_sequence_rename}
\end{figure}

When the action sequence is created we double click the default name ``Action sequence'', type in ``Demo Behavior.json'', and hit the Enter key.
This is shown in \autoref{fig:guide_sim_sequence_rename}.
Adding ``.json'' to the end of a node name is what makes it saveable to file.
Saving is done by right clicking the node and selecting ``Save to File'', pressing ``Ctrl + S'' while the mouse is hovering in the behavior tree panel, or using the file menu at the top of the panel.
An asterisk symbol (*) is shown next to the node name when changes are present that have not been saved.
When the tree is saved, the asterisk symbol should disappear.
The behavior can safely be saved at any time and we do it often to avoid losing work.

\subsection{Arm Actions}
\begin{figure}[H]
    \centering
    \includegraphics[width=.95\columnwidth]{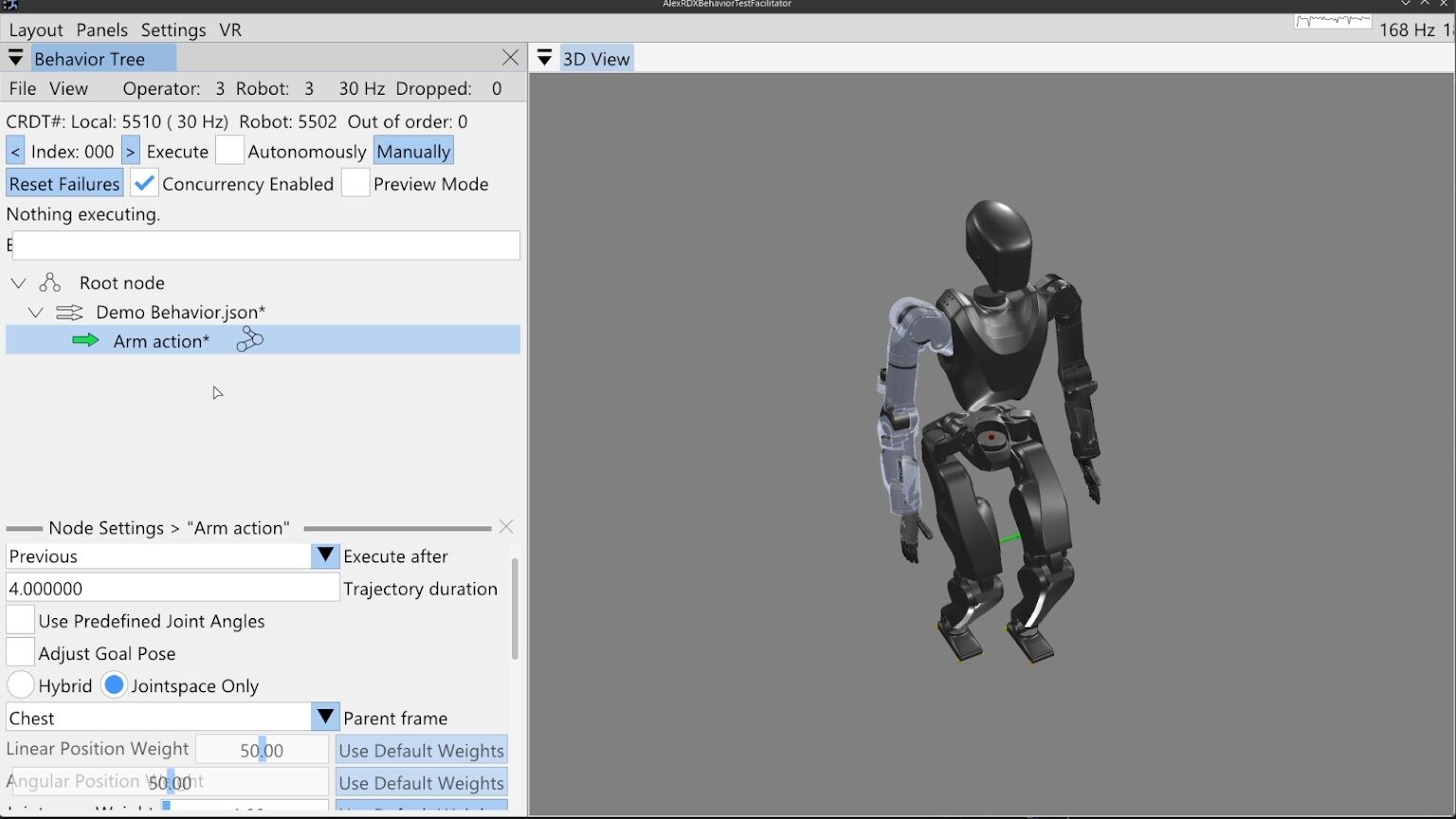}
    \caption{A right arm action has been created.}
    \label{fig:guide_sim_arm_action}
\end{figure}

We then add an arm action by right clicking the sequence node and clicking ``Add Child Node...'' as before, then click ``Right'' on the Arm row which instantiates a new arm action node with side set to right.
For sided actions, we currently don't allow changing the side after creation, however, that isn't an entirely purposeful design choice.
Now that the arm action has been created, the node can be seen in the tree, beneath our sequence node and indented to the right, signifying that it is a child of the sequence node.

When we single-click select the arm action node (anywhere except on the sideways arrow icon), the node's line is highlighted, as shown in \autoref{fig:guide_sim_arm_action}.
In the lower portion of the behavior tree panel, there is a node settings area.
This area can be closed using the ``X'' in the top right of its area, but will open whenever a node is selected, as we did above.
This setting area renders the settings of whichever node is selected, but only for one node at a time.

To make things go faster, we will reduce the arm action's trajectory duration to 1 second by double clicking the current value, typing 1, and hitting the Enter key.
There are two main ways to define this arm action: by adjusting the hand goal pose with a 3D pose gizmo and inverse kinematics solver, or by using sliders to define the arm's joint angles directly.
This is decided using the ``Use Predefined Joint Angles'' checkbox.

\begin{figure}[H]
    \centering
    \includegraphics[width=.95\columnwidth]{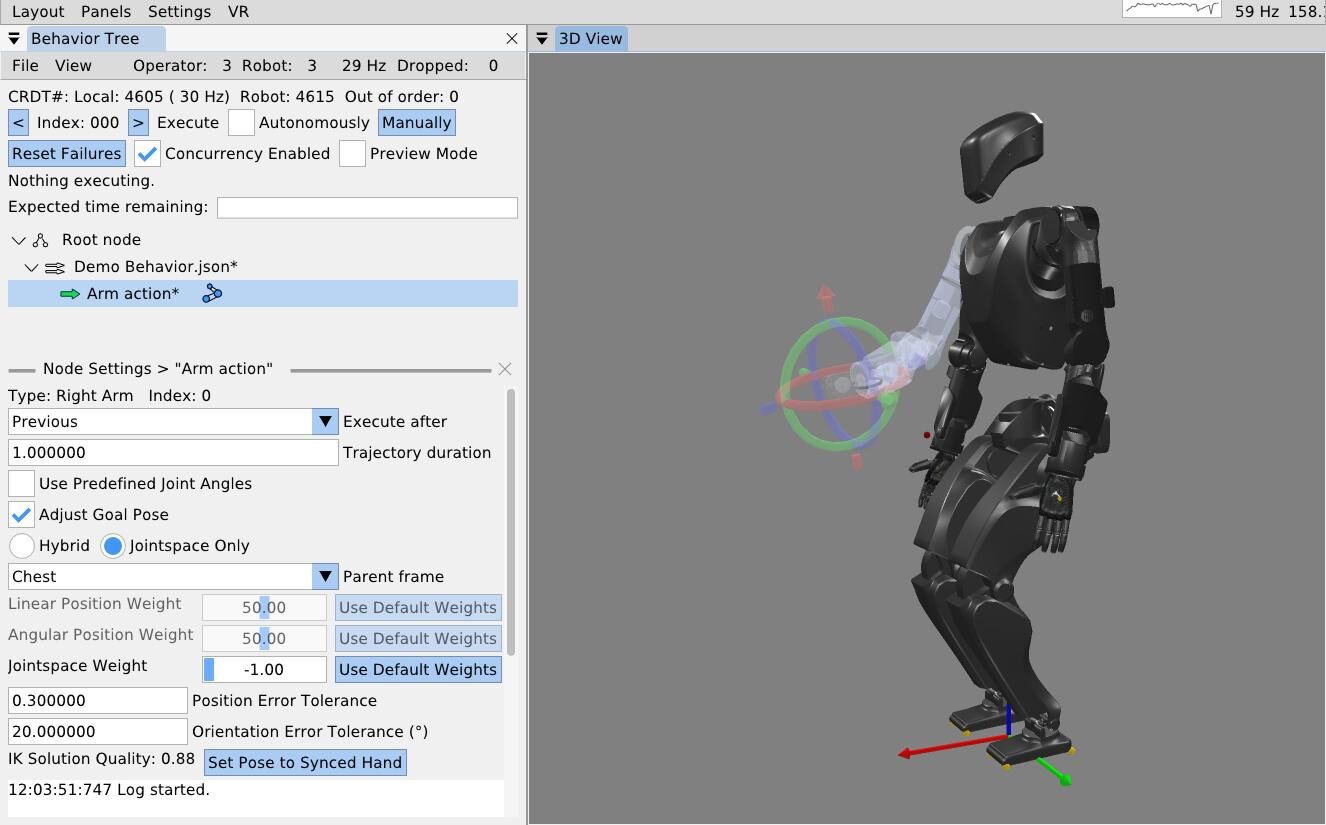}
    \caption{A right arm action with the 3D pose gizmo activated.}
    \label{fig:guide_sim_arm_gizmo}
\end{figure}

\subsection{3D Pose Gizmo}
As shown in \autoref{fig:guide_sim_arm_gizmo}, to adjust the arm action via hand pose, check the ``Adjust Goal Pose'' box.
A 3D pose gizmo appears in the 3D view.
The gizmo's axes are colored as Red, Green, and Blue, to match X, Y, and Z, and Roll, Pitch, and Yaw.
A way to remember it is ``RGB -> XYZ''.
The tori can be dragged with the mouse to adjust the orientation.
The arrow heads and tails can be dragged with the mouse to adjust the translation.
The gizmo can also be adjusted via the keyboard.
A key for gizmo keyboard controls is presented in \autoref{tab:pose_3d_gizmo_keys}.
Right clicking the gizmo will display a context menu that allows for numerical pose adjustment, fine and coarse increments, resetting to zero, and changing the modification frame.
The gizmo is, by default, modified in camera Z up frame so it translates laterally on the world X-Y plane and vertically on the world Z axis, similar to the focus based camera.

\begin{table}[H]
    \centering
        \renewcommand{\arraystretch}{1.2}
        \begin{tabular}{ll}
            \hline
            \textbf{Pose 3D gizmo} & \textbf{Key} \\
            \hline
            Fine adjustment modifier & Shift \\
            Manipulate axes & Left mouse drag \\
            Open context menu & Right mouse click \\
            Pitch adjustment + & Alt + Up arrow \\
            Pitch adjustment - & Alt + Down arrow \\
            Roll adjustment + & Alt + Right arrow \\
            Roll adjustment - & Alt + Left arrow \\
            Translation adjustment X+ & Up arrow \\
            Translation adjustment X- & Down arrow \\
            Translation adjustment Y+ & Left arrow \\
            Translation adjustment Y- & Right arrow \\
            Translation adjustment Z+ & Ctrl + Up arrow \\
            Translation adjustment Z- & Ctrl + Down arrow \\
            Yaw adjustment + & Ctrl + Left arrow \\
            Yaw adjustment + & Ctrl + Mouse scroll down \\
            Yaw adjustment - & Ctrl + Mouse scroll up \\
            Yaw adjustment - & Ctrl + Right arrow \\
            \hline
        \end{tabular}
        \caption{Pose 3D gizmo keyboard shortcuts}
        \label{tab:pose_3d_gizmo_keys}
\end{table}

\subsection{Frame-Relative Action}
When an arm action is defined by a pose, it is specified in a selectable parent frame.
By default, this frame is chest frame.
Changing this frame to an object class such as ``Door Lever'' allows for a pose that will be relative to that object, wherever that object is in the scene.
The ``Hybrid'' and ``Jointspace Only'' options and the weights can be ignored.
They are specific to how our model based whole body controller tracks the arm command.
What is important is to note that any whole body controller specific options you may have can be included in this arm action's definition.
It is meant to be extendable and flexible rather than a rigid specification.
The inverse kinematics solution quality is displayed where values from 0 to 1 are good, and >1 is bad.
Bad solutions mean the pose is unreachable and will not be achieved consistently with our implementation.
The transparent arm graphic will turn red to signify this.
A ``Set Pose to Synced Hand'' option is available to reset the pose to where the hand currently is on the real robot.

As seen in the bottom left of \autoref{fig:guide_sim_arm_gizmo}, position and orientation error tolerance settings are available.
If the hand's pose is not within the error tolerance of the goal pose by the end of the trajectory duration, the action will fail.
This setting can be adjusted based on the required precision of the motion.
This functionality is very context and controller dependent.
For example, a controller may continuously try to achieve the goal pose or it may stop trying once the trajectory duration is over.
As another example, when applying a force on something is desired, a position setpoint may be placed beyond an immovable object, resulting in a desired pose error.
In these cases the error tolerances may not make sense or may need to be large.
It may also be that we should add a timeout for achieving the motion in addition to the nominal trajectory duration.

\subsection{Jointspace Mode}
\begin{figure}[H]
    \centering
    \includegraphics[width=.95\columnwidth]{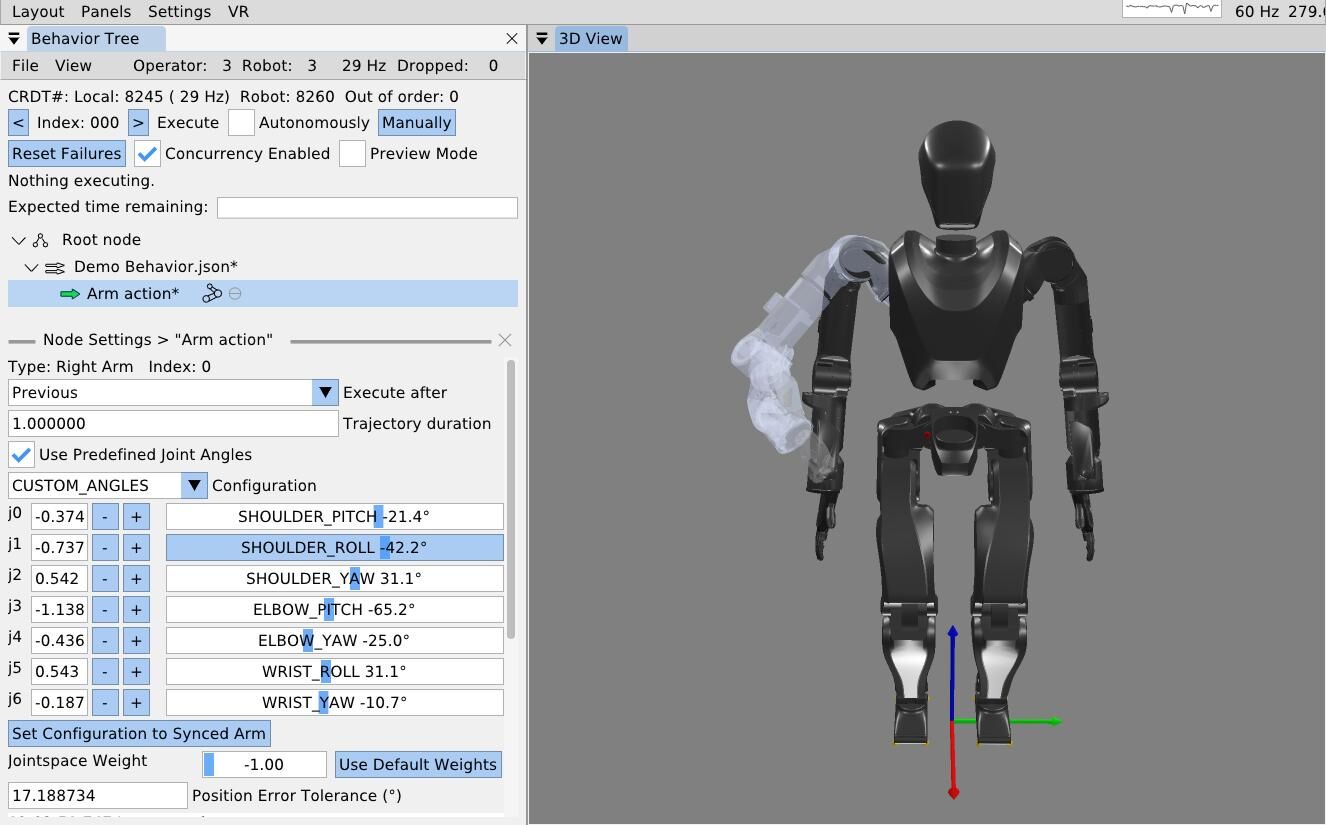}
    \caption{A right arm action being adjusted in jointspace.}
    \label{fig:guide_sim_arm_jointspace}
\end{figure}

The other way to define arm actions is by specifying the joint angles directly.
To do this, check the ``Use Predefined Joint Angles'' box.
This will dynamically change the available settings in the panel, now showing sliders for all the joints.
As shown in \autoref{fig:guide_sim_arm_jointspace}, these sliders can be used to set the joint angles.
As the sliders are dragged, the full arm 3D preview is updated to provide an interactive experience.
The sliders are bounded by the joint limits from the robot model.
A joint angle can also be input as a number manually in the input box next to the slider.
This mode has a ``Set Configuration to Synced Arm'' button to reset the values to match the real robot's current configuration.
It also has a position error tolerance setting which is the maximum allowable sum of joint angle errors throughout the arm.
In our example behavior, we roll the arm out from the body and pull the forearm up.

\subsection{Arm Action Execution}
To execute this arm action, we ensure the hollow arrow icon next to the action is green, meaning it is selected as the next action to execute.
To execute the action, we click the (Execute) ``Manually'' button at the top of the panel.
The simulated robot performs a 1 second trajectory to the goal configuration.

We would like to extend our arm action to support N-length trajectories, as we do with our screw primitive covered later.
One use case would be to record teleoperated motions, store them to a CSV file or something, and execute them with the arm action.
Another option would be to allow the behavior author to add and remove waypoints tuned using gizmos.
Using multi-waypoint trajectories would allow the arm to keep moving through poses instead of stopping at each one, as is currently enforced.

\subsection{Mirroring an Arm Action}
\begin{figure}[H]
    \centering
    \includegraphics[width=.95\columnwidth]{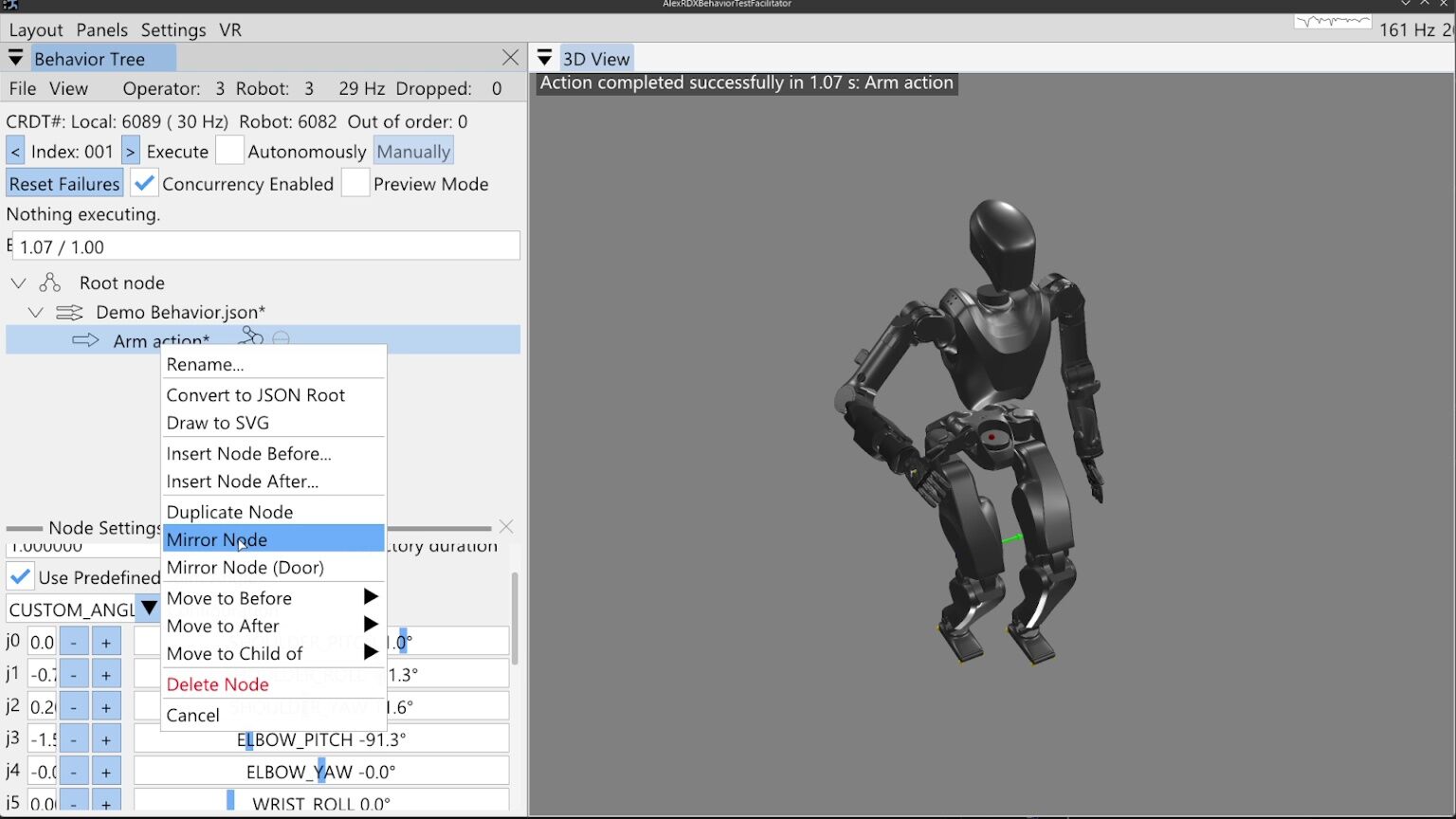}
    \caption{The node mirroring context menu option.}
    \label{fig:guide_sim_arm_mirroring}
\end{figure}

Next, we will mirror this action for the left arm using the ``Mirror Node'' context menu entry on the arm node, as shown in \autoref{fig:guide_sim_arm_mirroring}.
A second arm action appears, already in joint angle mode and mirroring the other arm action.
We manually execute this action.

\subsection{Walk Action}
\begin{figure}[H]
    \centering
    \includegraphics[width=.95\columnwidth]{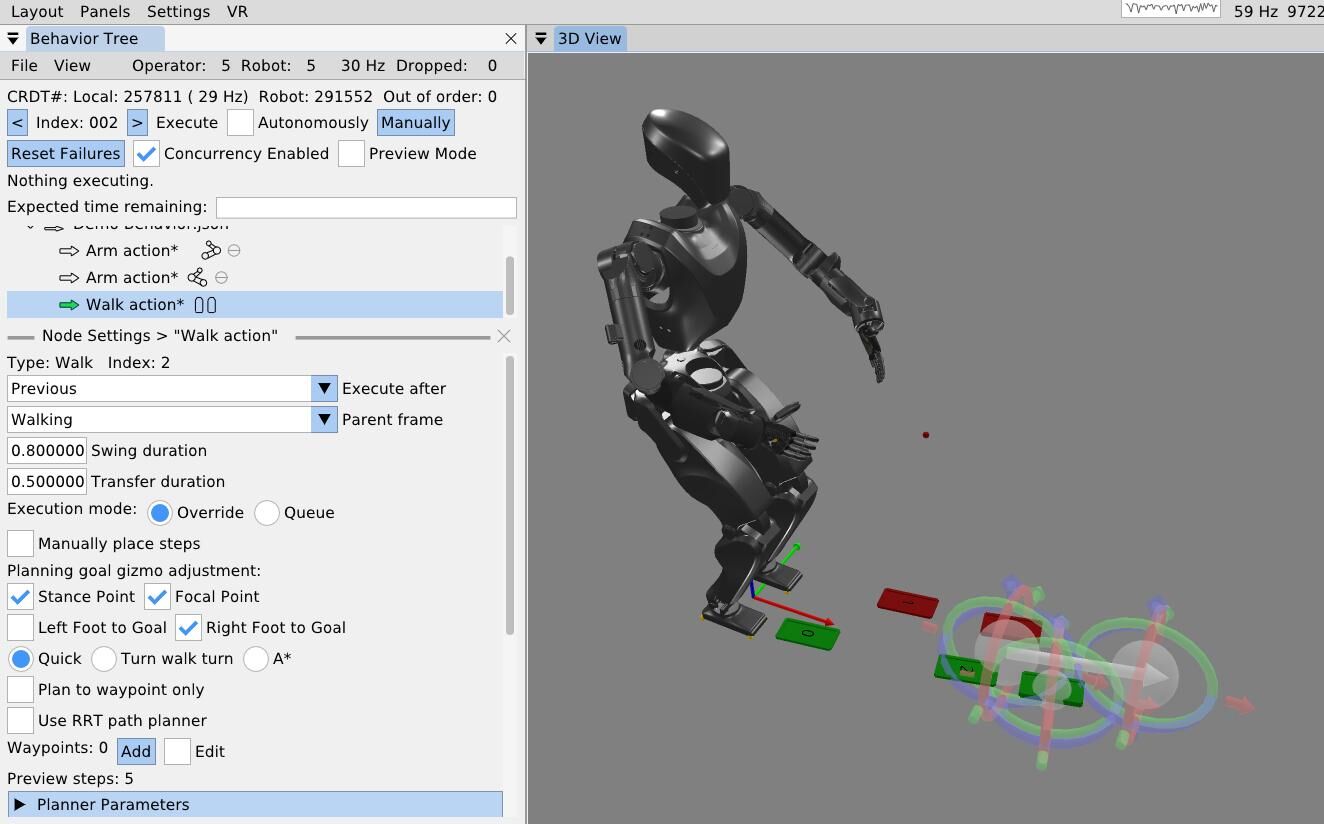}
    \caption{The walk action settings and goal tuning.}
    \label{fig:guide_sim_walk_settings}
\end{figure}

As the final action in this first example behavior, we will walk forward a little bit.
We right-click the second arm action, click ``Insert Node After...'', and select the walk action.
The walk action now appears in the tree.
We click it to access its settings, which are shown in \autoref{fig:guide_sim_walk_settings}.
For this example, we'll keep the frame set to ``Walking'' frame, which is a frame on the ground underneath and facing the direction of the pelvis.

The walk action supports setting controller specific settings such as the walking speed via the foot swing and double support transition durations.
An ``execution mode'' setting specifies whether the robot should finish any steps it may have queued versus overriding those and taking the first step of this walk action after the current step is completed.
This setting is also controller specific.

The walk action is currently implemented to dispatch a list of footsteps as 3D sole poses to a controller for execution.
The steps can be specified manually by adding and tuning them with gizmos or planned.
Footstep plans can be converted to a manually defined plan for action definition, but otherwise, planning will happen on action execution.

\subsection{Walk Goal Specification}
To define the planning goal, we use a mid-stance point and a focal point.
The robot walks to the stance location and ends facing the point.
We project the goal Z to the robot's current walking frame Z, which lies on the ground between the feet.
This format of goal specification works well to keep the robot from stepping in the air when on flat ground.
It also makes the robot's goal facing orientation easier to tune by placing the focal point farther from the stance point.
This separation acts as a ``lever arm'' for goal orientation precision.

The goal footsteps are also each tunable in this mid-stance goal frame.
By default they are even at the controller's default stance width for a squared up goal stance.
However, for many task approaches, such as pull doors, we require a staggered stance.
The ``Left Foot to Goal'' and ``Right Foot to Goal'' checkboxes toggle gizmos to tune the goal footsteps.
In \autoref{fig:guide_sim_walk_settings}, the stance point, focus point, and the right footstep are all being tuned.

\subsection{Footstep Planning}
We currently have three footstep planners available to the walk action: the quick footstep planner, the turn-walk-turn planner, and the A* planner.
The quick footstep planner is the newest option.
It is a procedural geometry-heuristic based planner that plans quickly and reliably.
It is designed to have as few failure modes as possible and reduce unnecessary steps.
Though the heuristics are general, we focused on ensuring the plan for approaching pull doors and getting into the staggered stance was high quality.
It supports walking to waypoints without specifying goal footsteps.
The option to not specify goal footsteps can speed up behaviors by removing an unnecessary square-up step.
It currently only supports flat ground, but we think it could be extended to plan over terrain maps.

When the quick footstep planner is selected, an option for RRT-Connect~\cite{kuffner2000rrt} path planning is available.
The current implementation will simply maintain a tunable distance from objects in the behavior scene while taking the shortest path.
This is an experimental mode that we would like to make more general using an occupancy map or something.

The turn-walk-turn is another procedural heuristic planner for flat ground, but often has many unnecessary steps in the plan, reducing overall behavior speed.
We don't use this one much.
It is mainly intended to be a backup option in case the other two don't work for some reason in certain situations.

The A* planner, as presented in~\cite{griffin2019footstepplanner}, is a search-based planner for flat ground and rough terrain.
It uses a large set of parameters that define the ideal and boundary step criteria and then searches over an SE2 (X, Y, and Yaw) lattice to find the optimal set of footsteps to the goal feet.
The planner can snap footsteps to planar regions and wiggle them as part of the search and in the pursuit of achieving a stable foothold.
It can also plan over a height map.
In this way, the A* planner gives the behavior system a rough terrain capability.
For flat ground, we don't use it as much to avoid the extra planning time and increased number of failure modes.

\subsection{Manual Footsteps}
The walk action also allows the operator to define a manual footstep plan by checking the ``Manually place steps'' box.
Then, footsteps can be added, removed, and tuned with gizmos.
A ``Select All Footsteps'' button is available to select all footstep gizmos in order to move the whole plan using the keyboard arrow keys.
A ``Reset footstep height'' option is available to reset all the footstep Z heights to the current robot height, which is useful for flat ground plans.

The manual footstep plan option is especially useful when the robot walks through the door frame, mainly because a planner has not been written for that case.
Door traversal footstep plans have to straddle the door frame in order to not hit the shoulder on the door frame.
It also helps to make the footstep plan narrower, to reduce side-to-side sway, which can reduce the severity of collisions with the door panel.

\subsection{Executing a Walk Action}
\begin{figure}[H]
    \centering
    \includegraphics[width=.95\columnwidth]{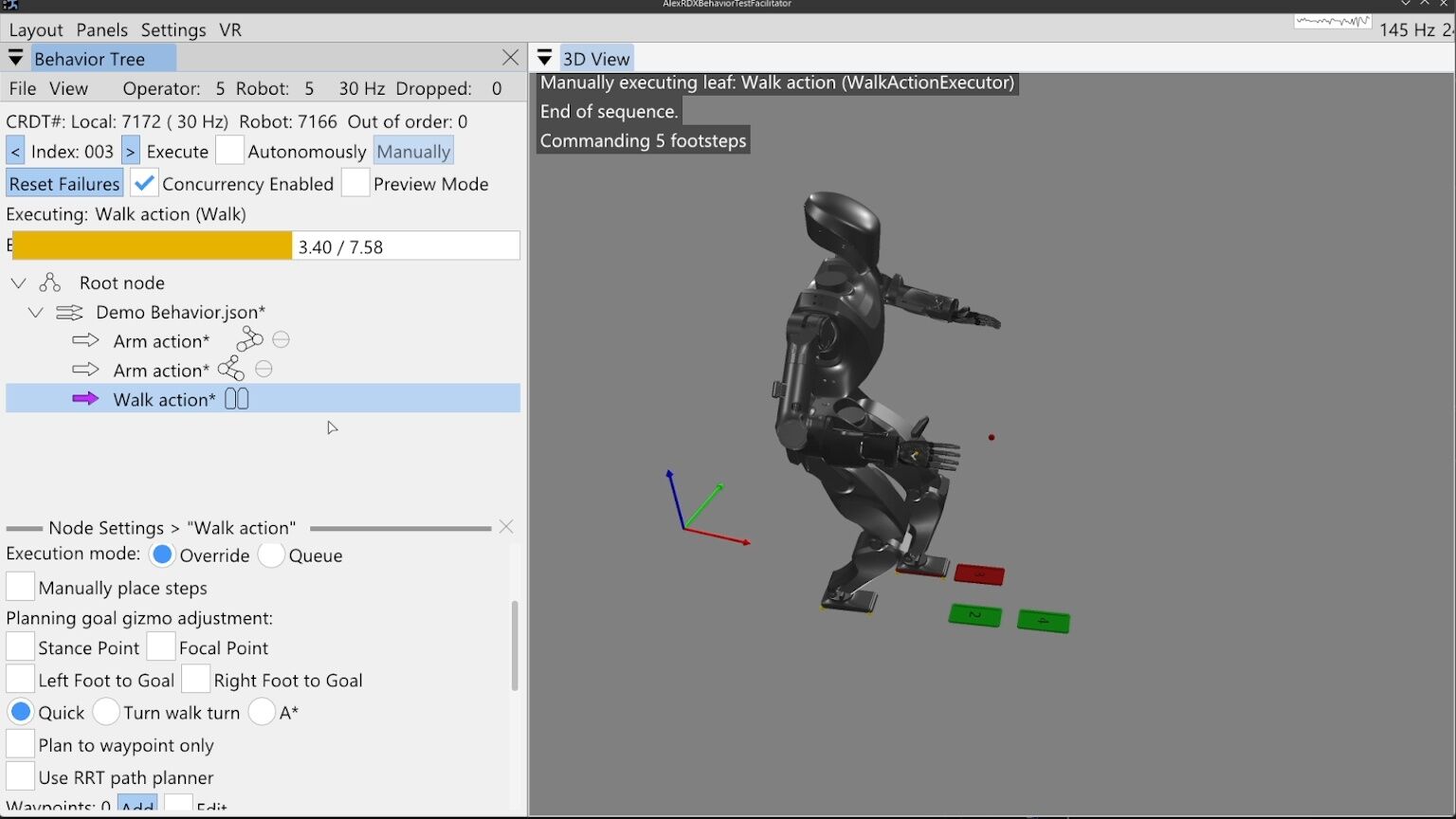}
    \caption{The walk action executing.}
    \label{fig:guide_sim_walk_executing}
\end{figure}

In \autoref{fig:guide_sim_walk_executing}, we execute the walk action to complete our first example.
A progress bar can be seen that is tracking the action.
The 7.58 second total is calculated by adding up all the transfer and swing times of the planned footsteps.
Virtual, numbered footsteps are displayed in the 3D view to show the robot controller's current queue.

\subsection{Simple Behavior JSON File}
\label{sec:simple_behavior_json_file}

A simplified version of the saved JSON for this simple example is presented in \autoref{fig:demo_behavior_json_part_1} and \autoref{fig:demo_behavior_json_part_2}.

\begin{figure}[t]
    \centering
    \begin{minipage}{0.98\columnwidth}
        \begin{Verbatim}[
            fontsize=\scriptsize,
            breaklines=true,
            breakanywhere=true,
            frame=single,
            rulecolor=\color{black!30},
            framesep=2mm
        ]
{
  "type" : "ActionSequenceDefinition",
  "name" : "Demo Behavior.json",
  "notes" : "",
  "children" : [ {
    "type" : "ArmActionDefinition",
    "name" : "Move Right Arm",
    "notes" : "",
    "children" : [ ],
    "executeAfterAction" : "Previous",
    "side" : "right",
    "trajectoryDuration" : 1.0,
    "usePredefinedJointAngles" : true,
    "preset" : "CUSTOM_ANGLES",
    "j0Degrees" : 40.1,
    "j1Degrees" : -22.92,
    "j2Degrees" : 22.92,
    "j3Degrees" : -108.86,
    "j4Degrees" : 0.0,
    "j5Degrees" : 0.0,
    "j6Degrees" : 0.0,
    "positionErrorTolerance" : 0.3,
    "jointspaceWeight" : -1.0
  }, {
    "type" : "ArmActionDefinition",
    "name" : "Move Left Arm",
    "notes" : "",
    "children" : [ ],
    "executeAfterAction" : "Previous",
    "side" : "left",
    "trajectoryDuration" : 1.0,
    "usePredefinedJointAngles" : true,
    "preset" : "CUSTOM_ANGLES",
    "j0Degrees" : 40.1,
    "j1Degrees" : 22.92,
    "j2Degrees" : -22.92,
    "j3Degrees" : -108.86,
    "j4Degrees" : 0.0,
    "j5Degrees" : 0.0,
    "j6Degrees" : 0.0,
    "positionErrorTolerance" : 0.3,
    "jointspaceWeight" : -1.0
  },
        \end{Verbatim}
    \end{minipage}
    \caption{Example behavior definition serialized as JSON (Part 1).}
    \label{fig:demo_behavior_json_part_1}
\end{figure}

\begin{figure}[t]
    \centering
    \begin{minipage}{0.98\columnwidth}
        \begin{Verbatim}[
            fontsize=\scriptsize,
            breaklines=true,
            breakanywhere=true,
            frame=single,
            rulecolor=\color{black!30},
            framesep=2mm
        ]
     {
    "type" : "WalkActionDefinition",
    "name" : "Walk action",
    "notes" : "",
    "children" : [ ],
    "executeAfterAction" : "Previous",
    "swingDuration" : 0.8,
    "transferDuration" : 0.5,
    "executionMode" : "OVERRIDE",
    "parentFrame" : "Walking",
    "goalStancePoint" : {
      "x" : 1.292,
      "y" : -0.013,
      "z" : 0.0
    },
    "goalFocalPoint" : {
      "x" : 2.2745,
      "y" : -0.002,
      "z" : 0.0
    },
    "leftGoalFootToGoal" : {
      "x" : 0.0,
      "y" : 0.11,
      "yawInDegrees" : 0.0
    },
    "rightGoalFootToGoal" : {
      "x" : 0.0,
      "y" : -0.11,
      "yawInDegrees" : 0.0
    },
    "planner" : "QUICK",
    "plannerParameters" : { }
  } ]
}
        \end{Verbatim}
    \end{minipage}
    \caption{Example behavior definition serialized as JSON (Part 2).}
    \label{fig:demo_behavior_json_part_2}
\end{figure}

\section{Concurrent Actions Example}

\subsection{Behavior Timeline}
\begin{figure}[H]
    \centering
    \includegraphics[width=.95\columnwidth]{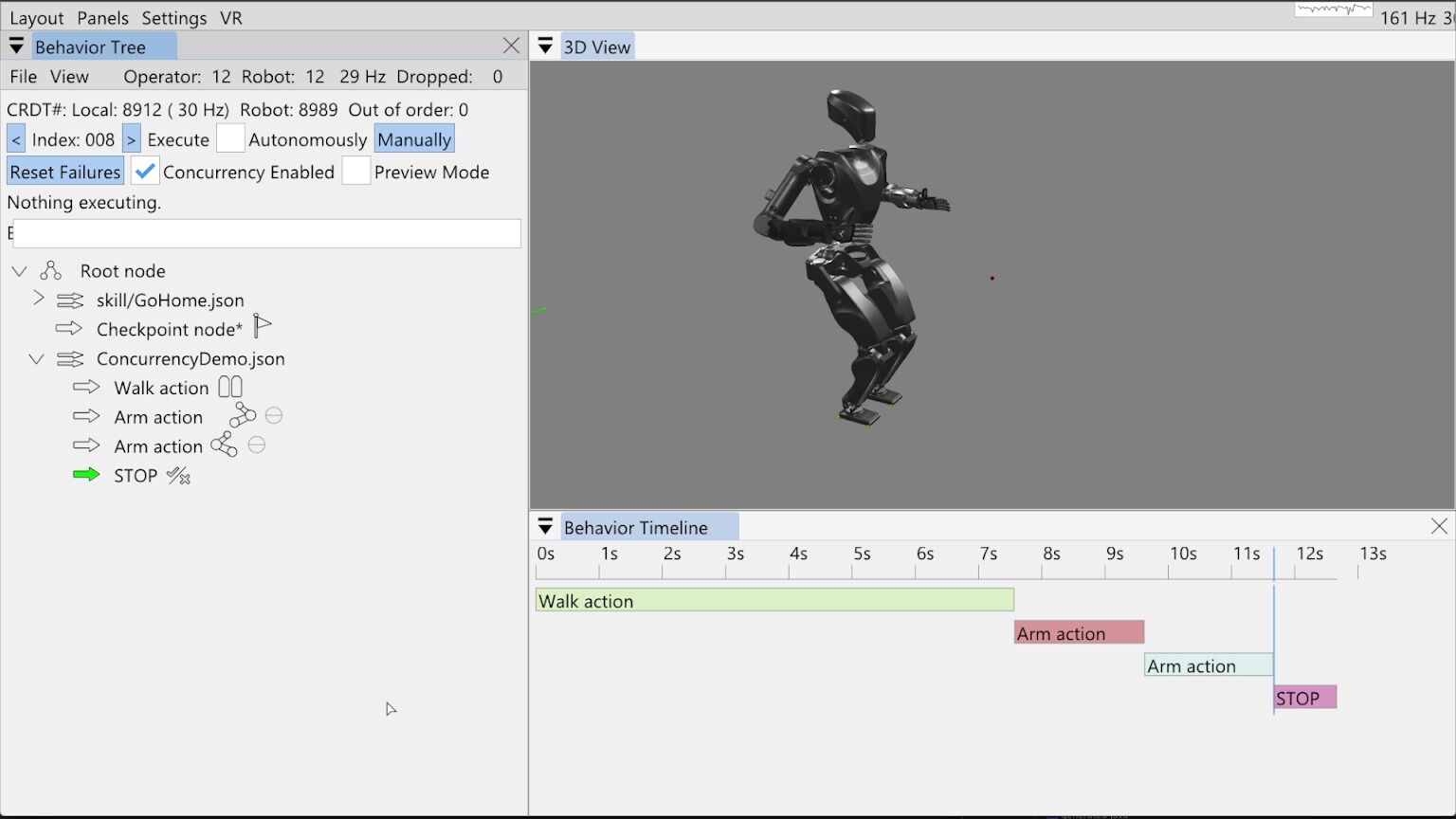}
    \caption{The non-concurrent starting behavior for the concurrency example.
    A video of this demonstration is available at \url{https://youtu.be/D53dKL-pgeo}.}
    \label{fig:guide_sim_nonconcurrent}
\end{figure}

In this example, we'll show how we can schedule arm motions while walking using our concurrent action layering system.
\autoref{fig:guide_sim_nonconcurrent} shows a behavior we have already built and run as the non-concurrent starting place.
In this tree, we have added the pre-existing ``Go Home'' skill from a file, which allows us to reset the robot to the home configuration.
This home configuration has the arms down by the sides in a natural position.
Then there is a checkpoint node and a savable concurrency demo behavior, with a walk action followed by two arm actions and an always failing ``STOP'' condition.

Our behavior timeline feature is presented in the bottom right of \autoref{fig:guide_sim_nonconcurrent}.
It is a rendering of an action sequence over time meant to help with understanding the timing and concurrency of behavior actions.
Here, it shows the tree as executed with each node running sequentially.

\subsection{``Execute After'' Layering Mechanism}
Concurrent action layering is implemented simply through an ``Execute after'' field in each action.
It is selectable via a drop-down menu in the node settings, which offers every prior node as an option.
By default this field is set to ``Previous'', which is a dynamic reference to whatever node is immediately before.
A dynamic ``Beginning'' reference is also available, which points to before the root node, though it may be removed because it is functionally the same as pointing to the ever-present root node.

When the behavior is running in autonomous mode, it triggers action execution in order.
When deciding whether to trigger execution of the next one, it first checks the referenced node to execute after.
If the node to execute after is currently executing, the next node is not triggered.
Importantly, action execution does not wait for any nodes that may be in between the ``execute after'' node and itself.
This is important for scheduling concurrent actions freely.

The full algorithm is more complicated because the behavior can also be run step-by-step by clicking the ``Manually'' button.
There is also a checkbox for disabling concurrency in the user interface, which effectively treats all ``Execute after'' fields as ``Previous''.
Further complexity is added to handle the fallback node's try and catch mechanism.
We present pseudocode in \autoref{alg:simplified_execution} without the fallback part.

\begin{algorithm}[H]
\caption{Simplified Action Execution Trigger Logic}
\label{alg:simplified_execution}
\SetAlgoLined
\DontPrintSemicolon
\SetKwFunction{FTickExecution}{TickExecution}
\SetKwProg{Fn}{Function}{:}{}

\Fn{\FTickExecution{$orderedLeaves$, $state$}}{
    \tcp{Halt if not autonomous and no manual step was requested}
    \If{$\neg state.\text{isAutonomous}() \land \neg state.\text{isManualStepRequested}()$}{
        \Return\;
    }

    $nextIndex \gets state.\text{getExecutionNextIndex}()$\;

    \For{$i \gets nextIndex$ \KwTo $orderedLeaves.\text{length} - 1$}{
        $node \gets orderedLeaves[i]$\;

        \If{$node.\text{isExecuting}()$}{
            \textbf{continue}\;
        }

        \tcp{Determine the required dependency}
        $afterIndex \gets node.\text{getExecuteAfterIndex}()$\;
        \If{$\neg state.\text{isConcurrencyEnabled}()$}{
            $afterIndex \gets i - 1$ \tcp*{Force sequential execution}
        }

        \tcp{Break if the action to execute after is still executing}
        \If{$afterIndex \geq 0 \land orderedLeaves[afterIndex].\text{isExecuting}()$}{
            \textbf{break}\;
        }

        \tcp{Trigger execution and advance the index tracker}
        $node.\text{triggerExecution}()$\;
        $state.\text{setExecutionNextIndex}(i + 1)$\;

        \tcp{If running manually step-by-step, consume the step and halt}
        \If{$\neg state.\text{isAutonomous}()$}{
            $state.\text{consumeManualStep}()$\;
            \textbf{break}\;
        }
    }
}
\end{algorithm}

\vspace{\baselineskip}

\subsection{Wait Nodes and Dependencies}
\label{sec:wait_nodes_dependencies}
\begin{figure}[H]
    \centering
    \includegraphics[width=.95\columnwidth]{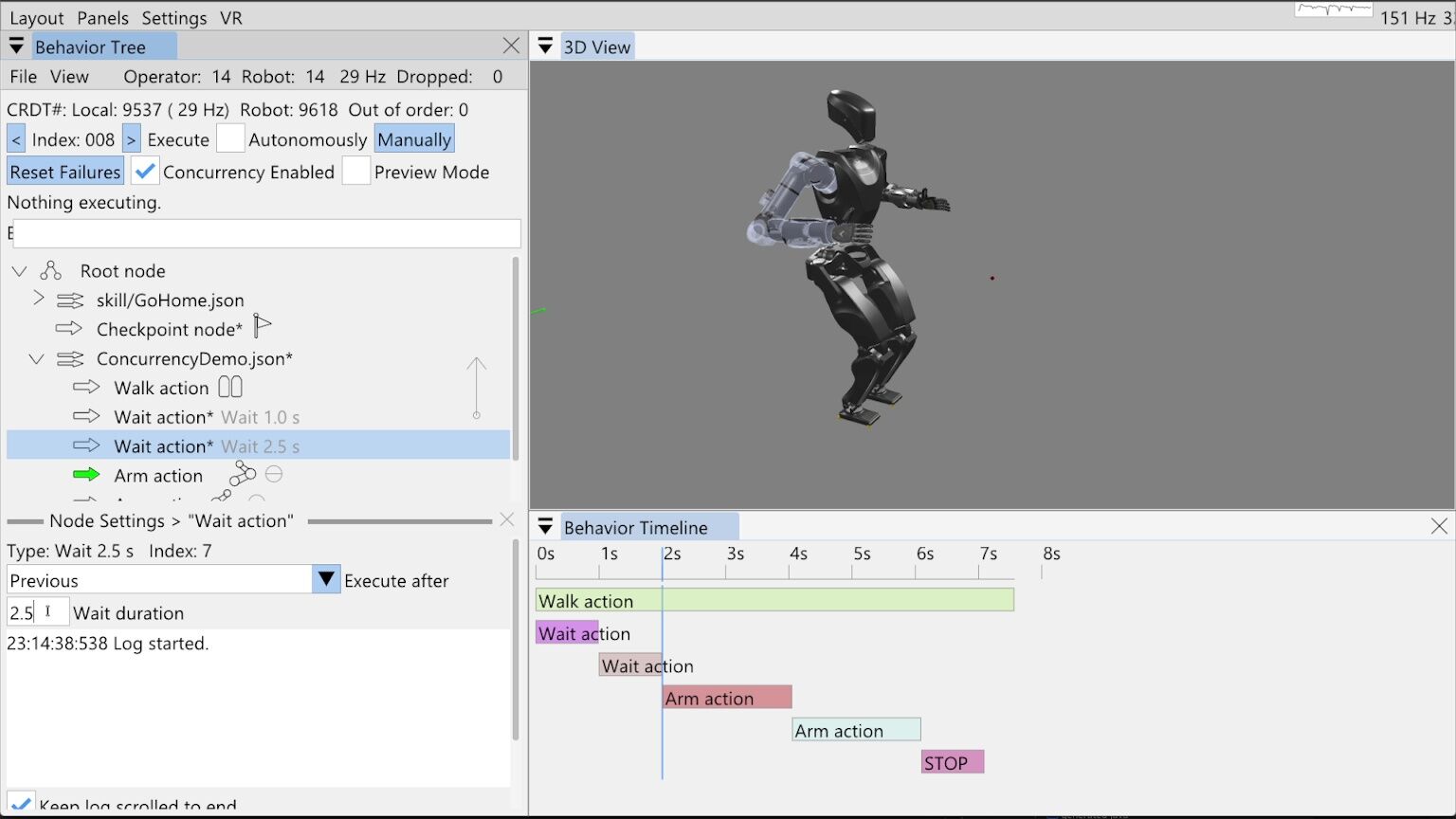}
    \caption{Adding wait nodes after the walk action.
    A video of this demonstration is available at \url{https://youtu.be/D53dKL-pgeo}.}
    \label{fig:guide_sim_concurrent_waits}
\end{figure}

Our example behavior in \autoref{fig:guide_sim_nonconcurrent} starts out as a non-concurrent sequence of a short walk and two arm motions.
Now we'll change it so the arm motions are scheduled during the walk.
In \autoref{fig:guide_sim_concurrent_waits}, we add two wait actions, one for 1 second and one for 2.5 seconds.
These waits are used to schedule the arm motions.
We'll move the right arm 1 second into the walk and the left arm 2.5 seconds into the walk.

\begin{figure}[H]
    \centering
    \includegraphics[width=.95\columnwidth]{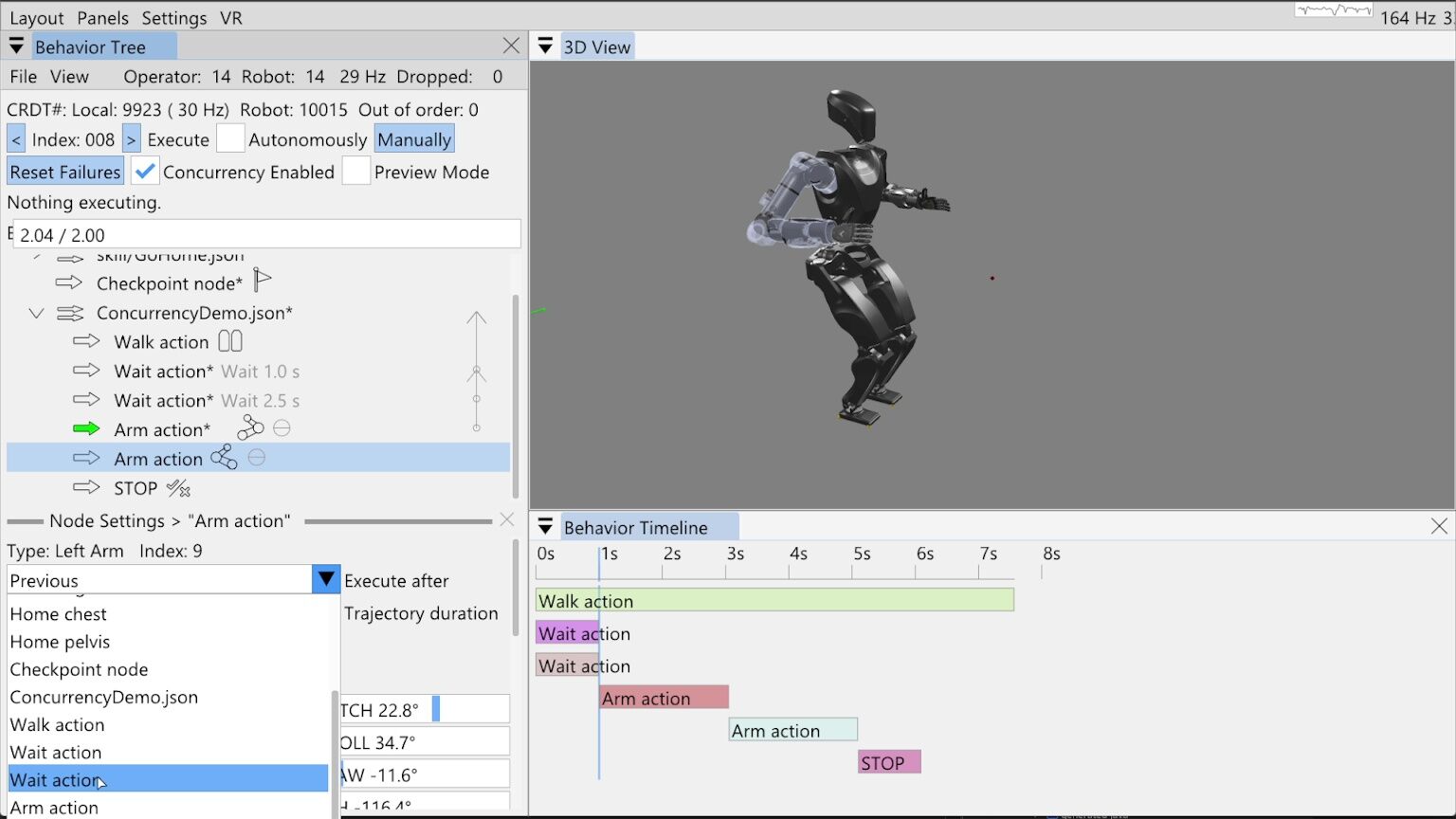}
    \caption{Setting the execute after fields of the arm actions to point to the wait actions.
    A video of this demonstration is available at \url{https://youtu.be/D53dKL-pgeo}.}
    \label{fig:guide_sim_concurrent_execute_after}
\end{figure}

The wait actions need to execute with the walk action, not after it.
To accomplish this, we set the execute after fields of the wait actions to the ``ConcurrencyDemo.json'' sequence node.
The sequence node is never executing, but pointing to it achieves the effect that the wait actions always start with the walk action.
This method also keeps the sequence logic self contained and reusable, as opposed to referencing a specific node earlier than the sequence node.
We then set the execute after field of the right arm action to the first wait action and the execute after field of the left arm action to the second wait action.
In \autoref{fig:guide_sim_concurrent_execute_after}, we show setting the execute after field of the second arm action to the second wait action.

The figure also shows the execute after pointers as arrows on the right side of the tree view that point up from the defining action to the dependency action.
You can see four such arrows that overlap.
When the mouse hovers over a node, the corresponding arrow bolds to make it easy to verify dependencies when they are partially overlapping.
At this point in time, the timeline view is updated to show the general structure of the new behavior configuration.
In this prototype implementation of the timeline view, all wait actions are rendered as 1 second long.
When we run the behavior, the timeline is redrawn with the actual timings.
We will do that now.

\subsection{Concurrent Result}
\begin{figure}[H]
    \centering
    \includegraphics[width=.95\columnwidth]{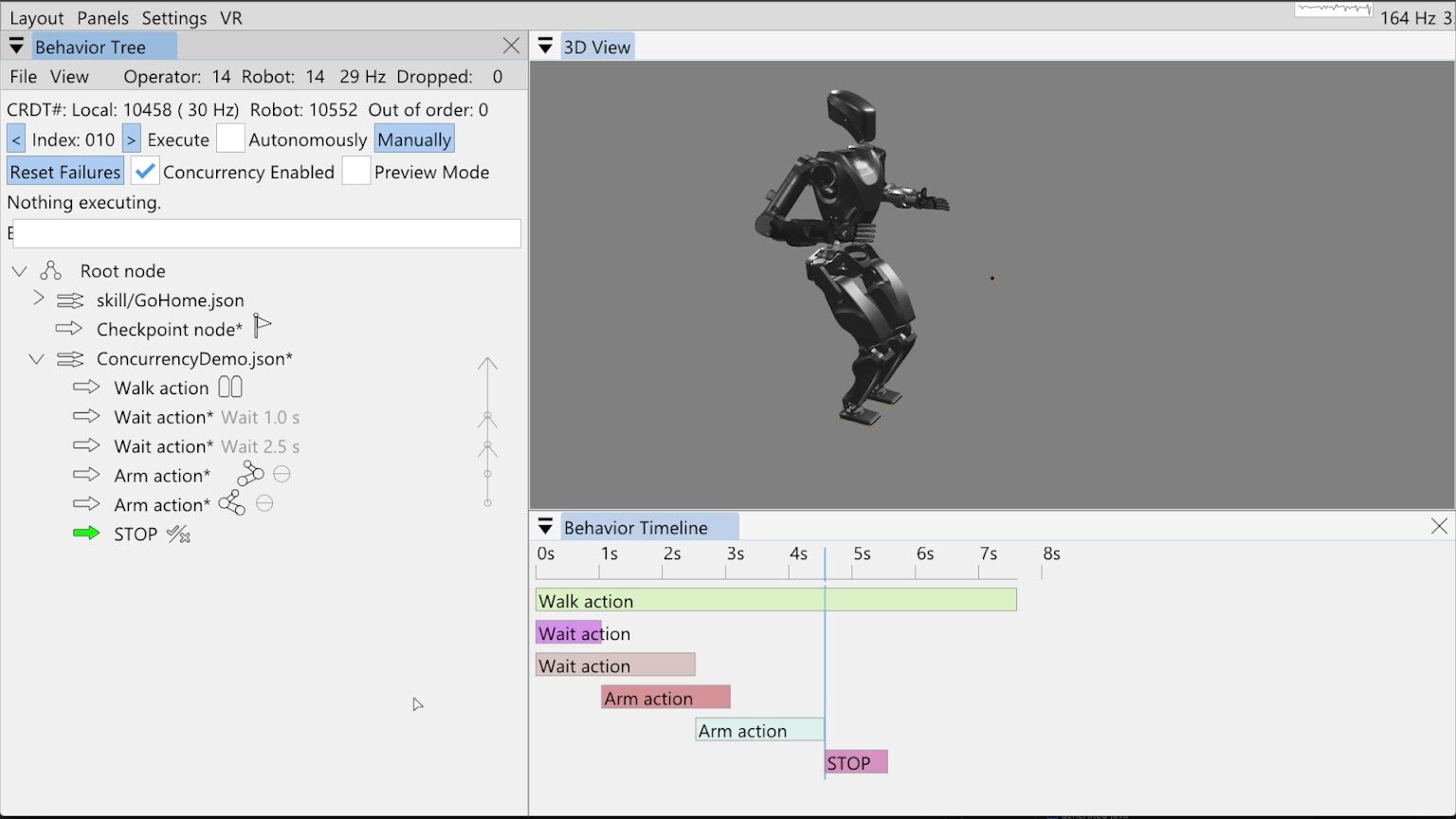}
    \caption{The concurrently executed result.
    A video of this demonstration is available at \url{https://youtu.be/D53dKL-pgeo}.}
    \label{fig:guide_sim_concurrent_result}
\end{figure}

In \autoref{fig:guide_sim_concurrent_result}, we show the result of executing our now concurrent behavior.
The timeline has been updated to reflect the actual action start and stop times of the simulated behavior execution.
As seen by comparing this timeline to the original, the behavior duration went from 11.7 seconds to 7.6 seconds, illustrating how this method can be used to speed up behaviors.

\section{Screw Primitives}

\subsection{Motion Model}
\begin{figure}[H]
    \centering
    \includegraphics[width=.95\columnwidth]{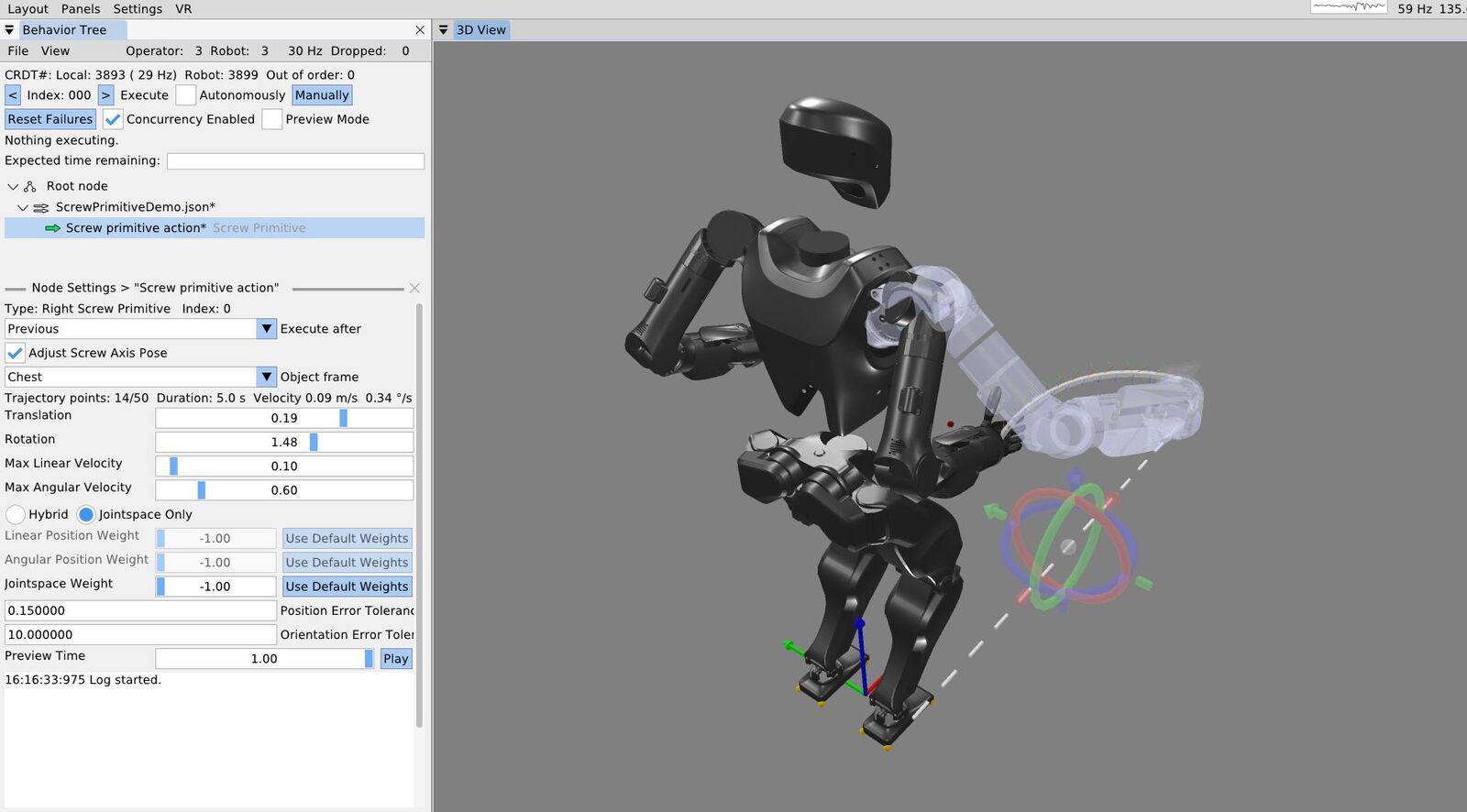}
    \caption{Tuning a screw primitive action.
    A video of this demonstration is available at \url{https://youtu.be/NkX0Jo9TlYs}.}
    \label{fig:guide_sim_screw_primitive}
\end{figure}

\autoref{fig:guide_sim_screw_primitive} shows a screw primitive action being authored.
Our screw primitive action is inspired by Pettinger's 2022 paper~\cite{Pettinger_2022} but differs in implementation.

Our screw primitive action is defined relative to a user-defined screw reference frame, where the axis of the screw is strictly aligned with the local x-axis ($\hat{\mathbf{x}}$).
The motion is parameterized by a total axial translation $\Delta x$ and a total rotation $\Delta \theta$.
\autoref{fig:guide_sim_screw_primitive} illustrates one parameterization for reference and the reference video shows varied parameterizations.

Let the initial position of the end-effector in the screw frame be $\mathbf{p}_0 = [x_0, y_0, z_0]^T$.
The radial distance $r$ from the screw axis to the end-effector is calculated by projecting the position onto the y-z plane:
\[ r = \sqrt{y_0^2 + z_0^2} \]

To compute motion bounds, the total Euclidean distance traversed by the end-effector, $D$, combines both the axial displacement and the tangential arc length:
\[ D = \sqrt{(r \Delta \theta)^2 + (\Delta x)^2} \]

The trajectory duration is determined by enforcing maximum velocity limits.
Given a maximum linear velocity $v_{max}$ and maximum angular velocity $\omega_{max}$, the base duration $T_{base}$ required to complete the motion is dominated by the most restrictive bound:
\[ T_{base} = \max\left( \frac{|\Delta \theta|}{\omega_{max}}, \frac{D}{v_{max}} \right) \]

The trajectory is discretized into $N$ waypoints (where $N$ is limited by the system's trajectory buffer size), yielding $N-1$ segments.
The nominal duration of a single segment is:
\[ \delta t = \frac{T_{base}}{N - 1} \]

To account for smooth acceleration and deceleration without requiring complex spline generation, the implementation applies a temporal boundary condition: the duration of the first and last trajectory segments is doubled (i.e., $2\delta t$), while all intermediate segments retain duration $\delta t$.
Consequently, the total movement duration expands to $T = T_{base} + 2\delta t$.

\subsection{Trajectory Construction}
The trajectory maintains constant base velocities throughout the central portion of the motion.
The scalar velocities are derived from the base duration:
\begin{itemize}
    \item Axial velocity: $v_a = \frac{\Delta x}{T_{base}}$
    \item Tangential velocity: $v_t = \frac{r \Delta \theta}{T_{base}}$
    \item Rotational velocity: $\omega_x = \frac{\Delta \theta}{T_{base}}$
\end{itemize}

At any intermediate waypoint $i$, let the current position of the end-effector in the screw frame be $\mathbf{p}_i = [x_i, y_i, z_i]^T$. The purely radial vector is $\mathbf{r}_{\perp, i} = [0, y_i, z_i]^T$.

The unit tangent vector $\mathbf{u}_{t,i}$ describing the instantaneous direction of rotation is found via the cross product of the screw axis and the radial vector:
\[ \mathbf{u}_{t,i} = \frac{\hat{\mathbf{x}} \times \mathbf{r}_{\perp, i}}{\|\hat{\mathbf{x}} \times \mathbf{r}_{\perp, i}\|} \]

The spatial velocity commands at waypoint $i$, expressed in the local screw frame, are formulated as:
\[ \mathbf{v}_i = v_a \hat{\mathbf{x}} + v_t \mathbf{u}_{t,i} \]
\[ \boldsymbol{\omega}_i = \omega_x \hat{\mathbf{x}} \]
Note that for the initial ($t=0$) and final ($t=T$) waypoints, the spatial velocities are strictly clamped to zero.

The rigid-body poses are iteratively generated by applying incremental transformations in the screw frame. For $K$ total segments (determined by spatial resolution limits), the incremental roll rotation $\delta \theta = \Delta \theta / K$ and translation $\delta x = \Delta x / K$ are calculated.

The pose of the end-effector at step $k$, denoted as $\mathbf{T}_k \in SE(3)$, is generated by prepending the incremental transformation to the previous pose:
\[ \mathbf{T}_k = \mathbf{T}_{k-1} \begin{bmatrix} \mathbf{R}_x(\delta \theta) & \begin{bmatrix} \delta x \\ 0 \\ 0 \end{bmatrix} \\ \mathbf{0}^T & 1 \end{bmatrix} \]
This sequence of generated poses and spatial velocities is finally fed into an inverse kinematics solver to compute the joint-space trajectories.

\subsection{Tuning and Preview}
As seen in \autoref{fig:guide_sim_screw_primitive}, the total axial translation, total rotation, max linear velocity, and max angular velocity can be specified using sliders.
The 3D trajectory visualization is updated in real time as the user drags these sliders to help dial in the motion.
As with the single pose trajectory arm action, the screw primitive has settings for the position and orientation tolerance.
If the final goal pose is not reached within these tolerances, the action fails.

In order to provide a rich intuition about the motion, there is also a motion preview slider for the screw primitive.
The slider may be dragged manually or played back at real time speed in a loop.
The arm is blue when the IK solution is good and red when it is bad.
The behavior author should strive to keep it blue throughout the motion.

\section{Behavior Scene Management Example}

\subsection{Scene Setup}
\begin{figure}[H]
    \centering
    \includegraphics[width=.95\columnwidth]{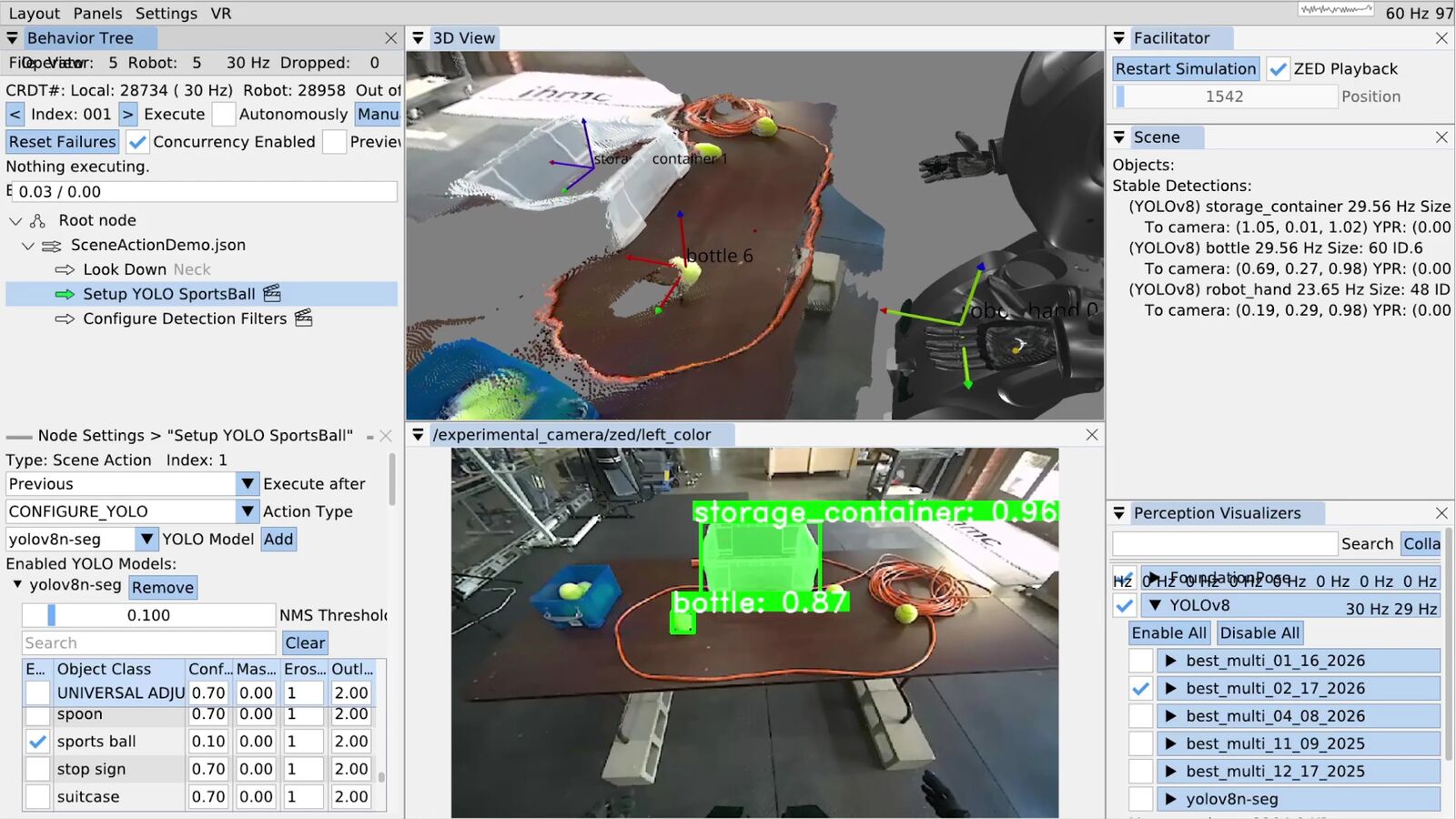}
    \caption{The beginning of our behavior scene demonstration.
    A video of this demonstration is available at \url{https://youtu.be/GN0_XoBxlv4}.}
    \label{fig:guide_sim_scene_action}
\end{figure}

In the next example, we'll cover basic behavior scene management by locking on to a tennis ball and authoring a grasp approach arm pose.
The point of our behavior scene functionality is to provide authorable perception which can increase the capability and adaptability of the robot-human team.
Authorable perception allows the human operator to encode context-specific and expert knowledge into the behavior.
The flexibility provided here can be used to work around several categories of issues both theoretical and practical and both environmental and digital.
Some examples of issues are: the hand occluding an object, a YOLO model having low confidence, and a latency or timing issue in the system.

\autoref{fig:guide_sim_scene_action} shows an example setting being played back in simulation but from a real data recording.
The real environment contains a table with some transparent storage containers and tennis balls on it.
An extension cord is on the table to prevent the balls from rolling off when the robot messes with them.

On the left side of the figure, we see the behavior tree view with a look down action to get the table centered in the camera field of view and two scene actions, denoted by the clapperboard icon inspired by the film industry.
In the bottom left, some scene action settings for the ``configure YOLO'' action type are shown.
In the top center is the 3D scene with a live colored point cloud from the ZED and 3D coordinate frames for the current stable detections.
The bottom center shows the first person ZED camera view with the YOLO detections and segmentations drawn over it.
The detections are annotated with the class name and the confidence of the detection.
In the top right, the behavior scene is shown with no privileged objects yet but with 3 stable detections: storage container, bottle, and robot hand.
Finally, in the bottom right, our YOLO module is able to be turned on and off and each available mode is shown, which can each be toggled on and off and expanded to tune class specific options.

\subsection{YOLO Configuration}
One of the first things we notice here is that the tennis ball is being detected as a bottle.
This is because the initial YOLO model that is enabled, ``best\_multi\_02\_17\_2026'', is one that we trained only for a specific set of classes which do not include the tennis ball.
We aren't sure why it's detecting the ball as a bottle at 87\% confidence.
In any case, to fix this, as shown in \autoref{fig:guide_sim_scene_action}, we author a scene action with type \texttt{CONFIGURE\_YOLO} to switch the YOLO model to the "yolov8n-seg" model.
Additionally, we disable all classes except for the ``sports ball'' class by unchecking the ``Enabled'' checkbox next to where it says ``UNIVERSAL ADJUSTER''.
Then, we scroll to find the sports ball row and check that box.

Unfortunately, the yolov8n-seg model isn't very confident in detecting our tennis balls.
To work around this, on the sports ball row, we lower the minimum ``Confidence'' value from 0.7 to 0.1.
The minimum confidence value acts as a gating filter for detections to be allowed into the persistent detections.
Lowering this value allows us to track the tennis balls at super low confidence levels.
This is okay because we can use this configure YOLO action to adjust the minimum confidence of each class individually.
In behaviors that grasp tennis balls, we make the low confidence threshold safe by requiring each tennis ball object to be within a reachable boundary before grasping it.

The configure YOLO action also supports enabling multiple YOLO models at once.
The behavior author picks the set of YOLO models that will be available at this time and all the class settings for each.
This is useful because sometimes we want to perform tasks involving objects which are detected by different YOLO models.

\subsection{Persistent Detections}
\begin{figure}[H]
    \centering
    \includegraphics[width=.95\columnwidth]{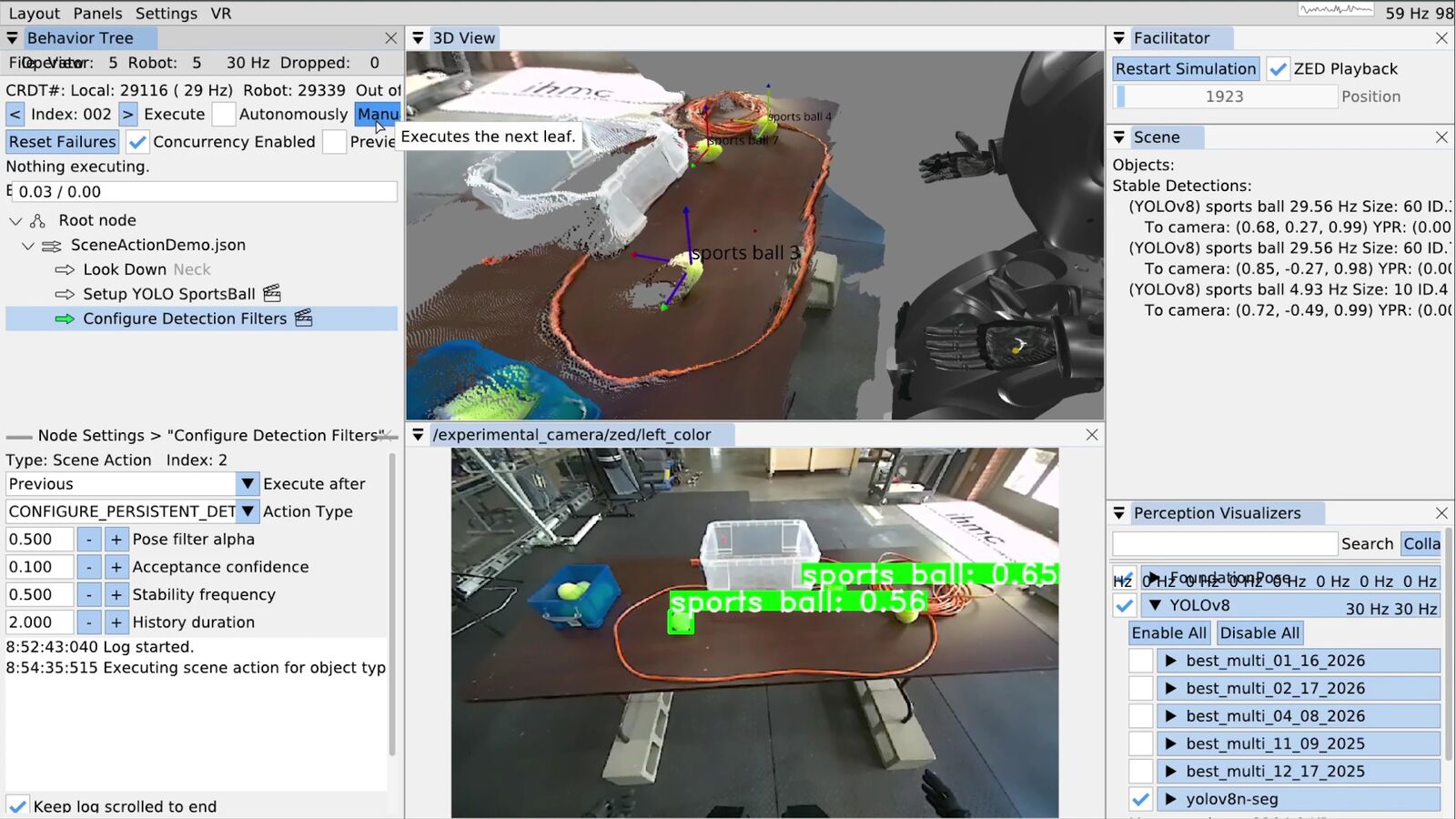}
    \caption{The behavior scene demonstration, after configuring YOLO, we are now configuring persistent detections.
    Persistent detections are referred to in the user interface as ``Stable Detections'', as we only show the stable ones.
    A video of this demonstration is available at \url{https://youtu.be/GN0_XoBxlv4}.}
    \label{fig:guide_sim_scene_action_persistent_detections}
\end{figure}

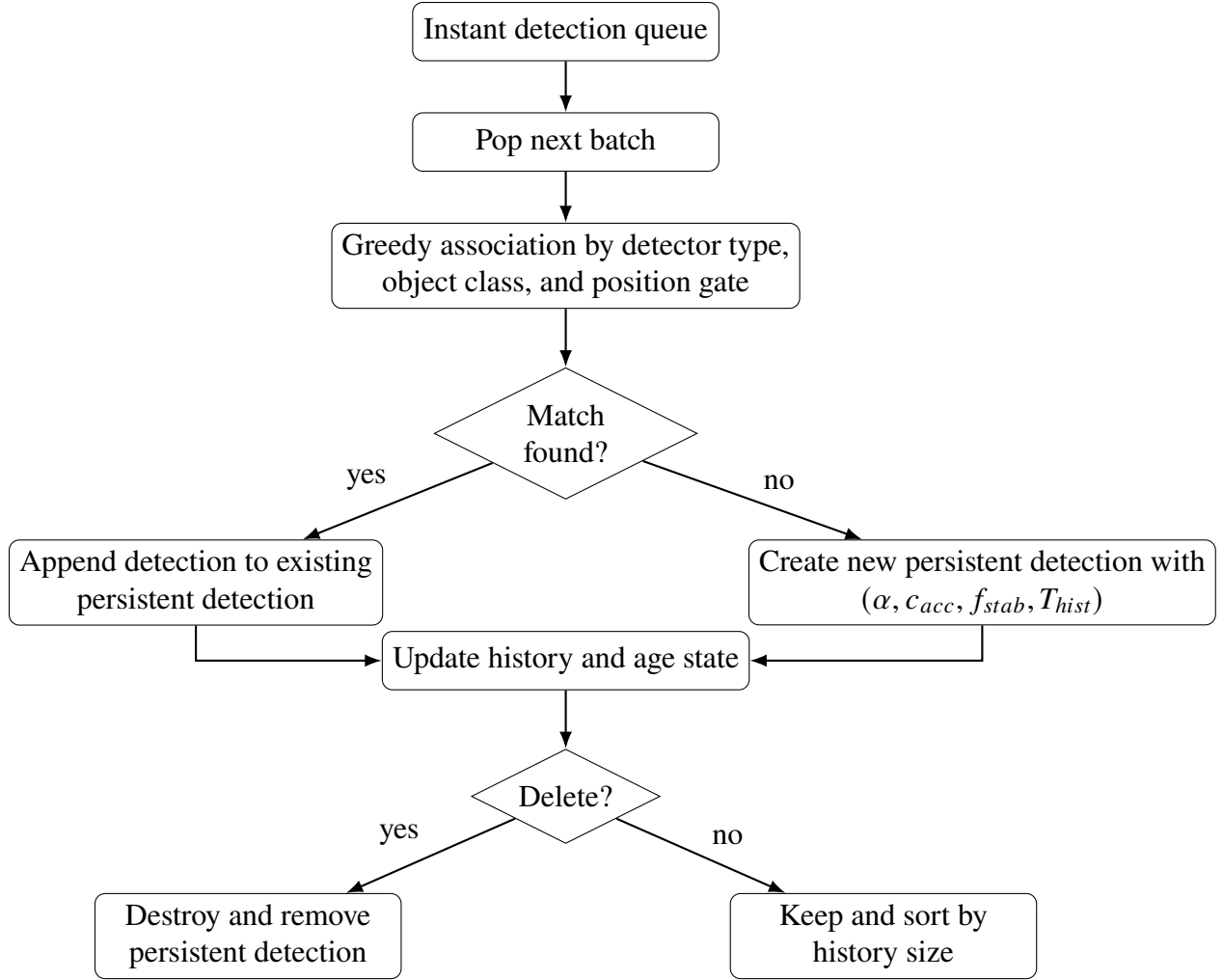
\begin{figure}[H]
        \centering
        \begin{tikzpicture}[
    node distance=7mm,
    block/.style={draw, rounded corners, align=center, minimum width=42mm, minimum height=8mm},
    decision/.style={draw, diamond, align=center, aspect=2, inner sep=1pt, text width=18mm},
    line/.style={-Latex, thick}
]
    \node[block] (queue) {Instant detection queue};
    \node[block, below=of queue] (batch) {Pop next batch};
    \node[block, below=of batch] (match) {Greedy association by detector type,\\ object class, and position gate};
    \node[decision, below=8mm of match] (found) {Match found?};
    \node[block, below left=10mm and 16mm of found] (update) {Append detection to existing\\ persistent detection};
    \node[block, below right=10mm and 16mm of found] (create) {Create new persistent detection with\\ $(\alpha, c_{acc}, f_{stab}, T_{hist})$};
    \node[block, below=18mm of found] (history) {Update history and age state};
    \node[decision, below=8mm of history] (delete) {Delete?};
    \node[block, below left=10mm and 16mm of delete] (remove) {Destroy and remove\\ persistent detection};
    \node[block, below right=10mm and 16mm of delete] (keep) {Keep and sort by\\ history size};

    \draw[line] (queue) -- (batch);
    \draw[line] (batch) -- (match);
    \draw[line] (match) -- (found);
    \draw[line] (found) -- node[above left] {yes} (update);
    \draw[line] (found) -- node[above right] {no} (create);
    \draw[line] (update) |- (history.west);
    \draw[line] (create) |- (history.east);
    \draw[line] (history) -- (delete);
    \draw[line] (delete) -- node[above left] {yes} (remove);
    \draw[line] (delete) -- node[above right] {no} (keep);
\end{tikzpicture}
        \caption{Lifecycle of persistent detections in the behavior scene. New instantaneous detections are greedily associated to existing persistent detections by comparing detector type, semantic class, and a fixed spatial gate. Unmatched detections spawn new persistent detections initialized with the configured filtering and temporal parameters.}
        \label{fig:persistent-detection-lifecycle}
    \end{figure}

Another thing we want to do is configure the persistent detection parameters themselves.
This is accomplished via the \texttt{CONFIGURE\_PERSISTENT\_DETECTIONS} scene action type, seen in \autoref{fig:guide_sim_scene_action_persistent_detections}.
There are four parameters for the persistent detections: pose filter alpha, acceptance confidence, stability frequency, and history duration.
\autoref{fig:persistent-detection-lifecycle} illustrates the instant and persistent detection management process, which happens continuously in the behavior scene.

The parameter $\alpha$ is the pose filter coefficient used when initializing each persistent detection, $c_{acc}$ is the acceptance confidence threshold, $f_{stab}$ is the required stability frequency, and $T_{hist}$ is the history duration maintained by the persistent detection.

For the tennis balls, we set the acceptance confidence threshold to 0.1 and the stability frequency to 0.5 and execute the scene action.
The result is shown in \autoref{fig:guide_sim_scene_action_persistent_detections} where 3 sports balls are tracked as persistent detections.
Note the confidence levels are not too bad in this case at around 50-60\% and the 3 labelled coordinate frames are shown in the 3D view.

\subsection{Setting up an Object}
\begin{figure}[H]
    \centering
    \includegraphics[width=.95\columnwidth]{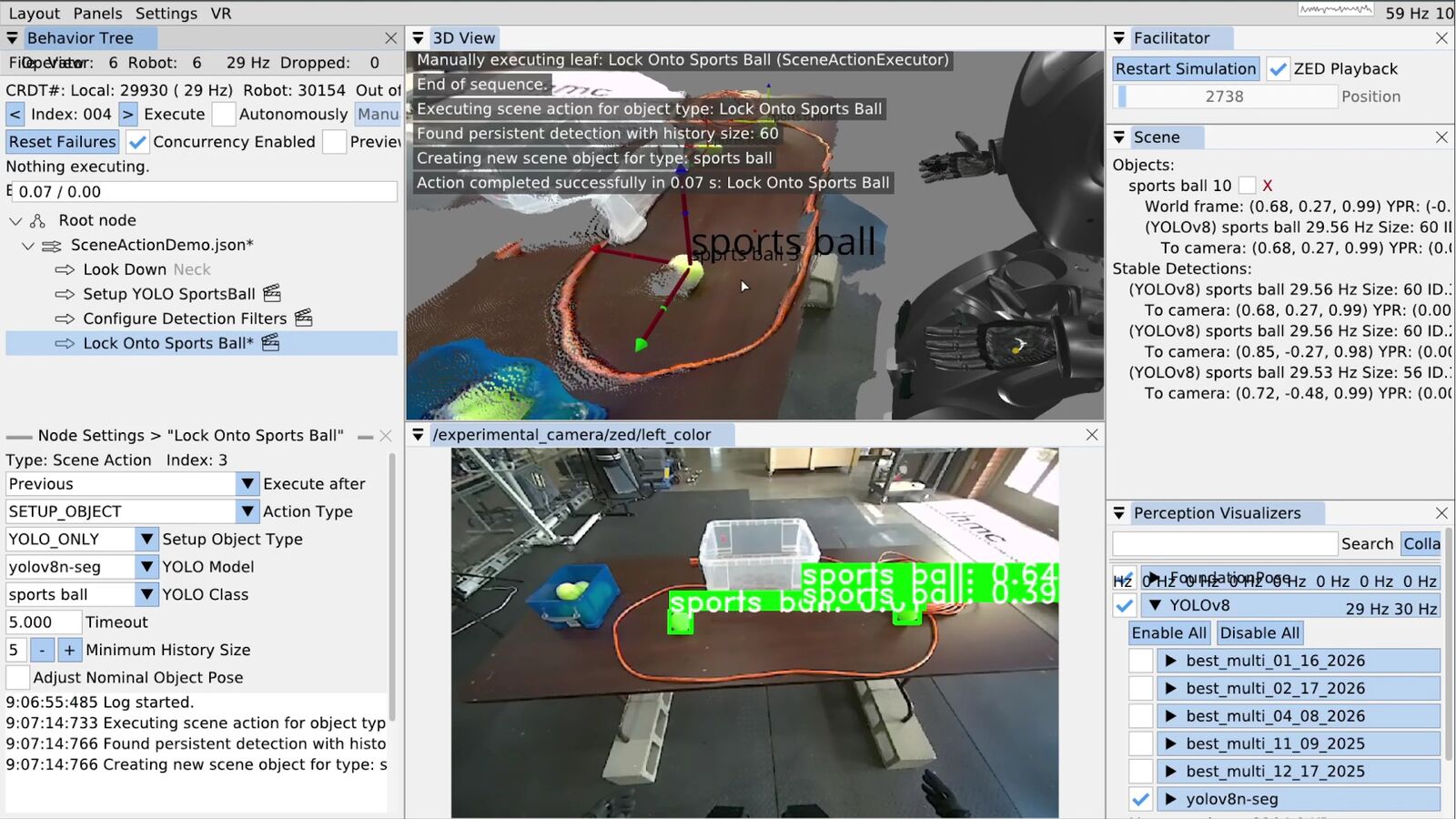}
    \caption{The behavior scene demonstration, after configuring persistent detections and setting up a sports ball object.
    A video of this demonstration is available at \url{https://youtu.be/GN0_XoBxlv4}.}
    \label{fig:guide_sim_scene_action_ball}
\end{figure}

Now that our list of stable persistent detections is populated, we will use a \texttt{SETUP\_OBJECT} scene action to "lock on" to the closest tennis ball.
In \autoref{fig:guide_sim_scene_action_ball} you can see we've added a scene action node called ``Lock Onto Sports Ball''.
In the settings area for that node, we've selected the \texttt{SETUP\_OBJECT} type, the yolov8n-seg YOLO model, and the sports ball YOLO class.
A timeout setting is also available.
Because there may not be matching stable detections in the scene when this action is run, the timeout is a way to wait for one.
The setup object scene action will wait up to the timeout while continuously searching for a match.
If it finds one, the action will immediately complete successfully.
Else, the action will fail when the timeout is reached.
A minimum history size setting is available to further filter our selection of persistent detections.
A shorter history size increases responsiveness but increases the risk of false positives and a longer history time requires tracking an object for longer before it becomes an object candidate.

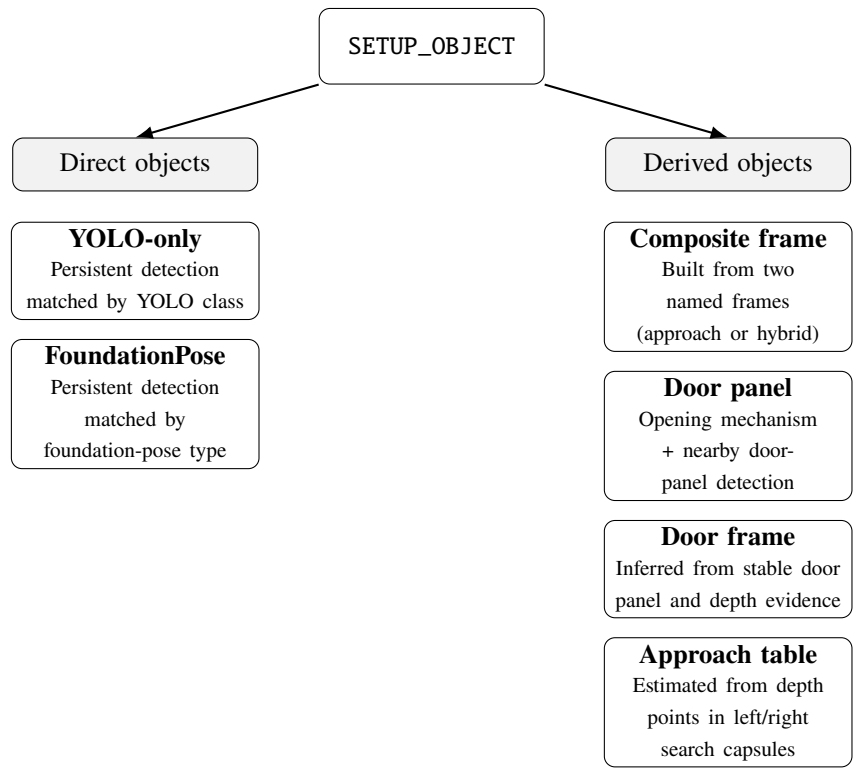
\begin{figure}[H]
    \centering
    \footnotesize
\begin{tikzpicture}[
    >=Latex,
    node distance=3mm,
    box/.style={draw, rounded corners, align=center, text width=31mm, minimum height=10mm, inner sep=2.5pt},
    header/.style={draw, rounded corners, align=center, text width=31mm, minimum height=7mm, inner sep=2pt, fill=black!5},
    line/.style={->, thick}
]
    \node[box, text width=28mm] (setup) {\texttt{SETUP\_OBJECT}};

    \node[header, below left=7mm and 8mm of setup] (direct) {Direct objects};
    \node[header, below right=7mm and 8mm of setup] (derived) {Derived objects};

    \draw[line] (setup.south west) -- (direct.north);
    \draw[line] (setup.south east) -- (derived.north);

    \node[box, below=4mm of direct] (yolo) {\textbf{YOLO-only}\\[-1pt]{\scriptsize Persistent detection matched by YOLO class}};
    \node[box, below=2.5mm of yolo] (fp) {\textbf{FoundationPose}\\[-1pt]{\scriptsize Persistent detection matched by foundation-pose type}};

    \node[box, below=4mm of derived] (composite) {\textbf{Composite frame}\\[-1pt]{\scriptsize Built from two named frames (approach or hybrid)}};
    \node[box, below=2.5mm of composite] (doorpanel) {\textbf{Door panel}\\[-1pt]{\scriptsize Opening mechanism + nearby door-panel detection}};
    \node[box, below=2.5mm of doorpanel] (doorframe) {\textbf{Door frame}\\[-1pt]{\scriptsize Inferred from stable door panel and depth evidence}};
    \node[box, below=2.5mm of doorframe] (table) {\textbf{Approach table}\\[-1pt]{\scriptsize Estimated from depth points in left/right search capsules}};
\end{tikzpicture}
    \caption{Scene object types supported by the \texttt{SETUP\_OBJECT} scene action. Direct objects are based on persistent detections, whereas derived objects are constructed from named frames, paired detections, or depth-based geometric calculations.}
    \label{fig:setup-object-types}
\end{figure}

The setup object scene action, when successful, creates or updates an ``object''.
Our concept of an object does not have to be an actual object, even though it often is.
They are privileged maintainers of reference frames that are subsequently usable for action definitions.
There are two types of objects: direct and derived.
The types are shown in \autoref{fig:setup-object-types}.
The derived types allow us to model articulated objects like doors and implement heuristic environmental feature extraction, such as for table edges.

\autoref{fig:guide_sim_scene_action_ball} also shows the result of executing the setup object scene action.
In the top right, a sports ball can be seen in the ``Objects:'' area.
For each object, we show its world frame pose information and any referenced persistent detections.
Here we can see the YOLOv8 persistent detection that's attached.
That persistent detection is still in the stable detections list, too -- it doesn't get removed.
However, when a persistent detection attached to an object stops tracking, unlike in the stable detections area, it doesn't get removed.
Instead, it renders as grayed out but is still usable by behavior actions in its last seen pose.
A larger coordinate frame is shown for the sports ball object in the 3D view.

\subsection{Freezing Scene Objects}
\begin{figure}[H]
    \centering
    \includegraphics[width=.95\columnwidth]{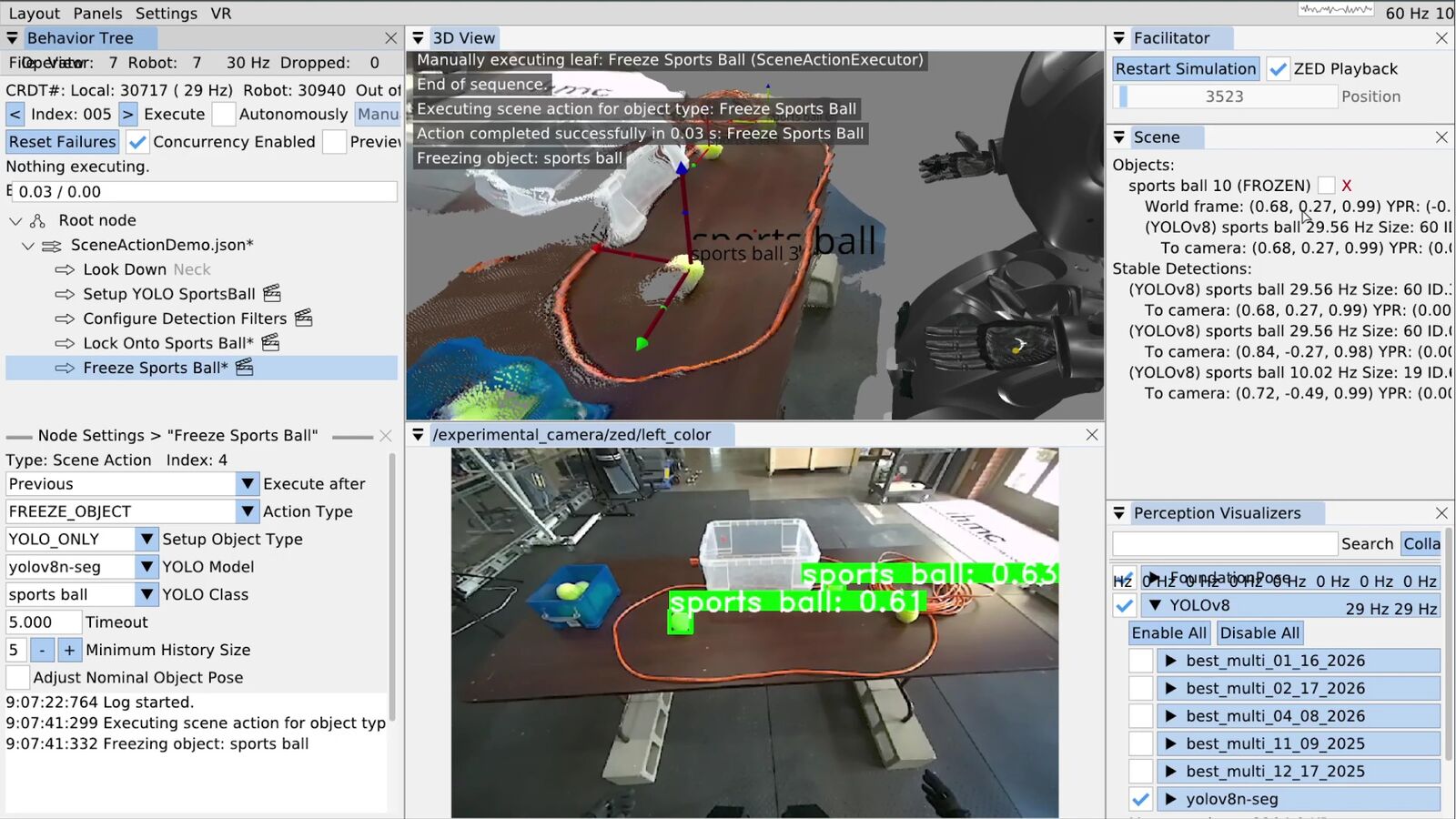}
    \caption{The behavior scene demonstration, after freezing the sports ball object.
    A video of this demonstration is available at \url{https://youtu.be/GN0_XoBxlv4}.}
    \label{fig:guide_sim_scene_action_freeze}
\end{figure}

Another scene action type is \texttt{FREEZE\_OBJECT}.
Freezing an object disables its pose from being updated, causing it to stay in place with respect to world frame regardless of further perceptual tracking.
The ability to freeze an object at any point in the course of a behavior is important for manipulation and dead reckoning.

When grasping an object, it's best to freeze it just before occluding it in any way.
This is because partial occlusion can corrupt the object's pose at the time when it matters most.
For example, we usually have several pre-grasp hand poses with a freeze in between.
The first pre-grasp gets the hand as close as possible without occluding the object.
The second pre-grasp pose gets the hand where the fingers can grab the object.
It's best to put a freeze action between these two.

The reason the pose will be corrupted is because our YOLO detections contain the 3D depth points that lie in the segmentation area.
A YOLO persistent detection is posed at the position of the centroid of those depth points.
If part of the object is occluded, the segmentation will be cropped, causing the centroid to move away from the hand.
As the hand closes in on the object, the position of the object can vary drastically, depending on the shape of the object, the hand, and the viewpoint.

Freezing objects can also be useful for dead reckoning with respect to objects that were seen in the past.
For example, our door traversal footstep plan is authored with respect to the door frame, which is ultimately based on the detection of the door opening mechanism which we can usually no longer see or choose not to see after the door opening.
The frozen frame from the handle pre-grasp is usually used throughout the remainder of the behavior.

Setting up a freeze scene action is easy.
As shown in \autoref{fig:guide_sim_scene_action_freeze}, we created a scene action named "Freeze Sports Ball", set the action type to \texttt{FREEZE\_OBJECT}, and set the YOLO model and class for the sports ball.
When executed, if there is a matched object in the list, it is frozen; otherwise it will wait until the timeout and fail if there is still no match.
In the top right of \autoref{fig:guide_sim_scene_action_freeze}, you can see the sports ball object is now marked as ``FROZEN''.

\subsection{Other Scene Action Types}

There are a few other simple scene action types at the moment.
\texttt{DELETE\_OBJECT} removes a matched object from the list and from being available for actions.
\texttt{CLEAR\_SCENE} removes all objects from the list.
\texttt{FREEZE\_SCENE} freezes every object in the scene.
\texttt{CONFIGURE\_FOUNDATION\_POSE} is an unimplemented placeholder for configuring FoundationPose.

\subsection{Authoring a Frame Based Arm Action}
\label{sec:authoring_a_frame_based_arm_action}

\begin{figure}[H]
    \centering
    \includegraphics[width=.95\columnwidth]{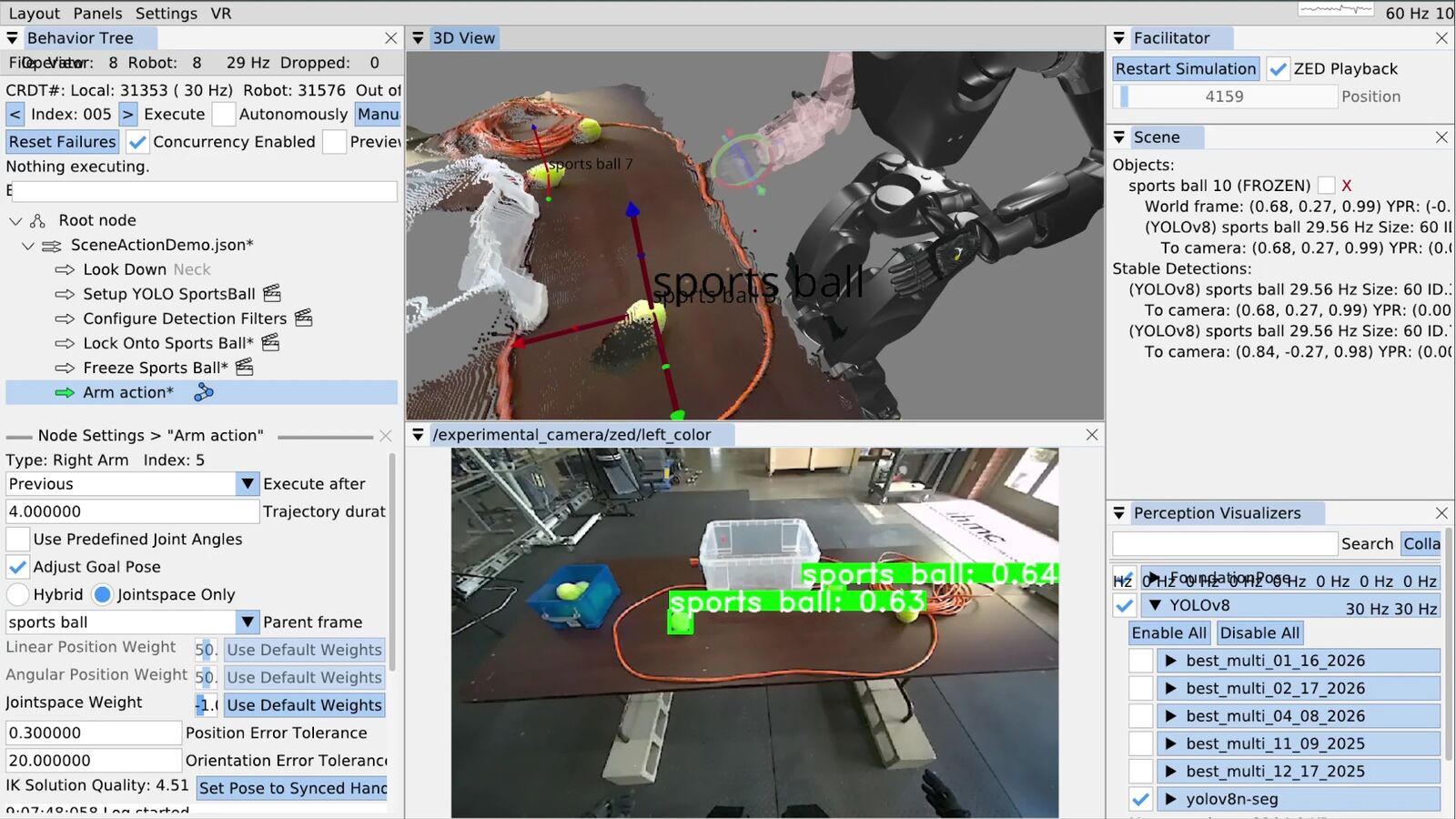}
    \caption{The behavior scene demonstration, authoring an arm action with respect to a sports ball.
    A video of this demonstration is available at \url{https://youtu.be/GN0_XoBxlv4}.}
    \label{fig:guide_sim_scene_action_arm}
\end{figure}

Now that we have locked onto a sports ball, we'll show how to move the arm with respect to it.
We'll go over a full grasp sequence later in the real robot example.
In \autoref{fig:guide_sim_scene_action_arm}, we've added an arm action.
When using the taskspace mode, as opposed to the ``Use Predefined Joint Angles'' mode, there is a ``Parent frame drop-down.
The list of available frames includes a frame for each privileged object in the scene.
In this case, it''s just the sports ball.
We select the sports ball from the list and that's it!
The action is now defined as a hand pose relative to the object.

For position-only objects like the YOLO sports ball, we use the orientation of the robot's chest so frame relative actions are always valid for the robot's current approach angle.
In \autoref{fig:guide_sim_scene_action_arm}, the sports ball frame is frozen, but when objects are not frozen and actively tracked, the arm goal pose will continuously update based on the object's current pose.
The inverse kinematics solution for the arm will likewise continuously update and be displayed in the 3D view.

When we save this behavior, the JSON will contain an object relative translation and rotation with respect to the object.
This makes the action reusable for later runs and future instances of this object.

\section{Door Traversal Footstep Plan Example}

\begin{figure}[H]
    \centering
    \includegraphics[width=.95\columnwidth]{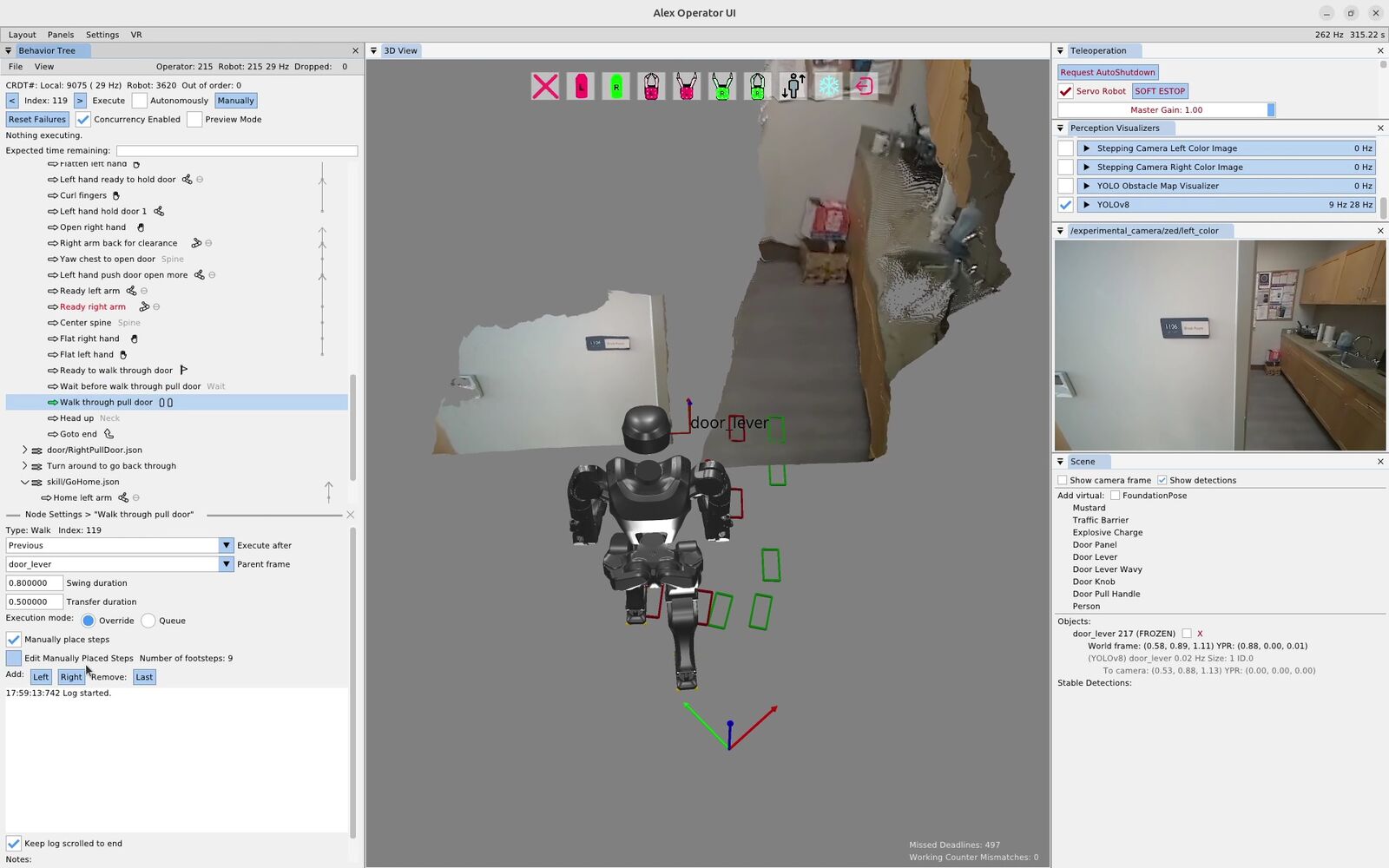}
    \caption{A demonstration of re-authoring a footstep plan online.
    A video of this demonstration is available at \url{https://youtu.be/YcX_Big6__Y}.}
    \label{fig:guide_real_footstep_authoring}
\end{figure}

In this section we'll recount an online, real robot session where we re-authored a set of door traversal footsteps.
In \autoref{fig:guide_real_footstep_authoring}, the robot is situated in front of a doorway and an existing footstep plan action is selected.
At this point, we had tried executing this footstep plan a couple of times, but the robot fell on each attempt.
Since this footstep plan was designed to be robust in the presence of a spring closer, it has an overly difficult side stepping sequence in the beginning.
In this case, the door was open and didn't have a spring closer, so in the interest of task success we decided to re-author an easier to execute footstep plan on the spot.

\subsection{Manually Defining a Footstep Plan}
In the walk action settings for the ``Manually place steps'' mode, there are several features that allow the operator to manage the footstep plan.
To enter the editing mode, which makes the feet graphics selectable in the 3D view with gizmos, the ``Edit Manually Placed Steps'' box is checked.
The settings area displays the current number of footsteps in the plan as a verification component and to be sure there isn't one off the screen somewhere.
On the next line, there are buttons to append a left or right footstep to the end of the plan.
There is also a button to remove the last step of the plan.
These editor features are simplistic but enough to get the job done.

\begin{figure}[H]
    \centering
    \includegraphics[width=.95\columnwidth]{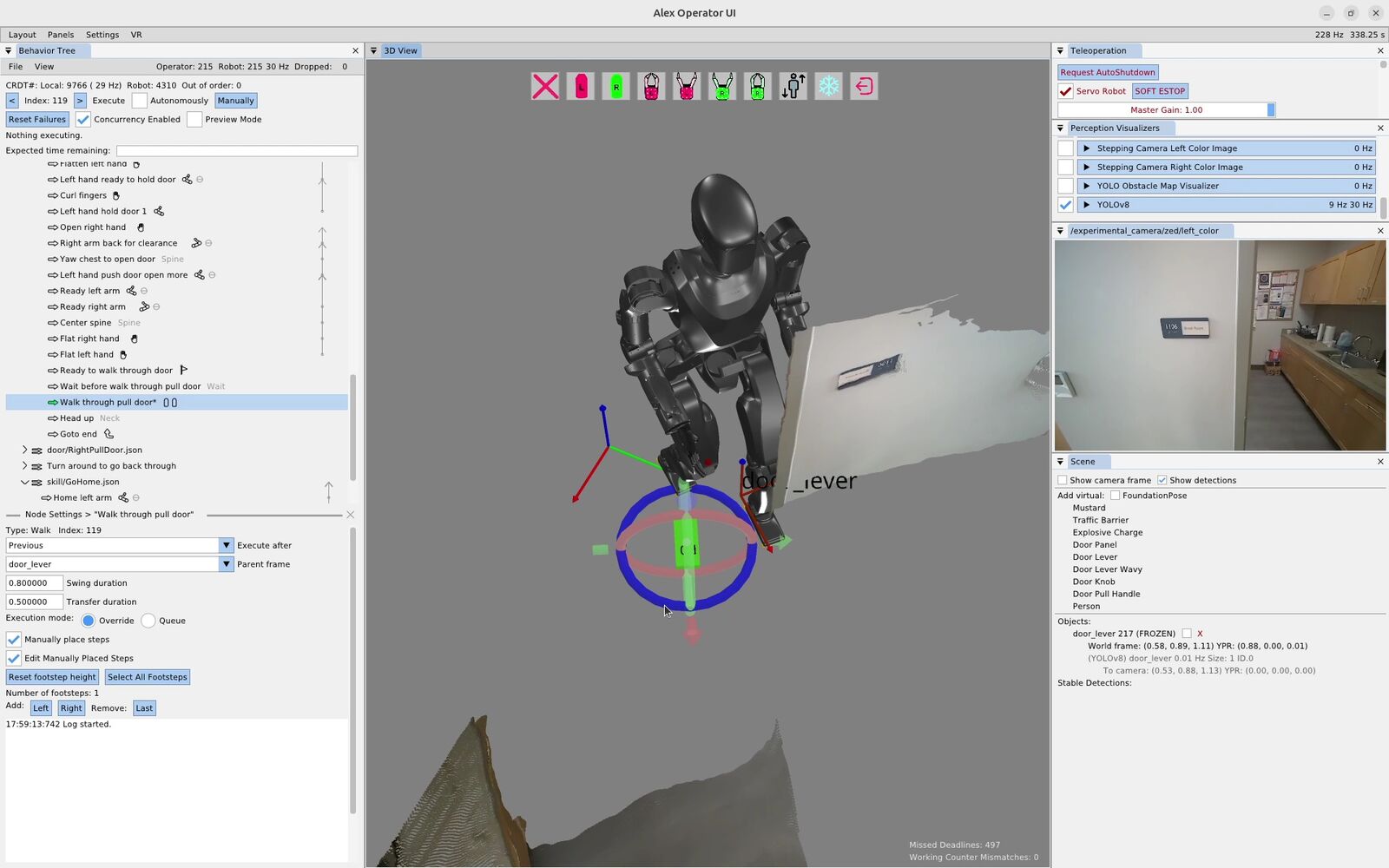}
    \caption{Adding the first step of a manual footstep plan.
    A video of this demonstration is available at \url{https://youtu.be/YcX_Big6__Y}.}
    \label{fig:guide_real_footstep_authoring_first_step}
\end{figure}

\begin{figure}[H]
    \centering
    \includegraphics[width=.95\columnwidth]{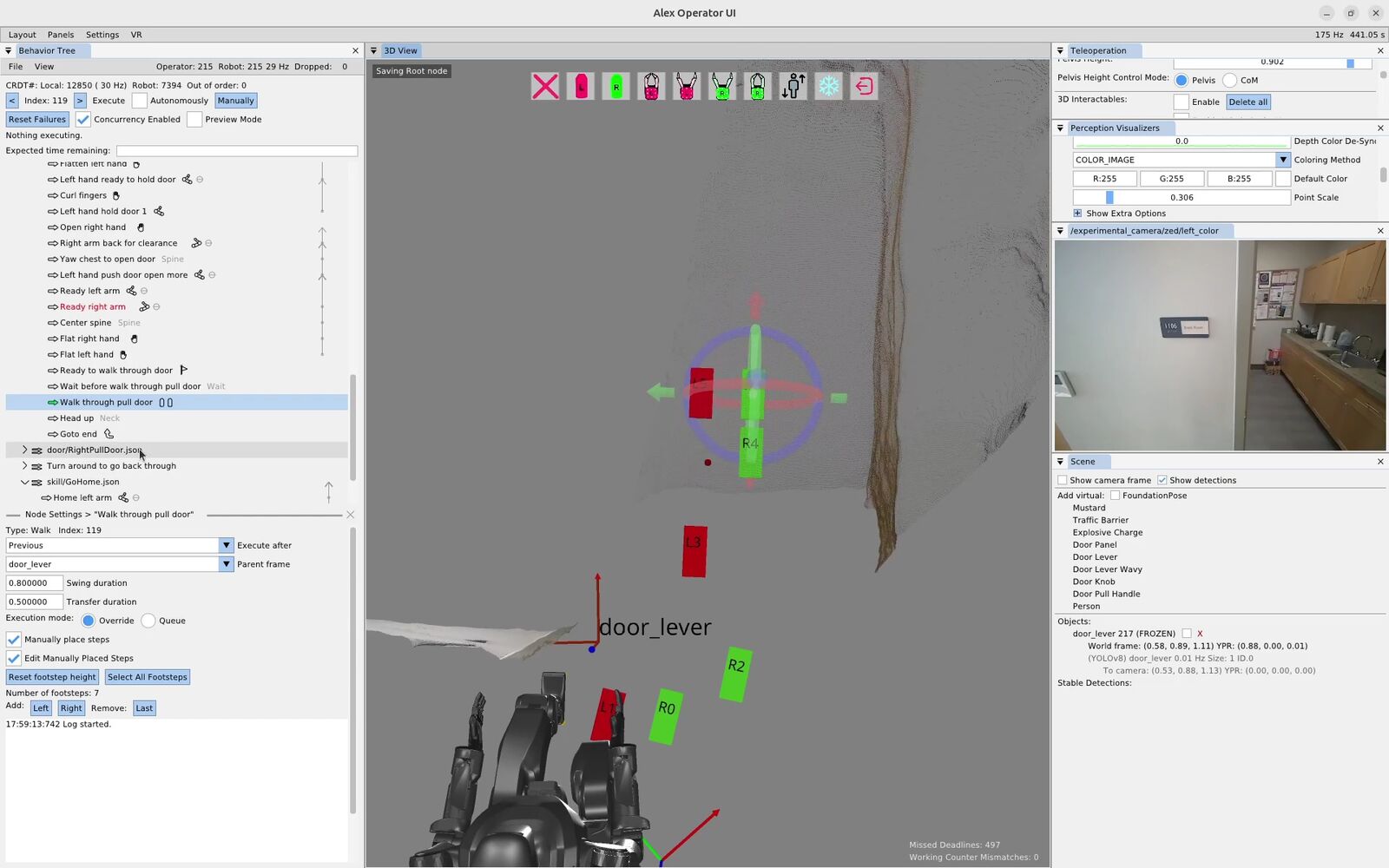}
    \caption{A fully authored manual footstep plan that was executed successfully by the robot.
    A video of this demonstration is available at \url{https://youtu.be/YcX_Big6__Y}.}
    \label{fig:guide_real_footstep_authoring_complete}
\end{figure}

To clear the plan, we first click the remove last button a bunch of times until the plan is empty.
Then, we click the ``Right'' button to add a right footstep to the plan.
We click the footstep in the 3D view and move it to the desired location using the gizmo, as seen in \autoref{fig:guide_real_footstep_authoring_first_step}.
This process is repeated until the plan is complete.
The completed door traversal footstep plan can be seen in \autoref{fig:guide_real_footstep_authoring_complete}.
This plan took about 1 minute and 30 seconds to create.

\section{Fallback Mechanism Example}
\label{sec:fallback_node}

\begin{figure}[H]
    \centering
    \includegraphics[width=.95\columnwidth]{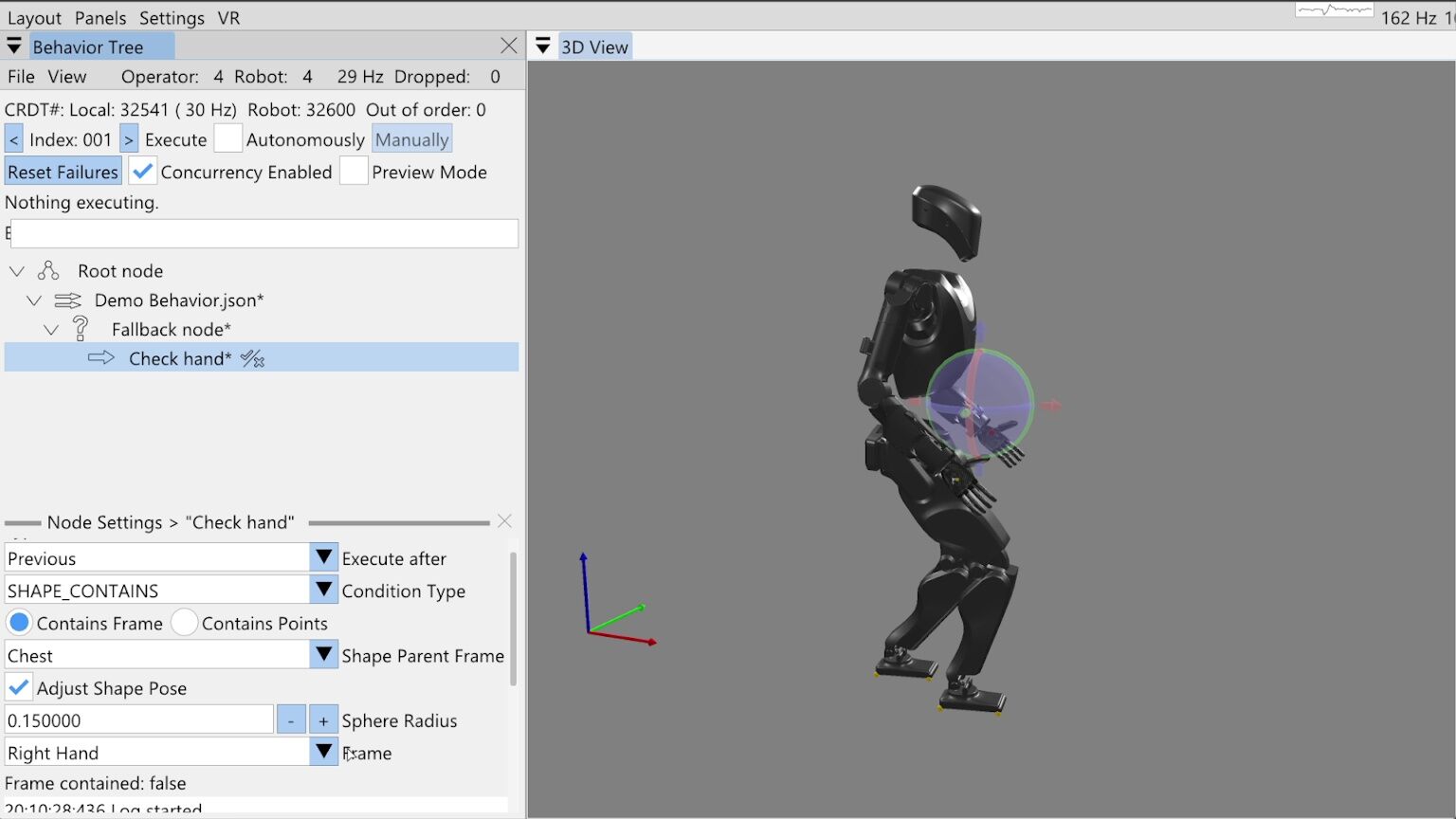}
    \caption{The beginning of our fallback node demonstration.
    A video of this demonstration is available at \url{https://youtu.be/mw4z4H4K9o0}.}
    \label{fig:guide_sim_fallback}
\end{figure}

Now we will cover fallback node operation.
A fallback node is like an if-else statement, but since the if condition evaluation is performing an action, we call it a try-catch.
A fallback node therefore has two parts: the try and the catch.
Each part can contain one or more actions, however, unlike a Behavior Tree fallback node from the literature~\cite{2018_colledanchise}, our fallback node has only one try clause and the catch clause is just an action sequence of unlimited length.

We also support concurrent actions in the fallback node.
To do this, we expand the try to contain the first concurrent action set.
This could be one or more actions with the rule that they must execute together.
If any action of the concurrent sequence fails, the try fails and the catch is executed once all try nodes are finished executing.
Else, if the entire concurrent try set succeeds, the catch is skipped and the node after the fallback is next for execution.

Our fallback node does not currently support nesting, but supporting it would be a desirable improvement.
Nesting try-catch blocks could increase behavior expressiveness.
In general, it would be good to look through the most used mechanisms in programming language and think about how they could be included as runtime-editable behavior elements.

In \autoref{fig:guide_sim_fallback}, we have started on a fallback example demonstration, where we have added a fallback node and a try node.
In this demonstration, we want to check if the right hand is raised to a certain position and raise it if it isn't.
Once the hand is raised, we will yaw the spine.
To check the hand position, we have added a ``shape contains'' condition node as the try action and named it ``Check hand''.

\subsection{Condition Nodes}

We currently have six types of condition nodes: \texttt{ALWAYS\_FAIL}, \texttt{ALWAYS\_SUCCEED}, \texttt{COUNTER}, \texttt{LLM}, \texttt{PROXIMITY}, and \texttt{SHAPE\_CONTAINS}.
A condition node is different from an action in that it does not directly perform an action, but instead is responsible for making a decision resulting in success or failure.
Success and failure for a condition node, however, have the same effect on behavior execution as actions: if the node succeeds, the behavior keeps going and, if the node fails, it halts execution by disabling autonomous mode.
The exception is that when a node fails in the try part of a fallback node, the behavior keeps going by executing the catch.

The ``always succeed'' and ``always fail'' condition node types are useful for testing and operation.
The ``always fail'' node is especially useful if you want to test out a part of the behavior in autonomous mode but have it stop at a certain point.
In this case, we'll often add a temporary ``always fail'' condition node named ``STOP'', as we did in \autoref{sec:wait_nodes_dependencies}.

The counter node maintains a persistent count and a tunable limit.
Each time it is executed, the count is increased.
If the count reaches the limit, the counter condition node fails.

The LLM condition node provides the ability to query a large language model to determine success or failure.
It has an authorable system prompt, repeated prompt, and boolean response matcher.
We covered the LLM node in more detail near \autoref{fig:2025_llm_condition_node}.

The proximity node checks the distance between two behavior reference frames.
If the distance is between the tunable minimum and maximum, the condition succeeds.
We also covered the proximity node in some more detail near \autoref{fig:2025_proximity_condition}.

For this example, we'll use the shape contains condition node to check that the right hand frame is within the bounds of a sphere.
In \autoref{fig:guide_sim_fallback}, our shape contains condition node is selected with the settings showing in the bottom-left.
There are two modes for the shape contains condition: contains frame and contains points.
We'll be using contains frame in this example, but the contains points functionality, which counts the number of live point cloud points in the virtual shape, is covered near \autoref{fig:alex_shape_contains_condition}.

\subsection{Shape Contains Frame Condition}
In \autoref{fig:guide_sim_fallback}, you'll see there are four main settings for the shape contains frame functionality.
The shape parent frame, the shape pose, the sphere radius, and the frame to check containment of.
We allow defining the shape's pose with respect to any robot or object frame just like we do for actions.
Checking the ``Adjust Shape Pose'' box enables the shape to be moved in the 3D view with a pose gizmo, as shown in the figure.
Currently, just spheres are supported, but adding more types of shapes would be a good improvement.

\begin{figure}[H]
    \centering
    \includegraphics[width=.95\columnwidth]{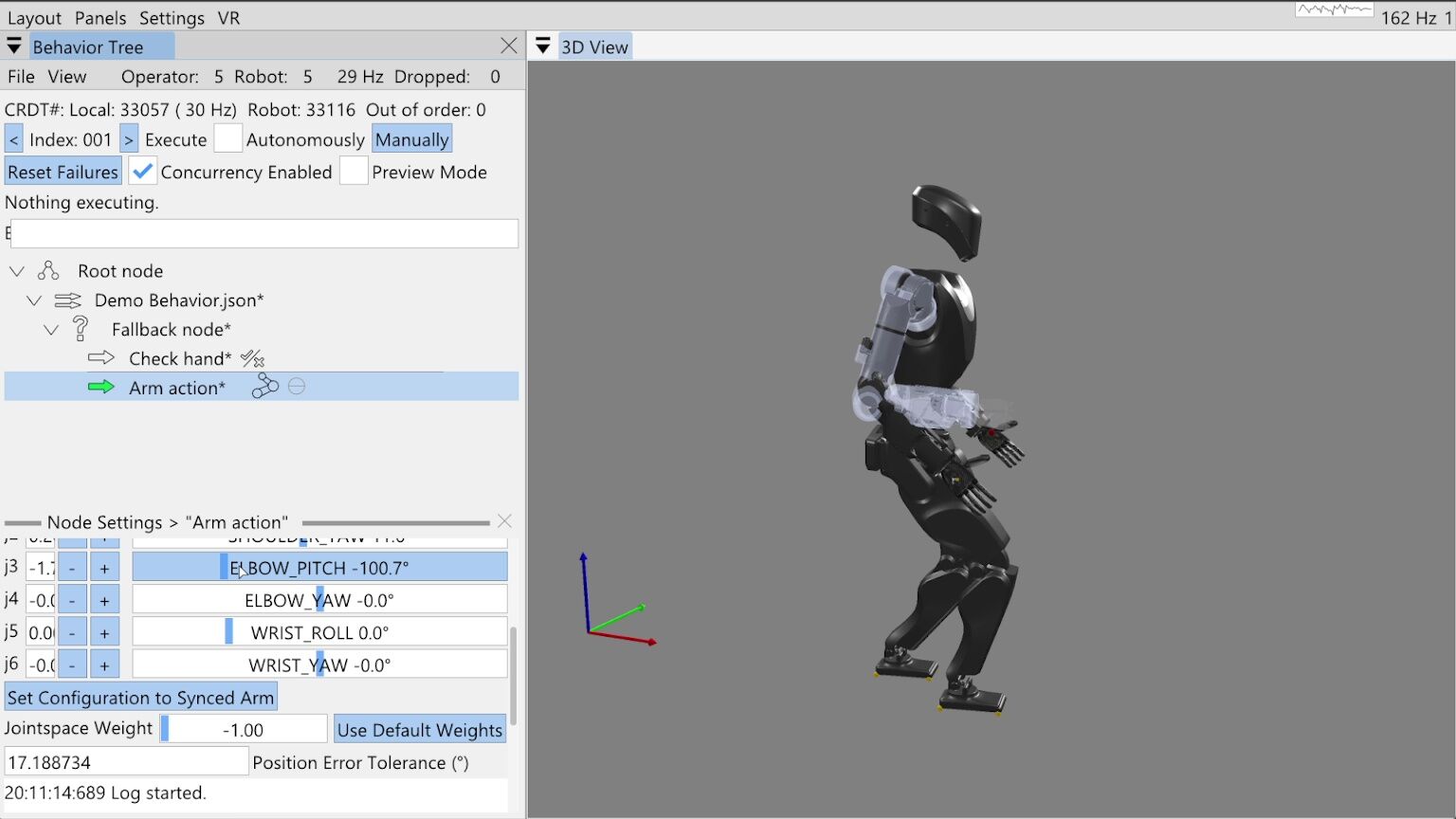}
    \caption{Adding an arm action to the fallback catch.
    A video of this demonstration is available at \url{https://youtu.be/mw4z4H4K9o0}.}
    \label{fig:guide_sim_fallback_arm}
\end{figure}

Pairing this condition node with our fallback gives us a mechanism to react to an unknown state, like where the hand is.
To robustify our behavior and apply corrective action if the hand is not where we want it, we will now add an arm action to the fallback catch, as shown in \autoref{fig:guide_sim_fallback_arm}.
When we add the catch action, because it is not concurrent with the condition in the try, we see a thin horizontal line in the tree editor.
This line, between the ``Check hand'' and ``Arm action'' nodes, shows us where the try-catch delineation is.

\subsection{The Goto Node}

\begin{figure}[H]
    \centering
    \includegraphics[width=.95\columnwidth]{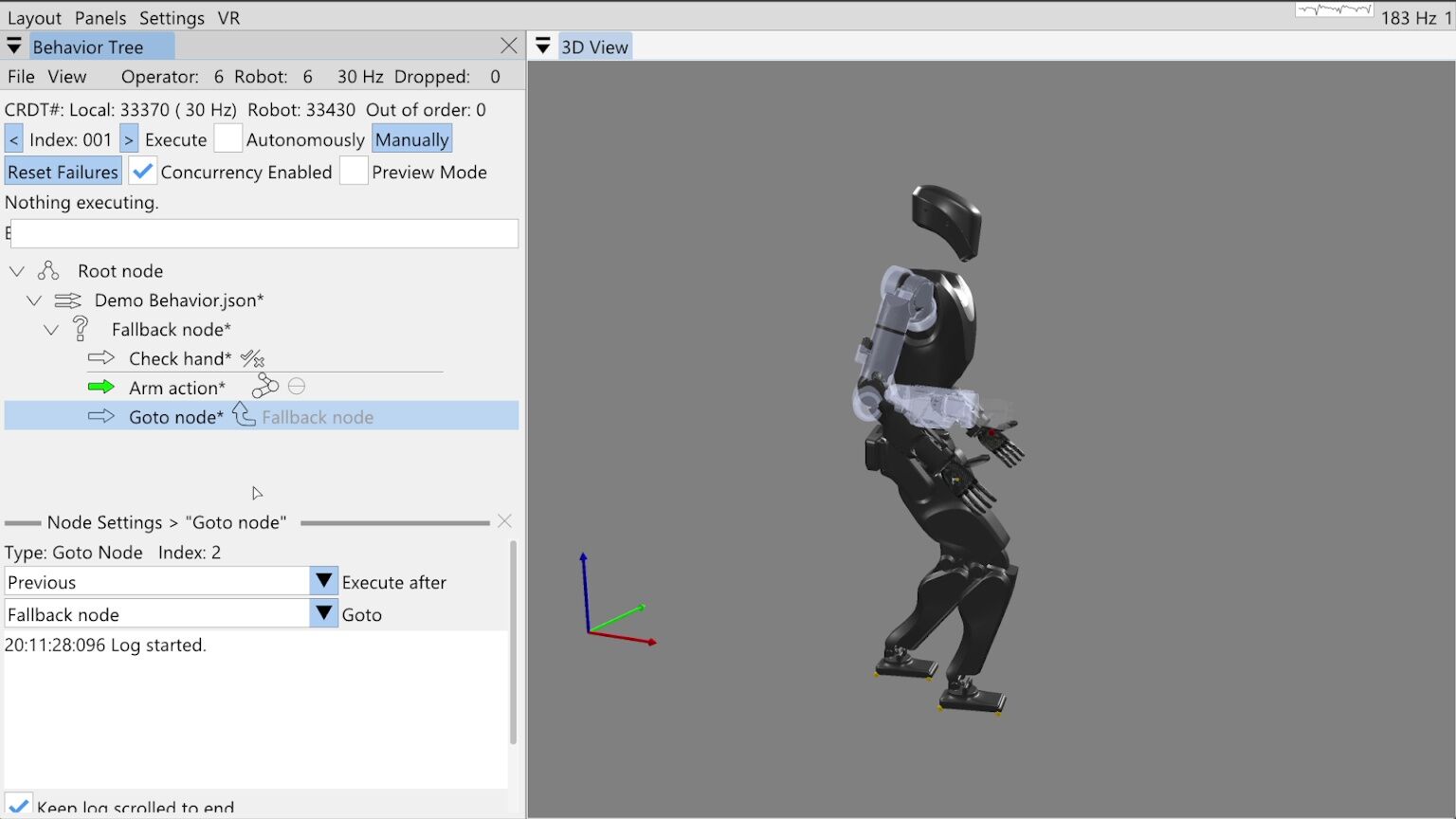}
    \caption{Adding a goto node to implement a while loop.
    A video of this demonstration is available at \url{https://youtu.be/mw4z4H4K9o0}.}
    \label{fig:guide_sim_fallback_while}
\end{figure}

To show how you can implement while-loop-like behavior, we add a goto node after the arm action in the fallback catch, as shown in \autoref{fig:guide_sim_fallback_while}.
The goto node has a goto field that refers to another node by name.
When the goto node is executed, the next execution index is set to that node.
Here we have selected the ``Fallback node'' as the node to goto, which is a general way of saying to reexecute the try, as the ``Fallback node'' is a control node, not an action.
When we point to control nodes, it's effectively the same as pointing to the next action.

Since we want to be able to save this goto reference by name in the JSON, there is the possibility of ambiguity in the node's name.
To handle the case when there are multiple nodes of the same name in the tree, we search for the node to goto by proximity to the goto node.
The closest matching node is chosen with levels of priority: children recursively come first, then siblings, then parents recursively.
Our choice of priority is somewhat arbitrary.
We recommend using unique node names where possible to avoid this ambiguity.

\subsection{Reactive Fallback Demonstration}
\begin{figure}[H]
    \centering
    \includegraphics[width=.95\columnwidth]{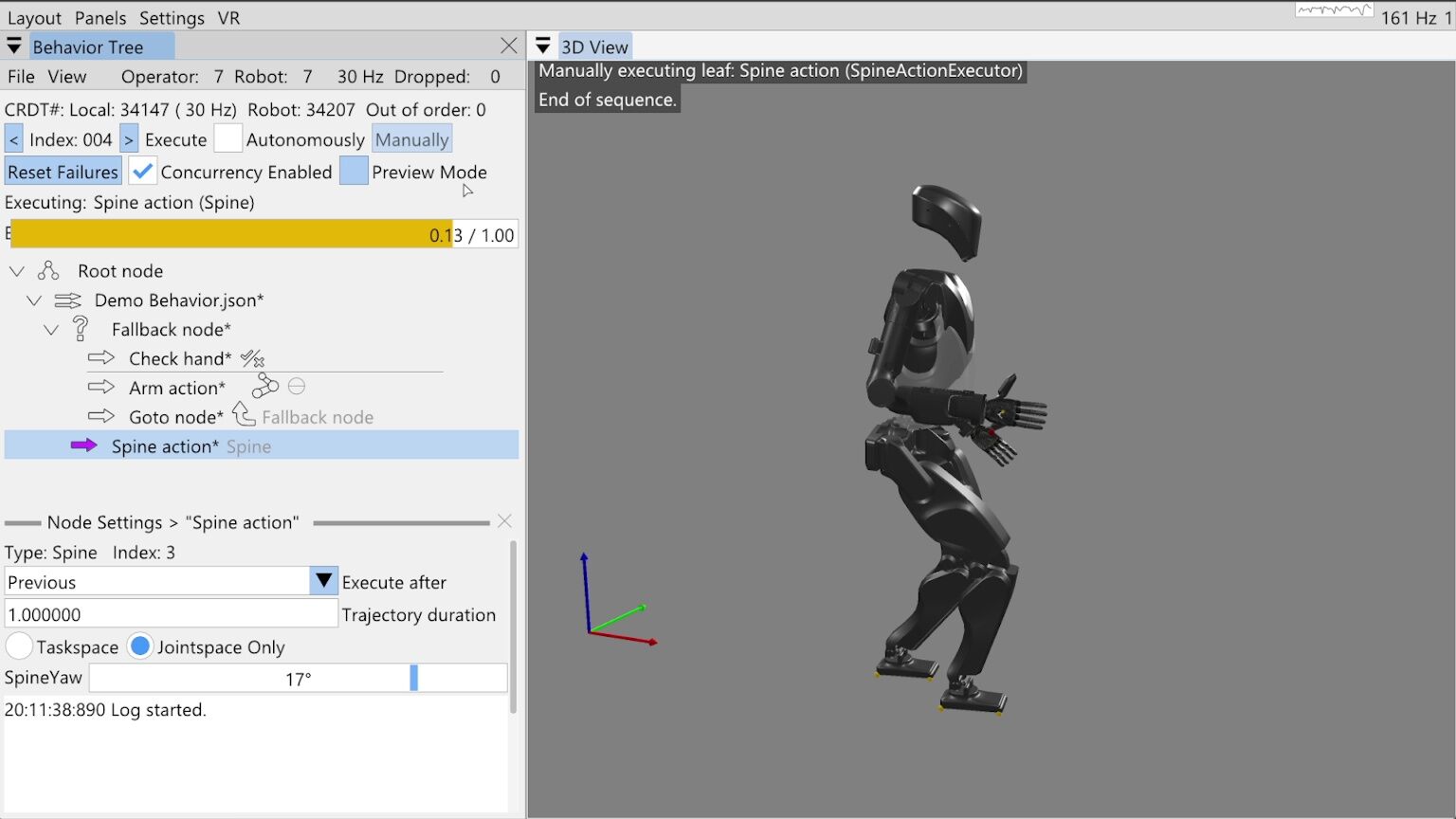}
    \caption{Completing our fallback demonstration by reactively moving the arm to the correct location and executing the spine yaw motion.
    A video of this demonstration is available at \url{https://youtu.be/mw4z4H4K9o0}.}
    \label{fig:guide_sim_fallback_end}
\end{figure}

Finally, we add our spine action to turn the spine 17 degrees to the left as shown in \autoref{fig:guide_sim_fallback_end}.
Notice this node is not a child of the fallback node, but a sibling.
We run through the behavior in the manual mode, inspecting the flow of execution.
We run check hand, it fails, then the arm action runs to move the hand into the correct location.
The goto node causes us to re-evaluate the check, which now succeeds, sending us on to the spine action, which executes successfully.

\section{Real Robot Example: Repeated Door Opening}

In this final section of the usage guide, we'll walk through a real robot authoring session in which we created a repeated door opening behavior from scratch in 32 minutes.
For this experiment, we used the Unitree H1-2 robot.
At the end of the authoring phase, we immediately entered a reliability test where we performed over 30 successful door openings with continuous autonomy.
We didn't run it until failure, just until we got bored.

\subsection{Starting Things Up}
Starting things up typically requires powering on the robot and the operator computer.
In this guide we'll assume the robot is tetherless but starts on a hoist to get it into its initial standing state.
We'll assume powering on the robot will also power the computers and that they turn on and that the motors have power.

After the robot's on-board computers have started, we will deploy the latest version of the control and autonomy software if necessary.
Once that is done, we launch the control and autonomy code on the robot via an SSH command line session and start the robot operator interface.
The next step is to set the robot down and transition the controller to walking.
We have a stand prep state and a walking state, with buttons in the operator interface to perform those transitions.

\subsection{Standing and the Operator Interface}
\begin{figure}[H]
    \centering
    \includegraphics[width=.95\columnwidth]{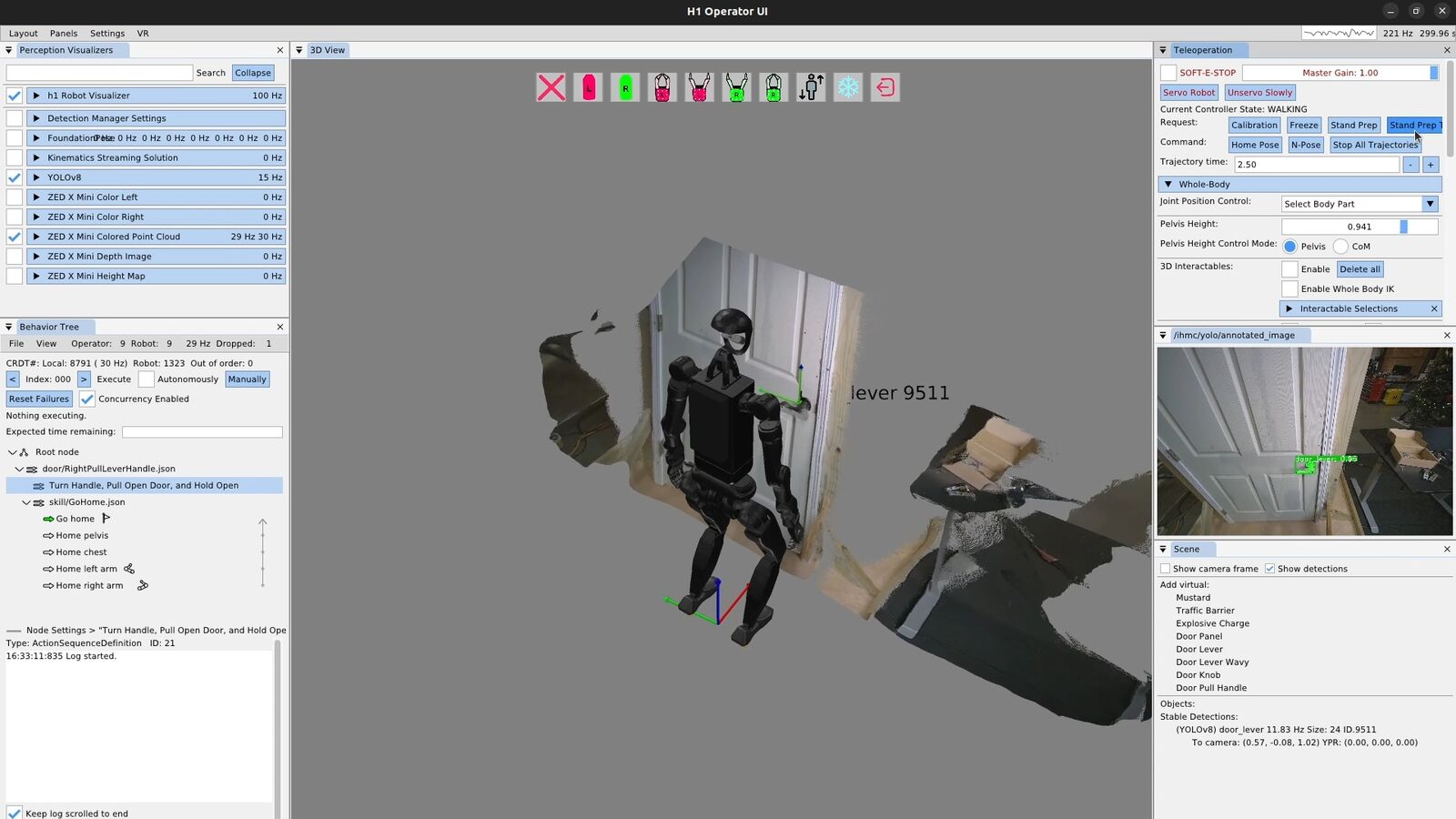}
    \caption{The initial state of the application after we've started the robot and the controller and autonomy processes and created a blank initial behavior action sequence.
        A video is available at \url{https://youtu.be/7VGFufJWaR4}.
    }
    \label{fig:guide_unitree_initial_setup}
\end{figure}

The robot should now be standing, and the operator interface should be showing the robot state and perception data.
It should look like \autoref{fig:guide_unitree_initial_setup}.
At this point we check a couple of things to make sure everything is ready to go:
\begin{enumerate}
    \item The robot visualizer is enabled via the checkbox, the robot is visible in the 3D scene, and the update frequency is as expected, in this case 100 Hz.
    \item YOLOv8 is enabled with the checkbox and is running at some frequency in the 5-20 Hz range.
    \item The ZED X Mini colored point cloud is enabled with the checkbox, is running at ~30 Hz and is being updated in the 3D scene.
    \item The YOLO annotated image is showing, with any detections shown that you would expect, in this case, the door lever and that it is at the expected confidence level.
    \item The behavior tree view shows the synchronization frequency at around 30 Hz.
    \item Stable detections appear for the YOLO detections in the scene panel.
\end{enumerate}

To start, we've created sequence nodes for our door opening behavior, with the top level being saved to JSON with the ``*.json'' extension.
Saving the behavior frequently is important to avoid losing work.
We've also loaded the go home skill as a convenient way to reset the robot when we need to.

\subsection{The Ability Hand Action}
\begin{figure}[H]
    \centering
    \includegraphics[width=.95\columnwidth]{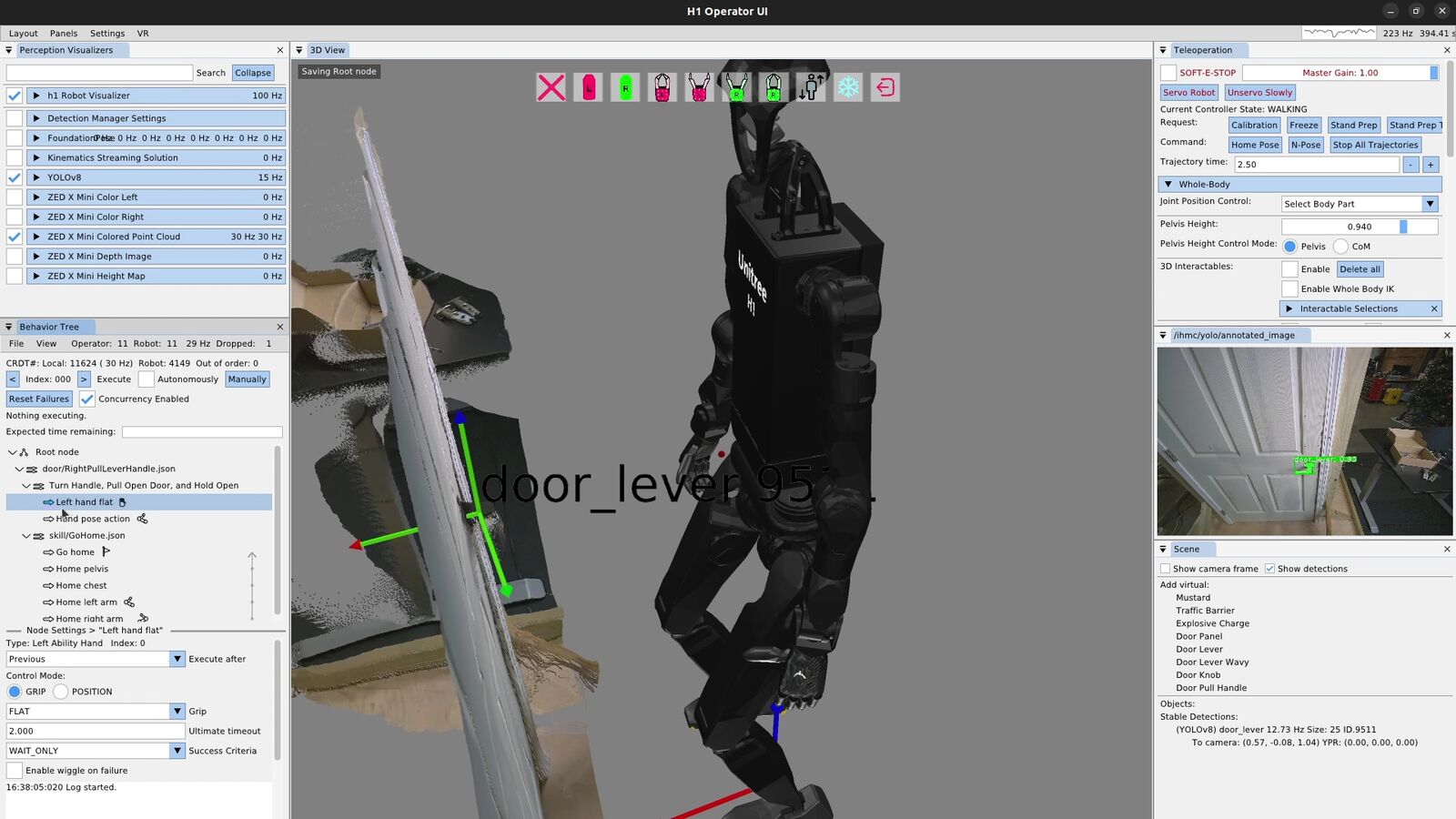}
    \caption{Adding a flat Ability Hand action.
        A video is available at \url{https://youtu.be/7VGFufJWaR4}.
    }
    \label{fig:guide_unitree_hand}
\end{figure}

We will be using the left arm and hand to turn the lever and open the door.
The first thing we need to do is specify that the finger configuration should be ``flat hand'' at the start.
To do this, we add a left Ability Hand action, as shown in \autoref{fig:guide_unitree_hand}, and set the "grip" field to \texttt{FLAT}.
There are several common grip types available as presets, including \texttt{OPEN}, \texttt{CLOSE}, \texttt{PINCH}, \texttt{FLAT}, \texttt{HOOK}, \texttt{RELAX}, \texttt{DOOR\_LEVER\_OPEN}, \texttt{DOOR\_LEVER\_CLOSE}, \texttt{DOOR\_LEVER\_CRUSH}, \texttt{KEY\_OPEN}, and \texttt{KEY\_CLOSE}.

\begin{figure}[H]
    \centering
    \includegraphics[width=.95\columnwidth]{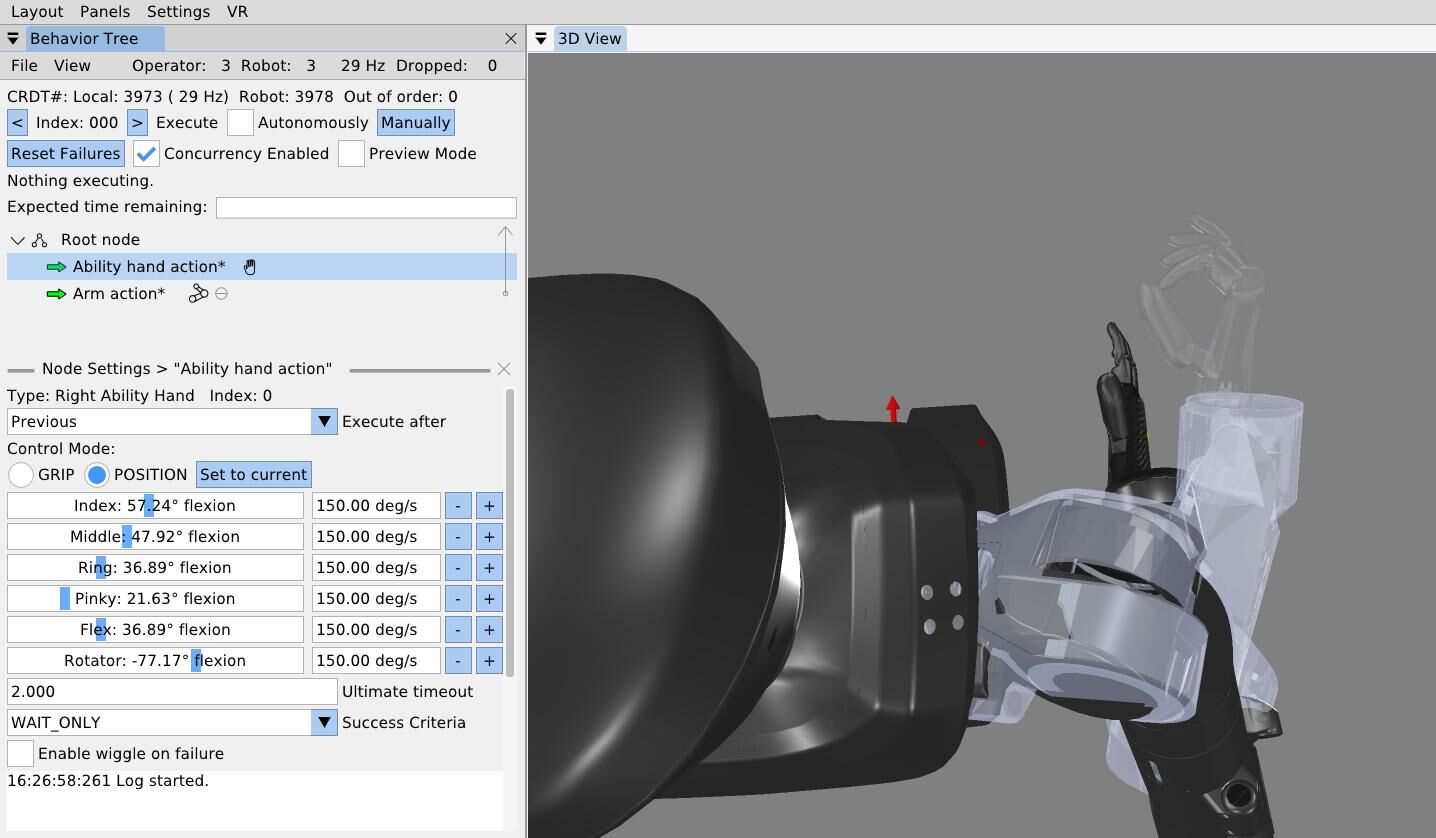}
    \caption{Specifying finger joint angles for an Ability Hand action.
    A video is available at \url{https://youtu.be/7VGFufJWaR4}.
    }
    \label{fig:guide_unitree_fingers}
\end{figure}

It is also possible to set the individual finger joint angles for an Ability Hand action, as shown in \autoref{fig:guide_unitree_fingers}.
The Ability Hand has six degrees of freedom, each with a slider and a velocity setting.
If the Ability Hand comes before an arm action and is concurrent with that arm action, it can be previewed as it is tuned, as shown in the figure.

\subsection{Arm Ready Action}
\begin{figure}[H]
    \centering
    \includegraphics[width=.95\columnwidth]{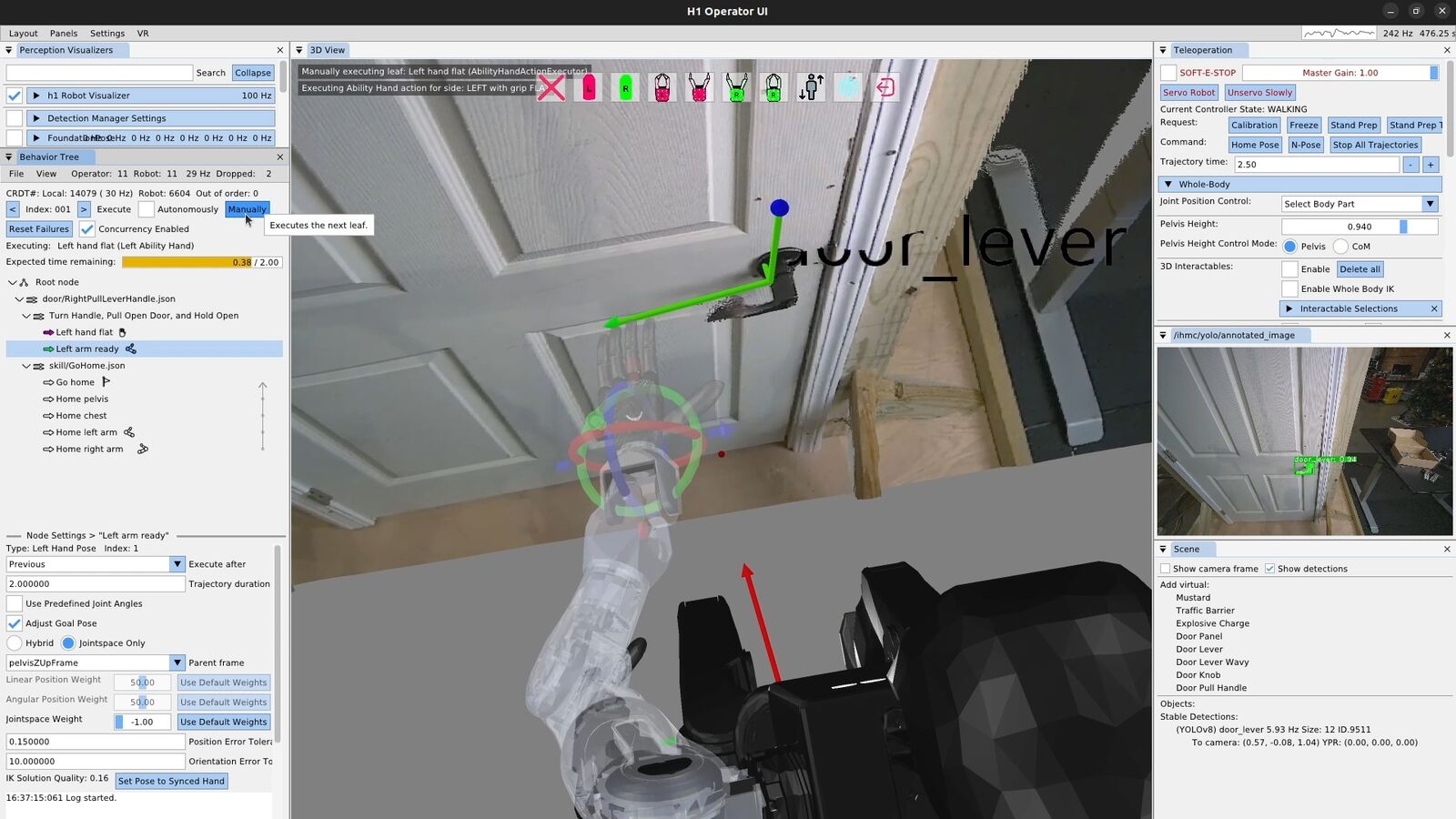}
    \caption{Authoring a ``left arm ready'' action, in preparation for pre-grasp.
    A video is available at \url{https://youtu.be/7VGFufJWaR4}.
    }
    \label{fig:guide_unitree_arm_ready}
\end{figure}

Next, we'll add a ``left arm ready'' action, as shown in \autoref{fig:guide_unitree_arm_ready}, which gets the hand and arm into a state where we can start to approach the handle from a known configuration.
Constraining the grasp approach in this way helps to robustify the behavior.
Otherwise, the hand might come from varying angles and may result in finger-handle collisions or unreliable inverse kinematics solutions.
The ready action brings the hand up from the robot's side and to roughly the orientation used for grasping the handle, but keeping a 10 cm or so distance away from the handle and any collisions.

Additionally, we typically will use joint angles or a robot-relative hand pose to define the arm ready actions, as they don't require a perceived object.
This helps us in using the action as a reset, regardless of the state of the environment.
We'll reset back to this action later when testing out the door opening components.
In this case, we define the hand pose in pelvis frame.

\begin{figure}[H]
    \centering
    \includegraphics[width=.95\columnwidth]{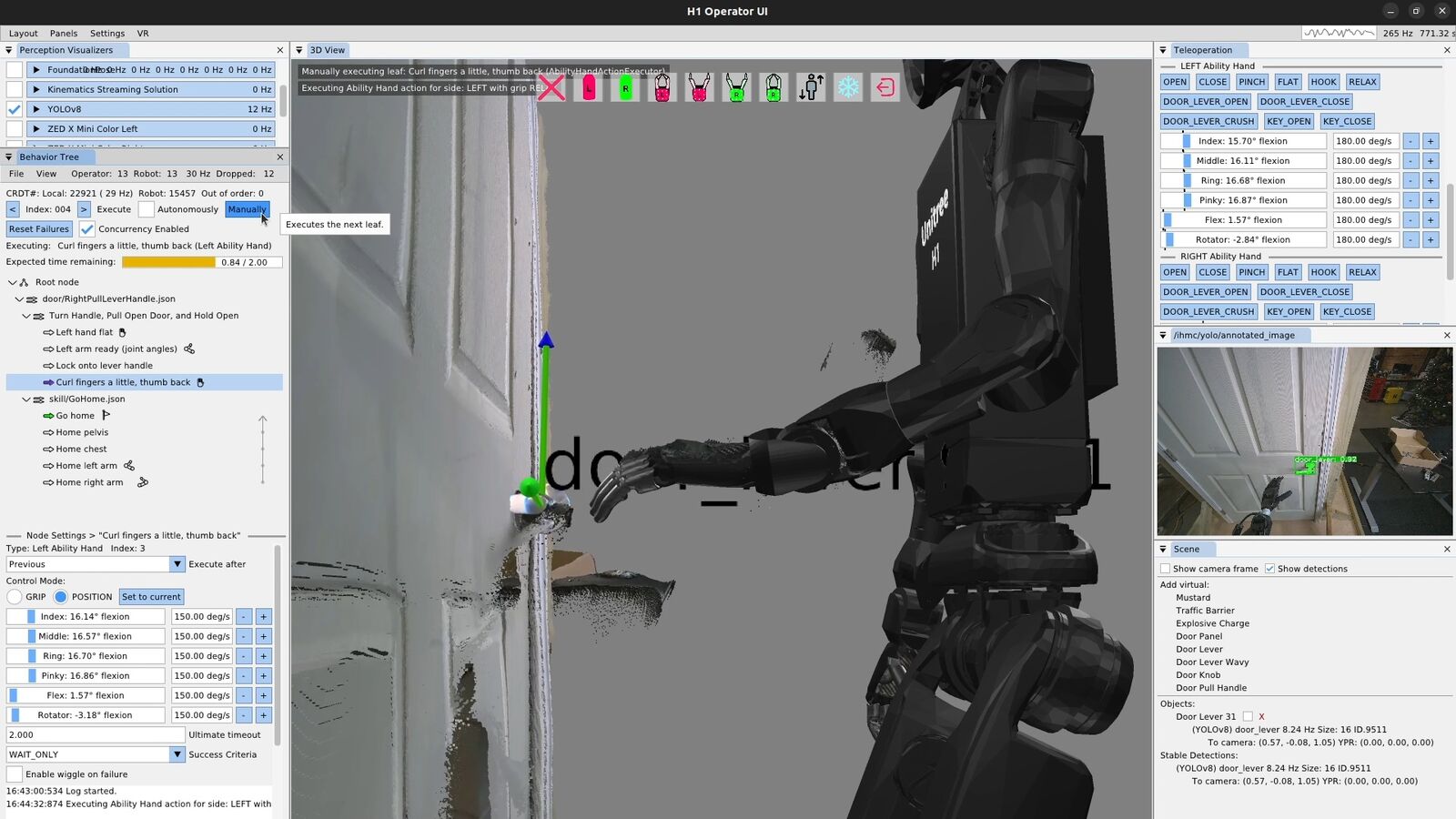}
    \caption{Locking onto the lever handle and adding a curl fingers action.
    A video is available at \url{https://youtu.be/7VGFufJWaR4}.
    }
    \label{fig:guide_unitree_lock_curl}
\end{figure}

Next, we'll create a scene action to lock on to the lever handle and another Ability Hand action to curl the fingers and pull the thumb out and back in preparation for handle contact.
In \autoref{fig:guide_unitree_lock_curl}, we've already run the scene action and have tuned the finger positions.
We used the sliders on the hand action to form our desired grasp for the lever handle.
The Ability Hand rubber is actually grippy enough that we won't need to close our fingers around the handle.
We then execute the Ability Hand action.

\subsection{Pre-Grasp Actions}
\begin{figure}[H]
    \centering
    \includegraphics[width=.95\columnwidth]{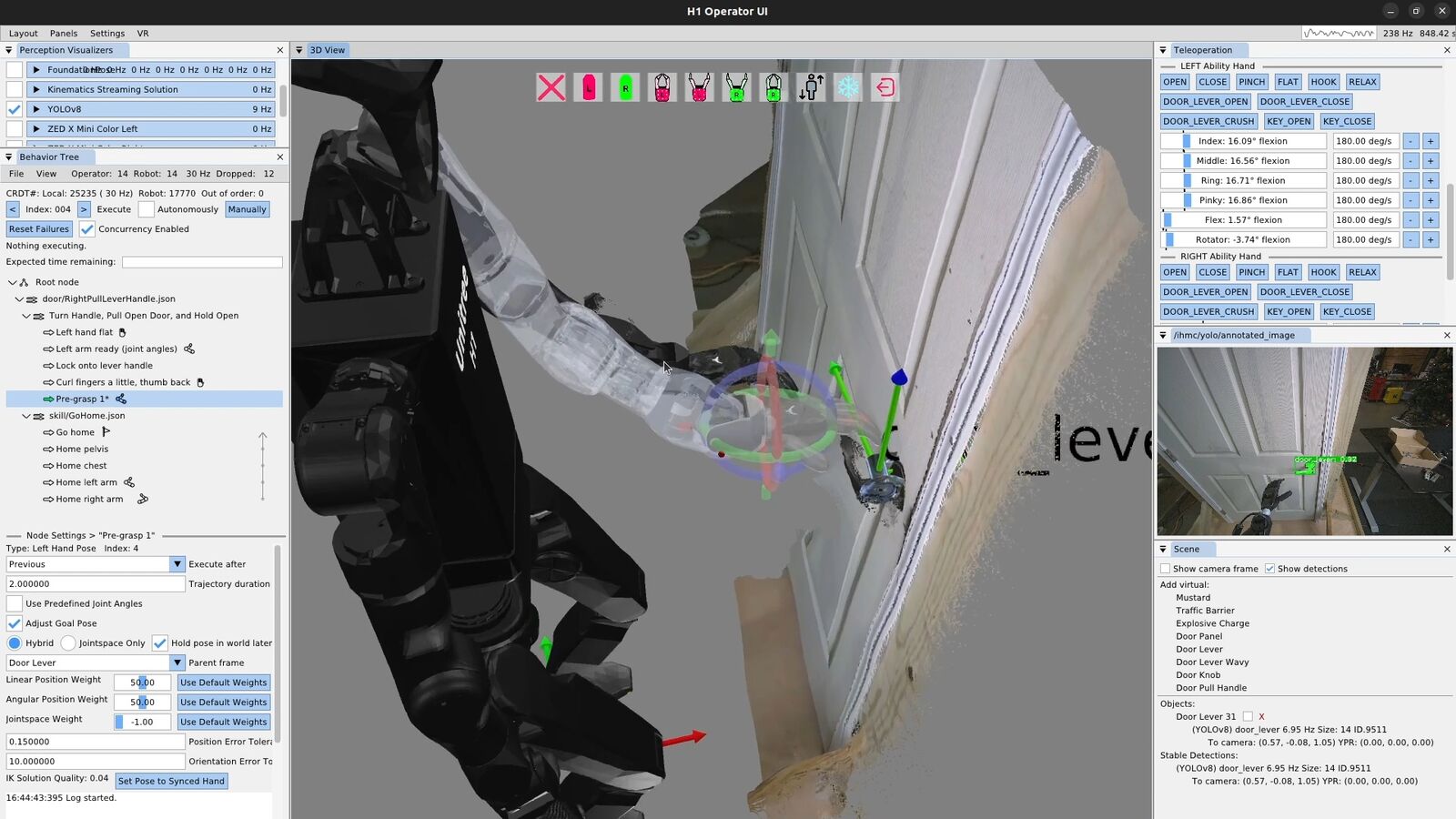}
    \caption{Authoring first pre-grasp hand pose.
    A video is available at \url{https://youtu.be/7VGFufJWaR4}.
    }
    \label{fig:guide_unitree_pregrasp1}
\end{figure}


In \autoref{fig:guide_unitree_pregrasp1}, we create our ``Pre-grasp 1'' action.
For this grasp, we will use two pre-grasp actions.
The first one hovers the hand 3-4 cm above the handle so the fingers don't hit the handle when getting into position.
Avoiding finger collisions is important with our system, as it will try hard to get to the desired position.
Given the somewhat delicate Ability Hand fingers, if you clip the fingers on things while moving the hands around, you run a real risk of breaking a finger.
Plus, it's just nice to not have unnecessary collision.

\begin{figure}[H]
    \centering
    \begin{tikzpicture}[
    node distance=2.2cm,
    font=\sffamily\small,
    action/.style={
        rectangle, rounded corners=6pt,
        draw=black!60, fill=blue!5, very thick,
        minimum width=4.2cm, minimum height=1.3cm,
        text centered, align=center
    },
    decision/.style={
        diamond, draw=black!60, fill=orange!5, very thick,
        aspect=2.2, minimum width=3.2cm,
        text centered, align=center, inner sep=3pt
    },
    terminal/.style={
        rectangle, rounded corners=8pt,
        draw=black!60, fill=green!5, very thick,
        minimum width=2.8cm, minimum height=1.0cm,
        text centered, align=center
    },
    arrow/.style={
        -{Stealth[length=3mm, width=2mm]},
        thick, black!70
    },
    label/.style={
        font=\sffamily\footnotesize\bfseries,
        text=black!80, inner sep=1pt
    }
]
    \node[action] (guess) {
        \textbf{Step 1}\\[2pt]
        Guess pose\\
        \textit{(3D pose gizmo)}
    };
    \node[action, below=of guess] (execute) {
        \textbf{Step 2}\\[2pt]
        Execute arm action
    };
    \node[decision, below=1.6cm of execute] (inspect) {
        \textbf{Step 3}\\[1pt]
        Visual\\
        inspection
    };
    \node[terminal, right=3.3cm of inspect] (done) {
        \textbf{Done}
    };
    \draw[arrow] (guess) -- (execute);
    \draw[arrow] (execute) -- (inspect);
    \draw[arrow] (inspect.west) -- ++(-1.5,0) coordinate (turn1)
    |- node[label, pos=0.25, above, sloped] {Adjust pose (gizmo)}
    (guess.west);
    \draw[arrow] (inspect) -- node[label, above] {Pose good enough} (done);
\end{tikzpicture}
    \caption{Iterative tuning loop for the first pre-grasp hand pose. The pose is guessed with the 3D gizmo, executed, inspected visually, and adjusted until it is good enough.}
    \label{fig:guide_unitree_pregrasp_tuning_loop}
\end{figure}
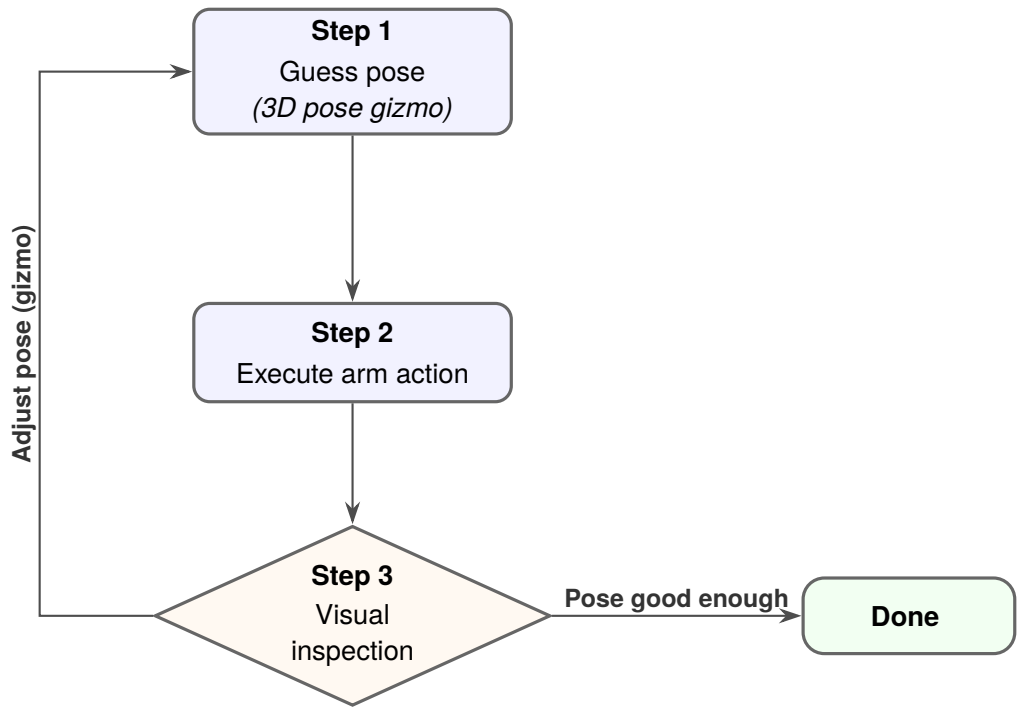

This pre-grasp action is the first time in this behavior we need to be fairly precise in the range of a centimeter or two.
It can be difficult to gauge the hand's pose with respect to the lever using the desktop monitor point cloud and first person view.
It's a little easier in VR with stereo vision.
Since we weren't using VR for this demo, we addressed the problem by sitting in line-of-sight to the robot.
The tuning process for this pre-grasp action is presented in \autoref{fig:guide_unitree_pregrasp_tuning_loop}.

\subsection{Manipulation Action Tuning Process}

\begin{figure}[H]
    \centering
    \includegraphics[width=.95\columnwidth]{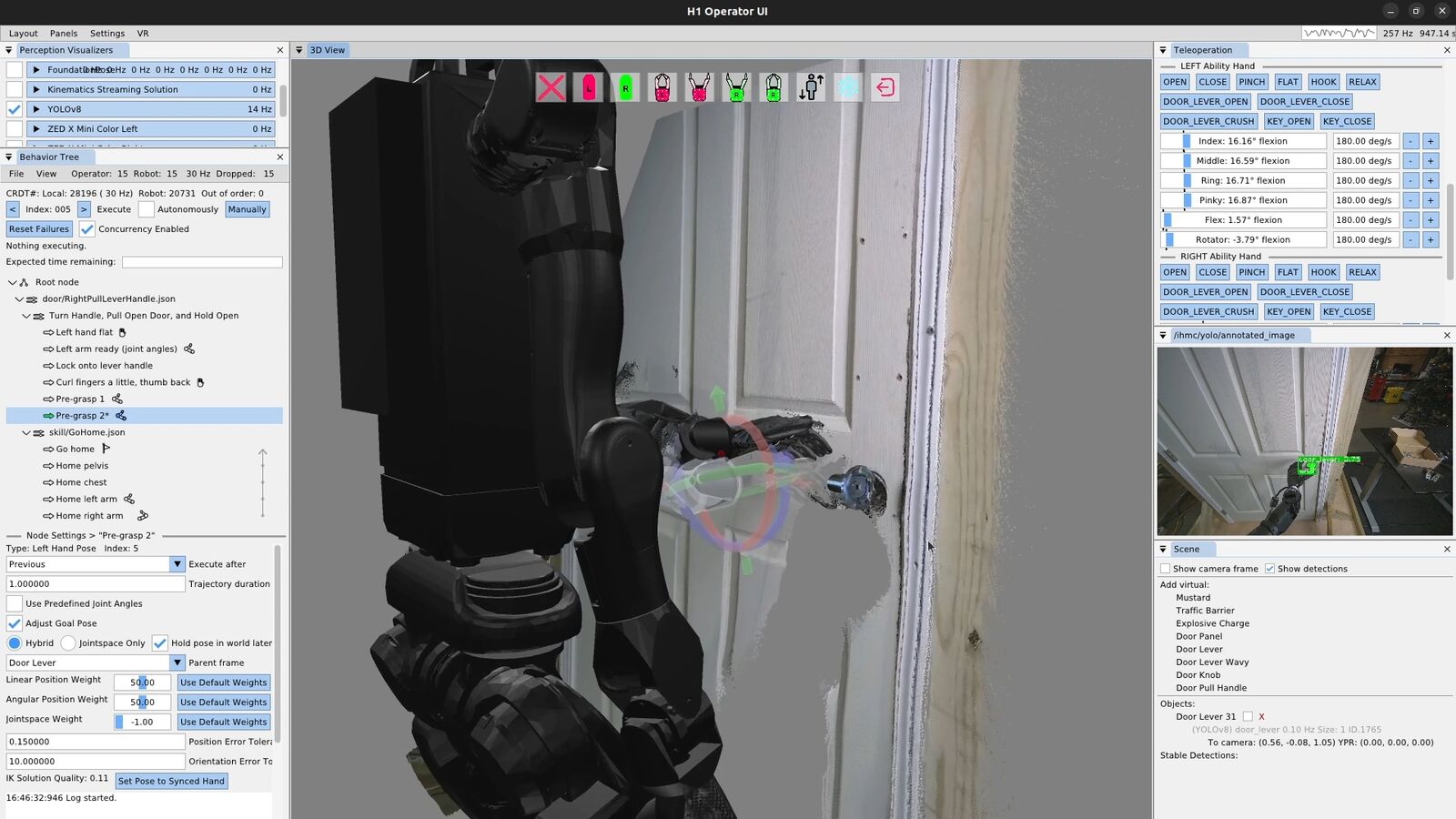}
    \caption{Authoring second pre-grasp hand pose, which contacts the handle.
    A video is available at \url{https://youtu.be/7VGFufJWaR4}.
    }
    \label{fig:guide_unitree_pregrasp2}
\end{figure}


We tune the first pre-grasp action until the only thing left to do to grasp the handle is to move the hand directly down such that it rests on the lever handle.
\autoref{fig:guide_unitree_pregrasp2} shows us authoring this action.
We repeat the visual guess-execute-inspect process for this pre-grasp action.

\subsection{Turning a Door Handle}
\begin{figure}[H]
    \centering
    \includegraphics[width=.95\columnwidth]{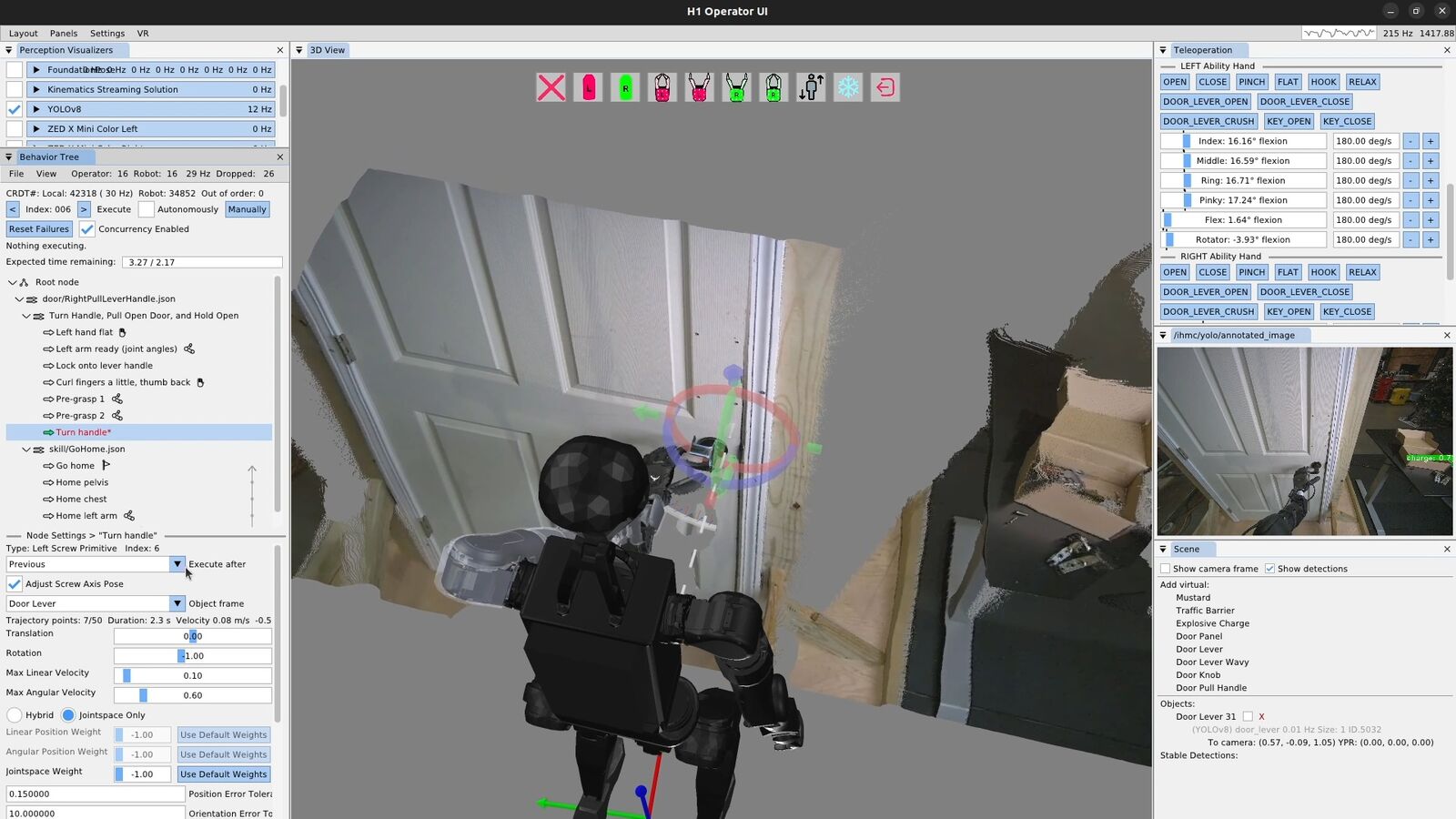}
    \caption{Authoring the handle turn screw primitive.
    A video is available at \url{https://youtu.be/7VGFufJWaR4}.
    }
    \label{fig:guide_unitree_screw_turn}
\end{figure}




Now that our hand is resting on the door lever handle, it's time to turn the lever and unlatch the door.
This is accomplished using a screw primitive action, as shown in \autoref{fig:guide_unitree_screw_turn}.
To tune a screw primitive, the first thing to do is set up the screw axis, which is the white dotted line shown in the figure.
It is moved using a pose 3D gizmo, just like the hand.
In general, the screw axis should be aligned with the rotational axis of the object you are manipulating.
In practice, we tend to have to move it some to get the desired robot motion, which is, in the end, the important part.
In addition to the axis, there are the translation and rotation amounts.
In practice, we just have to use a tuning loop like in \autoref{fig:guide_unitree_pregrasp_tuning_loop} to guess at these values and re-execute until we get the desired result.

For door lever turning we typically leave translation at zero and adjust the rotation to be more than necessary.
We also have to move the axis above and to the side opposite to the hand in order to result in a position controlled motion that achieves the necessary forces to turn the lever properly.

The iterative tuning process for a screw primitive requires an extra element for step two.
Since the screw primitive is a motion relative to the hand's current position, playing it back a second time will cause the hand to travel further and further along the helical motion profile, rather than resetting to the beginning.
For this reason, we need to reset back to the next previous arm action, in this case pre-grasp 2, in order to re-execute the door lever turn action.

Additionally, for the door lever turning action, we must make sure the handle is turned enough to unlatch the door sufficiently.
This is why we also perform a direct visual inspection of the latch as part of this tuning process.

\subsection{Opening a Door}
\begin{figure}[H]
    \centering
    \includegraphics[width=.95\columnwidth]{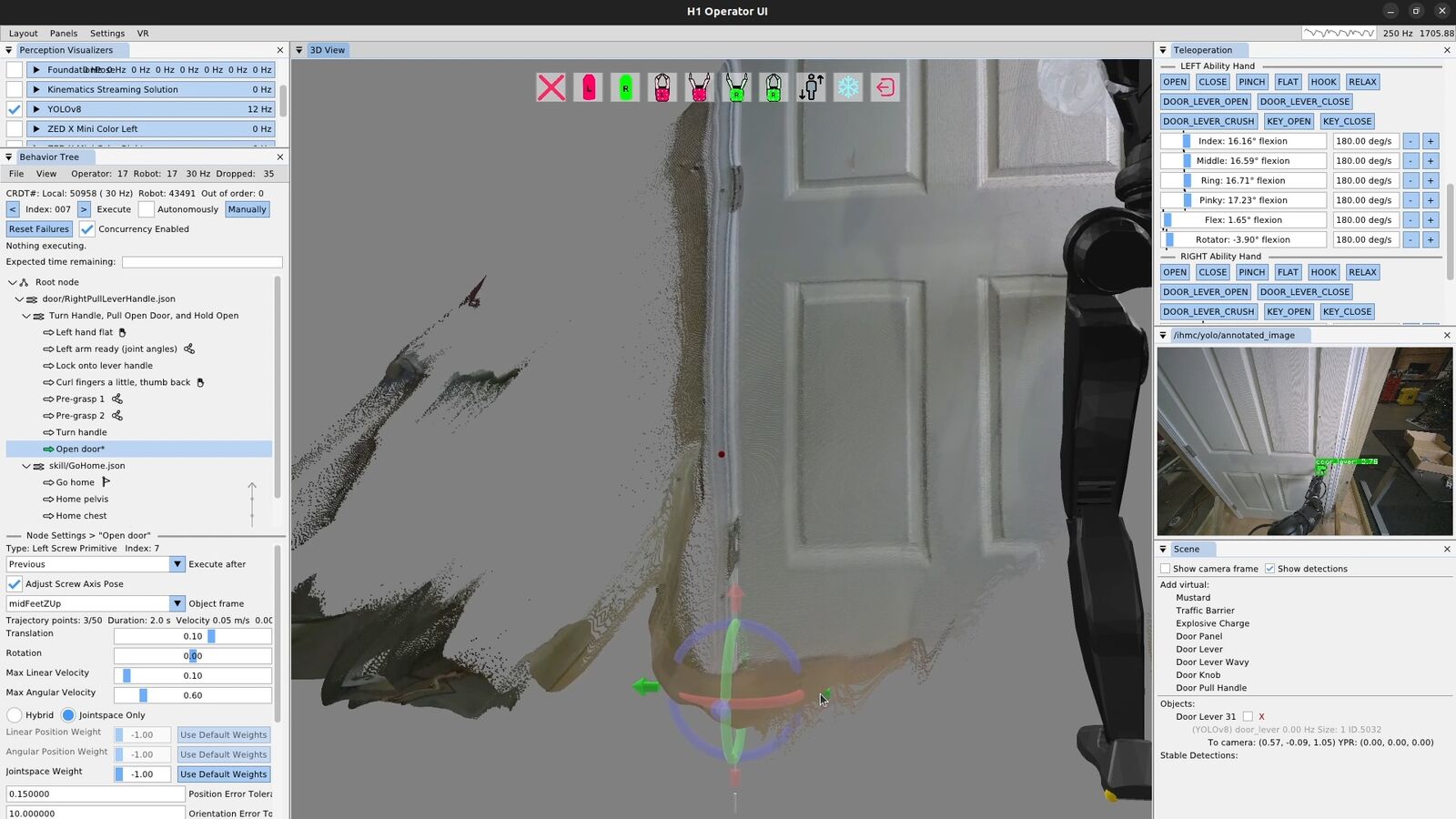}
    \caption{Authoring the open door screw primitive.
    A video is available at \url{https://youtu.be/7VGFufJWaR4}.
    }
    \label{fig:guide_unitree_screw_open}
\end{figure}


The door opening is done as another screw primitive action, but along a different axis: the door panel hinge.
Aligning the screw axis to the door panel hinge is done using the point cloud as shown in \autoref{fig:guide_unitree_screw_open}.
It doesn't have to be super precise -- within a few centimeters in the X-Y axis and 5 degrees rotationally is fine.

For door opening, the tuning process is a bit easier than the lever turning.
This is because the hand is fairly compliant to the door since it's applying a force on the handle and there is no unlatching requirement on this one.
The direction of the axis alignment will determine whether the screw rotation value for opening is positive or negative.


\subsection{Looping a Behavior}
Once we finish up the door opening action and execute it successfully, the repeated door opening behavior is pretty much done!
We set the execute after field of the pre-grasp 1 action to the scene action to curl the fingers at the same time.
Then, we add a goto action at the end that goes back to the beginning, to create an infinite loop.
We then check the execute ``Autonomously'' checkbox and let it spin.
This behavior executed 33 times in a row successfully before we stopped it.

\begin{figure}[H]
    \centering
    \includegraphics[width=.95\columnwidth]{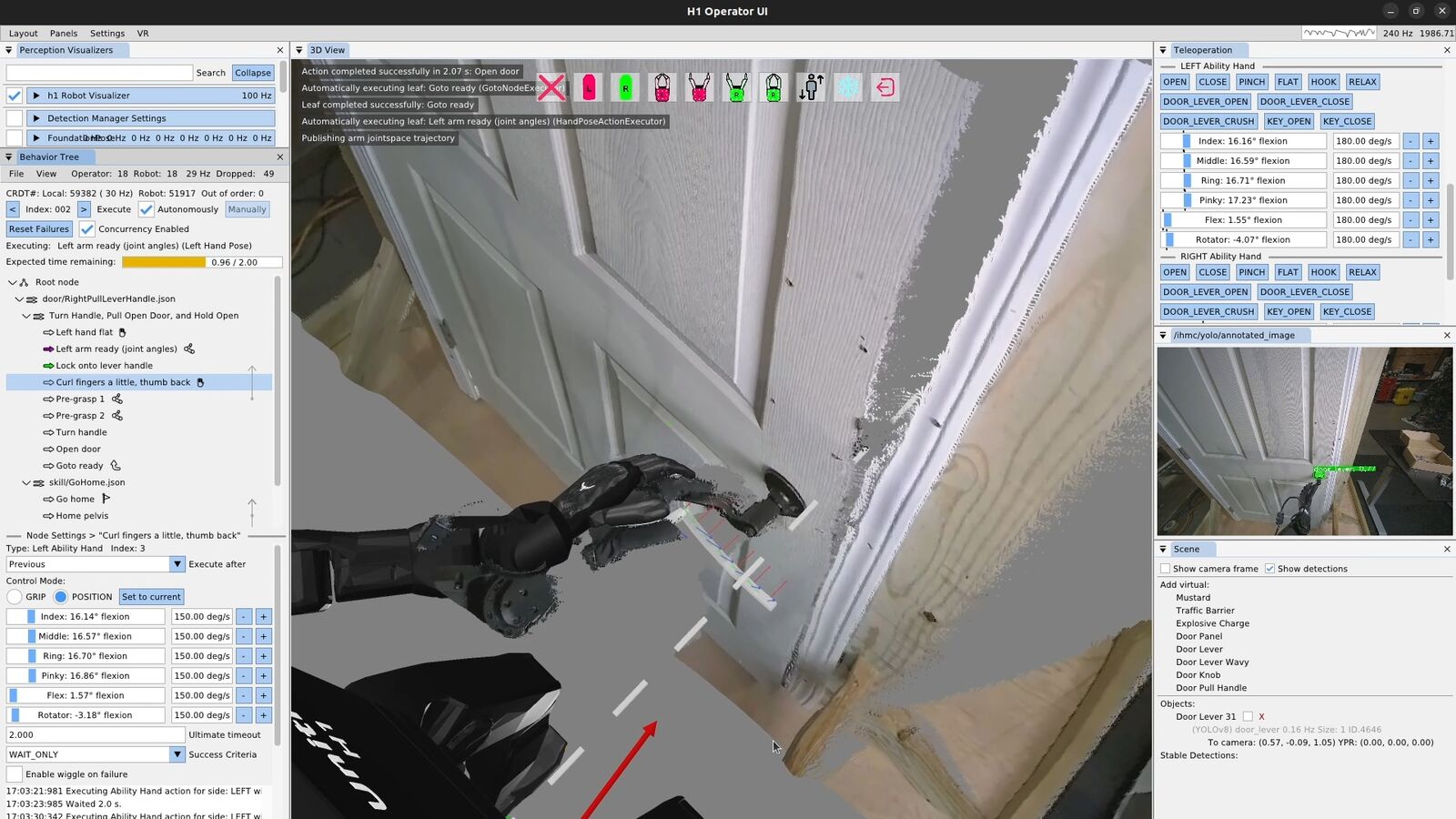}
    \caption{Running the repeated door opening behavior autonomously.
    A video is available at \url{https://youtu.be/7VGFufJWaR4}.
    }
    \label{fig:guide_unitree_done}
\end{figure}

In this guide, we have covered what all the main action types do and how to use them.
All behaviors presented in this thesis build on these basic concepts to create more complete and longer-horizon behaviors.
In \autoref{ch:defense}, we evaluate these behaviors against our hypotheses and the literature.

\chapter{Evaluation}
\label{ch:defense}

In this chapter, we'll go through our real robot demonstrations to evaluate our hypotheses, which we'll state here again:

\begin{enumerate}
    \item Robot-local execution with synchronized UI state, concurrent action layering, reactive tree logic, and behavior-time semantic perception yield door behaviors that are faster and more reliable than prior IHMC baselines and competitive with reported reinforcement learning systems on overlapping door tasks.
    \item Runtime-editable behaviors and perception modules reduce the iteration loop required to diagnose failures, modify logic, and re-test on the robot, relative to redeploy, restart, or retrain workflows.
    \item Decomposing behaviors into reusable primitives, subtrees, and scene actions allows new door and loco-manipulation variants to be brought up by editing a small part of a working behavior rather than rebuilding it from scratch.
\end{enumerate}

We'll present evidence that supports our first hypothesis with speed and reliability results for real robot tasks in \autoref{sec:h1_fast_loco_manipulation_behaviors}, our second hypothesis by showing the speed with which we can create and modify behaviors in \autoref{sec:h2_fast_behavior_authoring}, and our third hypothesis by showing that we can readily adapt existing behaviors to new tasks in \autoref{sec:h3_fast_adaptation}.
Each demonstration subsection states what that run shows for the relevant hypothesis.
We'll then do a comparative analysis against the literature to see how our results stack up in \autoref{sec:comparative_analysis}.
Finally, we'll summarize to what degree our hypotheses held in \autoref{sec:evaluation}.

\section{Hypothesis 1: Fast Loco-Manipulation Behaviors}
\label{sec:h1_fast_loco_manipulation_behaviors}

To support our first hypothesis, we'll present the demonstrations that highlight the system's capabilities and breadth while also executing with high speed relative to prior and comparative works.
This section also includes demonstrations of repeated-run reliability and resilience to external disturbances when trying to traverse doors and pick and place objects.

\subsection{Speed and Capability Evidence}

\begin{figure}[H]
    \centering
    {\singlespacing
    \resizebox{\columnwidth}{!}{%
        \input{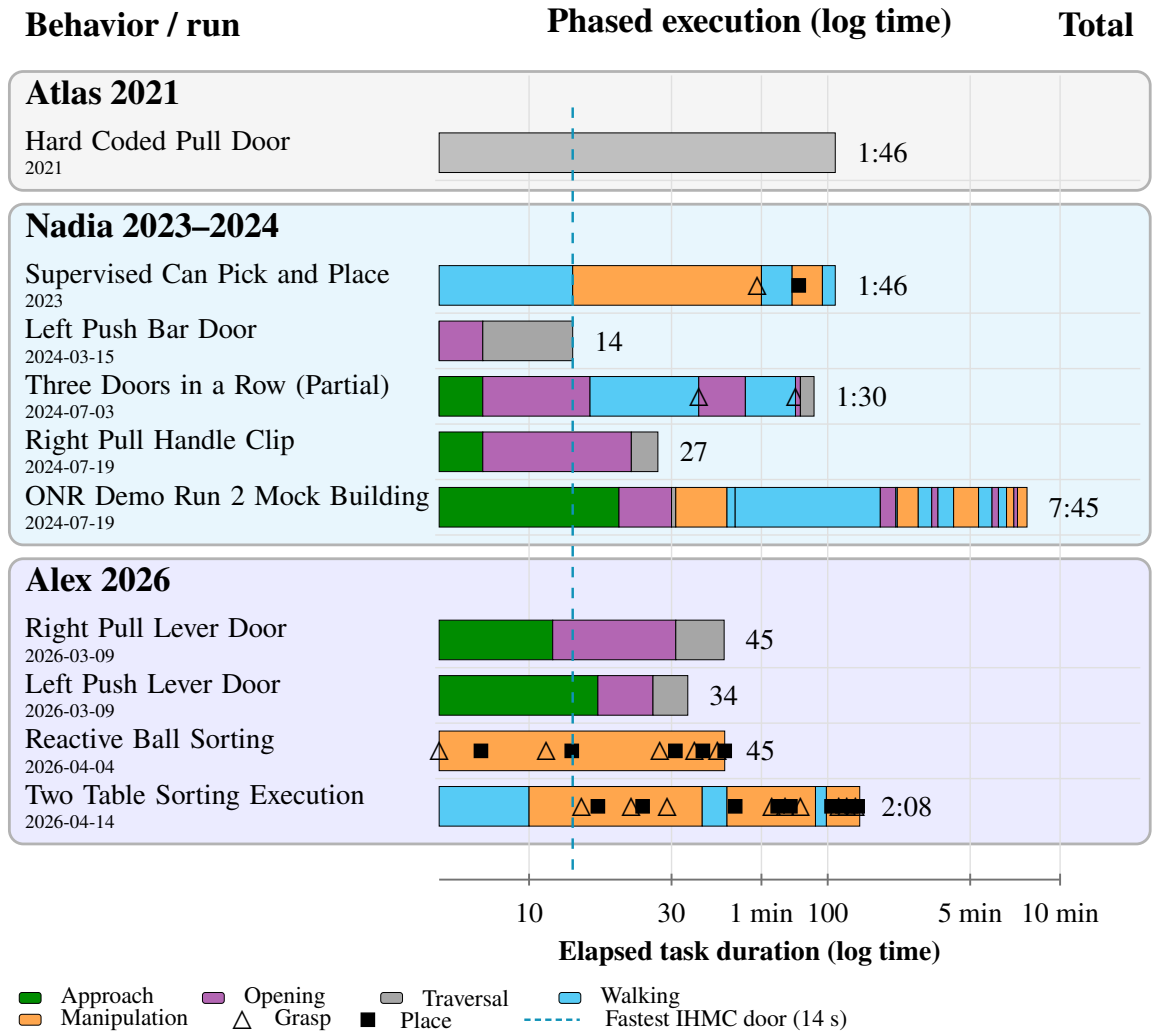}%
    }
    \par
    }
    \caption{
        In-house real robot task durations on a log-second axis with internal phase structure.
    }
    \label{fig:in_house_speed_phase_timeline}
\end{figure}

In \autoref{fig:in_house_speed_phase_timeline}, we present a representative sample of our loco-manipulation behaviors.
This figure shows that our door behaviors were very slow on Atlas, over a minute, when we had hard-coded behaviors.
Door behaviors were drastically sped up on Nadia with runtime-editable structure and concurrent action layering.
Later on Alex, we demonstrate new manipulation tasks and locomotion transitions in the same speed regime of seconds to minutes.

In the figure, the rows are color coded to show different phases of the behaviors.
For door behaviors, it shows approach, opening, and traversal phases, and for loco-manipulation behaviors it shows walking and manipulation phases.
Triangle and square markers indicate object grasps and placements.
The dashed vertical line marks the fastest IHMC door traversal on record (14~s, Nadia 2024 left push-bar).

\subsubsection{Look-and-Step Rough Terrain}

\begin{figure}[H]
    \centering
    \includegraphics[width=1.0\columnwidth]{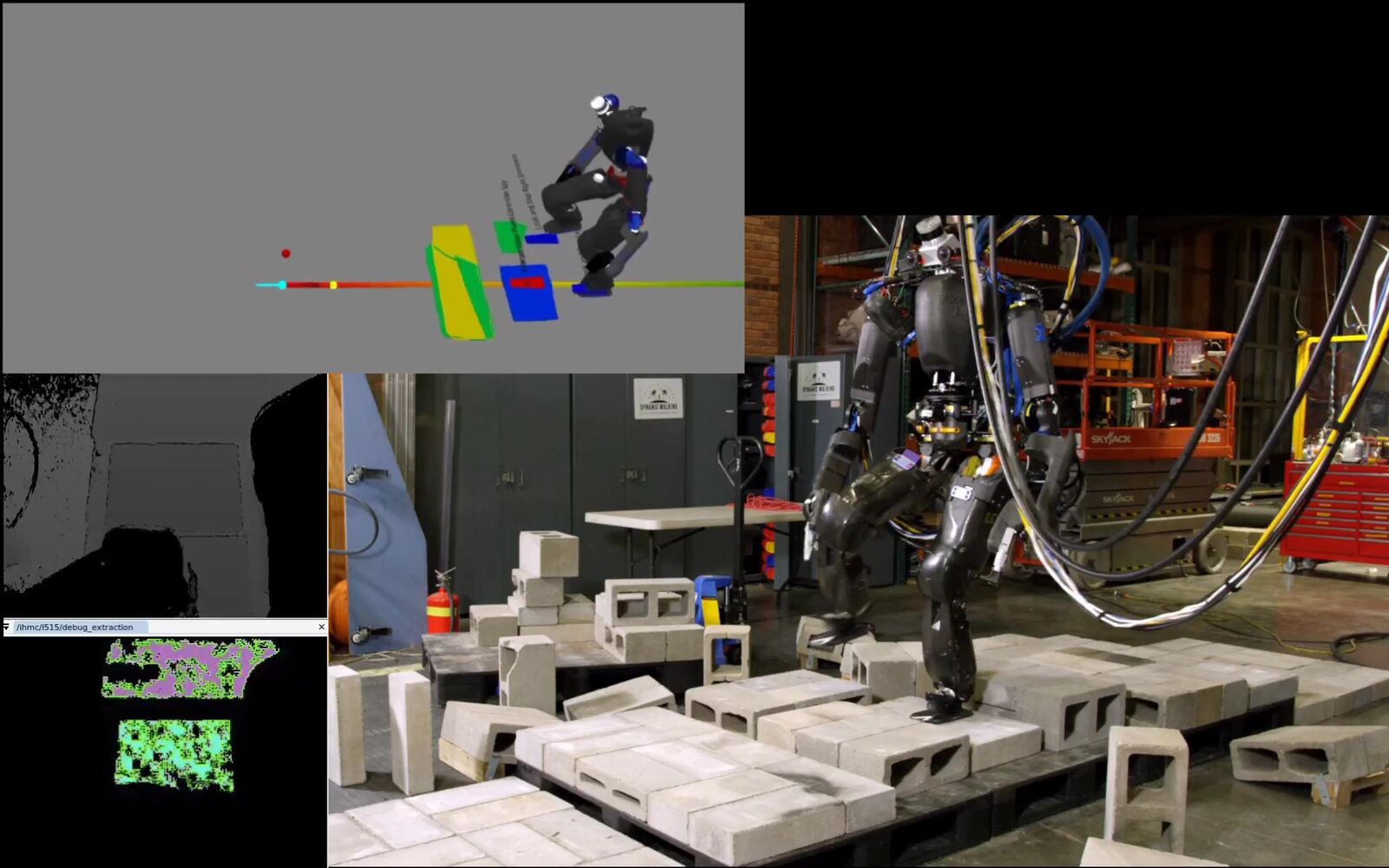}
    \caption{The Nadia humanoid performing a rough-terrain traversal with the \emph{look-and-step} behavior. The composite view shows the real robot on the terrain, the 3D footstep plan, the extracted planar regions, and the first-person planar-region segmentation view used during execution.
    A video is available at \url{https://youtu.be/nBMn1lJ57TU}.
    }
    \label{fig:look_and_step_nadia_continuity}
\end{figure}

An early speed result was in our perceptive locomotion work\cite{lookandstep}.
\autoref{fig:look_and_step_nadia_continuity} illustrates a 14-step key result, taking 14 autonomous steps in 37 seconds.
The speed of this behavior compared to prior work in the literature and an estimated human baseline is presented in \autoref{tab:rough_terrain_speed_comparison_ch4}.

\begin{table}[H]
    \caption{Rough-terrain speed comparison used in the first speed pillar claim.}
    \centering
    \begin{tabular}{l c c c}
        \hline
        System & Distance (m) & Avg. speed (m/s) & Relative speed \\
        \hline
        Fallon et al. \cite{Fallon_2015} & 5.5 & 0.023 & 1.0x \\
        Look-and-step \cite{lookandstep} & 5.3 & 0.079 & 3.4x \\
        Human baseline & - & 0.7 & 30x \\
        \hline
    \end{tabular}
    \label{tab:rough_terrain_speed_comparison_ch4}
\end{table}

This 14 step run averaged 0.079~m/s over 5.3~m (\autoref{tab:rough_terrain_speed_comparison_ch4}), about 3.4 times faster than the Fallon et al. baseline at 0.023~m/s.
This demonstration uses an earlier architecture than is tested in hypothesis 1.
It is still useful context because it shows an early focus as part of this thesis in achieving faster robot locomotion and onboard sensing.

\subsubsection{Atlas Hard-Coded Pull-Door Traversal}

\begin{figure}[H]
    \centering
    \includegraphics[width=0.95\columnwidth]{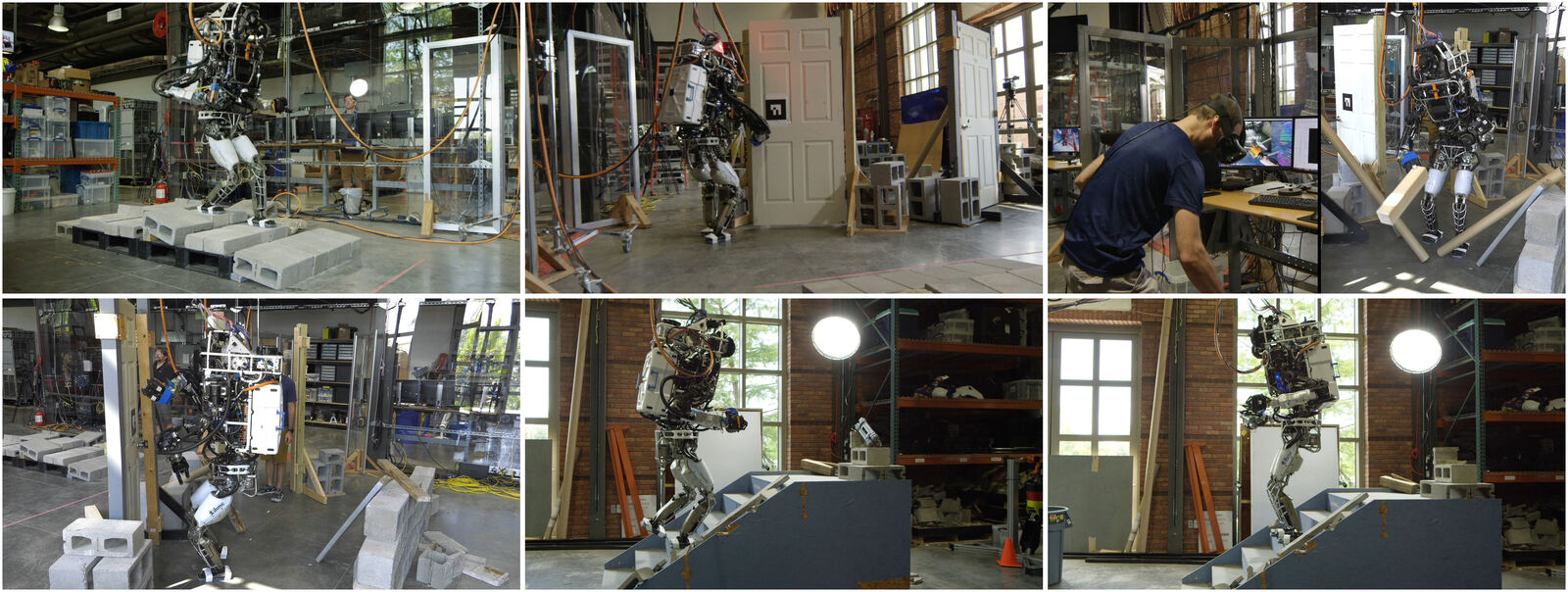}
    \caption{The June 23, 2021 building exploration demo on IHMC Atlas, including the hard-coded automatic pull-door behavior used as the 106~s speed anchor in \autoref{fig:in_house_speed_phase_timeline}.
    A video is available at \url{https://youtu.be/CFiFaO-ENPw}.}
    \label{fig:2021_building_exploration_demo_defense}
\end{figure}

Our earliest in-house door-speed anchor is the hard-coded IHMC Atlas pull-door traversal from the June 2021 building exploration era, shown in \autoref{fig:2021_building_exploration_demo_defense}.
That step-by-step behavior took 106~s from approach start to full traversal completion, with long pauses, fiducial dependence, and no robot-local authored recovery.

Compared with the July 2024 Nadia right pull handle traversal at 27~s (\autoref{tab:nadia_right_pull_handle_event_timeline}), this Atlas run is about 3.9 times slower on a pull side door task.
The later run removed fiducial dependence, used runtime authored structure, and was run on a different platform.
The comparison is not controlled, but still anchors how much faster our recent door behaviors became relative to this IHMC baseline.

\subsubsection{Supervised Can Pick-and-Place}

\begin{figure}[H]
    \centering
    \includegraphics[width=0.95\columnwidth]{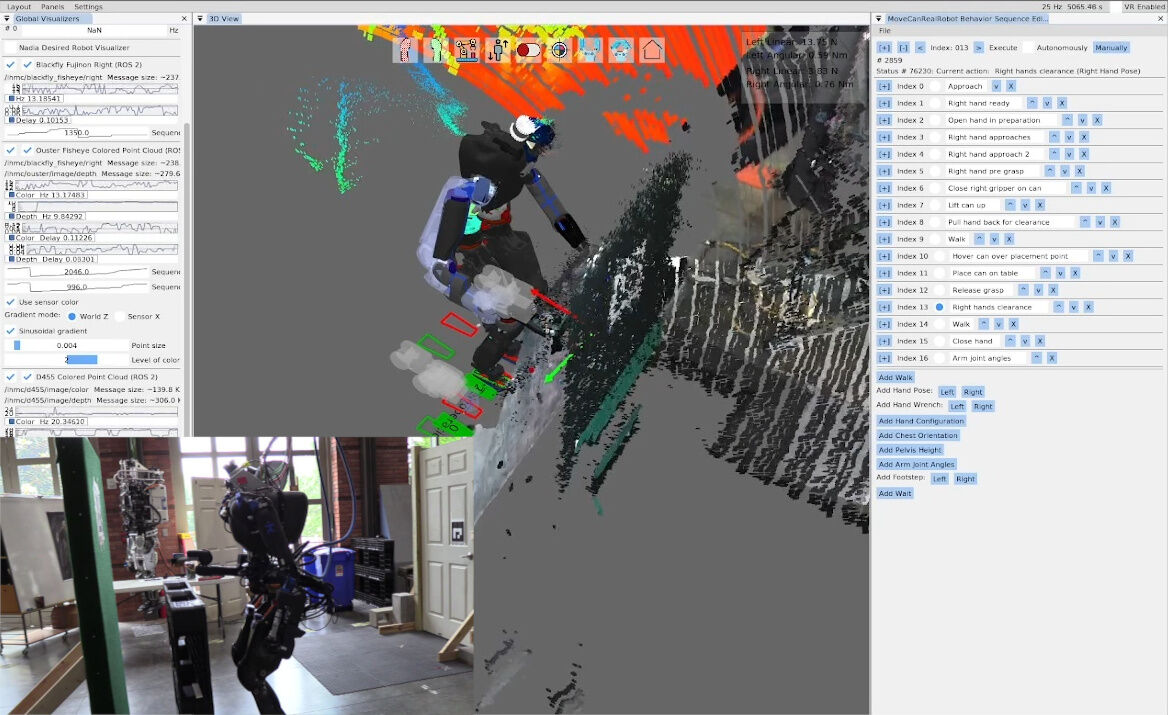}
    \caption{Nadia executing the supervised can-of-soup pick-and-place behavior on June 20, 2023.
    Videos are available at \url{https://youtu.be/V8jMvhVdP8k} and \url{https://youtu.be/ZBj8zs1wzik}.}
    \label{fig:nadia_pick_soup_defense}
\end{figure}

On June 20, 2023, we executed a supervised can-of-soup pick-and-place behavior on Nadia in 1 minute and 46 seconds, shown in \autoref{fig:nadia_pick_soup_defense}.
The run was operator-supervised, used an ArUco marker rather than direct can detection, and required manual gripper retries between actions.
Timestamps are presented in \autoref{tab:pick_and_place_execution_2023}.

\begin{table}[H]
\caption[Step-by-step supervised execution of picking and placing can of soup.]{Step-by-step supervised execution of picking and placing can of soup.}
\centering
\footnotesize
\begin{tabular}{c p{0.70\columnwidth}}
 \hline
 Time & Action completed \\
 \hline
 0:00 & Begin approach. \\
 0:11 & Approach table. \\
 0:14 & Right hand approaches can. \\
 0:53 & Pre-grasp hand pose. \\
 0:58 & Grasp can of soup. \\
 1:00 & Pull back hand with can of soup. \\
 1:16 & Step to the side. \\
 1:20 & Set down can. \\
 1:36 & Release grasp on can. \\
 1:46 & Back away from task. \\
 \hline
\end{tabular}
\label{tab:pick_and_place_execution_2023}
\end{table}

The 1 minute 46 second run shows an early manipulation capability on Nadia, but much of the timeline is operator supervised tuning rather than autonomous execution (\autoref{tab:pick_and_place_execution_2023}).
ArUco marker dependence and operator supervision make this demonstration an early speed reference rather than a testable result for hypothesis 1.
It does show that the architecture could, albeit with great difficulty, execute pick and place on real hardware before YOLO based scene authoring was in place.

\subsubsection{14~s Nadia Left Push-Bar Traversal}

\begin{figure}[H]
    \centering
        \includegraphics[width=0.74\columnwidth,height=0.32\textheight,keepaspectratio]{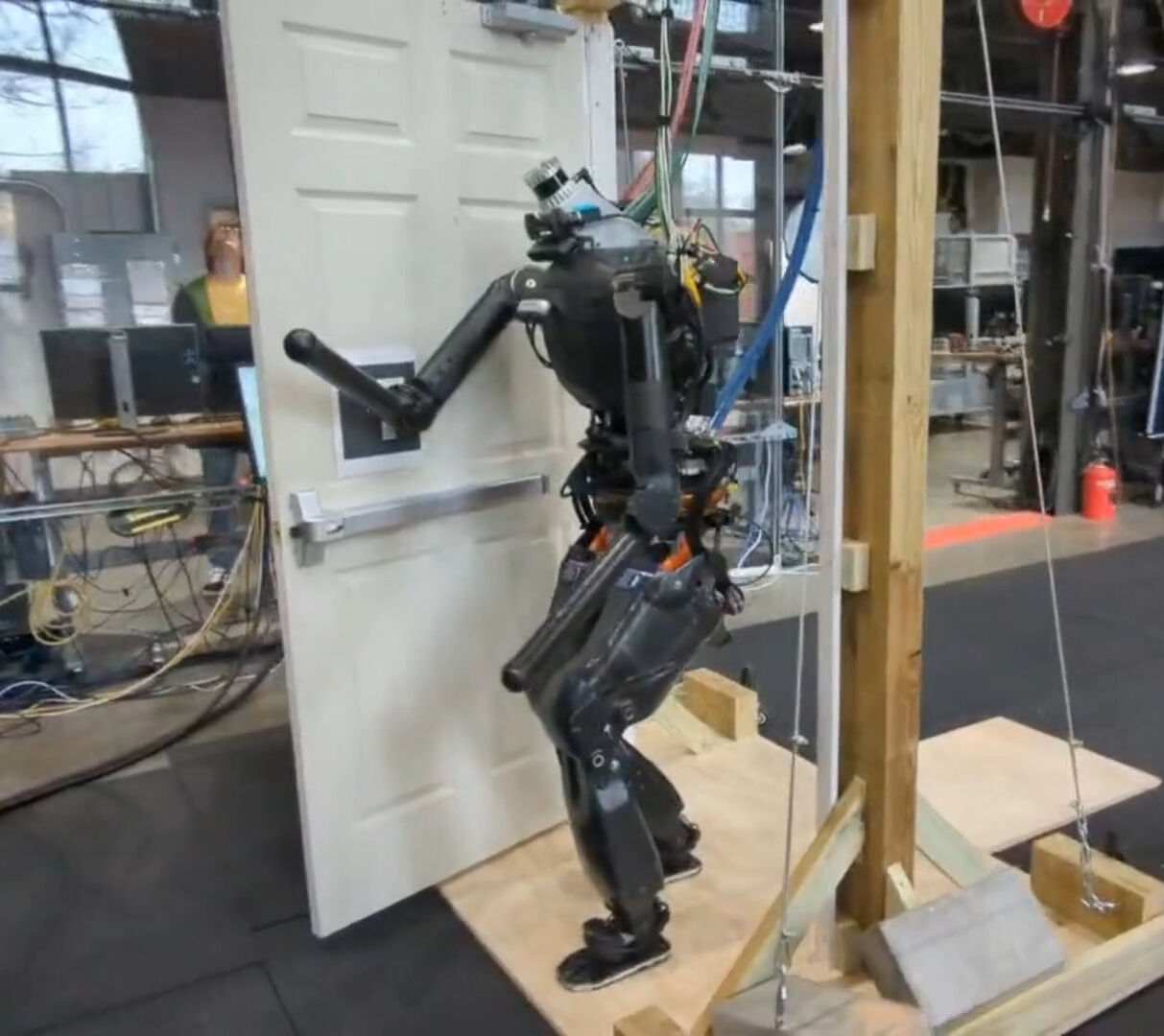}
    \caption{March~15,~2024 Nadia left push-bar traversal on a door with a closer.
    This continuous-walking run executed in 14~s, our fastest door traversal on record.
    A video is available at \url{https://youtu.be/DrR2Ng3ft5Y}.}
    \label{fig:20240315_NadiaLeftPushBarWithCloserContinuousWalking}
\end{figure}

Our fastest door traversal ever was on March 15, 2024, and is presented in \autoref{fig:20240315_NadiaLeftPushBarWithCloserContinuousWalking}.
The robot walked continuously during this run, using concurrent arm actions to open the push bar door during the traversal.
Timestamps are presented in \autoref{tab:nadia_left_push_bar_event_timeline}.

\begin{table}[H]
\caption{Event times for the March~15,~2024 Nadia left push-bar traversal shown in \autoref{fig:20240315_NadiaLeftPushBarWithCloserContinuousWalking}.}
\centering
\begin{tabular}{c p{0.72\columnwidth}}
 \hline
 Time & Event \\
 \hline
 0:00 & Approach begins. \\
 0:04 & Push bar is unlatched. \\
 0:07 & Shoulders square with the door frame. \\
 0:14 & Traversal is complete. \\
 \hline
\end{tabular}
\label{tab:nadia_left_push_bar_event_timeline}
\end{table}

The robot unlatched the push bar at 4~s and cleared the doorway at 14~s (\autoref{tab:nadia_left_push_bar_event_timeline}).
Continuous walking with concurrent arm actions produced our fastest door traversal on record and is marked by the dashed line in \autoref{fig:in_house_speed_phase_timeline}.
Relative to the 106~s Atlas anchor, this run is direct evidence that concurrent action layering removed long execution pauses.

\subsubsection{Three Doors in a Row}

\begin{figure}[H]
    \centering
    \includegraphics[width=0.95\columnwidth]{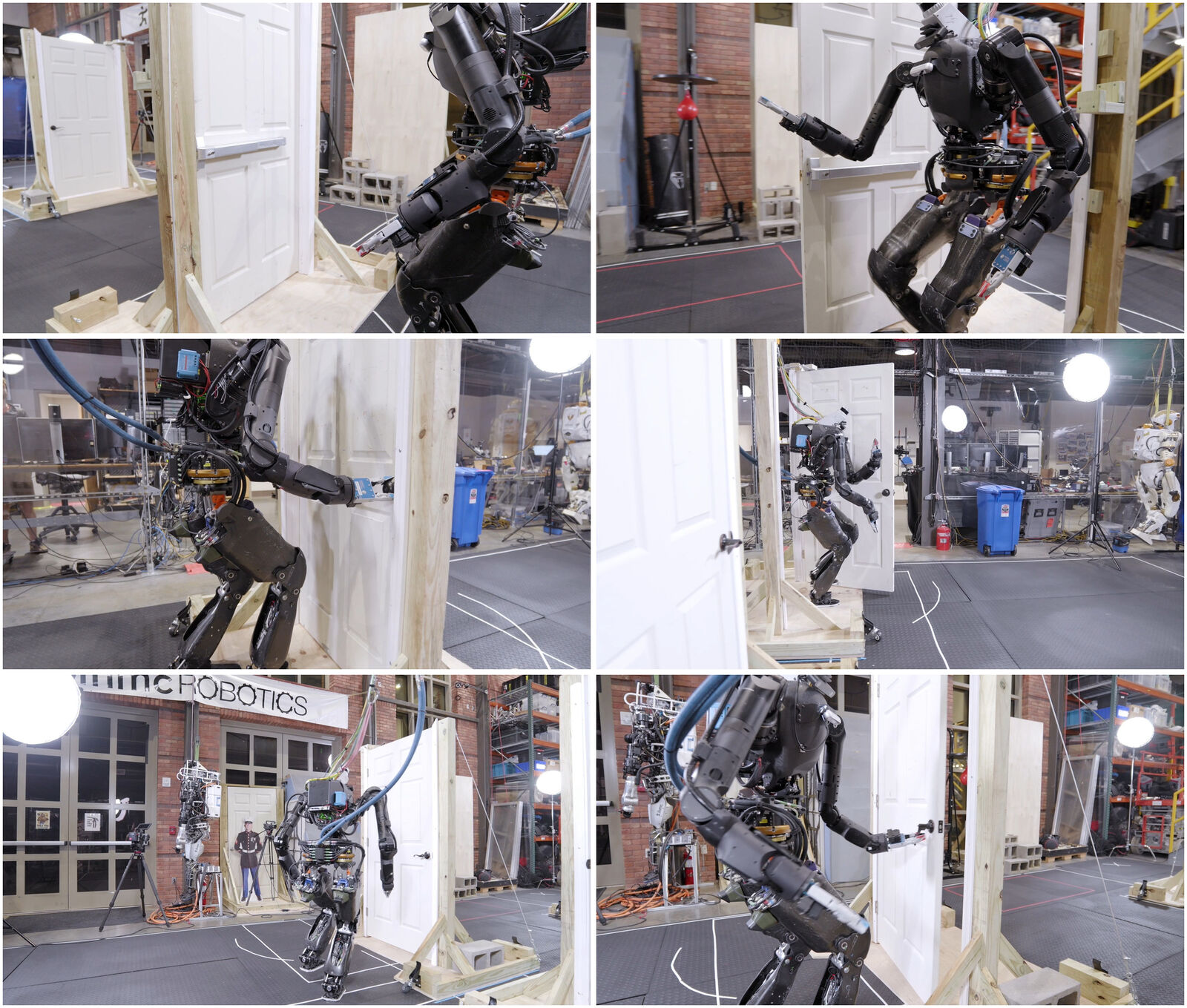}
    \caption{The July 3, 2024 Nadia run attempting three consecutive door traversals with continuous autonomy.
    The robot cleared the first two doors but fell after opening the third.
    A video is available at \url{https://youtu.be/cd-lo-l7pPI}.}
    \label{fig:2024_3_doors_defense}
\end{figure}

On July 3, 2024, we attempted to traverse three lab doors consecutively in one continuous run, shown in \autoref{fig:2024_3_doors_defense}.
The robot cleared the first two doors but fell during traversal of the third; the timed run ended at 1 minute and 30 seconds.
Timestamps are presented in \autoref{tab:three_doors_event_timeline}.

\begin{table}[H]
\caption{Event times for the July~3,~2024 Nadia three-door run shown in \autoref{fig:2024_3_doors_defense}.}
\centering
\footnotesize
\begin{tabular}{c p{0.72\columnwidth}}
 \hline
 Time & Event \\
 \hline
 0:02 & Start approaching door 1 (left push bar with spring closer). \\
 0:07 & Start pushing door 1 open. \\
 0:16 & Clear door 1; start approaching door 2. \\
 0:37 & Start grasping door 2 knob (right push knob door). \\
 0:42 & Open door 2. \\
 0:53 & Clear door 2; start approaching door 3. \\
 1:18 & Start grasping door 3 lever handle (right pull lever handle door). \\
 1:21 & Open door 3. \\
 1:30 & Fall during door 3 traversal. \\
 \hline
\end{tabular}
\label{tab:three_doors_event_timeline}
\end{table}

The robot cleared the first door at 16~s and the second at 53~s, then fell during the third traversal at 1 minute 30~s (\autoref{tab:three_doors_event_timeline}).
The run still supports capability breadth across push bar, push knob, and pull handle doors in one continuous behavior.
The fall shows that full traversal reliability remained a weak point, which we talk about in the reliability results below.

\subsubsection{Nadia Right Pull-Handle Traversal}

\begin{figure}[H]
    \centering
        \includegraphics[width=0.66\columnwidth,height=0.32\textheight,keepaspectratio]{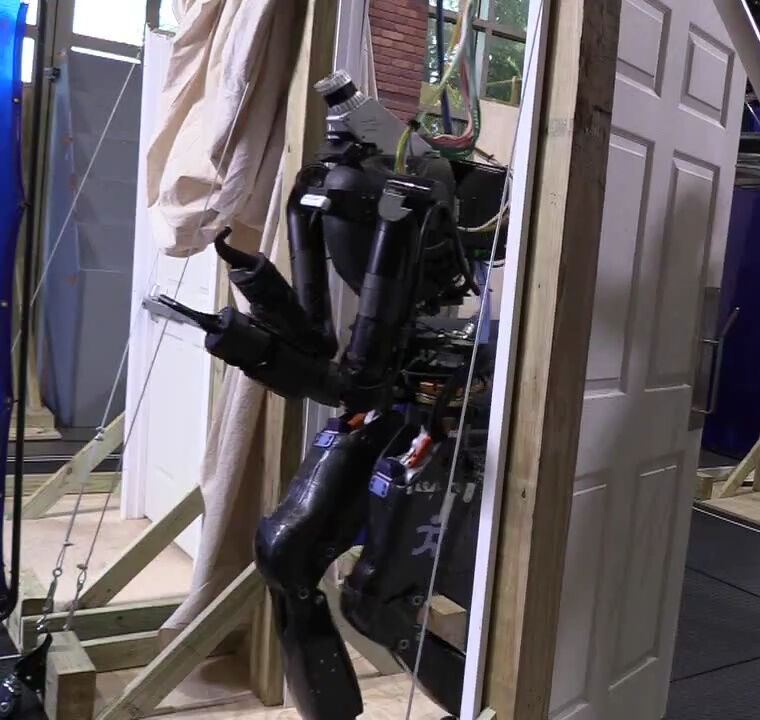}
    \caption[July~19,~2024 Nadia hook-hands right pull-handle traversal.]{July~19,~2024 Nadia right pull-handle traversal with hook hands.
    This representative pull-side run completed in 27~s from approach start.
    A video of the full mock-building demo run is available at \url{https://youtu.be/D0gylAJEdZw}.}
    \label{fig:20240719_ONRDemoRun2PullHandleDoorNadia}
\end{figure}

On July 19, 2024, Nadia completed a representative right pull-handle door traversal with hook hands in 27 seconds, shown in \autoref{fig:20240719_ONRDemoRun2PullHandleDoorNadia}.
Timestamps are presented in \autoref{tab:nadia_right_pull_handle_event_timeline}.

\begin{table}[H]
\caption{Event times for the July~19,~2024 Nadia hook-hands right pull-handle traversal shown in \autoref{fig:20240719_ONRDemoRun2PullHandleDoorNadia}.}
\centering
\begin{tabular}{c p{0.72\columnwidth}}
 \hline
 Time & Event \\
 \hline
 0:00 & Approach begins. \\
 0:07 & Door is opened. \\
 0:22 & Shoulders square with the door frame. \\
 0:27 & Traversal is complete. \\
 \hline
\end{tabular}
\label{tab:nadia_right_pull_handle_event_timeline}
\end{table}

The 27~s total matches the pull lever traversal measurement in \autoref{fig:in_house_speed_phase_timeline}.
Door opening finished at 7~s, but panel clearance kept the robot from getting through the frame until 22~s (\autoref{tab:nadia_right_pull_handle_event_timeline}).
\autoref{fig:DoorTraversalTimingsSpeed} shows the same push versus pull timing pattern across Nadia runs.

\subsubsection{ONR Mock-Building Demo Run 2}

\begin{figure}[H]
    \centering
    \includegraphics[width=\columnwidth]{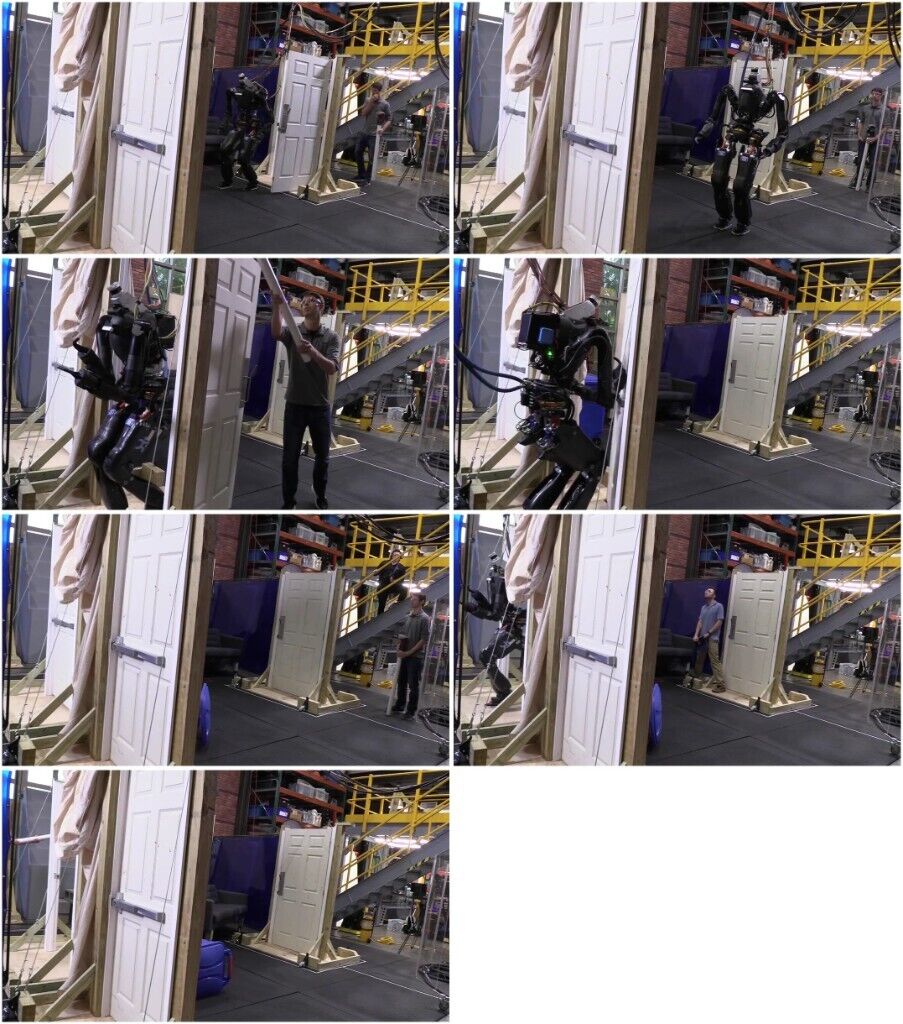}
    \caption{July~19,~2024 Nadia ONR mock-building demo run~2 collage.
    The run entered through a push-bar door, searched multiple rooms, cleared a blocked doorway, traversed additional doors, moved furniture to recover a hidden object, and ended with a salute behavior in 7 minutes and 45 seconds.
    A video is available at \url{https://youtu.be/D0gylAJEdZw}.}
    \label{fig:20240719_ONRDemoRun2}
\end{figure}

On July 19, 2024, we executed the second timed full run of our ONR mock-building search demo in 7 minutes and 45 seconds, shown in \autoref{fig:20240719_ONRDemoRun2}.
This run searched three rooms, cleared a blocked doorway, moved furniture, and ended with a salute behavior.
Timestamps are presented in \autoref{tab:onr_demo_run_2_event_timeline}.

\begin{table}[H]
\caption{Event times for the July~19,~2024 Nadia ONR mock-building demo run~2.}
\centering
\footnotesize
\setlength{\tabcolsep}{3pt}
\renewcommand{\arraystretch}{0.98}
\begin{tabular}{c p{0.72\columnwidth}}
 \hline
 Time & Event \\
 \hline
 0:00 & Robot starts from outside first room. \\
 0:20 & Starts pushing push-bar door. \\
 0:30 & Clears right push-bar door. \\
 0:31 & Starts room search; standing in place. \\
 0:46 & Finishes room search. \\
 0:49 & Begins approach to door~2 (right pull handle). \\
 2:30 & Starts pulling open door~2. \\
 2:49 & Clears walking through door~2. \\
 2:51 & Begins searching room~2; standing in place. \\
 3:21 & Search ended; object not found; leaving room. \\
 3:43 & Starts pushing door~2 (left push bar) open to go back through. \\
 3:54 & Cleared door~2; back in main room. \\
 4:24 & After walking to center of room; search again. \\
 5:20 & Clears recycling bin blocking door~3. \\
 5:55 & Starts pulling open door~3 (right pull door). \\
 6:13 & Clears door~3. \\
 6:37 & Searches room~3; object not found; turn around. \\
 7:00 & Starts pushing door~3 (left push bar) open to go back through. \\
 7:12 & Clears door~3. \\
 7:31 & Moves couch out of the way; finds object. \\
 7:45 & Salute behavior; end of demo. \\
 \hline
\end{tabular}
\label{tab:onr_demo_run_2_event_timeline}
\end{table}

The 7 minute 45 second run combined three room searches, five door passages, furniture manipulation, and a blocked doorway recovery (\autoref{tab:onr_demo_run_2_event_timeline}).
Much of the elapsed time went to searching and repositioning between doors, not to isolated door traversals.
The standalone 27~s pull handle run from the same day (\autoref{tab:nadia_right_pull_handle_event_timeline}) shows that individual door phases could be fast, but this chained demo mainly demonstrates breadth across many authored behaviors in one tree.

\subsubsection{Alex Right Pull Lever Traversal}

\begin{figure}[H]
    \centering
        \includegraphics[width=1.0\columnwidth]{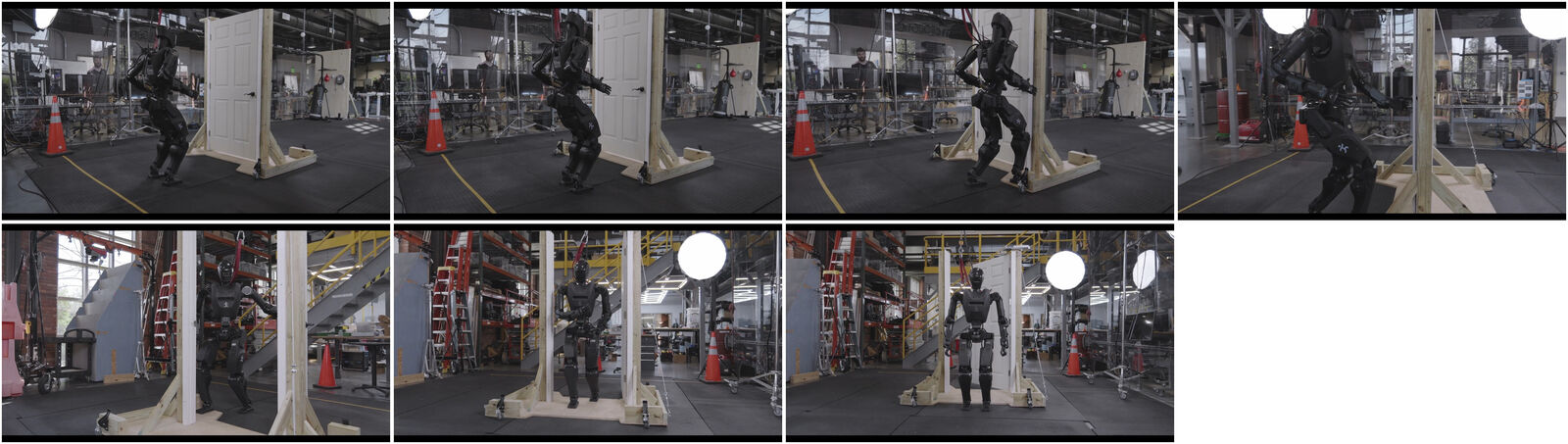}
    \caption[Current Alex right pull lever-handle traversal key frames.]{Current Alex right pull lever-handle traversal key frames from the March~9,~2026 run.
    From left to right and top to bottom: coarse approach begins at 0:02, fine approach begins at 0:09, opening stance is achieved at 0:14, the door is unlatched at 0:18, the door is opened and traversal begins at 0:33, the shoulders align with the frame at 0:42, and traversal completes at 0:47.
    A video is available at \url{https://youtu.be/fP-9DfGFvW8}.}
    \label{fig:AlexRightPullDoorProfessional}
\end{figure}

On March 9, 2026, Alex completed a right pull lever-handle door traversal in 45 seconds, shown in \autoref{fig:AlexRightPullDoorProfessional}.
Timestamps are presented in \autoref{tab:alex_right_pull_event_timeline}.

\begin{table}[H]
\caption{Timestamped events for the March~9,~2026 Alex right pull lever-handle traversal shown in \autoref{fig:AlexRightPullDoorProfessional}.}
\centering
\begin{tabular}{c p{0.72\columnwidth}}
 \hline
 Time & Event \\
 \hline
 0:02 & Coarse approach begins. \\
 0:09 & Fine approach begins. \\
 0:14 & Opening stance is achieved. \\
 0:18 & Door unlatching is completed. \\
 0:33 & Door opening is sufficient and traversal begins. \\
 0:42 & Shoulders align with the door frame. \\
 0:47 & Traversal is complete. \\
 \hline
\end{tabular}
\label{tab:alex_right_pull_event_timeline}
\end{table}

\begin{table}[H]
\caption{Per-phase timing breakdown for the March~9,~2026 Alex right pull lever-handle traversal.}
\centering
\setlength{\tabcolsep}{4pt}
\renewcommand{\arraystretch}{1.01}
\begin{tabular}{p{2.1cm} c c p{0.41\columnwidth}}
 \hline
 Phase & Time window & Duration & Description \\
 \hline
 Approach & 0:02--0:14 & 12~s & Coarse approach from the distant start, followed by the fine approach into the authored opening stance. \\
 Unlatch & 0:14--0:18 & 4~s & Final upper-body alignment, handle engagement, and latch release. \\
 Panel motion & 0:18--0:33 & 15~s & Pull the panel clear while staying outside the door swing. \\
 Traverse & 0:33--0:47 & 14~s & Commit through the frame, bring the shoulders into alignment by 0:42, and clear the doorway. \\
 \hline
\end{tabular}
\label{tab:alex_right_pull_phase_breakdown}
\end{table}

This 45~s run breaks down into 12~s approach, 4~s unlatch, 15~s panel motion, and 14~s traverse (\autoref{tab:alex_right_pull_phase_breakdown}).
Panel motion is the longest phase on pull doors because the robot must stay clear of the swing.
This timing places Alex pull traversals in the same tens of seconds band as our 2024 Nadia results and well below the minute scale classical literature entries in \autoref{fig:hero_speed_comparison}.

\subsubsection{Alex Left Push Lever Traversal}

\begin{figure}[H]
    \centering
        \includegraphics[width=1.0\columnwidth]{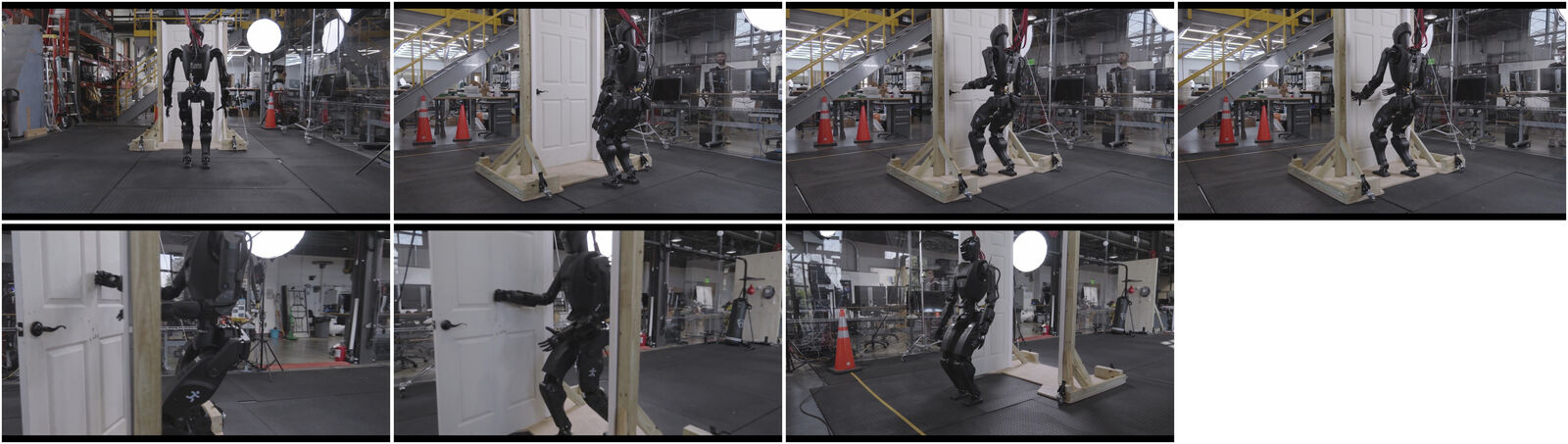}
    \caption[Current Alex left push lever-handle traversal key frames.]{Current Alex left push lever-handle traversal key frames from the March~9,~2026 run.
    From left to right and top to bottom: coarse approach begins at 0:03, fine approach begins at 0:12, opening stance is achieved at 0:17, the lever is unlatched at 0:20, the door is open and traversal begins at 0:26, the shoulders align with the frame at 0:27, and traversal completes at 0:34.
    A video is available at \url{https://youtu.be/rKOUzo2sZ70}.}
    \label{fig:AlexLeftPushDoorProfessional}
\end{figure}

On March 9, 2026, Alex completed a left push lever-handle door traversal in 34 seconds, shown in \autoref{fig:AlexLeftPushDoorProfessional}.
Timestamps are presented in \autoref{tab:alex_left_push_event_timeline}.

\begin{table}[H]
\caption{Timestamped events for the March~9,~2026 Alex left push lever-handle traversal shown in \autoref{fig:AlexLeftPushDoorProfessional}.}
\centering
\begin{tabular}{c p{0.72\columnwidth}}
 \hline
 Time & Event \\
 \hline
 0:03 & Coarse approach begins. \\
 0:12 & Fine approach begins. \\
 0:17 & Opening stance is achieved. \\
 0:20 & Door lever unlatching is completed. \\
 0:26 & Door opening is sufficient and traversal begins. \\
 0:27 & Shoulders align with the door frame. \\
 0:34 & Traversal is complete. \\
 \hline
\end{tabular}
\label{tab:alex_left_push_event_timeline}
\end{table}

\begin{table}[H]
\caption{Per-phase timing breakdown for the March~9,~2026 Alex left push lever-handle traversal.}
\centering
\setlength{\tabcolsep}{4pt}
\renewcommand{\arraystretch}{1.01}
\begin{tabular}{p{2.1cm} c c p{0.41\columnwidth}}
 \hline
 Phase & Time window & Duration & Description \\
 \hline
 Approach & 0:03--0:17 & 14~s & Coarse approach from the distant start, followed by the fine approach into the authored opening stance. \\
 Unlatch & 0:17--0:20 & 3~s & Final upper-body alignment, lever engagement, and latch release. \\
 Panel motion & 0:20--0:26 & 6~s & Push the panel clear while continuing to bias the body toward forward progress. \\
 Traverse & 0:26--0:34 & 8~s & Commit through the frame, align the shoulders with the opening by 0:27, and clear the doorway. \\
 \hline
\end{tabular}
\label{tab:alex_left_push_phase_breakdown}
\end{table}

The 34~s push side run is faster than the 45~s pull run on the same robot and week.
Approach and panel motion together take only 20~s, and shoulders align with the frame just 1~s after traversal begins (\autoref{tab:alex_left_push_phase_breakdown}).
This matches the progress curve pattern in \autoref{fig:DoorTraversalTimingsSpeed}, where push doors allow more continuous forward motion.

\subsubsection{Reactive Single-Table Ball Sorting}
\label{sec:reactive_ball_sorting}

\begin{figure}[H]
    \centering
    \includegraphics[width=1.0\columnwidth]{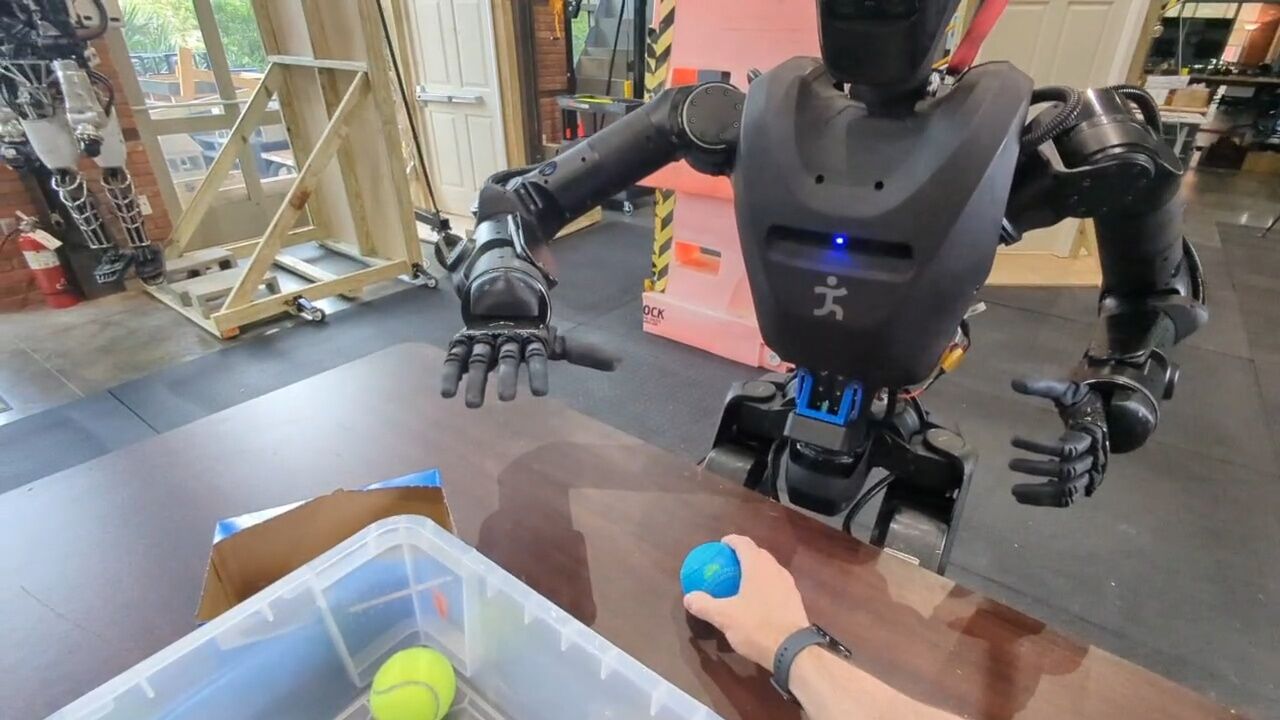}
    \caption[Alex reactive single-table ball sorting.]{Alex reactive single-table ball sorting on April 4, 2026.
    The run includes human disturbances; the tree abandons a stale pick and returns to search when the hand-centered containment check shows that the object is no longer in the hand.
    A video is available at \url{https://youtu.be/9DhgbjIRynM}.}
    \label{fig:ReactiveRobustBallSortingDisturbance}
\end{figure}

On April 4, 2026, Alex completed a reactive single-table ball-sorting run in 45.2 seconds while humans continuously placed and sometimes removed balls on the table, shown in \autoref{fig:ReactiveRobustBallSortingDisturbance}.
Timestamps are presented in \autoref{tab:AlexReactiveRobustBallSortingTimeline}.

\begin{table}[H]
    \caption{Timeline of the Alex reactive ball-sorting demonstration. Times are relative to the first human ball placement, which occurs at 1.2~s in the source video.}
    \centering
    \small
    \begin{tabular}{r p{10.1cm}}
        \hline
        Time from start (s) & Event \\
        \hline
        0.0 & Human places ball 1 (yellow) on the table. \\
        4.5 & Robot grasps ball 1. \\
        6.9 & Robot places ball 1 in container 1. \\
        7.2 & Human places ball 2 (yellow) on the table. \\
        11.4 & Robot grasps ball 2. \\
        13.9 & Robot places ball 2 in container 1. \\
        14.5 & Human places ball 3 (yellow) on the table. \\
        17.8 & Human removes ball 3 from the scene. \\
        20.6 & Robot identifies that the pick failed. \\
        21.9 & Human places ball 4 (blue) on the table. \\
        27.4 & Robot grasps ball 4. \\
        29.8 & Human places ball 5 (yellow) on the table. \\
        30.9 & Robot places ball 4 in container 2. \\
        32.2 & Human places ball 6 (yellow) on the table. \\
        35.8 & Robot grasps ball 5. \\
        38.2 & Robot places ball 5 in container 1. \\
        42.8 & Robot grasps ball 6. \\
        45.2 & Robot places ball 6 in container 1. \\
        \hline
    \end{tabular}
    \label{tab:AlexReactiveRobustBallSortingTimeline}
\end{table}

The 45.2~s run sorted six balls while humans continuously placed new balls on the table (\autoref{tab:AlexReactiveRobustBallSortingTimeline}).
This rate of action supports hypothesis 1 on manipulation speed under a dynamic scene.
Resilience to the ball removal disturbance at 17.8~s is analyzed in \autoref{sec:reactive_ball_sorting_disturbance}.

\subsubsection{Two-Table Loco-Manipulation Sorting}

\begin{figure}[H]
    \centering
    \includegraphics[width=0.95\columnwidth]{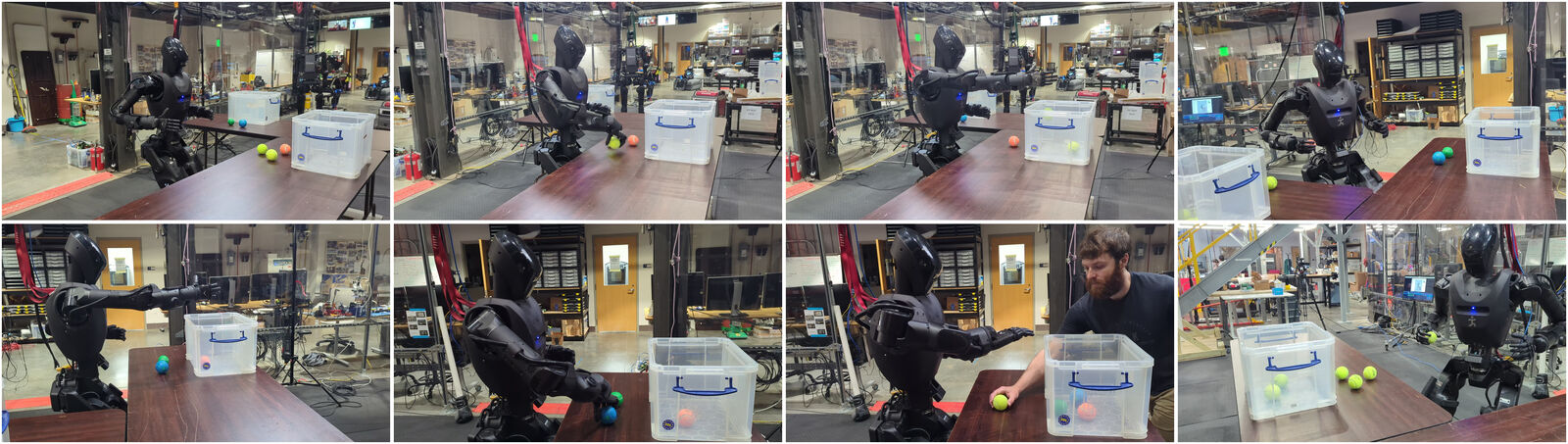}
    \caption{The April 14, 2026 Alex two-table loco-manipulation ball-sorting task.
    A video is available at \url{https://youtu.be/KxIihoDiMtc}.}
    \label{fig:alex_two_table_sorting_defense}
\end{figure}

On April 14, 2026, we extended the reactive ball-sorting behavior to a two-table loco-manipulation task, shown in \autoref{fig:alex_two_table_sorting_defense}.
In the timed execution run used in \autoref{fig:in_house_speed_phase_timeline}, the robot sorted nine balls across two tables in 2 minutes and 8 seconds.
Timestamps are presented in \autoref{tab:alex_two_table_sorting_event_timeline}.

\begin{table}[H]
\caption{Event times for the April~14,~2026 Alex two-table loco-manipulation sorting run shown in \autoref{fig:alex_two_table_sorting_defense}.}
\centering
\footnotesize
\begin{tabular}{c p{0.72\columnwidth}}
 \hline
 Time & Event \\
 \hline
 0:00 & Run begins; walk to table~A. \\
 0:10 & Begin ball sorting at table~A. \\
 0:15 & Grasp ball. \\
 0:17 & Place ball in container. \\
 0:22 & Grasp ball. \\
 0:24 & Place ball in container. \\
 0:29 & Grasp ball. \\
 0:38 & Walk to table~B. \\
 0:46 & Begin ball sorting at table~B. \\
 0:49 & Place ball carried from table~A. \\
 1:05 & Grasp ball. \\
 1:08 & Place ball in container. \\
 1:12 & Grasp ball. \\
 1:15 & Place ball in container. \\
 1:21 & Grasp ball. \\
 1:31 & Walk to table~A. \\
 1:39 & Begin ball sorting at table~A. \\
 1:43 & Place ball in container. \\
 1:49 & Grasp ball. \\
 1:52 & Place ball in container. \\
 1:56 & Grasp ball. \\
 2:00 & Place ball in container. \\
 2:04 & Grasp ball. \\
 2:06 & Place ball in container. \\
 2:08 & Run complete. \\
 \hline
\end{tabular}
\label{tab:alex_two_table_sorting_event_timeline}
\end{table}

Nine balls were sorted between two tables in 2 minutes 8 seconds, with locomotion between tables at 0:00, 0:38, and 1:31 (\autoref{tab:alex_two_table_sorting_event_timeline}).
This demonstration shows a loco-manipulation capability beyond door traversal and exploration tasks.
The timed run appears in \autoref{fig:in_house_speed_phase_timeline}; the behavior extension that produced it is analyzed under hypothesis 3.

\subsubsection{Nadia Door-Traversal Progress Curves}

\begin{figure}[H]
    \centering
        \includegraphics[width=0.8\columnwidth]{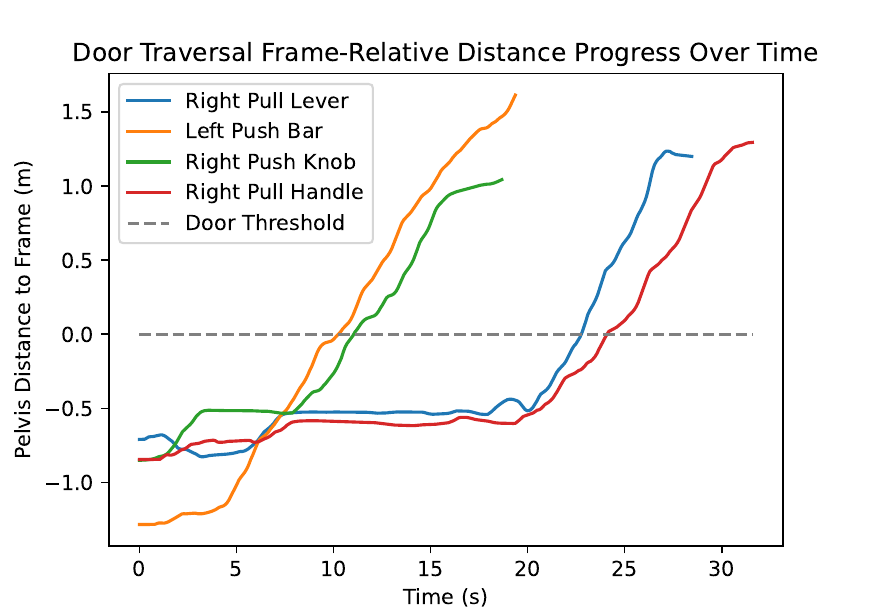}
    \caption{Distance progress through the door frame for four representative 2024 Nadia real-robot door traversals. Push-door behaviors are faster because the robot can keep making forward progress. Pull-door behaviors take longer because the robot must stay clear of the panel while fully opening it before traversal.}
    \label{fig:DoorTraversalTimingsSpeed}
\end{figure}

\autoref{fig:DoorTraversalTimingsSpeed} illustrates representative forward progress curves for door traversals on Nadia, showing the difference between push and pull doors.
Pull doors can take longer because the robot must stay clear of the panel as it opens.

The four Nadia curves separate push doors, which keep moving forward, from pull doors, which stall progress while the panel opens.
The 14~s push bar and 27~s pull handle runs above match these two curve shapes.
The Alex push and pull traversals above follow the same qualitative pattern at 34~s and 45~s, even though they are not plotted in this figure.

\autoref{fig:in_house_speed_phase_timeline} summarizes the speed arc from the 106~s Atlas anchor through 14 to 45~s door traversals and minute scale manipulation tasks.
The individual runs above supply the timed events behind each bar.
Together they support the speed and capability parts of hypothesis 1.

\subsection{Reliability and Resilience Evidence}

\begin{figure}[H]
    \centering
    {\singlespacing
    \resizebox{\columnwidth}{!}{%
        \input{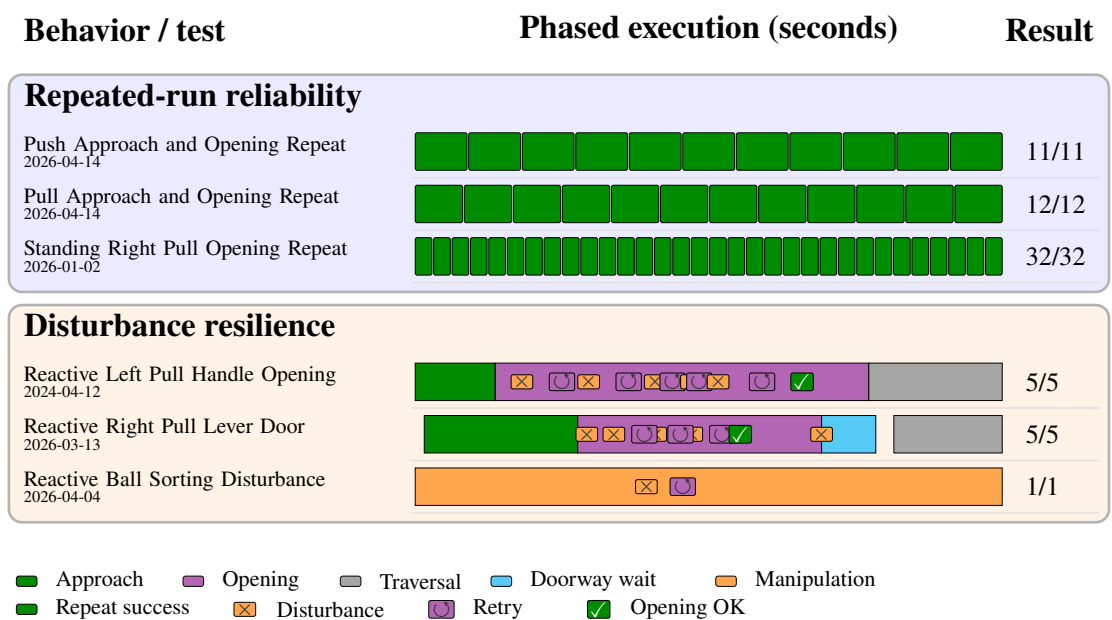}%
    }
    \par
    }
    \caption{
        In-house reliability and resilience evidence at a glance.
        The repeated run band encodes each successful trial as a green cell (11/11, 12/12, and 32/32).
        The disturbance resilience band shows elapsed time phases with disturbances, retries, and opening success marked at measured timestamps (Nadia reactive pull 44~s, Alex reactive pull 65~s, ball sort 45.2~s).
    }
    \label{fig:in_house_reliability_resilience_summary}
\end{figure}

In \autoref{fig:in_house_reliability_resilience_summary}, we summarize our main in-house reliability and resilience anchors.
The top band shows consecutive successful trials for repeated door approach and opening on Alex and Unitree H1-2.
The bottom band shows disturbed runs where the robot retried and recovered.
Photos and event tables for each run follow in the subsections below.

\subsubsection{Alex Left Push Approach and Opening Repeat}

\begin{figure}[H]
    \centering
        \includegraphics[width=1.0\columnwidth]{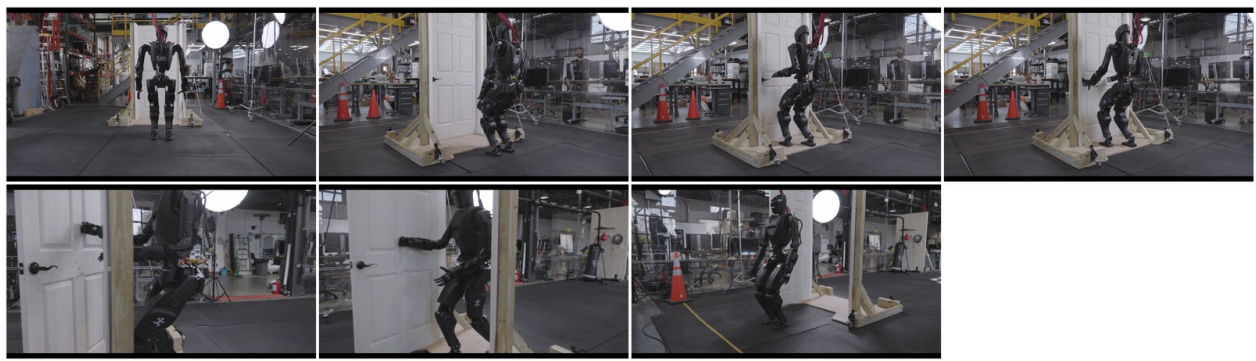}
    \caption{April~14,~2026 Alex left push door approach and opening reliability test.
    Key frames from the behavior repeated in the test.
    A video is available at \url{https://youtu.be/N4rzKOKUjIA}.}
    \label{fig:alex_push_reliability_20260414}
\end{figure}

On April 14, 2026, we ran the left push door behavior on Alex with a goal of at least 10 approach and opening trials.
We lost count during the session and kept going so we would not stop short.
The test ended with 11 successes in a row.
All trials used the same lab left push door on the same night.
An operator was present at the UI.
We did not run the test to failure.
Each run completed the approach and opening phases without failure.
We did not repeat full doorway traversal in this test.
The test is shown in \autoref{fig:alex_push_reliability_20260414}.
Trial outcomes are presented in \autoref{tab:alex_push_reliability_runs}.

\begin{table}[H]
\caption{Trial outcomes for the April~14,~2026 Alex left push approach and opening reliability test shown in \autoref{fig:alex_push_reliability_20260414}.}
\centering
\footnotesize
\begin{tabular}{c c p{0.58\columnwidth}}
 \hline
 Run & Outcome & Phase completed \\
 \hline
 1 & Success & Approach and opening \\
 2 & Success & Approach and opening \\
 3 & Success & Approach and opening \\
 4 & Success & Approach and opening \\
 5 & Success & Approach and opening \\
 6 & Success & Approach and opening \\
 7 & Success & Approach and opening \\
 8 & Success & Approach and opening \\
 9 & Success & Approach and opening \\
 10 & Success & Approach and opening \\
 11 & Success & Approach and opening \\
 \hline
\end{tabular}
\label{tab:alex_push_reliability_runs}
\end{table}

Eleven consecutive approach and opening trials completed without failure (\autoref{tab:alex_push_reliability_runs}).
We did not test full traversal because the walking controller was not working well for traversal at the time of these experiments.
However, our repeated trials still demonstrate reliable approach and opening.
This 11/11 result is the Alex left push row in \autoref{fig:in_house_reliability_resilience_summary}.

\subsubsection{Alex Right Pull Approach and Opening Repeat}

\begin{figure}[H]
    \centering
        \includegraphics[width=1.0\columnwidth]{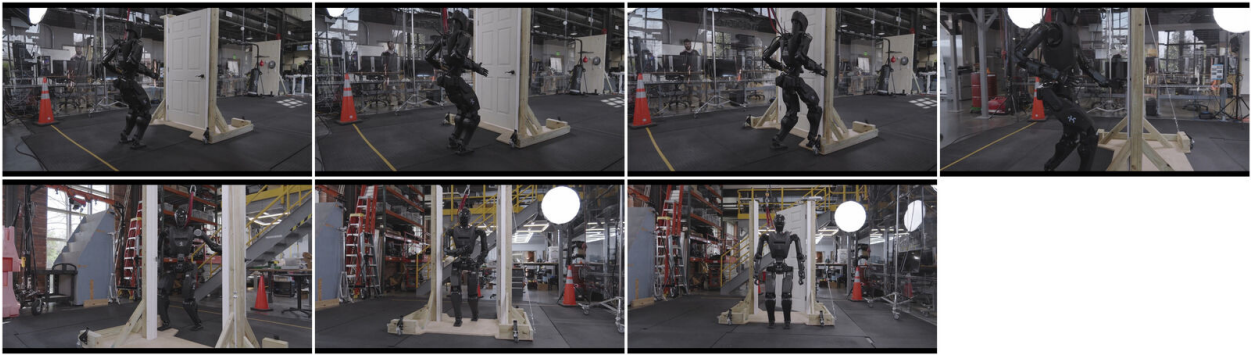}
    \caption{April~14,~2026 Alex right pull door approach and opening reliability test.
    Key frames from the behavior repeated in the test.
    A video is available at \url{https://youtu.be/l2piKf40ea4}.}
    \label{fig:alex_pull_reliability_20260414}
\end{figure}

On April 14, 2026, we ran the right pull door behavior on Alex with a goal of at least 10 approach and opening trials.
We lost count during the session and kept going so we would not stop short.
The test ended with 12 successes in a row.
All trials used the same lab right pull door on the same night as the repeated push test.
An operator was present at the UI.
We did not run the test to failure.
Each run completed the approach and opening phases without failure.
We did not repeat full doorway traversal in this test.
The test is shown in \autoref{fig:alex_pull_reliability_20260414}.
Trial outcomes are presented in \autoref{tab:alex_pull_reliability_runs}.

\begin{table}[H]
\caption{Trial outcomes for the April~14,~2026 Alex right pull approach and opening reliability test shown in \autoref{fig:alex_pull_reliability_20260414}.}
\centering
\footnotesize
\begin{tabular}{c c p{0.58\columnwidth}}
 \hline
 Run & Outcome & Phase completed \\
 \hline
 1 & Success & Approach and opening \\
 2 & Success & Approach and opening \\
 3 & Success & Approach and opening \\
 4 & Success & Approach and opening \\
 5 & Success & Approach and opening \\
 6 & Success & Approach and opening \\
 7 & Success & Approach and opening \\
 8 & Success & Approach and opening \\
 9 & Success & Approach and opening \\
 10 & Success & Approach and opening \\
 11 & Success & Approach and opening \\
 12 & Success & Approach and opening \\
 \hline
\end{tabular}
\label{tab:alex_pull_reliability_runs}
\end{table}

Twelve consecutive approach and opening trials completed without failure on the same night as the repeated push test (\autoref{tab:alex_pull_reliability_runs}).
Again, the scope was approach and opening only.
This 12/12 result is the Alex right pull row in \autoref{fig:in_house_reliability_resilience_summary}.

\subsubsection{Unitree H1-2 Standing Right Pull Opening Repeat}

\begin{figure}[H]
    \centering
    \includegraphics[width=0.92\columnwidth]{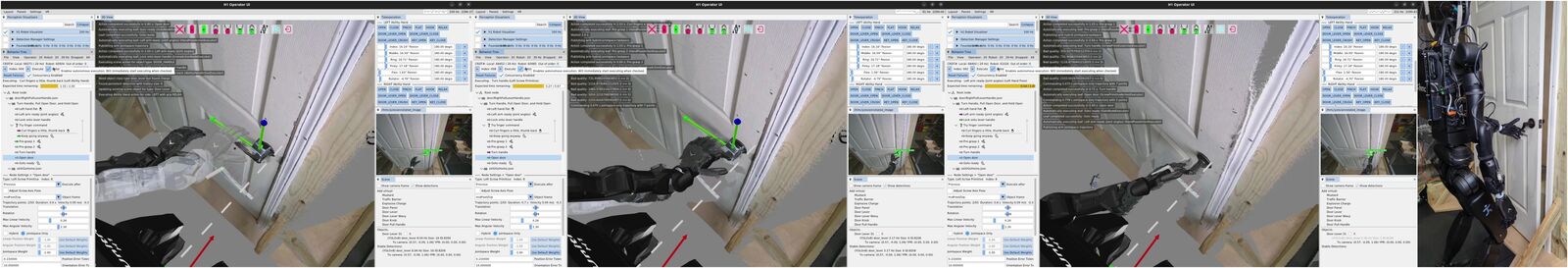}
    \caption{January~2,~2026 Unitree H1-2 right pull opening reliability test.
    The left three panels show operator UI key frames.
    The right panel shows a third person view.
    The behavior opened the door 32 times in a row from a standing start.
    The test took about five minutes.
    A video is available at \url{https://youtu.be/fQCNuXEF9xg}.}
    \label{fig:unitree_pull_opening_reliability_20260102}
\end{figure}

On January 2, 2026, we ran the standing right pull lever opening behavior 32 times in a row on Unitree H1-2.
All trials used the same lab right pull door during the same session.
An operator was present at the UI.
We stopped after 32 successes; we did not run the test to failure.
The robot opened the door from a standing start each time.
The full test took about five minutes.
The test is shown in \autoref{fig:unitree_pull_opening_reliability_20260102}.
The test protocol is presented in \autoref{tab:unitree_pull_opening_reliability_protocol}.

\begin{table}[H]
\caption{Protocol summary for the January~2,~2026 Unitree H1-2 standing right pull opening repeat test shown in \autoref{fig:unitree_pull_opening_reliability_20260102}.}
\centering
\begin{tabular}{p{0.28\columnwidth} p{0.62\columnwidth}}
 \hline
 Item & Description \\
 \hline
 Start condition & Standing at the door with the opening loop behavior loaded \\
 Same door & One lab right pull lever door for all 32 trials \\
 Operator & Present at the UI throughout the session \\
 Test scope & Opening only; no approach or doorway traversal \\
 Stop rule & Stopped after 32 successes; not run to failure \\
 Loop structure & Goto action returns to the start of the opening sequence after each success \\
 Measured result & 32/32 successful openings \\
 Elapsed time & About five minutes \\
 \hline
\end{tabular}
\label{tab:unitree_pull_opening_reliability_protocol}
\end{table}

The robot opened the same lab door 32 times in a row from a standing start in about five minutes (\autoref{tab:unitree_pull_opening_reliability_protocol}).
This repeated trial was autonomous even between repetitions: the goto loop returned to the same opening sequence after each success.
It also serves as evidence that our system's reliability extends beyond a single humanoid robot platform.
The 32/32 result is the Unitree row in \autoref{fig:in_house_reliability_resilience_summary}.

\subsubsection{Nadia Reactive Left Pull Handle Opening}

\begin{figure}[H]
    \centering
    \includegraphics[width=0.95\columnwidth]{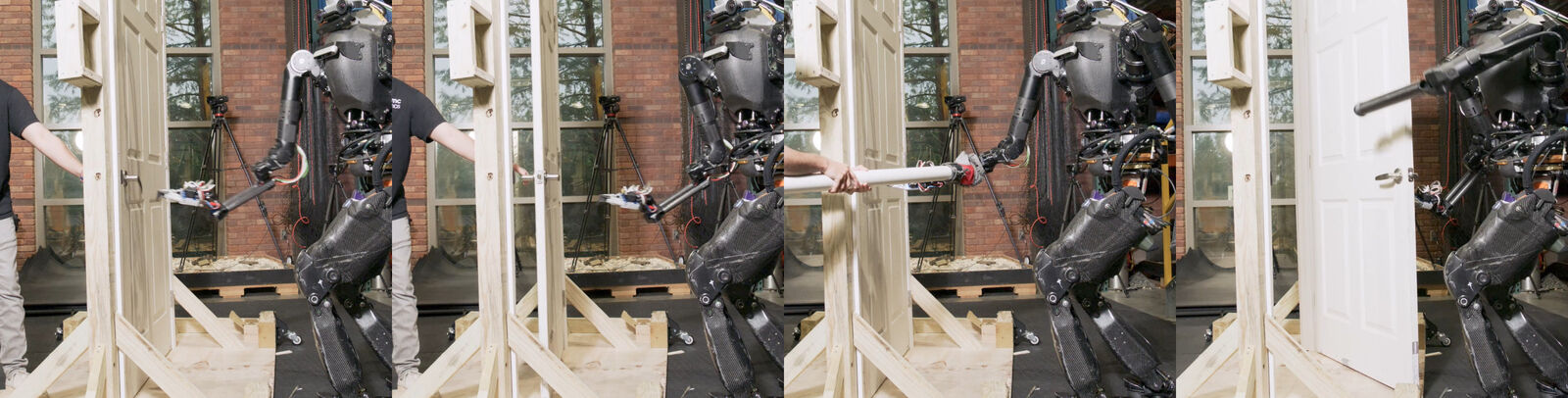}
    \caption{April~12,~2024 Nadia reactive left pull handle opening under human disturbance.
    The robot retried five times and completed the full door traversal.
    A video is available at \url{https://youtu.be/j_rzh5cAP2E}.}
    \label{fig:nadia_reactive_pull_20240412}
\end{figure}

On April 12, 2024, we disturbed a Nadia left pull door opening five times during one run.
The hard coded door traversal node detected each failure and retried until the door opened.
The robot finished the full traversal without operator intervention.
The run is shown in \autoref{fig:nadia_reactive_pull_20240412}.
Timestamps are presented in \autoref{tab:nadia_reactive_pull_event_timeline}.

\begin{table}[H]
\caption{Event times for the April~12,~2024 Nadia reactive left pull handle opening shown in \autoref{fig:nadia_reactive_pull_20240412}.}
\centering
\footnotesize
\begin{tabular}{c p{0.72\columnwidth}}
 \hline
 Time & Event \\
 \hline
 0:00 & Robot begins approach to left pull door. \\
 0:06 & Robot begins to grasp and turn door handle. \\
 0:08 & Human holds door closed; robot gripper slips off handle (disturbance 1). \\
 0:11 & Robot attempts 2nd grasp and turn of handle. \\
 0:13 & Human allows partial opening, then pulls door back; robot hand slips off again (disturbance 2). \\
 0:16 & With door closed again, robot attempts 3rd grasp and turn. \\
 0:18 & Human allows partial opening, pulls back; grasp slips off again (disturbance 3). \\
 0:20 & Human pushes robot arm back during 4th reach attempt, preventing grasp (disturbance 4). \\
 0:22 & Robot arm pushed during reach for 5th grasp attempt in the same way (disturbance 5). \\
 0:26 & Robot begins 6th grasp attempt. \\
 0:29 & Robot is allowed to successfully open the door. \\
 0:34 & Robot completes opening the door; begins traversal steps. \\
 0:39 & Robot shoulders even with door frame. \\
 0:44 & Robot fully clears doorway. \\
 \hline
\end{tabular}
\label{tab:nadia_reactive_pull_event_timeline}
\end{table}

Five human disturbances during opening caused grasp slips or blocked reaches (\autoref{tab:nadia_reactive_pull_event_timeline}).
A hard coded traversal node retried until the door opened at 29~s and the robot cleared the doorway at 44~s without operator intervention.
This is an earlier resilience mechanism than the authored fallback trees used in later Alex behaviors, but it still produced the Nadia reactive pull row in \autoref{fig:in_house_reliability_resilience_summary}.

\subsubsection{Alex Reactive Right Pull Lever Door}

\begin{figure}[H]
    \centering
    \includegraphics[width=1.0\columnwidth]{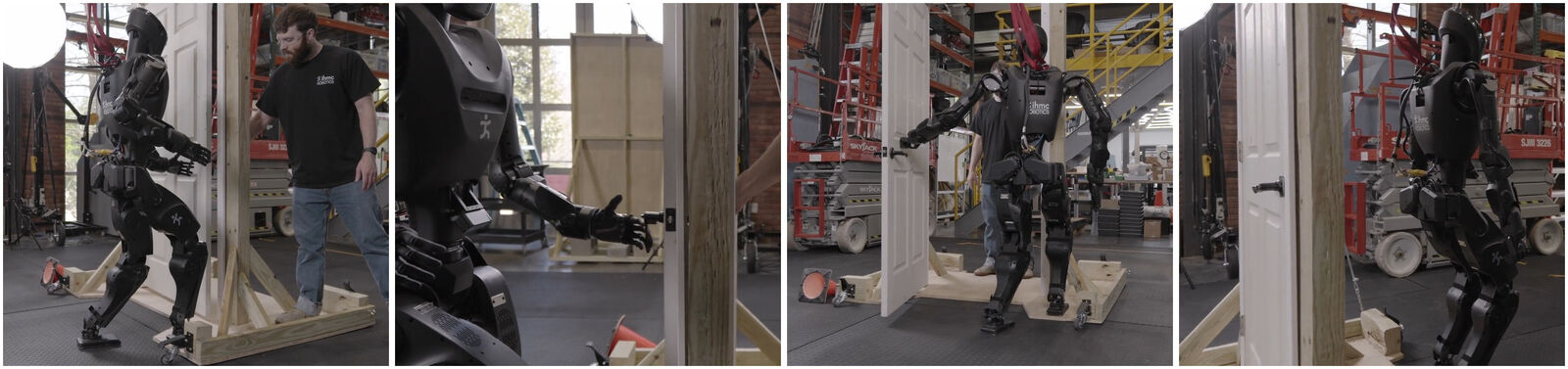}
    \caption{March~13,~2026 Alex right pull door reactivity demonstration.
    The opening retries and blocked doorway wait are authored in the behavior tree.
    A video is available at \url{https://youtu.be/pWxptD5j8b4}.}
    \label{fig:alex_reactive_pull_20260313}
\end{figure}

On March 13, 2026, we disturbed an Alex right pull lever door opening four times during the opening phase.
A human also blocked the doorway before traversal.
The behavior retried the opening and waited until the doorway was clear.
The run is shown in \autoref{fig:alex_reactive_pull_20260313}.
Timestamps are presented in \autoref{tab:alex_reactive_pull_event_timeline}.

\begin{table}[H]
\caption{Event times for the March~13,~2026 Alex reactive right pull lever door run shown in \autoref{fig:alex_reactive_pull_20260313}.}
\centering
\footnotesize
\begin{tabular}{c p{0.72\columnwidth}}
 \hline
 Time & Event \\
 \hline
 0:01 & Robot begins coarse door approach. \\
 0:10 & Robot begins fine door approach. \\
 0:18 & Robot begins 1st grasp and turn lever handle attempt. \\
 0:19 & Door partially opens; human pulls door back shut; robot grasp slips (disturbance 1). \\
 0:22 & Robot begins 2nd grasp attempt; human holds door shut; robot grasp slips (disturbance 2). \\
 0:26 & Robot begins 3rd grasp attempt; human allows partial opening again; robot grasp slips (disturbance 3). \\
 0:30 & Robot begins 4th grasp attempt; human pulls door back again; robot grasp slips (disturbance 4). \\
 0:34 & Robot begins 5th grasp attempt. \\
 0:36 & Robot is allowed to open the door. \\
 0:45 & Robot completes opening door all the way; human is now blocking doorway. \\
 0:51 & Human moves out of doorway. \\
 0:53 & Robot begins door traversal steps. \\
 1:01 & Robot shoulders even with door frame. \\
 1:05 & Robot fully clears doorway. \\
 \hline
\end{tabular}
\label{tab:alex_reactive_pull_event_timeline}
\end{table}

Four disturbances during opening and a blocked doorway before traversal triggered retries and a wait for clearance (\autoref{tab:alex_reactive_pull_event_timeline}).
The run finished in 65~s overall.
Unlike the 2024 Nadia case, the retry and wait logic here was authored in the behavior tree with fallback and condition nodes, and it appears as the Alex reactive pull row in \autoref{fig:in_house_reliability_resilience_summary}.

\subsubsection{Reactive Single Table Ball Sorting Disturbance}
\label{sec:reactive_ball_sorting_disturbance}

For the previously presented 45.2~s sorting run in \autoref{sec:reactive_ball_sorting}, the ball removal at 17.8~s is the main resilience event.
The reactive tree logic abandoned the stale pick at 20.6~s and still placed all six balls by 45.2~s.
That recovery is the ball sorting row in \autoref{fig:in_house_reliability_resilience_summary}.

The repeated run and disturbance results above support reliability and resilience within hypothesis 1.
We do not have comparable repeat trial data for prior IHMC door baselines.
Our tests also stopped short of repeated full door traversals because walk through performance was not yet trustworthy enough to risk the hardware.

\section{Hypothesis 2: Fast Behavior Authoring}
\label{sec:h2_fast_behavior_authoring}

To support our second hypothesis, we'll show how quickly new behaviors can be authored on the real robot using the runtime edit-test loop, rather than redeploy, restart, or retrain workflows.

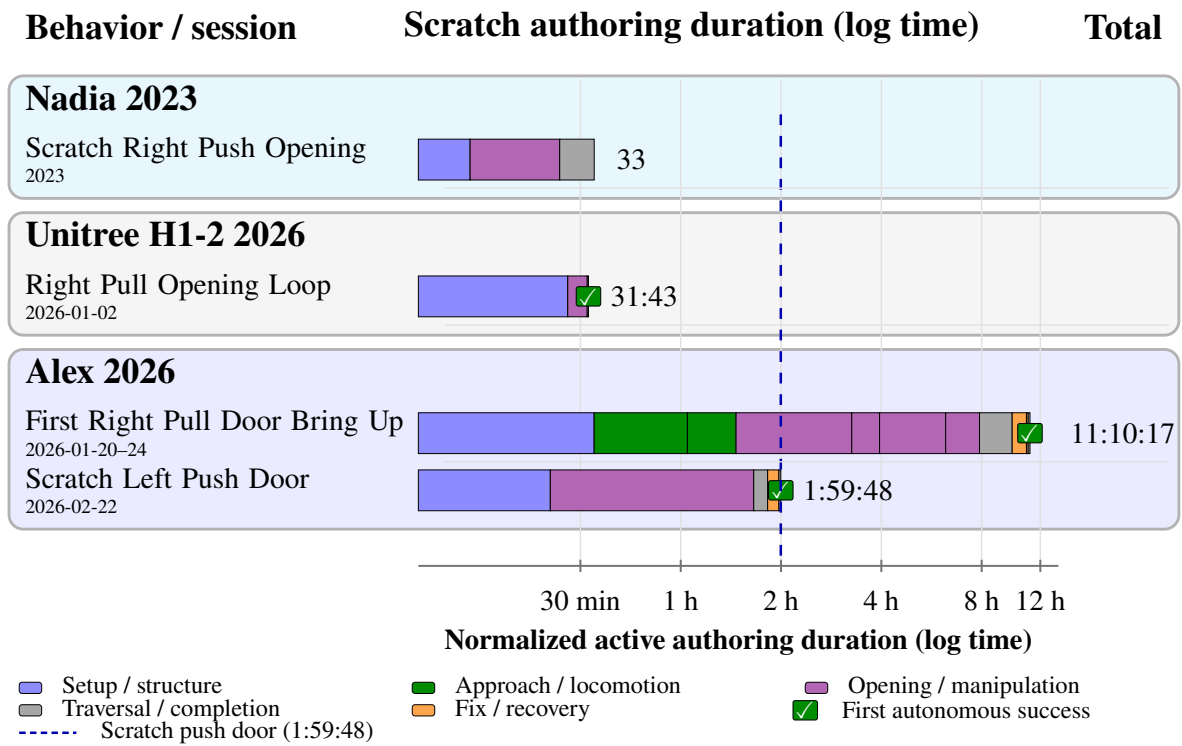
\begin{figure}[H]
    \centering
    {\singlespacing
    \resizebox{\columnwidth}{!}{%
        \newcommand{\inauthlabel}[2]{#1\\[-2.4pt]{\fontsize{5.5}{6.2}\selectfont #2}}
\newcommand{\inauthrowlabel}[3]{%
  \node[rowlbl, anchor=west] at (\systemleft + 0.02, #1) {\inauthlabel{#2}{#3}};}
\newcommand{\inauthnum}[3]{%
  \pgfmathparse{#2 > 9.8 ? 1 : 0}%
  \ifnum\pgfmathresult=1
    \node[numlbl, anchor=east] at (\numright - 0.10, #1) {#3};%
  \else
    \pgfmathsetmacro{\inauthnumx}{#2 + 0.08 - \authbarshift}%
    \node[numlbl, anchor=west] at (\inauthnumx, #1) {#3};%
  \fi}
\newcommand{\authbarphase}[5]{%
  \filldraw[#5, draw=black, line width=0.28pt]
    (#3, {#1 - #2}) rectangle (#4, {#1 + #2});}
\newcommand{\authbarsuccess}[2]{%
  \node[draw=black, line width=0.28pt, fill=green!55!black, rounded corners=0.6pt, inner xsep=0.8pt, inner ysep=0.4pt, font=\fontsize{6.0}{6.4}\selectfont\bfseries, text=white] at (#2, #1) {$\checkmark$};}
\newcommand{\inauthswatch}[3]{%
  \filldraw[#1, rounded corners=0.8pt] (#2, {\inlegendy - 0.055}) rectangle ++(0.22, 0.11);%
  \node[legendtxt, anchor=west] at ({#2 + 0.28}, \inlegendy) {#3};}

\begin{tikzpicture}[
    x=1cm,
    y=1cm,
    every node/.style={outer sep=0pt},
    rowlbl/.style={font=\fontsize{8.2}{9.4}\selectfont, anchor=west, align=left, text width=4.20cm},
    numlbl/.style={font=\fontsize{8.2}{8.8}\selectfont, anchor=west, align=left, text width=0.82cm},
    bandtitle/.style={font=\footnotesize\bfseries, anchor=north west, inner xsep=1.2pt, inner ysep=0.8pt},
    legendtxt/.style={font=\fontsize{6.8}{7.4}\selectfont}
  ]
  \def\barh{0.20}

  \def\systemw{4.25}
  \def\barw{6.35}
  \def\numw{0.82}
  \def\colgap{0.02}
  \def\headerh{1.20}
  \def\rowh{0.56}
  \def\bandgap{0.14}
  \def\bandtitleh{0.54}
  \def\bandpadbottom{0.10}
  \def\bottomregionh{2.45}

  \pgfmathsetmacro{\systemleft}{0}
  \pgfmathsetmacro{\systemright}{\systemleft + \systemw}
  \pgfmathsetmacro{\barleft}{\systemright + \colgap}
  \pgfmathsetmacro{\barright}{\barleft + \barw}
  \pgfmathsetmacro{\numleft}{\barright + \colgap}
  \pgfmathsetmacro{\numright}{\numleft + \numw}
  \pgfmathsetmacro{\figurewidth}{\numright}

  \pgfmathsetmacro{\bandah}{\bandtitleh + \rowh + \bandpadbottom}
  \pgfmathsetmacro{\bandbh}{\bandtitleh + \rowh + \bandpadbottom}
  \pgfmathsetmacro{\bandch}{\bandtitleh + 2 * \rowh + \bandpadbottom}
  \pgfmathsetmacro{\figureheight}{\headerh + \bandah + \bandgap + \bandbh + \bandgap + \bandch + \bottomregionh}

  \pgfmathsetmacro{\headertop}{\figureheight}
  \pgfmathsetmacro{\headerbottom}{\headertop - \headerh}
  \pgfmathsetmacro{\bandatop}{\headerbottom}
  \pgfmathsetmacro{\bandabottom}{\bandatop - \bandah}
  \pgfmathsetmacro{\bandbtop}{\bandabottom - \bandgap}
  \pgfmathsetmacro{\bandbbottom}{\bandbtop - \bandbh}
  \pgfmathsetmacro{\bandctop}{\bandbbottom - \bandgap}
  \pgfmathsetmacro{\bandcbottom}{\bandctop - \bandch}

  \pgfmathsetmacro{\rownadia}{\bandatop - \bandtitleh - 0.5 * \rowh}
  \pgfmathsetmacro{\rowunitree}{\bandbtop - \bandtitleh - 0.5 * \rowh}
  \pgfmathsetmacro{\rowalexpull}{\bandctop - \bandtitleh - 0.5 * \rowh}
  \pgfmathsetmacro{\rowalexpush}{\rowalexpull - \rowh}

  \def\authbarshift{0.30}

  \pgfmathsetmacro{\authaxisleft}{\barleft + 0.04}
  \pgfmathsetmacro{\authaxisright}{\barright - 0.04}
  \pgfmathsetmacro{\axisy}{\bandcbottom - 0.36}
  \def\legendentryh{0.11}
  \def\legendrowpitch{0.22}
  \pgfmathsetmacro{\axistitley}{\axisy - 0.48}
  \pgfmathsetmacro{\axistitlebottom}{\axistitley - 0.18}
  \pgfmathsetmacro{\legendgapbelowtitle}{0.48}
  \pgfmathsetmacro{\inlegendrowA}{\axistitlebottom - \legendgapbelowtitle - 0.5 * \legendentryh}
  \pgfmathsetmacro{\inlegendrowB}{\inlegendrowA - \legendrowpitch}
  \pgfmathsetmacro{\inlegendrowC}{\inlegendrowB - \legendrowpitch}
  \pgfmathsetmacro{\durationheaderx}{\authaxisleft - 0.28}
  \def\inauthlegendcolA{0.10}
  \def\inauthlegendcolB{3.95}
  \def\inauthlegendcolC{7.80}

  \path[draw=black!30, fill=cyan!8, rounded corners=4pt, line width=0.8pt]
    (0, \bandabottom) rectangle (\figurewidth, \bandatop);
  \path[draw=black!30, fill=gray!8, rounded corners=4pt, line width=0.8pt]
    (0, \bandbbottom) rectangle (\figurewidth, \bandbtop);
  \path[draw=black!30, fill=blue!8, rounded corners=4pt, line width=0.8pt]
    (0, \bandcbottom) rectangle (\figurewidth, \bandctop);

  \node[anchor=west, font=\footnotesize\bfseries] at (\systemleft + 0.02, \headerbottom + 0.48) {Behavior / session};
  \node[anchor=east, font=\footnotesize\bfseries] at (\numright - 0.02, \headerbottom + 0.48) {Total};

  \inauthrowlabel{\rownadia}{Scratch Right Push Opening}{2023}
  \inauthrowlabel{\rowunitree}{Right Pull Opening Loop}{2026-01-02}
  \inauthrowlabel{\rowalexpull}{First Right Pull Door Bring Up}{2026-01-20--24}
  \inauthrowlabel{\rowalexpush}{Scratch Left Push Door}{2026-02-22}

  \begin{scope}[shift={(-\authbarshift,0)}]
  \node[anchor=west, font=\footnotesize\bfseries] at (\durationheaderx, \headerbottom + 0.48) {Scratch authoring duration (log time)};

  \foreach \tickx in {5.8974,6.8801,7.8627,8.8453,9.8280,10.4028} {
    \draw[draw=black!12, line width=0.4pt] (\tickx, \axisy) -- (\tickx, \bandatop - 0.04);
  }

  \authbarphase{\rownadia}{\barh}{\authaxisleft}{4.8168}{blue!45}
  \authbarphase{\rownadia}{\barh}{4.8168}{5.6945}{violet!60}
  \authbarphase{\rownadia}{\barh}{5.6945}{6.0325}{black!35}

  \authbarphase{\rowunitree}{\barh}{\authaxisleft}{5.7732}{blue!45}
  \authbarphase{\rowunitree}{\barh}{5.7732}{5.9643}{violet!60}
  \authbarphase{\rowunitree}{\barh}{5.9643}{5.9763}{black!35}
  \authbarsuccess{\rowunitree}{5.9763}

  \authbarphase{\rowalexpull}{\barh}{\authaxisleft}{6.0318}{blue!45}
  \authbarphase{\rowalexpull}{\barh}{6.0318}{6.9455}{green!55!black}
  \authbarphase{\rowalexpull}{\barh}{6.9455}{7.4222}{green!55!black}
  \authbarphase{\rowalexpull}{\barh}{7.4222}{8.5552}{violet!60}
  \authbarphase{\rowalexpull}{\barh}{8.5552}{8.8280}{violet!60}
  \authbarphase{\rowalexpull}{\barh}{8.8280}{9.4755}{violet!60}
  \authbarphase{\rowalexpull}{\barh}{9.4755}{9.8070}{violet!60}
  \authbarphase{\rowalexpull}{\barh}{9.8070}{10.1271}{black!35}
  \authbarphase{\rowalexpull}{\barh}{10.1271}{10.2667}{orange!70}
  \authbarphase{\rowalexpull}{\barh}{10.2667}{10.2836}{orange!70}
  \authbarphase{\rowalexpull}{\barh}{10.2836}{10.3013}{black!35}
  \authbarsuccess{\rowalexpull}{10.3013}

  \authbarphase{\rowalexpush}{\barh}{\authaxisleft}{5.6026}{blue!45}
  \authbarphase{\rowalexpush}{\barh}{5.6026}{7.5950}{violet!60}
  \authbarphase{\rowalexpush}{\barh}{7.5950}{7.7316}{black!35}
  \authbarphase{\rowalexpush}{\barh}{7.7316}{7.8419}{orange!70}
  \authbarphase{\rowalexpush}{\barh}{7.8419}{7.8603}{black!35}
  \authbarsuccess{\rowalexpush}{7.8603}

  \draw[dashed, draw=blue!70!black, line width=0.7pt] (7.8603, \axisy + 0.10) -- (7.8603, \bandatop - 0.36);

  \draw[draw=black!55, line width=0.55pt] (\authaxisleft, \axisy) -- (\authaxisright, \axisy);
  \foreach \tickx/\ticklabel in {5.8974/30 min,6.8801/1 h,7.8627/2 h,8.8453/4 h,9.8280/8 h,10.4028/12 h} {
    \draw[draw=black!55, line width=0.55pt] (\tickx, \axisy - 0.06) -- (\tickx, \axisy + 0.06);
    \node[font=\scriptsize, anchor=north] at (\tickx, \axisy - 0.10) {\ticklabel};
  }
  \node[font=\scriptsize\bfseries, anchor=north] at ({0.5 * (\authaxisleft + \authaxisright)}, \axistitley) {Normalized active authoring duration (log time)};
  \end{scope}

  \inauthnum{\rownadia}{6.0325}{33}
  \inauthnum{\rowunitree}{5.9763}{31:43}
  \inauthnum{\rowalexpull}{10.3013}{11:10:17}
  \inauthnum{\rowalexpush}{7.8603}{1:59:48}

  \foreach \rowy in {\rownadia,\rowunitree,\rowalexpull} {
    \draw[draw=black!12, line width=0.4pt] (\barleft, {\rowy - 0.5 * \rowh}) -- (\figurewidth - 0.10, {\rowy - 0.5 * \rowh});
  }

  \node[bandtitle, fill=cyan!8] at (0.12, \bandatop - 0.06) {Nadia 2023};
  \node[bandtitle, fill=gray!8] at (0.12, \bandbtop - 0.06) {Unitree H1-2 2026};
  \node[bandtitle, fill=blue!8] at (0.12, \bandctop - 0.06) {Alex 2026};

  \pgfmathsetmacro{\inlegendy}{\inlegendrowA}
  \inauthswatch{blue!45, draw=black, line width=0.28pt}{\inauthlegendcolA}{Setup / structure}
  \inauthswatch{green!55!black, draw=black, line width=0.28pt}{\inauthlegendcolB}{Approach / locomotion}
  \inauthswatch{violet!60, draw=black, line width=0.28pt}{\inauthlegendcolC}{Opening / manipulation}

  \pgfmathsetmacro{\inlegendy}{\inlegendrowB}
  \inauthswatch{black!35, draw=black, line width=0.28pt}{\inauthlegendcolA}{Traversal / completion}
  \inauthswatch{orange!70, draw=black, line width=0.28pt}{\inauthlegendcolB}{Fix / recovery}
  \node[draw=black, line width=0.28pt, fill=green!55!black, rounded corners=0.6pt, inner xsep=0.8pt, inner ysep=0.4pt, font=\fontsize{6.0}{6.4}\selectfont\bfseries, text=white] at (\inauthlegendcolC, \inlegendy) {$\checkmark$};
  \node[legendtxt, anchor=west] at ({\inauthlegendcolC + 0.22}, \inlegendy) {First autonomous success};

  \pgfmathsetmacro{\inlegendy}{\inlegendrowC}
  \draw[dash pattern=on 1.8pt off 1.2pt, draw=blue!70!black, line width=0.7pt] (\inauthlegendcolA, \inlegendy) -- ++(0.56, 0);
  \node[legendtxt, anchor=west] at ({\inauthlegendcolA + 0.66}, \inlegendy) {Scratch push door (1:59:48)};
\end{tikzpicture}%
    }
    \par
    }
    \caption{
        In-house real-robot from-scratch authoring durations on a log-second axis with internal milestone structure.
        Rows are grouped by robot era.
        The Nadia 2023 row does a stand-in-place opening with ArUco perception and teleoperated traversal in the final segment.
        Checkmarks mark first autonomous success.
        Times are normalized active authoring; gaps between daily sessions are excluded for the Alex bring-up.
    }
    \label{fig:in_house_authoring_from_scratch_timeline}
\end{figure}

In \autoref{fig:in_house_authoring_from_scratch_timeline}, we summarize measured from-scratch authoring sessions on real hardware.
Each row shows normalized active authoring time on a log-second axis with internal milestone structure recovered from screen recordings.
Checkmarks mark first autonomous success.
Gaps between daily sessions are excluded for the multi-day Alex bring-up.

The subsections below walk through each row in \autoref{fig:in_house_authoring_from_scratch_timeline} with a representative still, the authoring-time table, and a link to the screen recording.

\subsection{Scratch Right Push Opening (Nadia 2023)}

\begin{figure}[H]
    \centering
    \includegraphics[width=0.95\columnwidth]{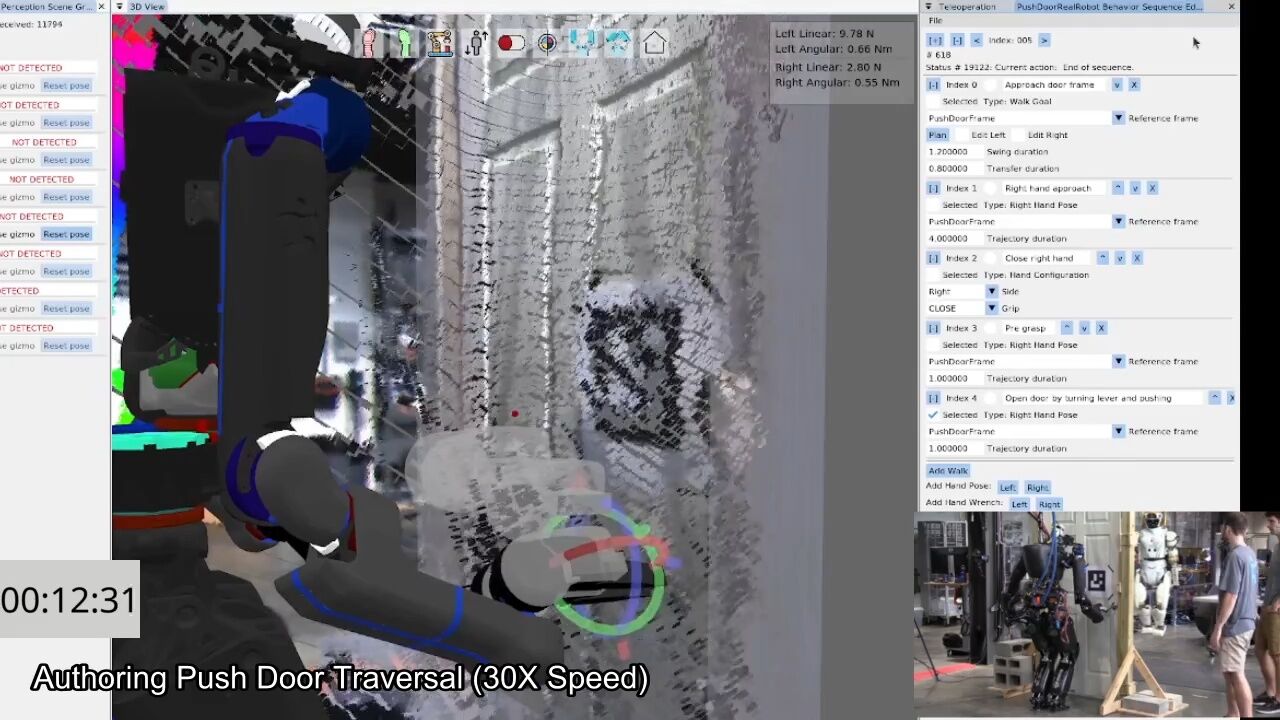}
    \caption{Nadia during the 2023 scratch right push door authoring session.
    A video is available at \url{https://youtu.be/SbBGpRHY_eE}.}
    \label{fig:h2_nadia_scratch_push_authoring}
\end{figure}

This row is our earliest measured scratch door opening on Nadia, at 33 minutes.
However, the behavior used ArUco marker perception, a squared up approach, and teleoperation for the final traversal steps.
The session is shown in \autoref{fig:h2_nadia_scratch_push_authoring}.
Authoring times are presented in \autoref{tab:h2_nadia_scratch_push_authoring_timeline}.

\begin{table}[H]
    \caption[2023 Nadia scratch right push authoring milestones.]{Historical 2023 Nadia right push door authoring milestones for the session summarized in \autoref{fig:in_house_authoring_from_scratch_timeline}.}
    \centering
    \footnotesize
    \setlength{\tabcolsep}{4pt}
    \renewcommand{\arraystretch}{0.97}
    \begin{tabular}{>{\RaggedLeft\arraybackslash}p{1.9cm} >{\RaggedRight\arraybackslash}p{\dimexpr\textwidth-1.9cm-6\tabcolsep\relax}}
        \hline
        Elapsed time & Authoring milestone \\
        \hline
        0:00 & Create new action sequence. \\
        0:02 & Approach door. \\
        0:05 & Right hand approaches handle. \\
        0:07 & Pre-grasp hand pose. \\
        0:09 & First handle-turn contact. \\
        0:14 & Latch disengaged. \\
        0:20 & Door pushed open with right hand. \\
        0:24 & Door pushed open more with left hand. \\
        0:25 & Door pushed open all the way with left hand. \\
        0:26 & Arms in collision-avoidance configuration. \\
        0:32 & Step forward a little. \\
        0:33 & Walk through the door frame. \\
        \hline
        \multicolumn{2}{l}{Total measured active authoring: 33 min.} \\
    \end{tabular}
    \label{tab:h2_nadia_scratch_push_authoring_timeline}
\end{table}

\autoref{tab:h2_nadia_scratch_push_authoring_timeline} is our earliest measured scratch door session at 33 minutes.
The session used ArUco perception and teleoperated traversal steps rather than full autonomy.
It still documents runtime editing on the robot, but it should not be directly compared to later sessions that produced fully autonomous behaviors.

\subsection{Right Pull Opening Loop (Unitree H1-2, 2026-01-02)}

\begin{figure}[H]
    \centering
    \includegraphics[width=1.0\columnwidth]{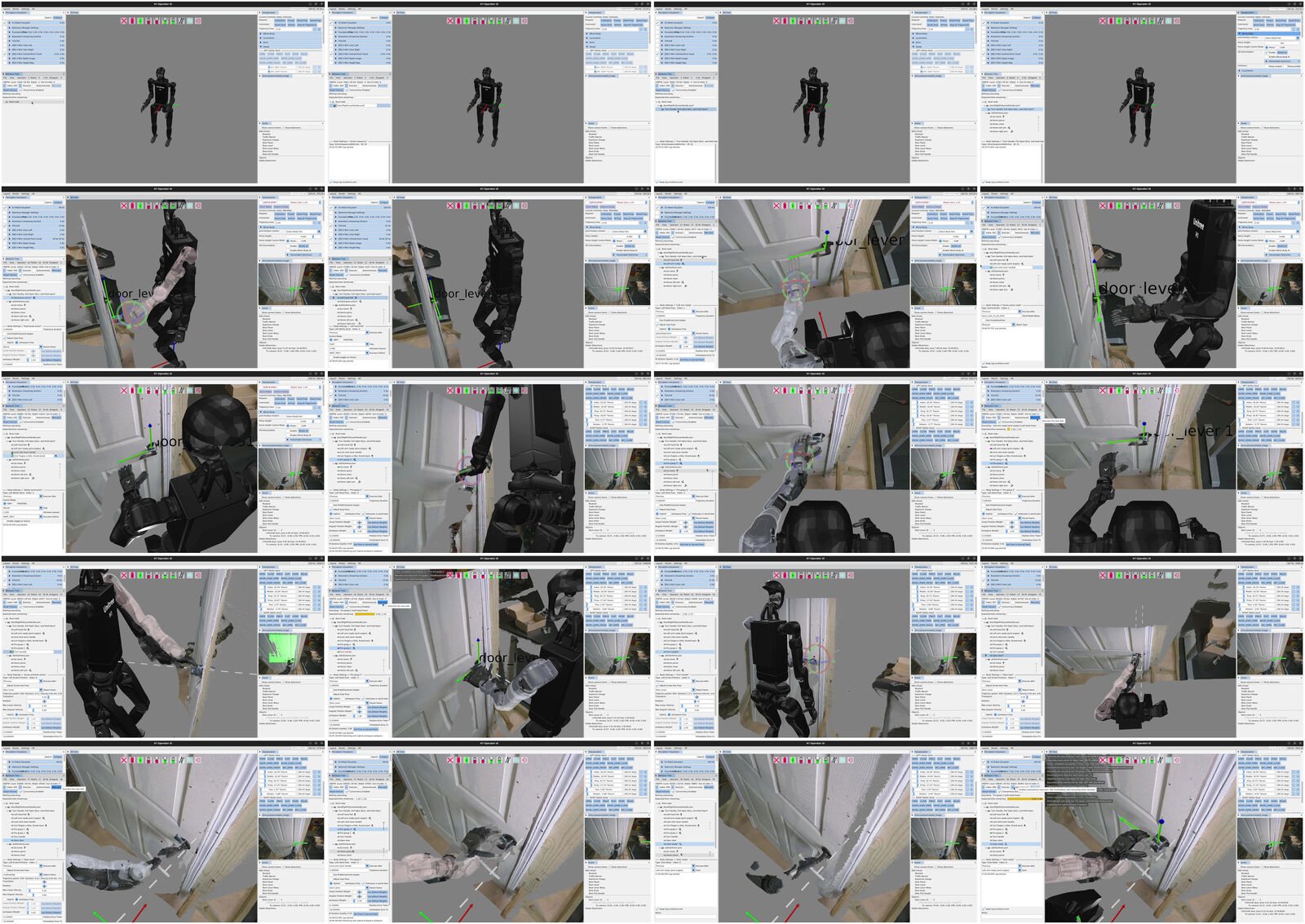}
    \caption{Key frames from the Unitree H1-2 scratch authoring session on January 2, 2026, from sequence creation through screw primitive tuning to looping autonomous opening.
    A video is available at \url{https://youtu.be/7VGFufJWaR4}.}
    \label{fig:h2_unitree_scratch_authoring}
\end{figure}

On January 2, 2026, we authored a looping standing right pull lever handle opening behavior on the Unitree H1-2 from an empty behavior tree.
The session reached first autonomous looping execution in 31 minutes and 43 seconds of active authoring time.
The progression is shown in \autoref{fig:h2_unitree_scratch_authoring}.
Authoring times are presented in \autoref{tab:h2_unitree_scratch_authoring_timeline}.

\begin{table}[H]
    \caption[Unitree scratch right pull authoring milestones.]{Scratch authoring milestones for the January~2,~2026 Unitree H1-2 looping right pull lever handle opening behavior shown in \autoref{fig:h2_unitree_scratch_authoring}.}
    \centering
    \footnotesize
    \renewcommand{\arraystretch}{0.97}
    \begin{tabular}{r p{10.7cm}}
        \hline
        Time & Authoring milestone \\
        \hline
        1:15 & Sequence \texttt{door/RightPullLeverHandle.json} created. \\
        6:26 & Ability hand and arm pose actions created. \\
        8:28 & Scene action node \texttt{Lock onto lever handle} added. \\
        14:45 & \texttt{Pre-grasp 1} arm action tuned. \\
        16:48 & Pre-grasps complete; handle-turn screw primitive created. \\
        27:29 & Handle-turn screw primitive tuned; \texttt{Open door} screw primitive created. \\
        30:28 & \texttt{Pre-grasp 1} set to execute concurrently with finger curl for speed. \\
        31:27 & Goto action added at the end to loop the opening sequence indefinitely. \\
        31:43 & Behavior set to autonomous execution; repeated-run testing begins. \\
        \hline
    \end{tabular}
    \label{tab:h2_unitree_scratch_authoring_timeline}
\end{table}

The first autonomous looping opening came at 31 minutes 43 seconds (\autoref{tab:h2_unitree_scratch_authoring_timeline}).
The goto loop at 31:27 immediately began the 32/32 repeated opening test in \autoref{fig:unitree_pull_opening_reliability_20260102}.
This session shows the diagnose, edit, and retest loop in one sitting, which is what hypothesis 2 claims relative to redeploy or retrain workflows.

\subsection{First Right Pull Door Bring Up (Alex, 2026-01-20 to 24)}

\begin{figure}[H]
    \centering
    \includegraphics[width=0.95\columnwidth]{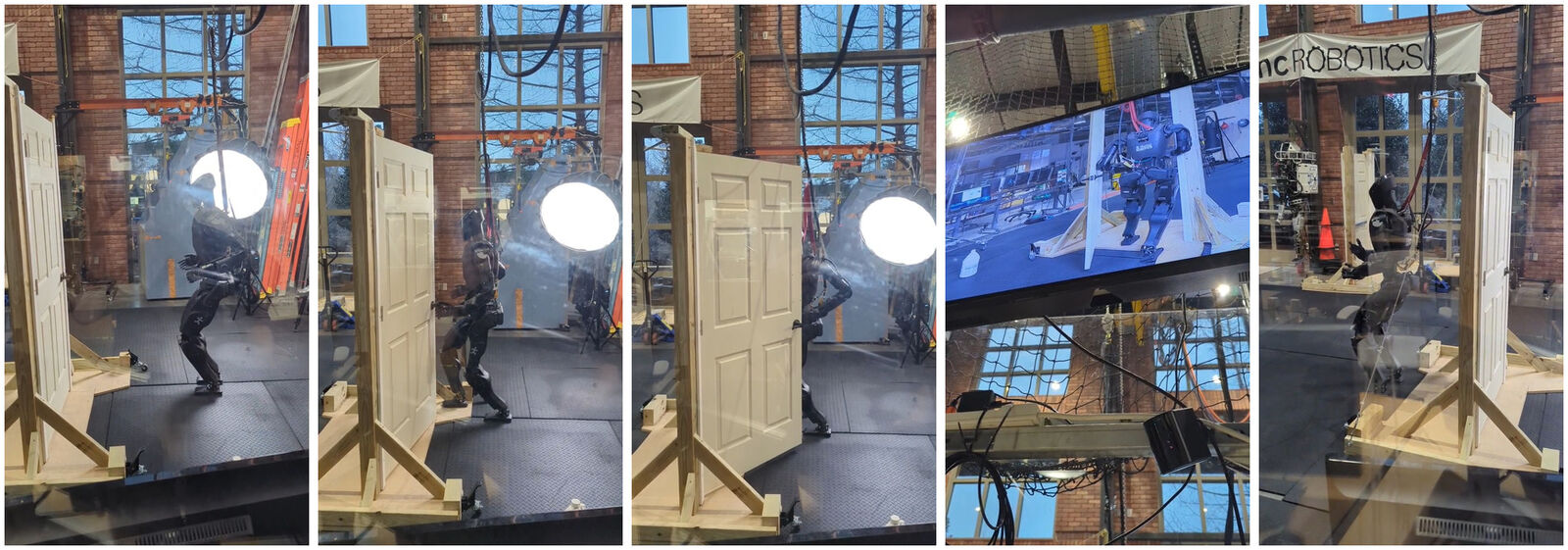}
    \caption{Alex during the January 20--24, 2026 first right pull door bring up, culminating in the first successful full traversal.
    A video is available at \url{https://youtu.be/zYYnatMz3wU}.}
    \label{fig:h2_alex_first_pull_authoring}
\end{figure}

From January 20 through 24, 2026, we brought up the first full Alex pull lever door traversal from scratch.
The measured active authoring total is 11 hours, 10 minutes, and 17 seconds.
That total is cumulative across five daily sessions and excludes inactive gaps between sessions.
A frame from the bring up period is shown in \autoref{fig:h2_alex_first_pull_authoring}.
Authoring times are presented in \autoref{tab:h2_alex_first_pull_authoring_timeline}.

\begin{table}[H]
    \caption[First Alex right pull authoring milestones.]{Normalized authoring timeline for the January~20--24,~2026 first Alex right pull lever handle door traversal shown in \autoref{fig:h2_alex_first_pull_authoring}.
    The time column is cumulative across days but excludes inactive gaps between daily sessions.}
    \centering
    \footnotesize
    \setlength{\tabcolsep}{4pt}
    \renewcommand{\arraystretch}{0.97}
    \begin{tabular}{>{\RaggedRight\arraybackslash}p{1.4cm} >{\RaggedRight\arraybackslash}p{1.95cm} >{\RaggedRight\arraybackslash}p{\dimexpr\textwidth-3.35cm-6\tabcolsep\relax}}
        \hline
        Day & Time & Authoring milestone \\
        \hline
        Day~1 & 0:00:00 & Create \texttt{door/RightPullLeverHandle.json}. \\
        Day~1 & 0:32:59 & Closed unlatch-and-open loop from a squared-up stance. \\
        Day~2 & 1:02:50 & Approach sequence created; footstep-planning issues encountered. \\
        Day~3 & 1:27:57 & Approach finalized; pre-grasp work begins. \\
        Day~3 & 3:15:35 & Knee-clearance check against the swinging panel. \\
        Day~3 & 3:57:05 & First open from a staggered approach stance. \\
        Day~4 & 5:32:44 & Pull-and-hold sequence begins. \\
        Day~4 & 6:14:21 & Door opens farther but not fully by end of session. \\
        Day~5 & 7:52:57 & Door fully open; first traversal step attempted. \\
        Day~5 & 9:52:47 & Full traversal footstep plan authored. \\
        Day~5 & 10:54:06 & Hold-open fallback added after force-related trouble. \\
        Day~5 & 11:01:58 & Pre-authored \texttt{Pull right arm in} clearance pose added. \\
        Day~5 & 11:10:17 & First successful full pull-door traversal. \\
        \hline
    \end{tabular}
    \label{tab:h2_alex_first_pull_authoring_timeline}
\end{table}

The 11 hour 10 minute total spans five days of incremental trial-and-error authoring (\autoref{tab:h2_alex_first_pull_authoring_timeline}).
Major milestones include the first open from a staggered stance on day 3 and the hold open fallback added on day 5 after force related trouble.
These Alex authoring sessions show the cost of bringing up the first door traversal behavior on a new hardware platform.

\subsection{Scratch Left Push Door (Alex, 2026-02-22)}

\begin{figure}[H]
    \centering
    \includegraphics[width=0.95\columnwidth]{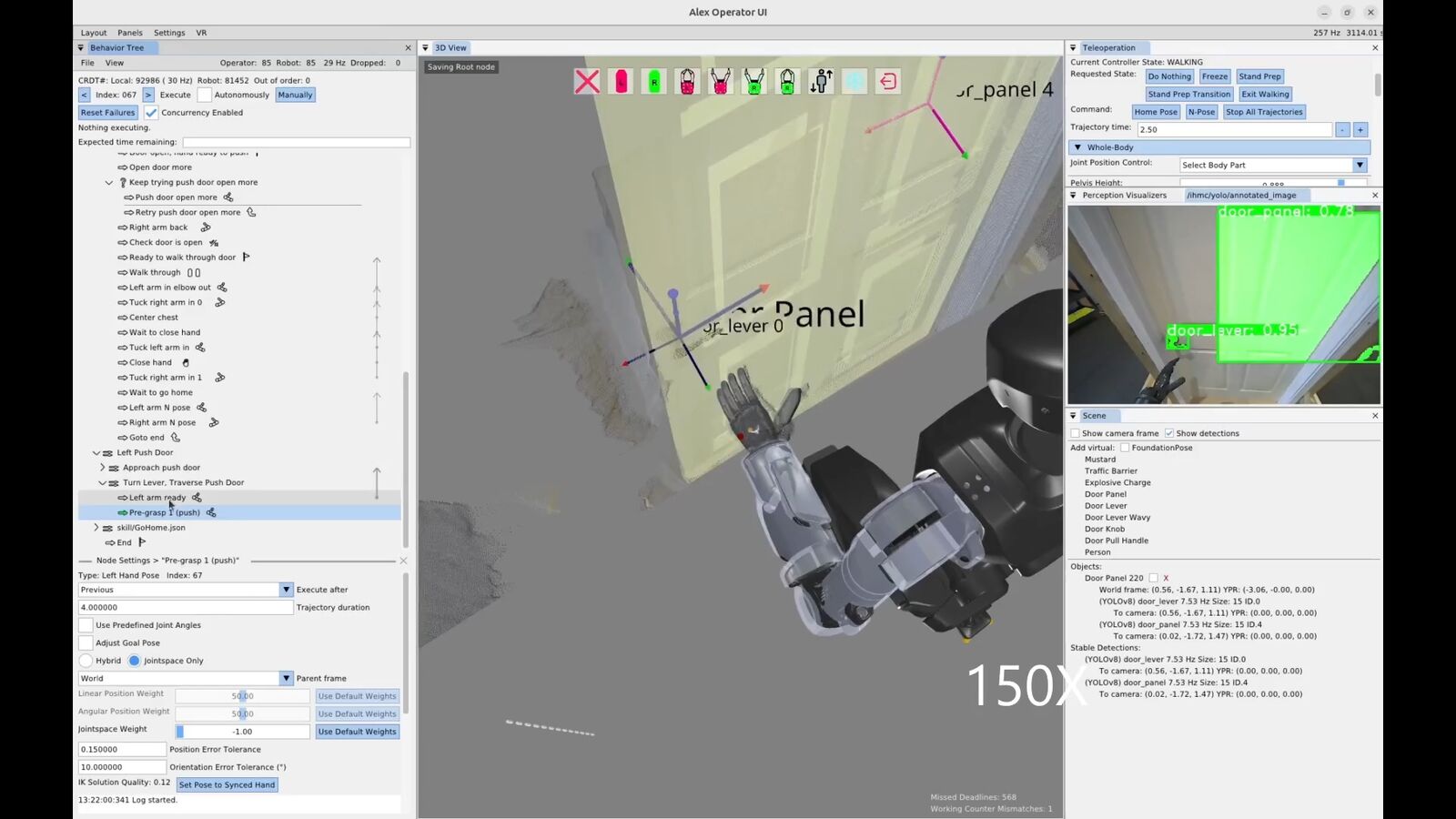}
    \caption{Screenshot from the February 22, 2026 scratch left push door authoring session on Alex.
    A video is available at \url{https://youtu.be/bLHQmV4GEFA}.}
    \label{fig:h2_scratch_left_push_authoring}
\end{figure}

On February 22, 2026, we authored a left push door traversal on Alex from an empty \texttt{Left Push Door} sequence to first fully automatic success in 1 hour, 59 minutes, and 48 seconds of active authoring.
This is the measured scratch case used in the comparative authoring process figure.
The session is shown in \autoref{fig:h2_scratch_left_push_authoring}.
Authoring times are presented in \autoref{tab:h2_scratch_left_push_authoring_timeline}.

\begin{table}[H]
    \caption[Scratch left push door authoring milestones.]{Normalized scratch authoring timeline for the February~22,~2026 left push door behavior shown in \autoref{fig:h2_scratch_left_push_authoring}.}
    \centering
    \footnotesize
    \setlength{\tabcolsep}{4pt}
    \renewcommand{\arraystretch}{0.97}
    \begin{tabular}{>{\RaggedLeft\arraybackslash}p{1.9cm} >{\RaggedRight\arraybackslash}p{\dimexpr\textwidth-1.9cm-6\tabcolsep\relax}}
        \hline
        Elapsed time & Authoring milestone \\
        \hline
        0:00:00 & Create the \texttt{Left Push Door} sequence. \\
        0:14:33 & Door-panel scene action and squared-up pre-approach stance added. \\
        0:24:22 & Staggered approach stance executed. \\
        1:21:14 & First handle turn. \\
        1:24:37 & Door opened. \\
        1:39:21 & Right arm pushes door fully open. \\
        1:49:24 & Full traversal footstep set authored. \\
        1:50:11 & Traversal taken; arm task-space error sends the right arm backward. \\
        1:51:13 & Right-arm fix: bring the arm in for traversal. \\
        1:58:15 & \texttt{shape contains points} stop condition added if the door does not open. \\
        1:59:48 & First fully automatic push-door success. \\
        \hline
    \end{tabular}
    \label{tab:h2_scratch_left_push_authoring_timeline}
\end{table}

This is our measured fully autonomous scratch case for literature comparison: 1 hour 59 minutes 48 seconds from empty sequence to first automatic success (\autoref{tab:h2_scratch_left_push_authoring_timeline}).
Failure at 1:50:11 from arm task space error was fixed in about 9 minutes by bringing the right arm in using jointspace for traversal.
That short edit and retest loop is the kind of workflow \autoref{fig:hero_authoring_process_comparison} contrasts against multi day retraining pipelines.

Across the four rows in \autoref{fig:in_house_authoring_from_scratch_timeline}, measured active authoring ranged from ~30 minutes to ~11 hours.
The 33 minute session from 2023 on Nadia resulted in a behavior that was only partially autonomous, while the February 22, 2026 Alex left push row is the fairest fully autonomous scratch comparison at just under 2 hours.
Each session documented failure diagnosis, logic edits, and retesting on the robot without a redeploy or retraining cycle.

\section{Hypothesis 3: Fast Adaptation, Extension, and Combination of Behaviors}
\label{sec:h3_fast_adaptation}

To support our third hypothesis, we'll show measured sessions where existing behaviors were adapted, composed, or extended rather than rebuilt from scratch.

\begin{figure}[H]
    \centering
    {\singlespacing
    \resizebox{\columnwidth}{!}{%
        \newcommand{\inadaptlabel}[2]{#1\\[-2.4pt]{\fontsize{5.5}{6.2}\selectfont #2}}
\newcommand{\inadaptrowlabel}[3]{%
  \node[rowlbl, anchor=west] at (\systemleft + 0.02, #1) {\inadaptlabel{#2}{#3}};}
\newcommand{\inadaptnum}[3]{%
  \pgfmathparse{#2 > 9.8 ? 1 : 0}%
  \ifnum\pgfmathresult=1
    \node[numlbl, anchor=east] at (\numright - 0.10, #1) {#3};%
  \else
    \pgfmathsetmacro{\inadaptnumx}{#2 + 0.08 - \adaptbarshift}%
    \node[numlbl, anchor=west] at (\inadaptnumx, #1) {#3};%
  \fi}
\newcommand{\adaptbarphase}[5]{%
  \filldraw[#5, draw=black, line width=0.28pt]
    (#3, {#1 - #2}) rectangle (#4, {#1 + #2});}
\newcommand{\adaptbarsuccess}[2]{%
  \node[draw=black, line width=0.28pt, fill=green!55!black, rounded corners=0.6pt, inner xsep=0.8pt, inner ysep=0.4pt, font=\fontsize{6.0}{6.4}\selectfont\bfseries, text=white] at (#2, #1) {$\checkmark$};}
\newcommand{\inadaptswatch}[3]{%
  \filldraw[#1, rounded corners=0.8pt] (#2, {\inlegendy - 0.055}) rectangle ++(0.22, 0.11);%
  \node[legendtxt, anchor=west] at ({#2 + 0.28}, \inlegendy) {#3};}

\begin{tikzpicture}[
    x=1cm,
    y=1cm,
    every node/.style={outer sep=0pt},
    rowlbl/.style={font=\fontsize{8.2}{9.4}\selectfont, anchor=west, align=left, text width=4.20cm},
    numlbl/.style={font=\fontsize{8.2}{8.8}\selectfont, anchor=west, align=left, text width=0.82cm},
    bandtitle/.style={font=\footnotesize\bfseries, anchor=north west, inner xsep=1.2pt, inner ysep=0.8pt},
    legendtxt/.style={font=\fontsize{6.8}{7.4}\selectfont}
  ]
  \def\barh{0.20}

  \def\systemw{4.25}
  \def\barw{6.35}
  \def\numw{0.82}
  \def\colgap{0.02}
  \def\headerh{1.20}
  \def\rowh{0.56}
  \def\bandtitleh{0.54}
  \def\bandpadbottom{0.10}
  \def\bottomregionh{2.55}

  \pgfmathsetmacro{\systemleft}{0}
  \pgfmathsetmacro{\systemright}{\systemleft + \systemw}
  \pgfmathsetmacro{\barleft}{\systemright + \colgap}
  \pgfmathsetmacro{\barright}{\barleft + \barw}
  \pgfmathsetmacro{\numleft}{\barright + \colgap}
  \pgfmathsetmacro{\numright}{\numleft + \numw}
  \pgfmathsetmacro{\figurewidth}{\numright}

  \pgfmathsetmacro{\bandch}{\bandtitleh + 3 * \rowh + \bandpadbottom}
  \pgfmathsetmacro{\figureheight}{\headerh + \bandch + \bottomregionh}

  \pgfmathsetmacro{\headertop}{\figureheight}
  \pgfmathsetmacro{\headerbottom}{\headertop - \headerh}
  \pgfmathsetmacro{\bandctop}{\headerbottom}
  \pgfmathsetmacro{\bandcbottom}{\bandctop - \bandch}

  \pgfmathsetmacro{\rowbottle}{\bandctop - \bandtitleh - 0.5 * \rowh}
  \pgfmathsetmacro{\rowbreak}{\rowbottle - \rowh}
  \pgfmathsetmacro{\rowtwotable}{\rowbreak - \rowh}

  \def\adaptbarshift{0.30}

  \pgfmathsetmacro{\adaptaxisleft}{\barleft + 0.04}
  \pgfmathsetmacro{\adaptaxisright}{\barright - 0.04}
  \pgfmathsetmacro{\axisy}{\bandcbottom - 0.36}
  \def\legendentryh{0.11}
  \def\legendrowpitch{0.30}
  \pgfmathsetmacro{\axistitley}{\axisy - 0.48}
  \pgfmathsetmacro{\axistitlebottom}{\axistitley - 0.18}
  \pgfmathsetmacro{\legendgapbelowtitle}{0.48}
  \pgfmathsetmacro{\inlegendrowA}{\axistitlebottom - \legendgapbelowtitle - 0.5 * \legendentryh}
  \pgfmathsetmacro{\inlegendrowB}{\inlegendrowA - \legendrowpitch}
  \pgfmathsetmacro{\inlegendrowC}{\inlegendrowB - \legendrowpitch}
  \pgfmathsetmacro{\durationheaderx}{\adaptaxisleft - 0.28}
  \def\inadaptlegendcolA{0.10}
  \def\inadaptlegendcolB{3.95}
  \def\inadaptlegendcolC{7.80}

  \path[draw=black!30, fill=blue!8, rounded corners=4pt, line width=0.8pt]
    (0, \bandcbottom) rectangle (\figurewidth, \bandctop);

  \node[anchor=west, font=\footnotesize\bfseries] at (\systemleft + 0.02, \headerbottom + 0.48) {Behavior / session};
  \node[anchor=east, font=\footnotesize\bfseries] at (\numright - 0.02, \headerbottom + 0.48) {Total};

  \inadaptrowlabel{\rowbottle}{Bottle Carry Composition}{2026-03-18}
  \inadaptrowlabel{\rowbreak}{Break Room Door Adaptation}{2026-03-26}
  \inadaptrowlabel{\rowtwotable}{Two Table Sorting Extension}{2026-04-13--14}

  \begin{scope}[shift={(-\adaptbarshift,0)}]
  \node[anchor=west, font=\footnotesize\bfseries] at (\durationheaderx, \headerbottom + 0.48) {Adaptation duration (log time)};

  \foreach \tickx in {4.3100,6.4000,8.4900,10.5800} {
    \draw[draw=black!12, line width=0.4pt] (\tickx, \axisy) -- (\tickx, \bandctop - 0.04);
  }

  \adaptbarphase{\rowbottle}{\barh}{\adaptaxisleft}{6.0749}{orange!70}
  \adaptbarphase{\rowbottle}{\barh}{6.0749}{7.4796}{violet!60}
  \adaptbarphase{\rowbottle}{\barh}{7.4796}{8.4231}{violet!60}
  \adaptbarphase{\rowbottle}{\barh}{8.4231}{8.4766}{black!35}
  \adaptbarphase{\rowbottle}{\barh}{8.4766}{9.1149}{black!35}
  \adaptbarsuccess{\rowbottle}{9.1149}

  \adaptbarphase{\rowbreak}{\barh}{\adaptaxisleft}{5.7083}{violet!60}
  \adaptbarphase{\rowbreak}{\barh}{5.7083}{6.3731}{orange!70}
  \adaptbarphase{\rowbreak}{\barh}{6.3731}{6.5455}{violet!60}
  \adaptbarphase{\rowbreak}{\barh}{6.5455}{8.2176}{violet!60}
  \adaptbarphase{\rowbreak}{\barh}{8.2176}{9.2507}{black!35}
  \adaptbarphase{\rowbreak}{\barh}{9.2507}{10.2623}{black!35}
  \adaptbarphase{\rowbreak}{\barh}{10.2623}{10.4077}{green!55!black}
  \adaptbarphase{\rowbreak}{\barh}{10.4077}{10.4267}{black!35}
  \adaptbarsuccess{\rowbreak}{10.4267}

  \adaptbarphase{\rowtwotable}{\barh}{\adaptaxisleft}{4.4089}{cyan!40}
  \adaptbarphase{\rowtwotable}{\barh}{4.4089}{5.3150}{green!55!black}
  \adaptbarphase{\rowtwotable}{\barh}{5.3150}{5.4921}{violet!60}
  \adaptbarphase{\rowtwotable}{\barh}{5.4921}{6.1577}{violet!60}
  \adaptbarphase{\rowtwotable}{\barh}{6.1577}{7.2206}{violet!60}
  \adaptbarphase{\rowtwotable}{\barh}{7.2206}{7.3881}{green!55!black}
  \adaptbarphase{\rowtwotable}{\barh}{7.3881}{7.4679}{violet!60}
  \adaptbarphase{\rowtwotable}{\barh}{7.4679}{7.5365}{orange!70}
  \adaptbarphase{\rowtwotable}{\barh}{7.5365}{8.1480}{blue!45}
  \adaptbarphase{\rowtwotable}{\barh}{8.1480}{8.9049}{green!55!black}
  \adaptbarphase{\rowtwotable}{\barh}{8.9049}{9.0053}{black!35}
  \adaptbarphase{\rowtwotable}{\barh}{9.0053}{9.2552}{violet!60}
  \adaptbarphase{\rowtwotable}{\barh}{9.2552}{10.3363}{violet!60}
  \adaptbarsuccess{\rowtwotable}{10.3363}

  \draw[draw=black!55, line width=0.55pt] (\adaptaxisleft, \axisy) -- (\adaptaxisright, \axisy);
  \foreach \tickx/\ticklabel in {4.3100/15 min,6.4000/30 min,8.4900/1 h,10.5800/2 h} {
    \draw[draw=black!55, line width=0.55pt] (\tickx, \axisy - 0.06) -- (\tickx, \axisy + 0.06);
    \node[font=\scriptsize, anchor=north] at (\tickx, \axisy - 0.10) {\ticklabel};
  }
  \node[font=\scriptsize\bfseries, anchor=north] at ({0.5 * (\adaptaxisleft + \adaptaxisright)}, \axistitley) {Normalized active adaptation duration (log time)};
  \end{scope}

  \inadaptnum{\rowbottle}{9.1149}{1:13:49}
  \inadaptnum{\rowbreak}{10.4267}{1:54:03}
  \inadaptnum{\rowtwotable}{10.3363}{1:50:41}

  \foreach \rowy in {\rowbottle,\rowbreak} {
    \draw[draw=black!12, line width=0.4pt] (\barleft, {\rowy - 0.5 * \rowh}) -- (\figurewidth - 0.10, {\rowy - 0.5 * \rowh});
  }

  \node[bandtitle, fill=blue!8] at (0.12, \bandctop - 0.06) {Alex 2026};

  \pgfmathsetmacro{\inlegendy}{\inlegendrowA}
  \inadaptswatch{blue!45, draw=black, line width=0.28pt}{\inadaptlegendcolA}{Setup / structure}
  \inadaptswatch{green!55!black, draw=black, line width=0.28pt}{\inadaptlegendcolB}{Approach / locomotion}
  \inadaptswatch{violet!60, draw=black, line width=0.28pt}{\inadaptlegendcolC}{Opening / manipulation}

  \pgfmathsetmacro{\inlegendy}{\inlegendrowB}
  \inadaptswatch{black!35, draw=black, line width=0.28pt}{\inadaptlegendcolA}{Traversal / completion}
  \inadaptswatch{orange!70, draw=black, line width=0.28pt}{\inadaptlegendcolB}{Fix / recovery}
  \node[draw=black, line width=0.28pt, fill=green!55!black, rounded corners=0.6pt, inner xsep=0.8pt, inner ysep=0.4pt, font=\fontsize{6.0}{6.4}\selectfont\bfseries, text=white] at (\inadaptlegendcolC, \inlegendy) {$\checkmark$};
  \node[legendtxt, anchor=west] at ({\inadaptlegendcolC + 0.22}, \inlegendy) {First autonomous success};

  \pgfmathsetmacro{\inlegendy}{\inlegendrowC}
  \inadaptswatch{cyan!40, draw=black, line width=0.28pt}{\inadaptlegendcolA}{Off-robot pre-authoring}
\end{tikzpicture}%
    }
    \par
    }
    \caption{
        In-house real-robot adaptation durations on a log-second axis with internal milestone structure.
        Rows are grouped by robot era and ordered chronologically.
        The bottle session composes existing behaviors; the break-room session adapts a mirrored lab behavior to a real door; the two-table session extends an existing manipulation behavior with locomotion and routing logic.
        The cyan segment marks off-robot pre-authoring counted in the two-table total.
        Checkmarks mark first autonomous success.
        Times are normalized active adaptation.
    }
    \label{fig:in_house_adaptation_timeline}
\end{figure}
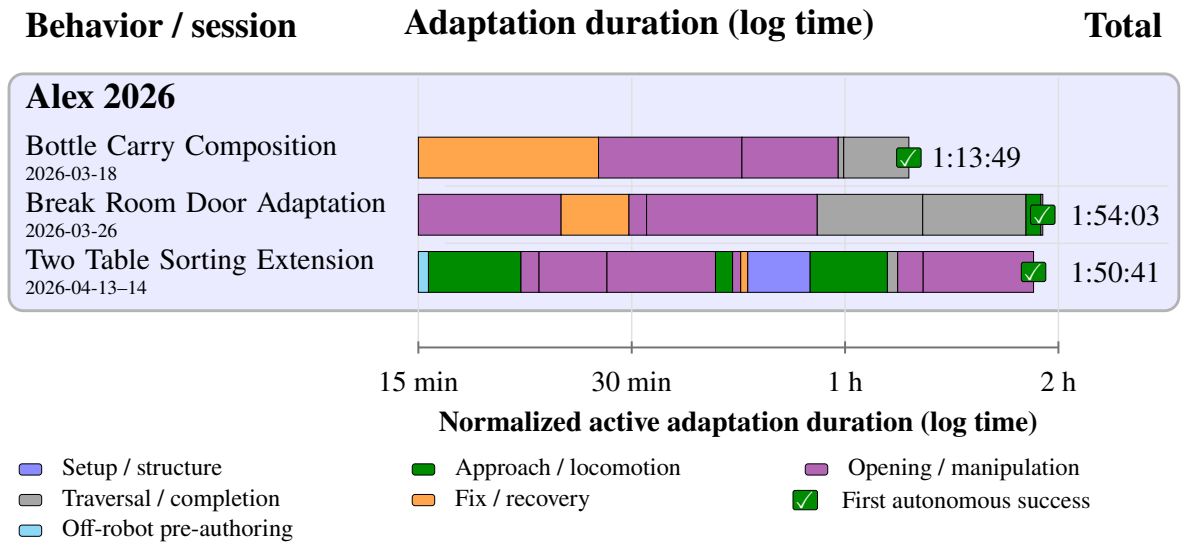

In \autoref{fig:in_house_adaptation_timeline}, we summarize three measured adaptation sessions on Alex.
Each row shows normalized active adaptation time on a log-second axis with internal milestone structure recovered from screen recordings.
The bottle carry-through session composes existing pickup and door behaviors into one long-horizon task.
The break-room session adapts a lab right-pull traversal to a real left-pull door: door-specific mirroring was only the starting transfer, and most time went to retuning opening, fallbacks, traversal, and footsteps.
The two-table sorting session extends the open-house ball-return behavior with locomotion, table-specific routing, and color-dependent branches; its total includes an off-robot pre-authoring pass.
Checkmarks mark first autonomous success.

The subsections below walk through each row in \autoref{fig:in_house_adaptation_timeline} with a representative still, the authoring-time table, and a link to the screen recording.

\subsection{Bottle Carry Through Composition (Alex, 2026-03-18)}

\begin{figure}[H]
    \centering
    \includegraphics[width=0.95\columnwidth]{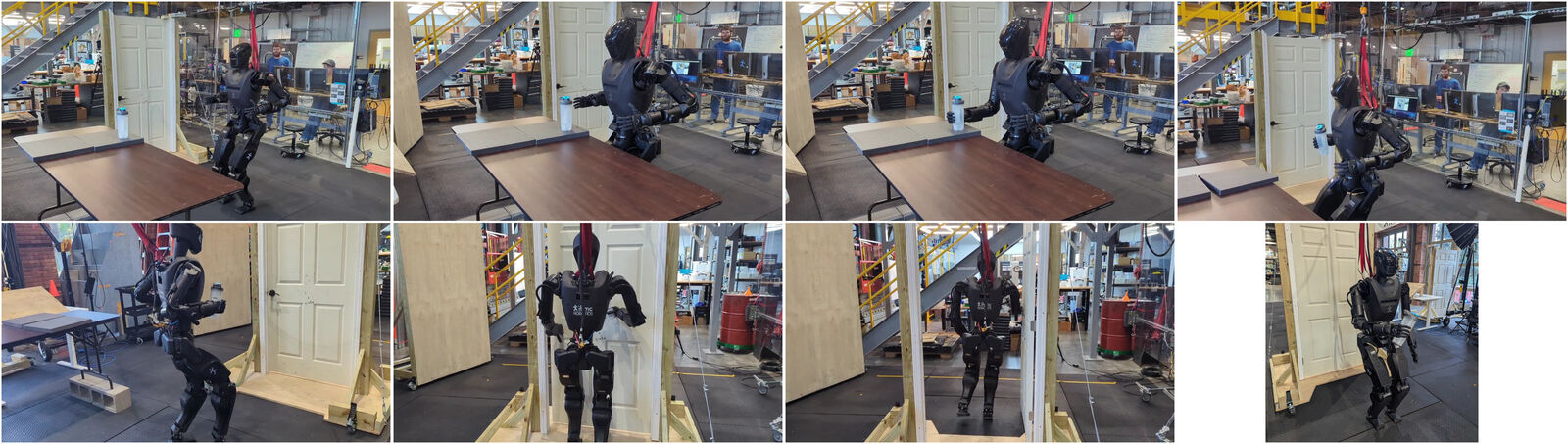}
    \caption{Alex after the first fully autonomous bottle pickup, walk, and door traversal with the bottle in hand on March 18--19, 2026.
    A video is available at \url{https://youtu.be/AH2dFMW-IbY}.}
    \label{fig:h3_bottle_carry_through_authoring}
\end{figure}

On March 18, 2026, we composed an existing bottle pickup behavior with the left push door traversal so Alex could carry a bottle through the door in one autonomous run.
Most of the session wired the two behaviors together, recovered from a tooling failure while loading the door subtree, and tuned grasp and traversal so the bottle stayed in hand.
First fully autonomous success came at 1 hour, 13 minutes, and 49 seconds of measured active adaptation time.
The result is shown in \autoref{fig:h3_bottle_carry_through_authoring}.
Authoring times are presented in \autoref{tab:h3_bottle_carry_through_authoring_timeline}.

\begin{table}[H]
    \caption[Bottle carry-through composition timeline.]{Cumulative adaptation timeline for the March~18,~2026 bottle carry-through composition session shown in \autoref{fig:h3_bottle_carry_through_authoring}.}
    \centering
    \footnotesize
    \setlength{\tabcolsep}{4pt}
    \renewcommand{\arraystretch}{0.97}
    \begin{tabular}{>{\RaggedLeft\arraybackslash}p{1.9cm} >{\RaggedRight\arraybackslash}p{\dimexpr\textwidth-1.9cm-6\tabcolsep\relax}}
        \hline
        Elapsed time & Composition milestone \\
        \hline
        0:00:00 & Start from existing bottle-pickup and left push door traversal behaviors; robot is at the table holding the bottle. \\
        0:01:34 & Backup-from-table and face-door walk action added. \\
        0:02:43 & Tooling issue while loading the large door subtree into the tree. \\
        0:26:56 & Door subtree loaded by manual JSON edit and UI/autonomy relaunch. \\
        0:42:55 & \texttt{Retry grasp bottle} fallback added with a near-hand point check. \\
        0:58:41 & Right-hand finger and arm motions disabled in the push-door behavior so the bottle grasp is preserved through traversal. \\
        0:59:44 & Door fully traversed with the bottle, but discontinuously. \\
        1:13:49 & First fully autonomous run: approach, pickup, walk, traverse with bottle in hand. \\
        \hline
    \end{tabular}
    \label{tab:h3_bottle_carry_through_authoring_timeline}
\end{table}

Most of the 1 hour 13 minute 49 second session went to wiring existing pickup and door subtrees and preserving the bottle grasp through traversal (\autoref{tab:h3_bottle_carry_through_authoring_timeline}).
We did not author a new door behavior from scratch.
This supports hypothesis 3 because the work edited connections and parameters between existing behaviors rather than rebuilding either subtree.

\subsection{Break Room Door Adaptation (Alex, 2026-03-26)}

\begin{figure}[H]
    \centering
    \includegraphics[width=0.95\columnwidth]{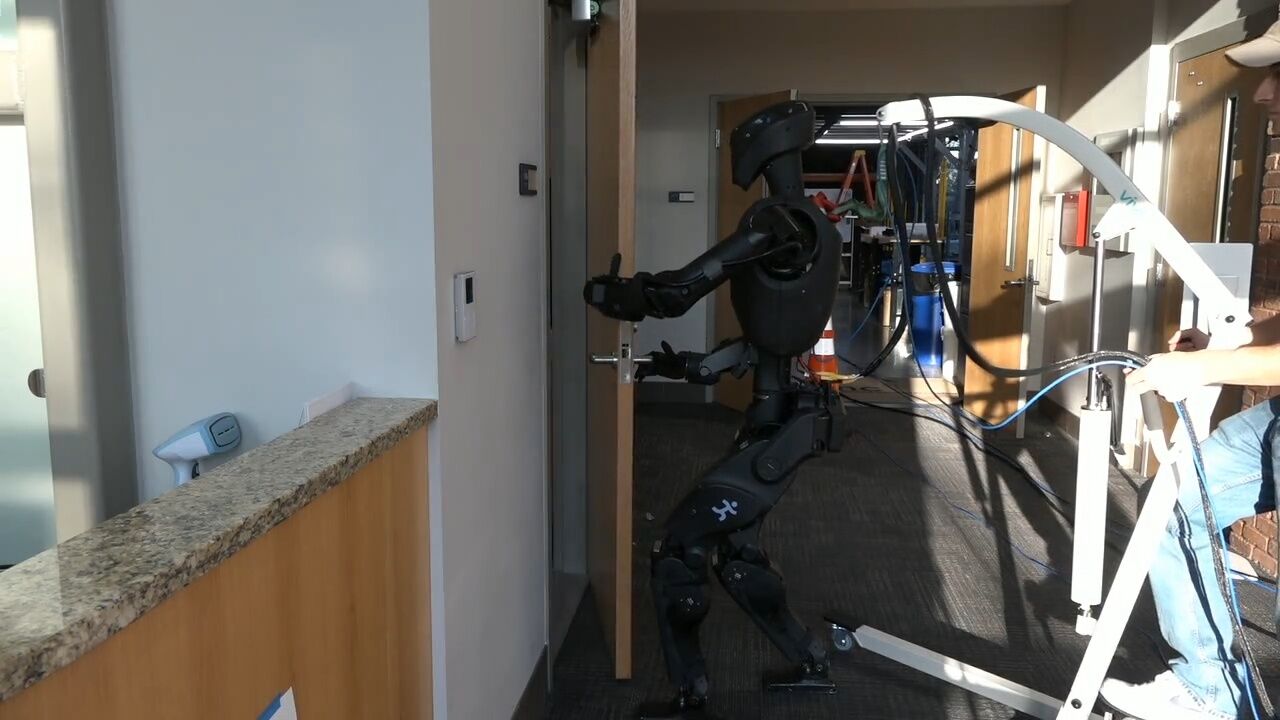}
    \caption{Alex opening the break room left pull door on March 26, 2026 after mirrored adaptation from the lab right pull behavior.
    A video is available at \url{https://youtu.be/HcJPLoRfs3Q}.}
    \label{fig:h3_break_room_adaptation}
\end{figure}

On March 26, 2026, we adapted the lab right pull door traversal to the break room left pull door on Alex.
Door mirroring was only the starting point.
Most of the 1 hour, 54 minutes, and 3 seconds of measured active adaptation time went to retuning opening, fallbacks, traversal, and footsteps on the real door.
The successful run is shown in \autoref{fig:h3_break_room_adaptation}.
Authoring times are presented in \autoref{tab:h3_break_room_adaptation_timeline}.

\begin{table}[H]
    \caption[Break-room adaptation timeline.]{Continuous authoring timeline for the March~26,~2026 break room door adaptation shown in \autoref{fig:h3_break_room_adaptation}.}
    \centering
    \footnotesize
    \setlength{\tabcolsep}{4pt}
    \renewcommand{\arraystretch}{0.97}
    \begin{tabular}{>{\RaggedLeft\arraybackslash}p{1.9cm} >{\RaggedRight\arraybackslash}p{\dimexpr\textwidth-1.9cm-6\tabcolsep\relax}}
        \hline
        Elapsed time & Adaptation milestone \\
        \hline
        0:00:33 & Load \texttt{DoorTraversal.json} (mirrored left pull, untested on a real door). \\
        0:01:56 & Door type and door-relative poses identified from scene objects. \\
        0:23:51 & First real-door opening after retuning the lever-turn motion. \\
        0:29:44 & Point-check fallback added to retry door opening when the pull is unreliable. \\
        0:31:29 & Door fully opened. \\
        0:54:49 & Spine yaw added after the lever turn for consistent opening. \\
        1:17:13 & Traversal arm motions simplified under the no-spring-closer assumption. \\
        1:48:00 & Traversal footsteps re-authored; shoulder-frame clearance checked in Preview Mode. \\
        1:53:20 & Reset and run the behavior autonomously from the top. \\
        1:54:03 & First successful break-room door traversal. \\
        \hline
    \end{tabular}
    \label{tab:h3_break_room_adaptation_timeline}
\end{table}

Our behavior mirroring function supplied the starting tree, but 1 hour 54 minutes 3 seconds went to retuning on the real left pull door (\autoref{tab:h3_break_room_adaptation_timeline}), which was more difficult to open and did not have a spring closer.
Point check fallbacks, spine yaw, and re-authored footsteps were the main edits rather than a new behavior file.
This supports hypothesis 3 because most of the effort went into localized changes inside an existing traversal behavior.

\subsection{Two Table Sorting Extension (Alex, 2026-04-13 to 14)}

\begin{figure}[H]
    \centering
    \includegraphics[width=0.95\columnwidth]{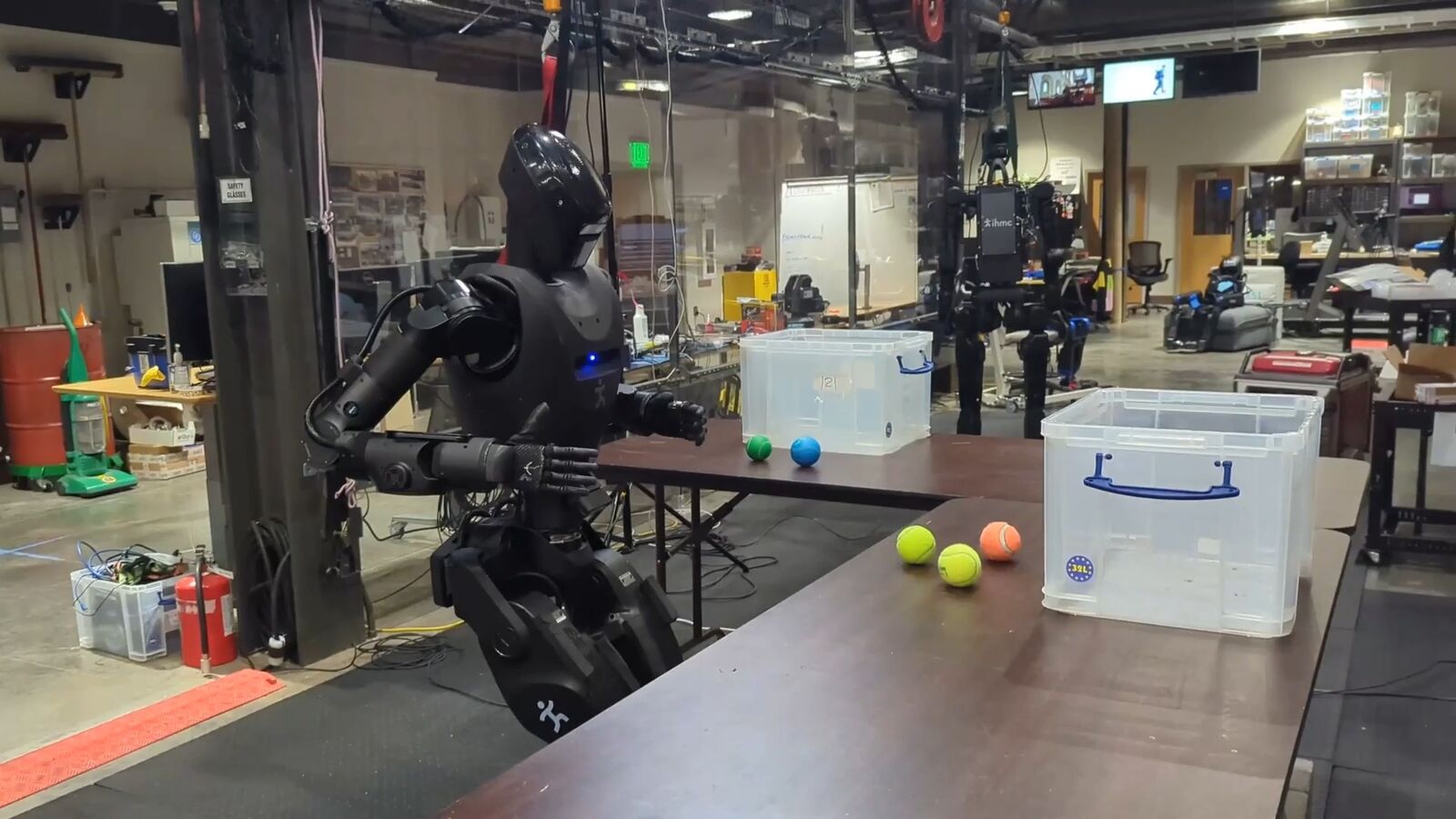}
    \caption{The two table ball sorting task on April 14, 2026.
    A video is available at \url{https://youtu.be/_3lqvcd_5WE}.}
    \label{fig:h3_two_table_sorting}
\end{figure}

From April 13 through 14, 2026, we extended the open house ball return behavior to sort colored balls between two tables with locomotion between stations.
The measured total is 1 hour, 50 minutes, and 41 seconds, including 15 minutes and 30 seconds of off-robot pre-authoring before the real-robot follow-up session.
The documented demonstration sorted nine of nine balls correctly, with three table approaches and two transitions between tables.
A still from the task setup is shown in \autoref{fig:h3_two_table_sorting}.
Authoring times are presented in \autoref{tab:h3_two_table_sorting_authoring_summary}.

\begin{table}[H]
    \caption[Two-table sorting extension authoring summary.]{Authoring summary for the April~13--14,~2026 two table sorting extension shown in \autoref{fig:h3_two_table_sorting}.}
    \centering
    \footnotesize
    \setlength{\tabcolsep}{4pt}
    \renewcommand{\arraystretch}{0.97}
    \begin{tabular}{>{\RaggedRight\arraybackslash}p{3.2cm} >{\RaggedRight\arraybackslash}p{2.2cm} >{\RaggedRight\arraybackslash}p{\dimexpr\textwidth-5.4cm-6\tabcolsep\relax}}
        \hline
        Phase & Duration & Milestone \\
        \hline
        Off-robot pre-authoring & 15~min~30~s & Fork \texttt{demo/BallsInBins.json} to \texttt{eval/TwoTableSorting.json}; restructure loop and color routing. \\
        Real-robot follow-up & 1~h~35~min~11~s & Reach documented 9/9 two-table sorting demonstration. \\
        Total measured & 1~h~50~min~41~s & First successful table-B placement at 26~min~8~s within the real-robot session. \\
        \hline
    \end{tabular}
    \label{tab:h3_two_table_sorting_authoring_summary}
\end{table}

The extended behavior reused \texttt{demo/BallsInBins.json} and added routing logic in 15 minutes 30 seconds off robot, followed by 1 hour 35 minutes on real robot (\autoref{tab:h3_two_table_sorting_authoring_summary}).
The documented execution run sorted 9/9 balls with walks between tables in 2 minutes 8 seconds (\autoref{tab:alex_two_table_sorting_event_timeline}).
This supports hypothesis 3 because locomotion and ball color-based branches were added to an existing sorting behavior instead of authoring a new one from an empty tree.

\section{Comparative Analysis}
\label{sec:comparative_analysis}

In this section, we'll compare our work with others in the literature across speed, capability, reliability, resilience, and the speed of creating and adapting behaviors.
The per run analyses above supply the measured evidence; here we place those numbers beside classical and learned references.

\subsection{Speed and Capability}

\begin{figure}[H]
\centering
{\singlespacing
\input{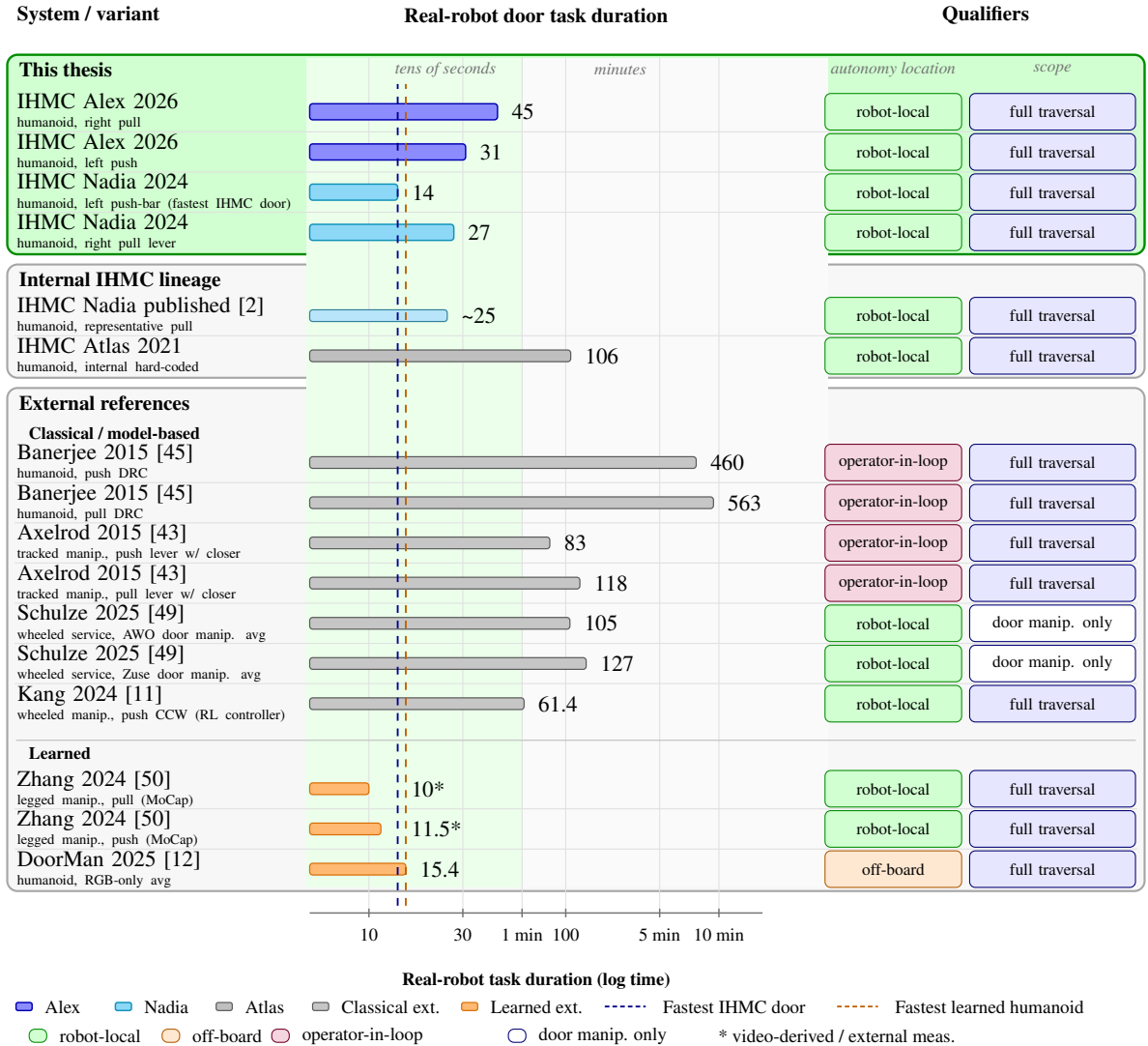}%
\par
}
\caption{Literature comparison of real-robot door task duration on a log-second axis.}
\label{fig:hero_speed_comparison}
\end{figure}

\autoref{fig:hero_speed_comparison} presents a comparison of the durations of robot door traversals across this thesis and the literature.
This represents every timed door traversal in the literature that we are aware of for both wheeled and legged robots.
Our door traversals are dramatically faster than the classical model-based results, all of which take more than one minute to traverse a door.
However, recent learned systems are quite fast.
Zhang et al.~\cite{zhang2024learningopentraversedoors}, with their ANYmal quadruped robot, demonstrate traversing a door in just 10 seconds.
DoorMan~\cite{xue2025opening} averages 15.4 seconds per door traversal.
However, as highlighted by the dotted line in the figure, we have one door traversal that executed in 14 seconds, just beating DoorMan's average.
This firmly places our system in the ``competitive with learned systems'' regime.

We think one of the ways Zhang et al achieves such a high speed is through their robot's kinematics.
They have a single, long arm mounted on the front of their quadruped robot which also has a high yaw range of motion that is something like 270 degrees or more.
These design elements give the robot a massive reachability workspace.
This allows it to open doors from a distance and swing the door panels all the way open in one sweeping motion.

There are two important caveats with the learned systems, however.
Zhang et al.~used external motion capture and fiducial markers rather than robot vision.
Xue et al.~(DoorMan) used an off-board computer to run their vision based neural inference.
This is in contrast to our system which uses only two color cameras on the robot for vision and all autonomous functionality is computed on board.

\subsection{Reliability and Resilience}

\begin{figure}[H]
\centering
\vspace*{-1.55cm}
\input{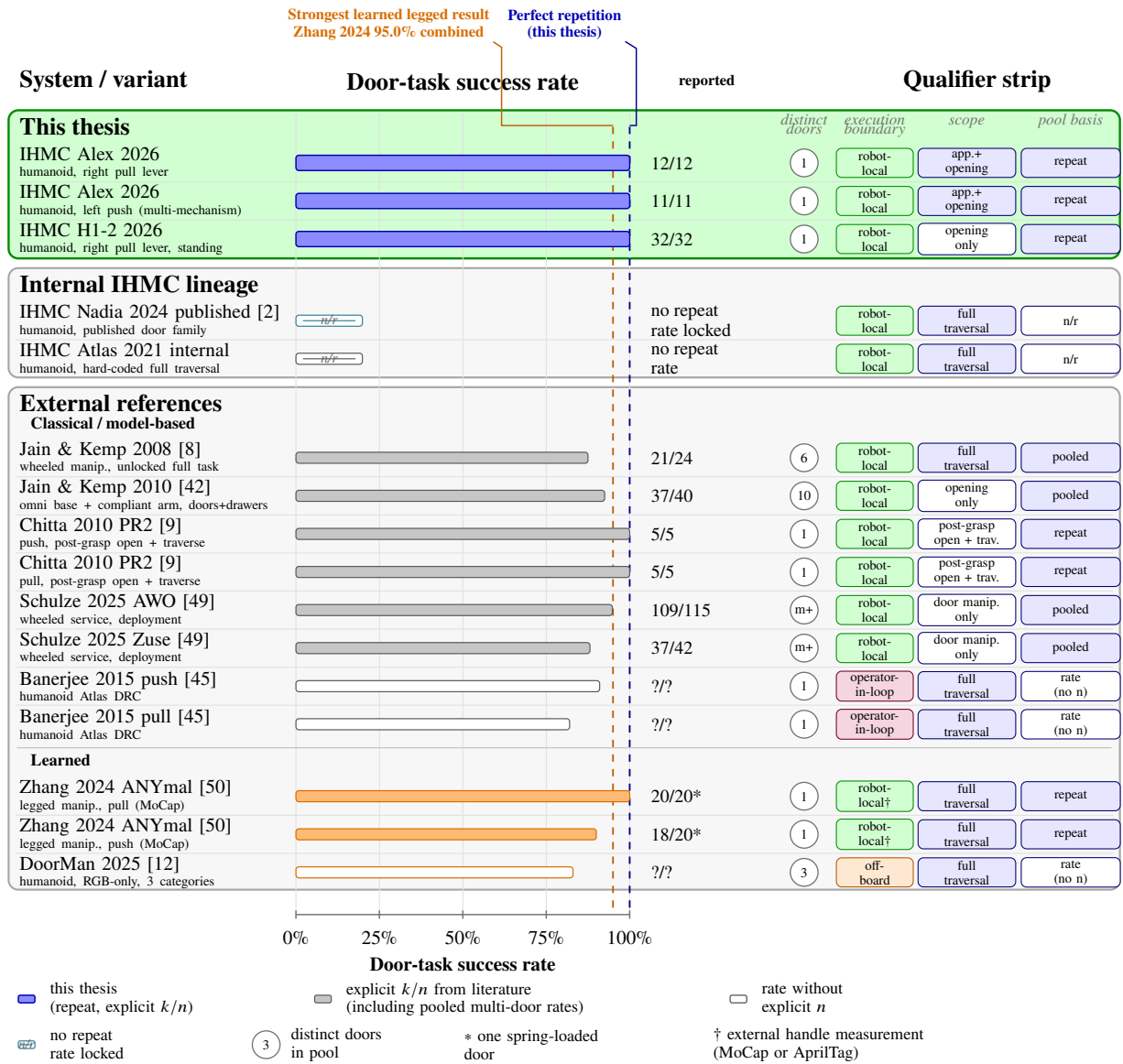}
\caption{Literature comparison of real-robot door-task success rate.}
\label{fig:hero_reliability_comparison}
\end{figure}

\autoref{fig:hero_reliability_comparison} presents a comparison of our results against all door traversal reliability measurements that appear in the literature that we are aware of.
Our premiere reliability tests were not full traversals, but were loco-manipulation behaviors.
We did not measure repeated full door traversals because our walk through component was not reliable in the time we had to test it.
Testing it could have damaged the robot beyond repair, preventing some of our more important results.
Although this is unfortunate for the full reliability comparison, we do feel that with a repeated approach and opening task, that we still show that our behavior architecture, perception, and manipulation control is capable of very high reliability.
The unreliability of the door walk through had more to do with the walking controller performance than the work presented in this thesis.

Most of the literature is in the same performance envelope of our repeated run reliability tests.
Although we reported no failures, this does not mean our system is 100\% reliable and it is not.
A similar statement can be made for the results in the literature.
Therefore, these repeated run measurements should be taken with a grain of salt, but they do serve as an indication that these systems are not super brittle ``one-off'' runs.

Among the learned systems, DoorMan~\cite{xue2025opening} actually stands out for saying ``83\%'' reliability without giving the number of attempts or a summary of what went into the calculation of that number.
It is for that reason that we put ``?/?'' in the diagram and leave out that percentage.
We do not feel like it is a reliable comparison to our thesis and the literature.


\vspace{2cm}

\subsection{Creation and Adaptability}

This thesis is essentially unique in its focus on reporting behavior authoring duration.
We focus on both the effort required to create new behaviors from scratch and modify existing ones, adapting them to new tasks.
We didn't find any documented results in the literature on the authoring durations of loco-manipulation behaviors on robots.

\begin{figure}[H]
\centering
\input{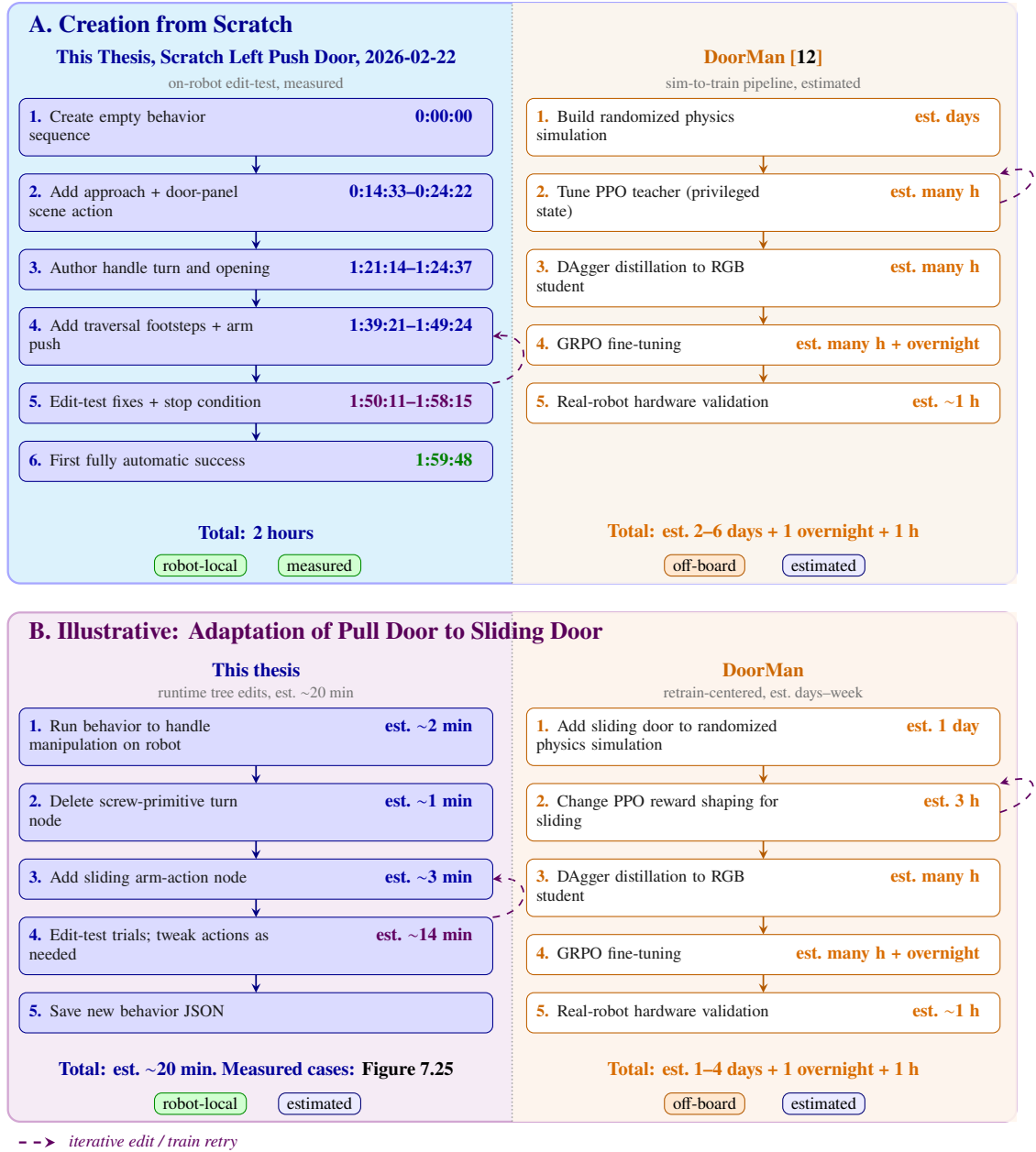}
\caption{This thesis versus DoorMan in from-scratch behavior creation (A) and a hypothetical behavior adaptation of an existing pull door traversal behavior to a sliding door behavior.}
\label{fig:hero_authoring_process_comparison}
\end{figure}

For this section, we'll simply compare against an estimate of what we consider to be our most relevant competition: DoorMan~\cite{xue2025opening}.
In \autoref{fig:hero_authoring_process_comparison} we give a step by step process of creating a door behavior from scratch and adapting an existing door behavior to a different door type.
We use an actual measurement for our push door creation from scratch: 6 steps in 2 hours.
For DoorMan, we use the steps presented in the paper and estimate the durations.

Two main things stand out when doing this comparison.
First, our system is estimated to be 50 to 100 times faster in creating and adapting behaviors.
Secondly, when an adapted behavior is needed, the process for our system is dramatically shortened, but the process for DoorMan still requires the same multi-day pipeline.

This stark contrast is due to the fact that our system is decomposed into reusable parts that have a high degree of independence from each other.
For example, the difference between a pull door and a sliding door might just be adjusting a few actions at runtime.
However, DoorMan's architecture requires a simulation refit, PPO policy tuneup, and an overnight retraining.

\section{Evaluation}
\label{sec:evaluation}

So how well did our hypotheses hold up?
The experiment subsections above analyzed each demonstration individually.
This section summarizes what that evidence supports when taken together.
But first, let's state the hypotheses again:

\begin{enumerate}
    \item Robot-local execution with synchronized UI state, concurrent action layering, reactive tree logic, and behavior-time semantic perception yield door behaviors that are faster and more reliable than prior IHMC baselines and competitive with reported reinforcement learning systems on overlapping door tasks.
    \item Runtime-editable behaviors and perception modules reduce the iteration loop required to diagnose failures, modify logic, and re-test on the robot, relative to redeploy, restart, or retrain workflows.
    \item Decomposing behaviors into reusable primitives, subtrees, and scene actions allows new door and loco-manipulation variants to be brought up by editing a small part of a working behavior rather than rebuilding it from scratch.
\end{enumerate}

Hypothesis 1 is supported on speed and capability, with partial and qualified support on reliability.
\autoref{sec:h1_fast_loco_manipulation_behaviors} traced the arc from the 106~s Atlas hard coded anchor through 14~s and 27~s Nadia door traversals to 34~s and 45~s Alex traversals and the 2 minute 8 second two table sorting behavior.
The 2021 Atlas door behavior took 3.9x longer to execute than our 2024 Nadia pull door behavior.
\autoref{fig:hero_speed_comparison} placed those door results in the tens of seconds regime alongside Zhang et al.~\cite{zhang2024learningopentraversedoors} and DoorMan~\cite{xue2025opening}, well below classical minute scale references.

\autoref{sec:h1_fast_loco_manipulation_behaviors} also documented 11/11 and 12/12 Alex door approach and opening repetitions, 32/32 Unitree standing door openings, and disturbed recoveries on doors and ball sorting (\autoref{fig:in_house_reliability_resilience_summary}).
We did not repeat full door traversals, and we lack comparable repeat trial data for prior IHMC door baselines.
Zhang et al.~\cite{zhang2024learningopentraversedoors} show the most convincing reliability with their 20/20 and 18/20 trial sets for full door traversals.
Learned systems such as Zhang et al. report higher trial counts on full traversals, so reliability comparisons should be read with that scope difference in mind.
However, DoorMan~\cite{xue2025opening} claims ``83\%'' reliability, but does not share the data from which that number is calculated.
We can say that the reliability of our system is competitive with learned door systems, especially when the resilience capabilities are factored in, which are also presented in \autoref{fig:in_house_reliability_resilience_summary}.

Hypothesis 2 is supported by the measured authoring sessions in \autoref{sec:h2_fast_behavior_authoring}.
Runtime-editable behaviors and perception have reduced our behavior authoring times to the hours regime as shown in \autoref{fig:in_house_authoring_from_scratch_timeline}.
Scratch times ranged from 33 minutes for a partially autonomous Nadia session to about 11 hours for the first full Alex pull traversal, with a fully autonomous Alex left push door scratch authoring session at 1 hour 59 minutes 48 seconds.
Creating new behaviors from scratch has been shown to require failure diagnostics, logic modification, and re-testing on the robot, which we have shown in our authoring videos and tables and described.
While we don't have direct comparisons with prior IHMC baselines, \autoref{fig:ihmc_behavior_development_timeline} suggests a drastic reduction in the iteration loop required to create new behaviors.
This figure spans a decade of IHMC real-robot behavior milestones from the 2015 DRC Finals through the Alex loco-manipulation demos in 2026.
What we see is a drastic increase in the rate of new behavior demonstrations over time given the same or less resources allocated to creating them.

Hypothesis 3 is supported by the adaptation sessions in \autoref{sec:h3_fast_adaptation}.
A bottle-door carry composition session took 1 hour 13 minutes, lab to break room door adaptation took 1 hour 54 minutes, and the two table sorting extension took 1 hour 50 minutes including off robot pre-authoring.
We credit our adaptation speed to our architectural decisions where we decompose behaviors into reusable primitives, subtrees, and scene actions.
In each case, the work edited part of an existing behavior rather than rebuilding from an empty tree.
The ability to perform these edits and re-test the behavior at runtime speeds up the process even further.
\autoref{fig:hero_authoring_process_comparison} estimates a 50 to 100 times speedup over DoorMan's multi day pipeline for comparable create and adapt steps, but that literature comparison is inferred rather than directly measured.
The literature does not provide data on the creation or modification timelines of loco-manipulation behaviors.
Standing alone in measuring this is one of the most novel aspects of our work.

\begin{figure}[H]
    \centering
    {\singlespacing
    \resizebox{\columnwidth}{!}{%

\pgfmathdeclarefunction{timelinex}{1}{%
  \pgfmathparse{0.85 + ((#1)-2015)/11.35*(13.35-0.85)}%
}
\newcommand{\behfillfor}[1]{%
  \def\behfill{black!55}%
  \ifnum\pdfstrcmp{#1}{teleop}=0 \def\behfill{orange!85!black}\else
  \ifnum\pdfstrcmp{#1}{hybrid}=0 \def\behfill{yellow!70!orange}\else
  \ifnum\pdfstrcmp{#1}{hard_auto}=0 \def\behfill{blue!45!gray!70}\else
  \ifnum\pdfstrcmp{#1}{rt_door}=0 \def\behfill{blue!75!cyan!25}\else
  \ifnum\pdfstrcmp{#1}{rt_manip}=0 \def\behfill{green!48!black}\else
  \ifnum\pdfstrcmp{#1}{rt_loco_manip}=0 \def\behfill{violet!70!blue!40}\else
  \def\behfill{black!55}%
  \fi\fi\fi\fi\fi\fi}
\newcommand{\behdot}[3]{%
  \pgfmathsetmacro{\bx}{timelinex(#1) + \behdotxshift}%
  \behfillfor{#3}%
  \filldraw[\behfill, draw=black!55, line width=0.22pt] (\bx, #2) circle (0.055);%
}
\newcommand{\behmarkerflat}[4]{%
  \behdot{#1}{#2}{#3}%
  \pgfmathsetmacro{\bx}{timelinex(#1)}%
  \pgfmathsetmacro{\laby}{#2 + 0.24}%
  \node[font=\fontsize{4.4}{5.0}\selectfont, anchor=south, text=black!82, inner sep=0.6pt]
    at (\bx, \laby) {#4};%
  \draw[draw=black!38, line width=0.22pt] (\bx, \laby) -- (\bx, #2);%
}
\newcommand{\behmarkertopleft}[5]{%
  \pgfmathsetmacro{\bx}{timelinex(#1) + \behdotxshift}%
  \behfillfor{#3}%
  \filldraw[\behfill, draw=black!55, line width=0.22pt] (\bx, #2) circle (0.055);%
  \pgfmathsetmacro{\labx}{\bx - 0.22*cos(45) + (#5)}%
  \pgfmathsetmacro{\laby}{#2 + 0.22*sin(45)}%
  \draw[draw=black!38, line width=0.22pt] (\bx, #2) -- (\labx, \laby);%
  \node[font=\fontsize{4.4}{5.0}\selectfont, rotate=-45, anchor=south east, text=black!82, inner sep=0.6pt]
    at (\labx, \laby) {#4};%
}
\newcommand{\behlegendentry}[4]{%
  \behfillfor{#1}%
  \filldraw[\behfill, draw=black!55, line width=0.22pt] (#2, #3) circle (0.045);%
  \node[legendtxt, anchor=west] at ({#2 + 0.10}, #3) {#4};}

\begin{tikzpicture}[
    x=1cm,
    y=1cm,
    every node/.style={outer sep=0pt},
    lanetitle/.style={font=\fontsize{6.6}{7.4}\selectfont\bfseries, anchor=west, text=black!70},
    axistick/.style={font=\scriptsize, text=black!65},
    legendtxt/.style={font=\fontsize{6.2}{6.8}\selectfont, text=black!78},
  ]

  \def\laneAlex{3.18}
  \def\laneUnitree{2.34}
  \def\laneNadia{1.50}
  \def\laneAtlas{0.66}
  \def\axisleft{0.85}
  \def\axisright{13.35}
  \def\axistop{3.62}
  \def\axisbottom{0.18}
  \def\legendrowA{-0.92}
  \def\legendrowB{-1.28}

  \def\behdotxshift{0}

  \pgfmathsetmacro{\eraA}{timelinex(2015.0)}
  \pgfmathsetmacro{\eraB}{timelinex(2016.0)}
  \pgfmathsetmacro{\eraC}{timelinex(2022.0)}
  \pgfmathsetmacro{\eraD}{timelinex(2024.0)}
  \pgfmathsetmacro{\eraE}{timelinex(2025.5)}
  \pgfmathsetmacro{\eraF}{timelinex(2026.35)}
  \path[fill=orange!7, opacity=0.55, rounded corners=3pt]
    (\eraA, \axisbottom) rectangle (\eraB, \axistop);
  \path[fill=blue!6, opacity=0.55, rounded corners=3pt]
    (\eraB, \axisbottom) rectangle (\eraC, \axistop);
  \path[fill=cyan!8, opacity=0.55, rounded corners=3pt]
    (\eraC, \axisbottom) rectangle (\eraD, \axistop);
  \path[fill=gray!8, opacity=0.55, rounded corners=3pt]
    (\eraD, \axisbottom) rectangle (\eraE, \axistop);
  \path[fill=violet!7, opacity=0.55, rounded corners=3pt]
    (\eraE, \axisbottom) rectangle (\eraF, \axistop);

  \foreach \y in {2.76, 1.92, 1.08} {
    \draw[draw=black!10, line width=0.35pt] (\axisleft, \y) -- (\axisright, \y);
  }

  \node[lanetitle] at (0.08, \laneAlex) {Alex};
  \node[lanetitle] at (0.04, \laneUnitree) {Unitree H1-2};
  \node[lanetitle] at (0.08, \laneNadia) {Nadia};
  \node[lanetitle] at (0.08, \laneAtlas) {Atlas};

  \behmarkertopleft{2026.06}{\laneAlex}{rt_door}{First pull bring-up}{0}
  \behmarkertopleft{2026.15}{\laneAlex}{rt_door}{Scratch left push}{0.134}
  \behmarkertopleft{2026.18}{\laneAlex}{rt_door}{Push/pull speed runs}{0.334}
  \behmarkertopleft{2026.20}{\laneAlex}{rt_door}{Reactive right pull}{0.544}
  \behmarkertopleft{2026.21}{\laneAlex}{rt_loco_manip}{Bottle carry-through}{0.766}
  \behmarkertopleft{2026.24}{\laneAlex}{rt_door}{Break room left pull}{0.966}
  \behmarkertopleft{2026.25}{\laneAlex}{rt_manip}{Table ball pick-place}{1.188}
  \behmarkertopleft{2026.26}{\laneAlex}{rt_manip}{Reactive ball sort}{1.410}
  \behmarkertopleft{2026.28}{\laneAlex}{rt_manip}{Open house ball return}{1.620}
  \behmarkertopleft{2026.29}{\laneAlex}{rt_loco_manip}{Two-table loco-manip}{1.842}

  \behmarkertopleft{2025.42}{\laneUnitree}{rt_manip}{Manipulation demos}{0}
  \behmarkertopleft{2026.00}{\laneUnitree}{rt_door}{Scratch right-pull opening}{0}

  \behmarkertopleft{2022.68}{\laneNadia}{teleop}{Milestone 1 multi-task}{0}
  \behmarkertopleft{2022.88}{\laneNadia}{hard_auto}{Look-and-step rough terrain}{0.013}
  \behmarkertopleft{2023.42}{\laneNadia}{rt_door}{Scratch right push door}{0}
  \behmarkertopleft{2023.47}{\laneNadia}{hybrid}{Supervised can pick-place}{0.178}
  \behmarkertopleft{2023.48}{\laneNadia}{rt_door}{First auto push door}{0.400}
  \behmarkertopleft{2023.64}{\laneNadia}{rt_door}{Force-based push door}{0.456}
  \behmarkertopleft{2024.20}{\laneNadia}{rt_door}{14 s left push-bar door}{0.072}
  \behmarkertopleft{2024.28}{\laneNadia}{rt_door}{Reactive left pull handle}{0.217}
  \behmarkertopleft{2024.50}{\laneNadia}{rt_loco_manip}{Three doors in a row}{0.208}
  \behmarkertopleft{2024.55}{\laneNadia}{rt_loco_manip}{ONR mock-building run}{0.385}
  \behmarkertopleft{2024.551}{\laneNadia}{rt_door}{Hook-hands right pull}{0.617}

  \behmarkerflat{2015.433}{\laneAtlas}{teleop}{DRC Finals (8 tasks)}
  \behmarkertopleft{2018.04}{\laneAtlas}{hard_auto}{Rough terrain autonomy}{0}
  \behmarkertopleft{2018.17}{\laneAtlas}{hard_auto}{Floor ball pickup}{0.090}
  \behmarkertopleft{2018.45}{\laneAtlas}{teleop}{VR rough-terrain teleop}{0.014}
  \behmarkertopleft{2018.67}{\laneAtlas}{hard_auto}{Hard-coded pull door}{0.005}
  \def\behdotxshift{-0.0275}
  \behmarkertopleft{2021.48}{\laneAtlas}{hybrid}{Building exploration}{0}
  \def\behdotxshift{0.0275}
  \behmarkertopleft{2021.481}{\laneAtlas}{hard_auto}{Hard-coded pull door}{0.177}
  \def\behdotxshift{0}

  \draw[draw=black!55, line width=0.55pt] (\axisleft, \axisbottom) -- (\axisright, \axisbottom);
  \foreach \yr in {2015,2016,2017,2018,2019,2020,2021,2022,2023,2024,2025,2026} {
    \pgfmathsetmacro{\tickx}{timelinex(\yr)}
    \draw[draw=black!55, line width=0.45pt] (\tickx, \axisbottom) -- (\tickx, \axisbottom - 0.08);
    \node[axistick, anchor=north] at (\tickx, \axisbottom - 0.10) {\yr};
  }
  \node[font=\scriptsize\bfseries, anchor=north, text=black!70]
    at ({0.5 * (\axisleft + \axisright)}, \axisbottom - 0.48) {Calendar year};

  \behlegendentry{teleop}{0.95}{\legendrowA}{Teleoperated}
  \behlegendentry{hybrid}{4.35}{\legendrowA}{Hybrid or supervised}
  \behlegendentry{hard_auto}{7.95}{\legendrowA}{Hard-coded autonomous}
  \behlegendentry{rt_door}{0.95}{\legendrowB}{Runtime-editable door}
  \behlegendentry{rt_manip}{4.05}{\legendrowB}{Runtime-editable manipulation}
  \behlegendentry{rt_loco_manip}{7.55}{\legendrowB}{Runtime-editable loco-manipulation}

\end{tikzpicture}%
    }
    \par
    }
    \caption{
        Calendar timeline of IHMC real robot behavior milestones since the 2015 DRC Finals.
        Colors distinguish teleoperation, hybrid or supervised operation, legacy hard-coded autonomy, and runtime-authored door, manipulation, and loco-manipulation behaviors.
        Entries include teleoperated and non-loco-manipulation demos from the development story in \autoref{ch:building}.
    }
    \label{fig:ihmc_behavior_development_timeline}
\end{figure}

\subsection{Desirable Characteristics}

So how well did we achieve the desirable characteristics we defined in \autoref{ch:metrics}?
We think we did pretty well overall.

Our system is highly capable.
We demonstrated dozens of different task varieties, from door traversal types and sorting objects on tables to building exploration.
A taxonomy is illustrated in \autoref{fig:behavior_library_taxonomy}.
There are, however, classes of tasks we cannot do yet, such as grasping objects while the robot is walking, dynamic bracing, tasks that require proprioception, and fast manipulation tasks such as playing ping-pong.
We cannot yet do them because the necessary control and perception components do not yet exist.
Dancing, swimming, and sports are probably classes of tasks that this system is not well suited to address because they are too continuous, too dynamic, and even artistic.

Demonstrating our system was certainly feasible, as is shown by our results on real hardware.
However, we did not show that it was feasible to take our system off-site and reproduce the results.
Our system definitely supports fast behaviors, and we feel that the videos included of our real robot demonstrations are watchable at 1x speed.
It also supports parallelism through moving multiple parts of the body at once, including being able to manipulate a door panel while walking.

We showed that our system is capable of producing reliable repeated-run behaviors for door approach and opening and for sorting balls on tables.
Our walking controller was, however, not very reliable for the door traversal walk-throughs.
We also showed how our system supports robustness to environmental disturbances through our pull door reactivity demo and our ball sorting demo with human disturbance.
This goes hand in hand with our resilience capabilities.
By using fallback nodes with condition nodes, common failure modes can be handled.
For longer-tail resilience, we support this through operator-robot teaming.
When behaviors fail, the human operator might be able to solve the problem with edits or additions to the behavior.

Our system is independent from external systems when it is in autonomous mode.
It does not rely on external comms or compute in this mode, and our perception is entirely from on-board color vision.

Our system is designed around adaptability of the human-robot team.
We have shown this through our results in behavior authoring.
To achieve this, we made our system observable, predictable, and directable.
Our system is learnable, and we have a handful of trained expert operators.
We have provided a usage guide in \autoref{ch:tutorial}.
We hope the reader will find our system understandable through reading this thesis, watching the videos, and browsing the source code.
We think the interface is easy to use, as suggested by our fast from-scratch authoring times.

Our system can be analyzed after behavior runs.
We have done this via screen recordings of the operator interface and the log system shown in \autoref{sec:logged_data}.
It is also debuggable by looking at logged data and by rerunning behavior logic in an application we call the ``behavior test facilitator''.
This application uses a kinematic simulation combined with the playback of real perception data from a logged run.
Very often, bugs can be reproduced and fixed in this environment.
We also use this application to automate behavior tests.

Lastly, we view our system as being very extendable.
As laid out in \autoref{ch:building}, the story of building it has been feature after brainstormed feature.
This train of extension is still in active motion.

%



\chapter{Discussion}
\label{ch:pontification}

We are pretty happy with what we've been able to accomplish with a model based approach.
The combination of DARPA Robotics Challenge inspired Coactive Design for humanoid robot operation, affordance templates, behavior trees, and a behavior-managed perception scene has proved to work very well.
In a 10 year period, we went from a 20 person team working tirelessly over years to get a robot doing 8 tasks via teleoperation, to essentially one person being able to set up a fully-autonomous multi-station ball sorting behavior in hours.

Much of the field, and especially startup robotics companies, have now abandoned these fundamental approaches and have jumped ship for ``end-to-end'' learning.
However, we are convinced that a lot of tasks can be solved both ways.
In fact, with the increased prevalence of intelligent coding assistance, our more classical, model-based approach may even be able to keep pace with learned systems.

\section{Strong Points}

We think our robot-local runtime-editable design is really the strongest success of our system.
Running the compute locally robustifies the system by decoupling it from external digital disturbances such as communications failures and compute failures on remote servers.
Editing behaviors online has enabled rapid bringup of new behaviors and modification, composition, and extension of existing behaviors as we have shown.

We also think we did well with the interface between the behavior system and the robot controller.
The ability to schedule asynchronous footstep, arm, leg, spine, and neck commands enables a high level of behavioral expressiveness.
It also supports the integration of classical planners and AI assisted behavior composition.

The next most successful part of our system is probably the authorable behavior-time perception through scene actions.
Perception systems have a lot of important and non-intuitive quality levels.
Furthermore, these quality levels can vary with lighting, situation, and hardware differences.
The metrics of quality include depth accuracy, semantic object detection confidence, latency, and frequency.
Our design designates an expert human operator to evaluate and exploit available quality levels for each behavior situation to achieve the goal at hand.

We also think a strong point of our system is its ability to be combined with learned systems.
For example, as mentioned in \autoref{ch:building}, we built an action node that executed a learned mimic policy, such as a dance or a door traversal walk through.
More work in this direction could enable more robust door walkthroughs.
We think this is a good way to get the best of both worlds for the time being.

\section{Weak Points}

There are important caveats, however, with our current system.
For one, our object-on-table manipulation demos center around a bottle pickup and balls.
We haven't cracked the nut on perceiving orientation of objects.
This is why our premier sorting demos use spherical balls.
Since they are symmetrical objects, only position is required to grasp them from any direction.
We can even sometimes grasp the balls while they are moving.
We have preliminary results using FoundationPose for object orientation estimation, but it is not yet working well enough for repeatable tests.

Another major weak point is the lack of support for executing sequence subtrees that return when they are done.
This would solve two existing problems in the system.
For one, it would solve the problem of having two of the same JSON file loaded in the tree at once.
When the same JSON file is loaded twice in the tree, changes to one instance are not reflected in the other, and saving one can overwrite the other.
If sequences can be run as a subtree and returned when done, identical subtrees throughout the system can be instantiated once instead of duplicated.

We have also lacked the ability to call a subroutine with one line.
In general, making the behavior tree mirror the structure of computer programs more would greatly increase the expressiveness of the system.
We think a good strategy would be to keep thinking of the next most useful thing to add, guided by operator frustration in trying to create new behaviors.

We also have a scaling problem with large trees, both in the view and in the data synchronization.
We were able to use trees consisting of well over 100 nodes, as we presented in \autoref{ch:building} with the ONR Demo that consisted of 178 nodes.
However, we don't think this could grow much further without some optimization.
Two issues stand out.
First, when loading in large subtrees, there can be a loading delay that can cause a glitch in the network synchronization.
Second, for trees with a tall maximum depth, the behavior tree editor view entries near the leaves of the tree can become indented too much and hard to work with.
We think both of these issues are very solvable.

\section{Natural Next Steps}

Throughout this work, we really want to add force-based action primitives.
The codebase even includes a ``wrench action'' node, which was used to pick up boxes.
Ultimately, we just didn't have the bandwidth to approach it from a controls perspective.

The wrench actions for box lifting worked only under very special circumstances.
The way it worked was to depend on prior position based arm actions to place the hands on either side of the box.
Then, the wrench action would command inwards accelerations for the hands to result in a force on the sides of the box.
The wrench action was never completed with a termination condition, so it was always a hack when demonstrated.

It would be a natural next step to develop a force-based action definition in combination with good support for it in the whole body controller.
One area where we would like to use it is for turning door handles.
Door handles have different spring constants and might require a torque based ramp if the handle isn't moving.
A hybrid force-position action might do the trick well---force-controlled angular velocity along the screw trajectory until the end stop is hit.
Lots of tasks, like opening drawers and sliding things against surfaces, could take advantage of such an approach.

In general, force-controller primitives can be a more natural way to interact with the world.
In fact, it can even be used as a form of proprioception.
For example, whether doors are push or pull, and whether they are locked, is not always visually identifiable.
One often has to try pushing in various directions to figure out what is going on with articulated objects.

Another area of work that would be nice to explore would be grasp planning.
We have seen that there may be some promising development in grasp planning, but haven't explored it at all yet.
For example, when going to operate tools designed for humans, it would be helpful and, we would think, feasible to plan a 5-finger grasp for them instead of having to manually specify finger configurations.

The problem doesn't have to be solved generally.
For example, if the behavior authoring interface was adapted to virtual or augmented reality, the user could simply use their hands to demonstrate an example grasp technique to seed the planner.

Our screw primitive action, inspired by~\cite{Pettinger_2022}, is a very versatile manipulation planning abstraction.
We think there could be useful constructs like this that extend beyond positional sliding and turning.
There may be interesting ways to model forces, torques, or the way objects can deform or move, using a similar approach.
For example, there is some work like this on clothes folding where geometrical primitives are used to define folding patterns~\cite{Moletta2023VRClothFolding}.

Another fruitful area to explore would be navigation.
In the behaviors presented in this thesis, even for the ``building exploration'' ones, we didn't ever form a map and plan from point A to point B.
They were more like a slightly reactive scripting of which door to go through next, with some stand-in-place visual searches.
Our building exploration demos have been limited to a maximum of 3 doors separating spaces in an open lab environment.
The main reason for that is that we haven't had a fully tetherless robot to work with yet, so we couldn't take it into non-lab spaces.

We think our approach to navigation would follow a similar pattern to our behavior-time scene: keep a behavior-local map that has authorable behavior-time interactions.
This is because the behaviors are the primary consumer of this information and are the most informed about what they need and when.
We would likely maintain very different map representations for different task levels.
For navigating from room to room in a building, a topological graph structure would be the most useful.
For traversing a space, a heightmap or octree might be the most useful.
For locating objects in dense spaces, a detailed scene graph might be the most useful.
It would be exciting and interesting to explore options for extending our behavior scene to include both maps and behavior specific planners.

It would also be great if we could use virtual reality while authoring to record arm motions.
These recordings could then be used directly as arm actions.
This would enable the operator to quickly define complicated motions that are hard to specify with geometric authoring alone.
That capability could also be extended by filtering the motions or by collecting repeated demonstrations and training a vision-action policy.
It would be valuable to demonstrate and dispatch vision-action model training runs from within the authoring interface.
On training completion, the resulting policies could be used as an alternative mode of the arm action.

Taking a step back and looking at our system as a whole, there's also an important gap in handling open world environments.
We focus on a human operator's ability to author robot behaviors rather than robot-generated ones.
The lack of a generative ability on the robot, where it could ``author'' its own behavior, limits our system to well thought out human constructed scenarios.

\section{Why Not Fully Embrace Behavior Trees}

One interesting part of this thesis is that we use a ``behavior tree'' architecture, but left out some of the key mechanisms from the literature version~\cite{2018_colledanchise}.
The most important part of why is just that we built what appeared to work best for us.
The other aspect is that the lone result from the literature just wasn't that convincing~\cite{2023_centaur_bt}, being ultimately very slow and locomotion only.

The reason we don't tick from the root node is that, for authoring behavior at runtime, we have the robot in a particular configuration and doing a particular thing and we are trying to author a particular part.
This means we want a ``cursor'', just like in a text editor, that marks our current editing location.

It may be beneficial to introduce the root node tick system for autonomous mode.
This has the prospect still of enabling the behavior to drastically change goals or strategies at basically any point in the behavior.
However, behavior trees don't really address the full problem here, as the robot might be in a situation, such as when going through a door, that it can't immediately get out of.
In dynamical situations, the robot really has to finish what it's doing before aborting the mission.
In these cases, it needs to execute specific ``abort mission'' behaviors that are very dependent on current robot and task configuration.

This is where we guess Behavior Trees would suggest using the blackboard.
However, the Behavior Tree blackboard is an unstructured design component and we would strive for a more formal mechanism for doing things like this.
A promising theory for this sort of reactivity includes sequential composition~\cite{1999_burridge}.




\section{The Future}

We would like to see a near future where technical people can buy or build their own robots and put them towards useful work.
To us, it is most convincing if the system is open source, understandable, and tinkerable.
We would like to be able to modify and extend the capabilities of personally owned robots without being beholden to subscriptions or AI datacenters.
We think it is possible to build such a system using classical approaches for planning and control and simpler neural nets like YOLO and FoundationPose for vision.
This is in direct contrast to where much of the industry is going.
We imagine that if we extended our runtime-editable system to a portable device like an augmented reality headset, and with more development, we could author behaviors for common household chores and basic outdoor contracting work.
It could be that the real world is too diverse for anything other than large-scale end-to-end learned systems, but we'd like to explore and try to find out where the boundaries between classical and learned systems really are.

Given the above dream, we would definitely like to keep using our system into the future and keep expanding it to new capabilities and modalities.
We are passionate about the approach taken in this thesis and think it has a bright future.
A lot of smart people contributed to this system over more than a decade and we are honored to present such positive results.

Thank you to the reader for hanging in there for what turned out to be a very long thesis.
We hope we have sparked some interest in our unique approaches to problems and that you got some value out of it.
Cheers!

\renewcommand{\bibname}{\centering\bfseries\fontsize{12}{14.4}\selectfont References}
\printbibliography

\end{document}